\pdfoutput=1
\documentclass[11pt]{article}

\usepackage{microtype} % microtypography
\usepackage{booktabs}  % tables
\usepackage{multirow} 
\usepackage{url}  % urls
\usepackage{todonotes}
\usepackage{wrapfig}
\usepackage{amsmath}
\usepackage{amsthm}

\usepackage[preprint]{automl}

\usepackage{natbib}
\usepackage{amsmath,amsfonts,bm}

\def\eqref#1{equation~\ref{#1}}
\def\1{\bm{1}}

\def\rvx{{\mathbf{x}}}

\DeclareMathAlphabet{\mathsfit}{\encodingdefault}{\sfdefault}{m}{sl}
\SetMathAlphabet{\mathsfit}{bold}{\encodingdefault}{\sfdefault}{bx}{n}

\DeclareMathOperator*{\argmin}{arg\,min}

\title{BBOmix: A Tabular Benchmark for Hyperparameter Optimization of Unsupervised Biological Representation Learning
}

\author[1]{\nameemail{Luca~Thale-Bombien}{luca.thale-bombien@uni-leipzig.de}}
\author[1]{\nameemail{Jan~Ewald}{jan.ewald@uni-leipzig.de}}
\author[1]{\nameemail{Ralf~K\"onig}{ralfkoenig@gmx.de}}
\author[1,2]{\nameemail{Aaron~Klein}{aaron.klein@tue.ellis.eu}}

\affil[1]{Center for Scalable Data Analytics and Artificial Intelligence (ScaDS.AI) Dresden/Leipzig, Leipzig University}
\affil[2]{ELLIS Institute T\"ubingen}

\hypersetup{%
  pdfauthor={Luca Thale-Bombien, Jan Ewald, Ralf Koenig, Aaron Klein}, % will be reset to "Anonymous" unless the "final" package option is given
  pdftitle={BBOmix: A Tabular Benchmark for Hyperparameter Optimization of Unsupervised Biological Representation Learning},
  pdfsubject={The rapid advancement of high-throughput sequencing has led to large, high-dimensional omics datasets. Deep unsupervised learning architectures, particularly Autoencoders (AEs), are increasingly used for dimensionality reduction and representation learning in this domain. However, AEs are highly sensitive to architectural choices and hyperparameters, and unsupervised optimization typically relies on reconstruction loss, which may be a poor proxy for downstream utility. Exhaustive hyperparameter optimization (HPO) is computationally expensive, leading researchers to frequently rely on suboptimal default configurations. To democratize access to large-scale unsupervised HPO research, we introduce \textbf{BBOmix}, the first open-source tabular benchmark for unsupervised representation learning on real-world biological data. Our benchmark includes 105,000 evaluations across four AE architectures and seven multi-omics modalities from the TCGA and SCHC datasets. We quantify the correlation between reconstruction loss and downstream task performance and provide an extensive evaluation of state-of-the-art single-fidelity, multi-fidelity, and transfer learning HPO methods, establishing a rigorous baseline for future research in unsupervised biological representation learning.},
  pdfkeywords={AutoML, Hyperparameter Optimization, Unsupervised Biological Representation Learning}
}

\begin{document}
\maketitle

\begin{abstract}
The rapid advancement of high-throughput sequencing has led to large, high-dimensional omics datasets. Deep unsupervised learning architectures, particularly Autoencoders (AEs), are increasingly used for dimensionality reduction and representation learning in this domain. However, AEs are highly sensitive to architectural choices and hyperparameters, and unsupervised optimization typically relies on reconstruction loss, which may be a poor proxy for downstream utility. Exhaustive hyperparameter optimization (HPO) is computationally expensive, leading researchers to frequently rely on suboptimal default configurations. To democratize access to large-scale unsupervised HPO research, we introduce \textbf{BBOmix}, the first open-source tabular benchmark for unsupervised representation learning on real-world biological data. Our benchmark includes 105,000 evaluations across four AE architectures and seven multi-omics modalities from the TCGA and SCHC datasets. We quantify the correlation between reconstruction loss and downstream task performance and provide an extensive evaluation of state-of-the-art single-fidelity, multi-fidelity, and transfer learning HPO methods, establishing a rigorous baseline for future research in unsupervised biological representation learning.

\end{abstract}

\section{Introduction}

The rapid advancements in high-throughput sequencing technologies have caused a paradigm shift in the life sciences, leading to large, high-dimensional datasets \citep{mardis2008impact, reuter2015high}. 
This data, often referred to as \textit{omics} data, refers to a family of large-scale biological disciplines, like genomics or transcriptomics, each comprehensively characterizing a specific class of biological molecules and different layers of cellular processes. Each modality (DNA, RNA, etc.) reflects the molecular state of biological systems at multiple levels of resolution e.g. tissue or single-cell.

Given the inherent noise and sparsity of this \textit{omics} data, deep unsupervised learning architectures, especially Autoencoders (AE) have shown strong capabilities for dimensionality reduction, denoising and representation learning, which is a cornerstone for functional analysis in cell biology and identification of biomarkers, enabling modern personalized medicine \citep{kopf2021latent, selby2025visible}.
Despite their use for handling \textit{omics} data, they are notoriously fragile and highly sensitive to architectural choices (e.g. depth, latent space dimensionality) and hyperparameter configurations (e.g. learning rate, weight decay), hindering the establishment of principled and standardized approaches \citep{joas2025AUTOENCODIX}. 

Traditionally, hyperparameter optimization (HPO) is set in supervised learning settings, where the configuration choice is reflected directly in an objective metric, such as validation accuracy. In contrast, unsupervised learning typically optimizes the reconstruction loss, which can be a poor proxy for the actual usefulness and quality of a learned latent space for further downstream applications such as cell-type classification or disease prognosis. Exhaustive HPO on large-scale omics data is computationally expensive and thus researchers frequently rely on default hyperparameter choices or manual tuning based on previous experiences, resulting in suboptimal representations and potentially skewed downstream analysis.

In recent years, the machine learning community has developed numerous benchmarks for HPO and NAS on supervised classification tasks, including NAS-Bench-201 \citep{dong2020bench}, FCNet \citep{klein2019tabular}, and TabRepo \citep{salinas2023tabrepo}, yet no equivalent exists for unsupervised representation learning and downstream task evaluation on biological data. While recent efforts such as AUTOENCODIX \citep{joas2025AUTOENCODIX} and scSSL-Bench \citep{ovcharenko2025scssl} have established standardized frameworks for evaluating AE architectures and self-supervised methods on omics data, they do not provide the exhaustive, precomputed configuration evaluations needed to benchmark HPO algorithms themselves. Beyond this, the field has historically relied on synthetic data or computer vision datasets, such as MNIST and CIFAR-10 that fail to capture the high-dimensional noise, non-random sparsity, and complex regulatory patterns inherent in biological systems \citep{whalen2022navigating}. 

We address both limitations with \textbf{BBOmix}\footnote{Code available under: \url{https://github.com/Kavlahkaff/BBOmix}}, the first open-source tabular HPO benchmark for unsupervised representation learning on omics data, comprising exhaustive evaluations across multiple AE architectures and omics modalities to democratize access to large-scale unsupervised HPO research in biology.
Specifically our contributions are threefold:

\begin{itemize}
    \item We provide the first large-scale open-source dataset for HPO in an unsupervised omics setting, consisting of 105.000 evaluations across 35 tasks from 2 datasets.
    \item We systematically quantify the correlation between reconstruction loss and downstream task performance, providing insights into trade-offs that can function as a guideline for further research.
    \item Finally, we provide an extensive evaluation of state-of-the-art (SOTA) optimization methods, including single-fidelity, multi-fidelity and transfer learning approaches, thereby establishing a baseline for HPO performance in an unsupervised setting. Transfer Learning especially allows us to investigate how effective meta-learning works between different data modalities and architectural families.
\end{itemize}

\section{Related Work}

\paragraph{Hyperparameter Optimization} Hyperparameter optimization (HPO)~\citep{franceschi-arxiv25} seeks the hyperparameter configuration $\rvx_{\star} \in \argmin_{\rvx \in \mathbb{X}} f(\rvx)$ of a machine learning algorithm that optimizes the validation performance $f(\rvx)$. In practice~\citep{bischl2023hyperparameter}, the objective function $f$ is typically expensive to evaluate, has unknown closed-form solution, and can only be observed with noise, i.e., $y \sim f(\rvx) + \epsilon$, where $\epsilon \sim \mathcal{N}(0, \sigma)$.

A widely used approach for HPO is Bayesian optimization (BO)~\citep{garnett_bayesoptbook_2023}, which fits a probabilistic surrogate model, such as a Gaussian process~\citep{snoek-nips12a} or a tree-structured Parzen estimator~\citep{bergstra2011algorithms}, to approximate the objective function $f(\rvx)$ and uses an acquisition function to balance exploration and exploitation.
In many applications, inexpensive approximations of $f$ are available, e.g., by training for fewer epochs or on subsets of the data. 
Multi-fidelity methods exploit these approximations to discard poor configurations early and reduce computational cost. 
For example, successive halving~\citep{jamieson-aistats16} iteratively allocates increasing budgets to the better-performing half of configurations. 
Methods such as Hyperband~\citep{li2018hyperband} and BOHB~\citep{falkner2018bohb} combine random or BO-guided sampling with successive halving and often outperform standard BO under limited compute budgets. This is especially important in deep learning, where a single full evaluation may require hours of GPU time. Our benchmark records per-epoch reconstruction losses for all 105{,}000 runs, making it directly compatible with multi-fidelity optimization methods.

\paragraph{Transfer Learning} To reduce the cold-start cost of HPO, which typically begins from random or default hyperparameter configurations, transfer learning approaches initialize the search using knowledge from prior runs on related tasks \citep{feurer2015initializing}. 
For example, \citet{wistuba2015learning} employ a warm-starting strategy that reuses the most successful configurations from previous tasks as initial candidates. 
Another approach is to constrain the search space to a smaller region containing high-performing configurations identified during earlier runs~\citep{perrone2019learning}.

In supervised learning settings, transfer HPO has demonstrated consistent improvements on tabular benchmarks such as TabRepo \citep{salinas2023tabrepo}. 
However, to the best of our knowledge, there has been no systematic evaluation of whether similar gains can be achieved in unsupervised settings using omics data, where task similarity is more difficult to define and downstream metrics are not observed during training.

\paragraph{HPO Benchmarks} Reproducible benchmarking is central to AutoML research. To reduce the computational cost of HPO evaluation, tabular benchmarks decouple optimization from model training by exhaustively evaluating configurations offline and using table lookups during benchmarking. For continuous or high-dimensional search spaces, surrogate models can instead predict the performance of unseen configurations. Representative benchmarks include NAS-Bench-201 \citep{dong2020bench}, FCNet \citep{klein2019tabular}, LCBench \citep{zimmer2021auto}, and TabRepo \citep{salinas2023tabrepo}.

A key limitation of existing HPO benchmarks is their supervised setting, where the optimization objective directly matches the downstream task. In unsupervised representation learning however, models are typically optimized for reconstruction loss, while downstream utility is not observed during training. As shown in Section~\ref{subsec:downstream}, reconstruction loss does not necessarily correlate with downstream performance, and optimal configurations may differ substantially.
Existing unsupervised benchmarks do not address this mismatch. Prior work has largely focused on synthetic or computer vision datasets \citep{locatello2019challenging}, which do not capture key properties of biological omics data, such as extreme sparsity, modality-specific noise, high feature-to-sample ratios, and biologically structured covariance. To our knowledge, \textbf{BBOmix} is the first benchmark for exhaustive HPO evaluation on real-world biological datasets in the unsupervised setting. Although HPO is performed using reconstruction loss, we additionally record downstream supervised metrics for every configuration, enabling post-hoc analysis of the relationship between unsupervised objectives and biological utility.

\paragraph{Autoencoder for biological representation learning}

Advancements of AE based representation learning were developed simultaneous to the advent of high-throughput methods where individual layers, including DNA (genomics), RNA (transcriptomics) or proteins (proteomics), are measured. 
Additionally, techniques enabling single-cell resolution vastly increased the necessity of new methods for  representation learning \citep{mamoshina2016applications}.
In the past, numerous AE architectures have been developed. Firstly, variational AE (VAE) enable generative modelling for synthetic data. More complex architectures, e.g stacked VAEs (or hierachical) emerged for multi-omics integration \citep{simidjievski2019variational} or cross-modal VAE for modality translation \citep{yang2021multi}. 

Recent work incorporates prior-knowledge into the decoder structure to enforce latent space explainability and improve regularization. These biological-informed neural AEs, such as VEGA \citep{seninge2021vega}, expiMAP \citep{lotfollahi2023biologically} or OntoVAE \citep{doncevic2023biologically} rely on an hierarchical ontology describing the feature-to-latent relationship representing gene-to-function knowledge.
Other applications are data imputation or denoising with AE which is increasingly important as single-cell resolution data contain large number of drop-out events \citep{eraslan2019single}. Furthermore, recent AEs where covariates are disentangled via parallel training of individual discriminator networks show new directions  \citep{lotfollahi2023predicting}. 

Despite this, the field lacks standardization. Architectural comparisons are typically conducted on a single dataset with manually selected hyperparameters, making it unclear whether performance differences reflect architectural merit or hyperparameter sensitivity. \citet{hu2018parameter} established evidence that hyperparameter choices dominate outcomes in VAEs for single-cell transcriptomics stressing the necessity of rigorous HPO. More recently, the standardization issue was addressed by the development of the framework AUTOENCODIX for common AE architectures enabling HPO benchmarking \citep{joas2025AUTOENCODIX}.

\section{Methodology}\label{sec:method}

Leveraging AUTOENCODIX, we collected \textbf{BBOmix} as a large-scale dataset of hyperparameter configurations for four AE architectures trained on real-world multi-omics data. 
In the following, we describe the used datasets, selected AEs and outline the procedure to construct \textbf{BBOmix}.

\subsection{Training Datasets}\label{subsec:dataset}

We used two established multi-omics datasets, the Cancer Genome Atlas (TCGA) \citep{weinstein2013cancer} and the Single-Cell Human Cortex (SCHC) dataset \citep{zhu2023multi}, which we will briefly describe in this section. Both datasets were retrieved as pre-processed including mutation score calculation as described by \citet{joas2025AUTOENCODIX, acx2025rawdata}. All modalities were processed in an standardized way using YAML-files determining feature selection, data scaling and random splitting into train, test and validation splits.

\textbf{{The Cancer Genome Atlas}}
The TCGA pan-cancer multi-omics dataset comprises over 10,000 human patient samples and clinical annotations for 32 cancer types including survival. The dataset contains three data modalities:  transcriptomic data (RSEM normalized bulk-cell mRNA-seq), epigenetic data (DNA methylation), genomic data (a per-gene mutation score). The clinical annotations were used as labels in downstream classification tasks (see Appendix \ref{app:clinical_annotations}). 

\textbf{{Single-Cell Human Cortex}}
The SCHC dataset contains single-cell multi-omics measurements from the developing human cerebral cortex. The dataset has two data modalities and clinical annotations: single-cell RNA sequencing (scRNA-seq), chromatin accessibility profiling (scATAC-seq) and associated cell-level annotations. In total, the dataset contains over 45.000 rows (cells), with the annotations: cell-type, age group and sex. 

\subsection{Different Architectures}\label{subsec:architecture}

We consider four AE architectures spanning a range from standard AEs to biologically informed encoders. All architectures share a common backbone of fully connected layers with batch normalization, dropout and ReLU activation, differing in their latent space and decoder structure. Encoder and decoder are symmetric in depth, with hidden layer widths determined by an encoding factor applied successively from the input dimension down to the latent bottleneck.
\\
\textbf{Vanillix:} Standard fully-connected feed-forward AE with batch normalization, dropout, and ReLU. Encoder and decoder are symmetric and we minimizes MSE reconstruction loss for training.
\\
\textbf{Varix:} A standard $\beta-$Variational Auto-Encoder (VAE)~\citep{higgins-iclr17} which follows the same architecture as Vanillix. We minimize the standard ELBO loss formulation: $Total\_loss = reconstruction\_loss + \beta \times KL\_divergence$, where $\beta$ is an additional hyperparameter.
\\
\textbf{Disentanglix:} Extends VAE for higher latent space disentanglement by a decomposition of the KL-divergence into three terms to enable a higher penalty of the total-correlation. Implementation in AUTOENCODIX is based on the $\beta$-TCVAE of \citep{chen2018isolating} with three different weights for the loss: total correlation $\beta_{tc}$, mutual information $\beta_{mi}$, and per-dimension KL weight $\beta_{DimKL}$.
\\
\textbf{Ontix:} A biologically-informed VAE where the decoder is constrained by an ontology (pathways or chromosome-based). The decoder represents known gene-to-pathway (Reactome \citep{milacic2024reactome}) or gene-positional (Chromosome) relationships. Both ontologies differ in decoder sparseness, since genes act in multiple pathways, but gene location is unique. We analyze both independently.

\subsection{Downstream Evaluation}

Beyond reconstruction quality, we assess the utility of the learned latent spaces by evaluating on a suite of downstream tasks applied to the frozen embeddings of the trained encoders. 
Following the protocol of \cite{joas2025AUTOENCODIX}, our downstream classification tasks are derived from available clinical annotations (see Appendix \ref{app:clinical_annotations}). 
We fit a logistic regression for classification on the embeddings and report the area under the receiver operating curve (AUC-ROC) in a one-versus-one scheme. At the end, the final downstream performance is aggregated as an average over all downstream tasks and reported alongside the individual scores.

\subsection{Data Collection}\label{subsec:benchmark}

Table \ref{tab:search_spaces} shows the distinct search spaces for each architecture, reflecting the differences in model structure and inductive bias.
The search spaces contain diverse parameters, ranging from low to high-dimensional, discrete, categorical, and numerical values. For each architecture, we sampled $1000$ configurations uniformly at random  over the respective search space, yielding 5,000 unique architecture-configuration pairs in total.
\\
We trained each configuration for 300 epochs on each of the seven dataset-modality tasks, and record reconstruction loss at every epoch, to produce full learning curves. To account for stochastic variation stemming from randomness in the AUTOENCODIX pipeline, we ran each configuration with three independent random seeds. The complete experimental design therefore yields:
\\
$\text{\#architectures} \times \text{\#configurations} \times \text{\#seeds} \times \text{\#modalities} = 105,000\text{ individual training runs.}$
\\
We trained each network on Nvidia H100 GPUs, amounting to $\approx$10k GPU hours in total.
Per-epoch loss curves are retained for all runs, enabling the benchmark to support both final-performance evaluations and learning-curve-based HPO methods, such as multi-fidelity and early-stopping approaches.
The resulting dataset contains for each configuration the reconstruction loss, training runtime and accuracy on downstream tasks.
We used the tabular blackbox format of the Syne Tune library \citep{salinas2022syne} to serialize the dataset. 
Each architecture-modality pair defines a separate blackbox, containing the full configuration to performance mapping as a queryable lookup table with dimensions for hyperparameter configurations, training fidelities and random seeds yielding 35 blackboxes in total. 
This format allows any HPO optimizer supported by Syne Tune to be evaluated against the benchmark, facilitating the use of surrogates to predict objective values of any configurations in the search space \citep{eggensperger2015efficient}. We use a simple k-nearest neighbor regressor with $k=1$ for our benchmarks to minimize the introduction of additional bias.

\begin{table}[ht]
\tiny
\centering
\caption{Hyperparameter search spaces for each AE architecture. 
% Shared parameters are common to all four architectures. Architecture-specific parameters are listed separately. 
$\mathcal{U}$ denotes a uniform distribution, $\mathcal{LU}$ a log-uniform distribution, 
and $\mathcal{C}$ a categorical choice. } 
\label{tab:search_spaces}
\resizebox{\textwidth}{!}{%
\begin{tabular}{lllcccc}
\toprule
\multirow{2}{*}{\textbf{Parameter}} & 
\multirow{2}{*}{\textbf{Symbol}} & 
\multirow{2}{*}{\textbf{Type}} &
\multicolumn{4}{c}{\textbf{Range or Choices}} \\
\cmidrule(lr){4-7}
& & & \textsc{Vanillix} & \textsc{Varix} & \textsc{Disentanglix} & \textsc{Ontix} \\
\midrule
\multicolumn{7}{l}{\textit{Shared architectural hyperparameters}} \\
\midrule
Input filter size     & $D$                  & $\mathcal{C}$  
  & \multicolumn{4}{c}{$\{128,\ 256,\ 512,\ 1024,\ 2048,\ 4096\}$} \\
No.\ encoder layers   & $n_\text{layers}$    & $\mathcal{C}$  
  & \multicolumn{4}{c}{$\{2,\ 3,\ 4\}$} \\
Encoding factor       & $r$                  & $\mathcal{U}$  
  & \multicolumn{4}{c}{$(1,\ 4)$} \\
Latent dimensionality & $d_z$                & $\mathcal{C}$  
  & \multicolumn{3}{c}{$\{2,\ 4,\ 8,\ 16,\ 32,\ 64\}$} & --- \\
\midrule
\multicolumn{7}{l}{\textit{Shared training hyperparameters}} \\
\midrule
Learning rate         & $\eta$               & $\mathcal{LU}$ 
  & \multicolumn{4}{c}{$(10^{-5},\ 10^{-1})$} \\
Batch size            & $b_s$                & $\mathcal{C}$  
  & \multicolumn{4}{c}{$\{32,\ 64,\ 128,\ 256\}$} \\
Dropout probability   & $p_\text{drop}$      & $\mathcal{U}$  
  & \multicolumn{4}{c}{$(0,\ 0.9)$} \\
Weight decay          & $\lambda$            & $\mathcal{LU}$
  & \multicolumn{4}{c}{$(10^{-5},\ 10^{-1})$} \\
\midrule
\multicolumn{7}{l}{\textit{Architecture-specific hyperparameters}} \\
\midrule
KL divergence weight  & $\beta$              & $\mathcal{LU}$ 
  & --- & $(10^{-4},\ 10^{0})$ & --- & $(10^{-4},\ 10^{0})$ \\
\midrule
Disentanglement  & $\beta_{tc}$              & $\mathcal{LU}$ 
  & --- & --- & $(10^{-2},\ 10^{3})$ & --- \\
\midrule
Disentanglement  & $\beta_{mi}$              & $\mathcal{LU}$ 
  & --- & --- & $(10^{-4},\ 10^{0})$ & --- \\
\midrule
Disentanglement  & $\beta_{DimKL}$              & $\mathcal{LU}$ 
  & --- & --- & $(10^{-2},\ 10^{0})$ & --- \\
\bottomrule
\end{tabular}%
}
\end{table}

\section{Experiment Section}

We structure our experimental analysis around two questions: (1) how hyperparameter choices and their interactions shape model performance across architectures and tasks, and (2) how well different HPO optimizers exploit the benchmark's structure to find good configurations efficiently.

\subsection{Reconstruction Loss as a Proxy for Downstream Performance}\label{subsec:downstream}
A practical motivation for using reconstruction loss as a cheap HPO criterion
is that it is available during training, while downstream evaluation requires a
separate supervised pipeline evaluating multiple tasks after training. If reconstruction loss
were a reliable proxy, one could safely optimize it instead of additionally
computing the true objective, drastically reducing the cost of HPO.

\begin{figure*}[t]
    \centering
    \begin{subfigure}[h]{0.48\textwidth}
        \centering
        \includegraphics[width=\linewidth]{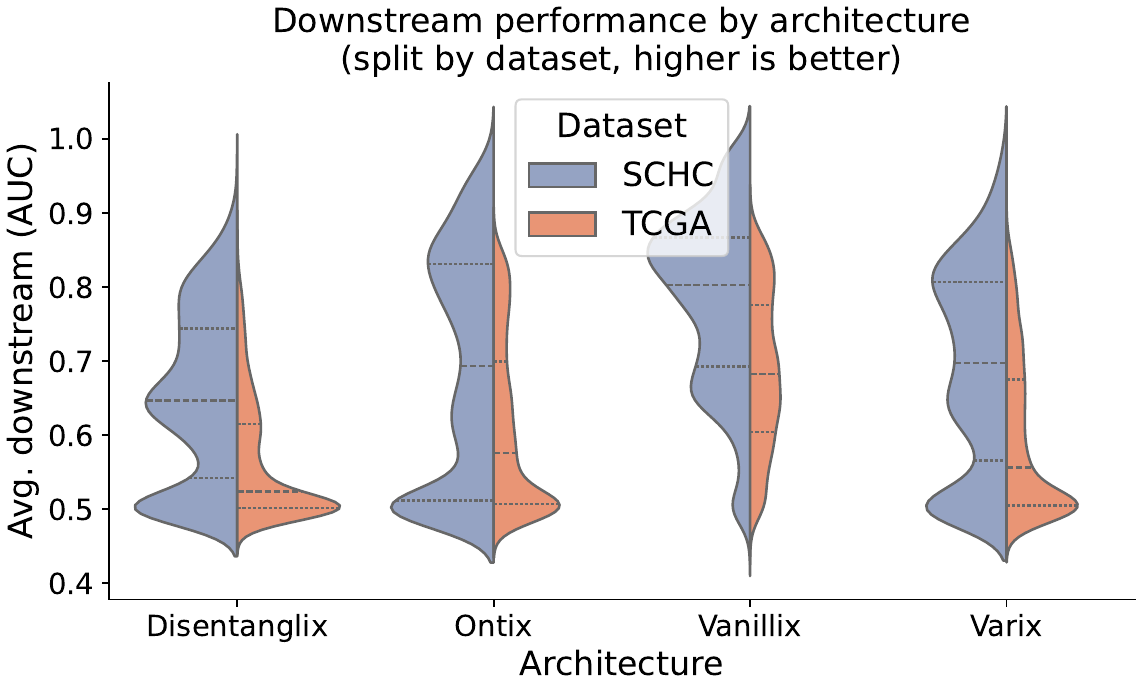}
        \caption{Downstream performance}
    \end{subfigure}
    \hfill
    \begin{subfigure}[h]{0.48\textwidth}
        \centering
        \includegraphics[width=\linewidth]{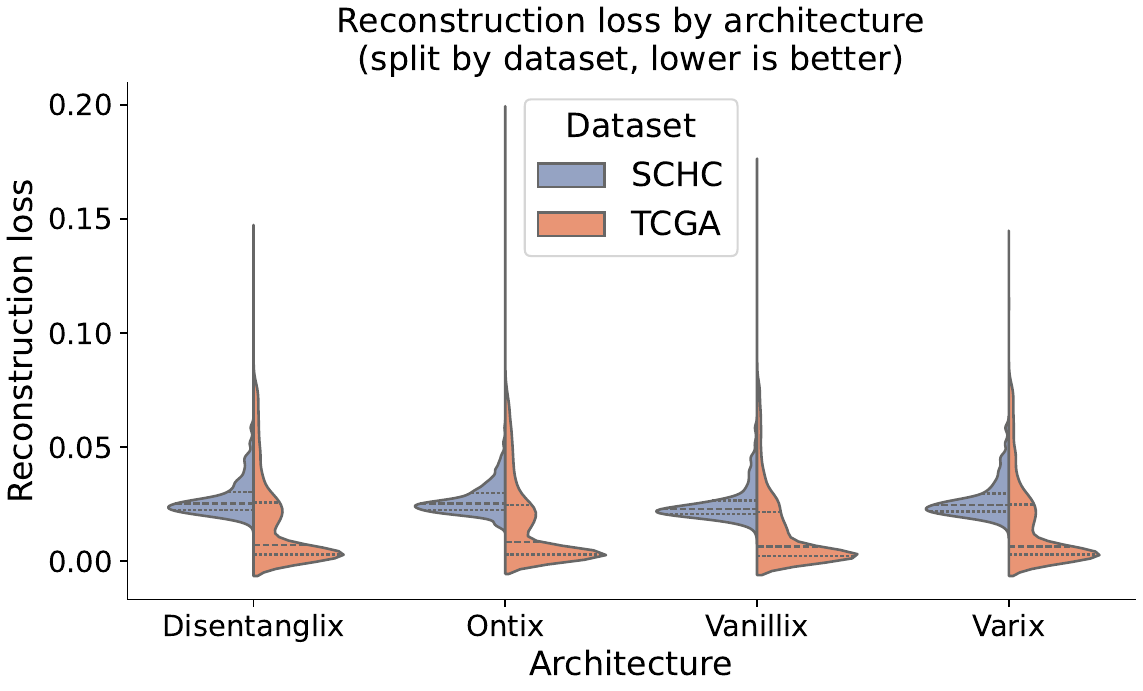}
        \caption{Reconstruction Loss}
    \end{subfigure}
    \caption{Violin plots of architecture performance with respect to dataset and performance metrics.}
    \label{fig:violin-plots}
\end{figure*}

To assess whether reconstruction loss is a reliable proxy for downstream
performance, we first examine the distributions of reconstruction loss and
downstream performance across hyperparameter configurations.
Figure~\ref{fig:violin-plots}a and Figure~\ref{fig:violin-plots}b show violin
plots of downstream performance and reconstruction loss, respectively, for each
architecture on the SCHC and TCGA dataset, illustrating the spread and variability
induced by hyperparameter choices within each architecture. For a breakdown per downstream task see Appendix \ref{app:per-task}.

We then we compute the Spearman rank correlation ($\rho$) between the reconstruction loss (to be minimized), and downstream task performance (to be maximized)
for each (architecture, task) pair in Figure~\ref{fig:hp-importance}b,
with Ontix cells split diagonally for its two ontology variants (chromosome /
reactome). A strong negative correlation ($\rho \approx -1$) indicates that
lower reconstruction loss consistently predicts better downstream performance,
validating its use as a proxy during HPO.

As shown in Figure~\ref{fig:hp-importance}b, reconstruction loss is a strong, but not perfect, predictor for downstream task performance across most architectures and modalities, except DNA (genomics).
Disentanglix, Vanillix, and Varix exhibit consistently strongest
correlations ($\rho$ between $-0.72$ and $-0.88$), with the
sole exception of the \textit{tcga\_DNA\_CLIN} modality, where correlations are
near zero or weakly positive. Ontix shows a similar pattern but with somewhat attenuated
magnitudes ($\rho$ ranging from $-0.37$ to $-0.72$), consistent across both the
chromosome and reactome ontology splits, again with \textit{tcga\_DNA\_CLIN} as
the primary outlier ($\rho \approx -0.00$ / $-0.41$).
Lower reconstruction loss consistently ranks hyperparameter configurations closer to
those achieving higher downstream performance, supporting its use as a low-cost
HPO criterion. The notable exception is the \texttt{tcga\_DNA\_CLIN} setting,
where the proxy relationship breaks down across all architectures, suggesting
that DNA methylation combined with clinical covariates in the TCGA cohort poses
a qualitatively different optimisation landscape. Practitioners should therefore
apply reconstruction-loss-based HPO with caution in this specific setting, while
treating it as a broadly valid selection criterion elsewhere.

\subsection{Hyperparameter Importance}
\label{subsec:hp_importance}

\begin{figure}
    \centering
    \begin{subfigure}[h]{0.48\linewidth}
        \centering
        \includegraphics[width=\linewidth]{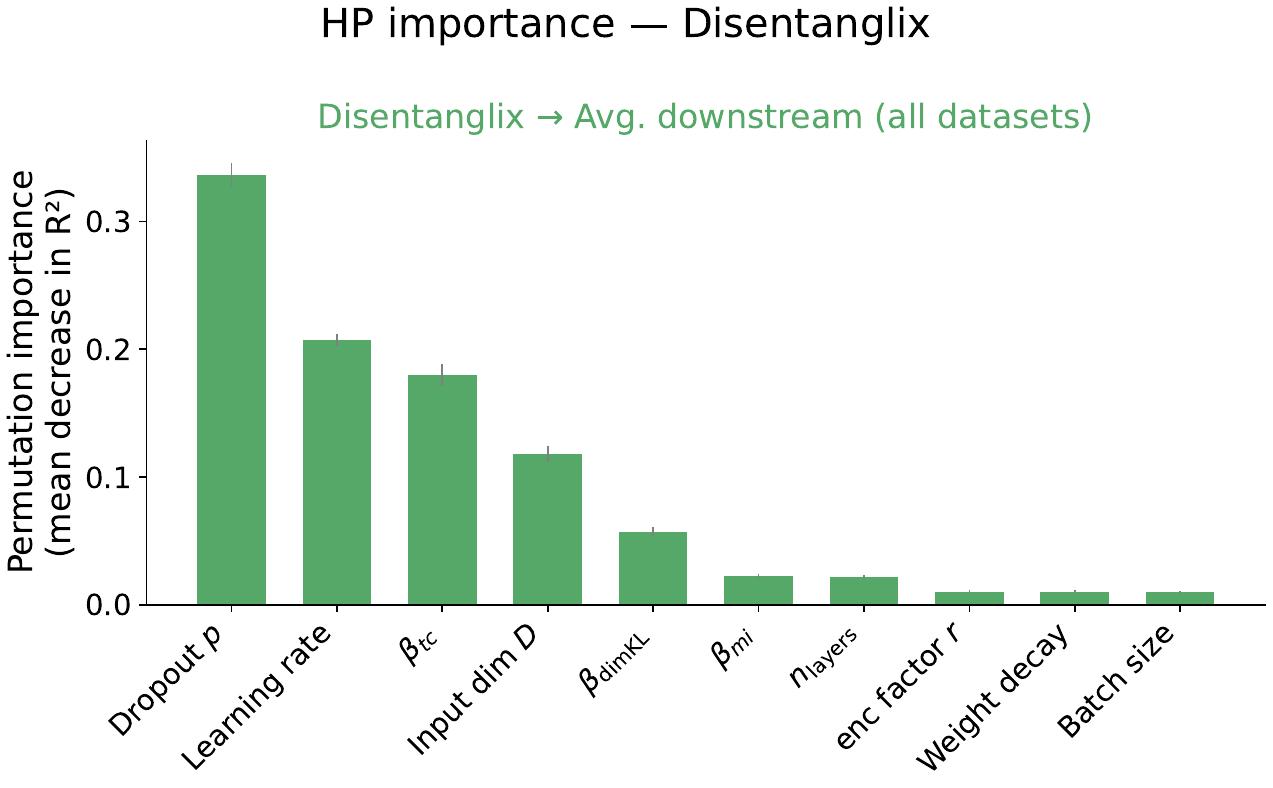}
        \caption{HP importance for Disentanglix downstream task performance.}
    \end{subfigure}
    \hfill
    \begin{subfigure}[h]{0.48\linewidth}
        \centering
        \includegraphics[width=\linewidth]{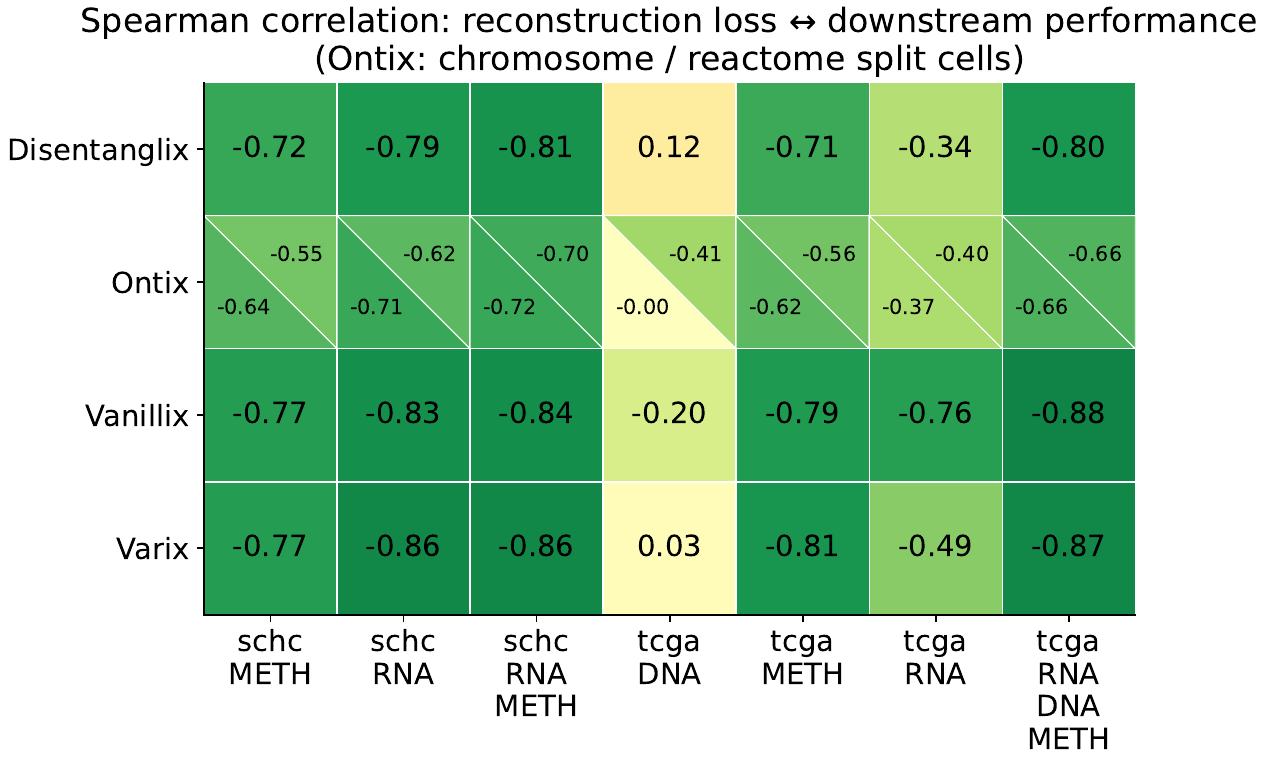}
        \caption{Spearman rank correlation of reconstruction loss and downstream performance.}
    \end{subfigure}
    \caption{Hyperparameter importance and proxy metric correlation.}
    \label{fig:hp-importance}
\end{figure}

To identify which hyperparameters most strongly influence to final performance, we train a random forest regressor on all hyperparameter configurations to predict each target and compute permutation importance scores, i.e.\ the mean decrease in out-of-sample $R^2$ when a single HP column is randomly shuffled \citep{fisher2019all}. 
We report results separately for (i) aggregate downstream performance, (ii) reconstruction loss, and (iii) each individual downstream target, yielding a comprehensive picture of which HPs matter and for which objective.
Figure~\ref{fig:hp-importance} shows the permutation importance scores exemplary for Disentanglix on average downstream task performance (see Appendix~\ref{app:hp-importance} for more results).

Across all architectures, \textbf{dropout $p$} is the single most influential
hyperparameter for downstream predictive performance: it ranks first for Vanillix,
Varix, and Disentanglix, and second only to learning rate for Ontix.
\textbf{Learning rate} is the dominant HP for Ontix on both SCHC and TCGA, while
playing a consistently strong secondary role elsewhere.
\textbf{Input dimensionality $D$} ranks among the top three for reconstruction loss
across all architectures, reflecting its direct effect on the AE's
compression bottleneck, but is less decisive for downstream classification and survival
tasks.
Architecture-specific regularisation terms, like $\beta$ for Varix and Ontix,
and $\beta_{tc}$ for Disentanglix, carry moderate importance for downstream
performance but are largely negligible for reconstruction loss, suggesting they shape
the latent geometry without strongly affecting reconstruction fidelity.
Structural HPs such as $n_{\text{layers}}$ and encoder factor $r$ or training HPs batch size, and weight
decay contribute only minor importance across objectives and architectures.

\subsection{Comparison of Optimizers}

\begin{figure*}[t]
    \centering
    \begin{subfigure}[h]{0.32\textwidth}
        \centering
        \includegraphics[width=\linewidth]{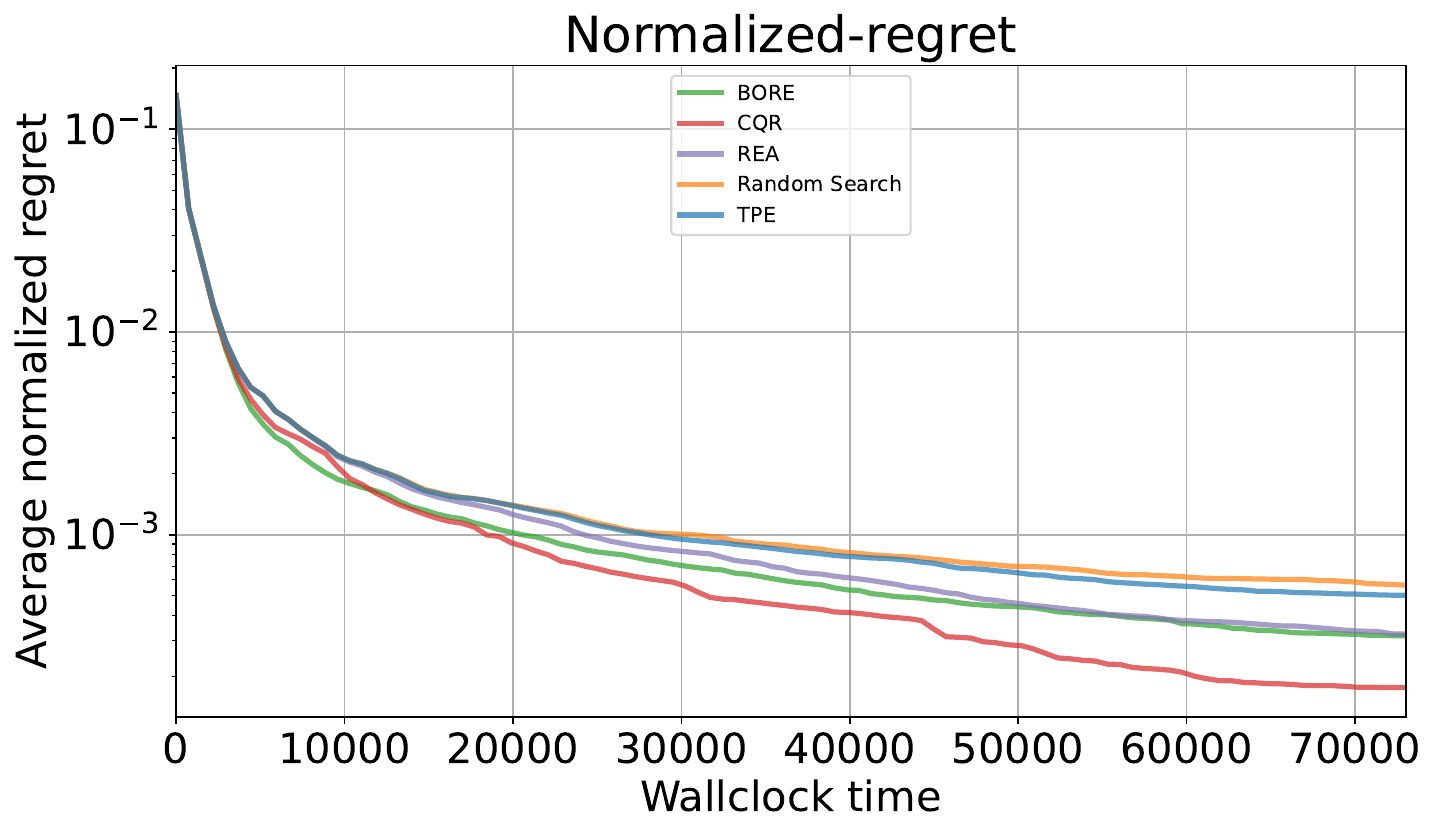}
        \caption{Single Fidelity}
    \end{subfigure}
    \begin{subfigure}[h]{0.32\textwidth}
        \centering
        \includegraphics[width=\linewidth]{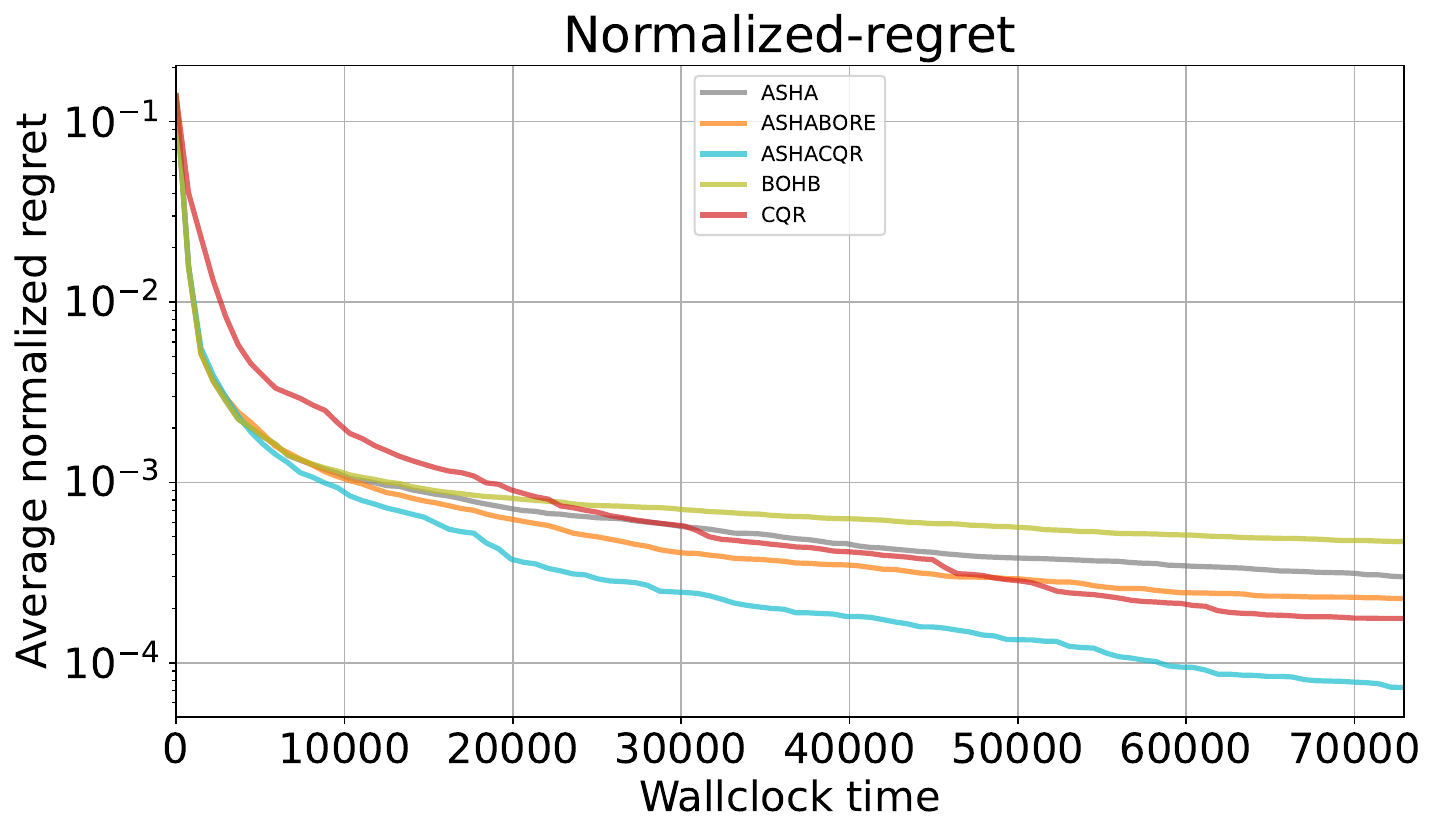}
        \caption{Multi Fidelity}
    \end{subfigure}
    \begin{subfigure}[h]{0.32\textwidth}
        \centering
        \includegraphics[width=\linewidth]{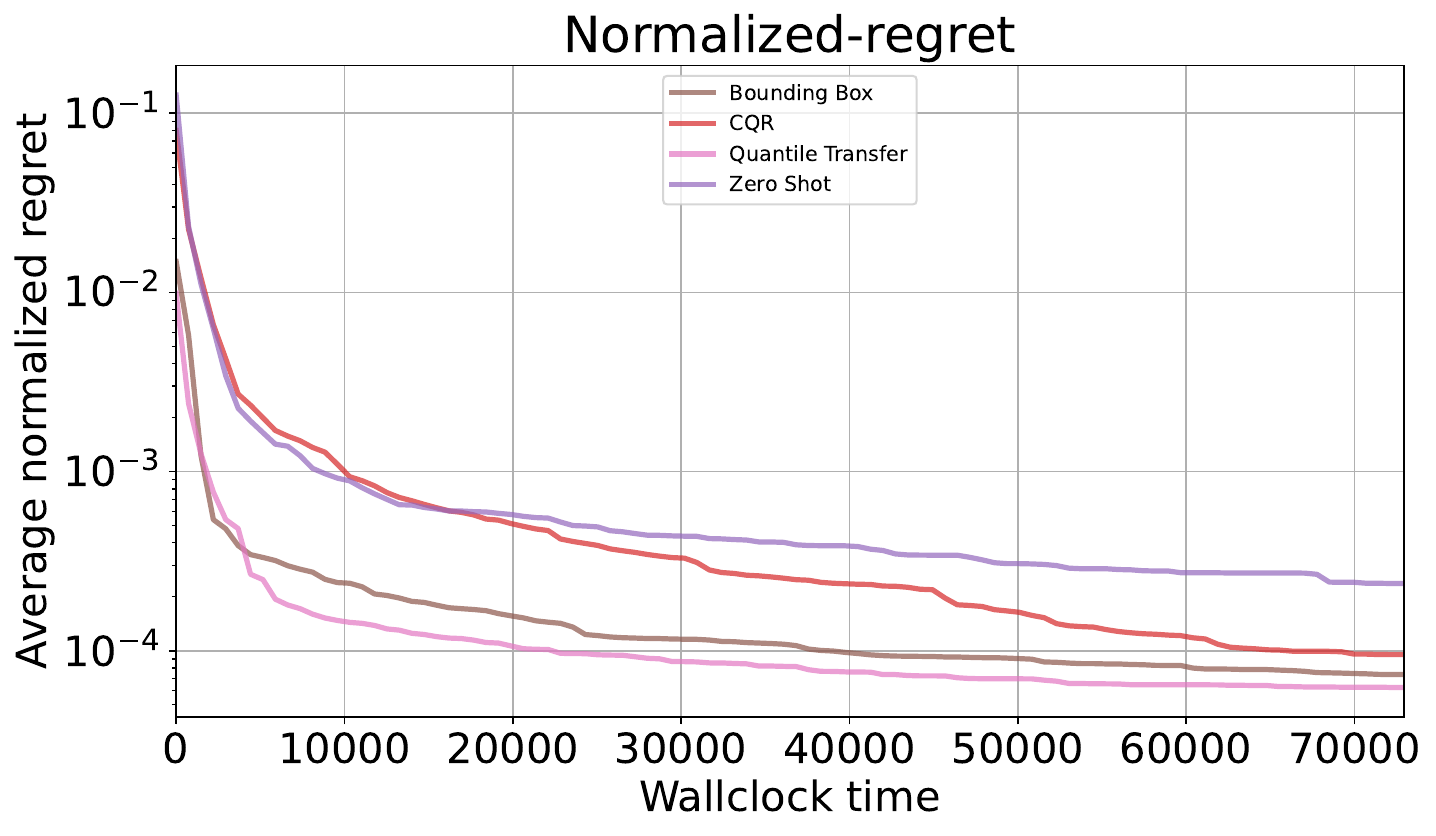}
        \caption{Transfer Learning}
    \end{subfigure}
    \hfill
    \begin{subfigure}[h]{0.32\textwidth}
        \centering
        \includegraphics[width=\linewidth]{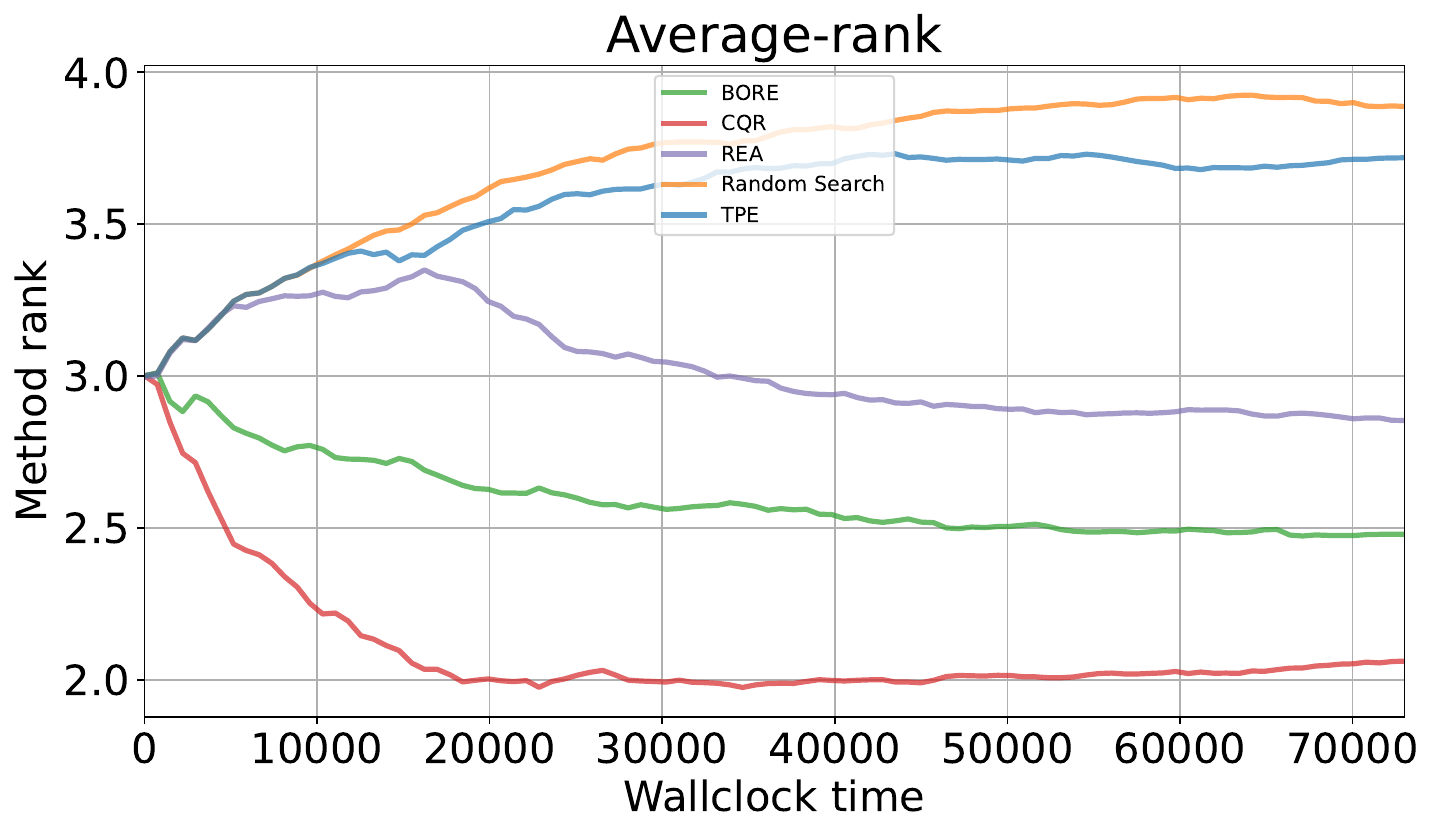}
        \caption{Single Fidelity}
    \end{subfigure}
    \hfill
    \begin{subfigure}[h]{0.32\textwidth}
        \centering
        \includegraphics[width=\linewidth]{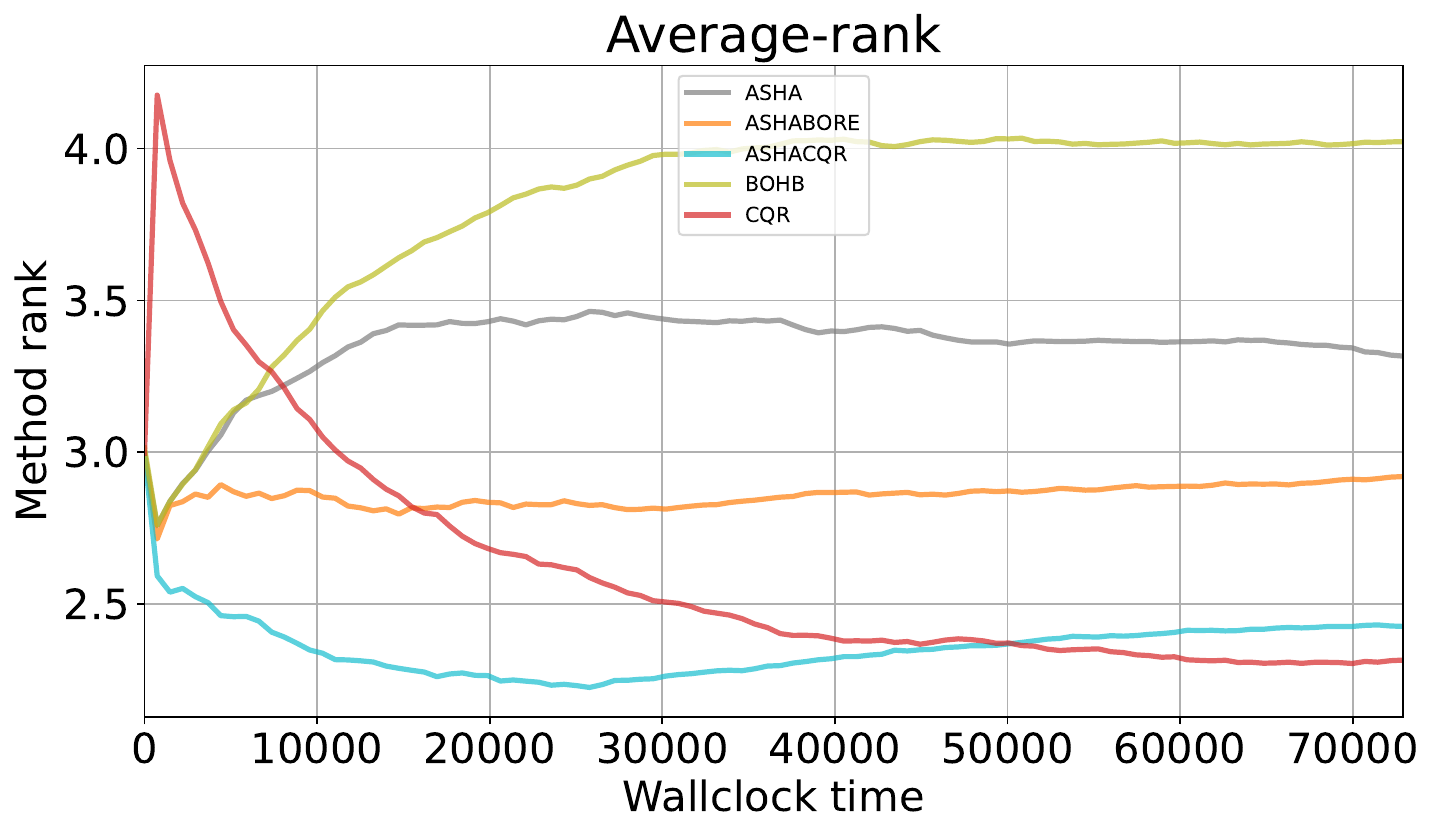}
        \caption{Multi Fidelity}
    \end{subfigure}
    \begin{subfigure}[h]{0.32\textwidth}
        \centering
        \includegraphics[width=\linewidth]{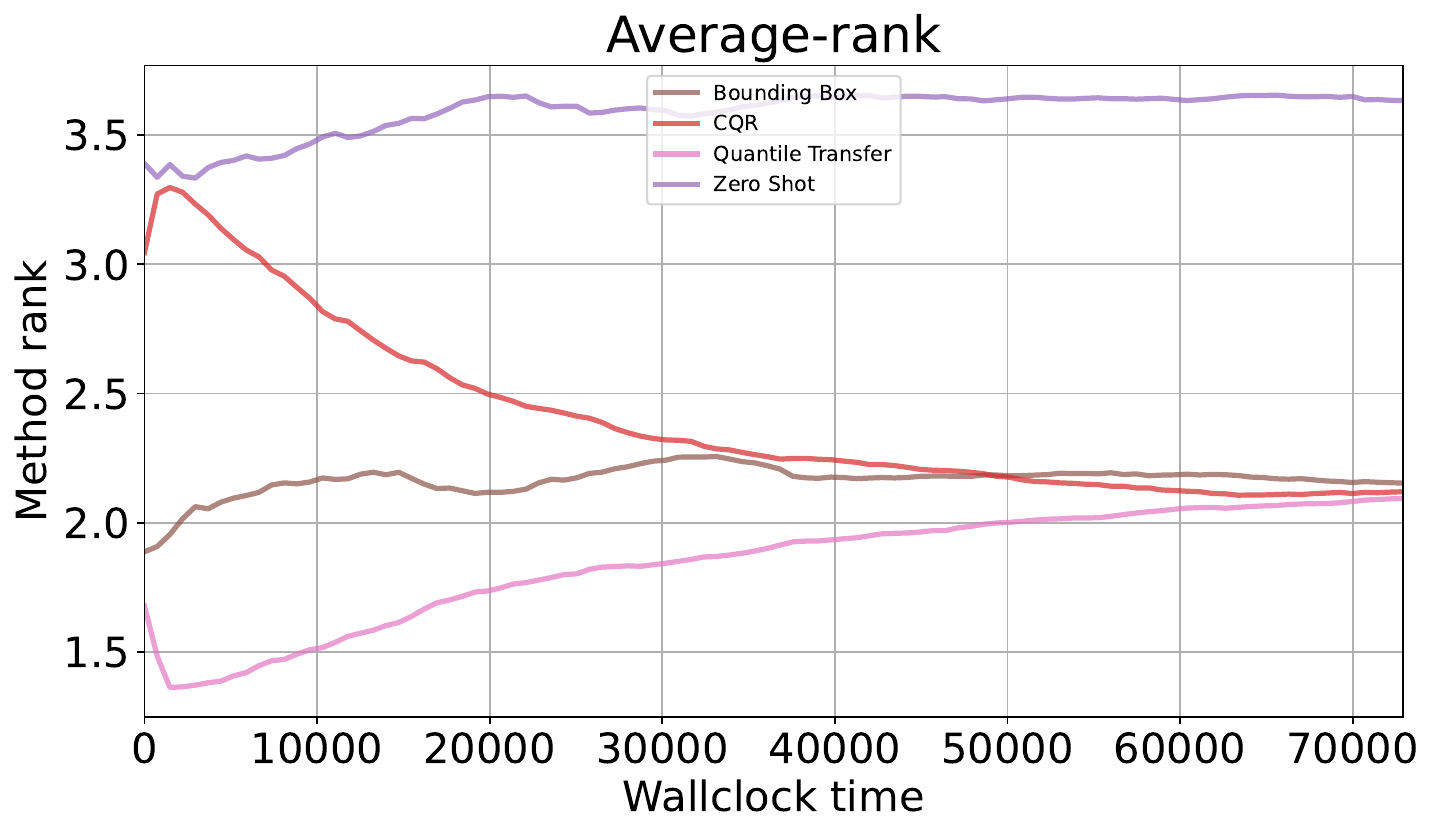}
        \caption{Transfer Learning}
    \end{subfigure}
    \hfill
    \caption{Normalized regret (a-c) and average rank (d-f) of single-fidelity, multi-fidelity and transfer-learning optimization algorithms, averaged over all architectures and tasks. 
    % All methods are plotted for a fixed budget of 72000 seconds wallclock-time and were evaluated over 30 seeds. The transfer learning trajectories are aggregated over four different transfer settings (one per architecture and one combined). We show multi-fidelity and transfer learning methods in comparison to CQR, since it achieved the best performance of the evaluated single-fidelity optimizers.
    }
    \label{fig:single-fidelity}
\end{figure*}

We benchmark methods across three paradigms and compare them on all 35 different blackboxes (see Figure \ref{fig:single-fidelity}). See Appendix~\ref{app:optimizers} for full descriptions of all methods. All methods use the default hyperparameters suggested by Syne Tune \citep{salinas2022syne}, with the exception of REA, where the population size was reduced to avoid sampling too many random configurations. Each method had a simulated wall-clock time budget of 72000 seconds. We repeated each optimization run 30 times with a different random seed. 
\\
\textbf{Single-fidelity} methods exhaust the full budget for each configuration. We include Random Search~\citep{bergstra2012random} as an unbiased baseline, TPE~\citep{bergstra2011algorithms}, BORE~\citep{tiao2020bayesian}, REA~\citep{real2019regularized}, and CQR~\citep{salinas2023optimizing}. Among these (Figure \ref{fig:single-fidelity}a and \ref{fig:single-fidelity}d), CQR achieves the lowest normalized regret and ranks most favorably across the budget range, achieving best performance across the board. BORE and REA perform slightly worse than CQR, but both clearly outperform TPE and Random Search throughout the optimization trajectory. Random Search, while informative as an unbiased baseline, falls behind as budget increases and structured methods are able to exploit the response surface more effectively. TPE occupies an intermediate position, offering modest but inconsistent gains over random search. The rank plots (Figure \ref{fig:single-fidelity}d) confirm the ordering observed in the regret curves. The optimization trajectory plots for individual tasks in Appendix~\ref{app:optimizers} show that all methods converge to similar final performance values given sufficient budget, but differ substantially in their early-budget behavior.
\\
\textbf{Multi-fidelity} methods exploit the full per-epoch learning curves available in the benchmark to enable early stopping. We evaluate ASHA~\citep{li2020system}, ASHABORE, ASHACQR, and BOHB~\citep{falkner2018bohb}. These methods (Figure \ref{fig:single-fidelity}b and \ref{fig:single-fidelity}e) discard unpromising configurations early, achieving comparable final performance to single-fidelity methods with substantially less wall-clock time. ASHA and ASHABORE reach low normalized regret levels with one to two orders of magnitude less budget consumed relative to their single-fidelity counterparts. BOHB performs competitively at early budgets but shows higher variance across tasks compared to ASHA-based methods. ASHACQR combines ASHA's aggressive early stopping with CQR's acquisition and achieves strong anytime performance, mirroring the advantage of CQR observed in the single-fidelity setting. The rank plots (Figure \ref{fig:single-fidelity}e) show ASHABORE and ASHACQR in the top rank across the budget range, with ASHA performing well at very low budgets where its successive halving strategy most aggressively filters configurations. CQR without multi-fidelity scheduling serves as a reference and is consistently outperformed by all multi-fidelity methods at equivalent wall-clock budgets, confirming the value of exploiting the learning curve structure available in the benchmark. Still, at the end, CQR is able to catch up to the other methods and in some cases even achieves higher end-performance.
\\
\textbf{Transfer learning} methods warm-start search from related tasks. We use BoundingBox~\citep{perrone2019learning}, ZeroShot~\citep{wistuba2015sequential}, and Quantile Transfer~\citep{salinas2020quantile}. These methods (Figure \ref{fig:single-fidelity}c and \ref{fig:single-fidelity}f) initialize the search using observations from related source tasks, reducing the cold-start cost compared to methods that start from scratch. BoundingBox and Quantile Transfer achieve substantially lower regret compared to CQR, though performance converges around the same end-performance. We can observe gains from transfer learning across architectures, suggesting that the learned HP importance structure is partially portable: the ranking of hyperparameters is stable enough across different biological data types and model families for source task experience to translate meaningfully to new target settings. This is consistent with the permutation importance analysis (Section \ref{subsec:hp_importance}), which revealed consistent dominance of learning rate across architectures and datasets, providing a stable signal for transfer. ZeroShot evaluates the single best-predicted configuration from a historical portfolio without performing any target-task evaluations. While it provides a strong immediate starting point at virtually zero cost, it lacks the ability to adapt dynamically to the target task's specific loss landscape over time. Consequently, it is quickly outperformed by methods like BoundingBox and Quantile Transfer, which actively update their surrogates based on online evaluations.

\section{Conclusions}

We have presented \textbf{BBOmix}, a large-scale tabular benchmark for unsupervised HPO of AEs on multi-omics data. The benchmark comprises 105,000 training runs organized into 35 black-box tasks, covering four AE architectures across two datasets and seven modalities.
Our analysis demonstrated that reconstruction loss serves as a mostly strong, yet not ideal, proxy for downstream performance across most biological domains, validating its use as a cheap HPO criterion. However, notable exceptions (such as TCGA-DNA reflecting a gene mutation score) exist, underscoring the need for careful evaluation. Furthermore, our hyperparameter importance analysis revealed that parameters such as learning rate and dropout rate consistently dictate generalization performance, regardless of the specific biological modality or architecture.

Finally, our evaluation of HPO optimizers highlighted the substantial effectiveness of transfer learning and multi-fidelity methods in this domain. These methods allow researchers to rapidly identify high-performing configurations and exploit learned structural similarities across biological datasets at a fraction of the computational cost of standard single-fidelity methods.
In sum, \textbf{BBOmix} provides a foundation for the community to develop and evaluate HPO methods in the underexplored unsupervised setting, and that the biological application domain motivates further HPO research.

\section{Limitations and Future Work}

Our benchmark, while extensive, is subject to certain limitations. First, our evaluation focuses on simple AE variants. Exploring convolutional, cross-modal, or graph-based architectures for spatially resolved or networked omics data remains an important next step. Second, while reconstruction loss serves as a generally robust proxy, it is fundamentally agnostic to downstream utility, and the proxy breakdown observed in the TCGA-DNA task highlights the necessity of developing more domain-aware unsupervised training metrics. While our current benchmark focuses on two comprehensive real-world multi-omics datasets (TCGA and SCHC), future work should expand this evaluation to encompass a broader variety of biological data modalities and clinical applications. Furthermore, our findings regarding the transferability of hyperparameter importance across architectures suggest a promising direction for meta-learning. Notably, our experiments exclusively consider cross-architecture transfer, and future work should systematically compare this against same-architecture, cross-modality and cross-dataset transfer to establish which axis of task relatedness most reliably predicts the portability of hyperparameter importance structure. We envision the development of a generalized, domain-specific HPO recommendation framework that leverages historical evaluations across diverse omics tasks to automatically suggest optimal architectures and hyperparameters for novel biological datasets, significantly reducing computational cost and lowering the barrier of entry for practitioners.

\begin{acknowledgements}
The authors gratefully acknowledge the computing time made available to them on the high-performance computer at the NHR Center of TU Dresden. This center is jointly supported by the Federal Ministry of Research, Technology and Space of Germany and the state governments participating in the NHR. Luca Thale-Bombien was further
supported by the BMFTR through a scholarship of DAAD project 57616814 (SECAI, School of Embedded and Composite AI) as part of the program Konrad Zuse Schools of Excellence in Artificial Intelligence.
 Aaron Klein acknowledges support by the EC under the grant No. 101195233 (OpenEuroLLM).
\end{acknowledgements}
% ==== Formatting Instructions
% The page limit for the main paper is 9 pages; this does not include the
% references or appendix.  The references and appendix are not limited in
% length. Accepted papers will be allowed to add an additional page of content
% to the main paper to react to reviewer feedback.

% This example section may be removed.

% ==== Bibliography
% print bibliography -- for bibtex / natbib, use:

% \bibliography{...}

% and for biber / biblatex, use:

% \printbibliography

% supplemental material -- everything hereafter will be suppressed during
% submission time if the hidesupplement option is provided!

\newpage

\newpage
\appendix

% \supplemental material can be placed here; this material will be hidden if the
% [hidesupplement] option is provided

\section{Clinical Annotations for Downstream Tasks}
\label{app:clinical_annotations}
To evaluate the downstream utility of the learned latent spaces, we employed classification tasks derived from the clinical and biological metadata accompanying each dataset. Performance was measured via the Area Under the Receiver Operating Characteristic curve (AUC-ROC).

\subsection{The Cancer Genome Atlas (TCGA)}
The TCGA dataset includes extensive patient-level clinical data. We utilized the following targets:
\begin{itemize}
    \item \textbf{Sex}: Binary classification of the patient's biological sex.
    \item \textbf{Cancer type}: Multi-class classification of the primary tumor type based on tissue origin (e.g., BRCA, LUAD).
    \item \textbf{Subtype}: Multi-class classification of the molecular or histological subtype.
    \item \textbf{Oncotree code}: Highly granular classification of the cancer lineage based on the OncoTree ontology.
    \item \textbf{AJCC stage}: Multi-class classification of the overall tumor stage (I, II, III, IV).
    \item \textbf{Path N stage}: Classification of regional lymph node involvement indicating tumor spread.
    \item \textbf{Grade}: Classification of the histological grade representing cellular abnormality.
    \item \textbf{OS status}: Binary classification of Overall Survival (living vs. deceased).
    \item \textbf{DSS status}: Binary classification of Disease-Specific Survival (death specifically attributed to the cancer).
\end{itemize}

\subsection{Single-Cell Human Cortex (SCHC)}
The SCHC dataset provides cell-level metadata. The downstream targets were:
\begin{itemize}
    \item \textbf{Cell type}: Multi-class classification of the specific neural or glial cell type (e.g., excitatory neurons, astrocytes) assigned by the original authors.
    \item \textbf{Age group}: Classification of the donor's developmental stage (e.g., fetal, adult).
    \item \textbf{Sex}: Binary classification of the donor's biological sex.
\end{itemize}

\section{Hyperparameter Importance per Architecture}
\label{app:hp-importance}

\begin{figure}[t]
    \centering
    \begin{subfigure}[t]{0.24\linewidth}
        \includegraphics[width=\linewidth]{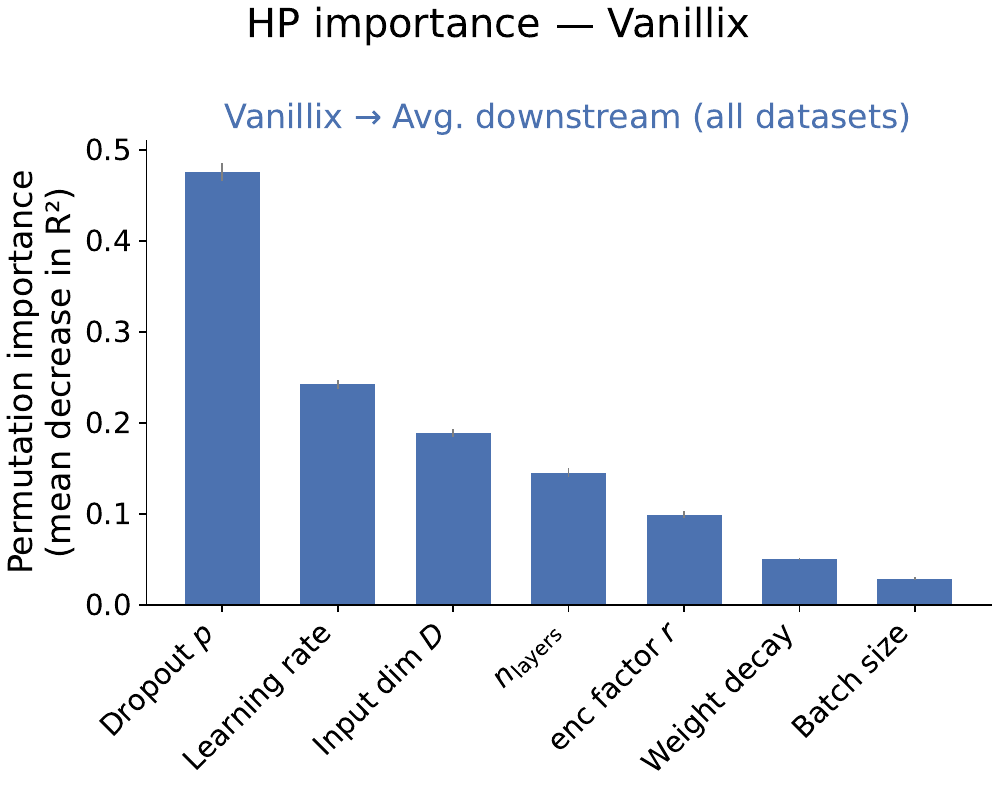}
        \caption{Downstream}
    \end{subfigure}\hfill
    \begin{subfigure}[t]{0.24\linewidth}
        \includegraphics[width=\linewidth]{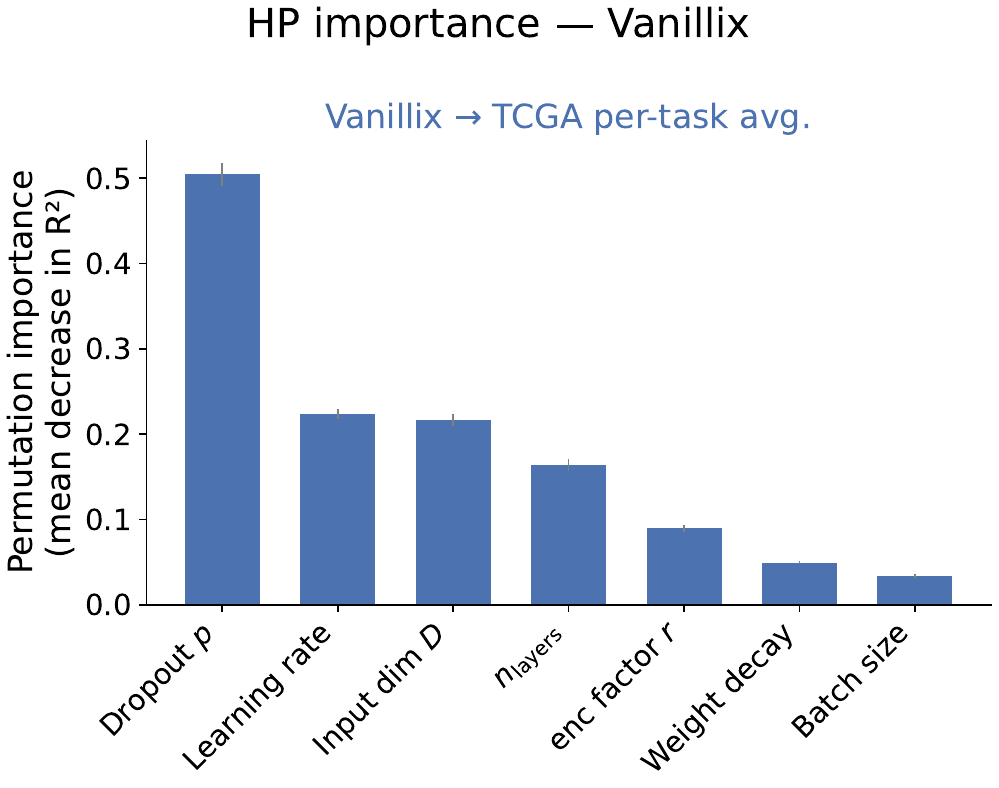}
        \caption{TCGA per-task avg.}
    \end{subfigure}\hfill
    \begin{subfigure}[t]{0.24\linewidth}
        \includegraphics[width=\linewidth]{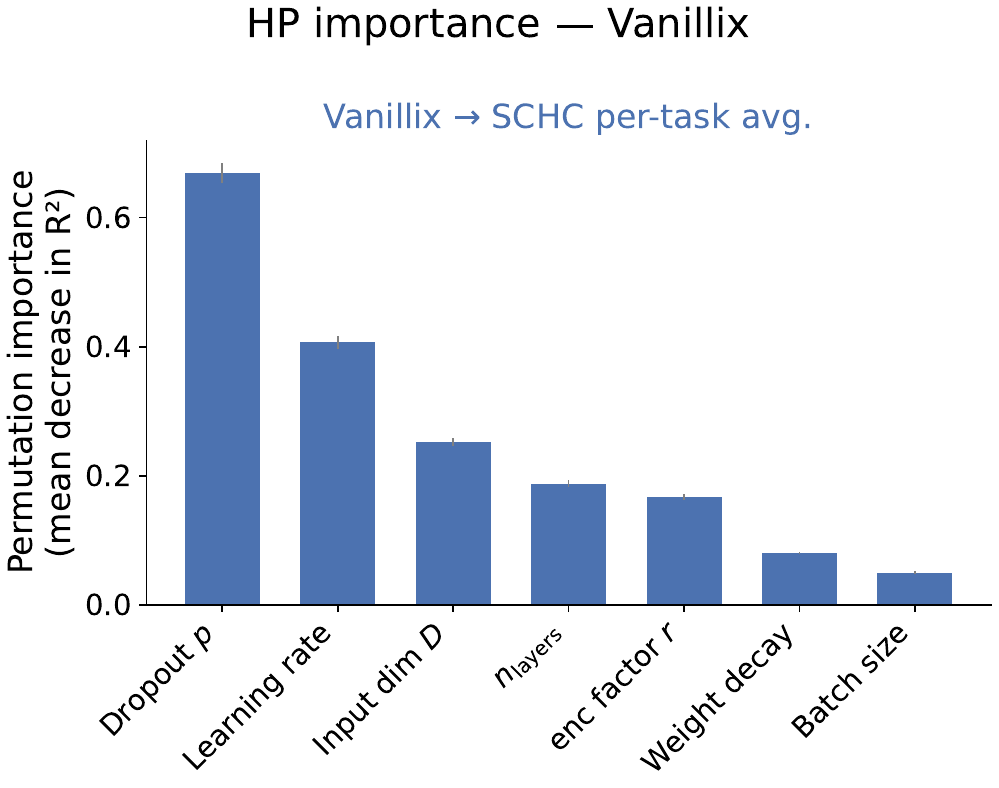}
        \caption{SCHC per-task avg.}
    \end{subfigure}\hfill
    \begin{subfigure}[t]{0.24\linewidth}
        \includegraphics[width=\linewidth]{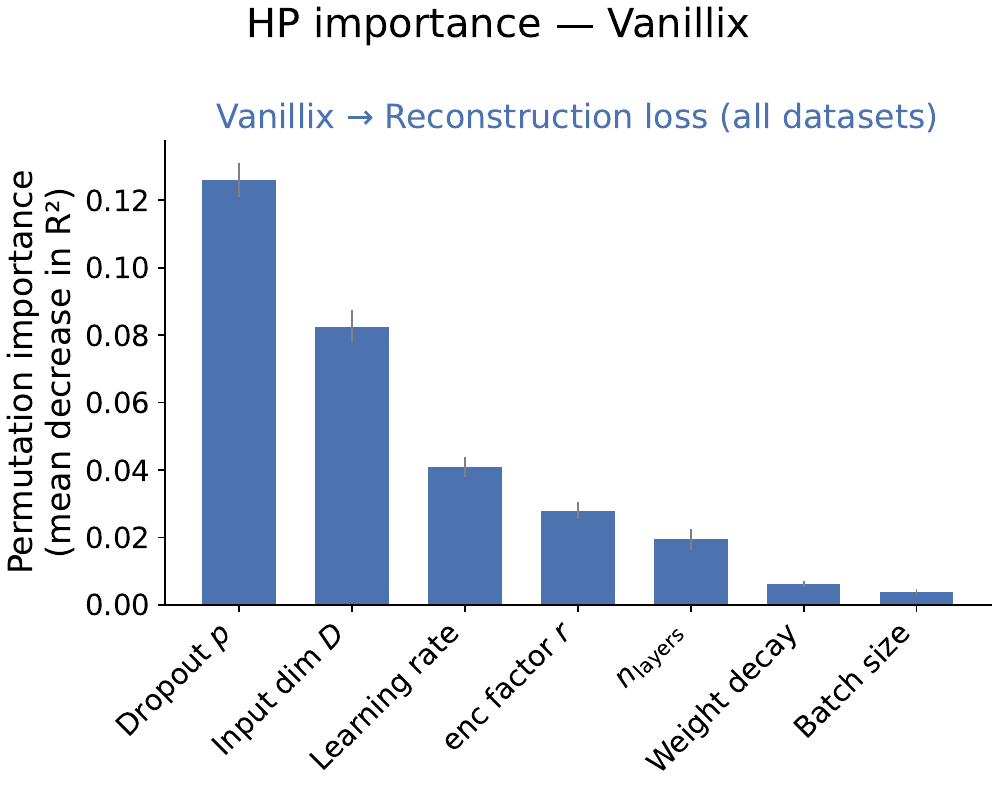}
        \caption{Reconstruction loss}
    \end{subfigure}
    \caption{HP importance Vanillix. Dropout $p$ dominates downstream performance
    across both datasets, followed by learning rate. Only for Reconstruction loss, the Input Dimensionality $D$ is ranked second.}
    \label{fig:app-hp-vanillix}
\end{figure}

\begin{figure}[t]
    \centering
    \begin{subfigure}[t]{0.24\linewidth}
        \includegraphics[width=\linewidth]{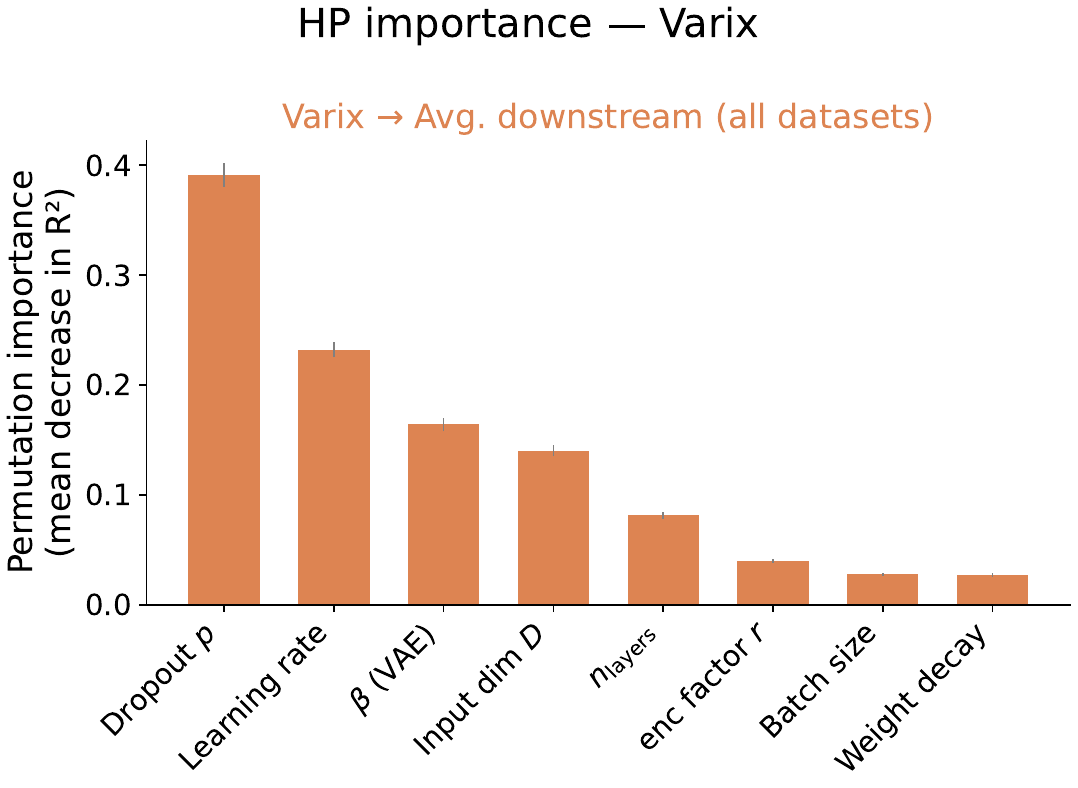}
        \caption{Downstream}
    \end{subfigure}\hfill
    \begin{subfigure}[t]{0.24\linewidth}
        \includegraphics[width=\linewidth]{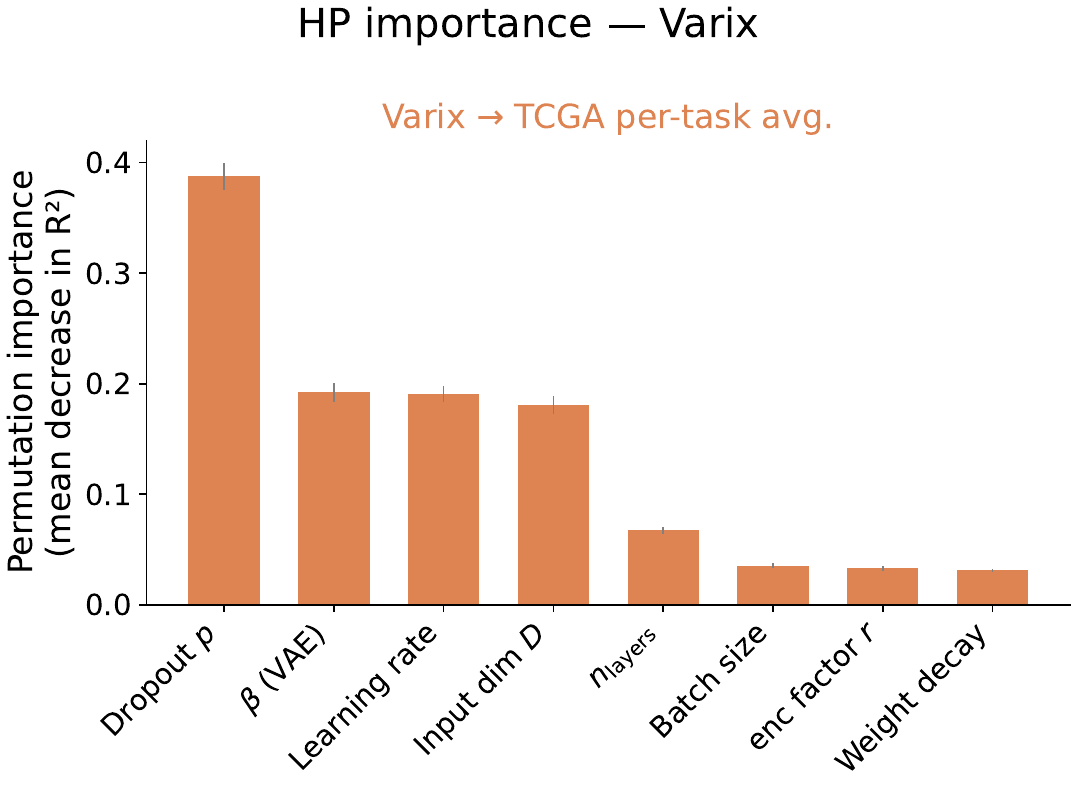}
        \caption{TCGA per-task avg.}
    \end{subfigure}\hfill
    \begin{subfigure}[t]{0.24\linewidth}
        \includegraphics[width=\linewidth]{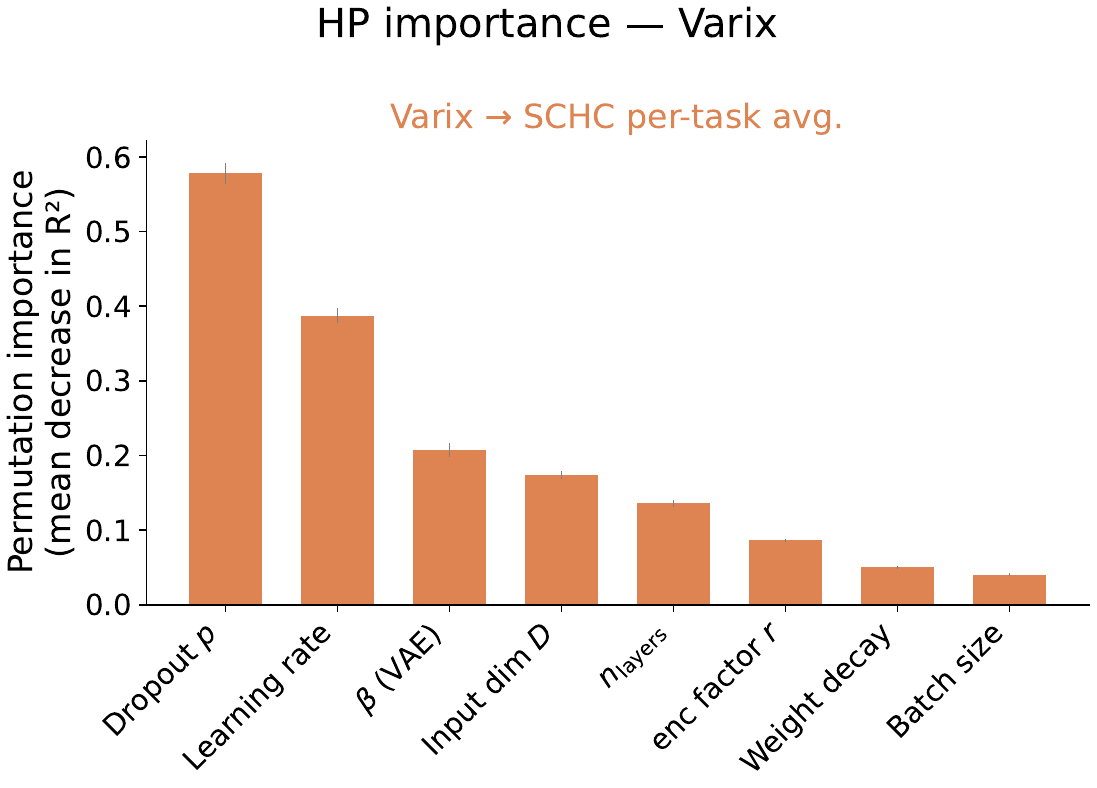}
        \caption{SCHC per-task avg.}
    \end{subfigure}\hfill
    \begin{subfigure}[t]{0.24\linewidth}
        \includegraphics[width=\linewidth]{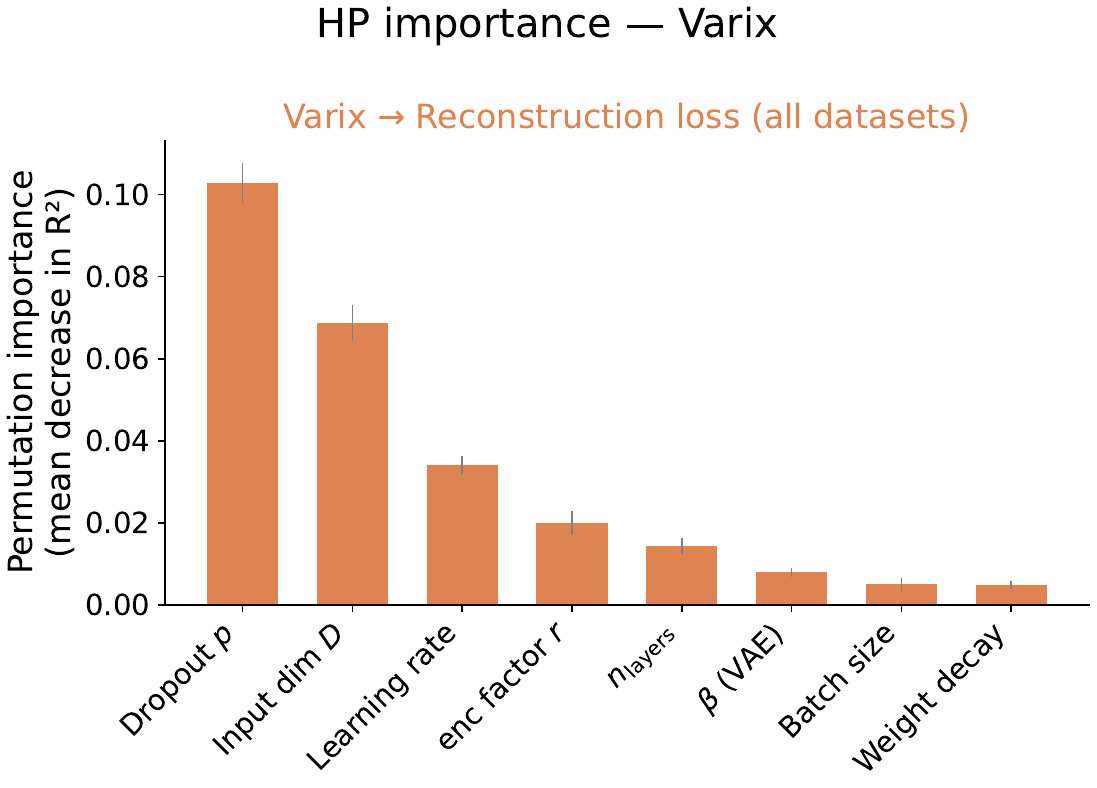}
        \caption{Reconstruction loss}
    \end{subfigure}
    \caption{HP importance Varix. Dropout $p$ is dominant for downstream
    performance, followed by the learning rate and the architecture specific parameter $\beta$. For reconstruction loss, input dim $D$ rises to second place,
    and the VAE-specific $\beta$ term contributes only marginally.}
    \label{fig:app-hp-varix}
\end{figure}

\begin{figure}[t]
    \centering
    \begin{subfigure}[t]{0.24\linewidth}
        \includegraphics[width=\linewidth]{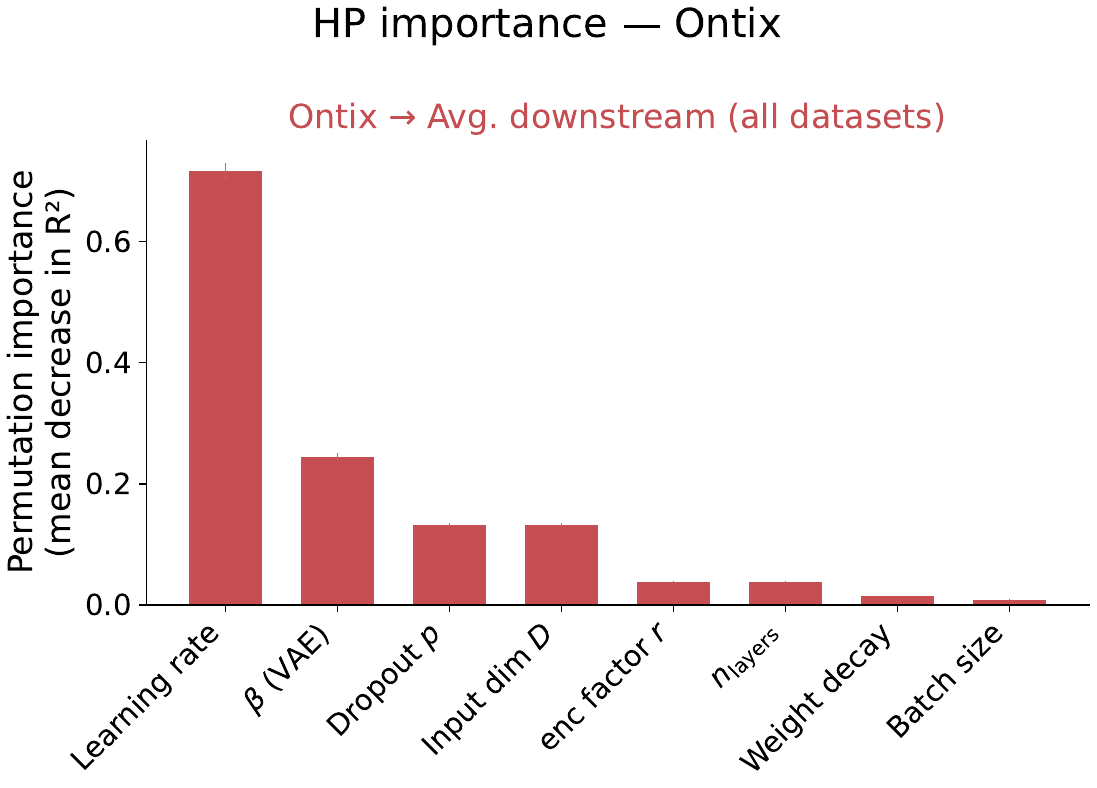}
        \caption{Downstream}
    \end{subfigure}\hfill
    \begin{subfigure}[t]{0.24\linewidth}
        \includegraphics[width=\linewidth]{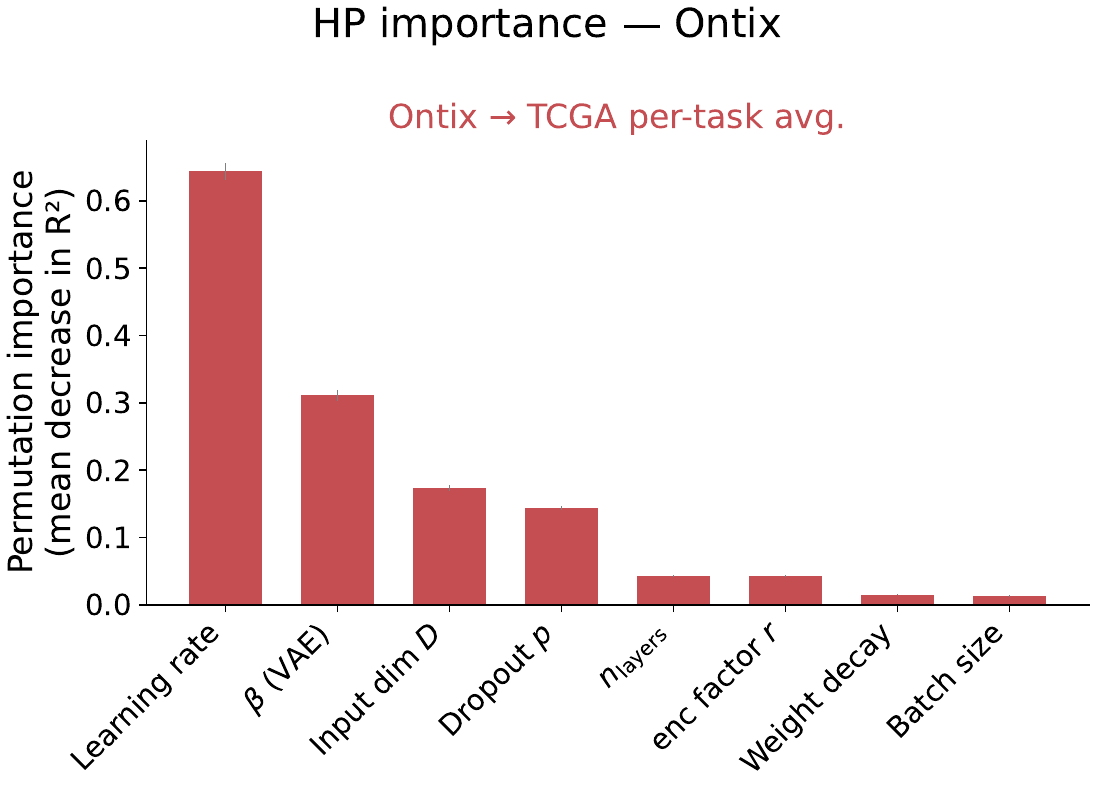}
        \caption{TCGA per-task avg.}
    \end{subfigure}\hfill
    \begin{subfigure}[t]{0.24\linewidth}
        \includegraphics[width=\linewidth]{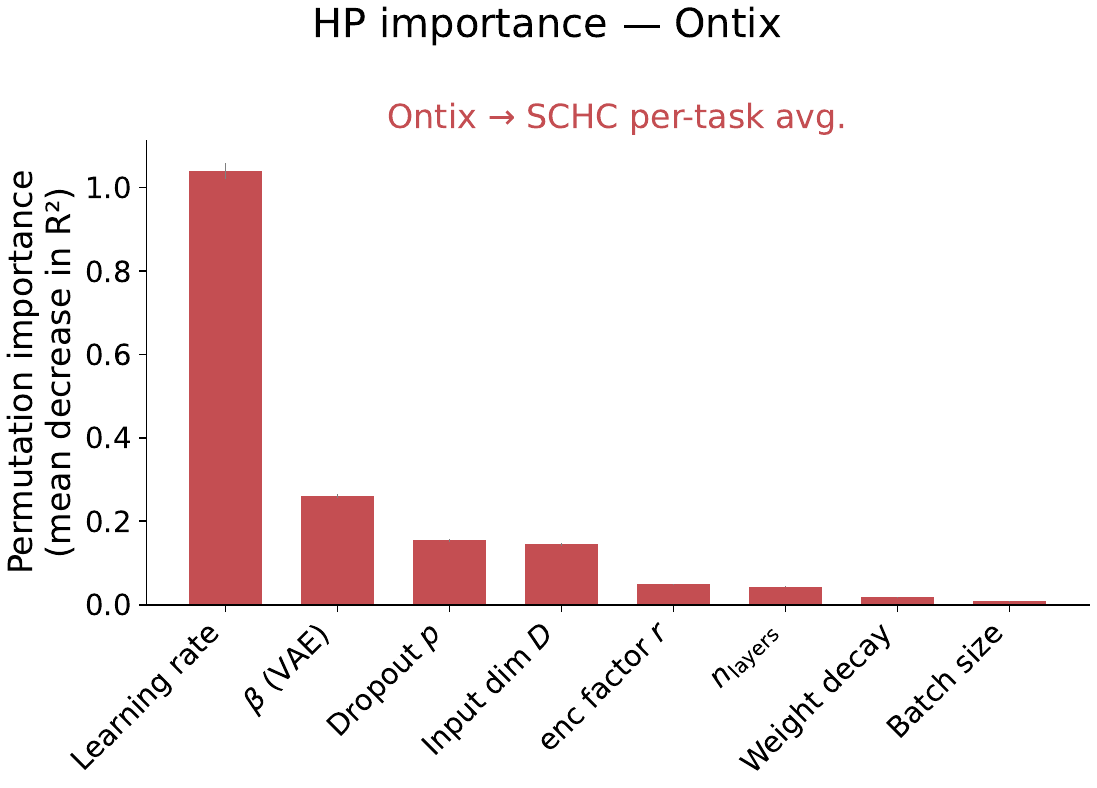}
        \caption{SCHC per-task avg.}
    \end{subfigure}\hfill
    \begin{subfigure}[t]{0.24\linewidth}
        \includegraphics[width=\linewidth]{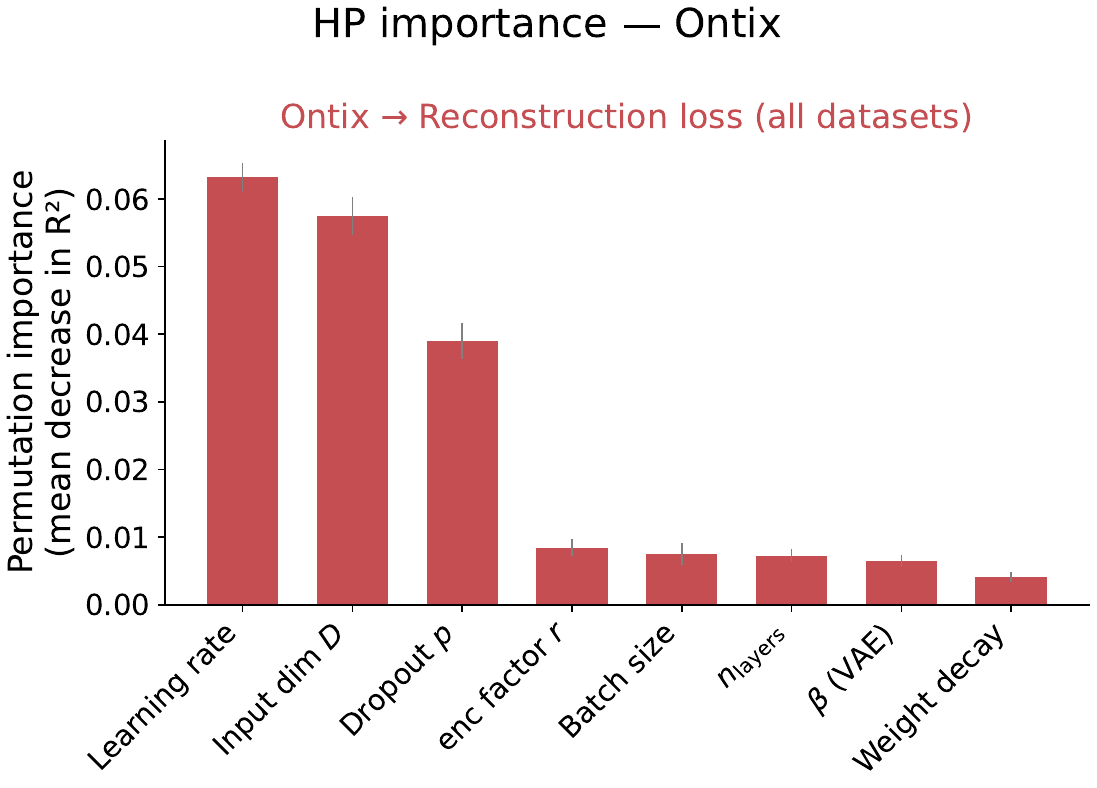}
        \caption{Reconstruction loss}
    \end{subfigure}
    \caption{HP importance Ontix. Learning rate is exceptionally dominant,
    accounting for the vast majority of explained variance in downstream performance
    on both SCHC and TCGA. $\beta$ (VAE) is the second most important HP, while
    dropout $p$ ranks third. For reconstruction loss, learning rate and input dim $D$
    are nearly equally important.}
    \label{fig:app-hp-ontix}
\end{figure}

\begin{figure}[t]
    \centering
    \begin{subfigure}[t]{0.24\linewidth}
        \includegraphics[width=\linewidth]{figures/hpo-figures/fig4_2_hp_importance_disentanglix_downstream.pdf}
        \caption{Downstream}
    \end{subfigure}\hfill
    \begin{subfigure}[t]{0.24\linewidth}
        \includegraphics[width=\linewidth]{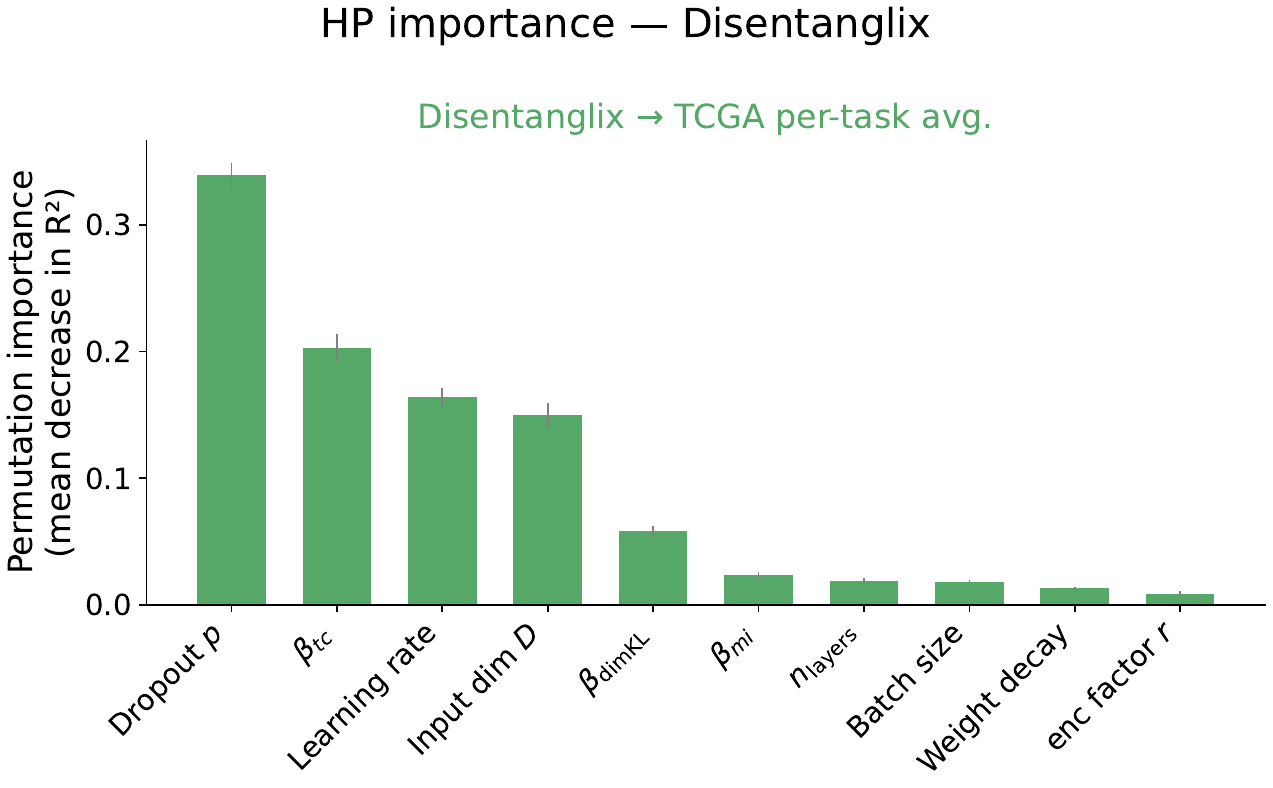}
        \caption{TCGA per-task avg.}
    \end{subfigure}\hfill
    \begin{subfigure}[t]{0.24\linewidth}
        \includegraphics[width=\linewidth]{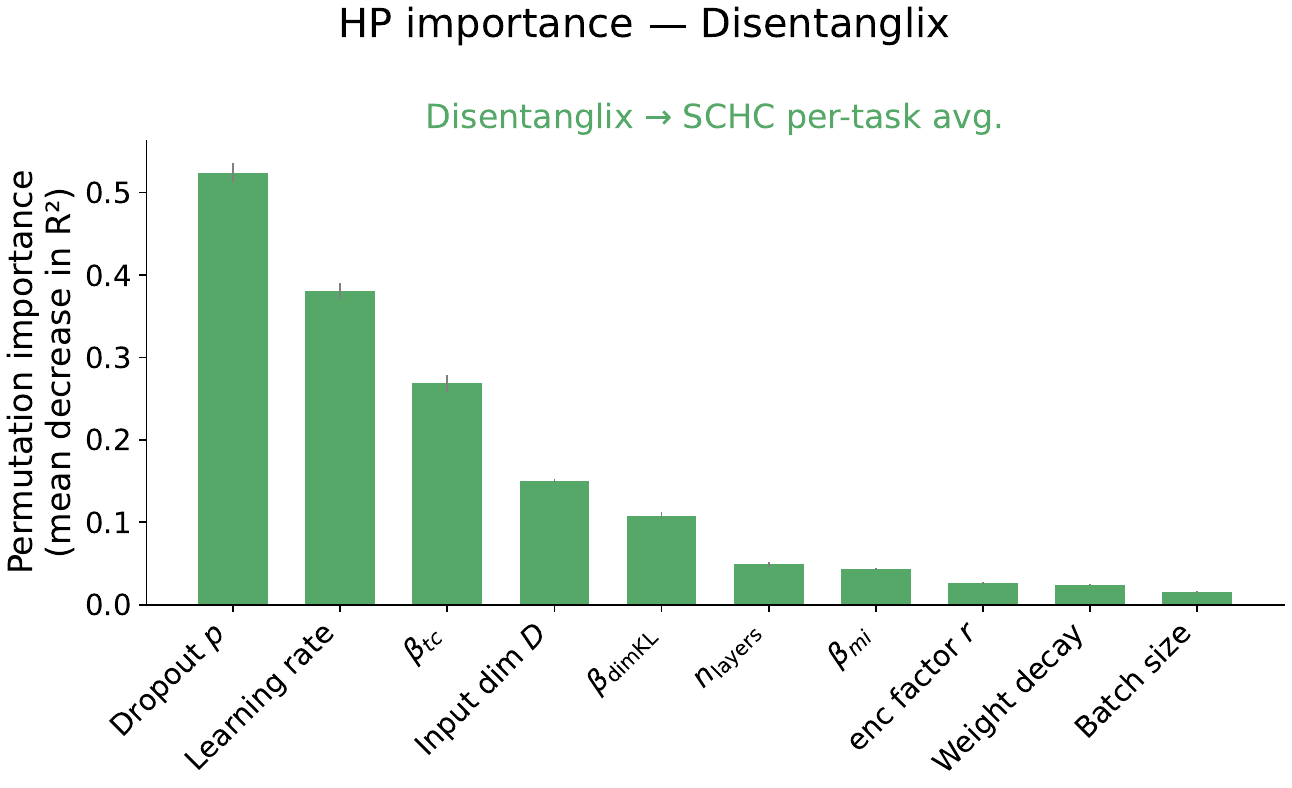}
        \caption{SCHC per-task avg.}
    \end{subfigure}\hfill
    \begin{subfigure}[t]{0.24\linewidth}
        \includegraphics[width=\linewidth]{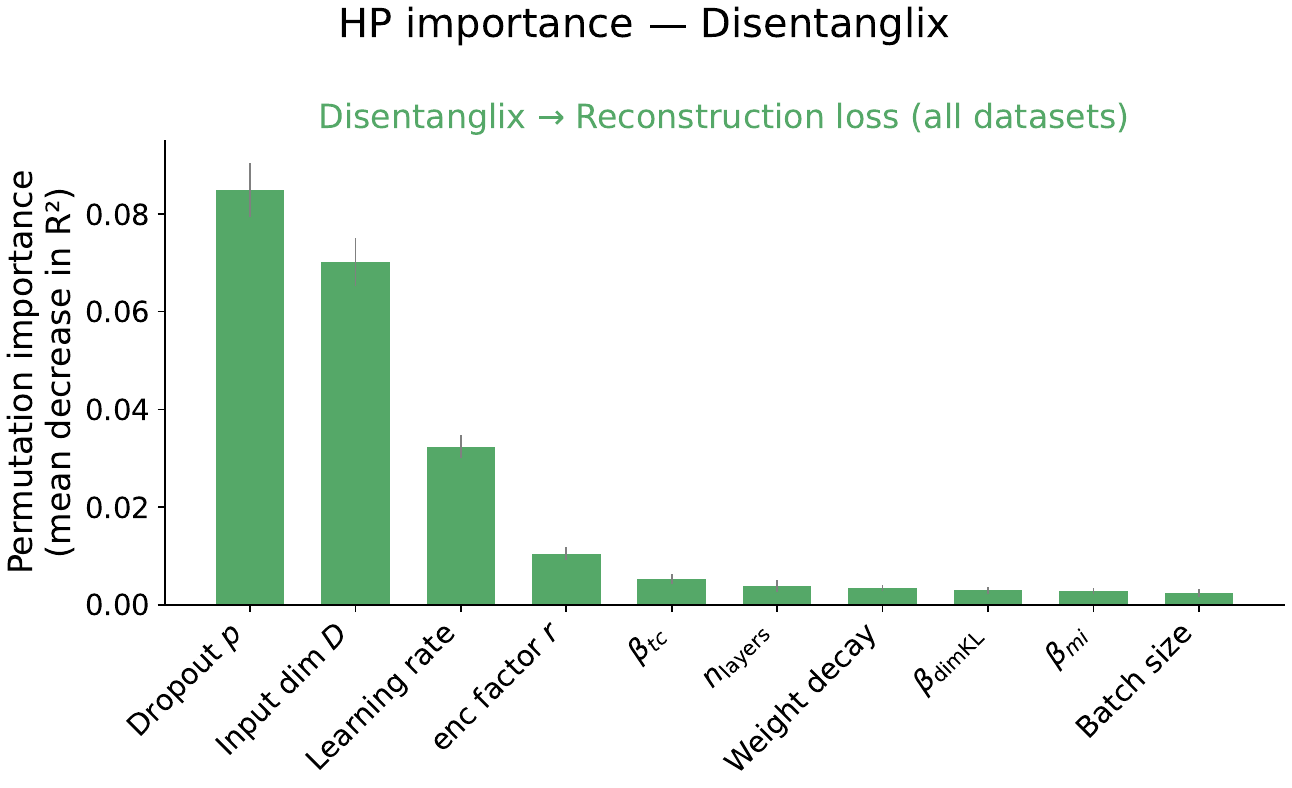}
        \caption{Reconstruction loss}
    \end{subfigure}
    \caption{HP importance Disentanglix. Dropout $p$ leads for downstream
    performance, followed by learning rate and $\beta_{tc}$, indicating that the
    total-correlation penalty meaningfully affects generalisation. For reconstruction
    loss, dropout $p$ and input dim $D$ dominate, while $\beta_{tc}$ is negligible.}
    \label{fig:app-hp-disentanglix}
\end{figure}

\begin{figure*}[t]
    \centering
    \includegraphics[width=0.24\linewidth]{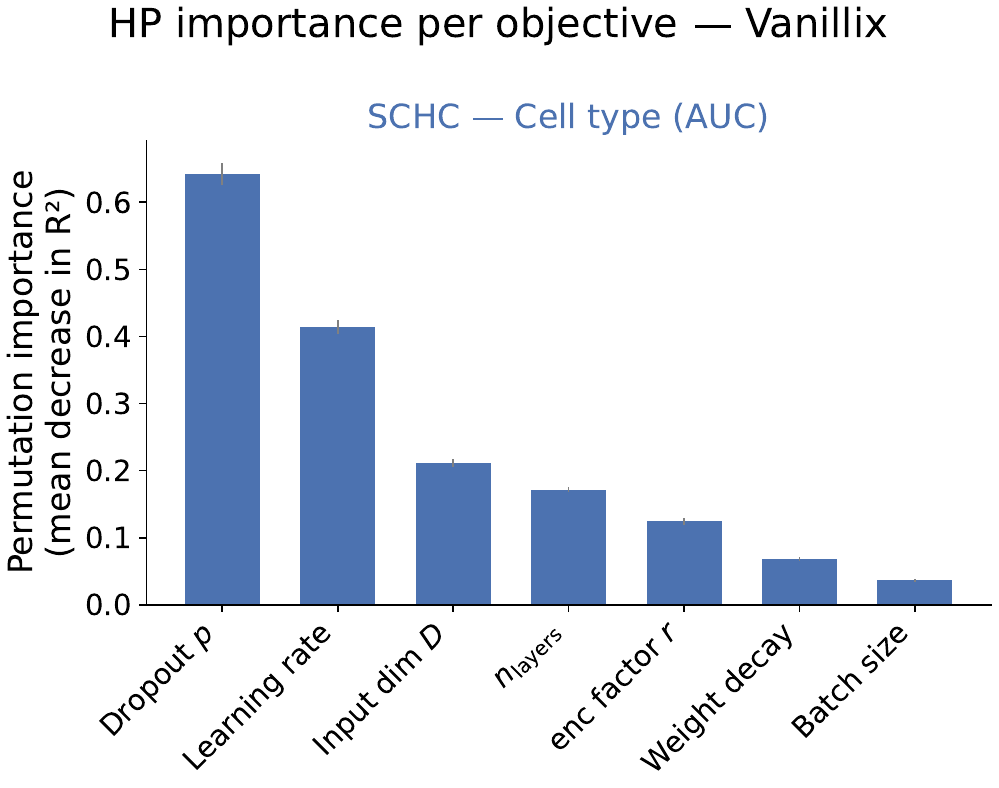}\hfill
    \includegraphics[width=0.24\linewidth]{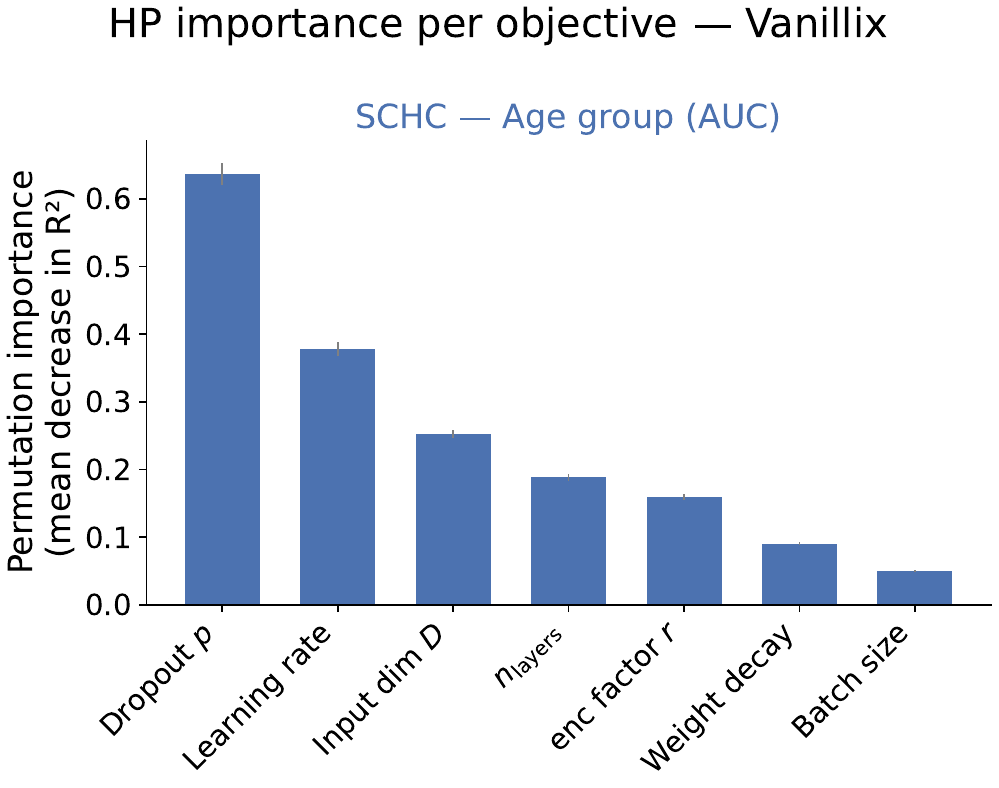}\hfill
    \includegraphics[width=0.24\linewidth]{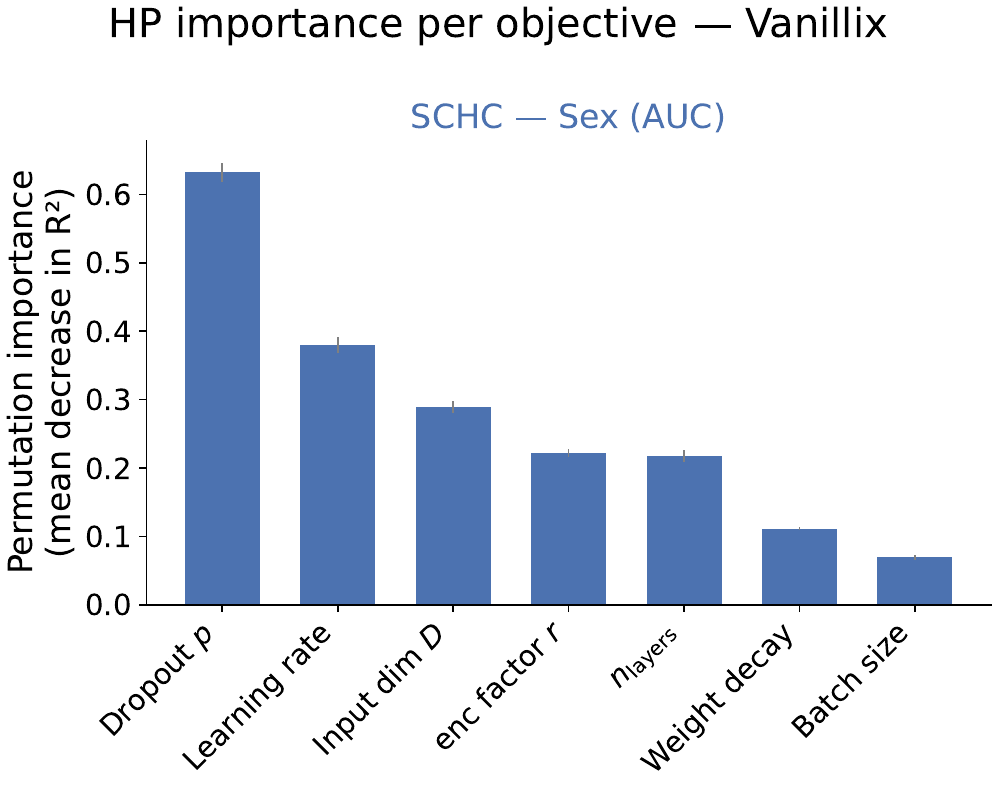}\hfill
    \includegraphics[width=0.24\linewidth]{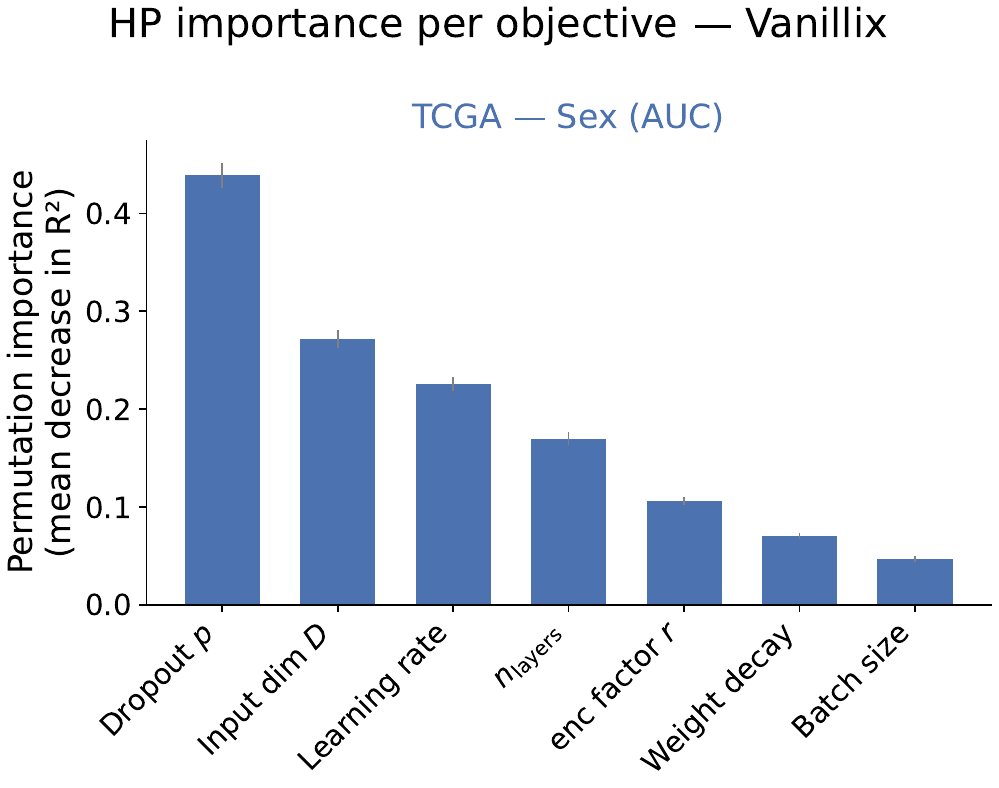}
    
    \vspace{0.2cm}
    \includegraphics[width=0.24\linewidth]{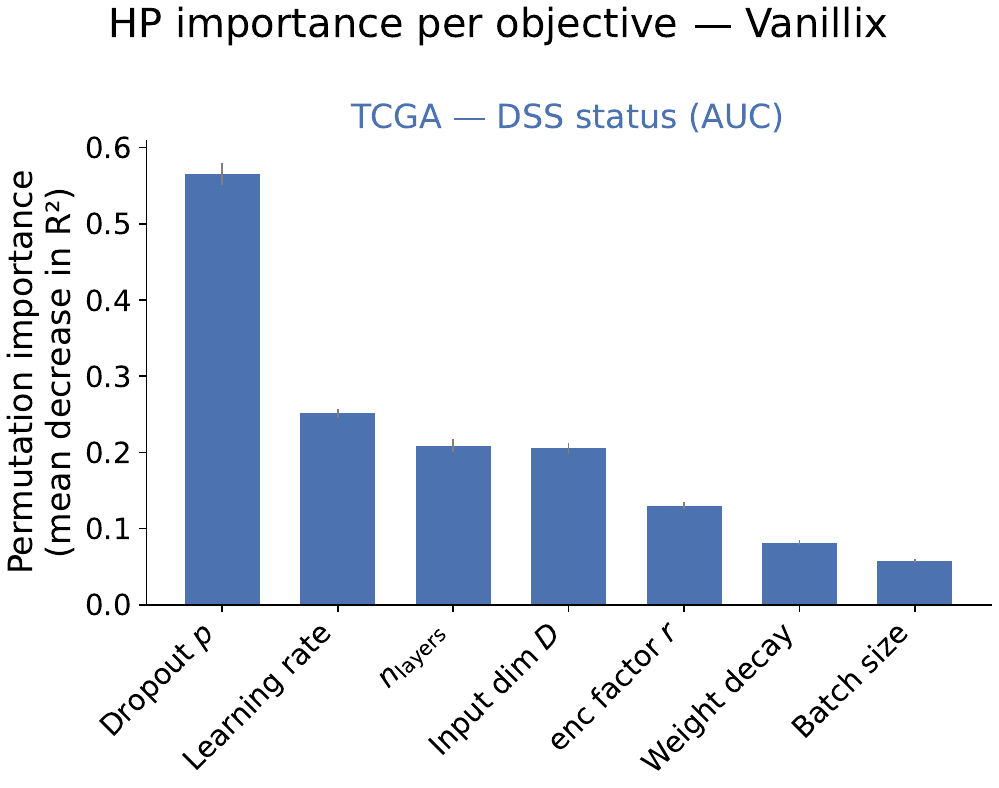}\hfill
    \includegraphics[width=0.24\linewidth]{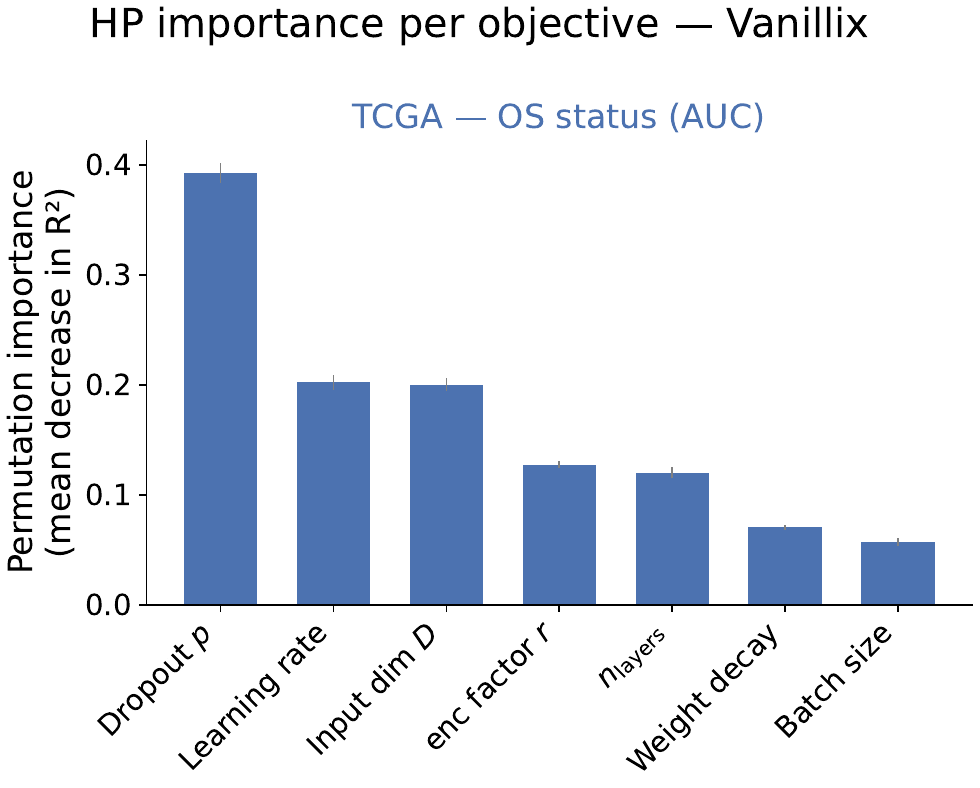}\hfill
    \includegraphics[width=0.24\linewidth]{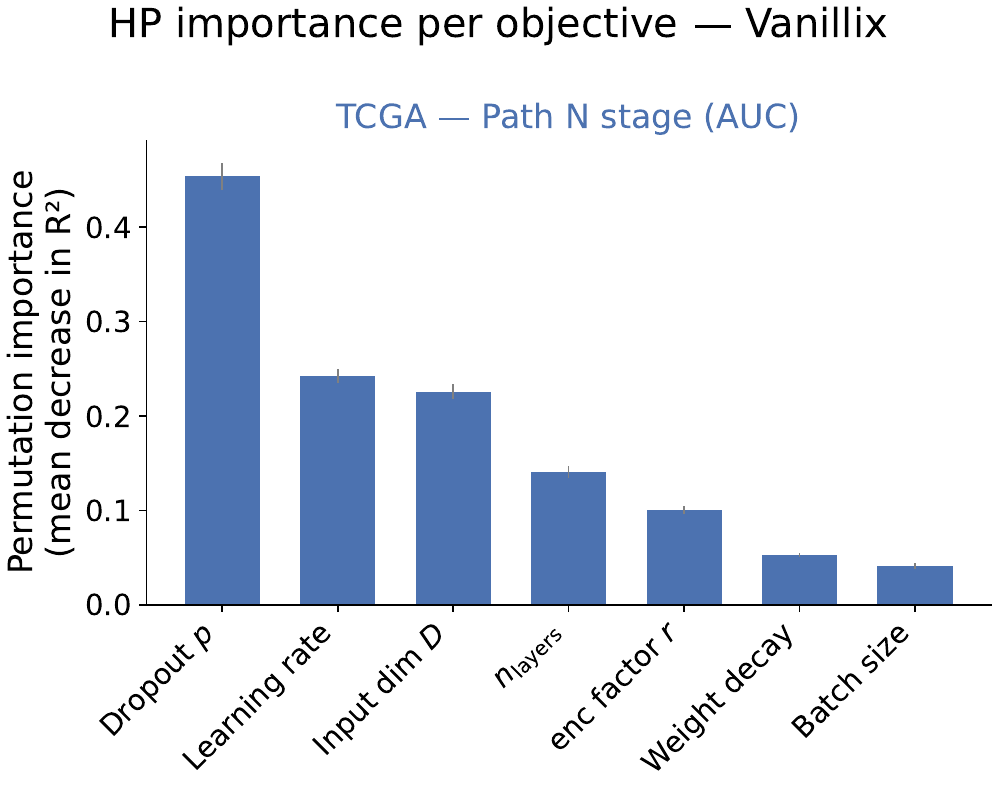}\hfill
    \includegraphics[width=0.24\linewidth]{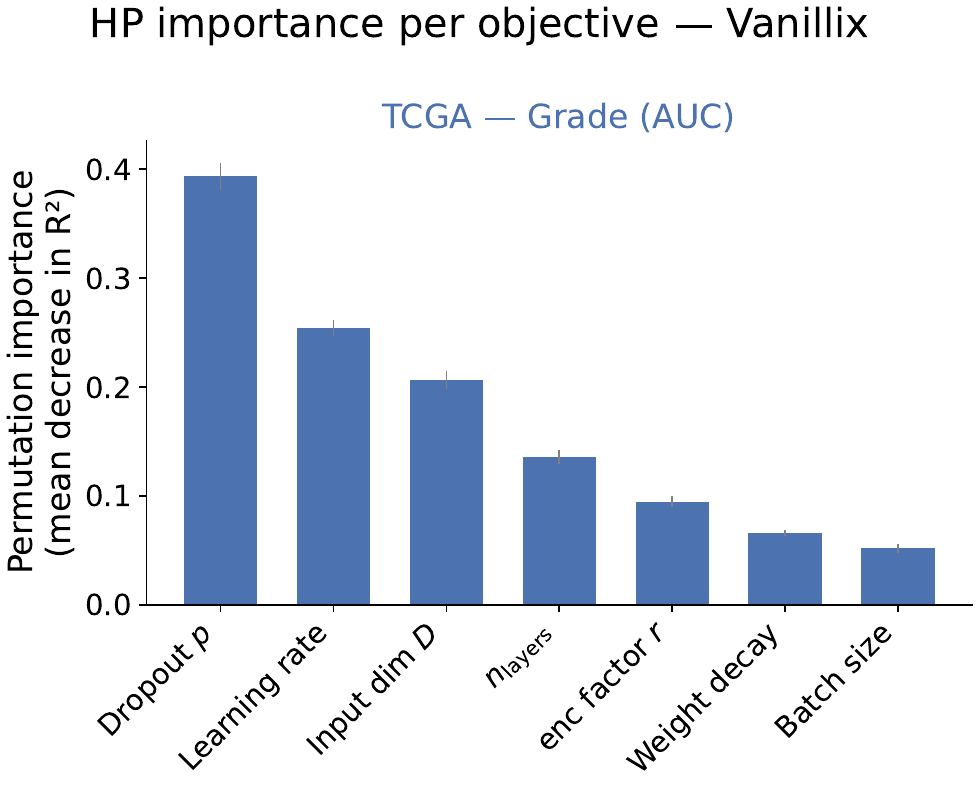}
    
    \vspace{0.2cm}
    \includegraphics[width=0.24\linewidth]{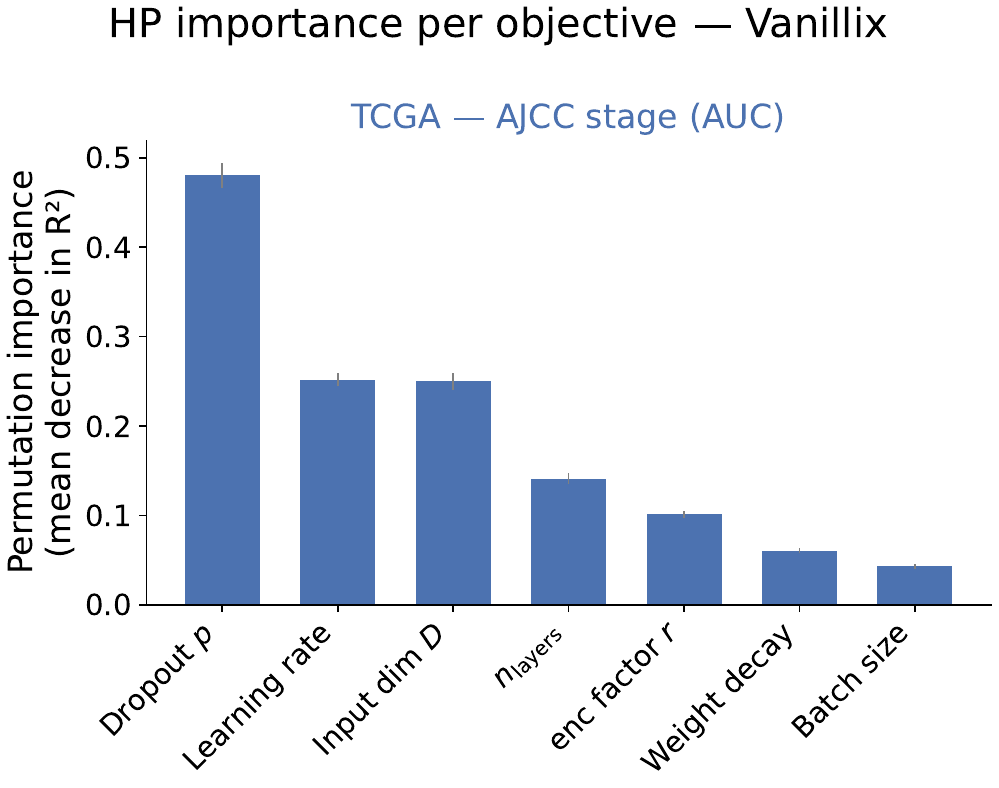}\hfill
    \includegraphics[width=0.24\linewidth]{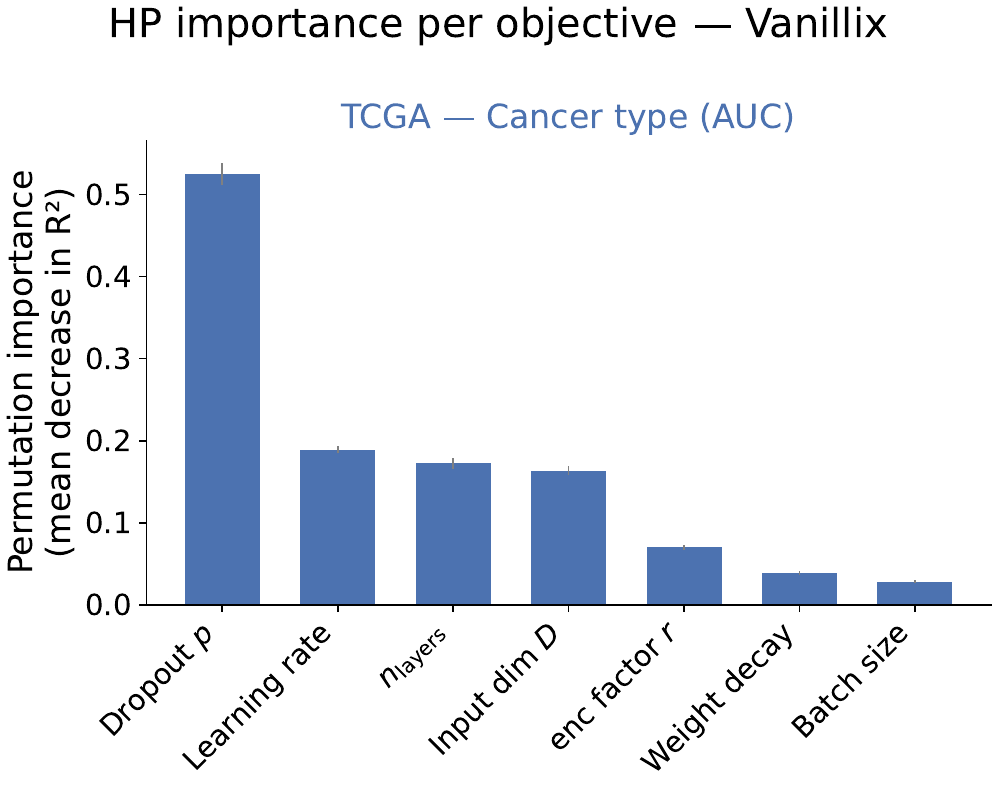}\hfill
    \includegraphics[width=0.24\linewidth]{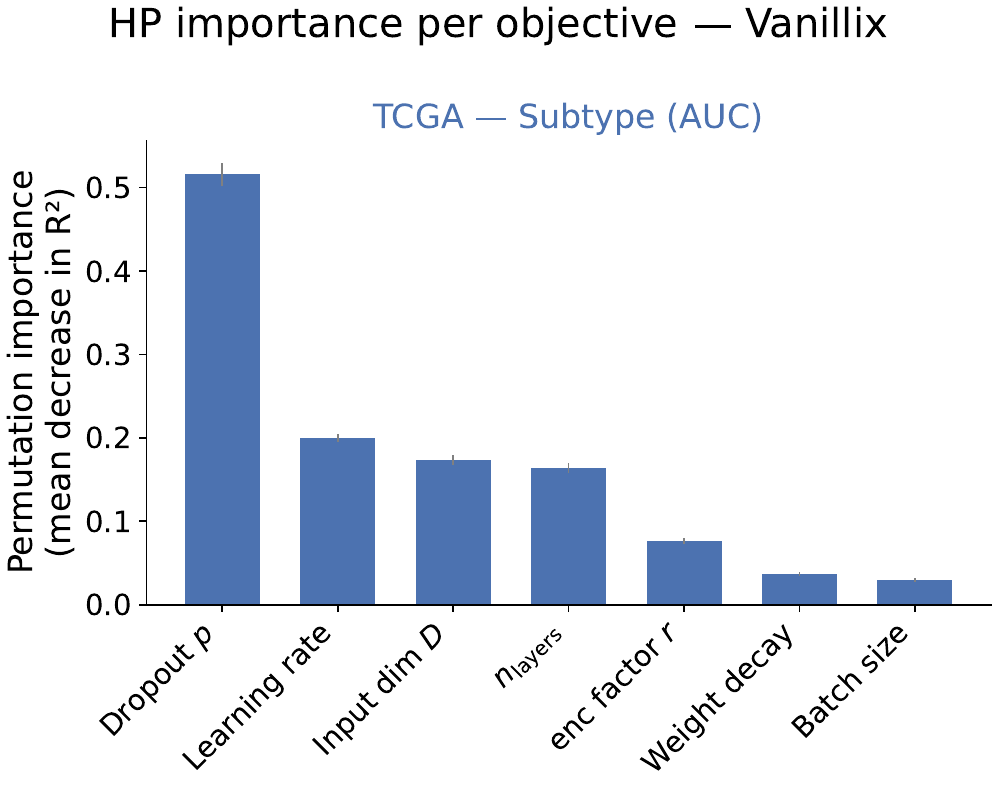}\hfill
    \includegraphics[width=0.24\linewidth]{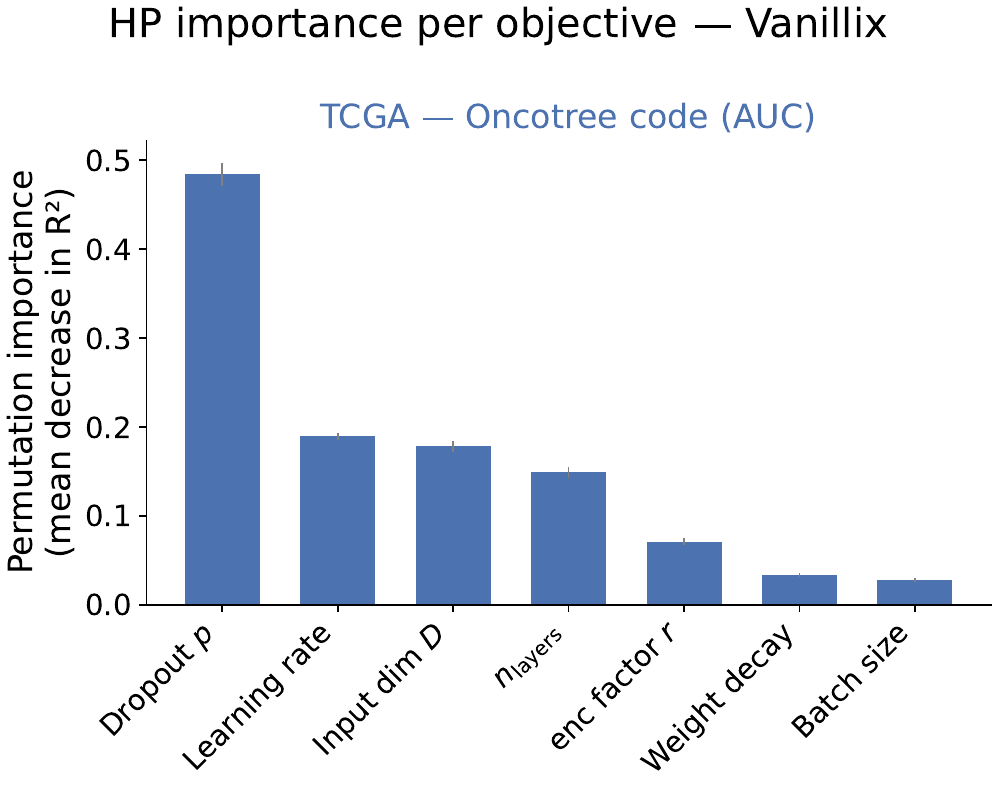}
    \caption{Per-task HP importance Vanillix. Dropout rate dominates consistently
    regardless of task, with $\beta$ (VAE) as a stable second contributor.}
    \label{fig:app-hp-pertask-vanillix}
\end{figure*}

\begin{figure*}[t]
    \centering
    \includegraphics[width=0.24\linewidth]{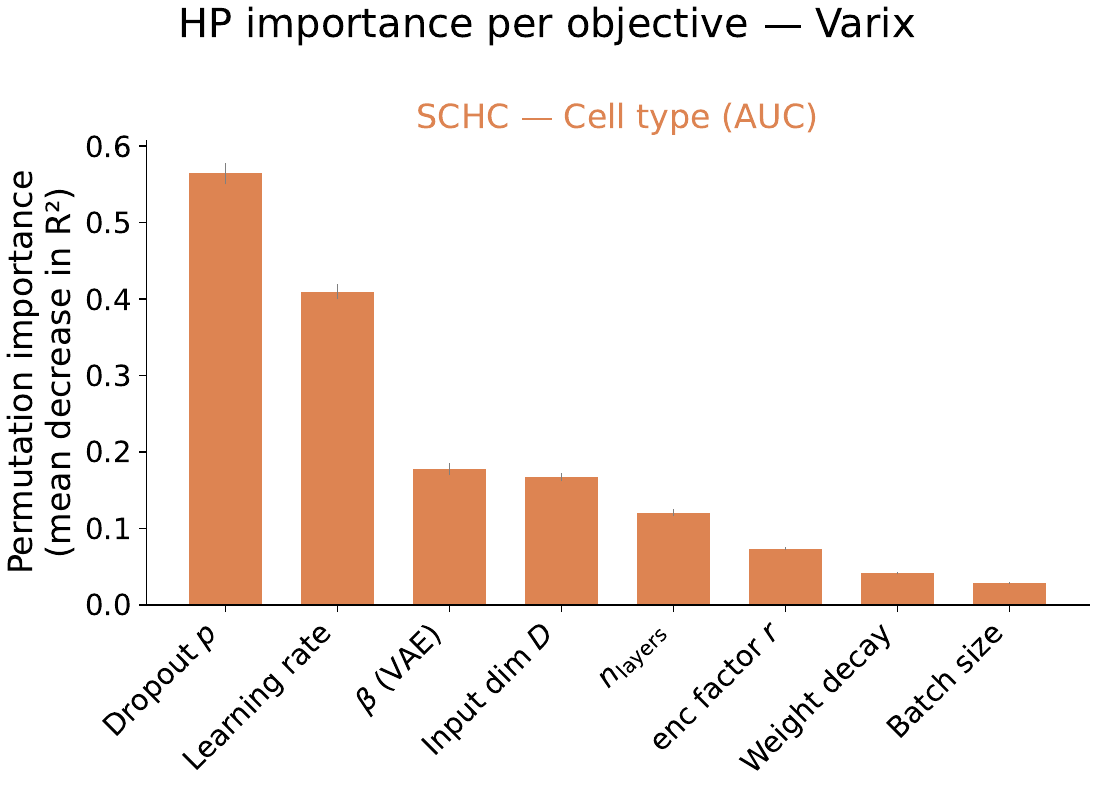}\hfill
    \includegraphics[width=0.24\linewidth]{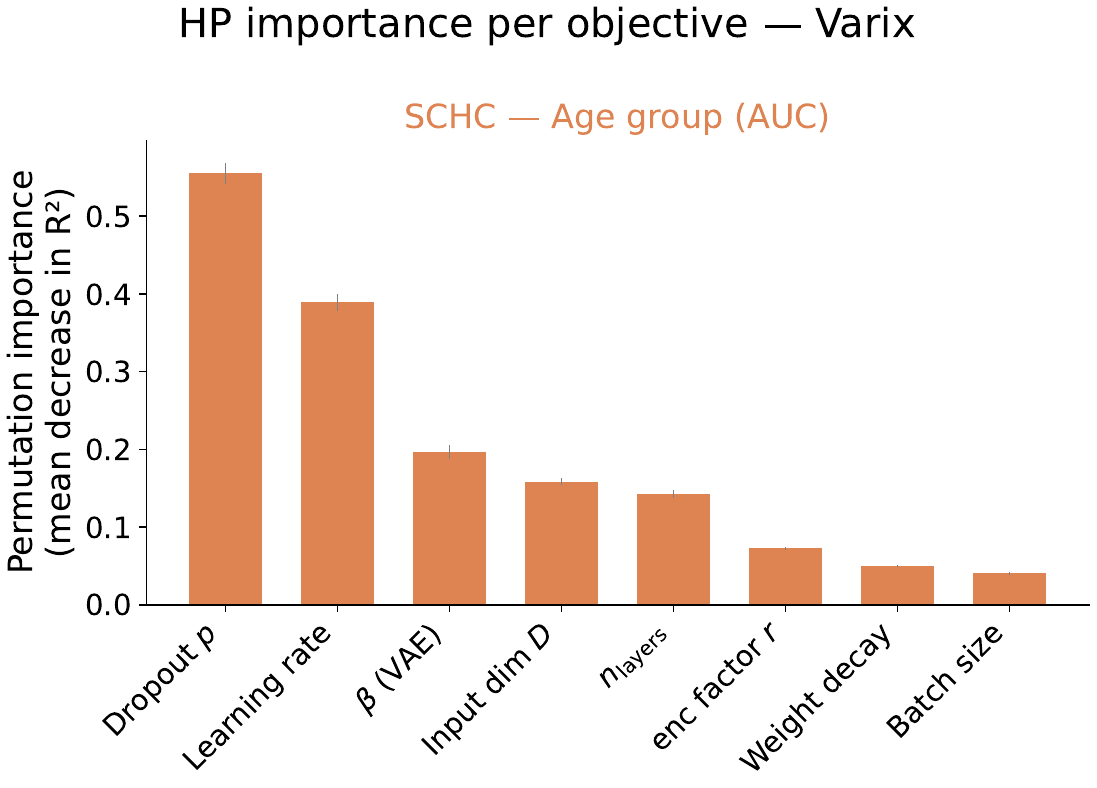}\hfill
    \includegraphics[width=0.24\linewidth]{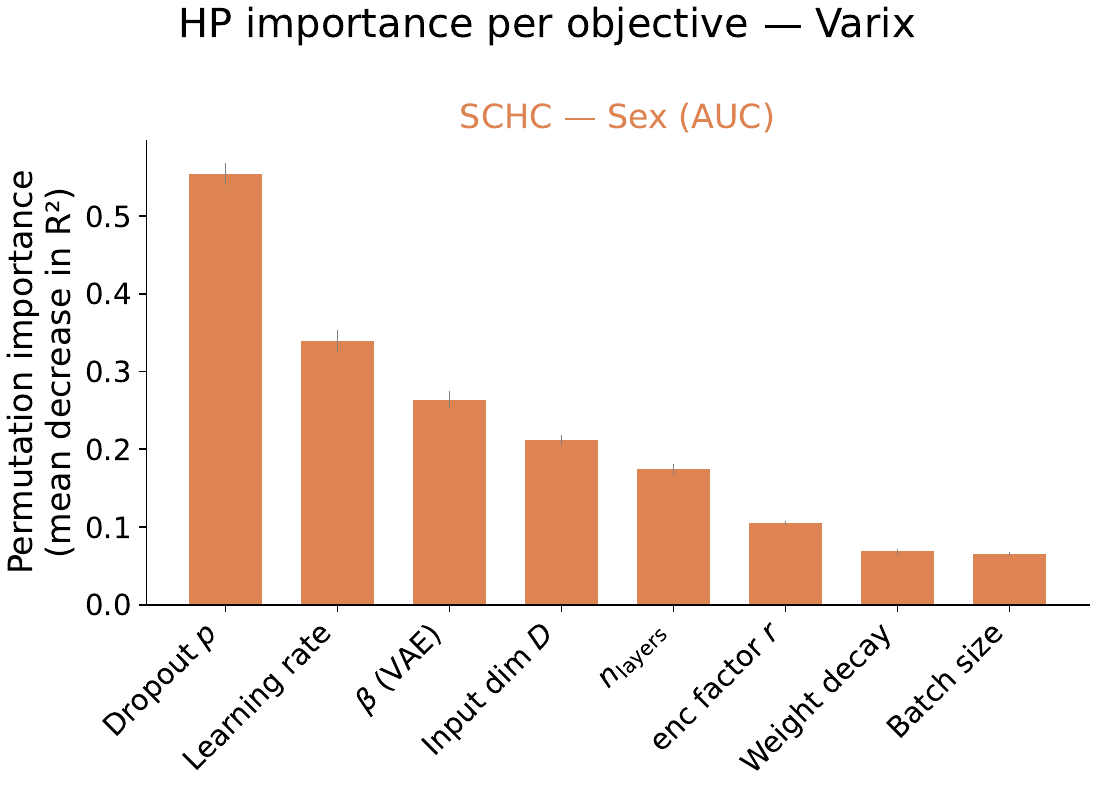}\hfill
    \includegraphics[width=0.24\linewidth]{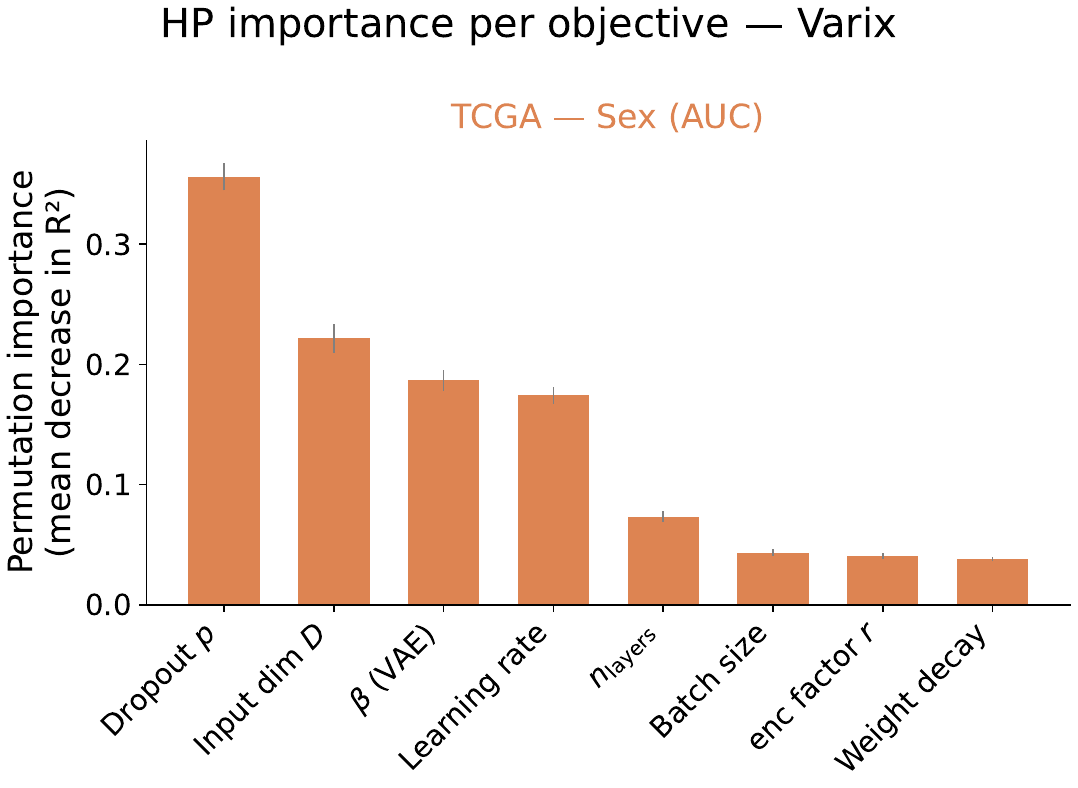}
    
    \vspace{0.2cm}
    \includegraphics[width=0.24\linewidth]{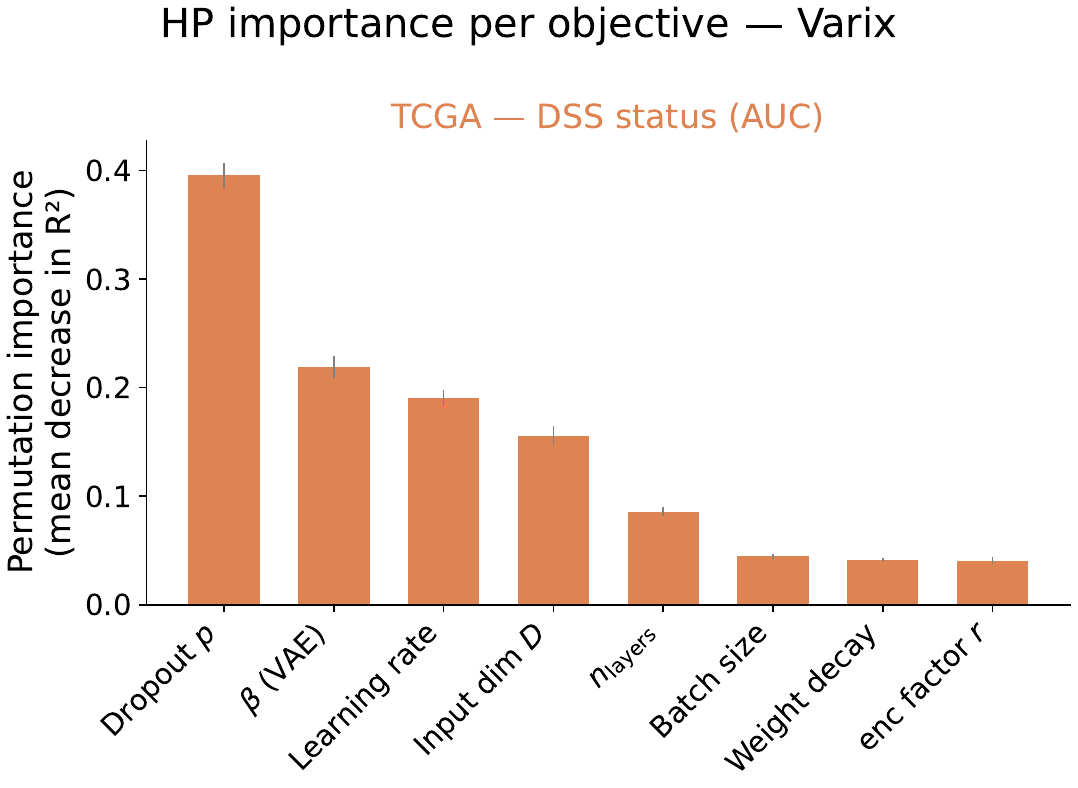}\hfill
    \includegraphics[width=0.24\linewidth]{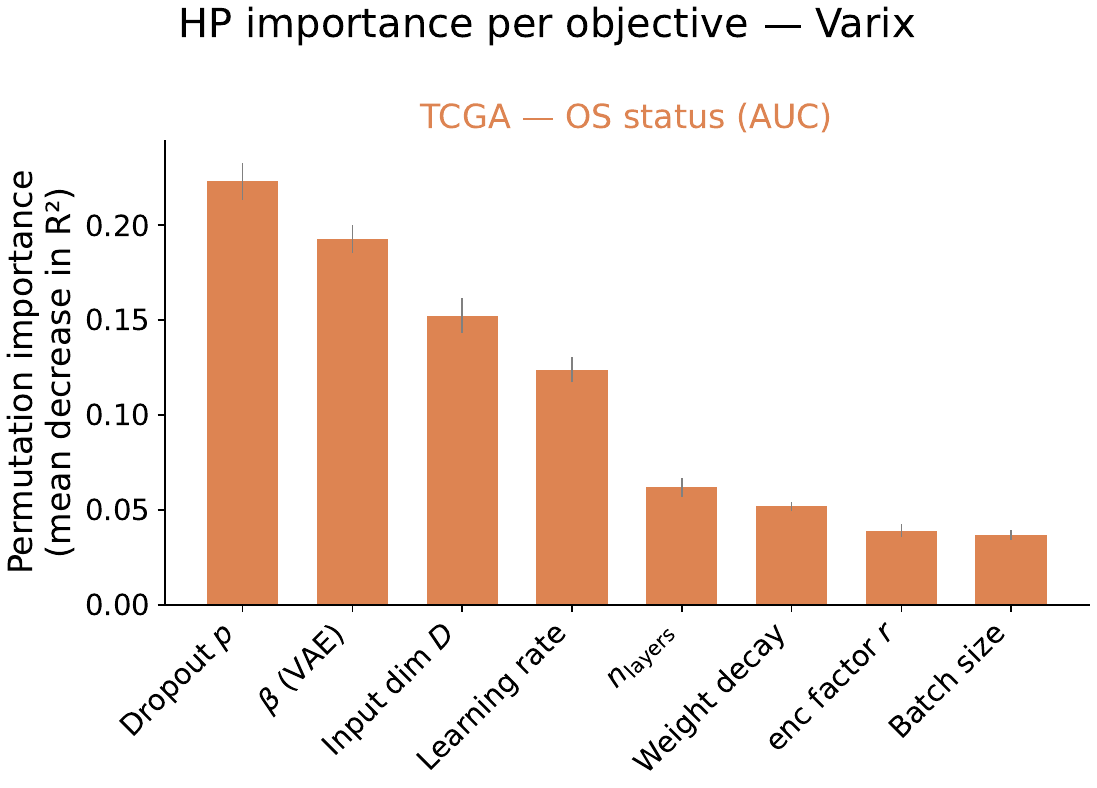}\hfill
    \includegraphics[width=0.24\linewidth]{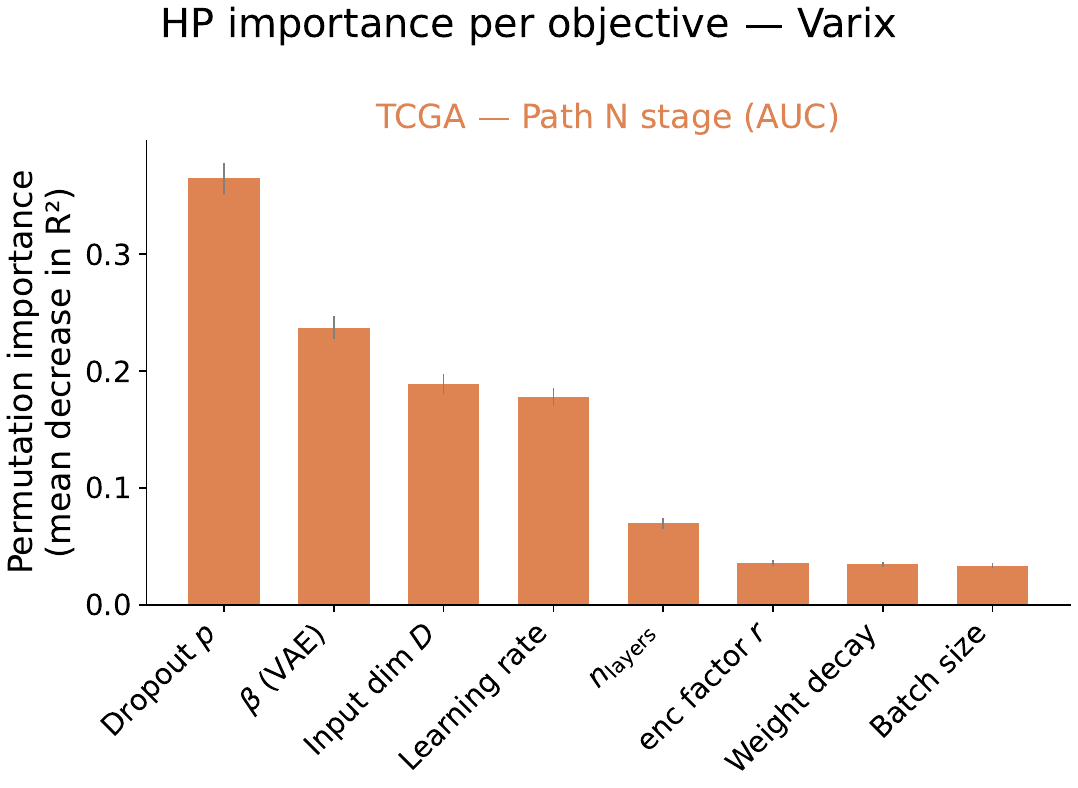}\hfill
    \includegraphics[width=0.24\linewidth]{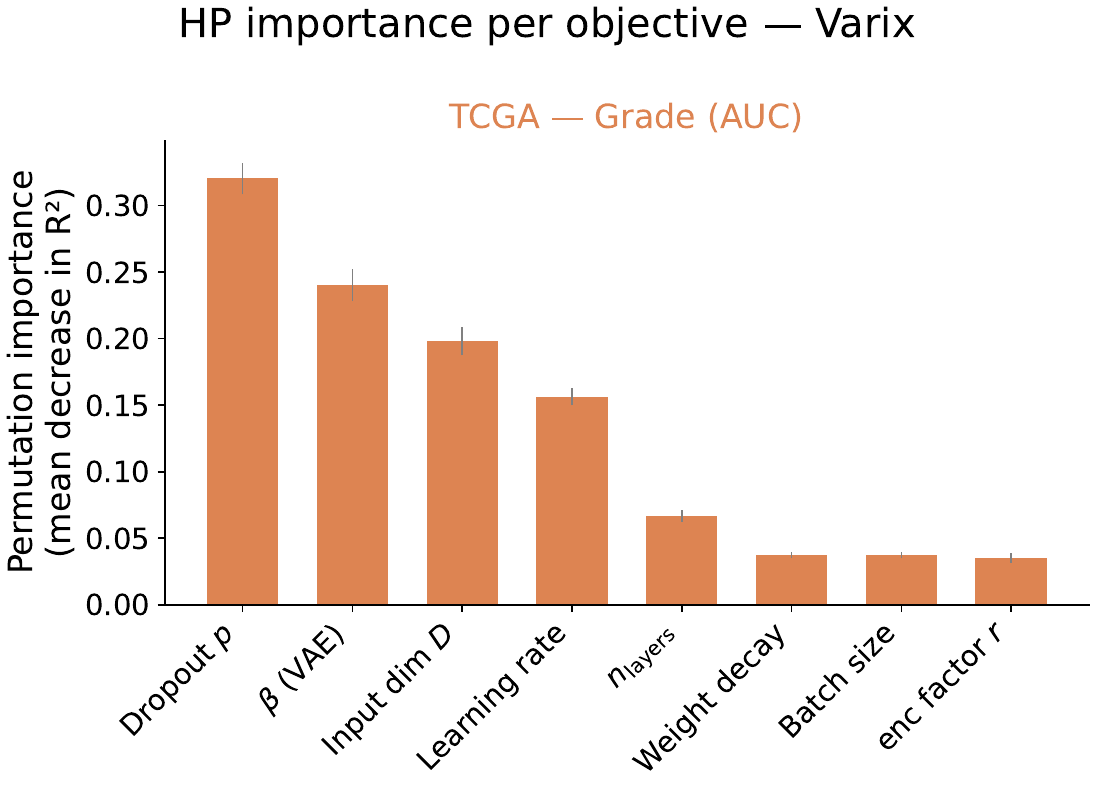}
    
    \vspace{0.2cm}
    \includegraphics[width=0.24\linewidth]{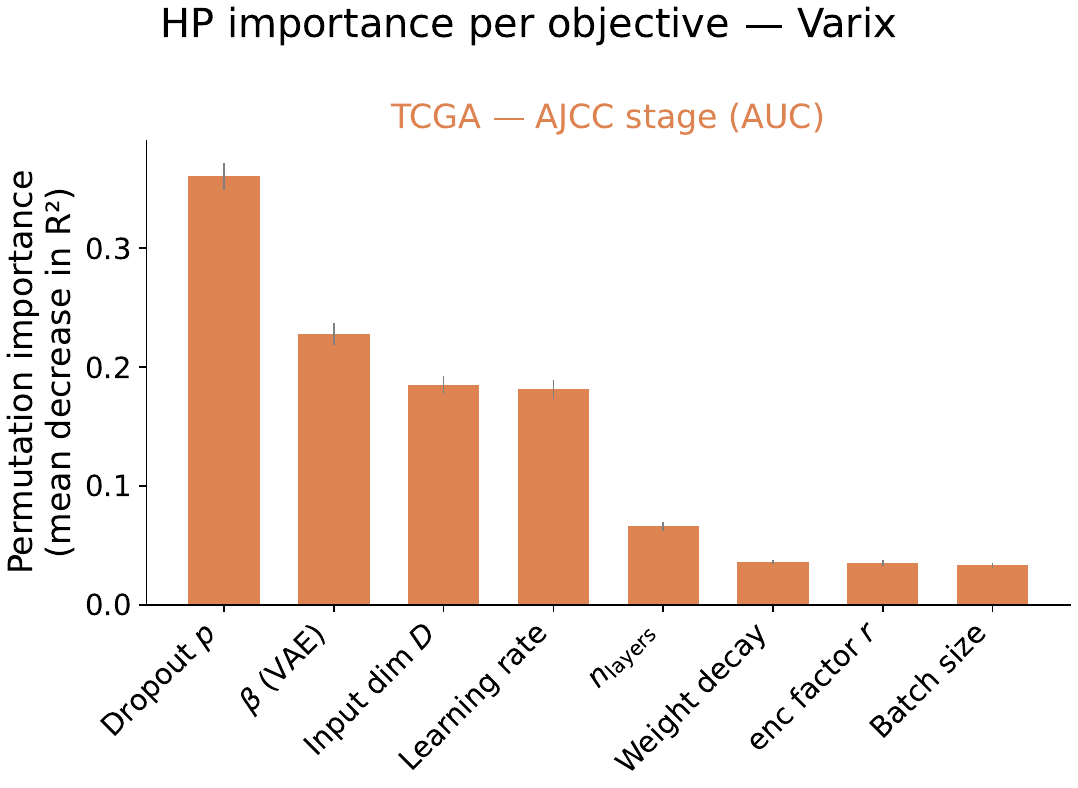}\hfill
    \includegraphics[width=0.24\linewidth]{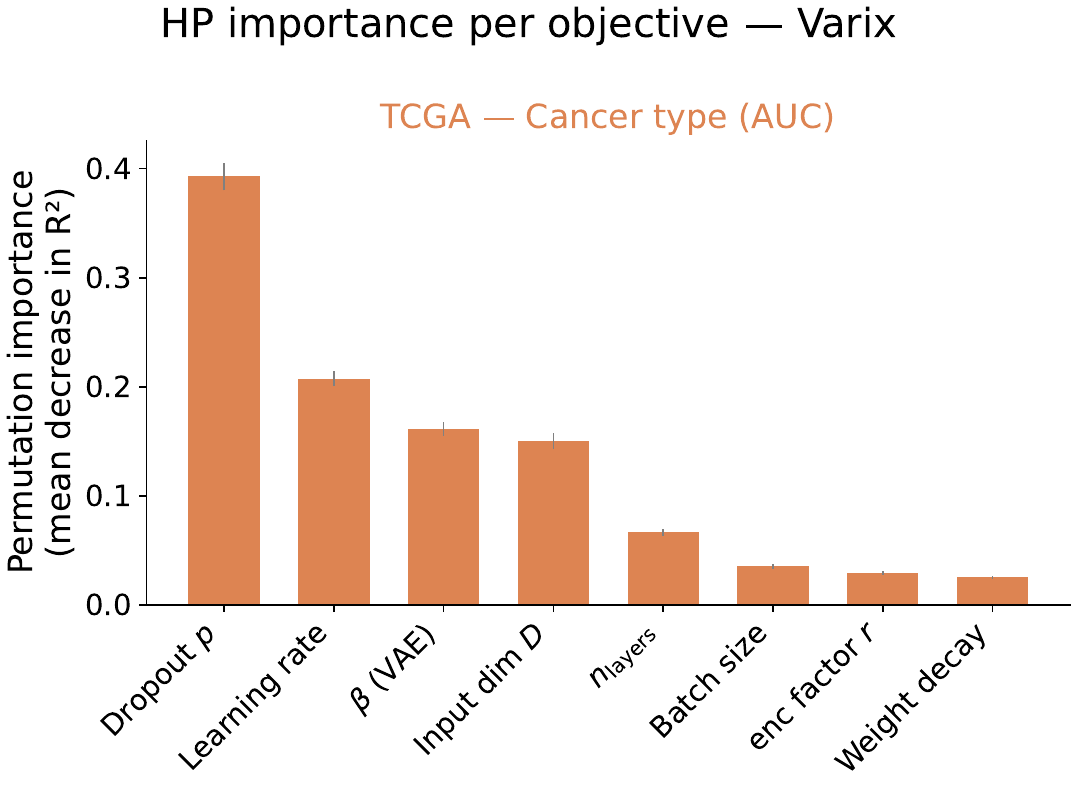}\hfill
    \includegraphics[width=0.24\linewidth]{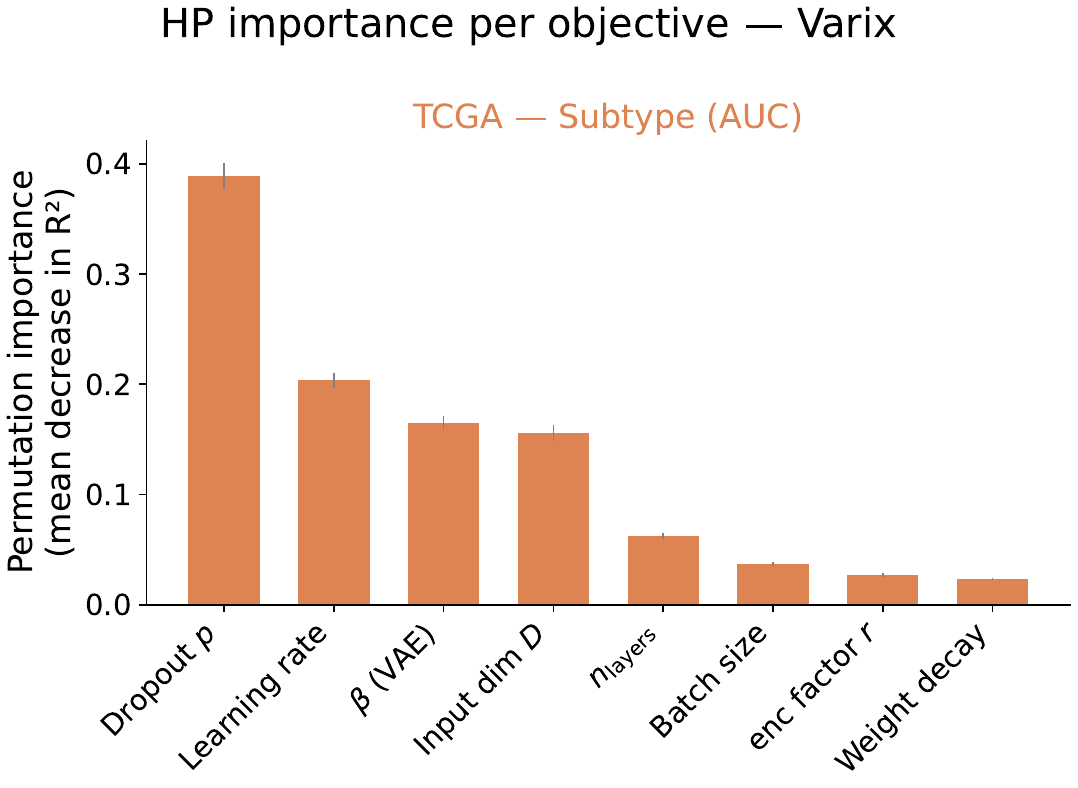}\hfill
    \includegraphics[width=0.24\linewidth]{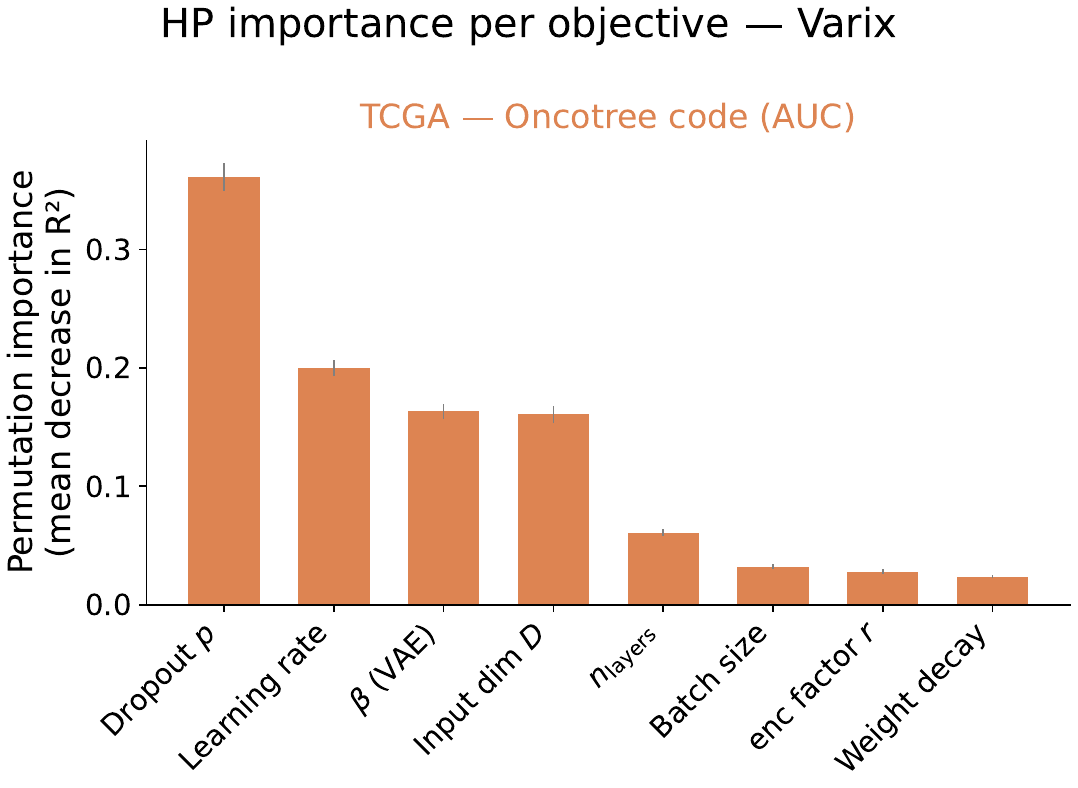}
    \caption{Per-task HP importance Varix. Dropout rate dominates consistently
    regardless of task, with $\beta$ (VAE) as a stable second contributor.}
    \label{fig:app-hp-pertask-varix}
\end{figure*}

\begin{figure*}[t]
    \centering
    \includegraphics[width=0.24\linewidth]{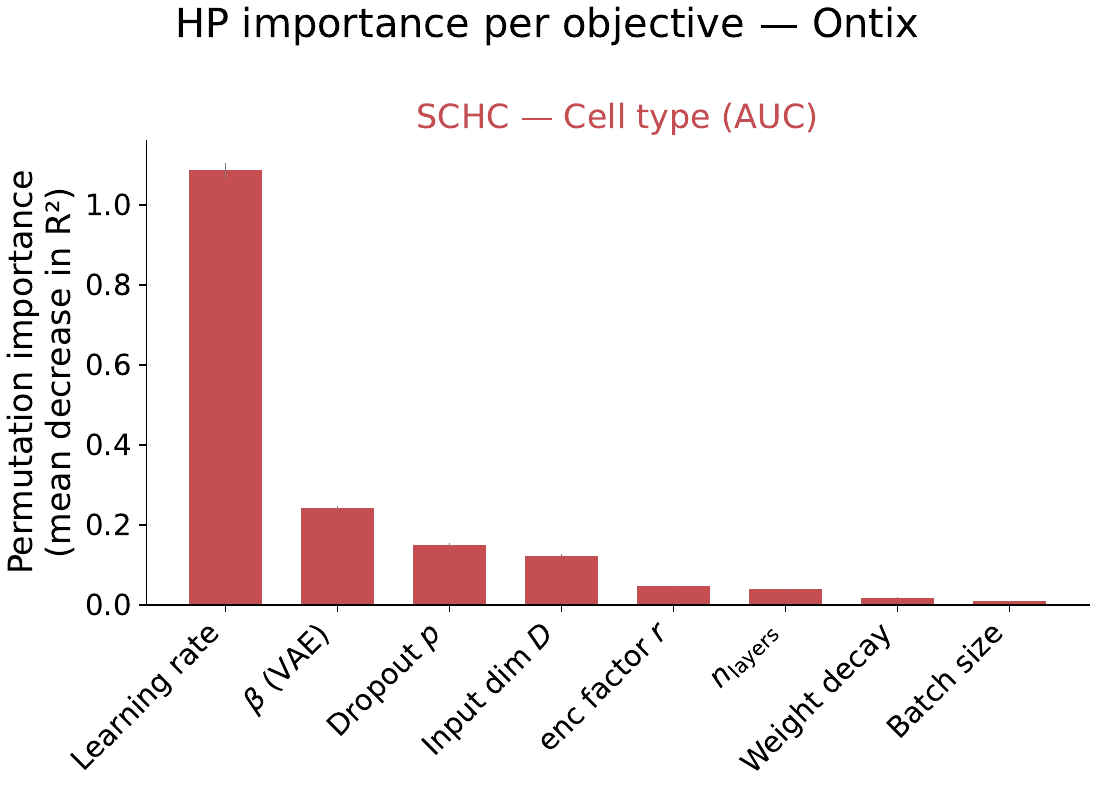}\hfill
    \includegraphics[width=0.24\linewidth]{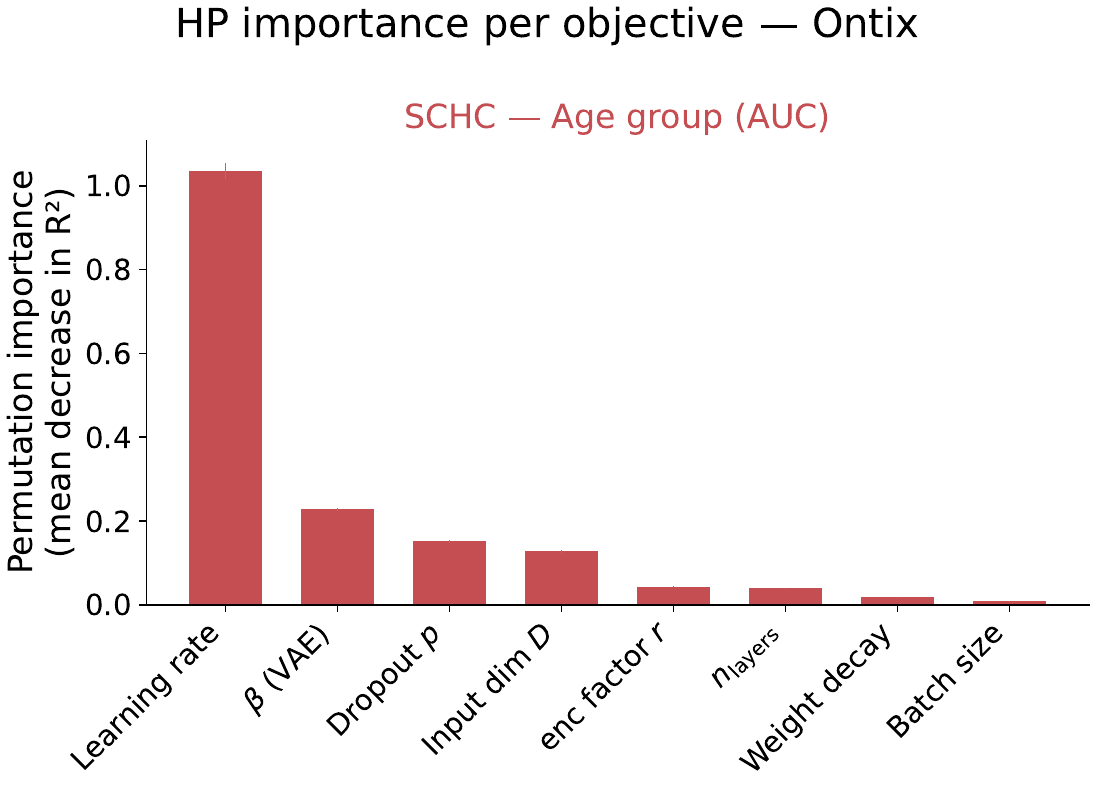}\hfill
    \includegraphics[width=0.24\linewidth]{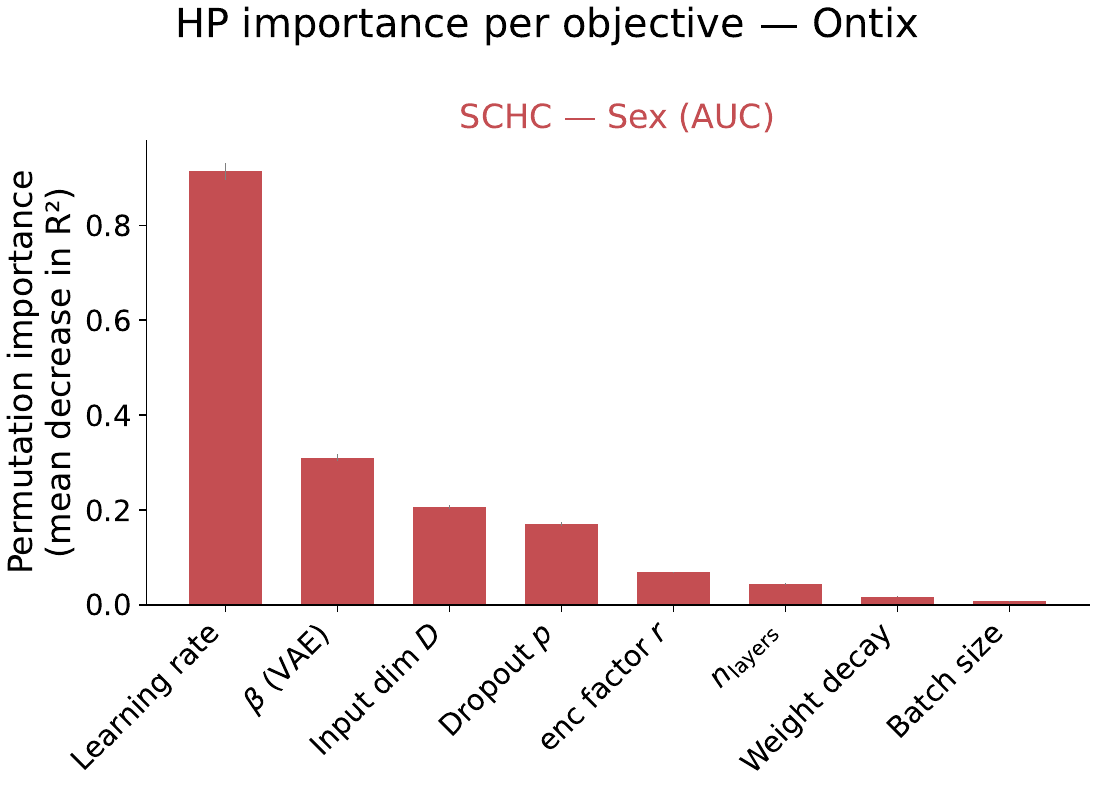}\hfill
    \includegraphics[width=0.24\linewidth]{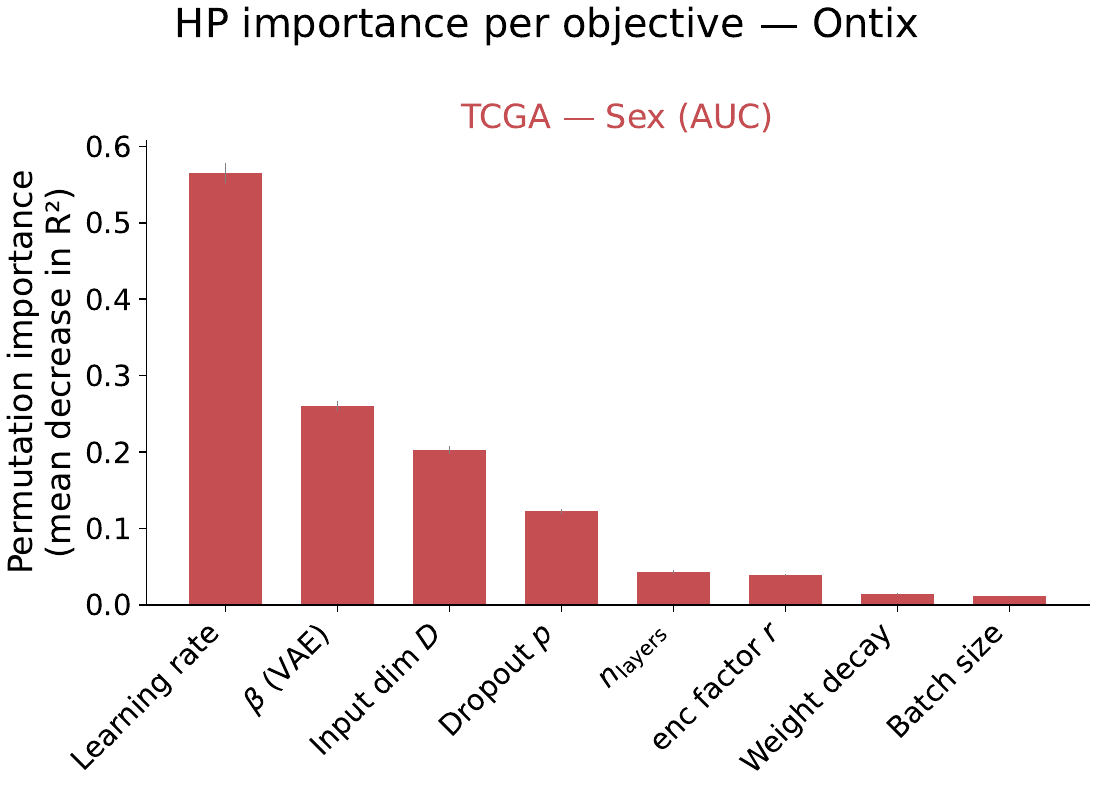}
    
    \vspace{0.2cm}
    \includegraphics[width=0.24\linewidth]{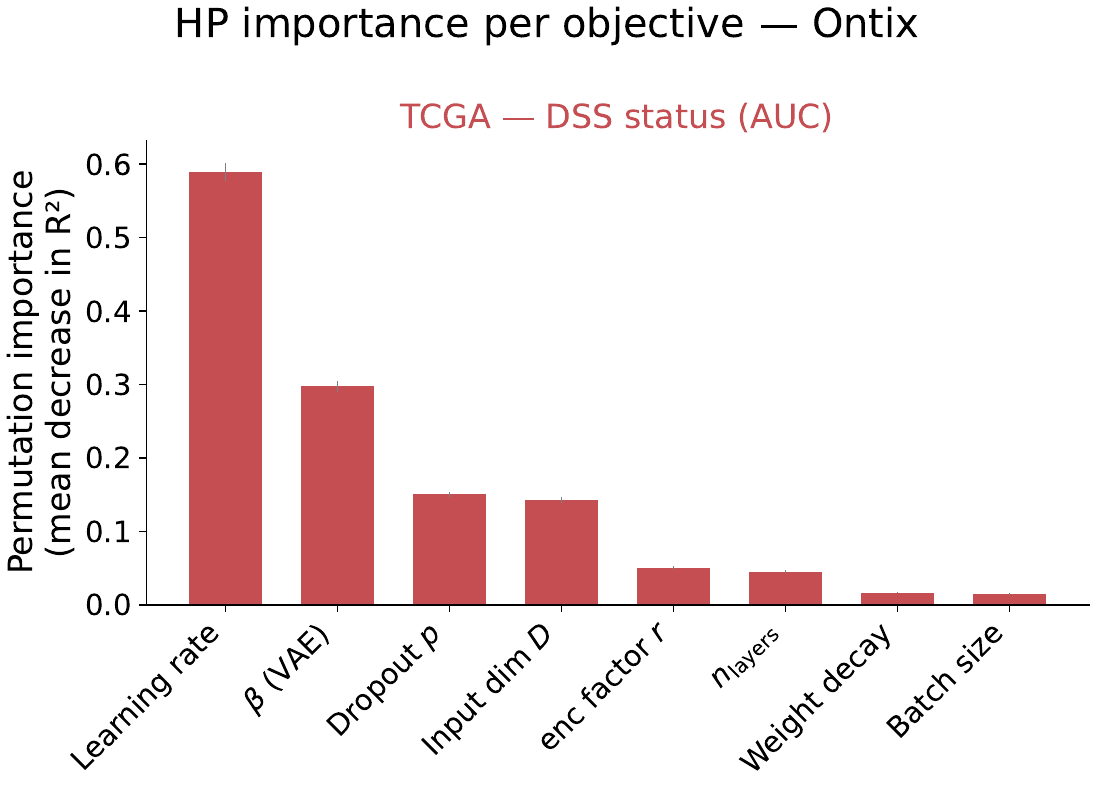}\hfill
    \includegraphics[width=0.24\linewidth]{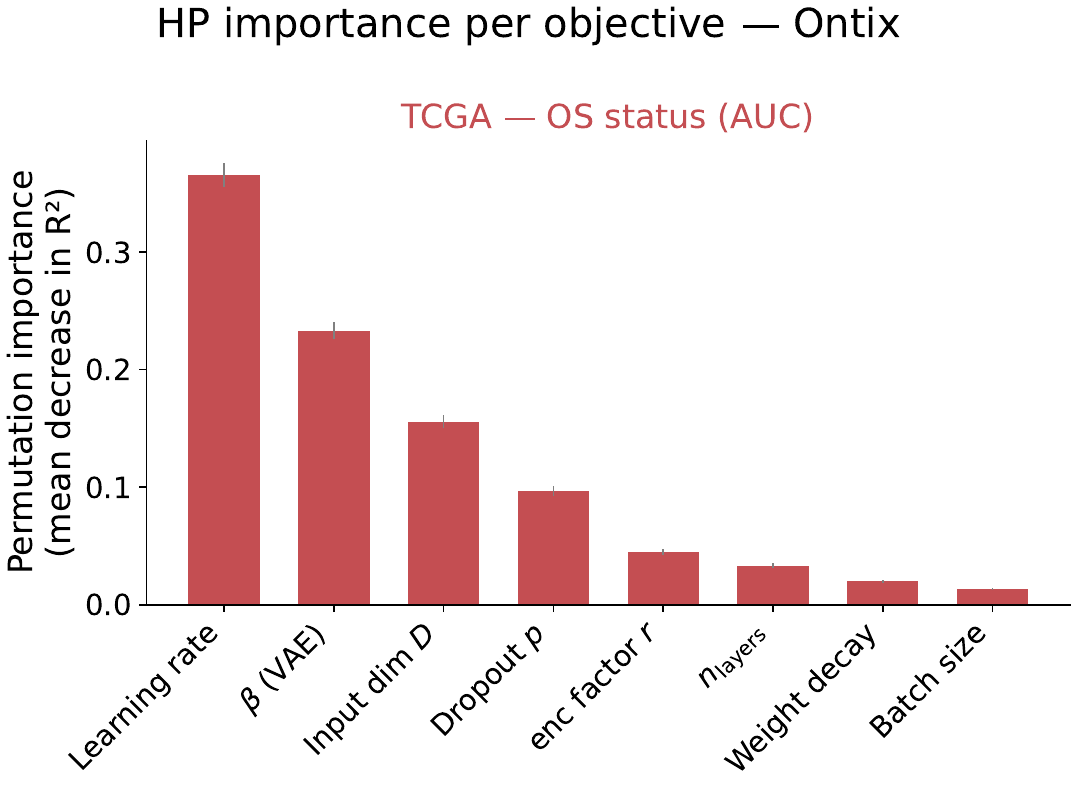}\hfill
    \includegraphics[width=0.24\linewidth]{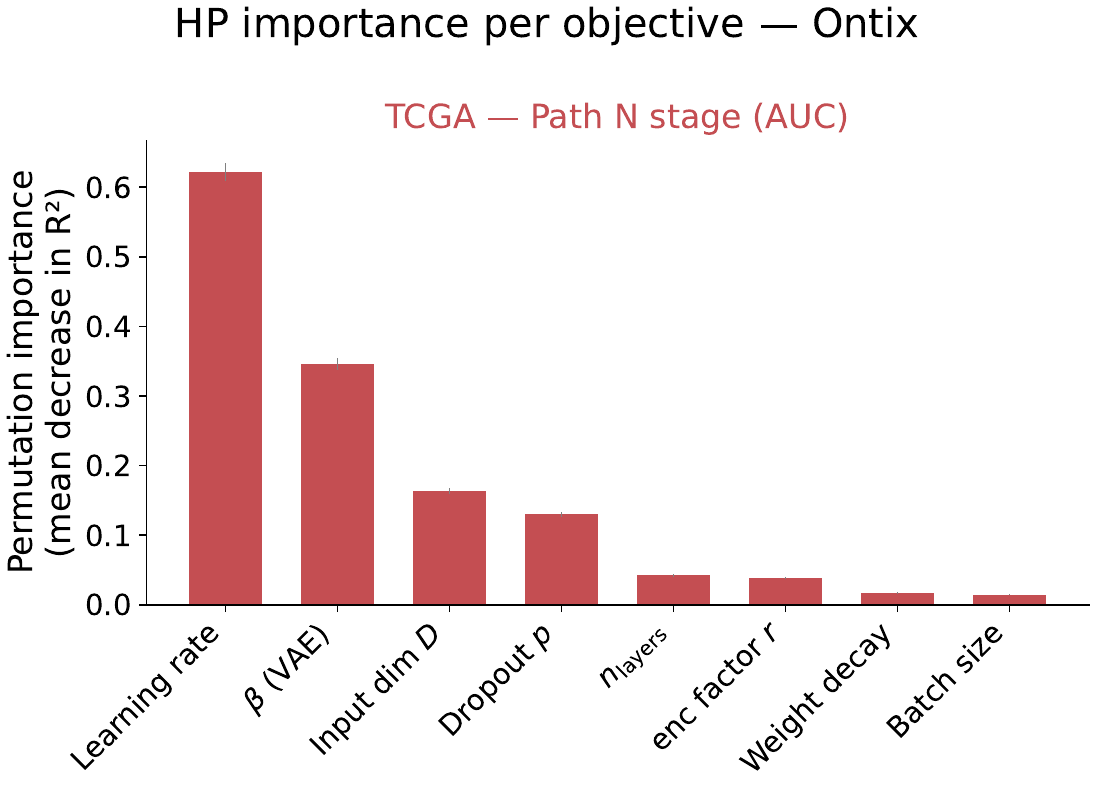}\hfill
    \includegraphics[width=0.24\linewidth]{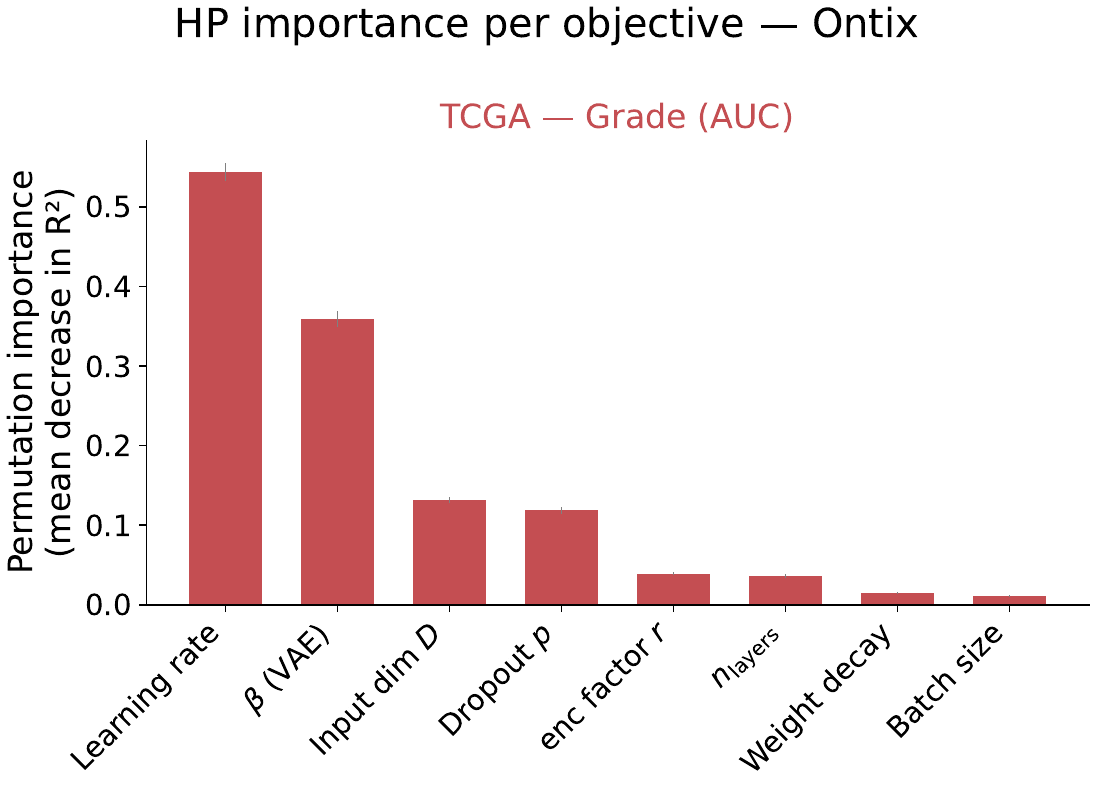}
    
    \vspace{0.2cm}
    \includegraphics[width=0.24\linewidth]{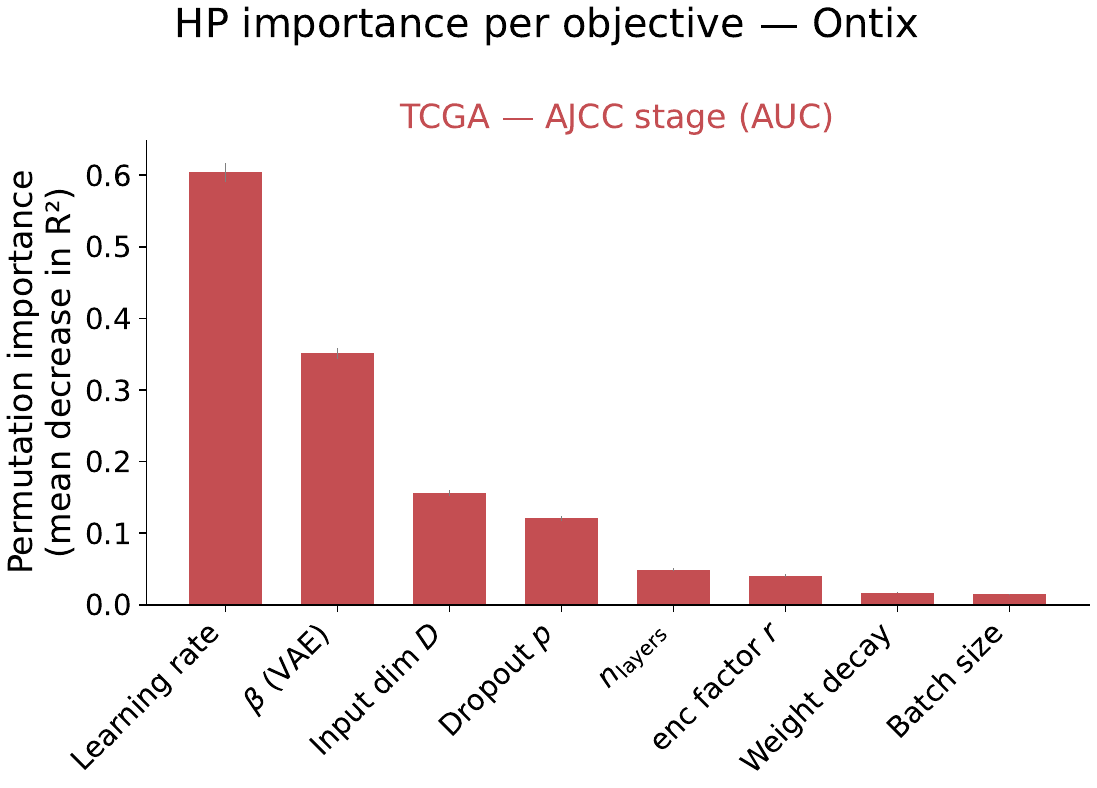}\hfill
    \includegraphics[width=0.24\linewidth]{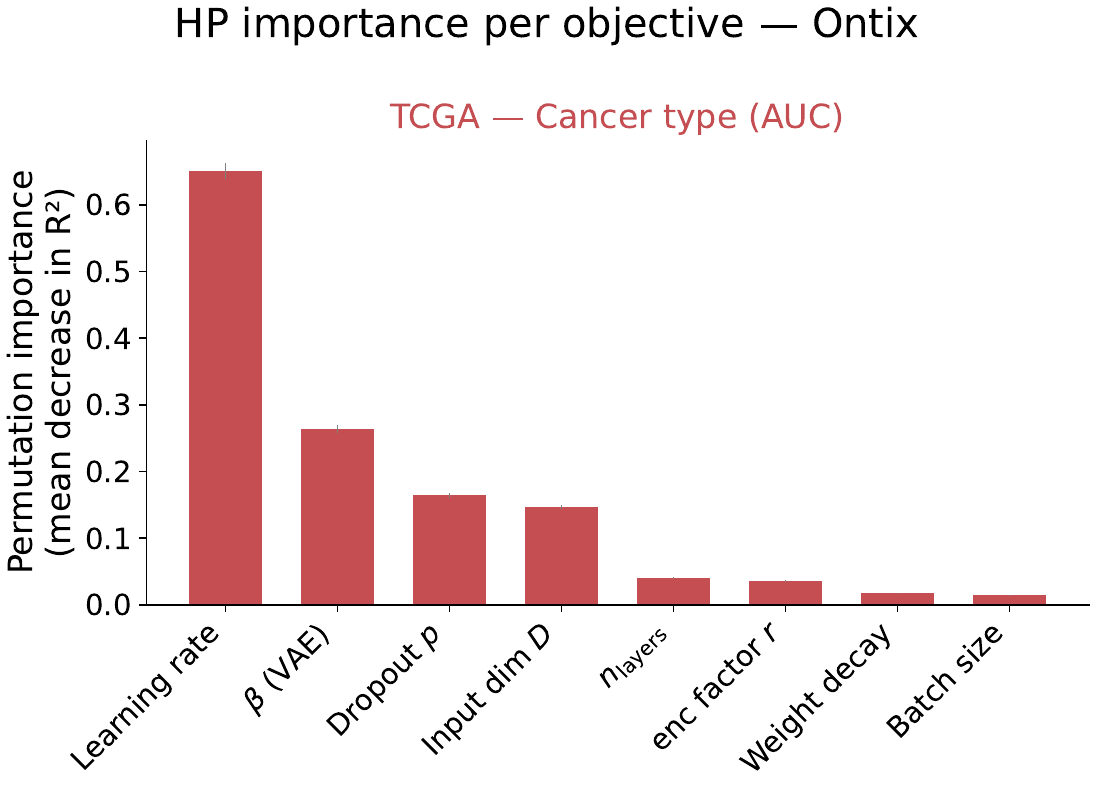}\hfill
    \includegraphics[width=0.24\linewidth]{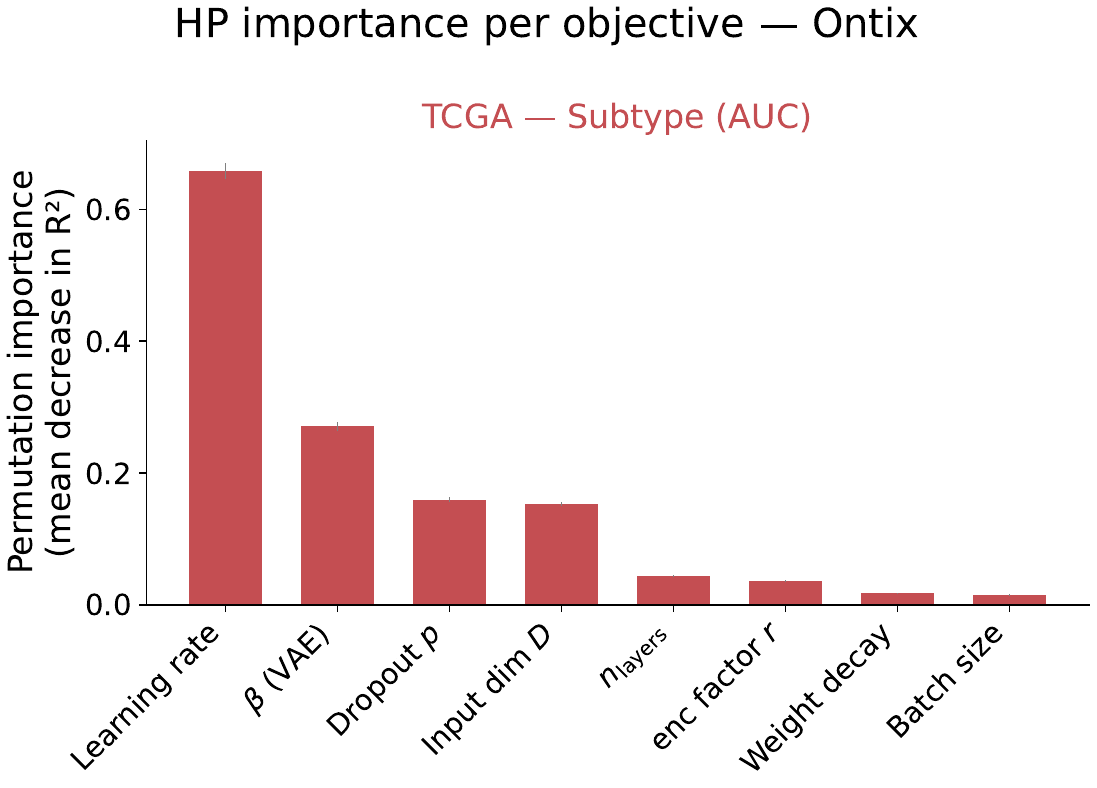}\hfill
    \includegraphics[width=0.24\linewidth]{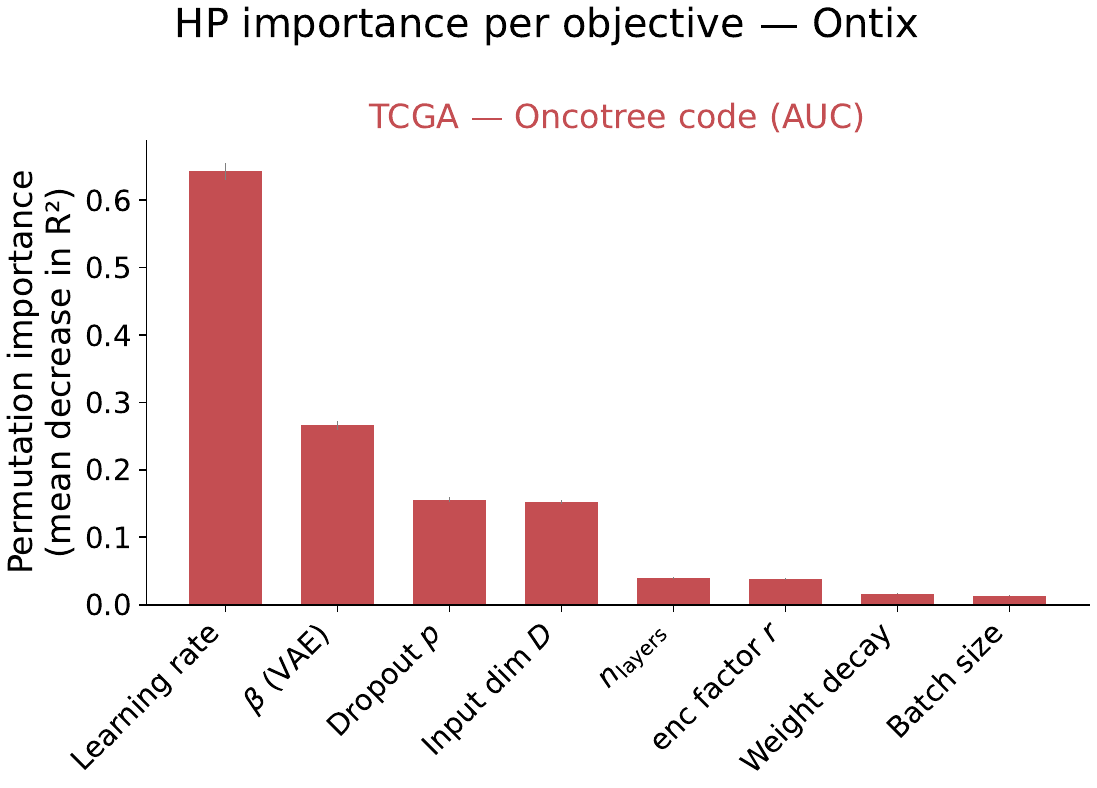}
    \caption{Per-task HP importance Ontix. Learning rate dominates consistently
    regardless of task, with $\beta$ (VAE) as a stable second contributor.}
    \label{fig:app-hp-pertask-ontix}
\end{figure*}

\begin{figure*}[t]
    \centering
    \includegraphics[width=0.24\linewidth]{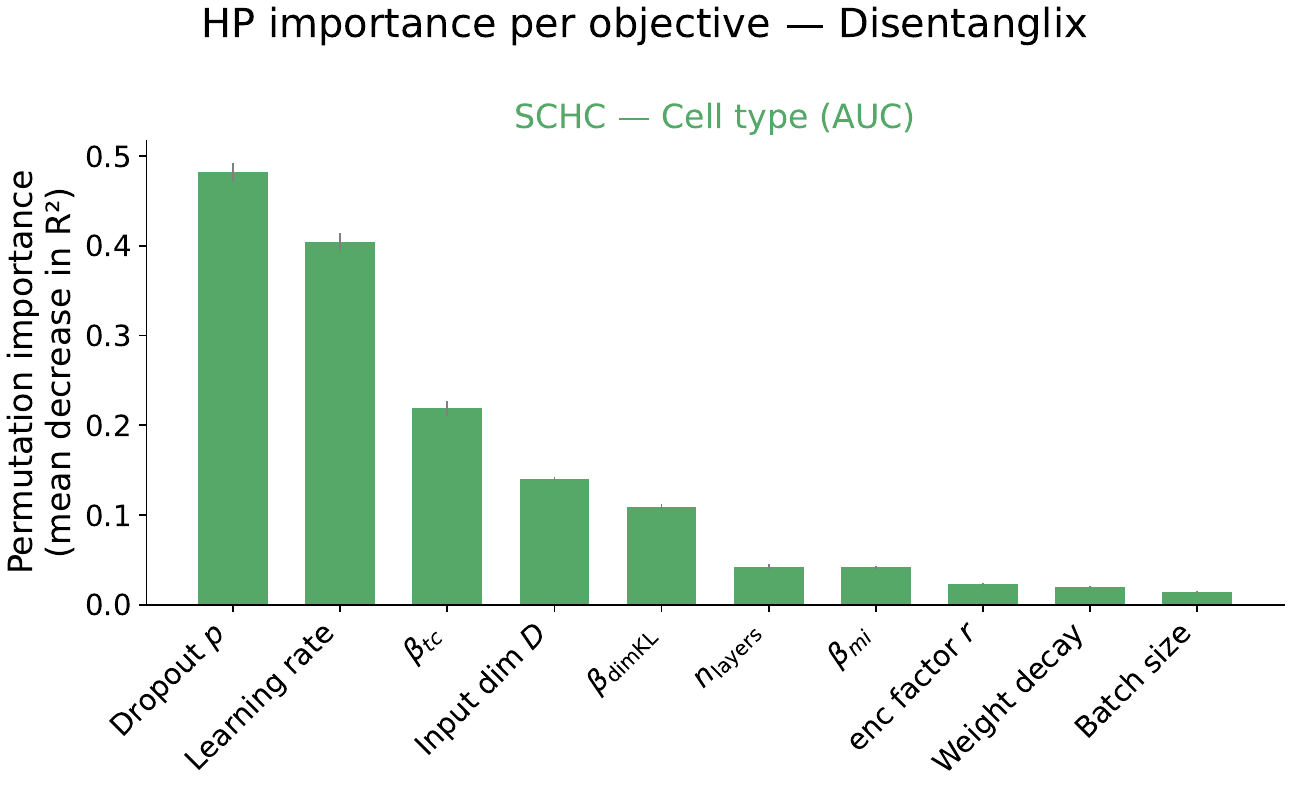}\hfill
    \includegraphics[width=0.24\linewidth]{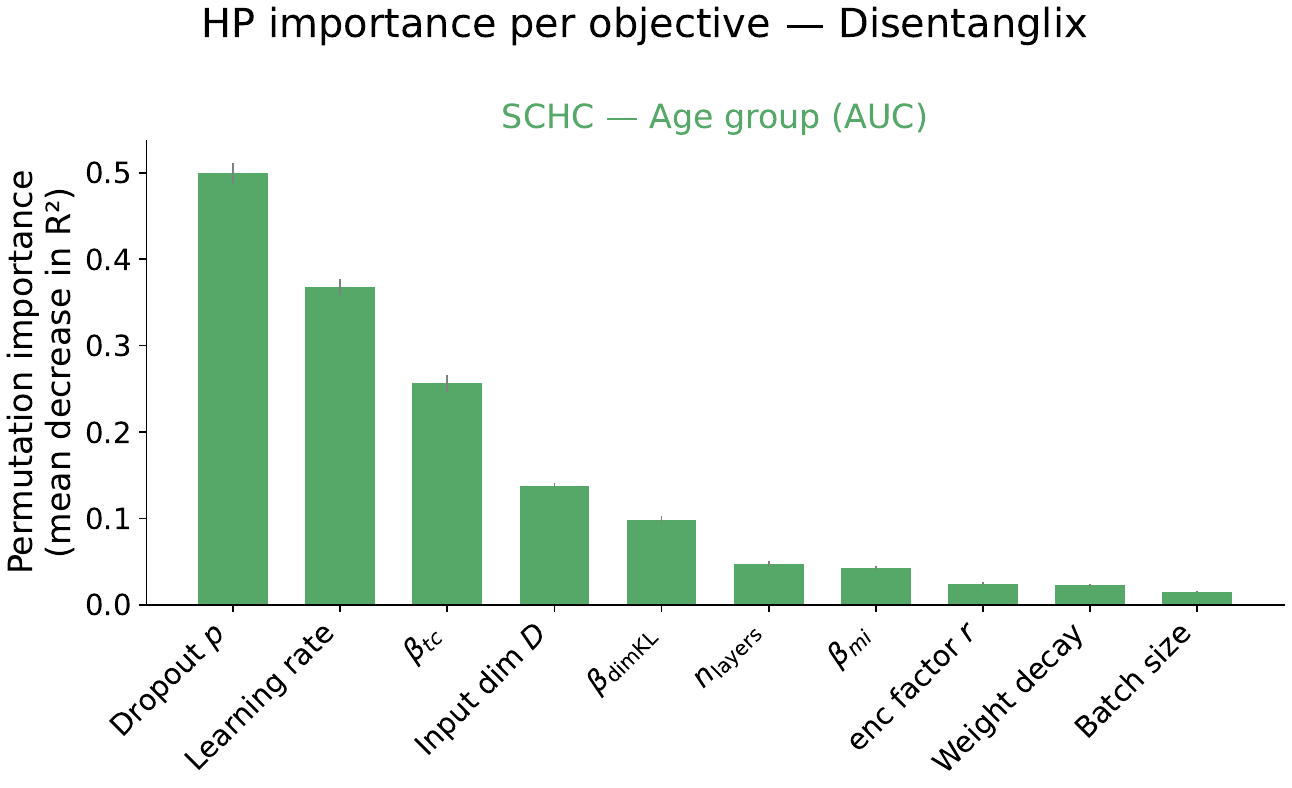}\hfill
    \includegraphics[width=0.24\linewidth]{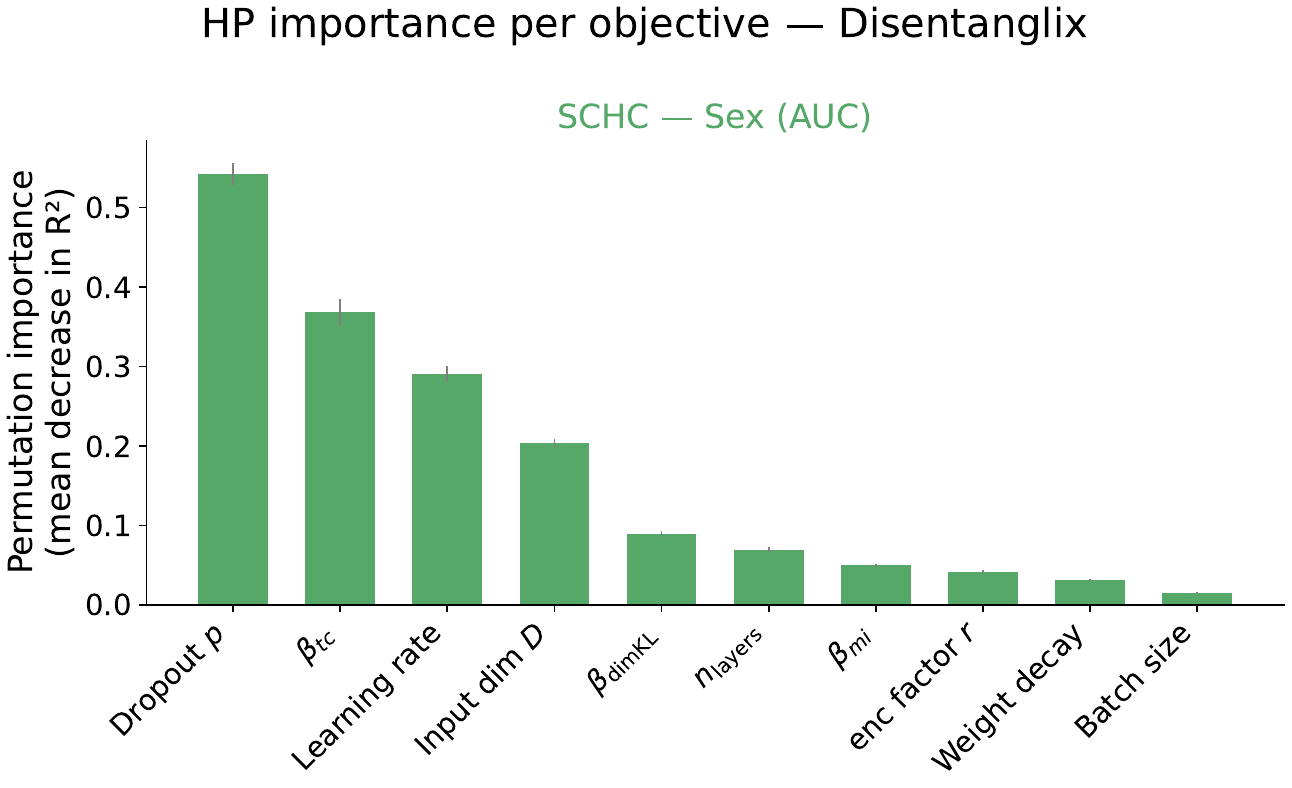}\hfill
    \includegraphics[width=0.24\linewidth]{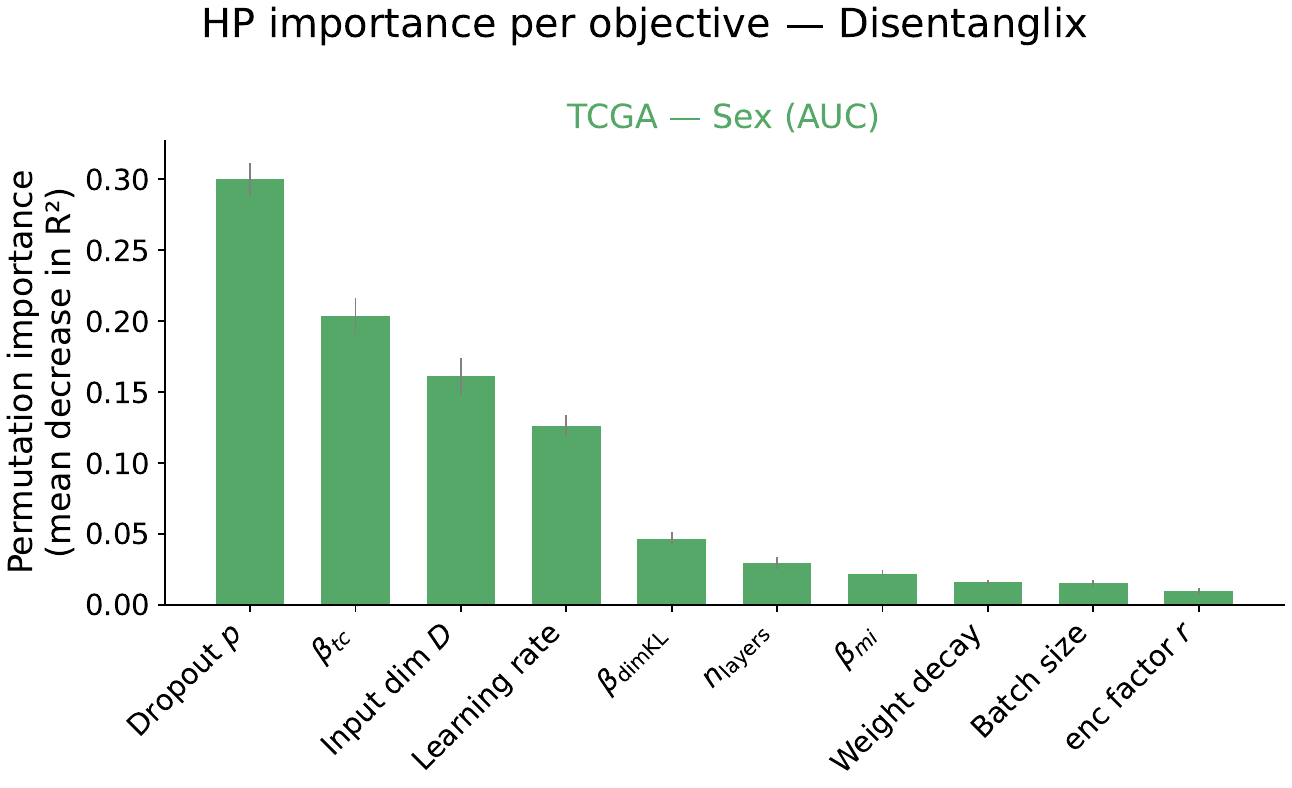}
    
    \vspace{0.2cm}
    \includegraphics[width=0.24\linewidth]{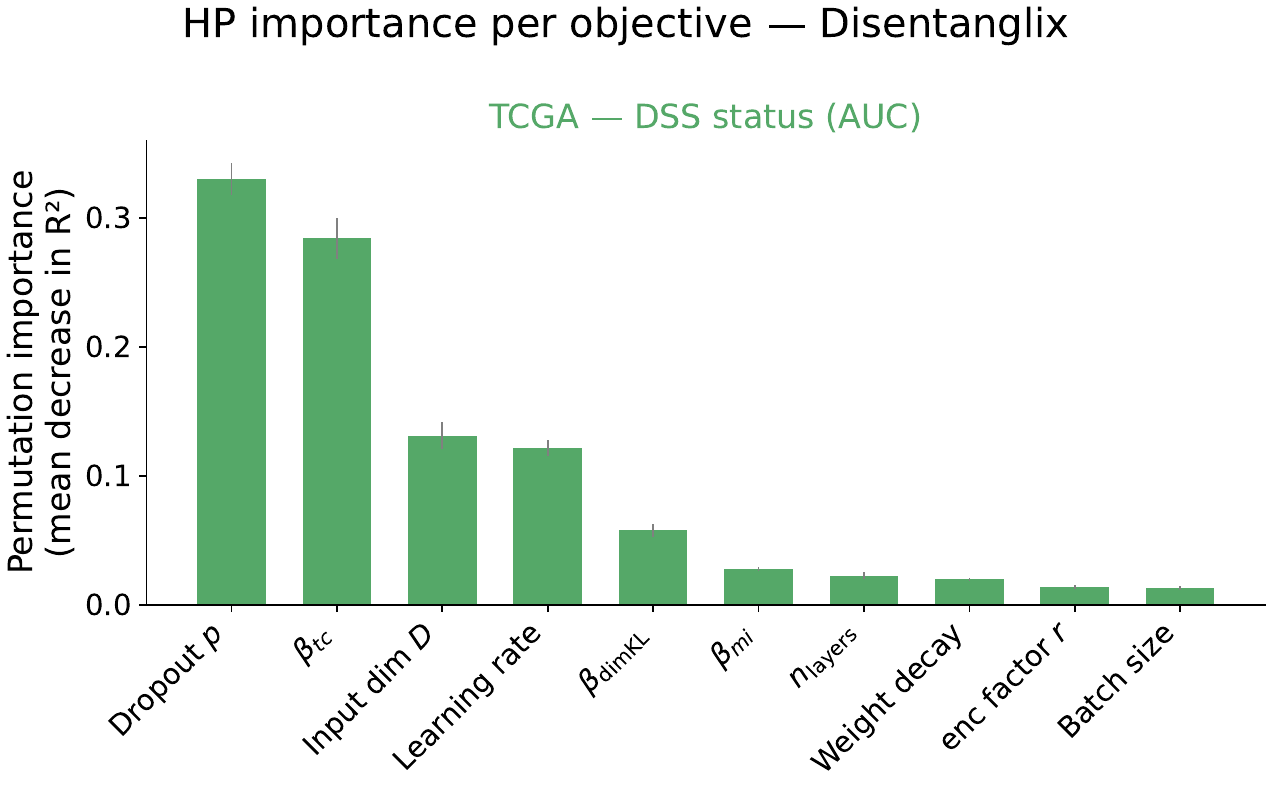}\hfill
    \includegraphics[width=0.24\linewidth]{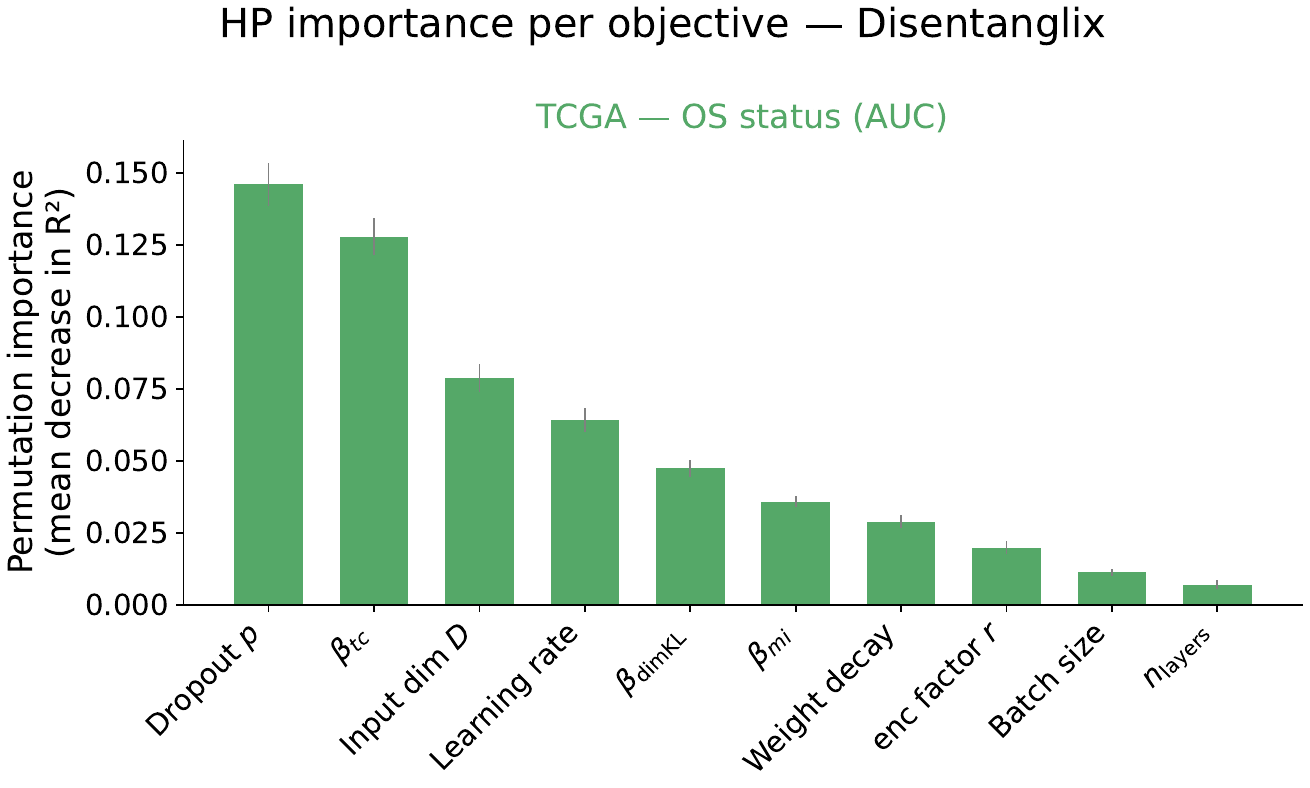}\hfill
    \includegraphics[width=0.24\linewidth]{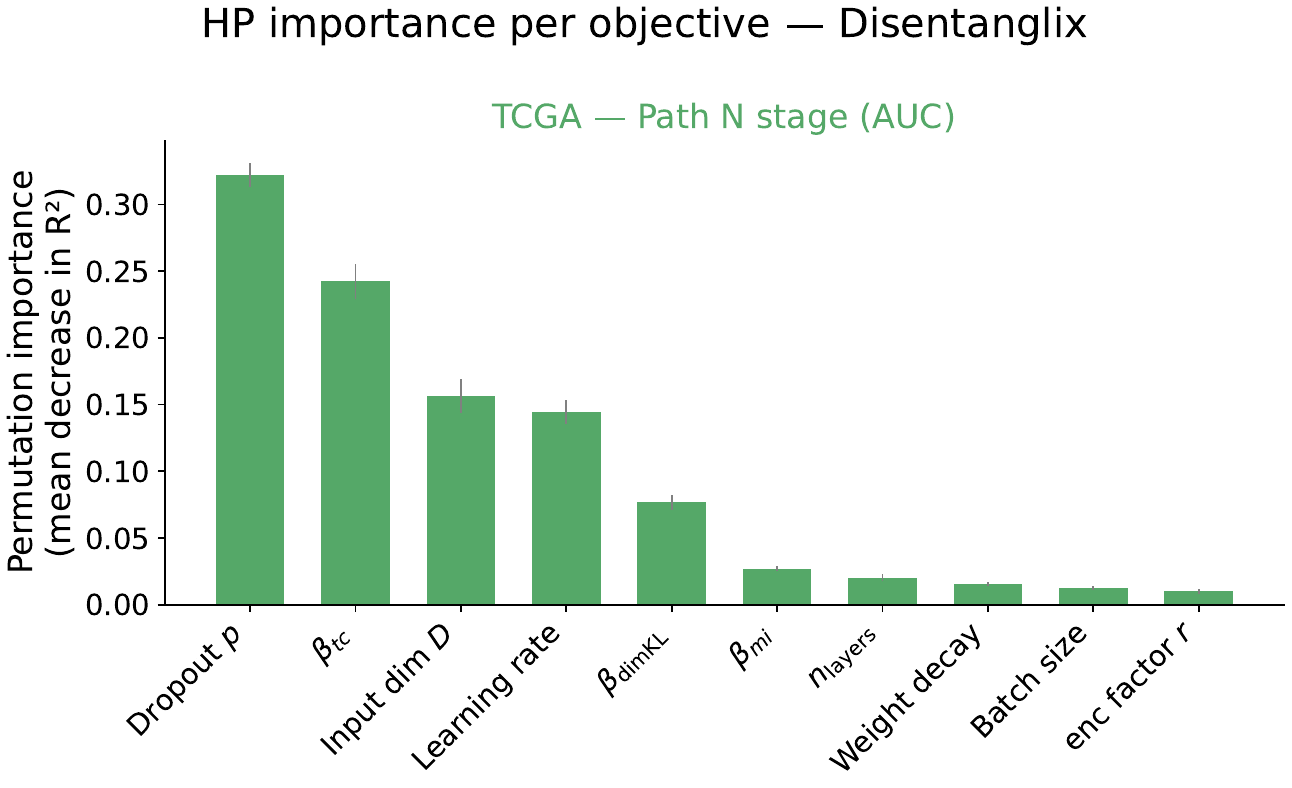}\hfill
    \includegraphics[width=0.24\linewidth]{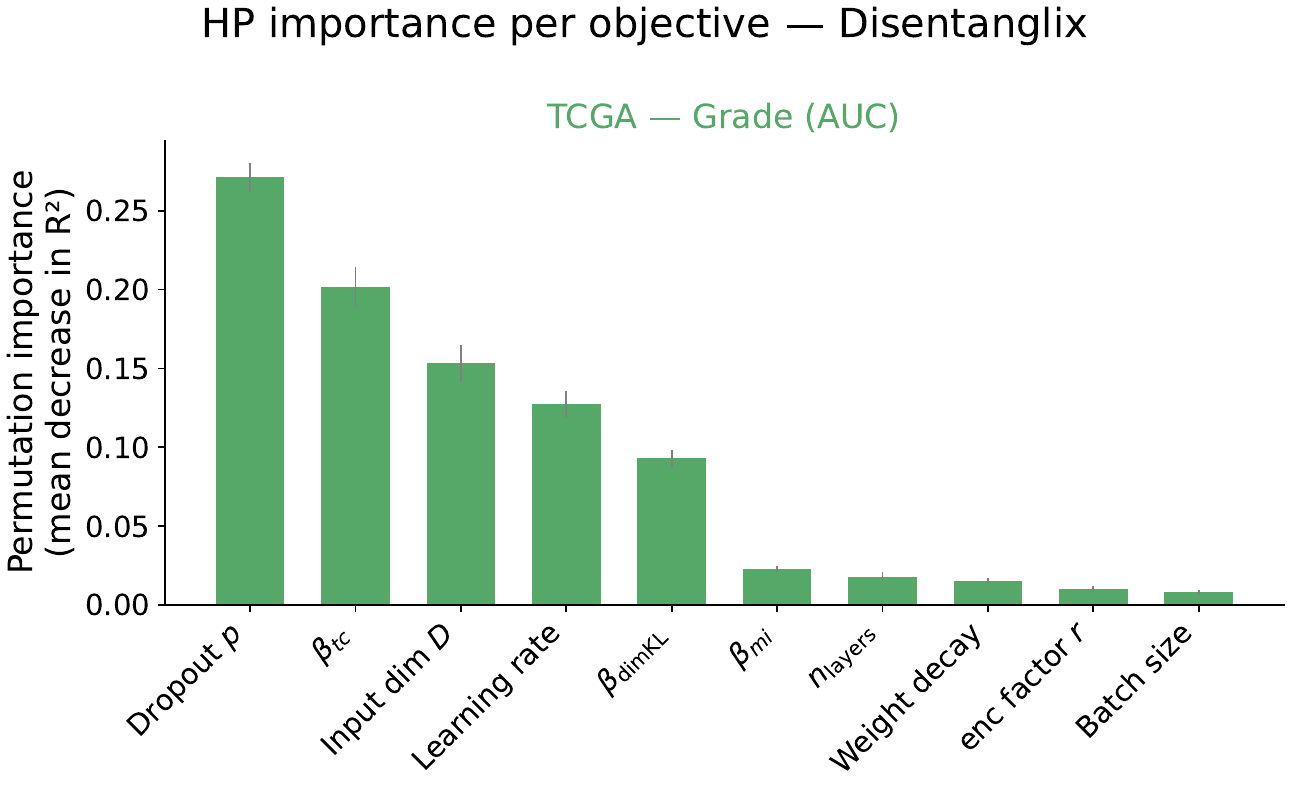}
    
    \vspace{0.2cm}
    \includegraphics[width=0.24\linewidth]{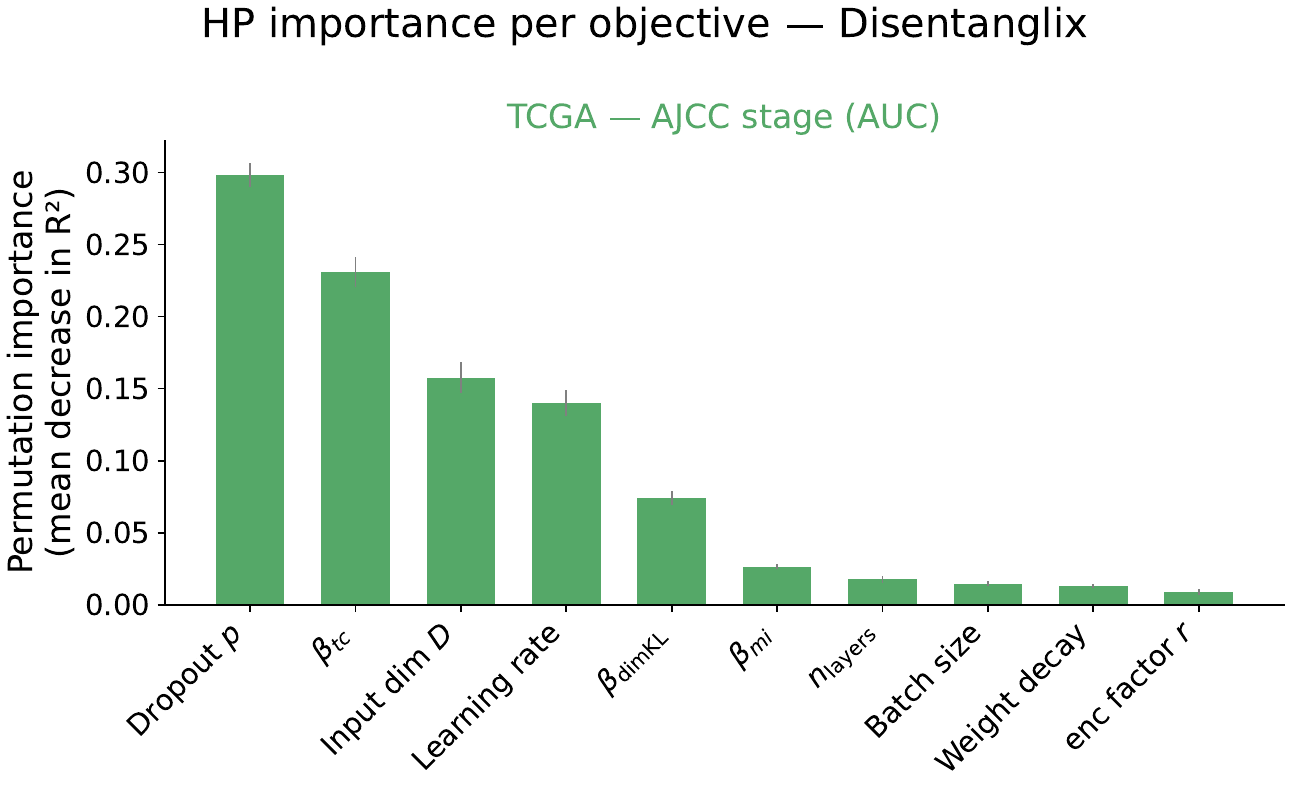}\hfill
    \includegraphics[width=0.24\linewidth]{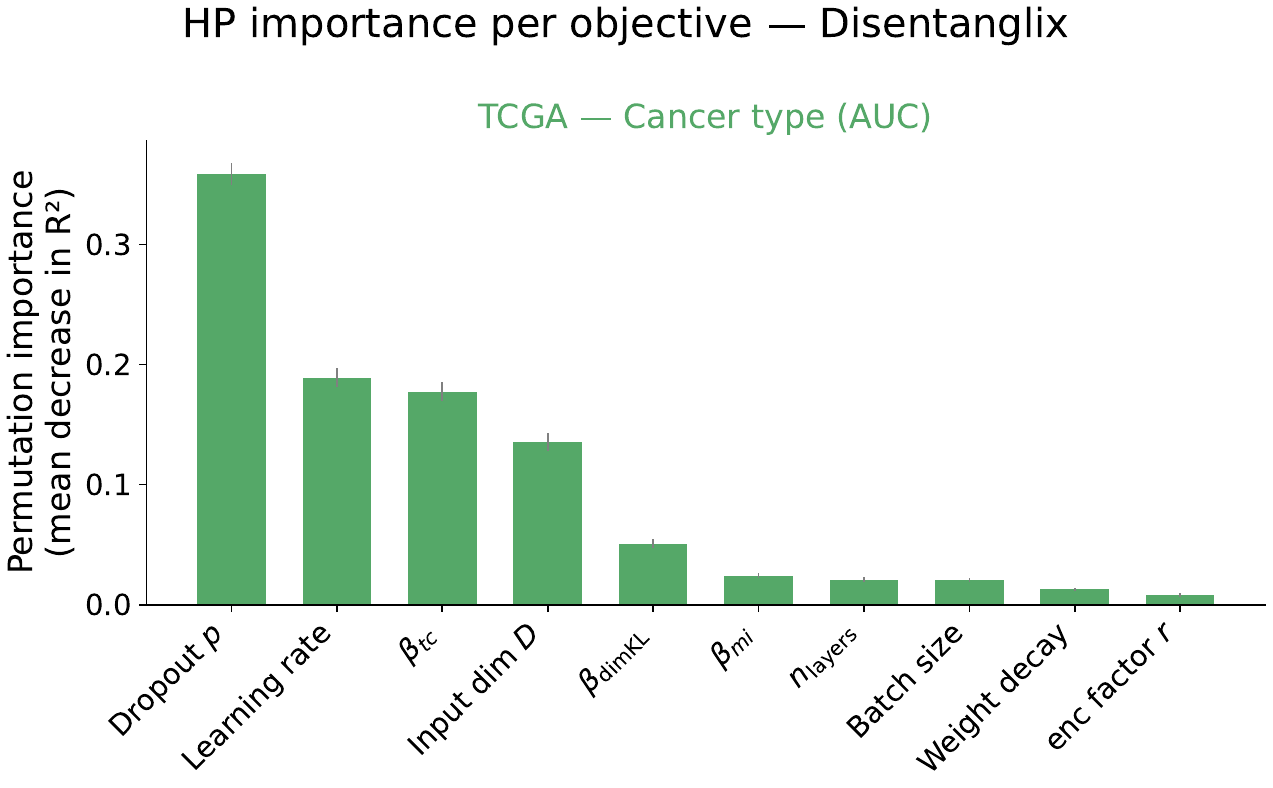}\hfill
    \includegraphics[width=0.24\linewidth]{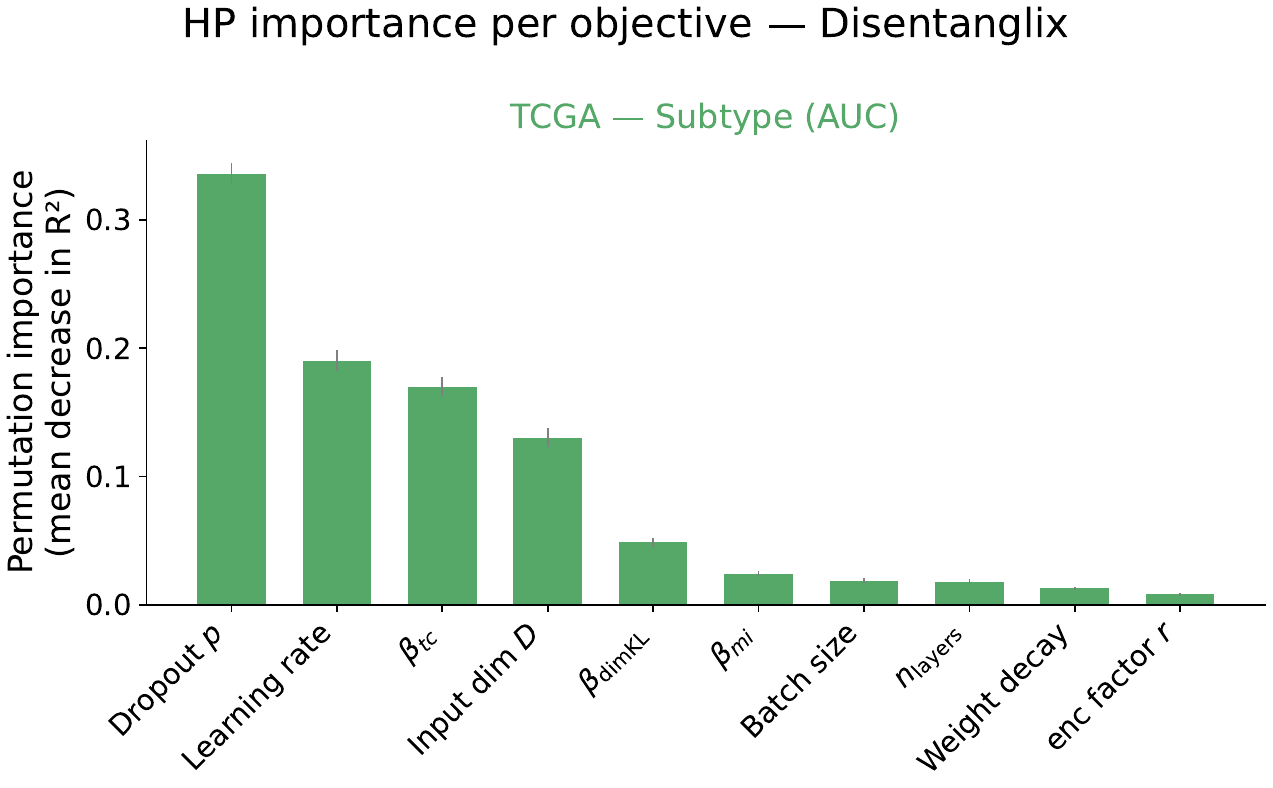}\hfill
    \includegraphics[width=0.24\linewidth]{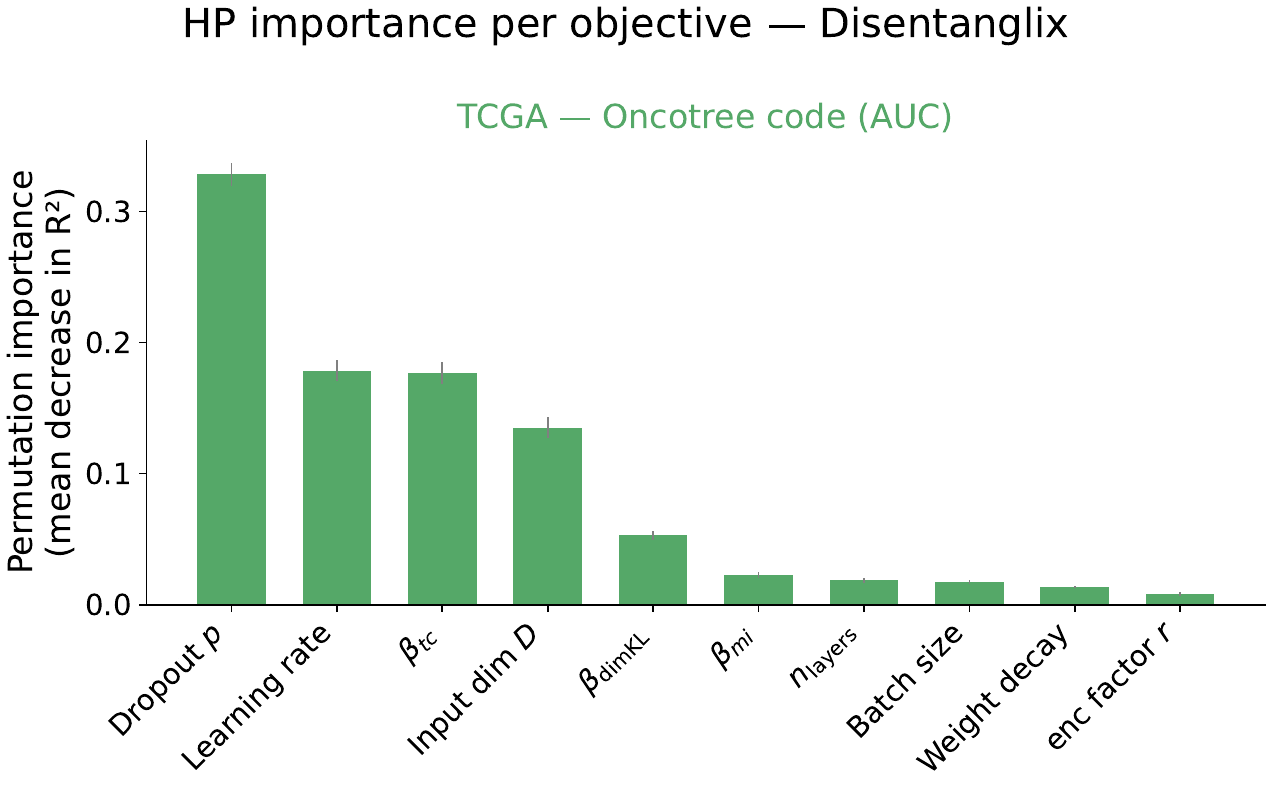}
    \caption{Per-task HP importance Disentanglix. Dropout $p$ and $\beta_{tc}$
    both contribute across tasks, with a more distributed importance profile than
    Ontix.}
    \label{fig:app-hp-pertask-disentanglix}
\end{figure*}

We report permutation importance scores for each architecture across three objectives:
aggregate downstream performance (averaged over all datasets), TCGA per-task average,
SCHC per-task average, and reconstruction loss.
Figures~\ref{fig:app-hp-vanillix}--\ref{fig:app-hp-pertask-disentanglix} show the results.

We also show hyperparameter importance for each downstream metric, task and architecture in figures \ref{fig:app-hp-pertask-vanillix} - \ref{fig:app-hp-pertask-disentanglix}.

\clearpage

\section{Description of Black-box Optimizers}
\label{app:optimizers}

We consider the following black-box optimization algorithms to construct our dataset:
\begin{itemize}
    \item \textbf{BORE}~\citep{tiao2020bayesian}: Bayesian Optimization by Density-Ratio
    Estimation recasts hyperparameter optimization as a binary classification task. A
    probabilistic classifier is trained to assign high scores to configurations likely
    to exceed a performance threshold. We used the Syne Tune default configuration.

    \item \textbf{CQR}~\citep{salinas2023optimizing}: Conformalized Quantile Regression
    is a surrogate model for Bayesian optimization that handles heteroskedastic
    objective noise by estimating conditional quantiles rather than assuming fixed
    Gaussian observation noise. Split conformal prediction is applied to calibrate
    these quantile intervals with an empirical offset $\gamma_j$, providing finite-sample
    coverage guarantees. The calibrated quantiles enable efficient Thompson sampling,
    selecting the next configuration as the candidate with the lowest randomly drawn
    quantile value. We used the Syne Tune default configuration.

    \item \textbf{REA}~\citep{real2019regularized}: Regularized Evolution is a
    population-based search method that modifies tournament selection via
    \textit{aging}: rather than eliminating the worst-performing individuals,
    the oldest configurations are removed from the population, encouraging
    broader exploration and reducing premature convergence. We used a \textit{population size} of 25 and \textit{sample size} of 5 instead of the Syne Tune defaults.

    \item \textbf{TPE}~\citep{bergstra2011algorithms}: Tree-structured Parzen Estimator
    is a Bayesian optimization method that models the hyperparameter--performance
    relationship through density estimation rather than direct regression. Two kernel
    density estimators are fit: $l(x) = p(x \mid y < y^*)$ over high-performing
    configurations and $g(x) = p(x \mid y \ge y^*)$ over the remainder, separated
    by a quantile threshold $y^*$. Acquisition proceeds by maximizing $l(x)/g(x)$,
    which is proportional to the Expected Improvement. We used the Syne Tune default configuration.

    \item \textbf{ASHA}~\citep{li2020system}: Asynchronous Successive Halving is a
    multi-fidelity hyperparameter optimization algorithm that allocates resources
    adaptively. Configurations are initially evaluated on small budgets, and only
    the most promising ones are promoted to higher fidelity levels, enabling
    efficient parallel search without requiring synchronization across workers. We used the Syne Tune default configuration.

    \item \textbf{ASHABORE}: A hybrid method that uses ASHA for multi-fidelity
    scheduling while replacing the default random promotion policy with BORE's
    density-ratio classifier to guide the selection of configurations at each
    rung level. We used the Syne Tune default configuration.

    \item \textbf{ASHACQR}: A hybrid method combining ASHA's successive halving
    schedule with CQR's conformalized quantile surrogate to select and promote
    configurations, inheriting both the budget efficiency of ASHA and the
    uncertainty calibration of CQR. We used the Syne Tune default configuration.
    \item \textbf{BOHB}~\citep{falkner2018bohb}: Bayesian Optimization with
    Hyperband combines the strong anytime performance of Hyperband with the
    sample efficiency of Bayesian optimization. It uses kernel density estimators
    (following TPE) to model the distribution of good and bad configurations, and
    couples this with successive halving to allocate compute budgets adaptively. We used a \textit{min\_bandwidth} of $1e^{-1}$.
    
    \item \textbf{Quantile Transfer}~\citep{salinas2023optimizing}: Extends CQR
    to the transfer learning setting by leveraging performance data from related
    tasks. Quantile regressors trained on source tasks are used to warm-start
    the surrogate on a new target task, reducing the number of evaluations
    required to identify well-performing configurations. We used the Syne Tune default configuration.

    \item \textbf{Zero Shot}~\citep{wistuba2015sequential}: Constructs a portfolio of
    configurations offline from historical data across tasks, without performing
    any online evaluations on the target task. At deployment time, the single
    best-predicted configuration from this portfolio is returned, making it
    entirely free of target-task function evaluations. We used the Syne Tune default configuration.

    \item \textbf{Bounding Box}~\citep{perrone2019learning}: Restricts the
    hyperparameter search space to a tighter \textit{bounding box} inferred
    from well-performing configurations observed on related tasks. Standard
    Bayesian optimization is then run within this reduced space, concentrating
    the search budget in more promising regions. We used the Syne Tune default configuration.

    \item \textbf{RS}~\citep{bergstra2012random}: Random Search draws configurations
    independently and uniformly from the search space, serving as a simple but
    competitive baseline. We used the Syne Tune default configuration.
\end{itemize}

\clearpage

\section{Per-Task Performance}
\label{app:per-task}

We show violin plots of average downstream performance for both, SCHC and TCGA and their corresponding downstream tasks. 

\label{app:per-task-performance}

\begin{figure}[h]
    \centering
    \includegraphics[width=0.75\linewidth]{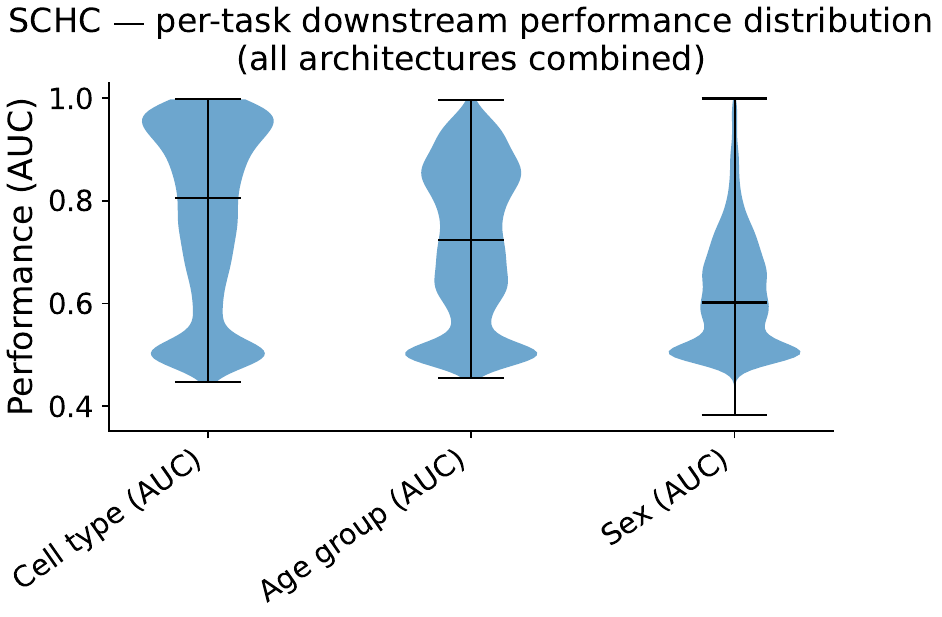}
    \caption{Per-task performance averaged over all architectures on the SCHC dataset.}
    \label{fig:task_performance_schc}
\end{figure}

\begin{figure}[h]
    \centering
    \includegraphics[width=0.75\linewidth]{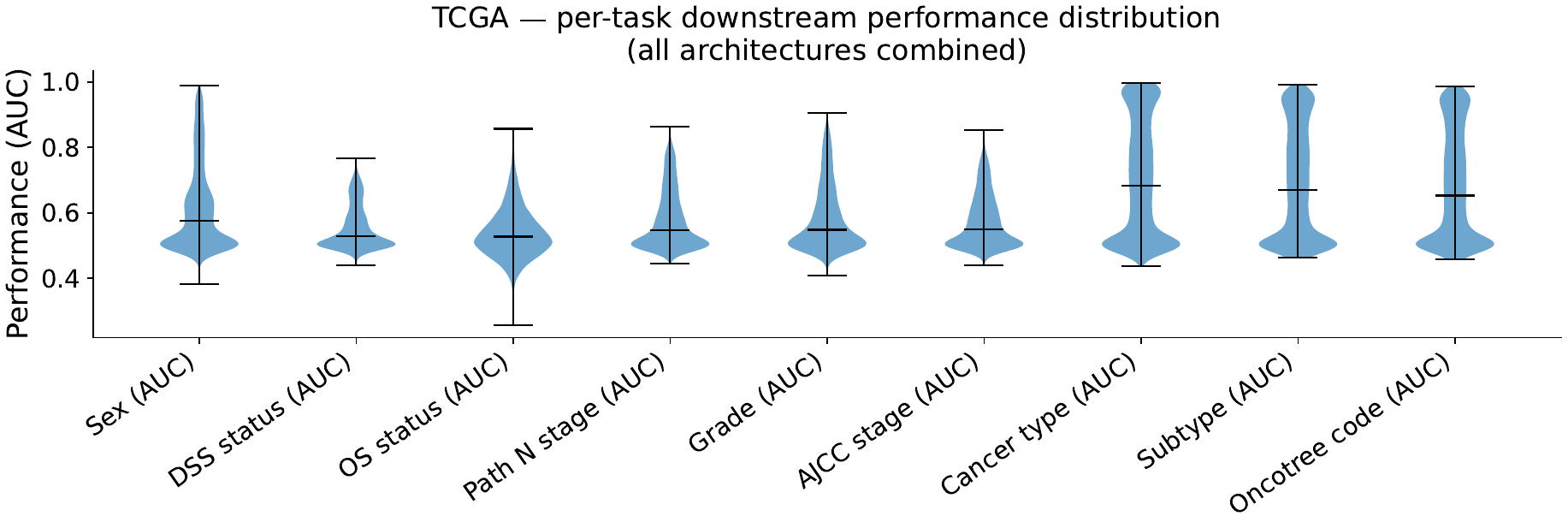}
    \caption{Per-task performance averaged over all architectures on the TCGA dataset.}
    \label{fig:task_performance_tcga}
\end{figure}

\clearpage

\section{Optimization Trajectories}
\label{sec:traj}

We show optimization trajectories for all methods used and all 35 blackbox tasks. Each optimizer was run with 30 different seeds and for 72000 seconds simulated wallclock time.

% Model: Disentanglix
\begin{figure}[htbp]
    \centering
    \setlength{\tabcolsep}{1pt}
    \begin{tabular}{ccc}
    \multicolumn{3}{c}{\textbf{autoencodix-disentanglix\_schc-schc-METH-CLIN}} \\
    \textbf{Single-Fidelity} & \textbf{Transfer Learning} & \textbf{Multi-Fidelity} \\
    \includegraphics[width=0.32\textwidth]{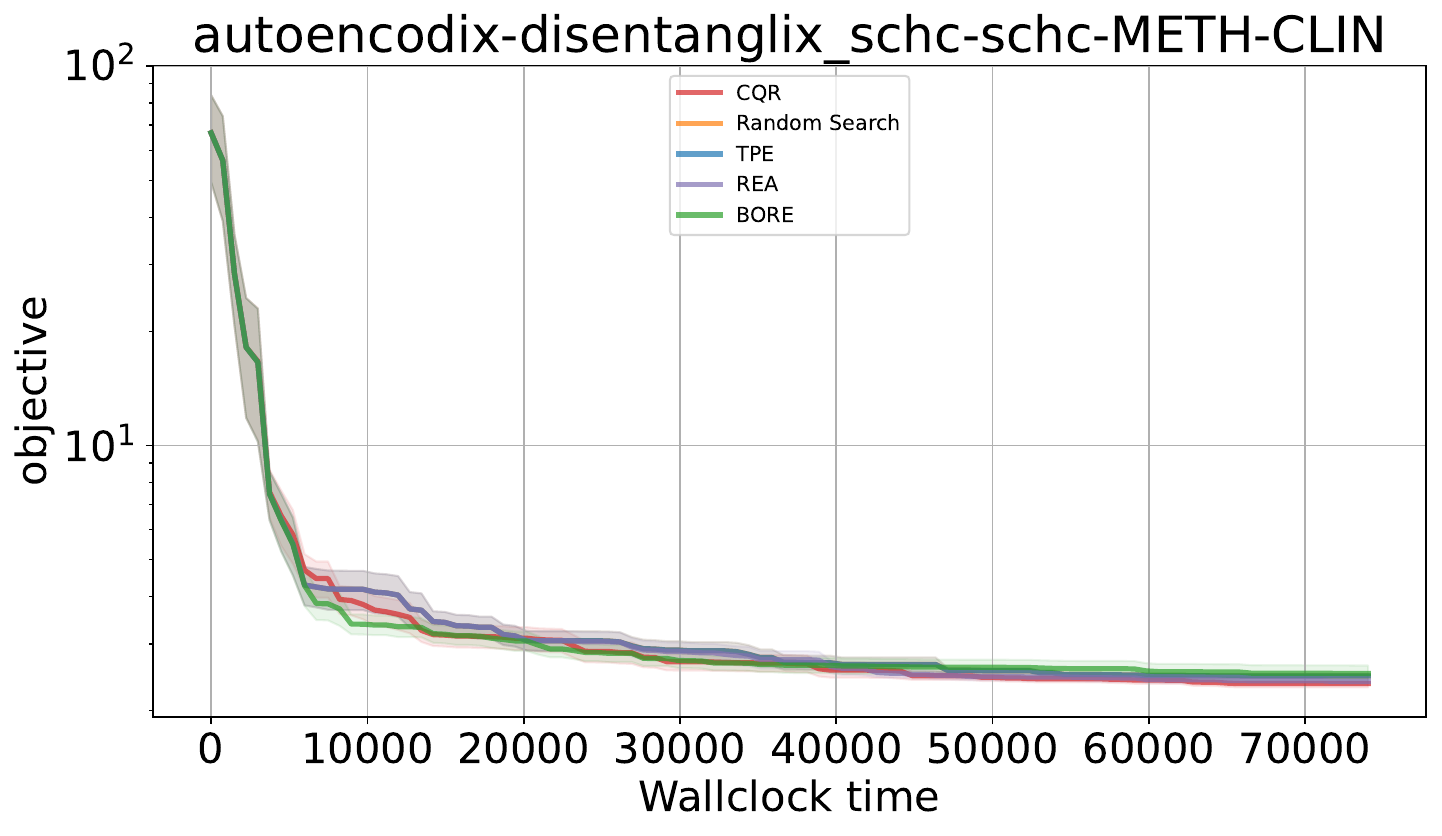} &
    \includegraphics[width=0.32\textwidth]{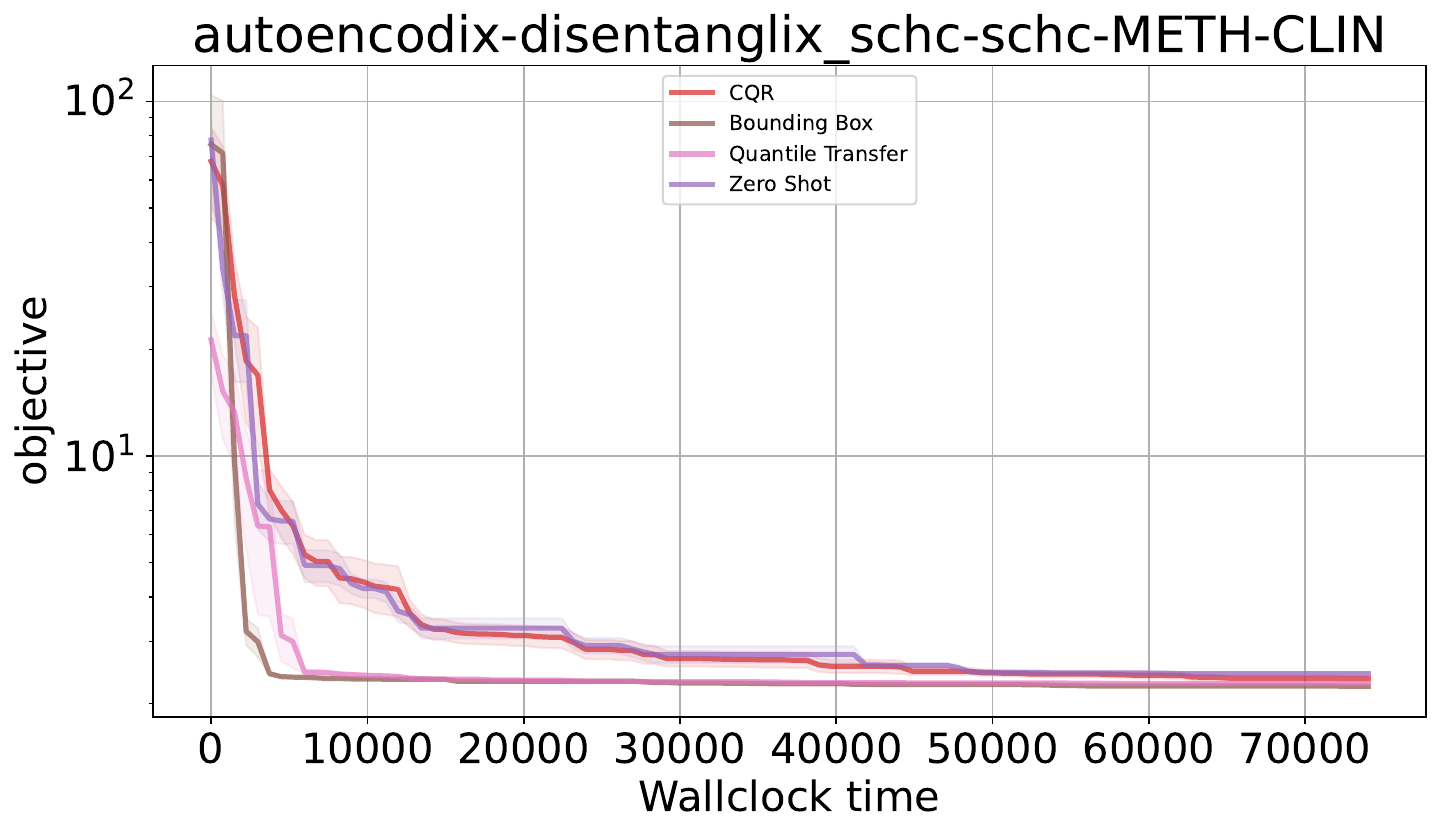} &
    \includegraphics[width=0.32\textwidth]{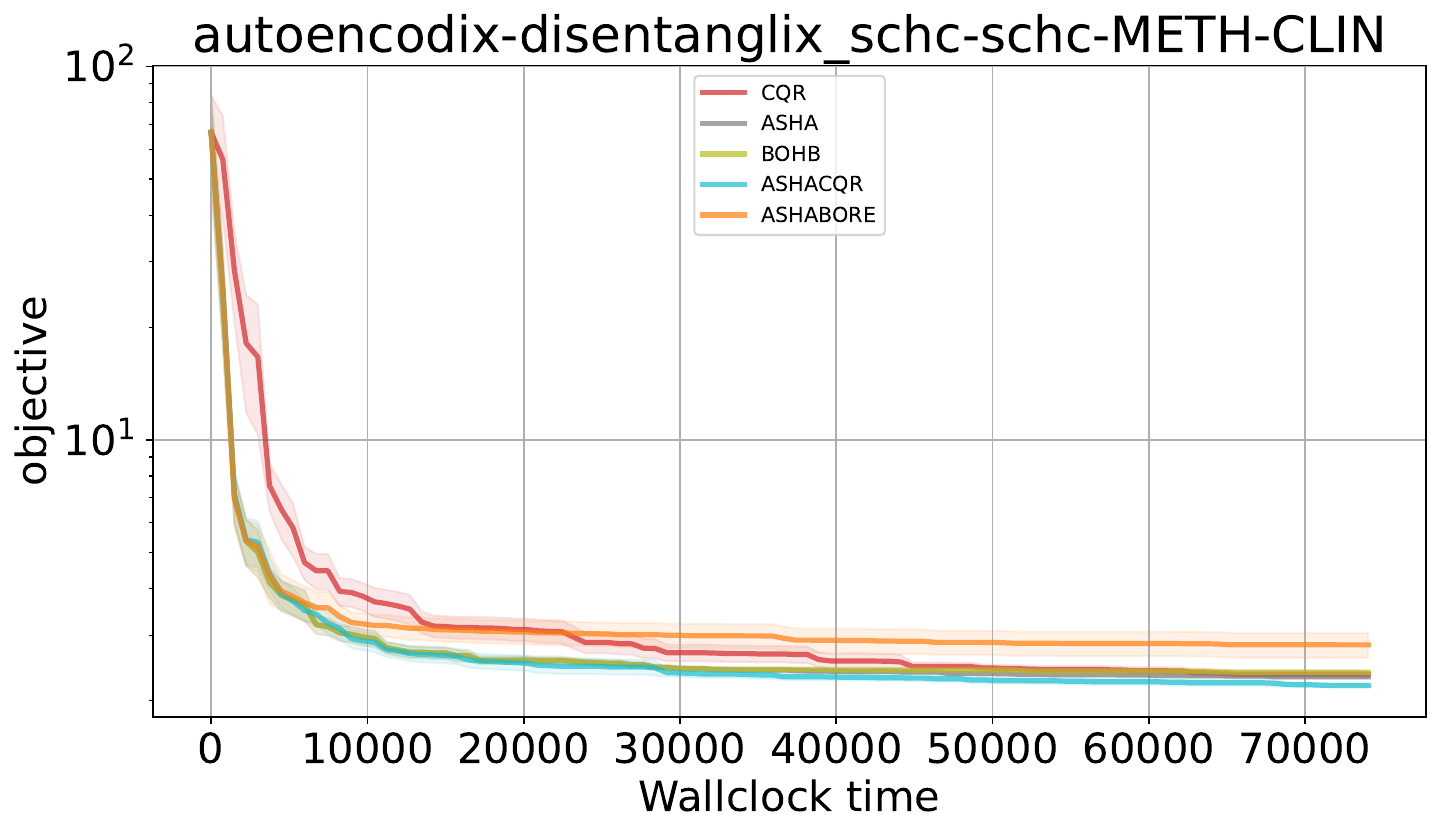} \\
    \includegraphics[width=0.32\textwidth]{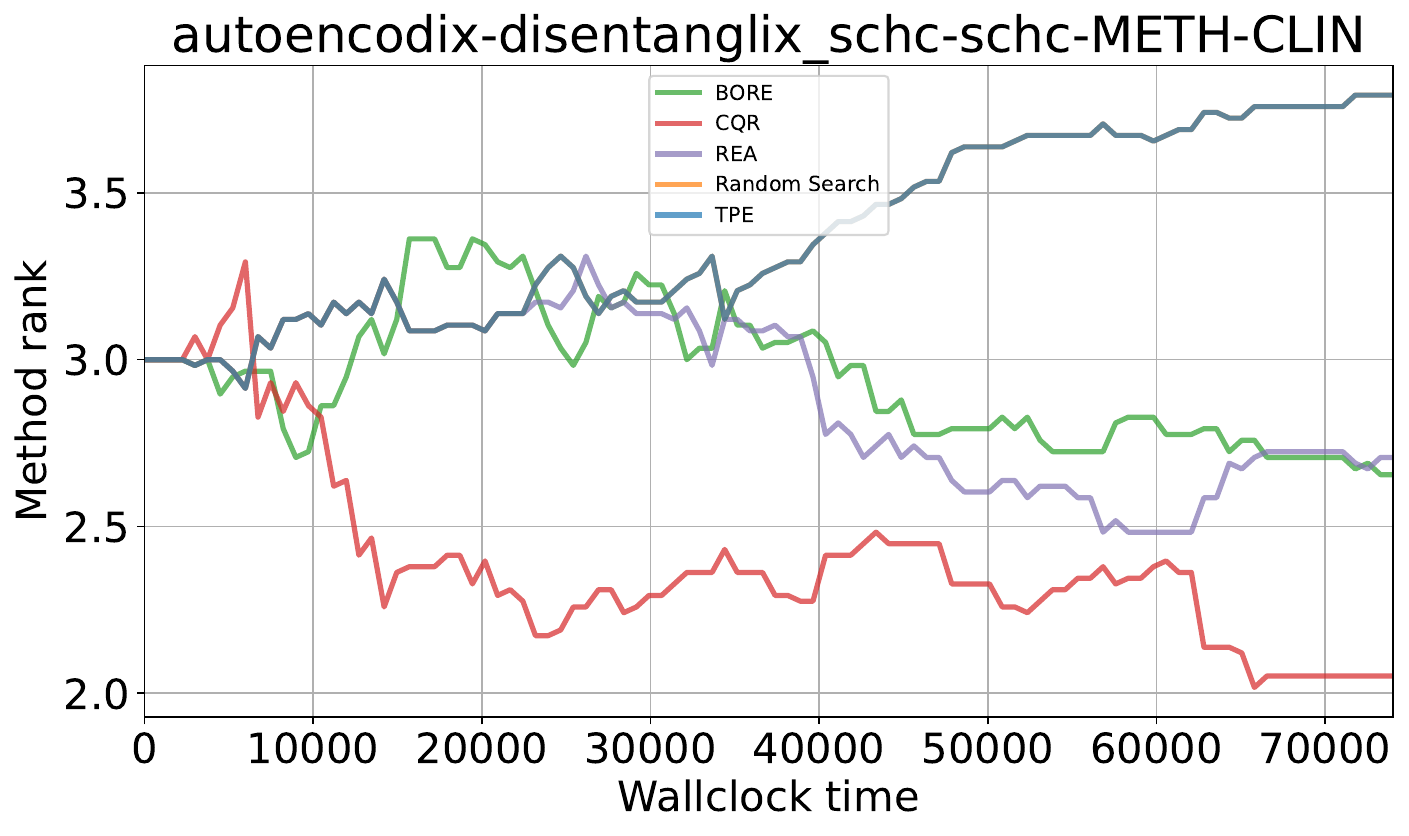} &
    \includegraphics[width=0.32\textwidth]{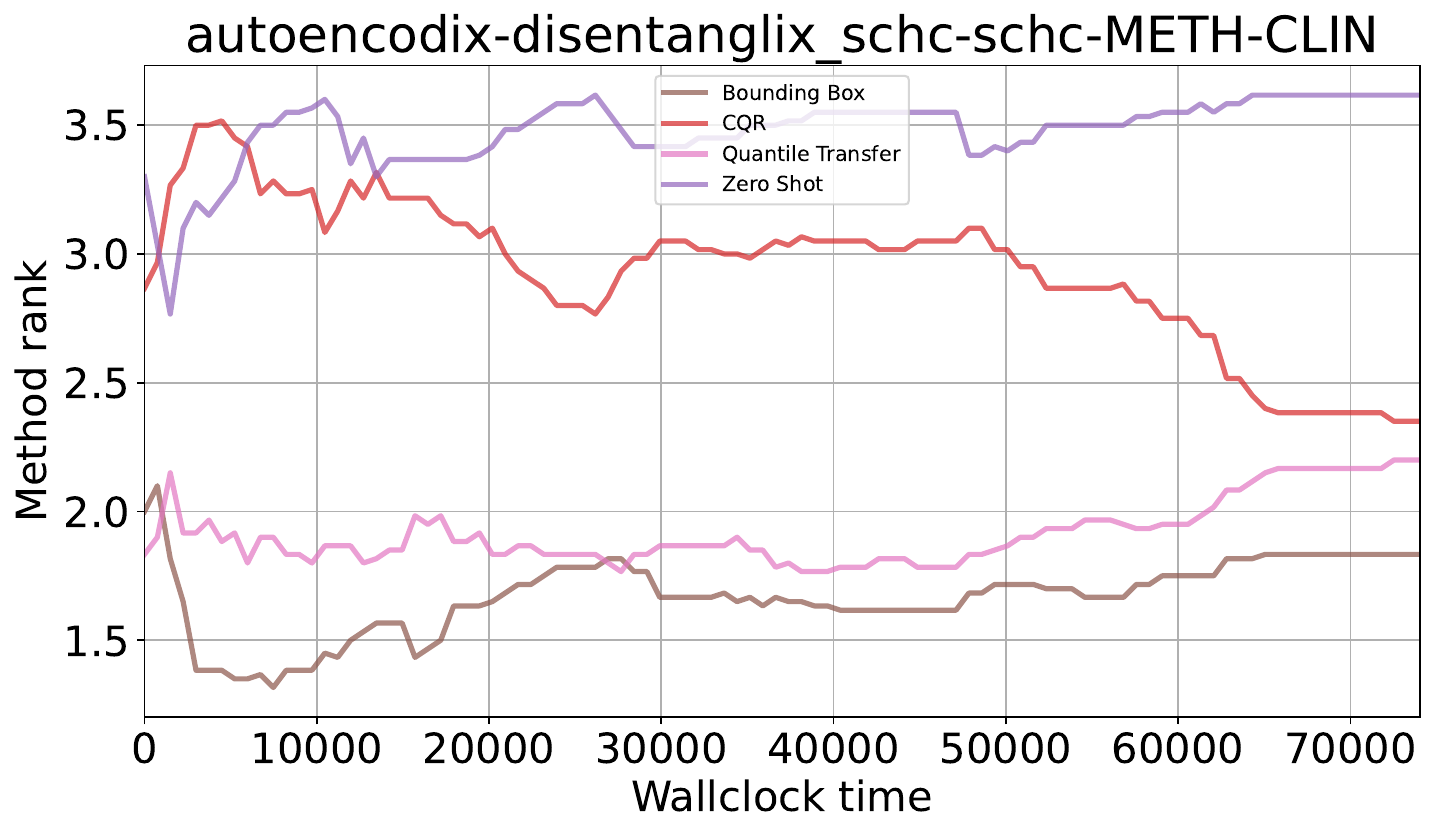} &
    \includegraphics[width=0.32\textwidth]{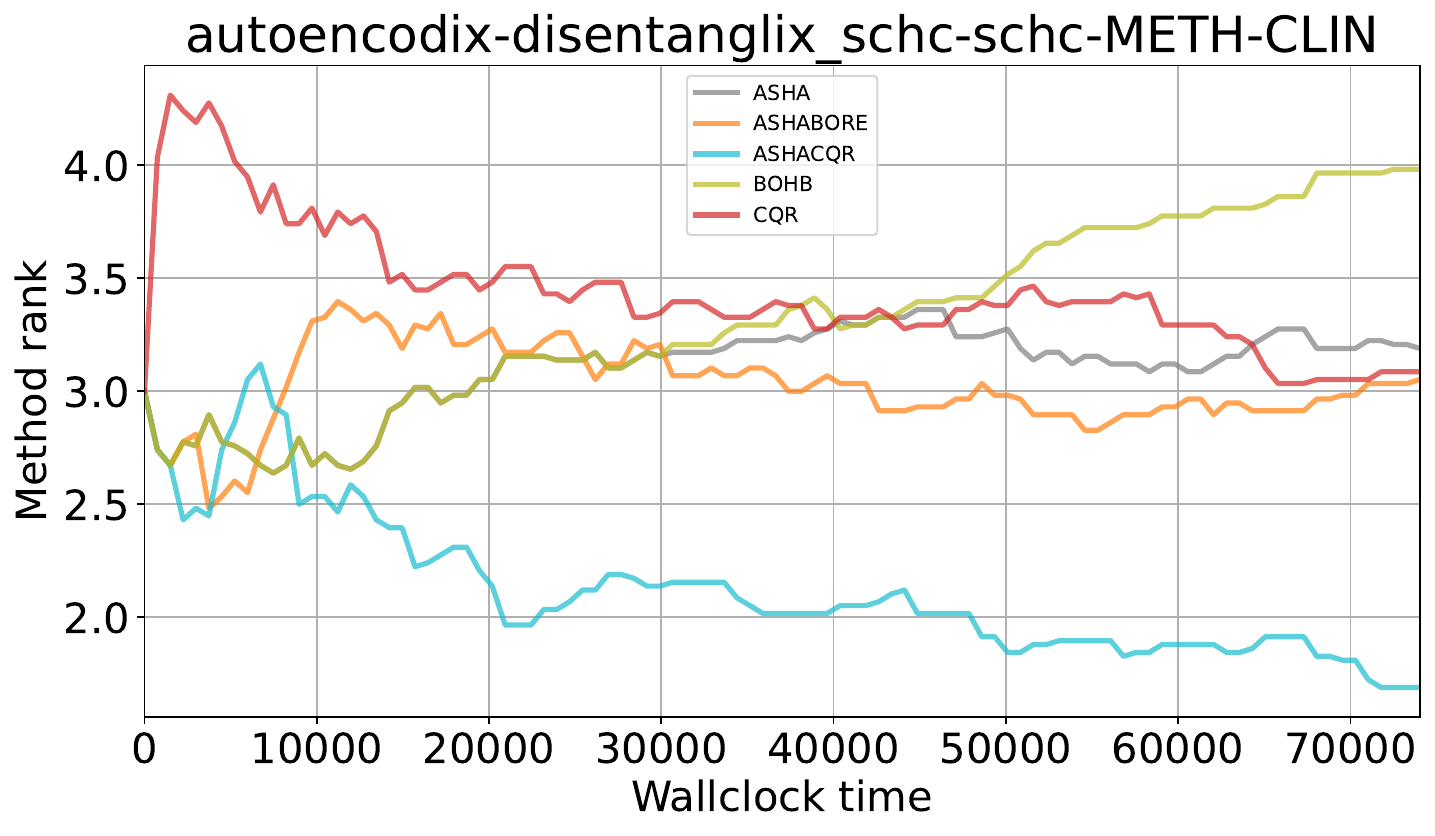} \\
    \midrule
    \multicolumn{3}{c}{\textbf{autoencodix-disentanglix\_schc-schc-RNA-CLIN}} \\
    \includegraphics[width=0.32\textwidth]{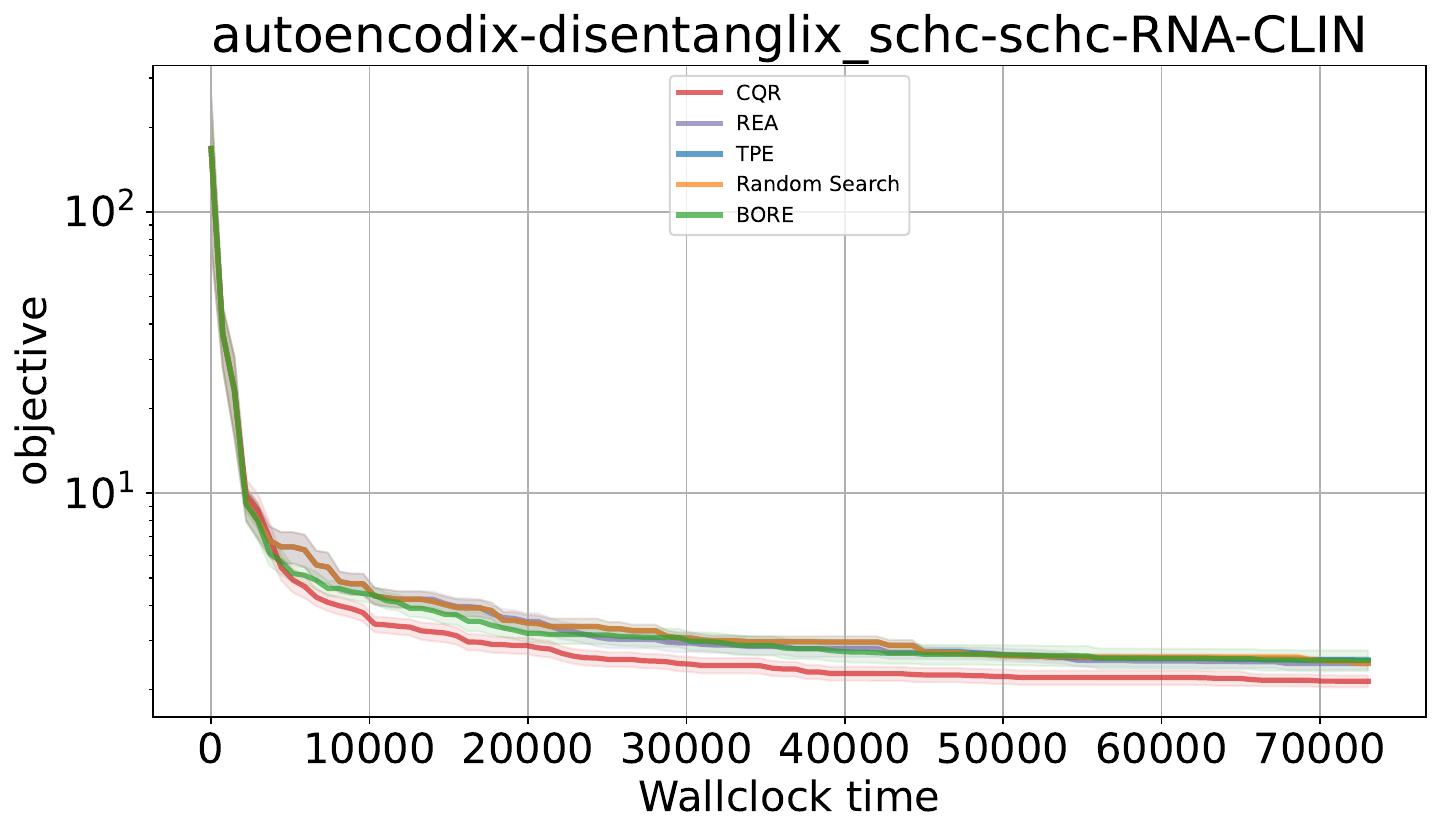} &
    \includegraphics[width=0.32\textwidth]{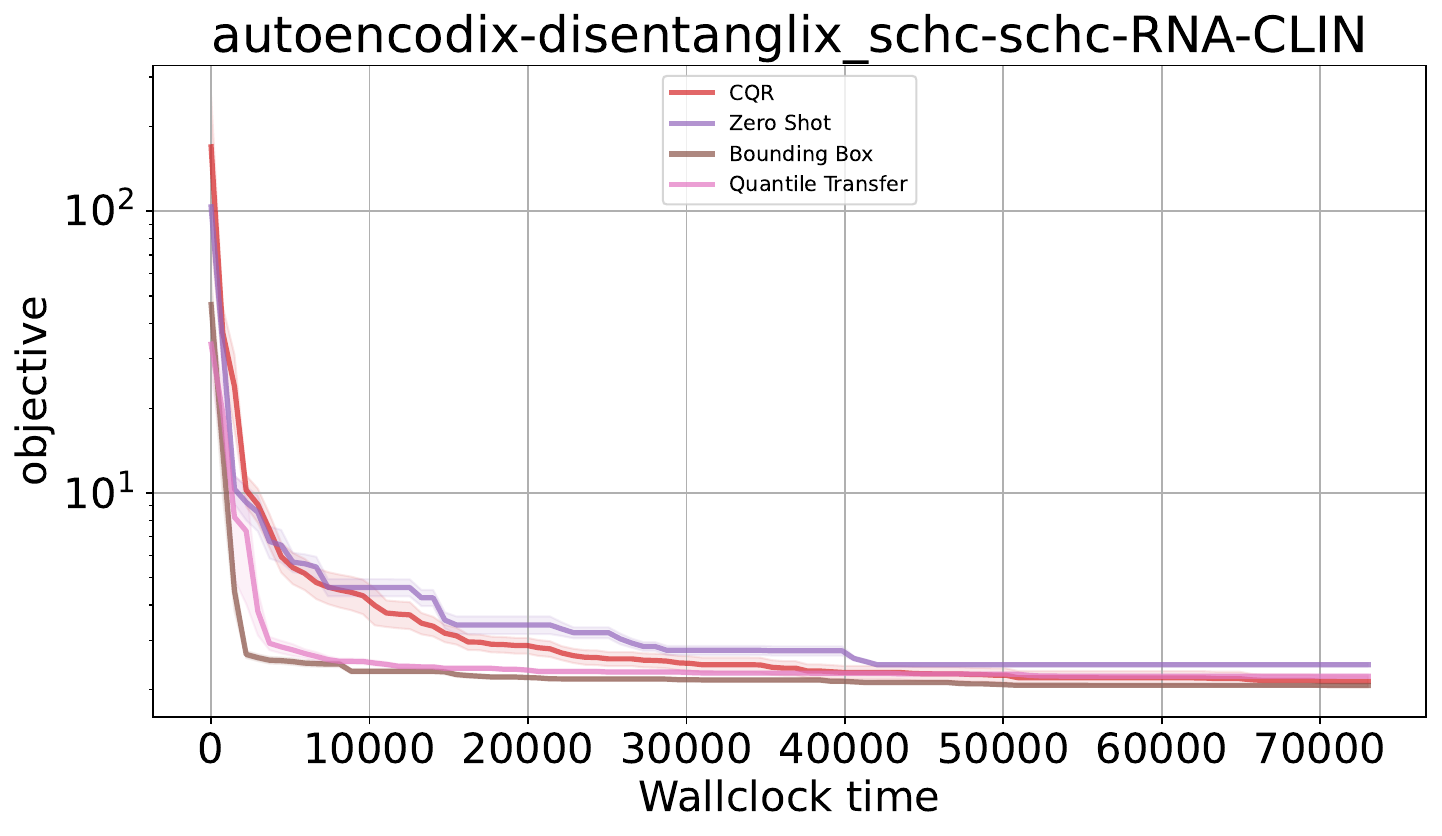} &
    \includegraphics[width=0.32\textwidth]{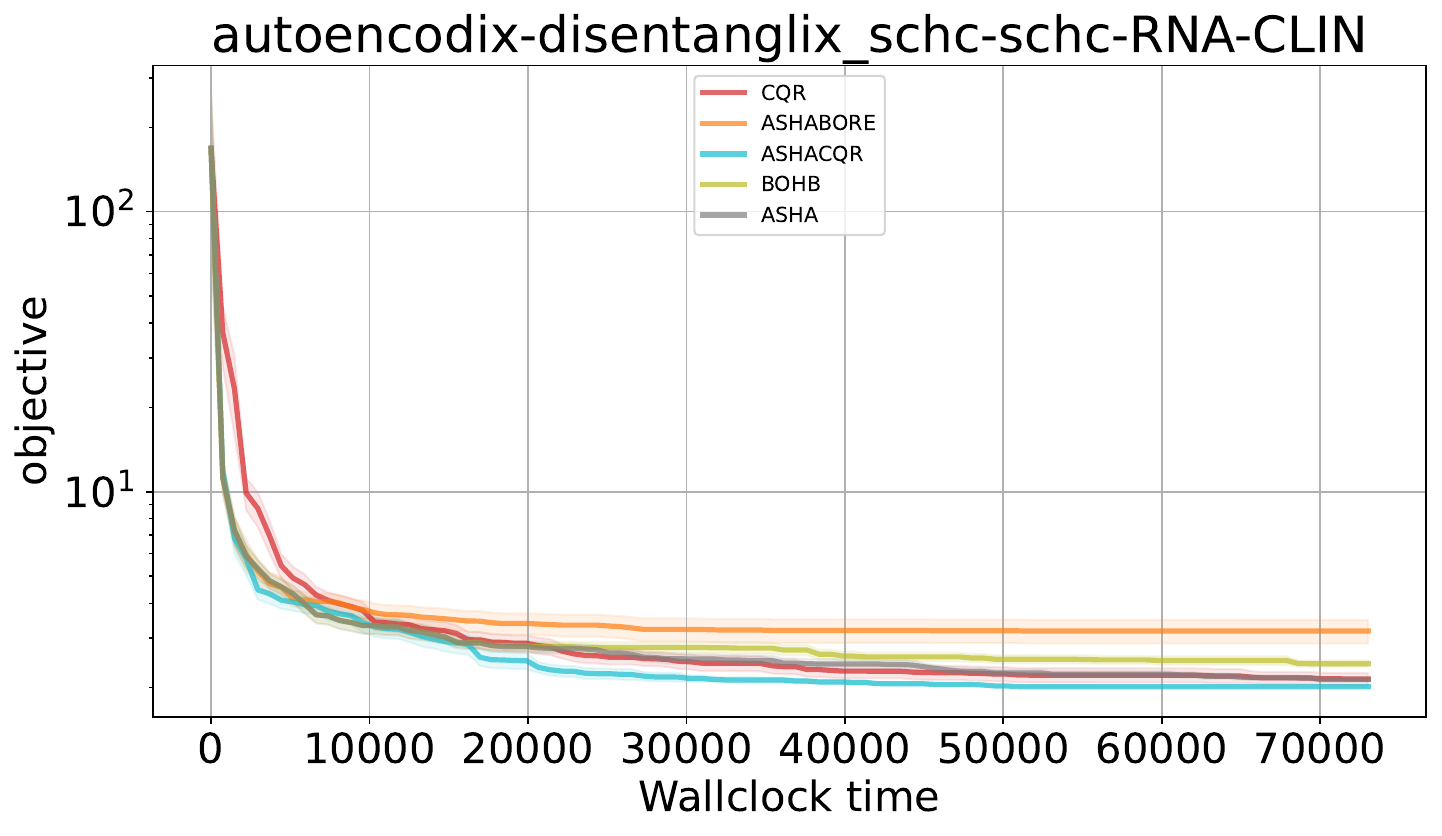} \\
    \includegraphics[width=0.32\textwidth]{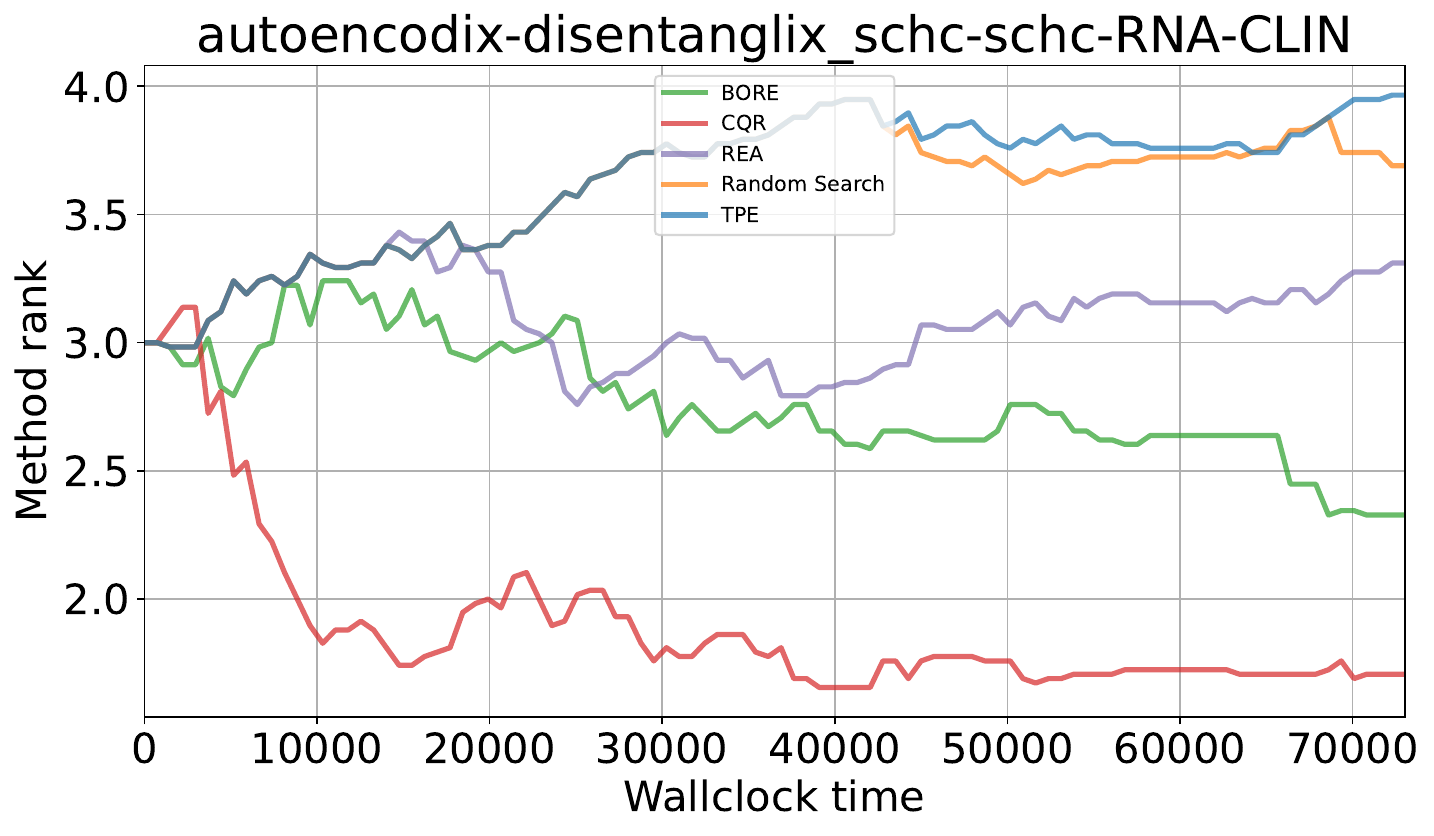} &
    \includegraphics[width=0.32\textwidth]{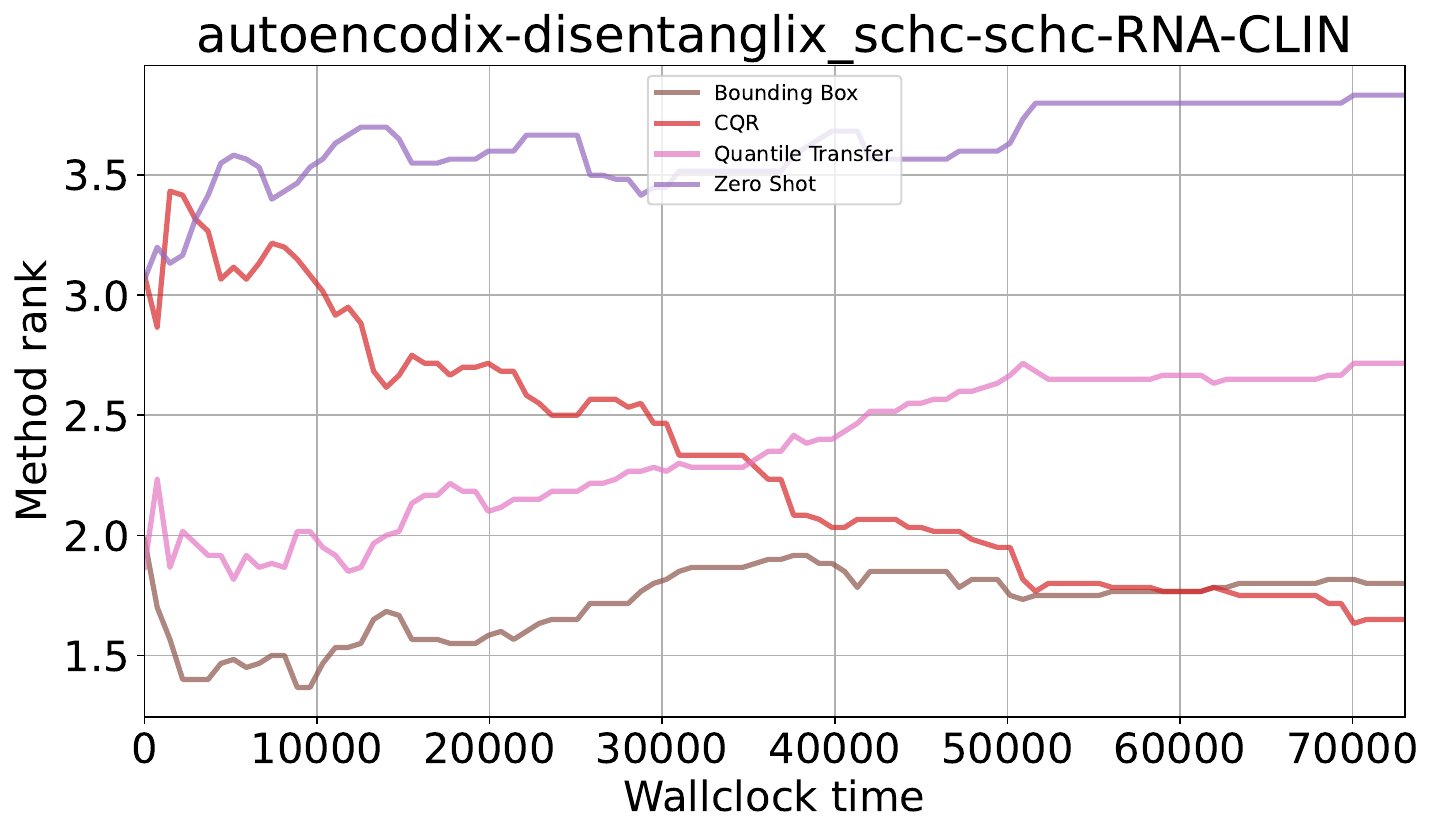} &
    \includegraphics[width=0.32\textwidth]{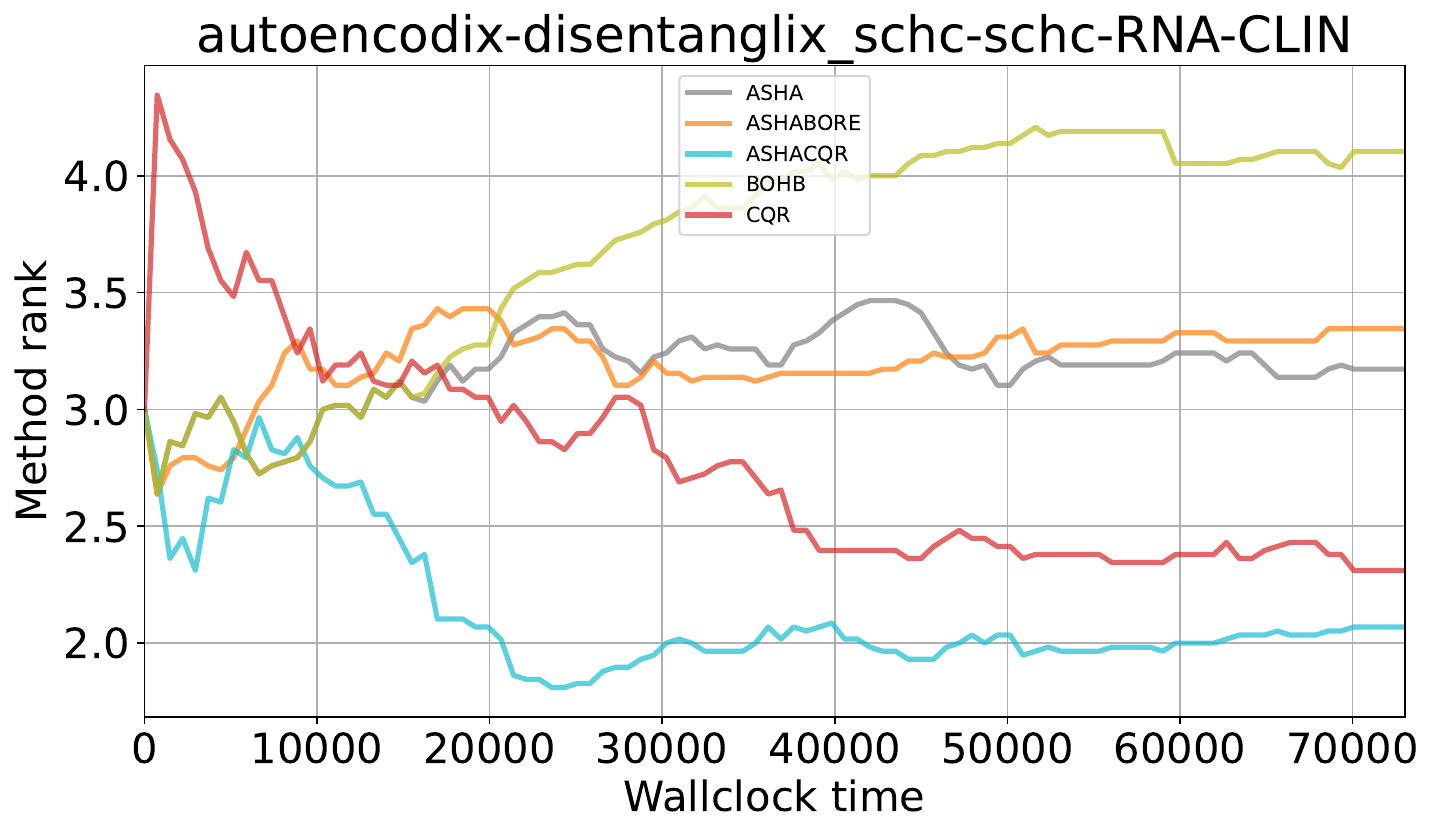} \\
    \midrule
    \multicolumn{3}{c}{\textbf{autoencodix-disentanglix\_schc-schc-RNA-METH-CLIN}} \\
    \includegraphics[width=0.32\textwidth]{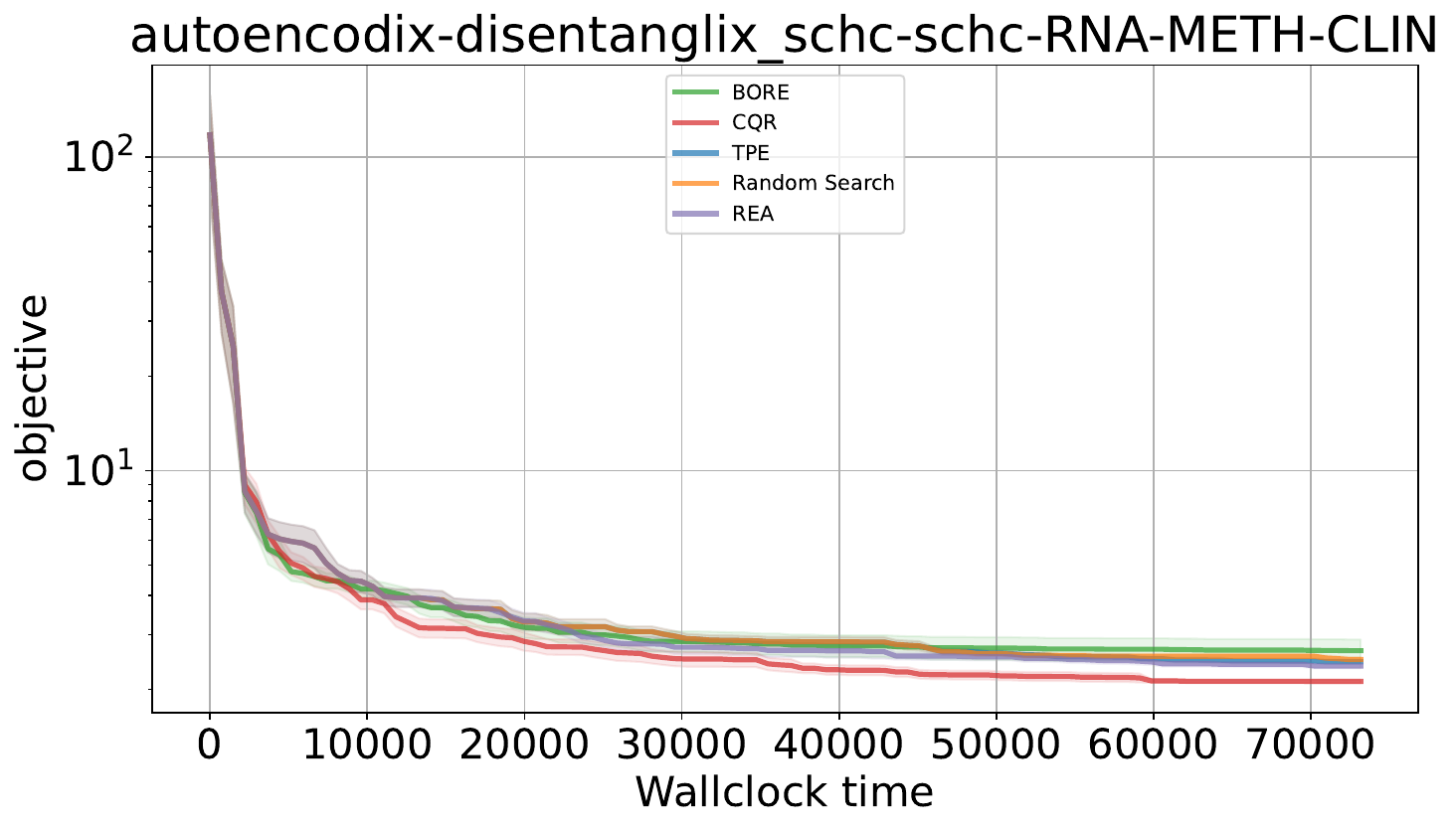} &
    \includegraphics[width=0.32\textwidth]{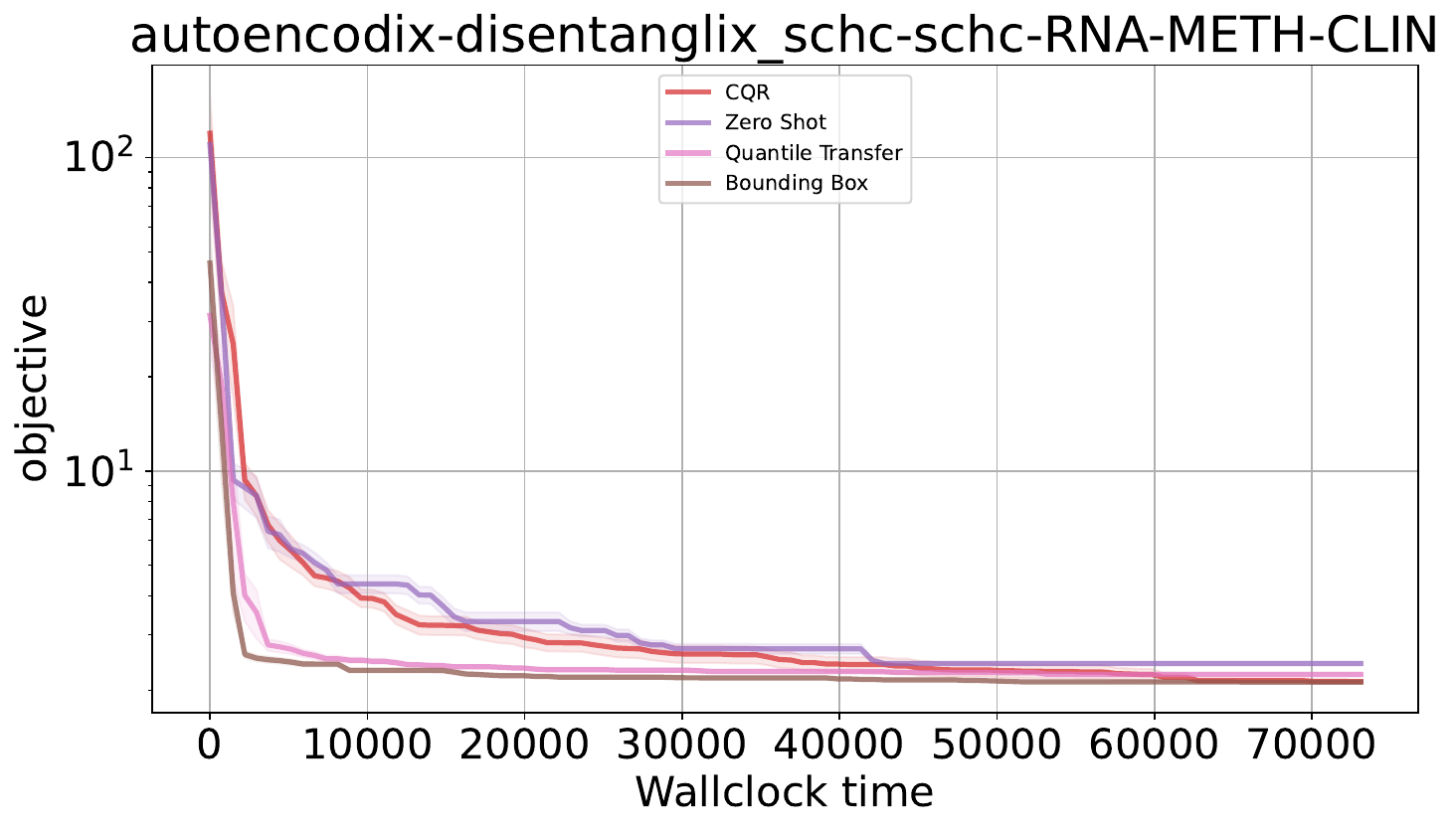} &
    \includegraphics[width=0.32\textwidth]{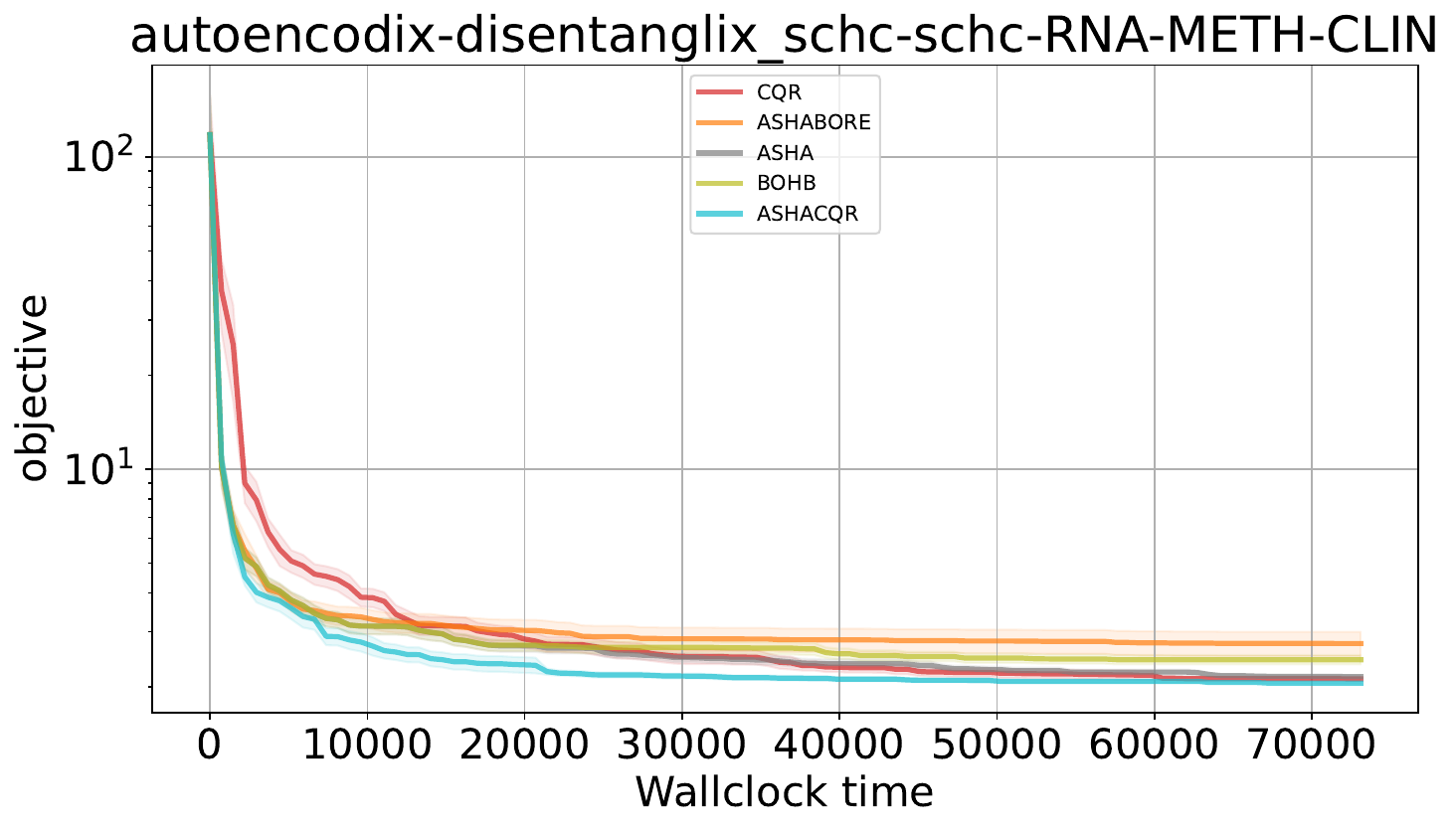} \\
    \includegraphics[width=0.32\textwidth]{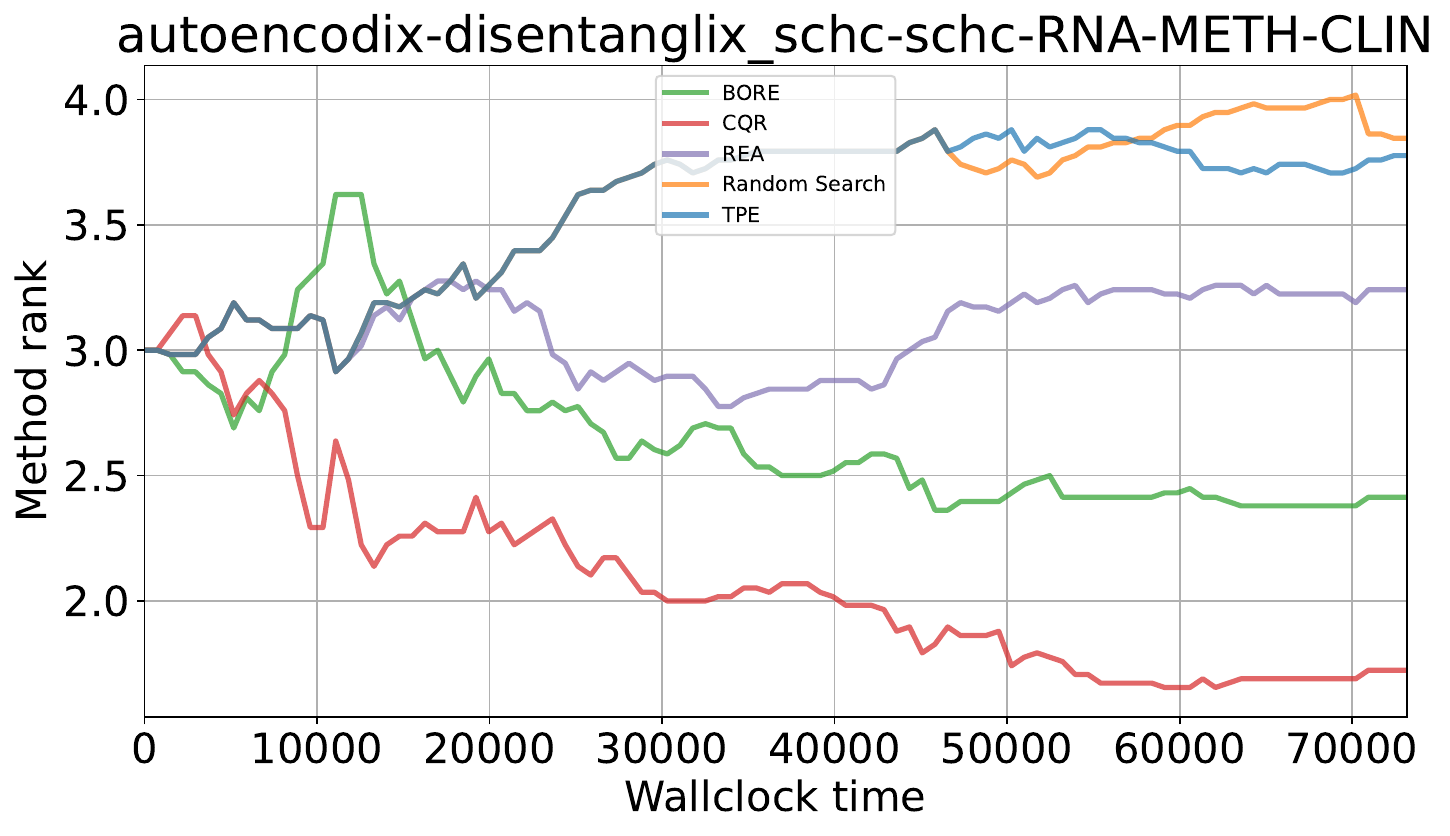} &
    \includegraphics[width=0.32\textwidth]{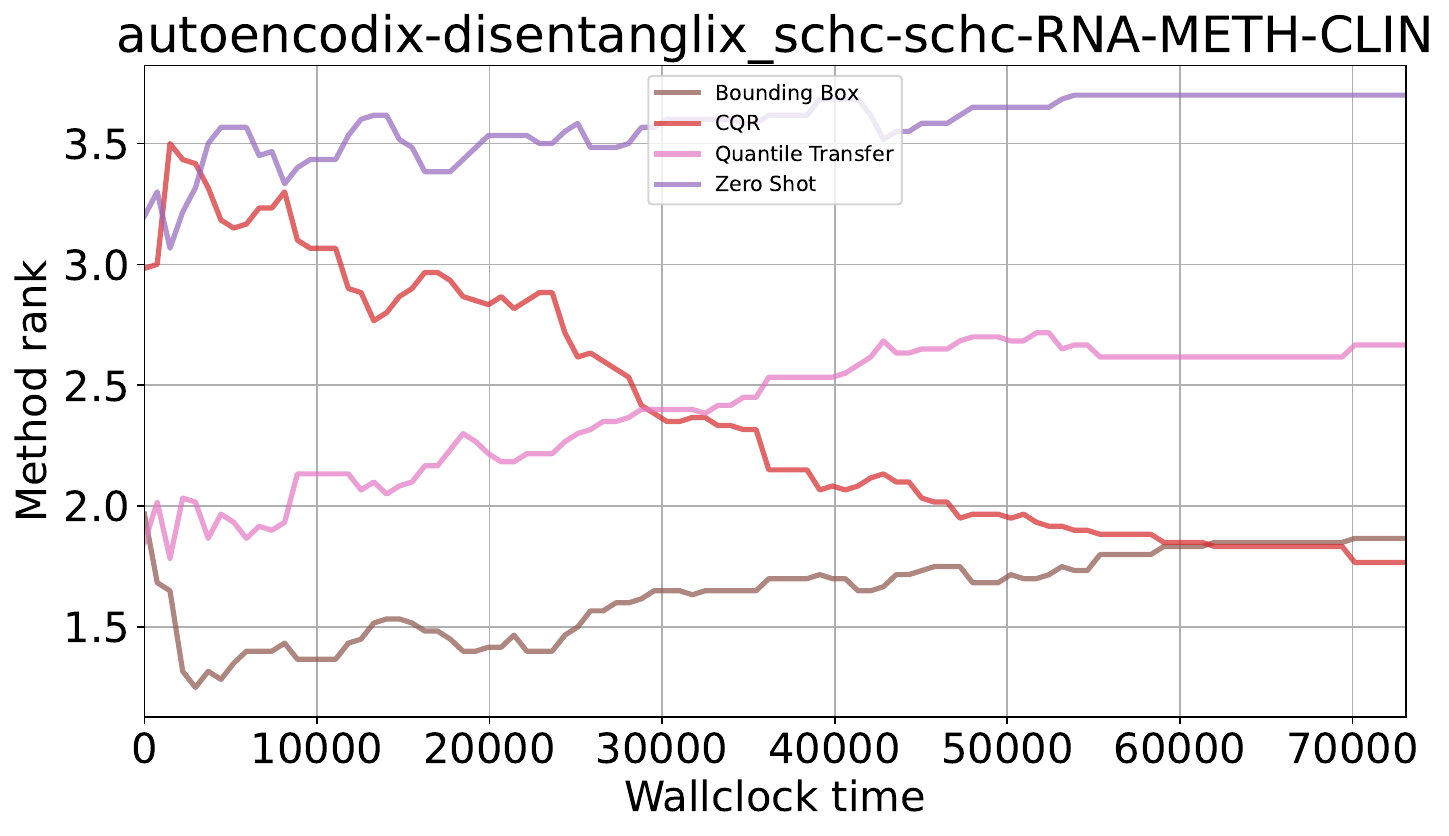} &
    \includegraphics[width=0.32\textwidth]{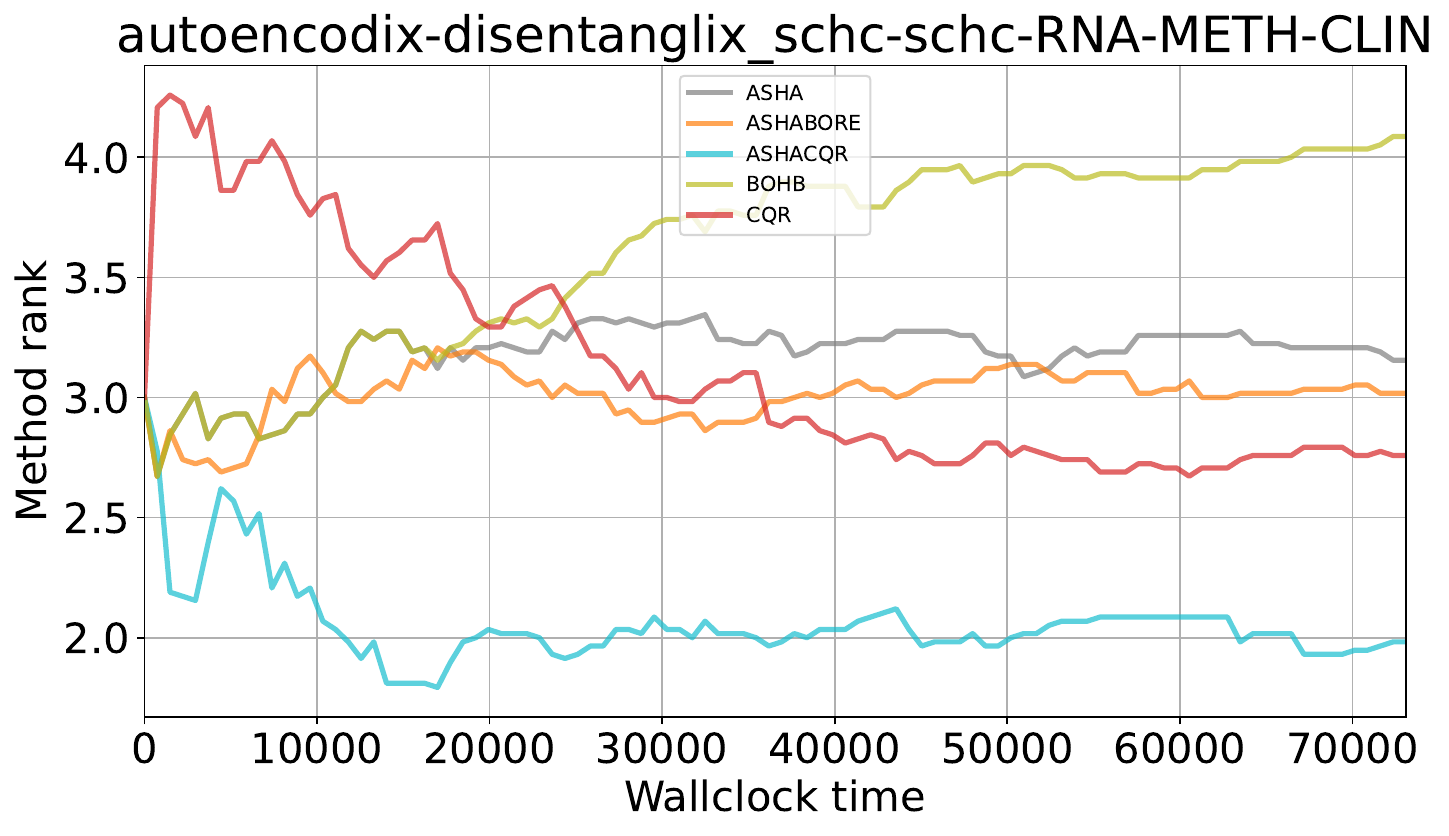} \\
    \end{tabular}
    \caption{Results for Disentanglix tasks (Part 1).}
    \label{fig:disentanglix_part1}
\end{figure}

\clearpage

\begin{figure}[htbp]
    \centering
    \setlength{\tabcolsep}{1pt}
    \begin{tabular}{ccc}
    \multicolumn{3}{c}{\textbf{autoencodix-disentanglix\_tcga-tcga-DNA-CLIN}} \\
    \textbf{Single-Fidelity} & \textbf{Transfer Learning} & \textbf{Multi-Fidelity} \\
    \includegraphics[width=0.32\textwidth]{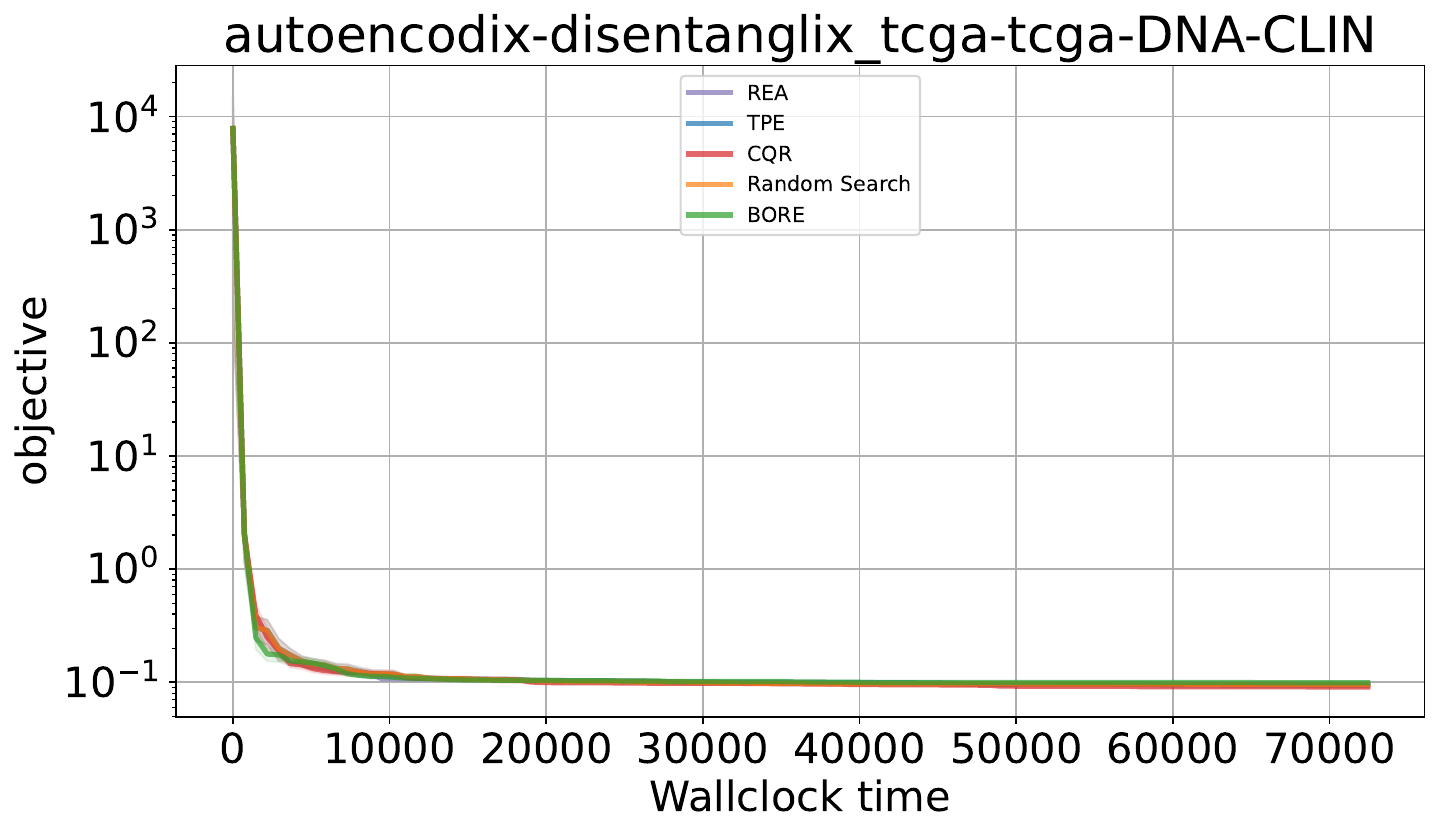} &
    \includegraphics[width=0.32\textwidth]{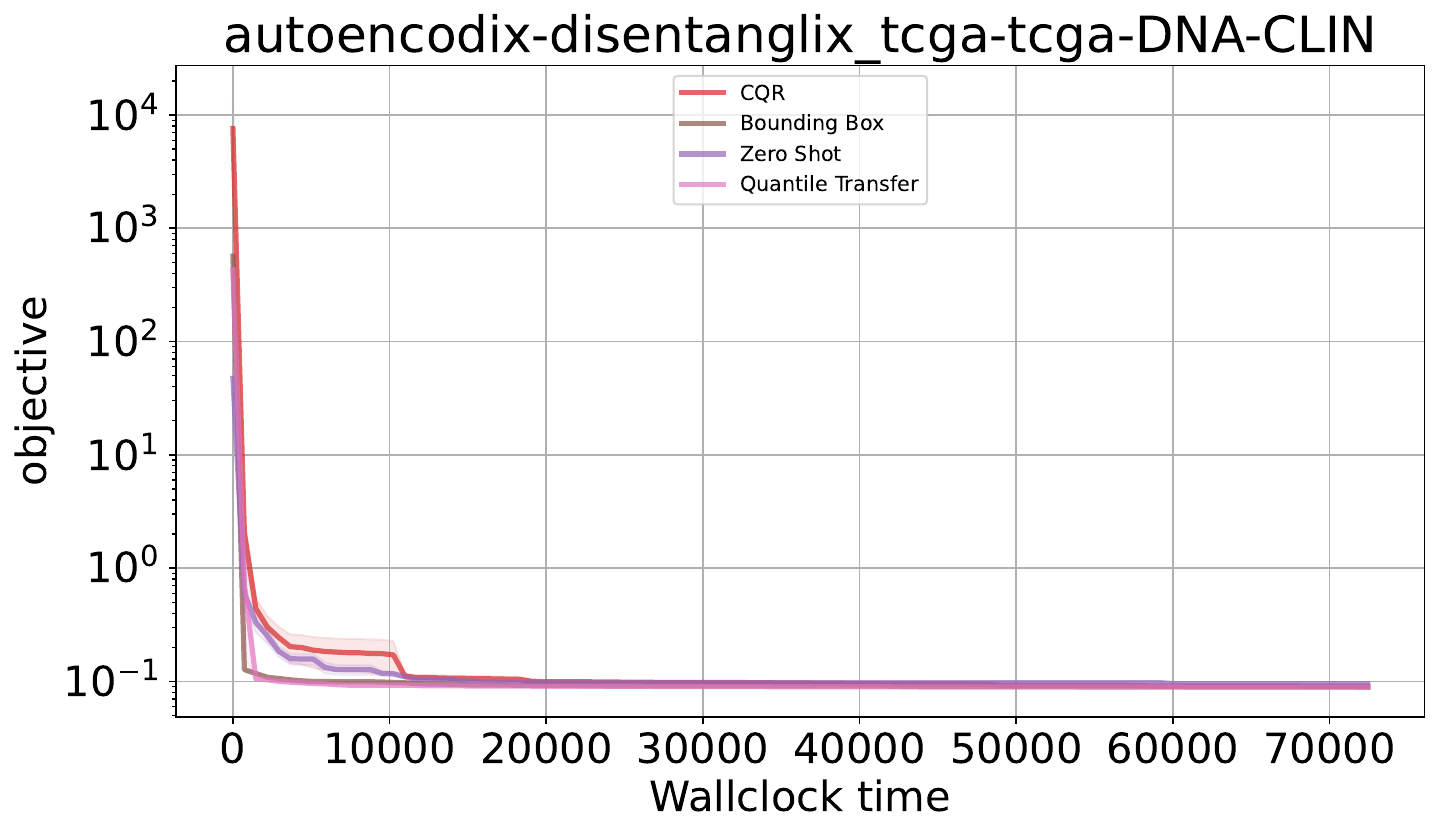} &
    \includegraphics[width=0.32\textwidth]{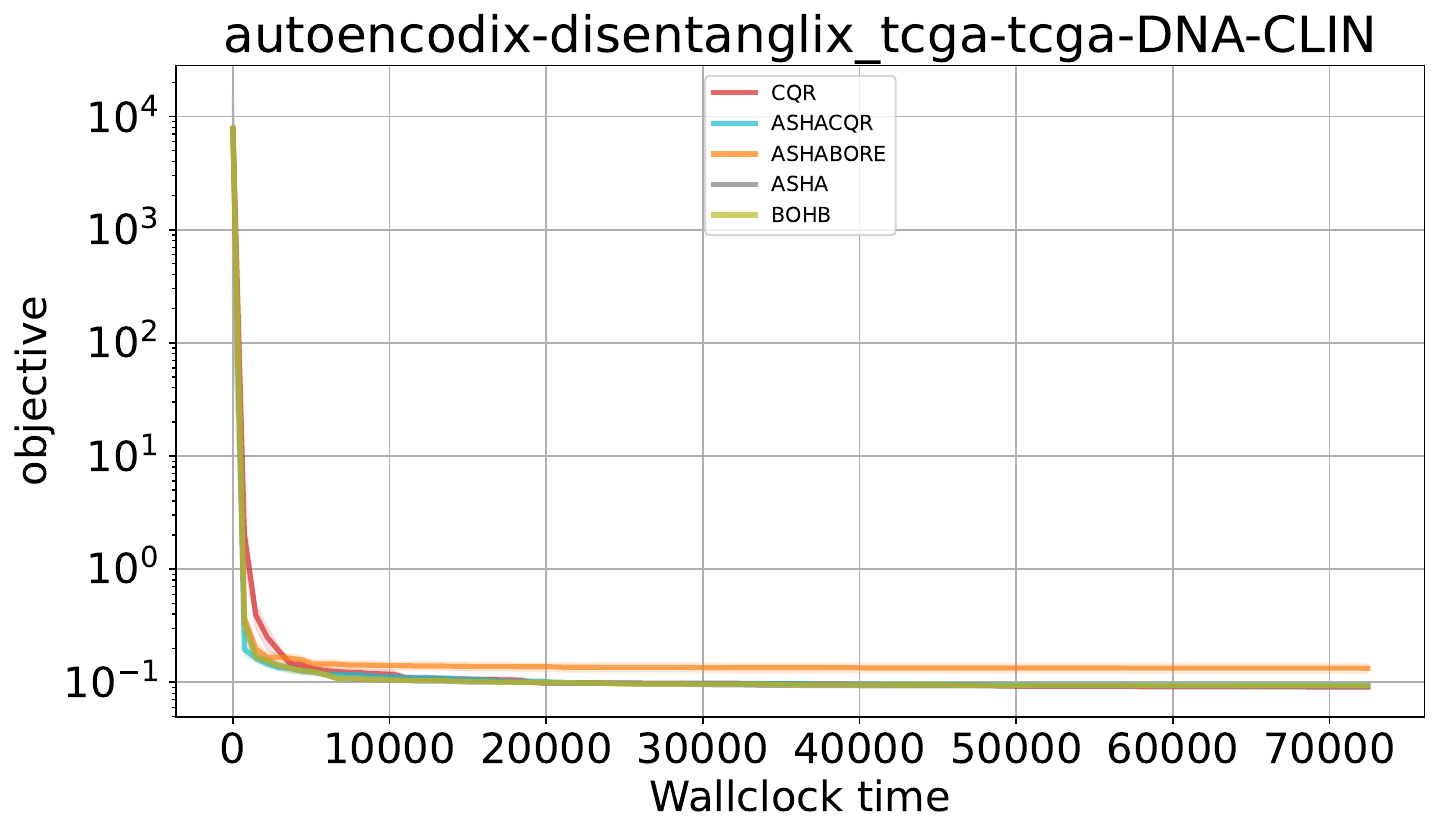} \\
    \includegraphics[width=0.32\textwidth]{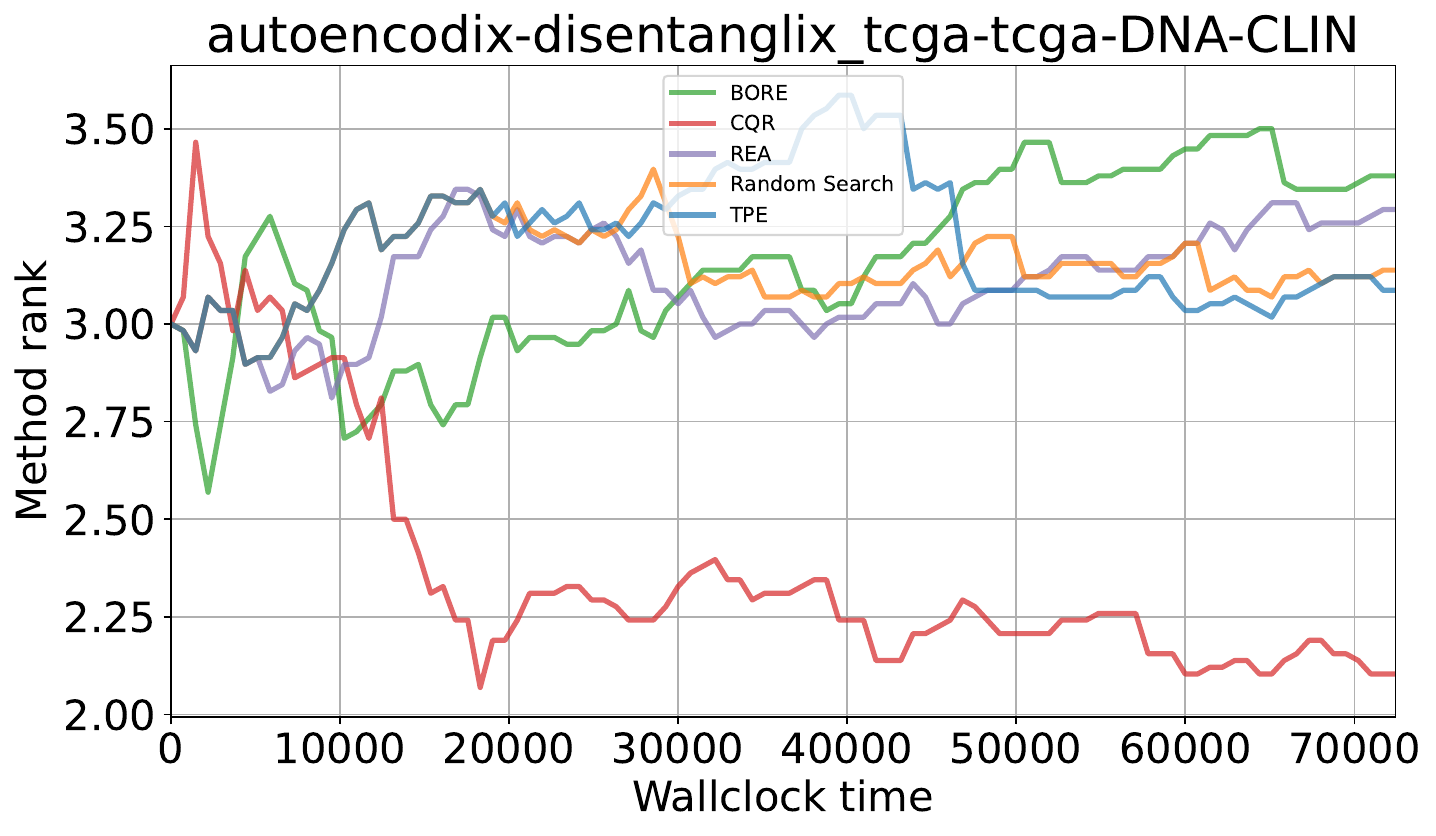} &
    \includegraphics[width=0.32\textwidth]{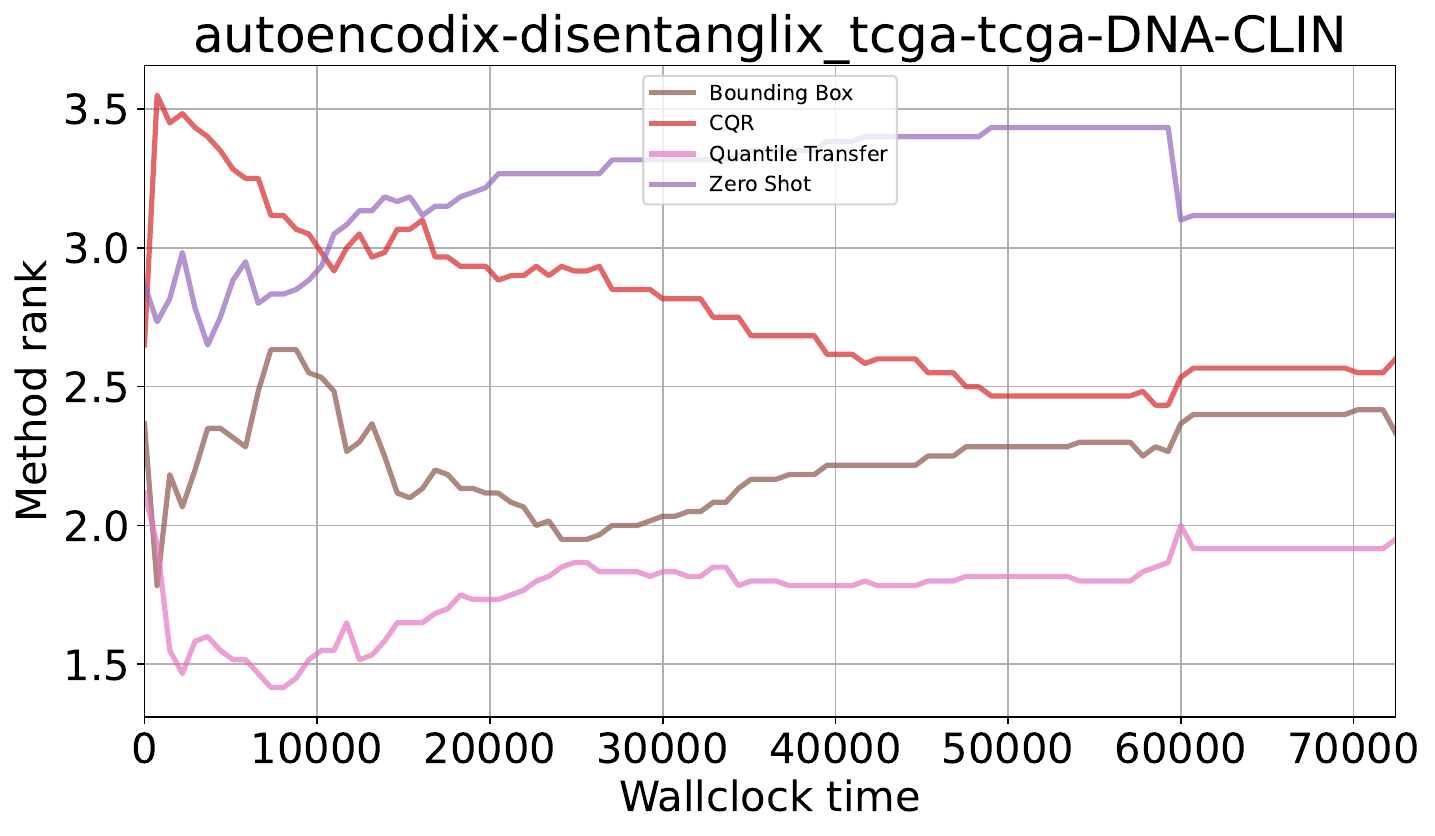} &
    \includegraphics[width=0.32\textwidth]{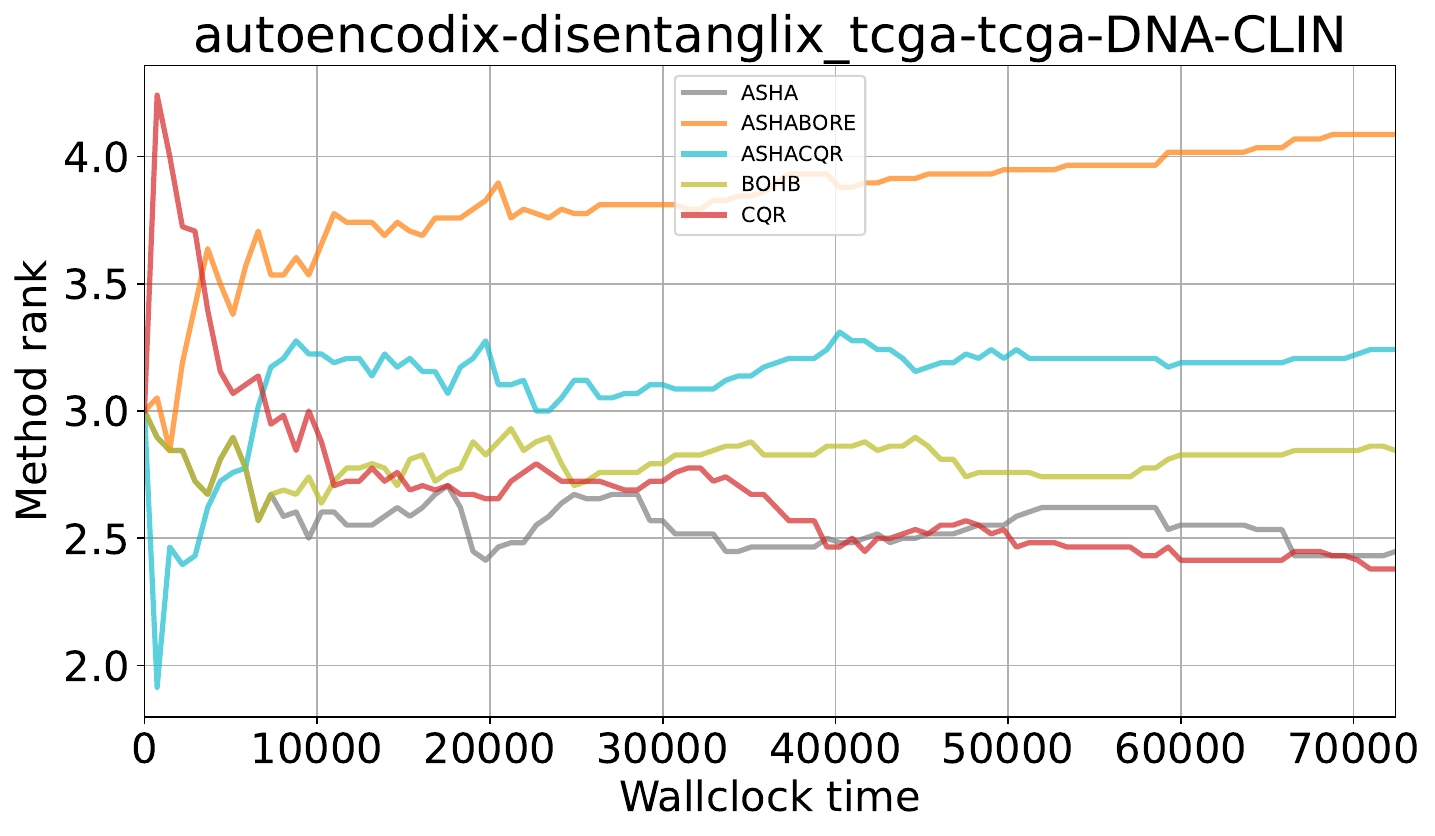} \\
    \midrule
    \multicolumn{3}{c}{\textbf{autoencodix-disentanglix\_tcga-tcga-METH-CLIN}} \\
    \includegraphics[width=0.32\textwidth]{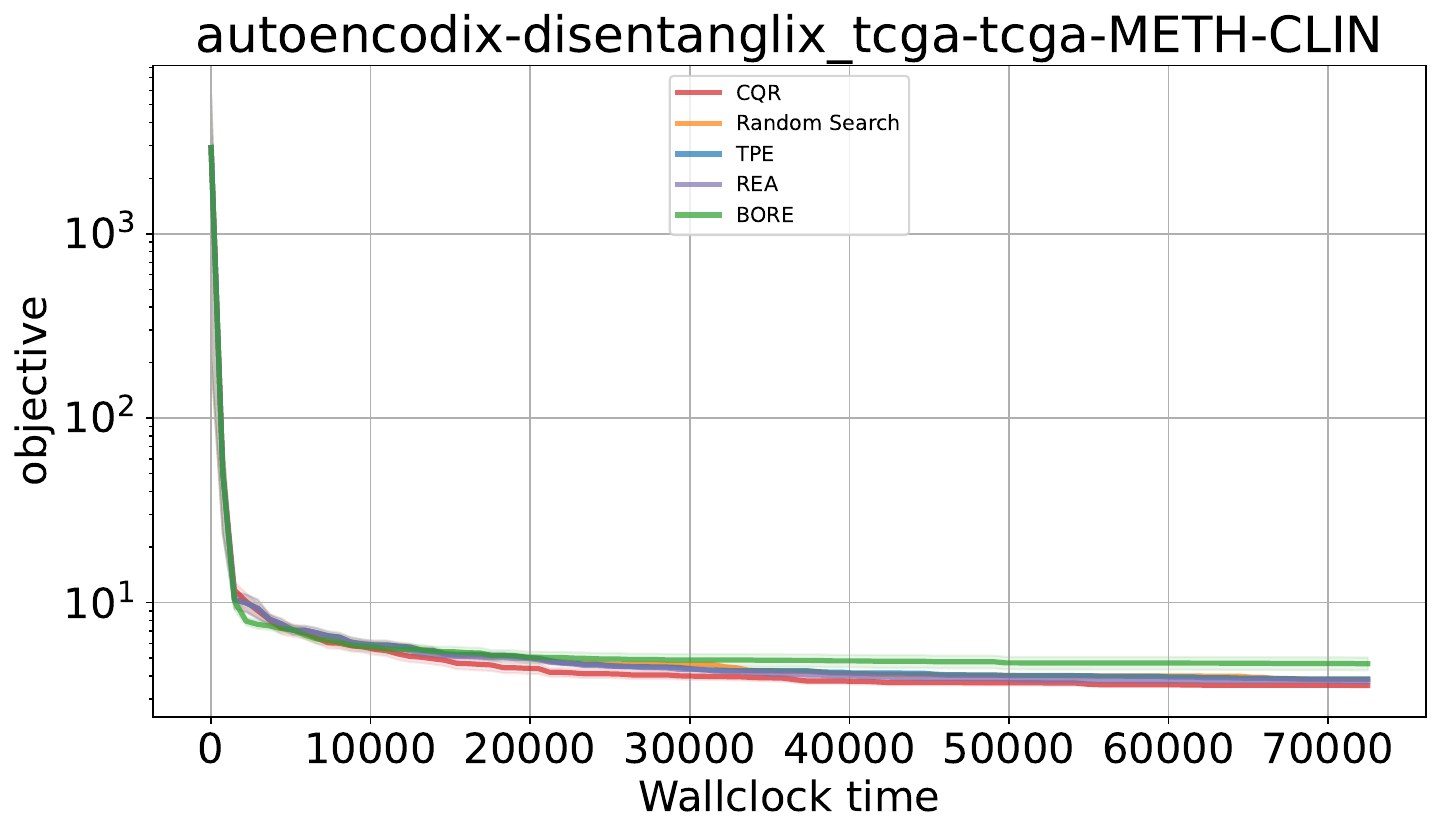} &
    \includegraphics[width=0.32\textwidth]{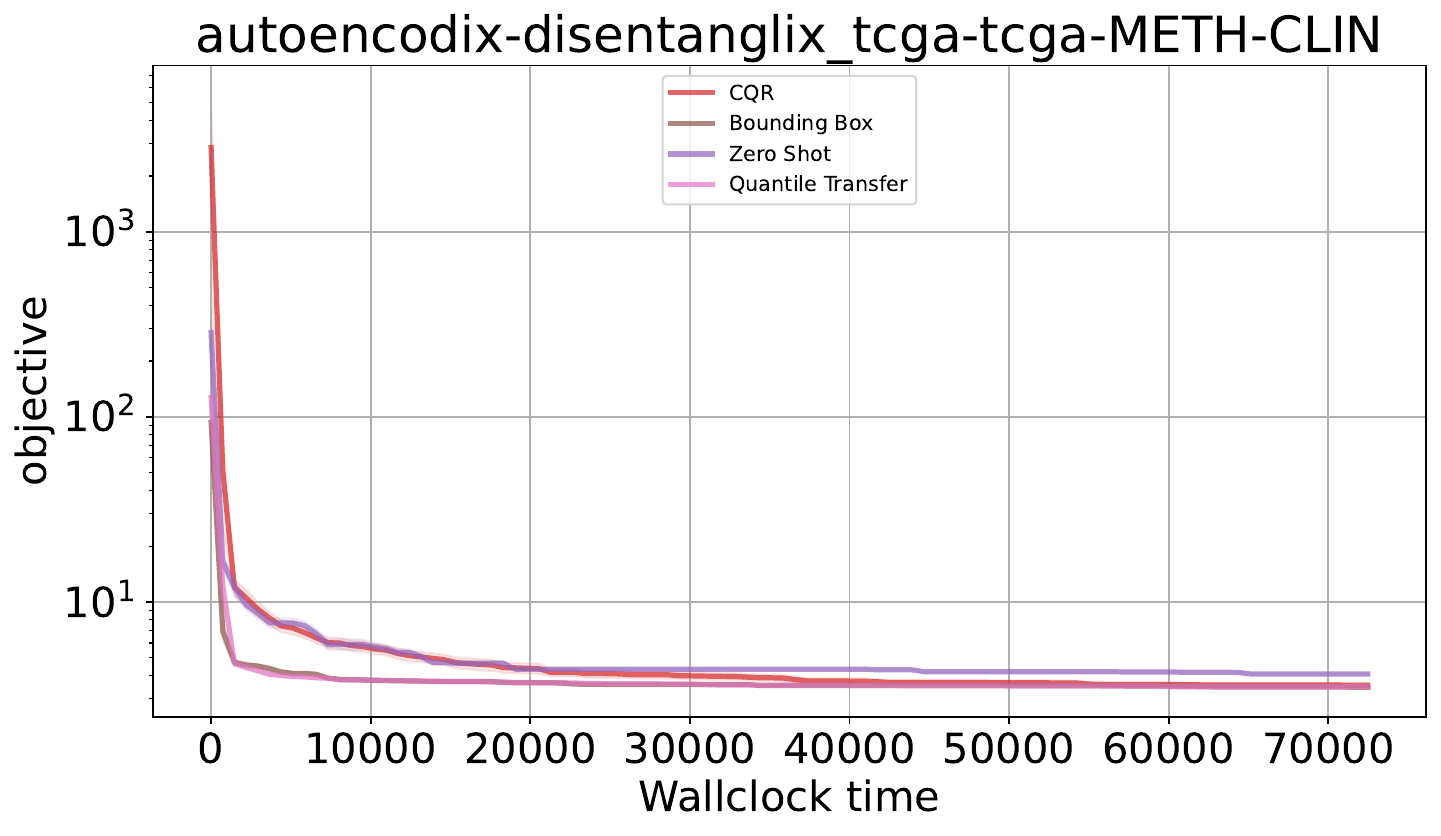} &
    \includegraphics[width=0.32\textwidth]{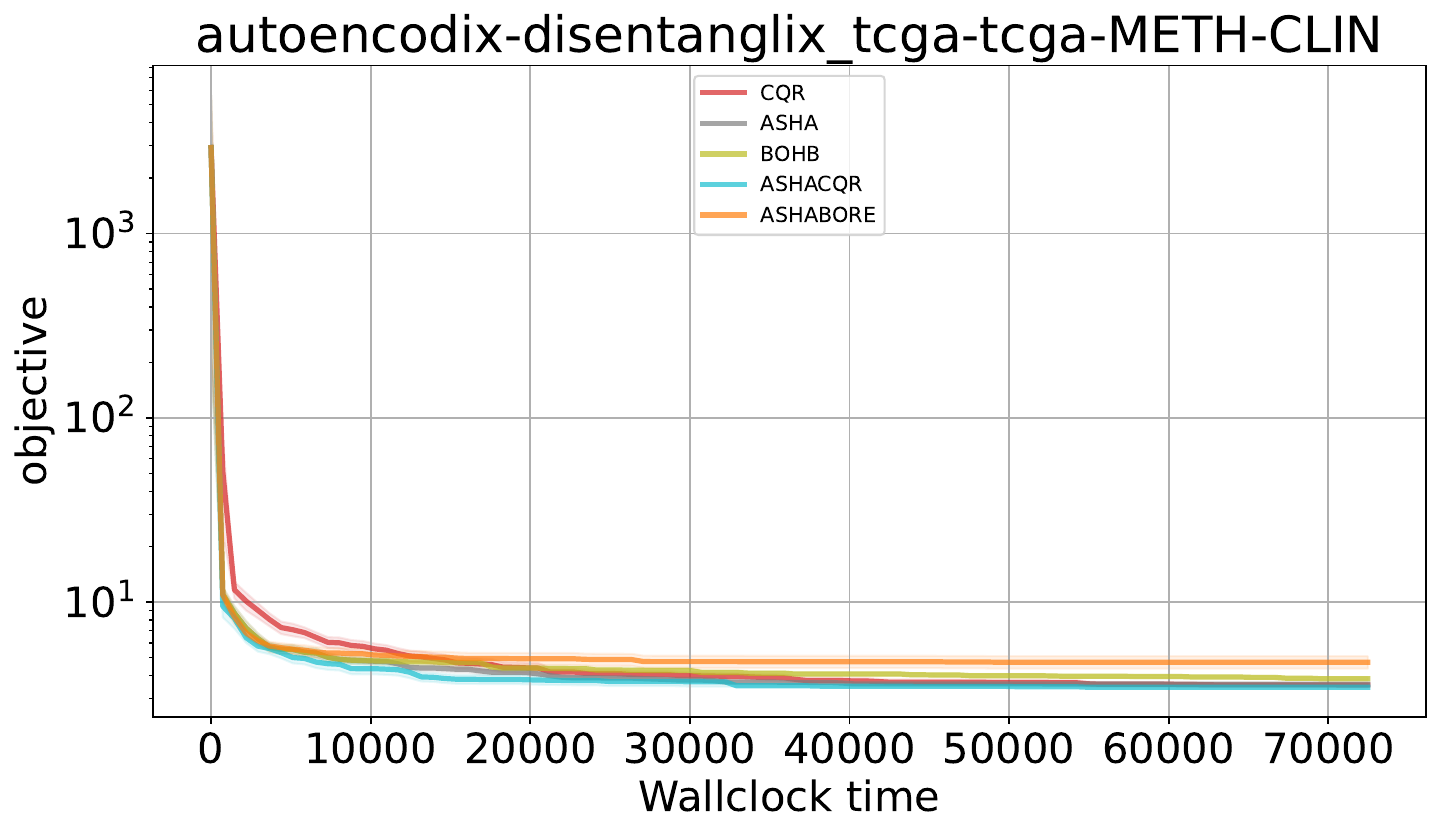} \\
    \includegraphics[width=0.32\textwidth]{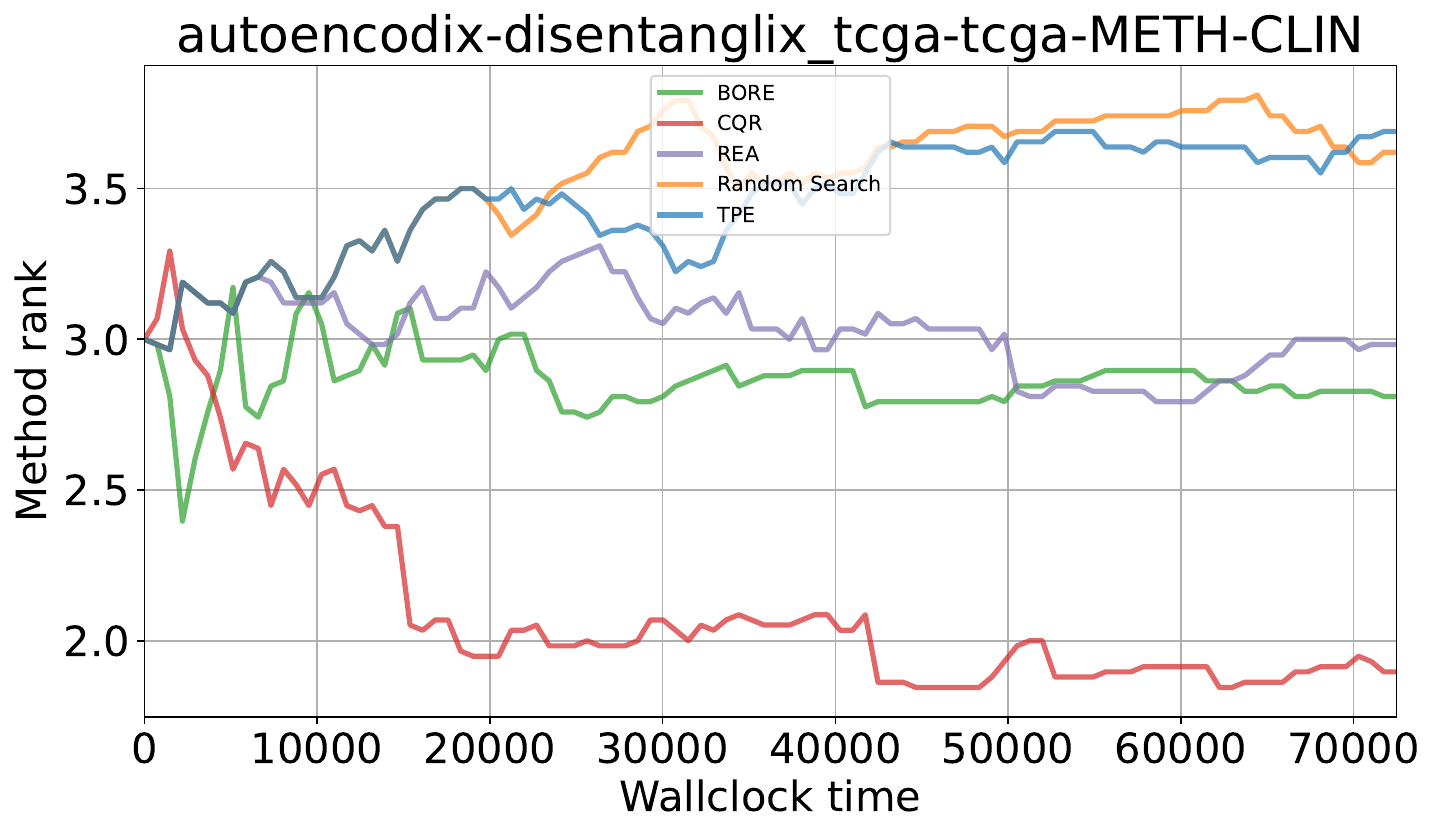} &
    \includegraphics[width=0.32\textwidth]{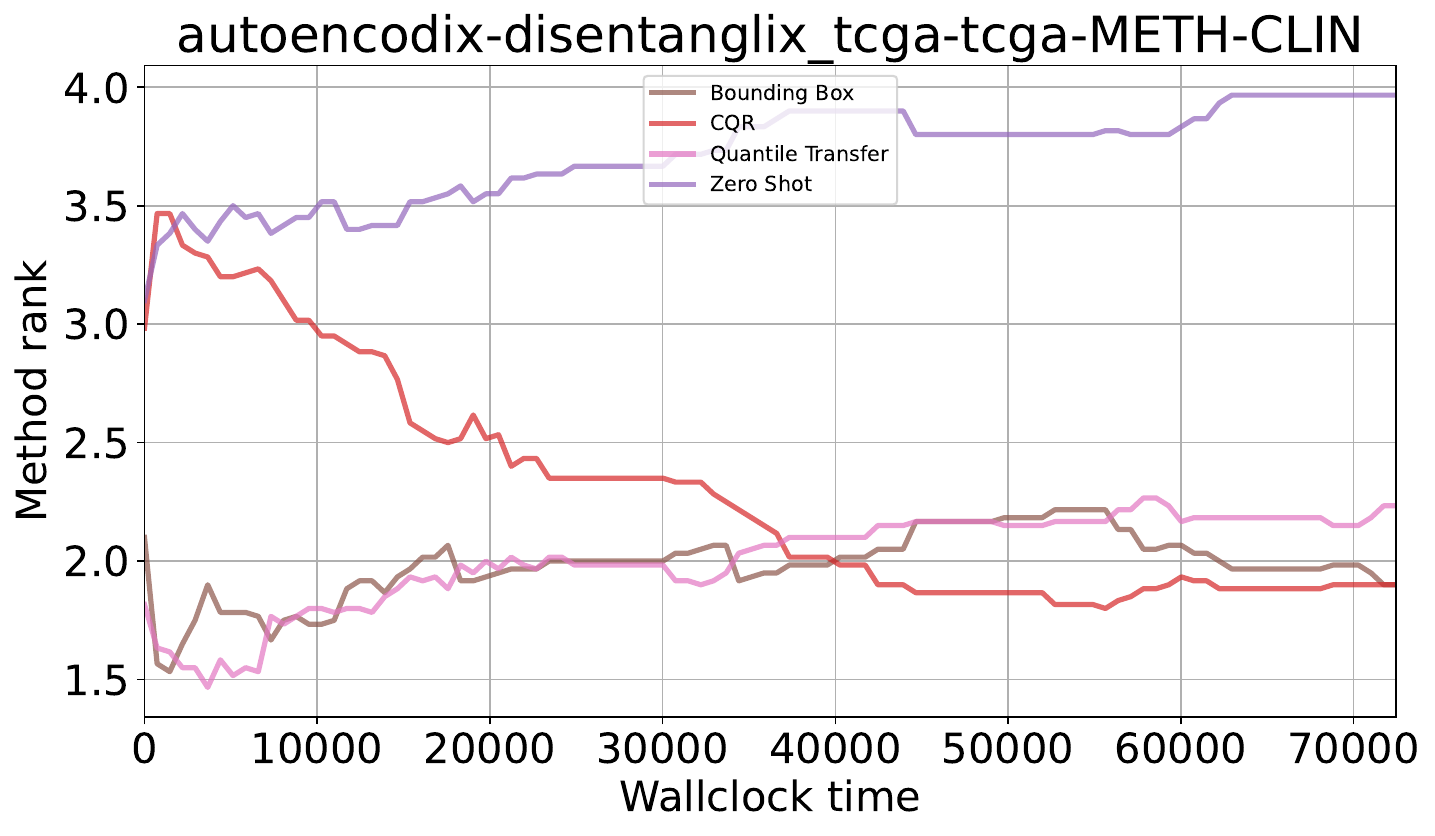} &
    \includegraphics[width=0.32\textwidth]{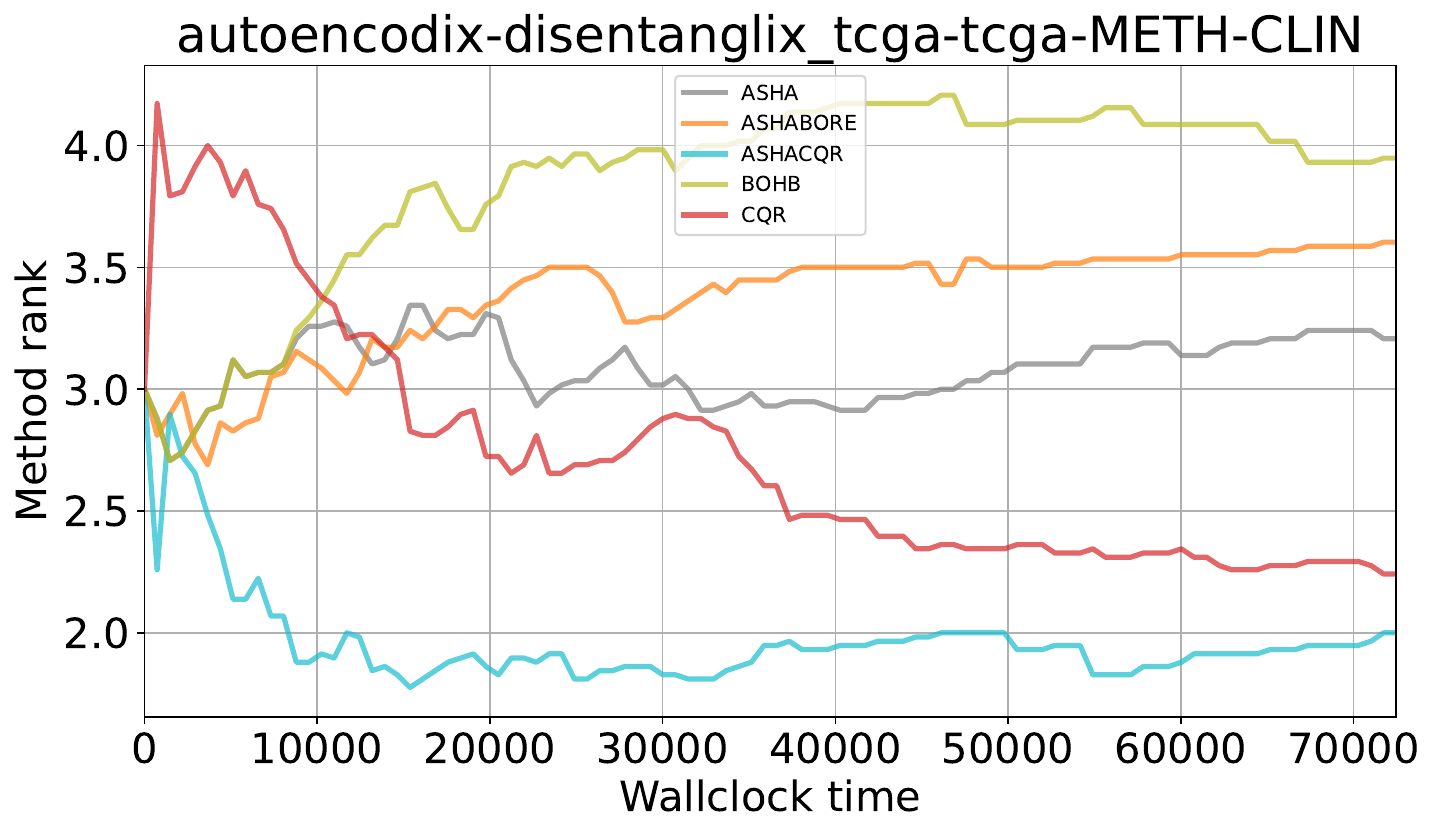} \\
    \midrule
    \multicolumn{3}{c}{\textbf{autoencodix-disentanglix\_tcga-tcga-RNA-CLIN}} \\
    \includegraphics[width=0.32\textwidth]{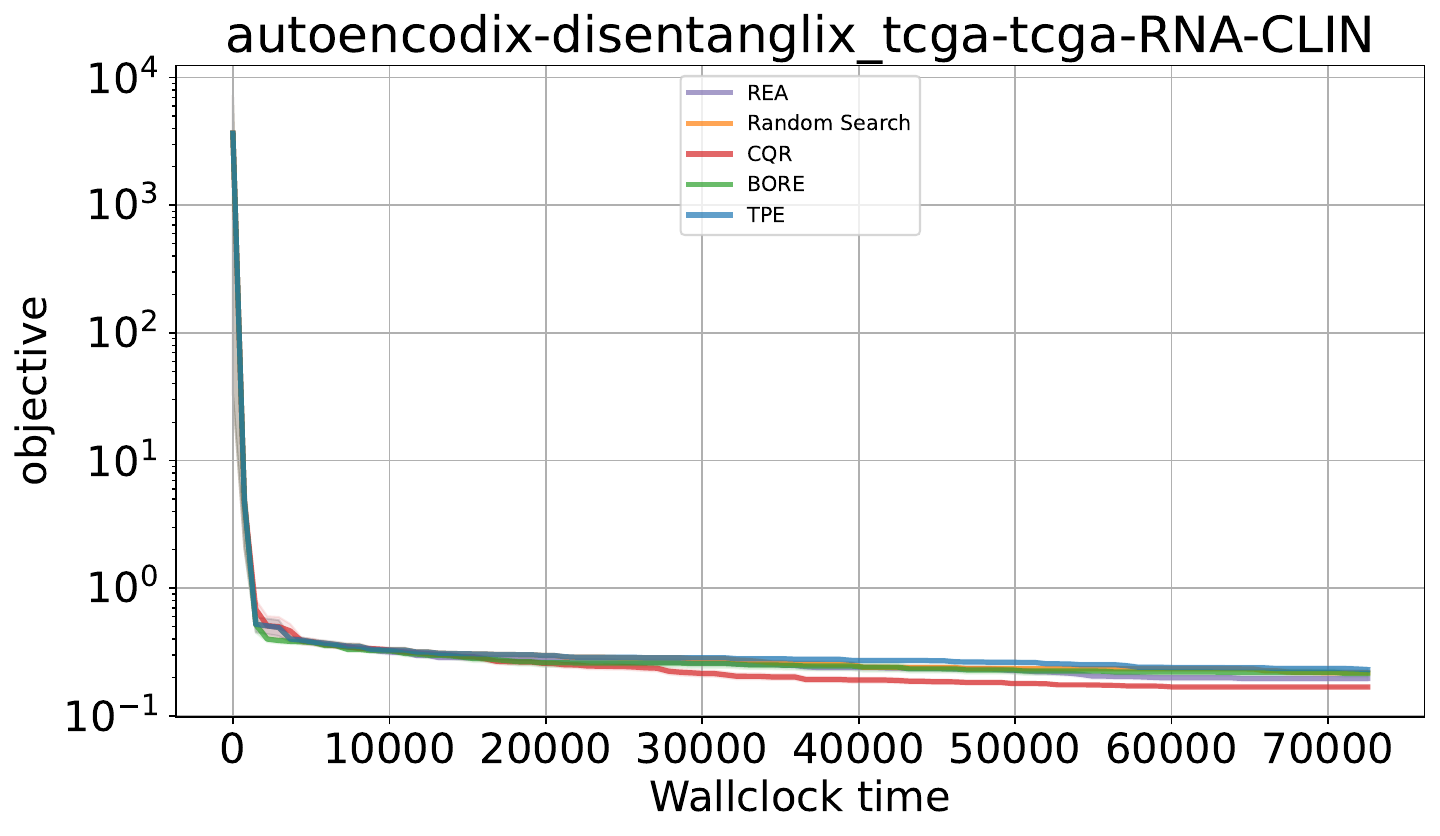} &
    \includegraphics[width=0.32\textwidth]{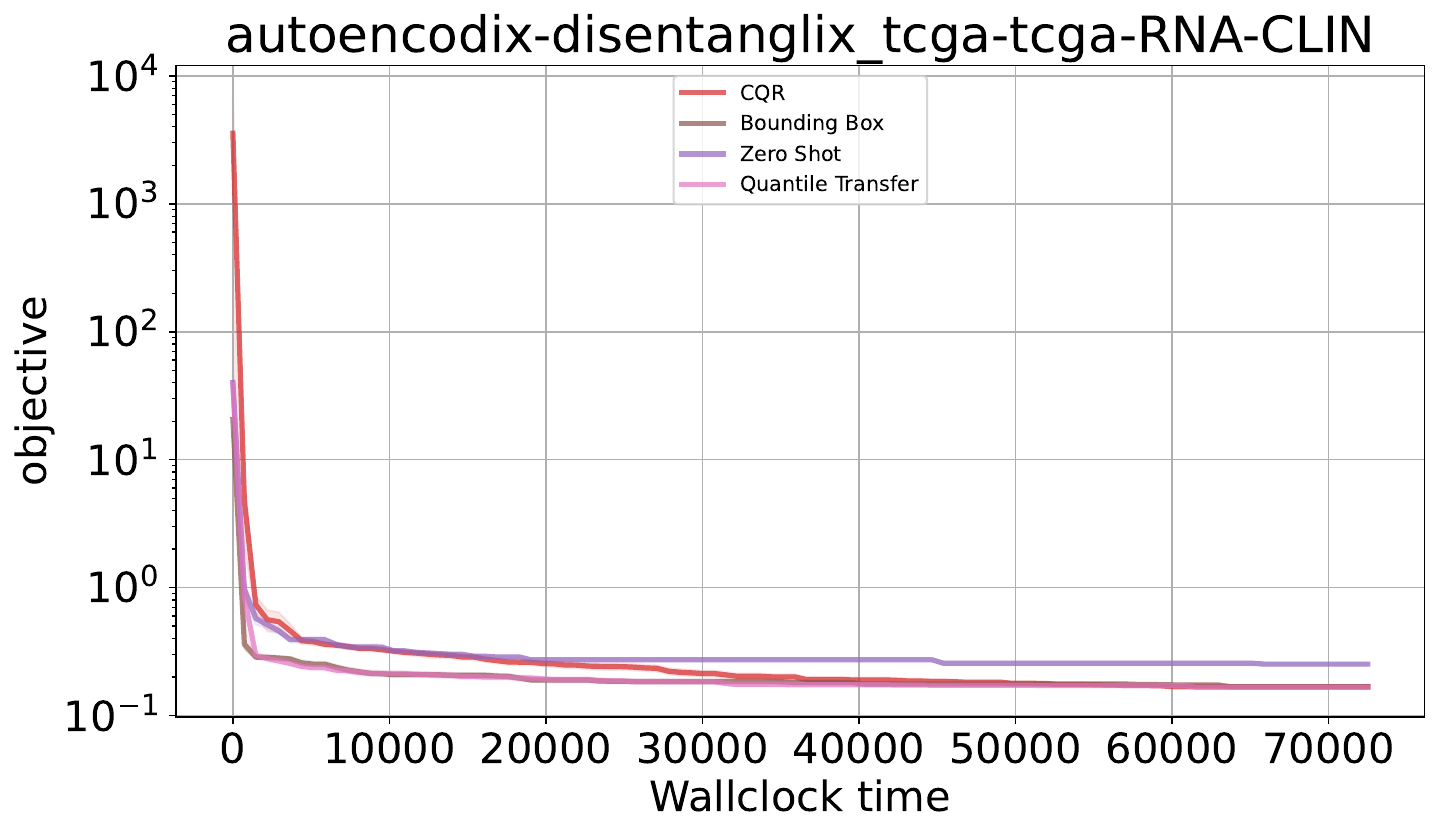} &
    \includegraphics[width=0.32\textwidth]{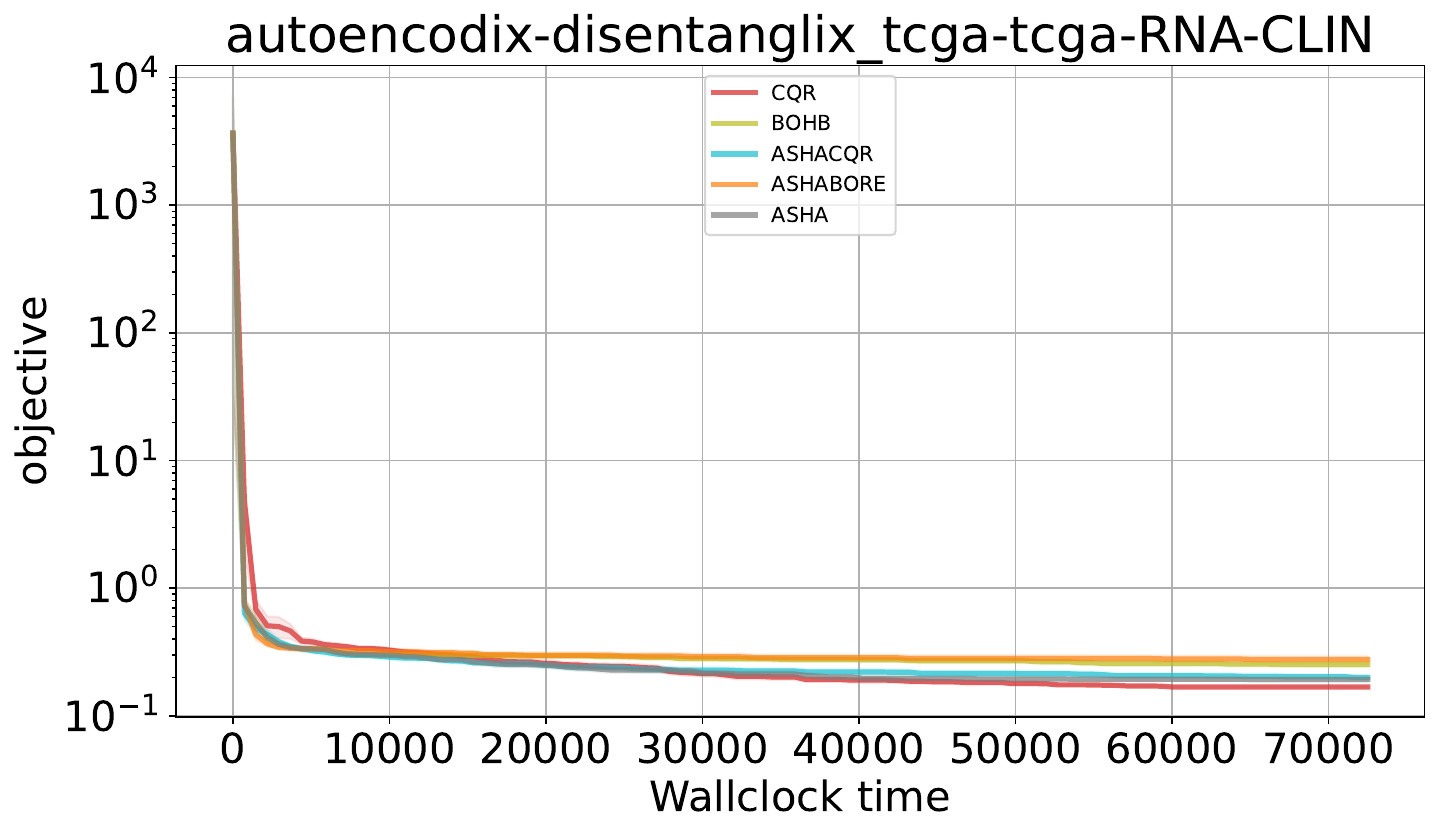} \\
    \includegraphics[width=0.32\textwidth]{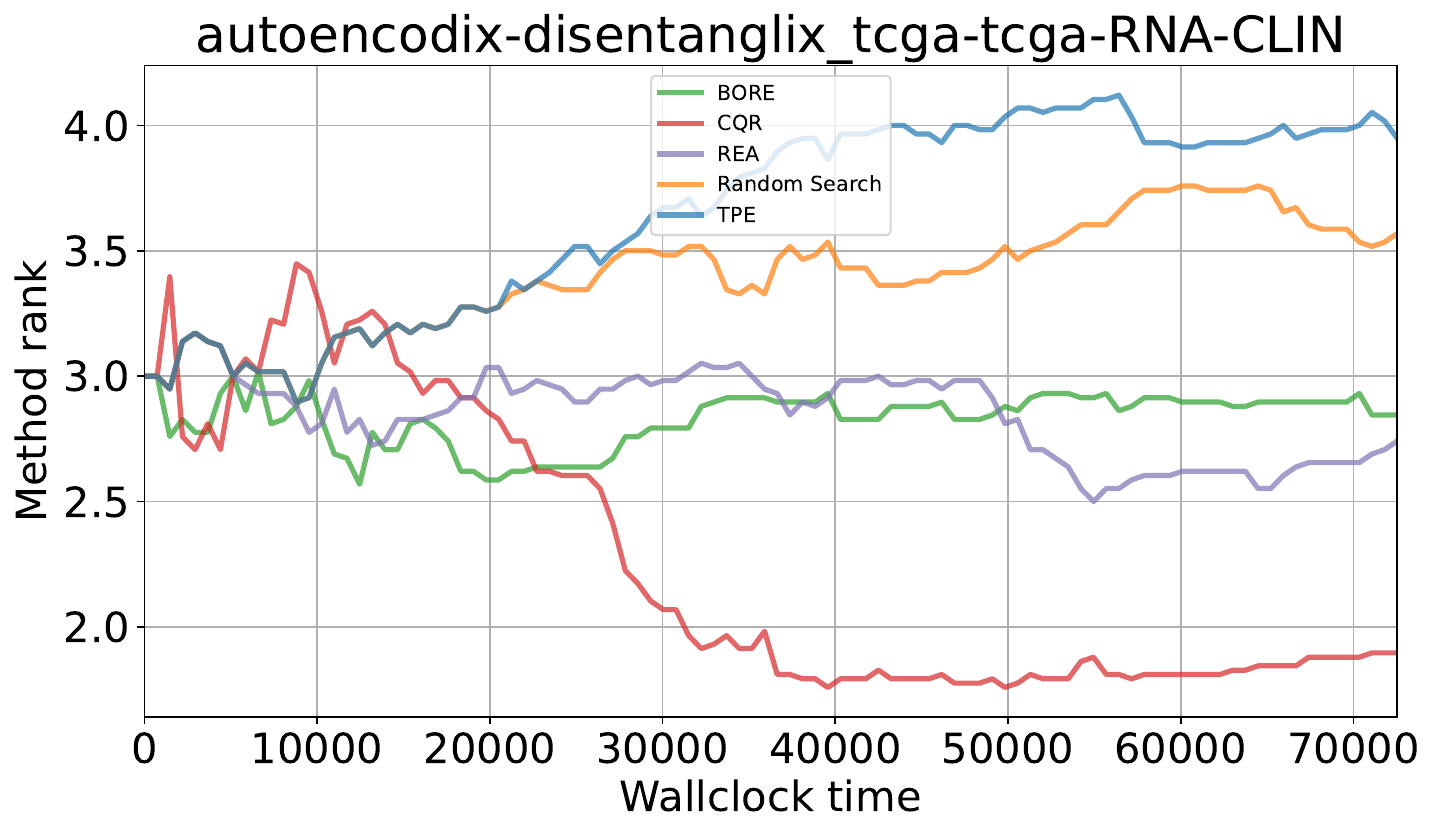} &
    \includegraphics[width=0.32\textwidth]{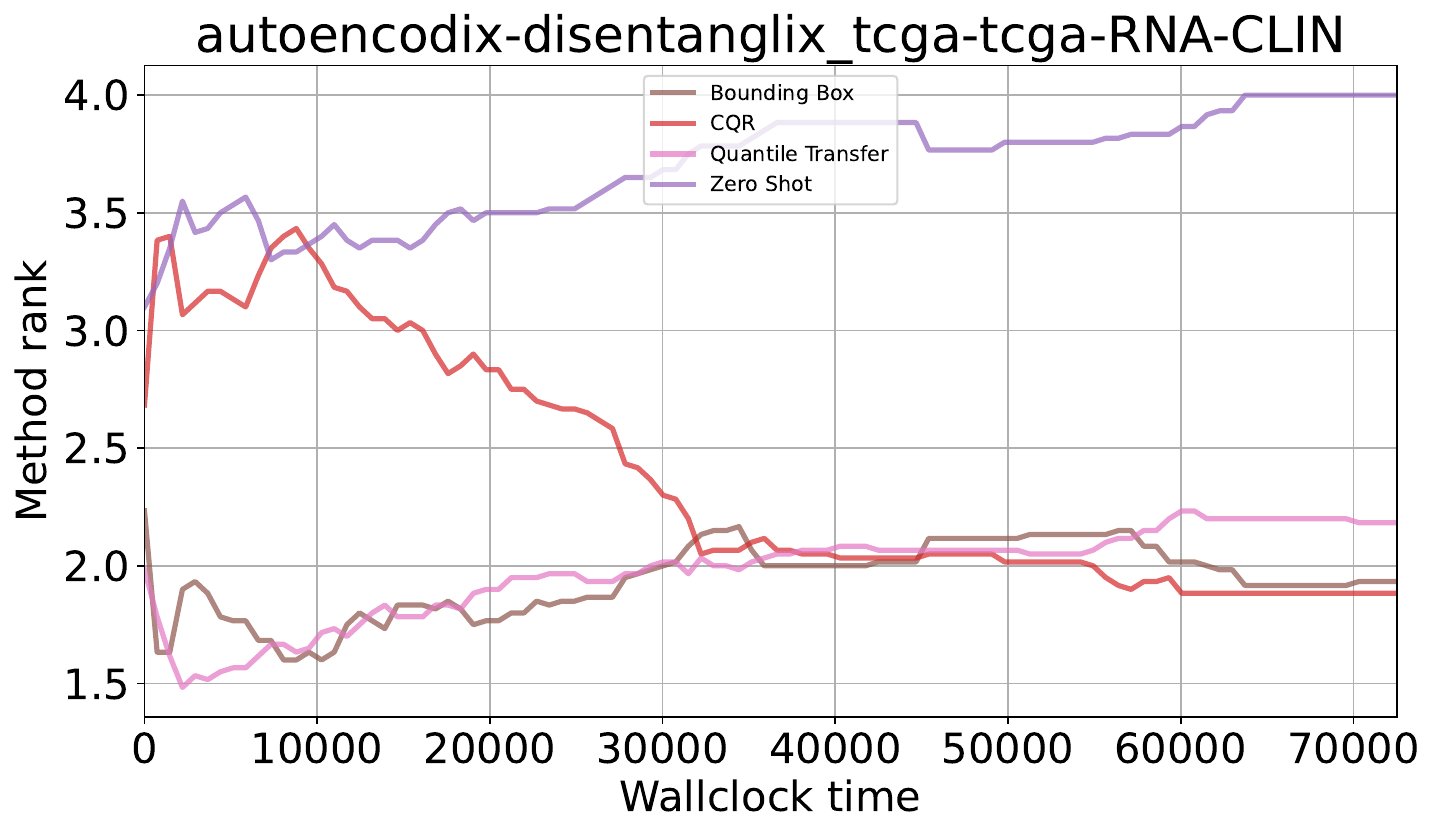} &
    \includegraphics[width=0.32\textwidth]{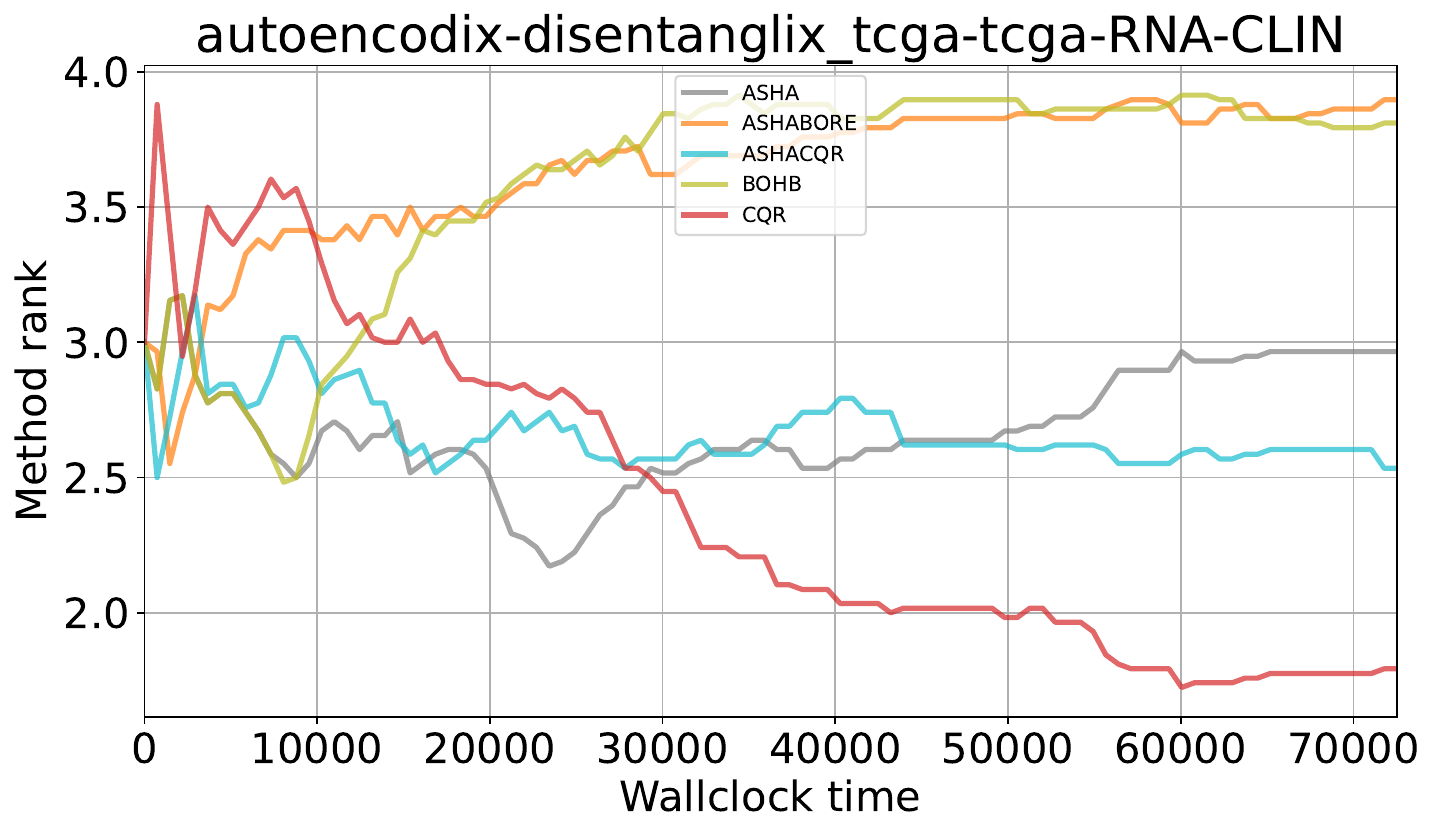} \\
    \end{tabular}
    \caption{Results for Disentanglix tasks (Part 2).}
    \label{fig:disentanglix_part2}
\end{figure}

\clearpage

\begin{figure}[htbp]
    \centering
    \setlength{\tabcolsep}{1pt}
    \begin{tabular}{ccc}
    \multicolumn{3}{c}{\textbf{autoencodix-disentanglix\_tcga-tcga-RNA-DNA-METH-CLIN}} \\
    \textbf{Single-Fidelity} & \textbf{Transfer Learning} & \textbf{Multi-Fidelity} \\
    \includegraphics[width=0.32\textwidth]{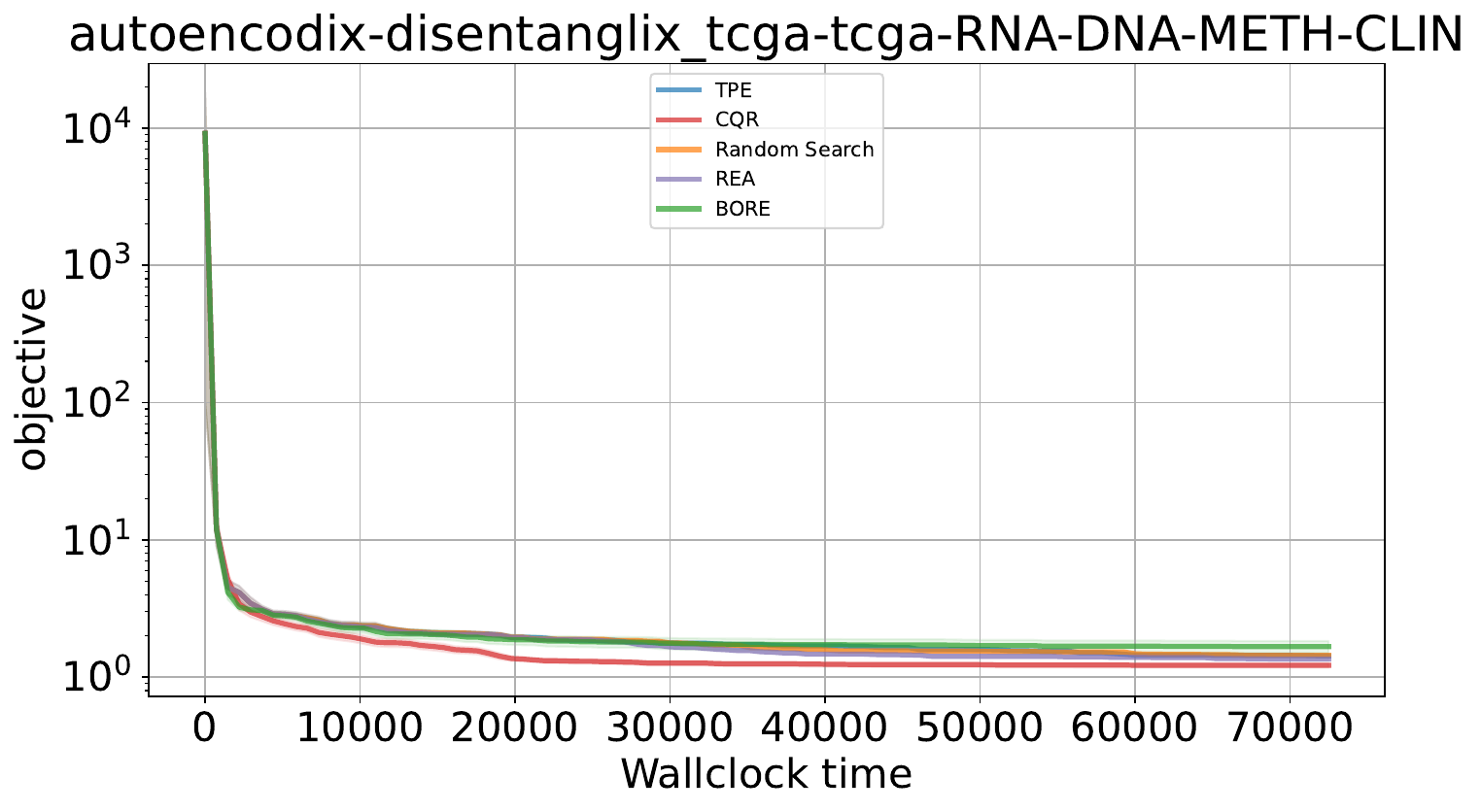} &
    \includegraphics[width=0.32\textwidth]{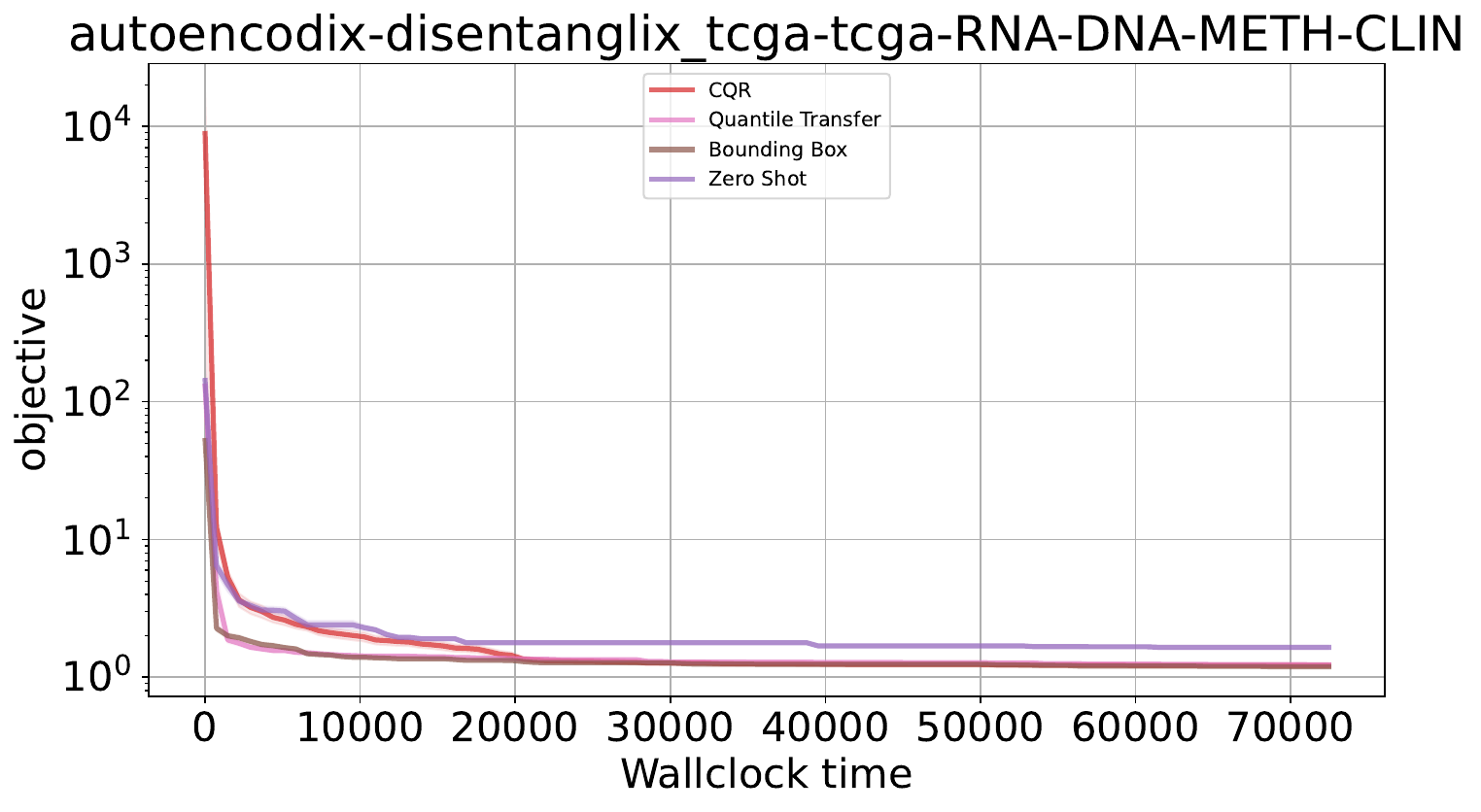} &
    \includegraphics[width=0.32\textwidth]{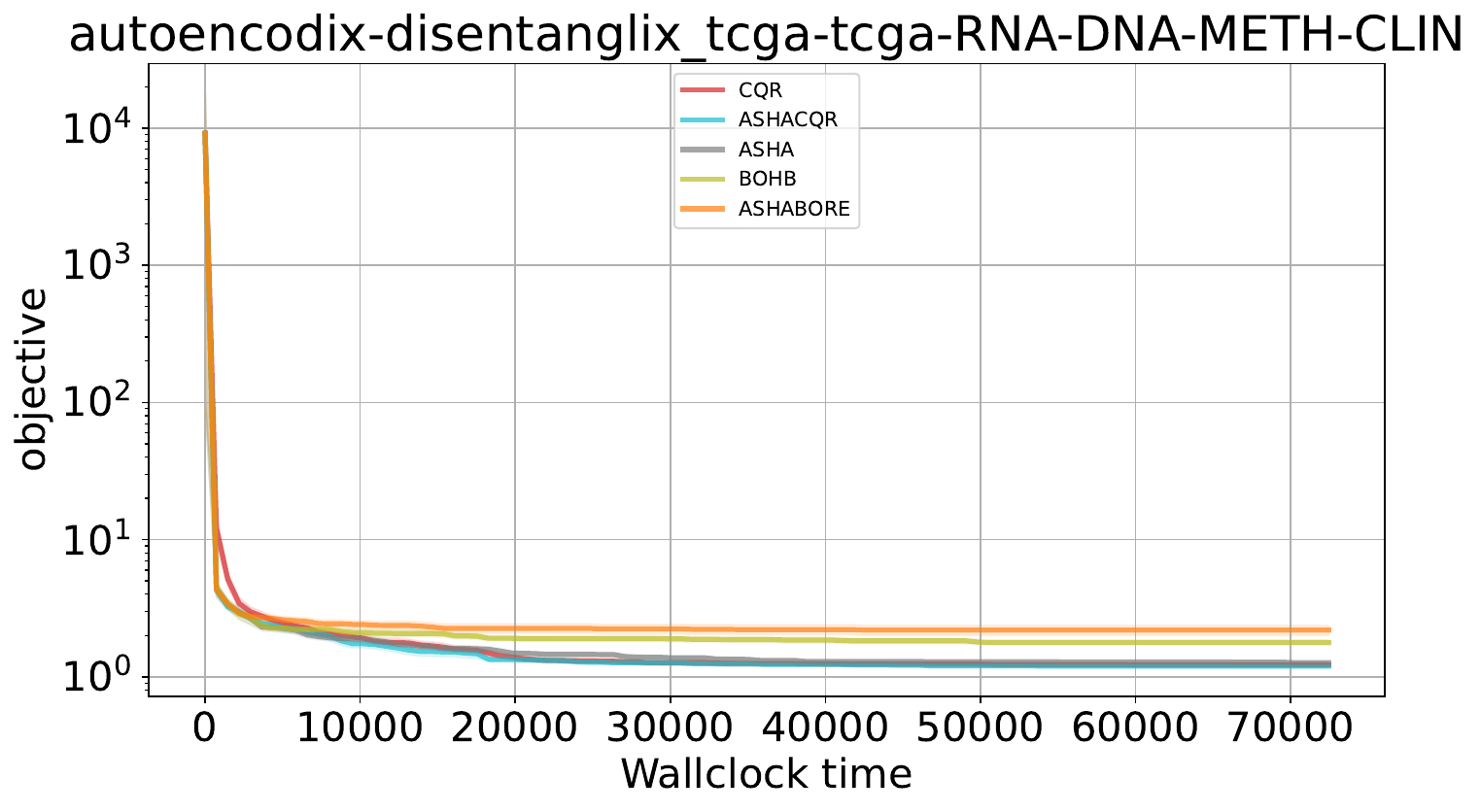} \\
    \includegraphics[width=0.32\textwidth]{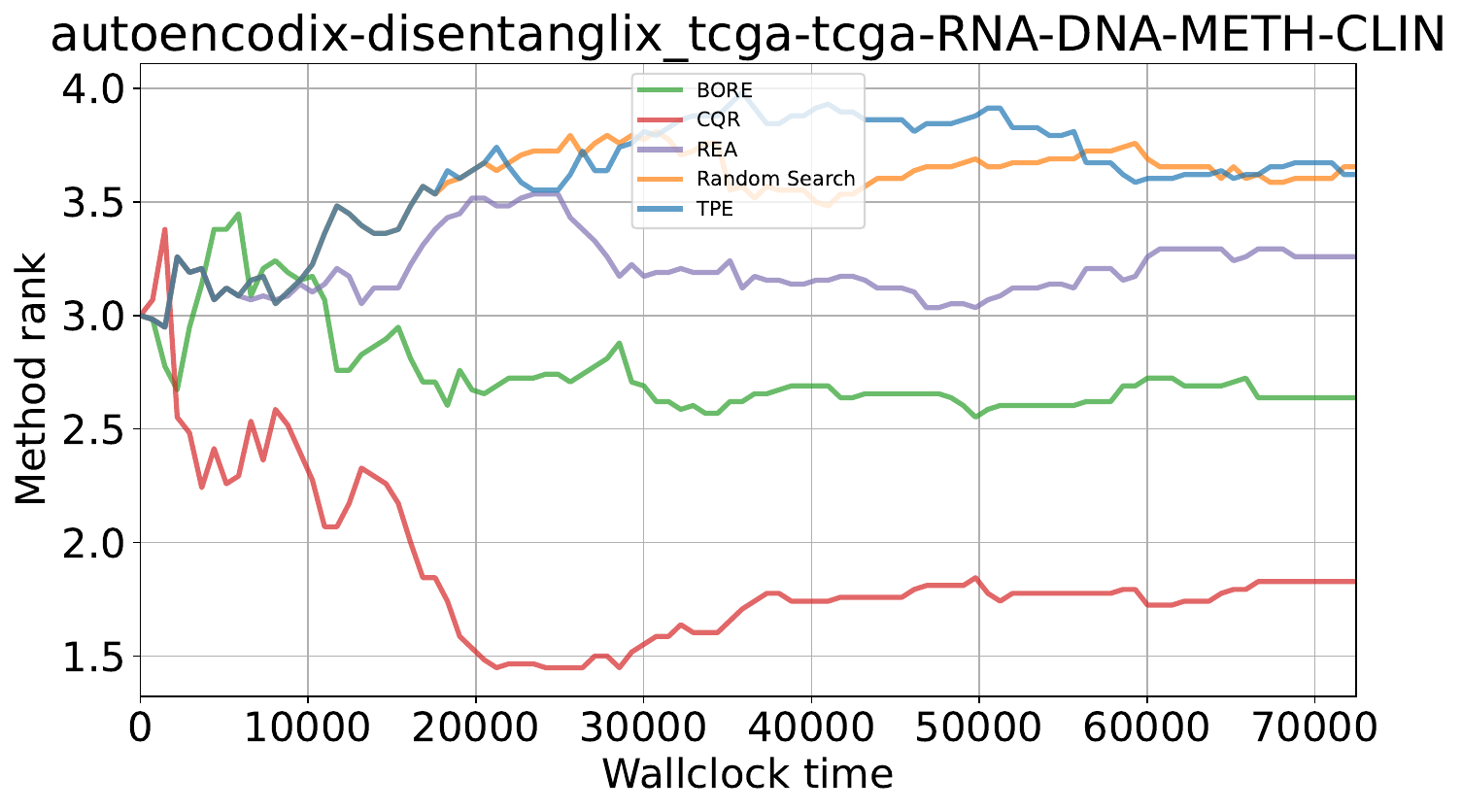} &
    \includegraphics[width=0.32\textwidth]{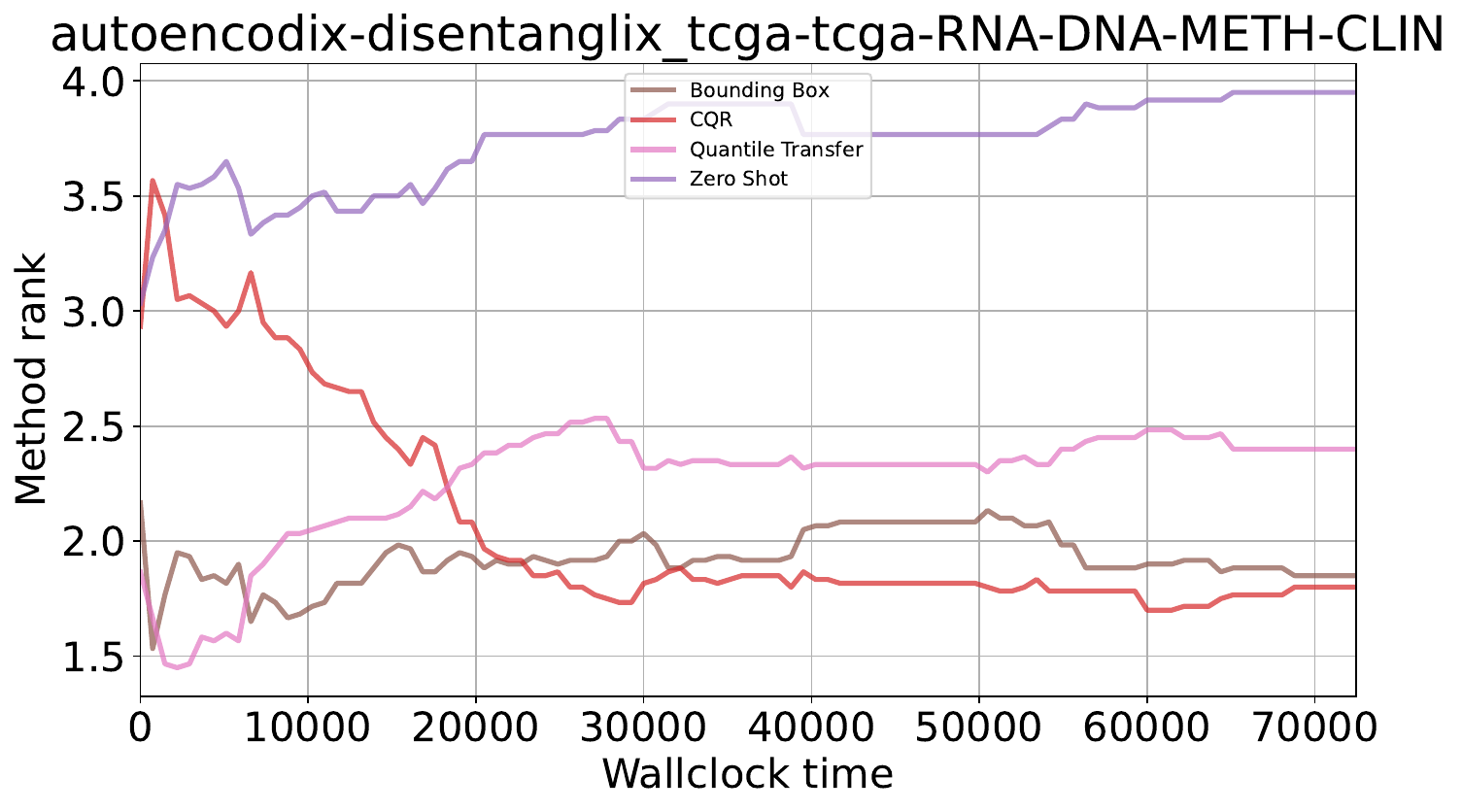} &
    \includegraphics[width=0.32\textwidth]{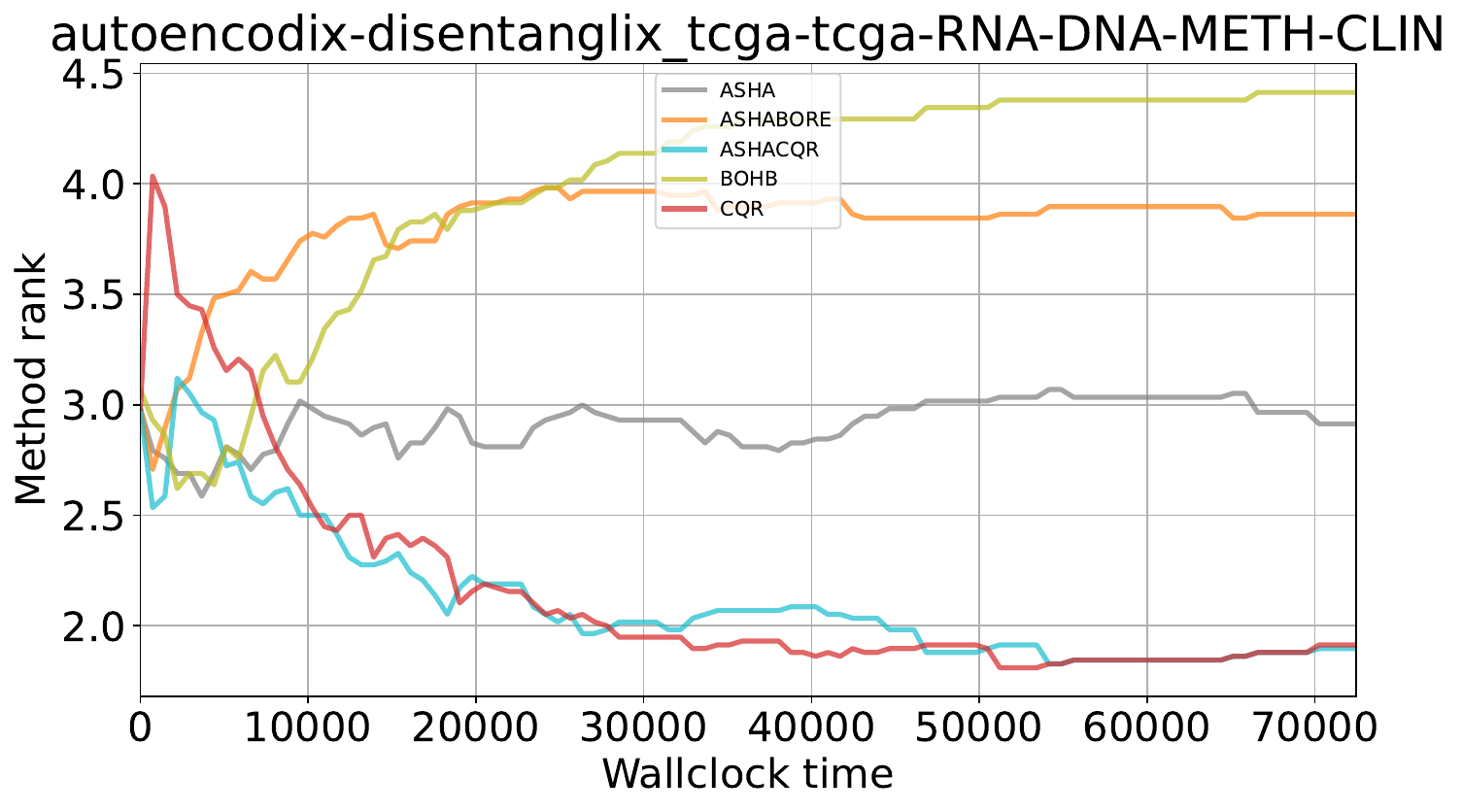} \\
    \end{tabular}
    \caption{Results for Disentanglix (autoencodix-disentanglix\_tcga-tcga-RNA-DNA-METH-CLIN).}
    \label{fig:disentanglix_part3}
\end{figure}

\clearpage

% Model: Ontix
\begin{figure}[htbp]
    \centering
    \setlength{\tabcolsep}{1pt}
    \begin{tabular}{ccc}
    \multicolumn{3}{c}{\textbf{autoencodix-ontix\_schc-schc-METH-CLIN-chromosome}} \\
    \textbf{Single-Fidelity} & \textbf{Transfer Learning} & \textbf{Multi-Fidelity} \\
    \includegraphics[width=0.32\textwidth]{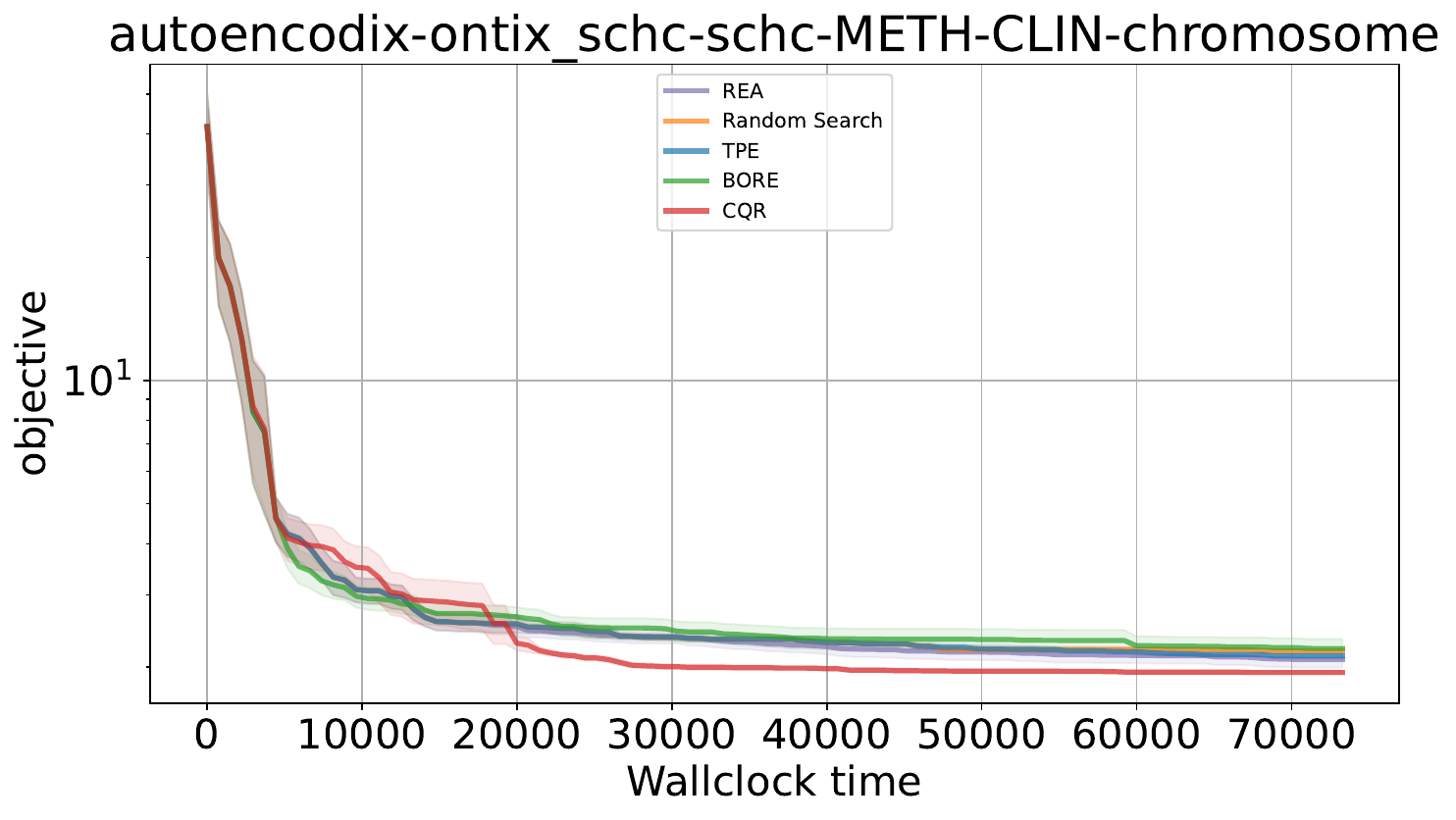} &
    \includegraphics[width=0.32\textwidth]{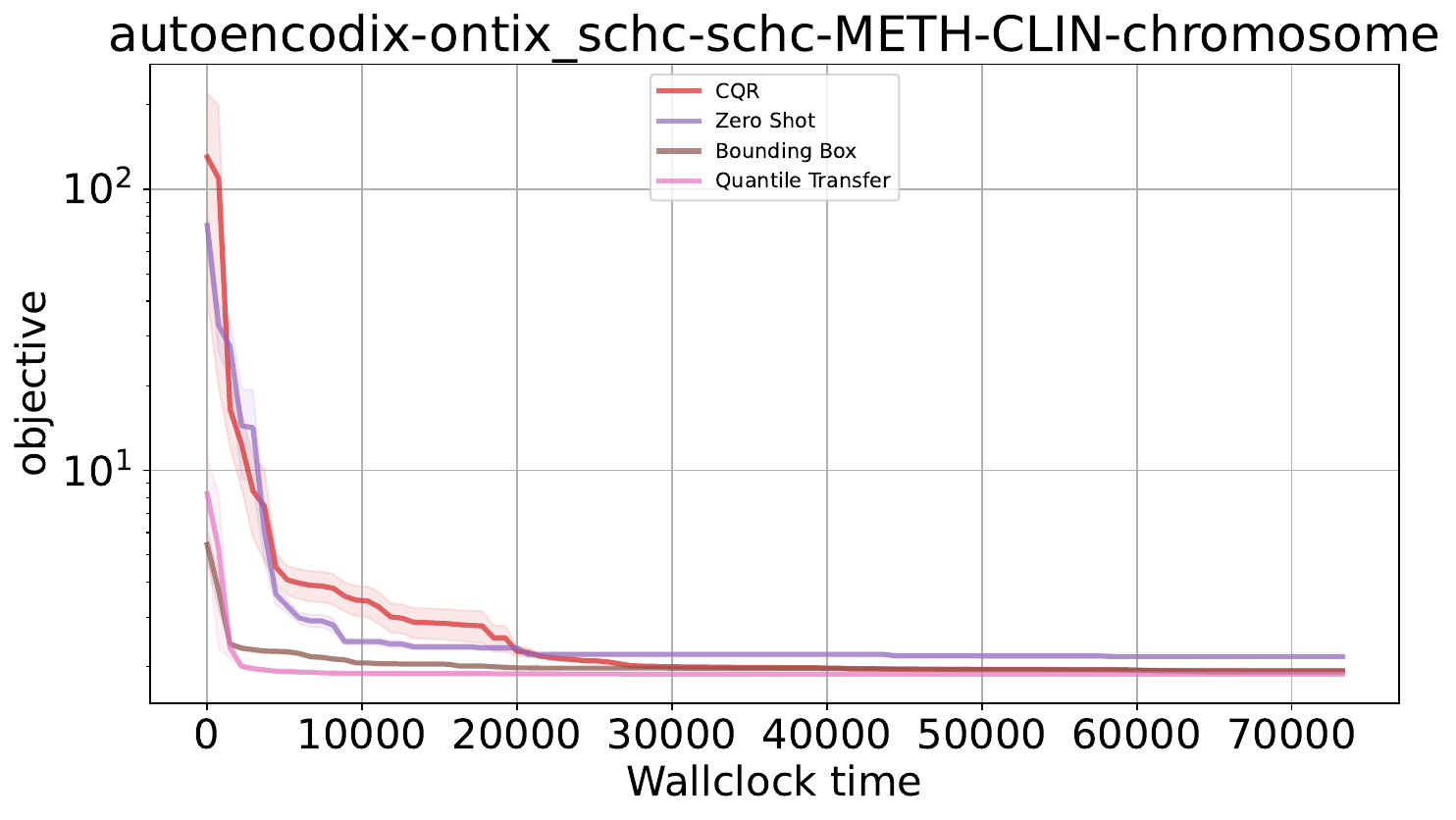} &
    \includegraphics[width=0.32\textwidth]{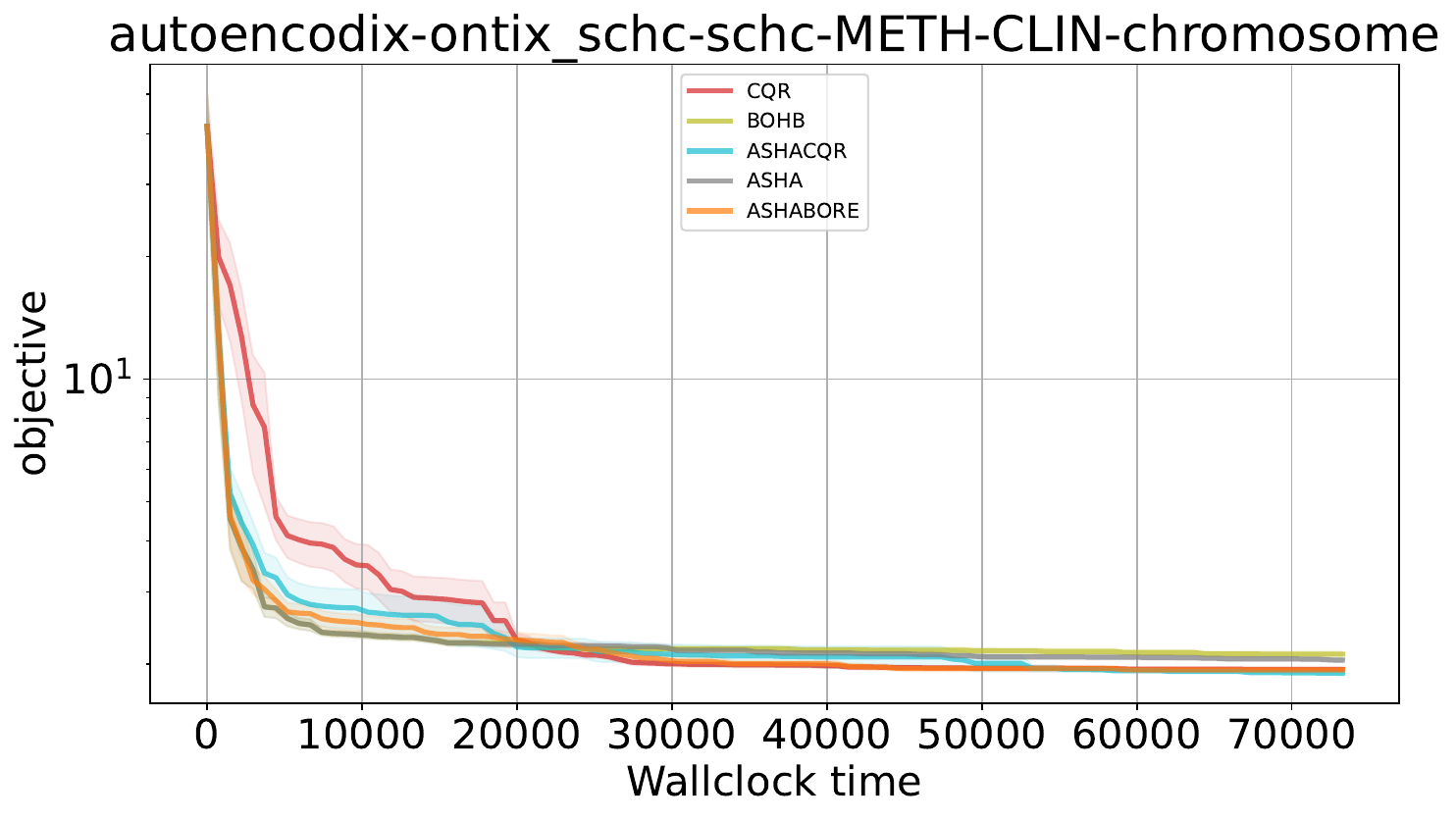} \\
    \includegraphics[width=0.32\textwidth]{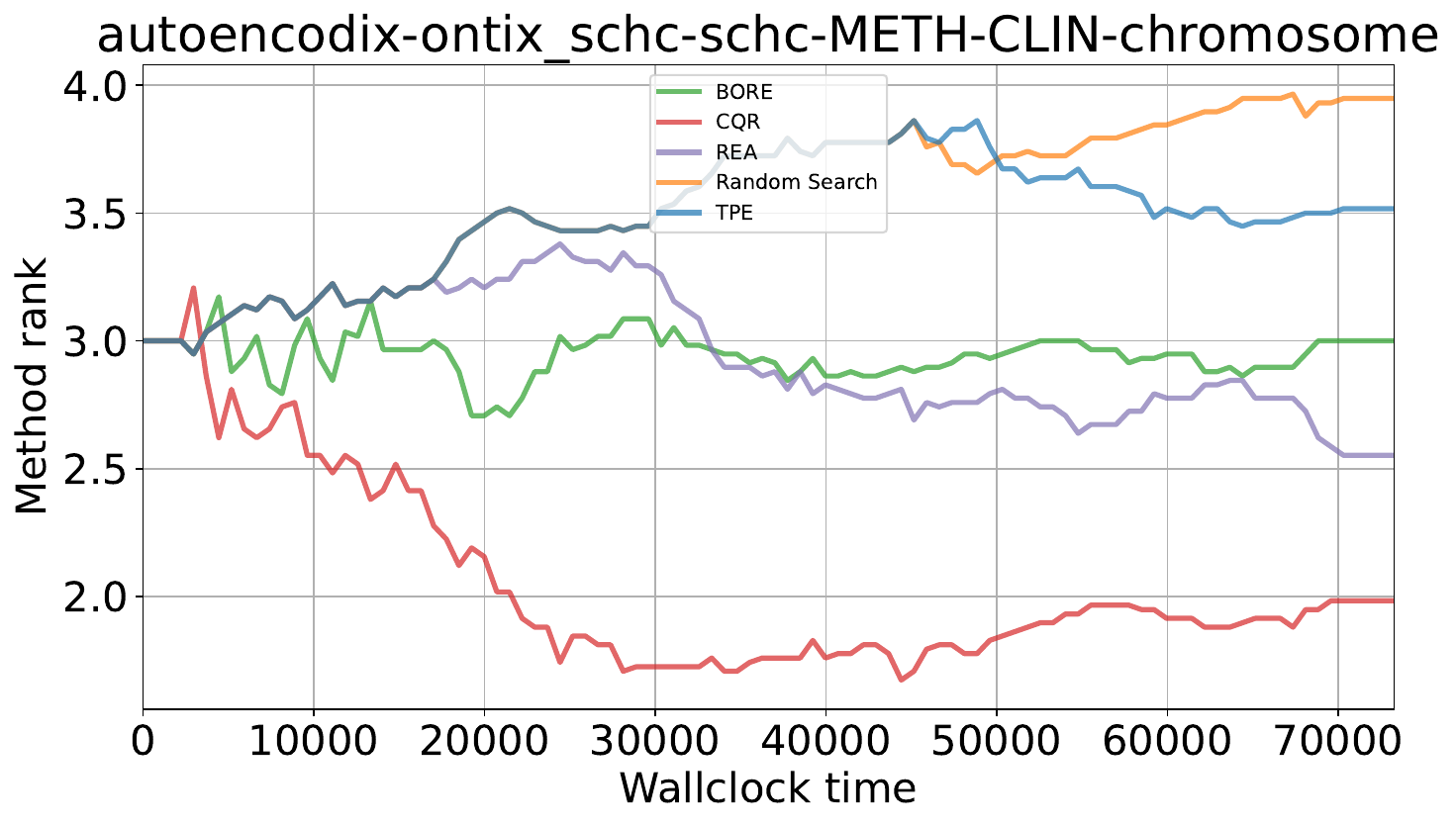} &
    \includegraphics[width=0.32\textwidth]{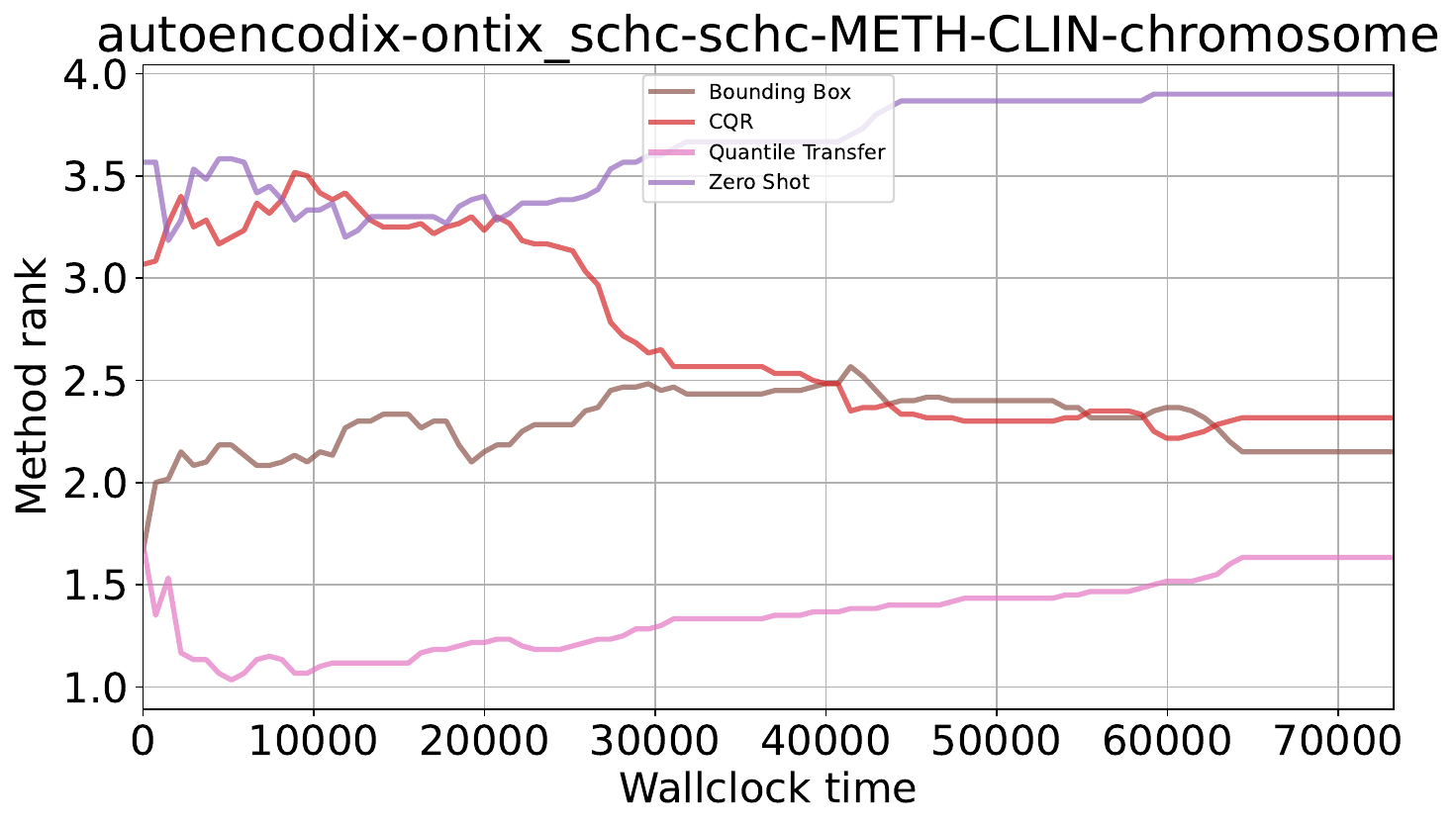} &
    \includegraphics[width=0.32\textwidth]{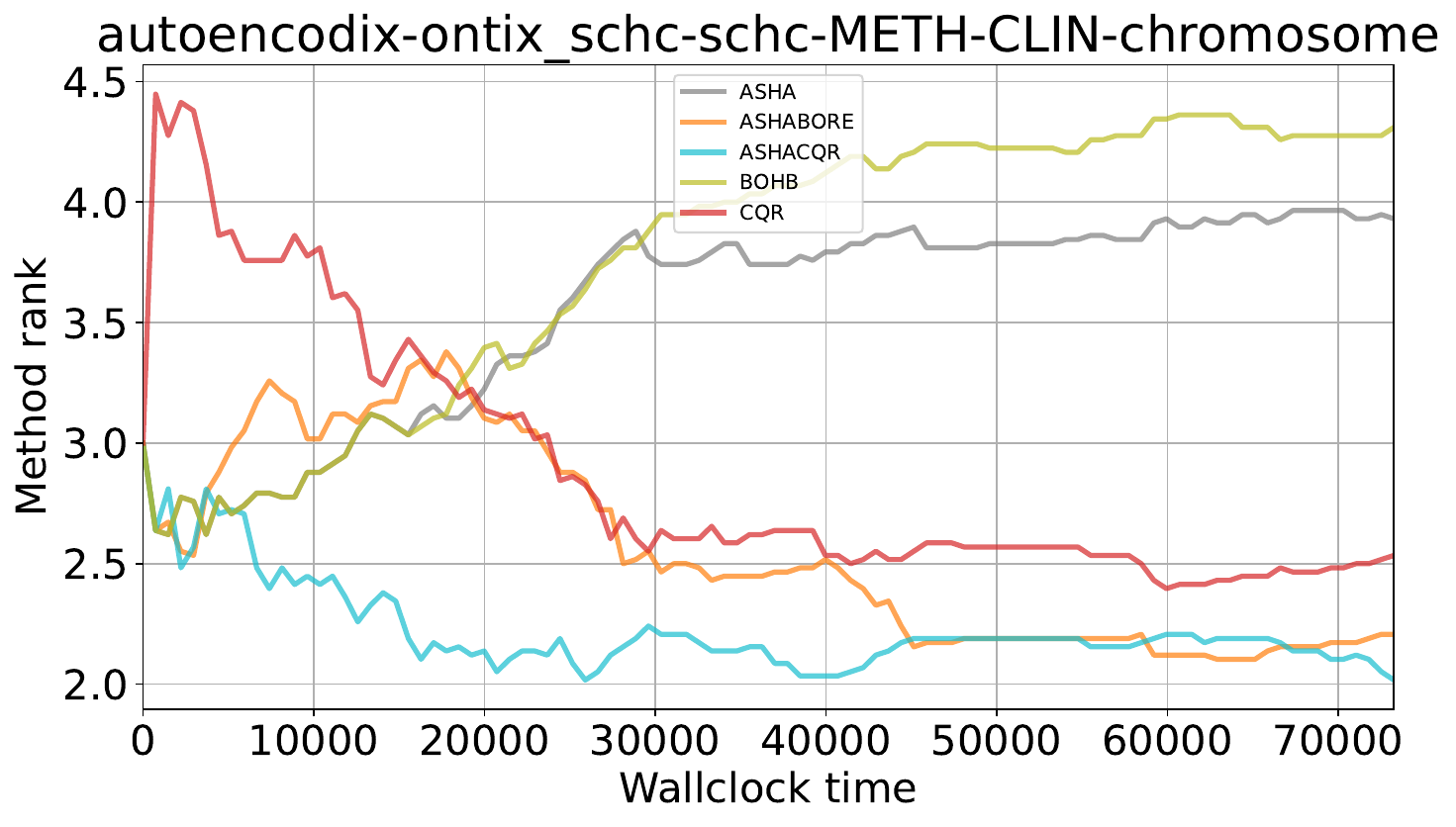} \\
    \midrule
    \multicolumn{3}{c}{\textbf{autoencodix-ontix\_schc-schc-METH-CLIN-reactome}} \\
    \includegraphics[width=0.32\textwidth]{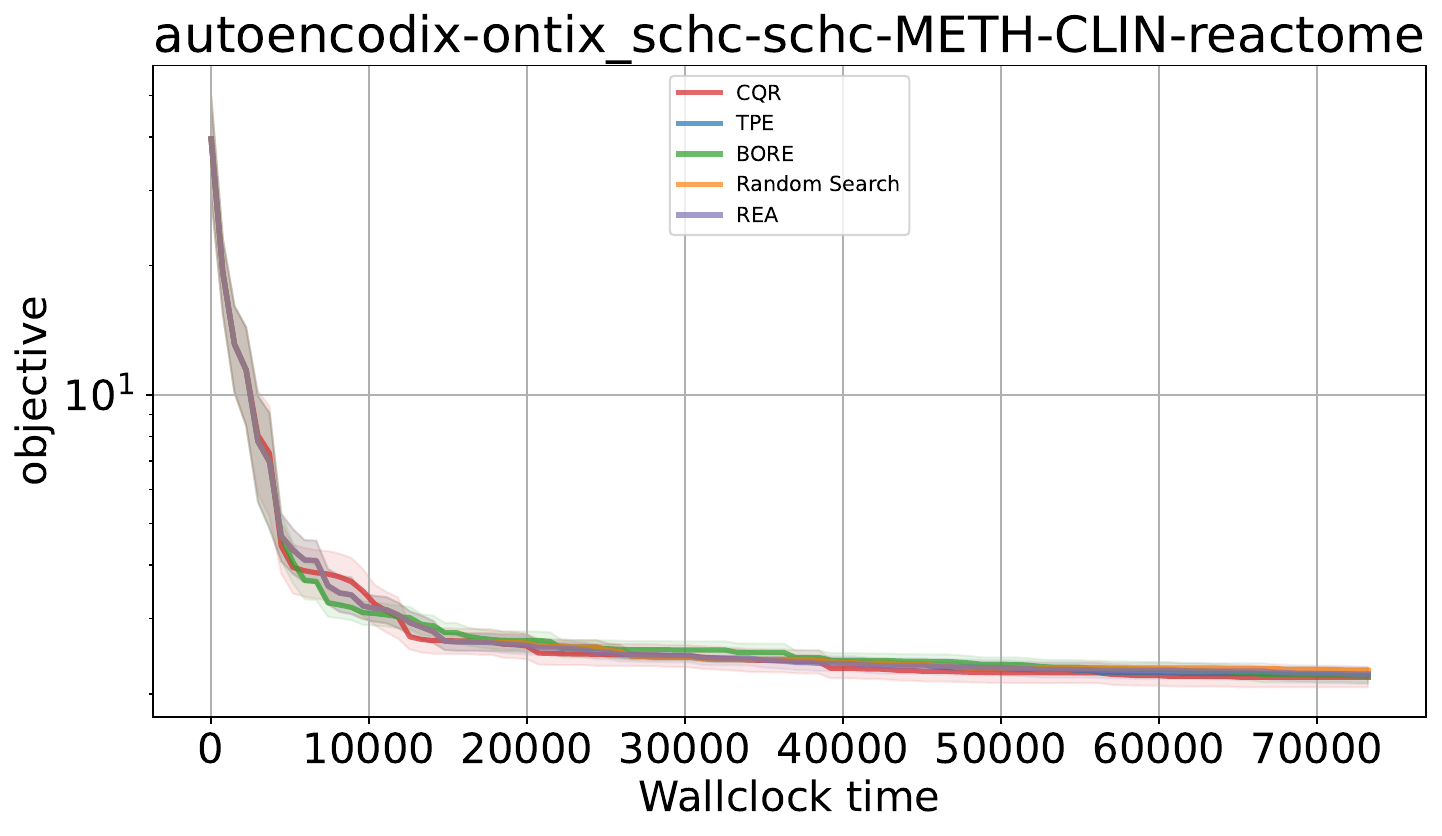} &
    \includegraphics[width=0.32\textwidth]{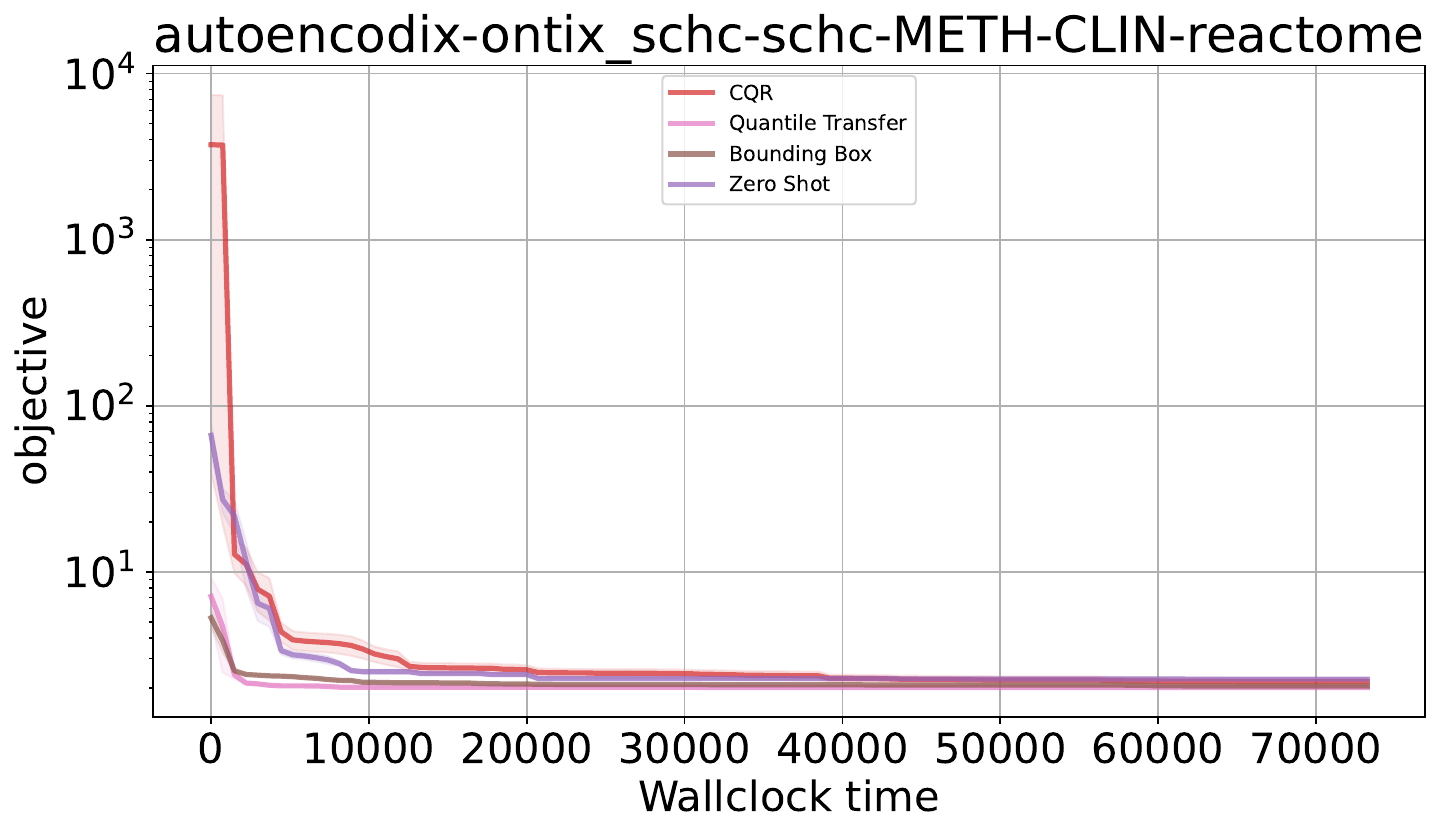} &
    \includegraphics[width=0.32\textwidth]{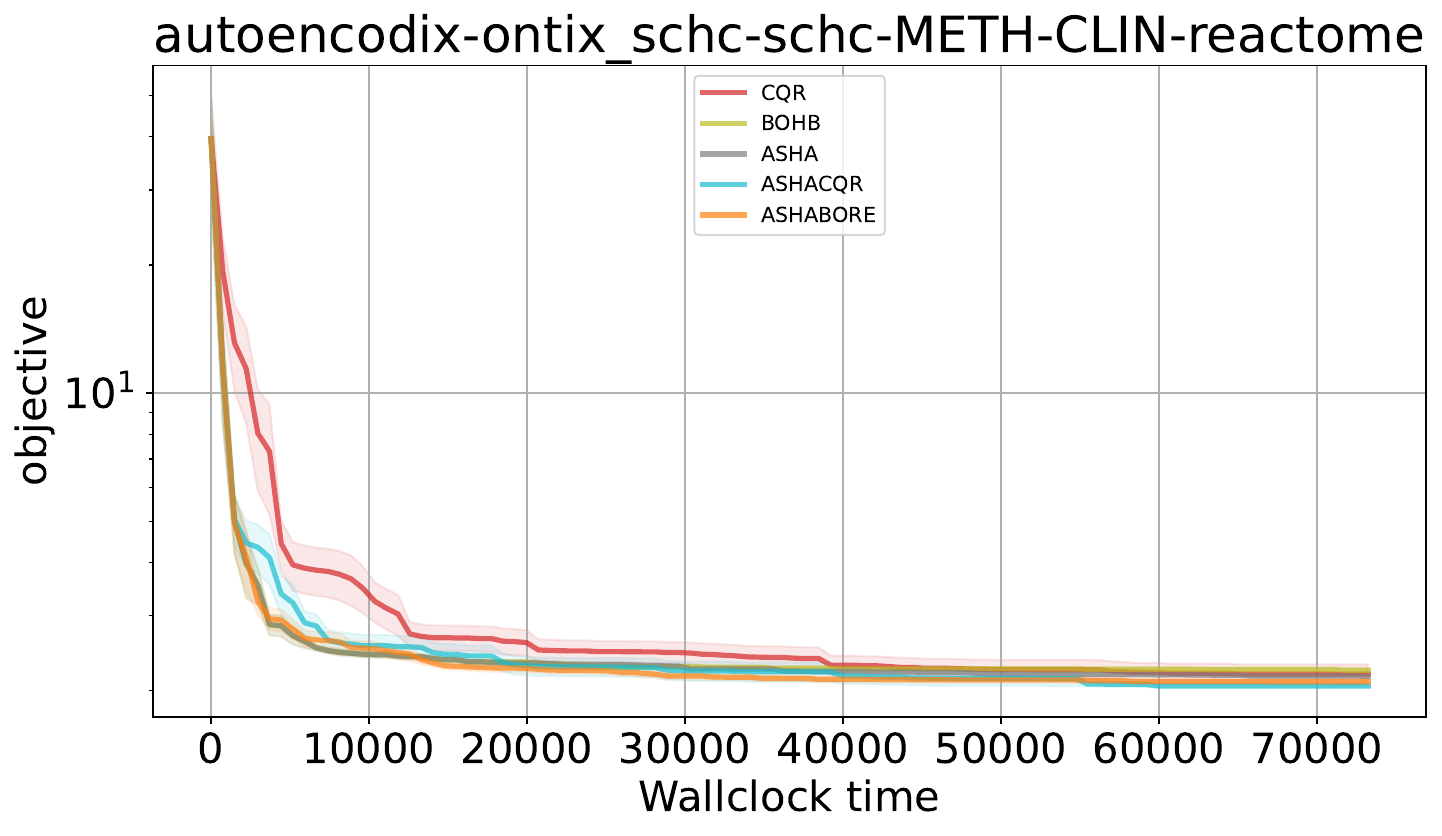} \\
    \includegraphics[width=0.32\textwidth]{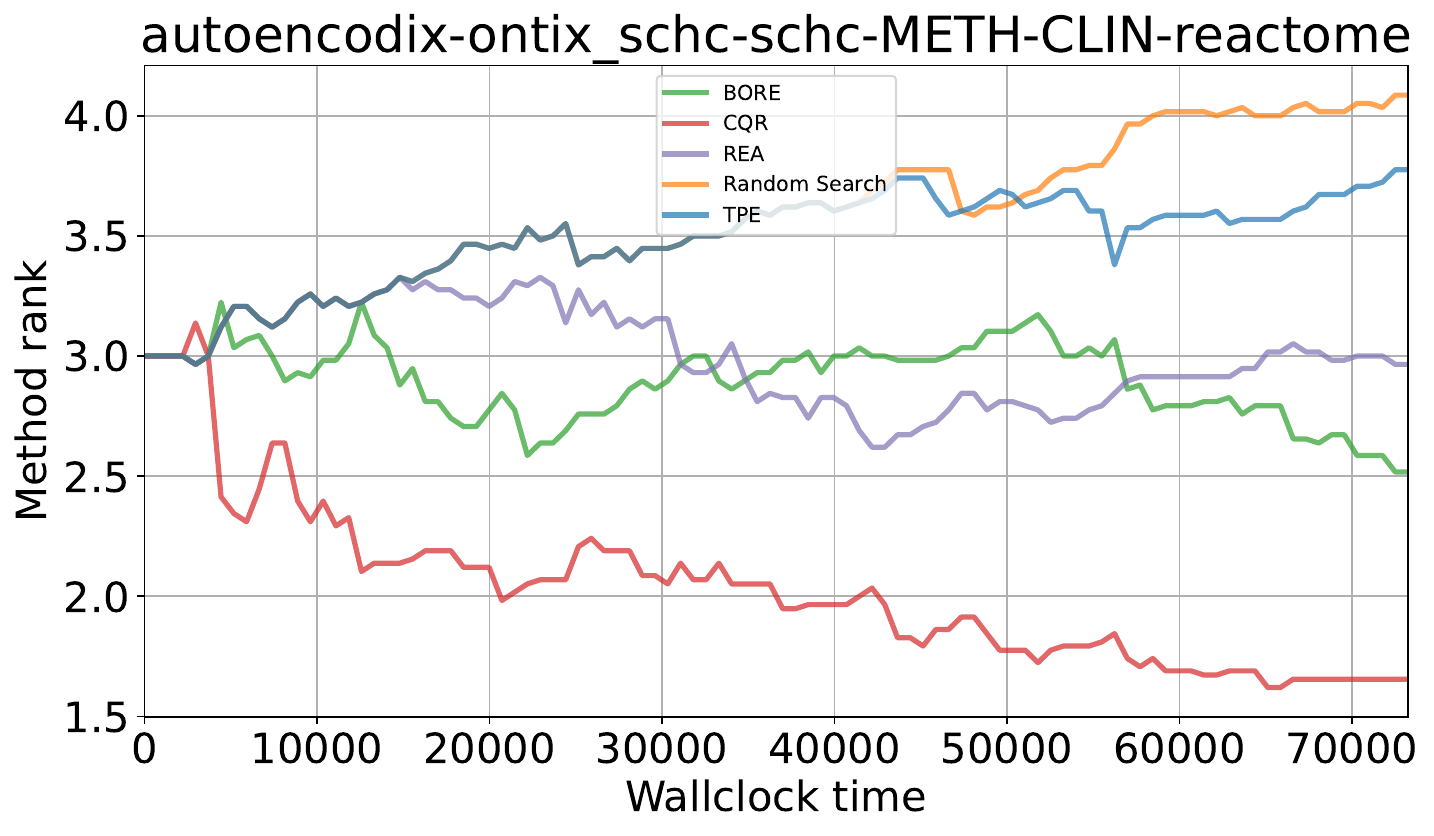} &
    \includegraphics[width=0.32\textwidth]{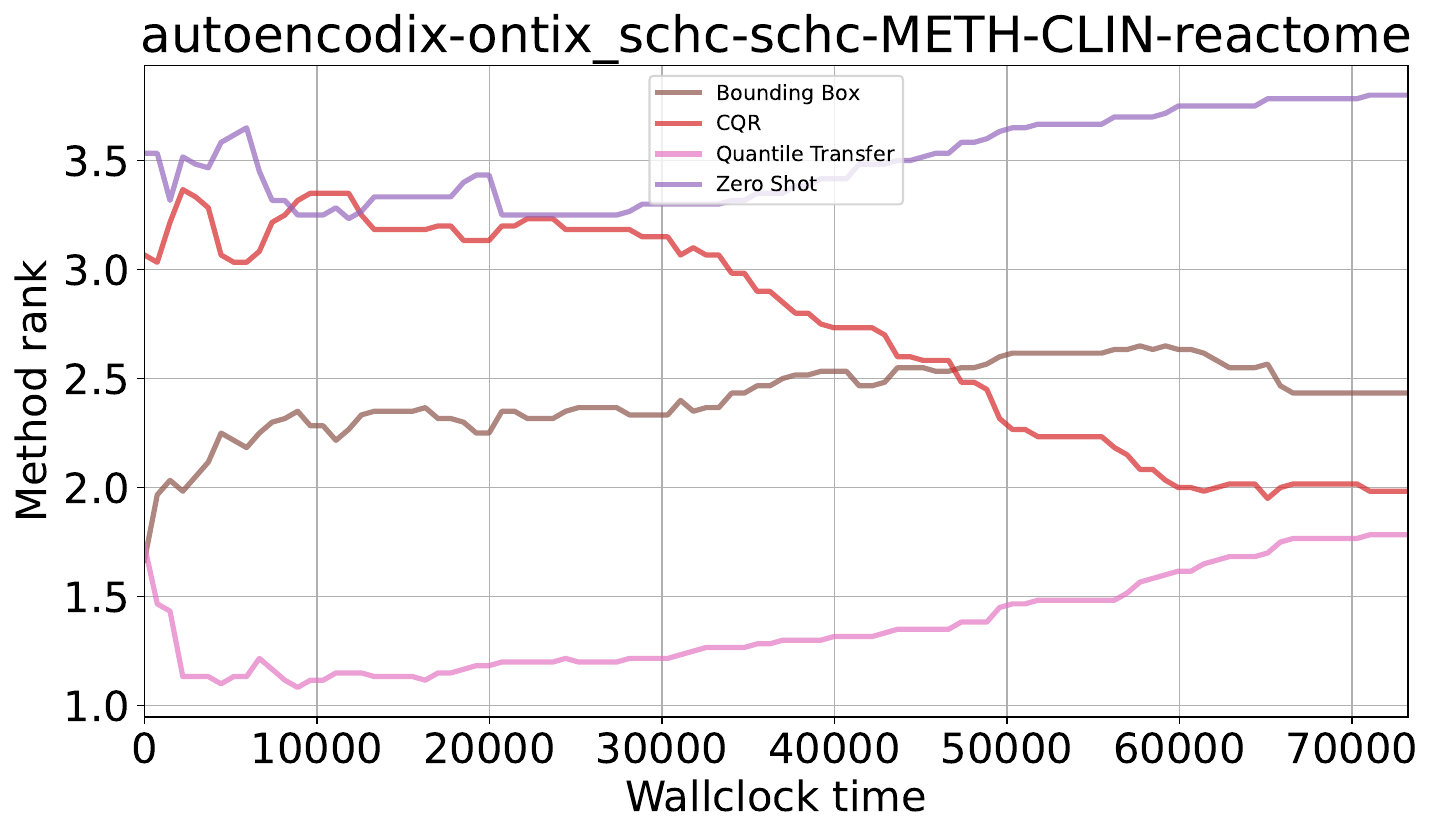} &
    \includegraphics[width=0.32\textwidth]{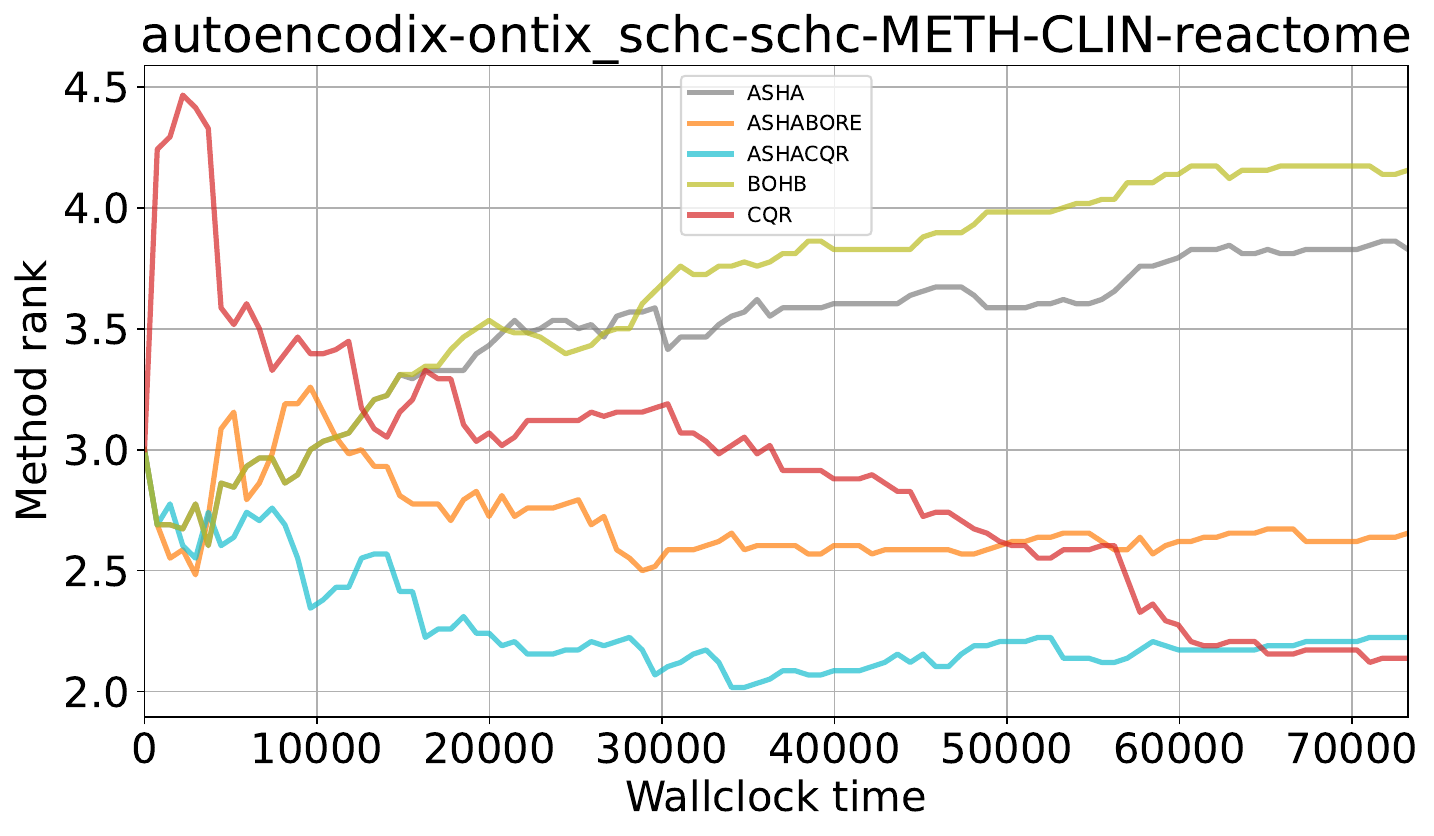} \\
    \midrule
    \multicolumn{3}{c}{\textbf{autoencodix-ontix\_schc-schc-RNA-CLIN-chromosome}} \\
    \includegraphics[width=0.32\textwidth]{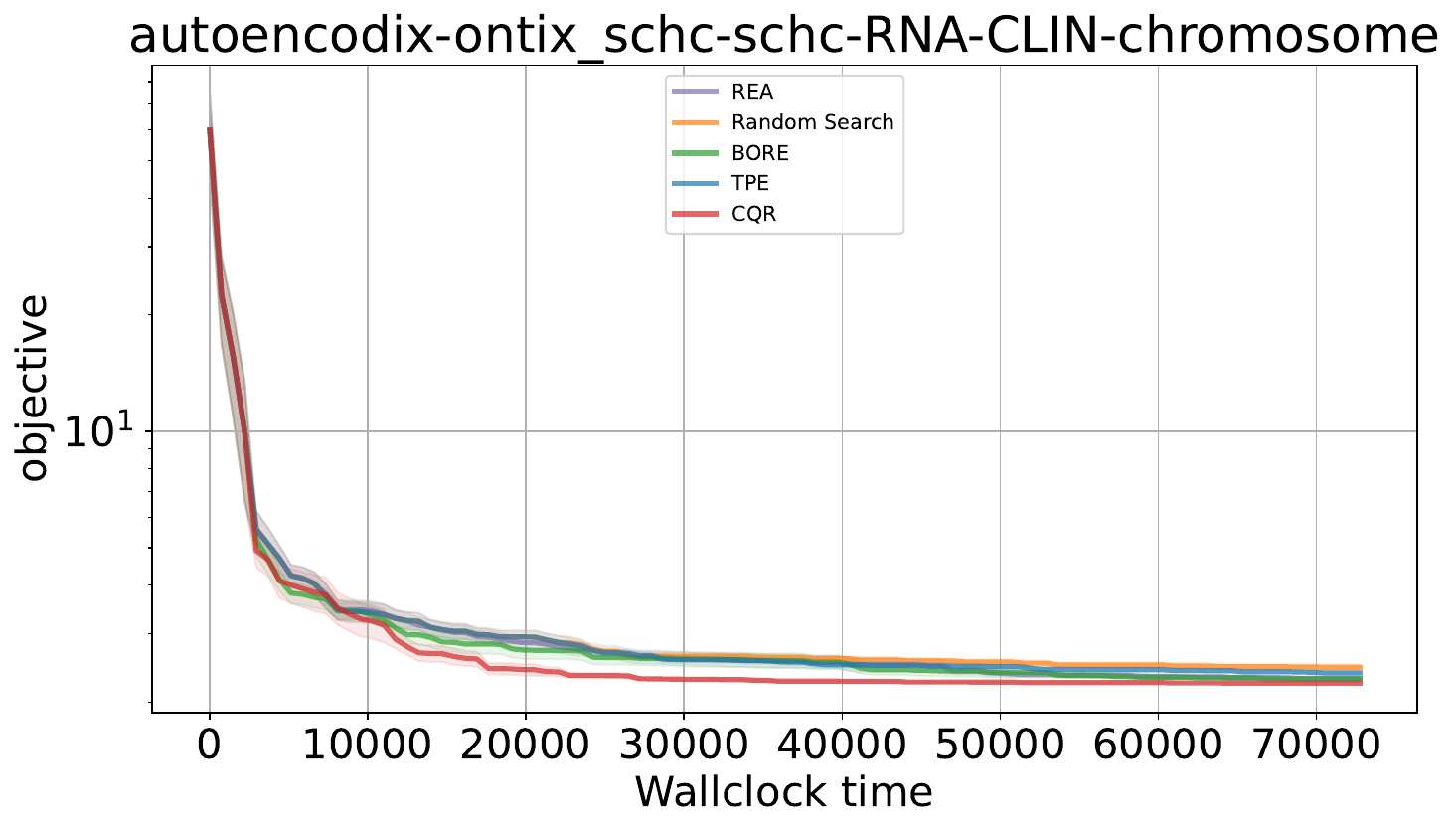} &
    \includegraphics[width=0.32\textwidth]{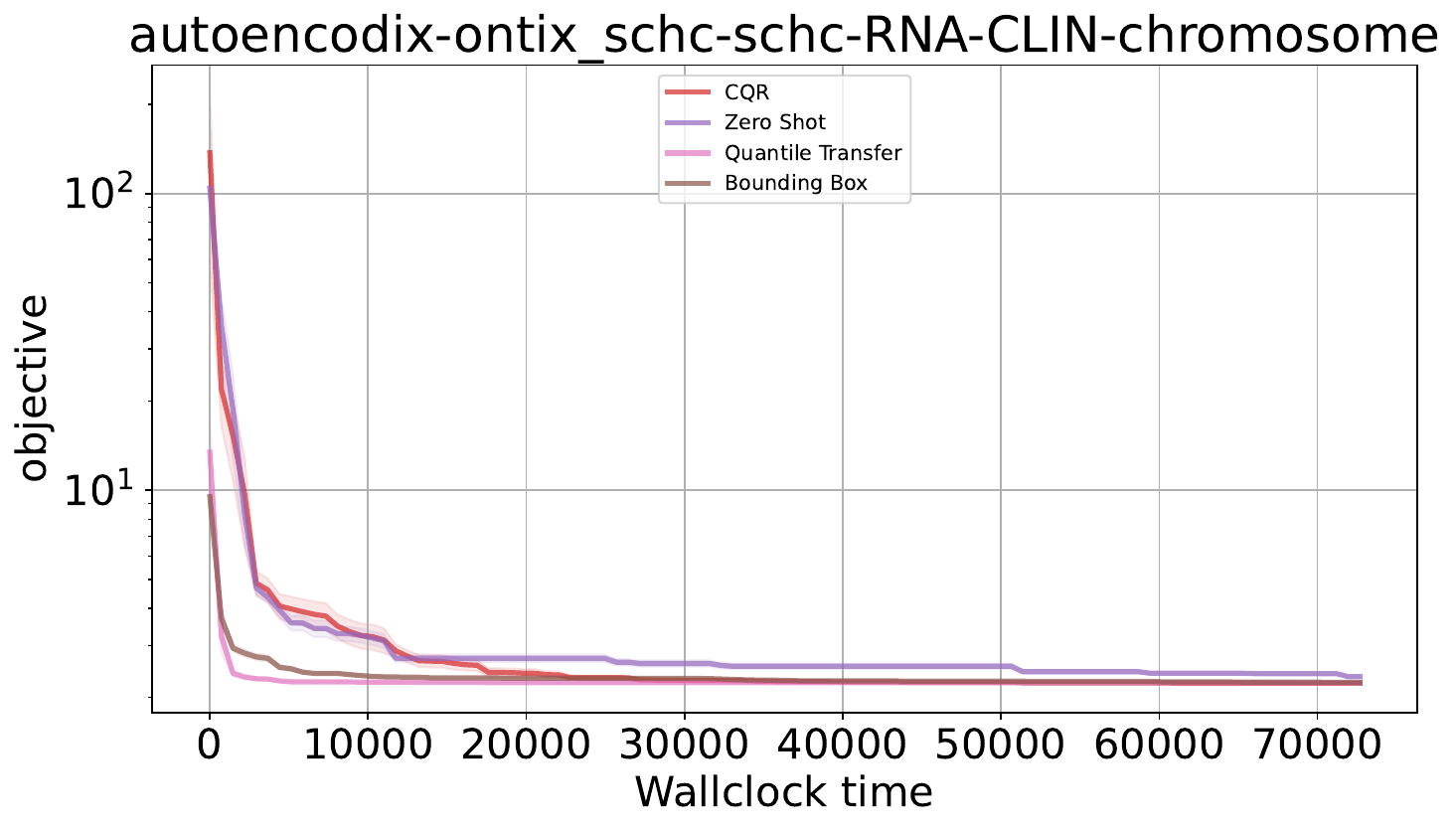} &
    \includegraphics[width=0.32\textwidth]{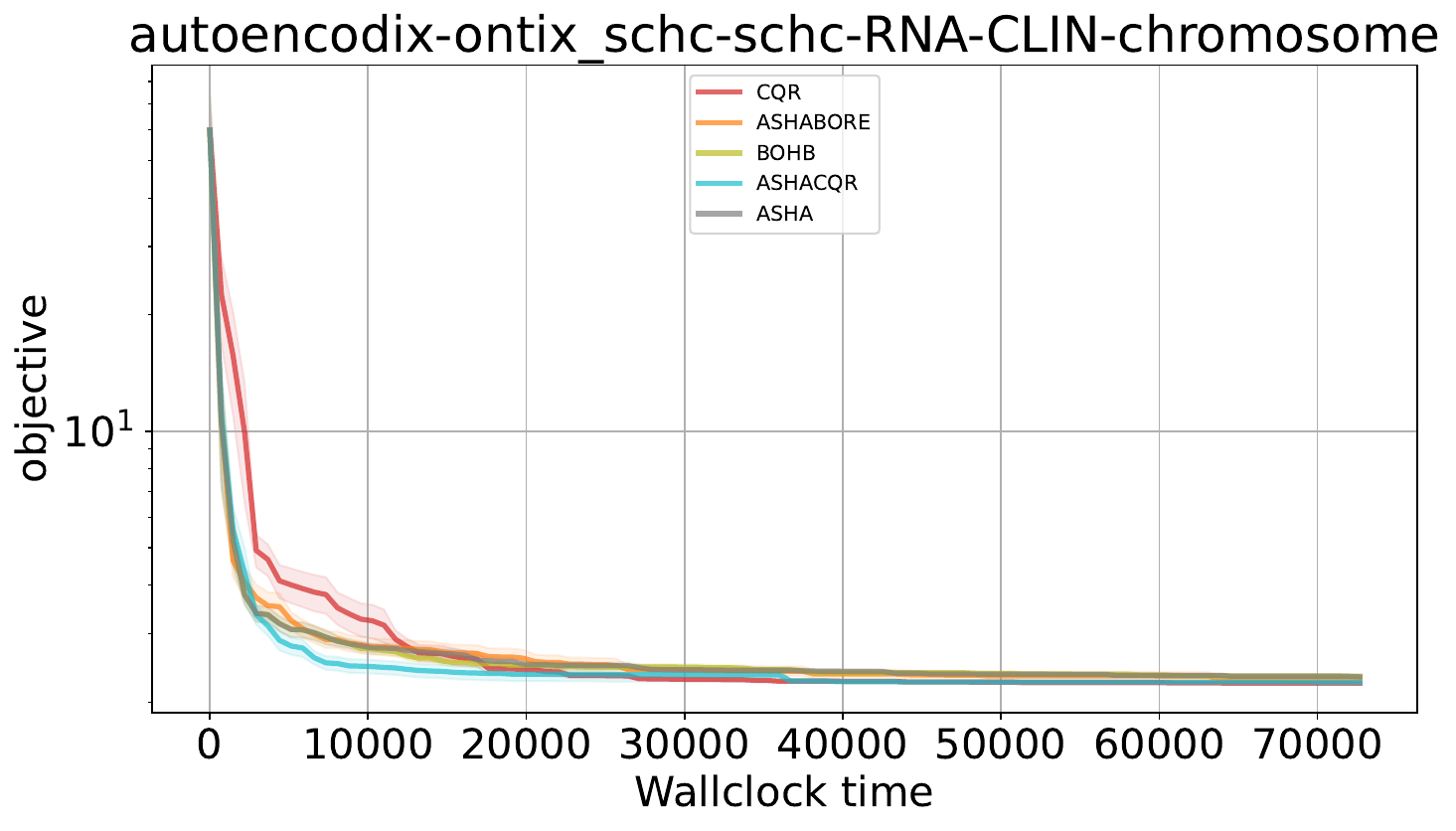} \\
    \includegraphics[width=0.32\textwidth]{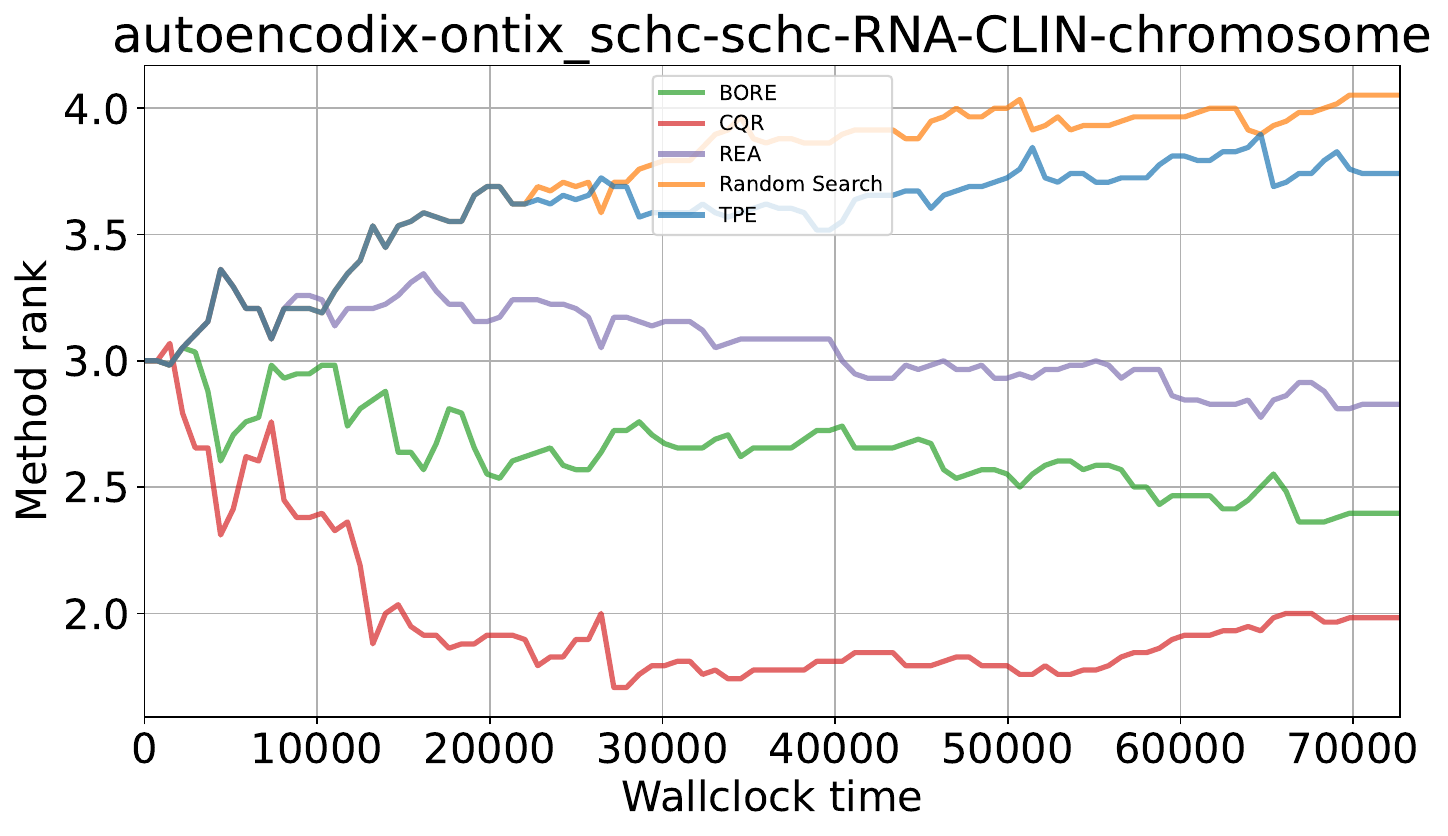} &
    \includegraphics[width=0.32\textwidth]{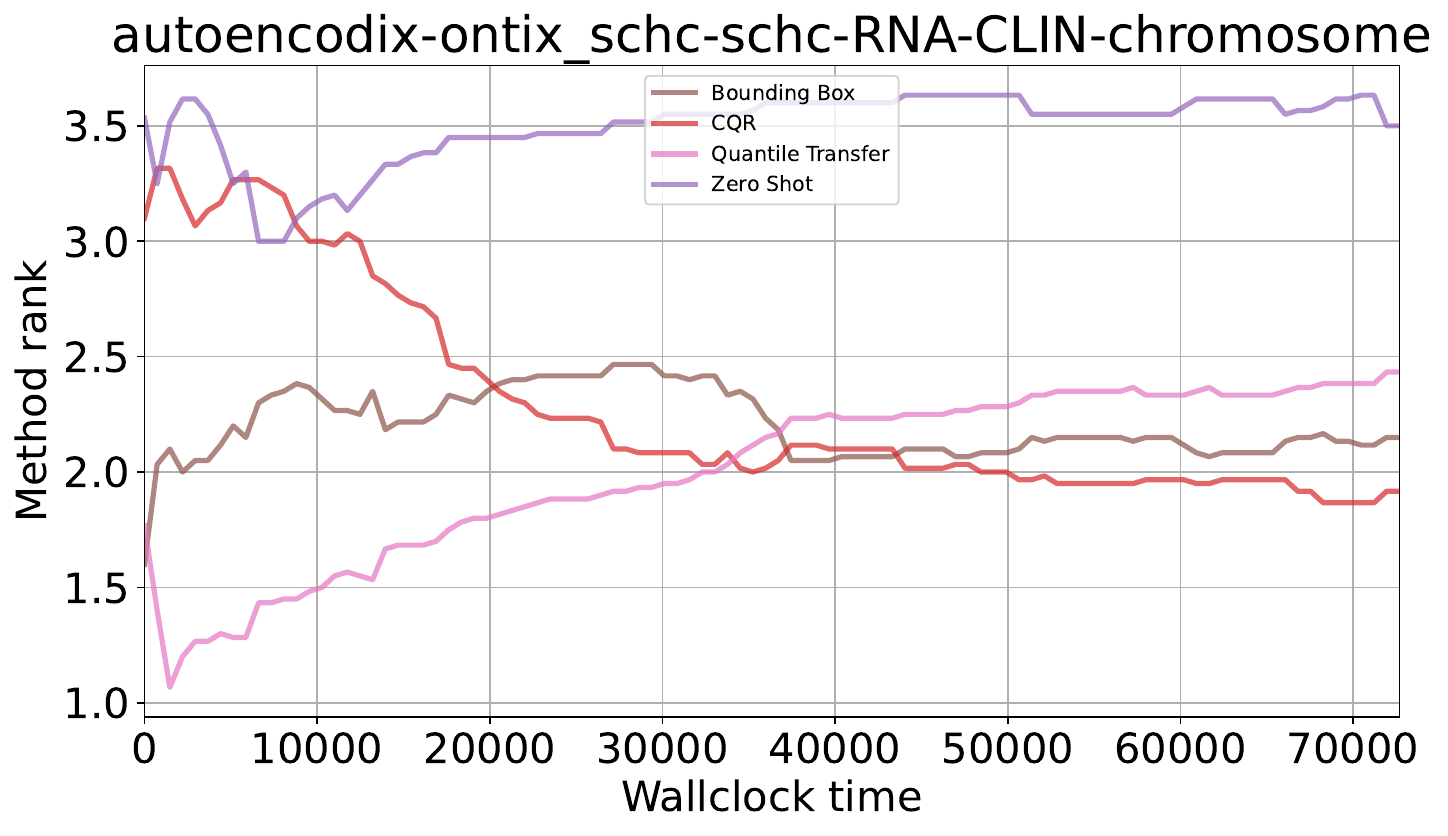} &
    \includegraphics[width=0.32\textwidth]{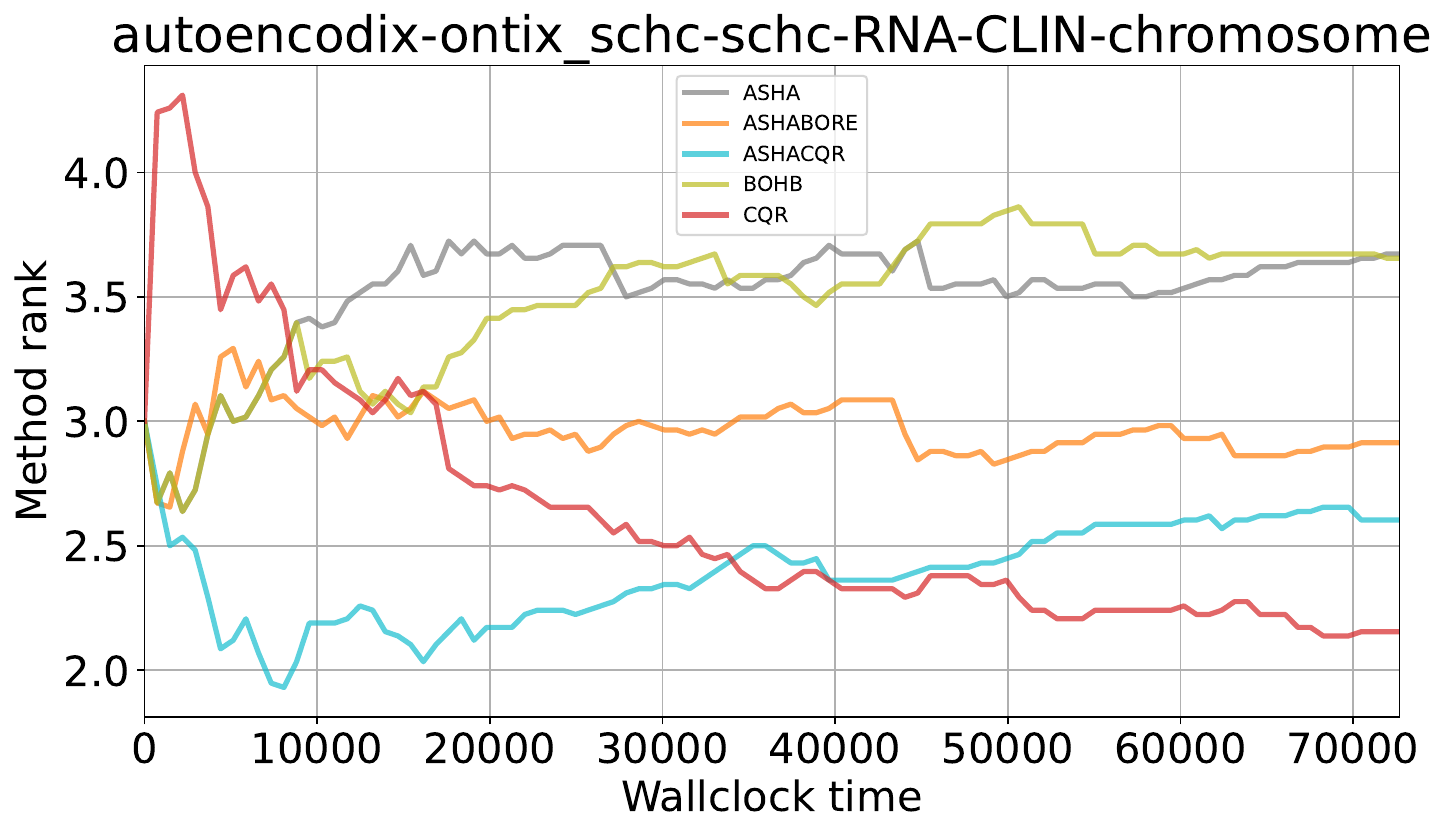} \\
    \end{tabular}
    \caption{Results for Ontix tasks (Part 1).}
    \label{fig:ontix_part1}
\end{figure}

\clearpage

\begin{figure}[htbp]
    \centering
    \setlength{\tabcolsep}{1pt}
    \begin{tabular}{ccc}
    \multicolumn{3}{c}{\textbf{autoencodix-ontix\_schc-schc-RNA-CLIN-reactome}} \\
    \textbf{Single-Fidelity} & \textbf{Transfer Learning} & \textbf{Multi-Fidelity} \\
    \includegraphics[width=0.32\textwidth]{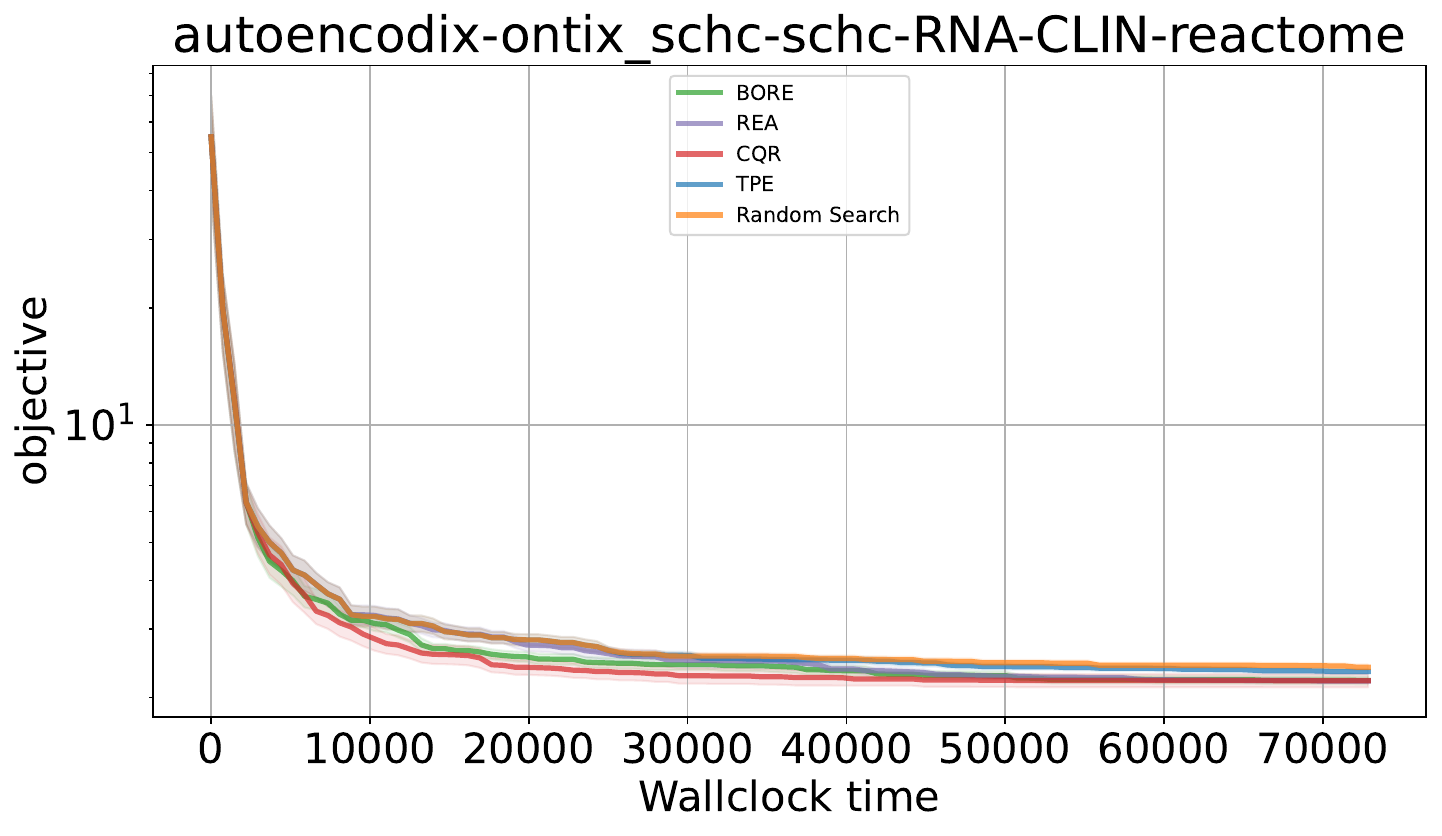} &
    \includegraphics[width=0.32\textwidth]{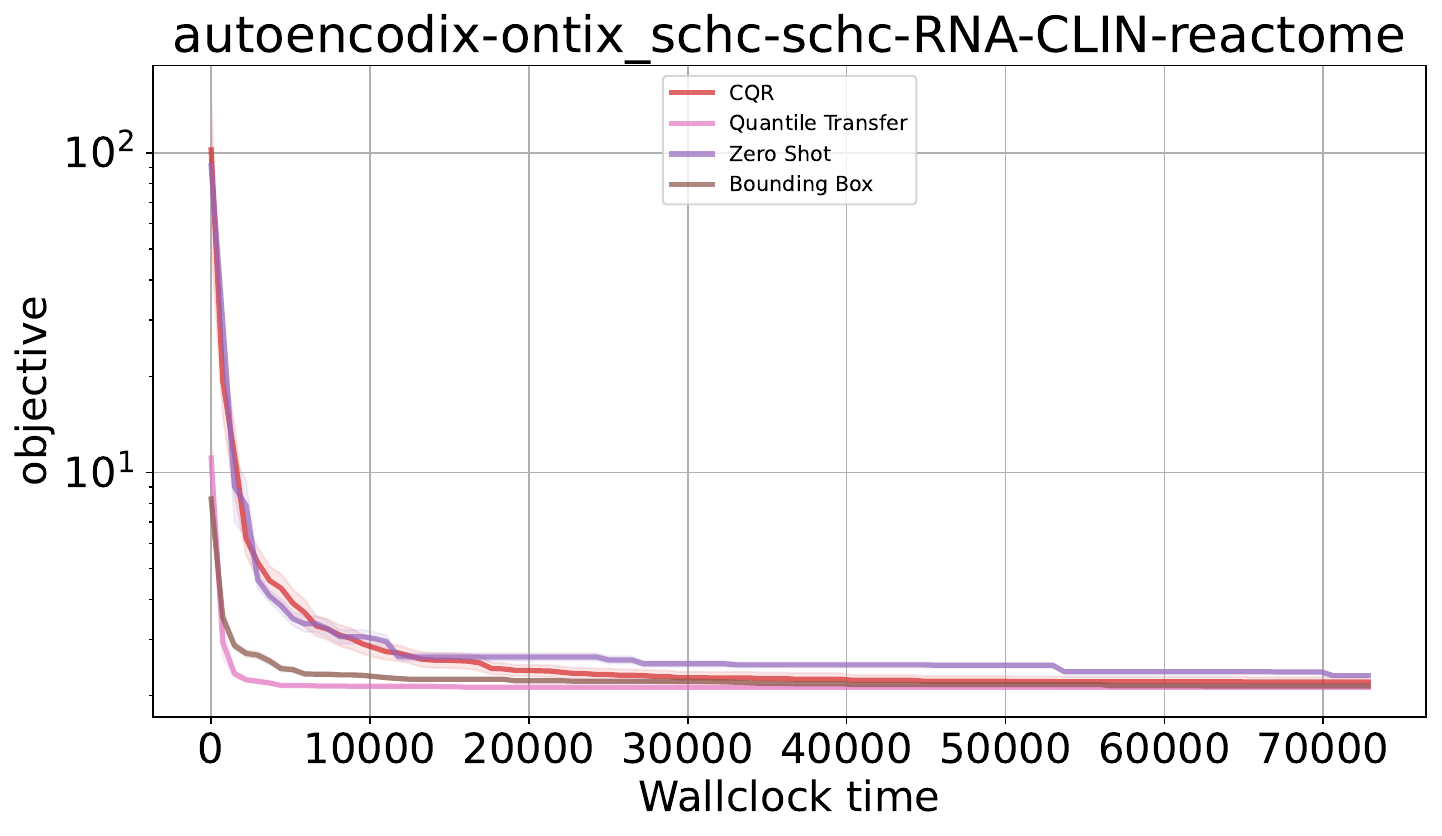} &
    \includegraphics[width=0.32\textwidth]{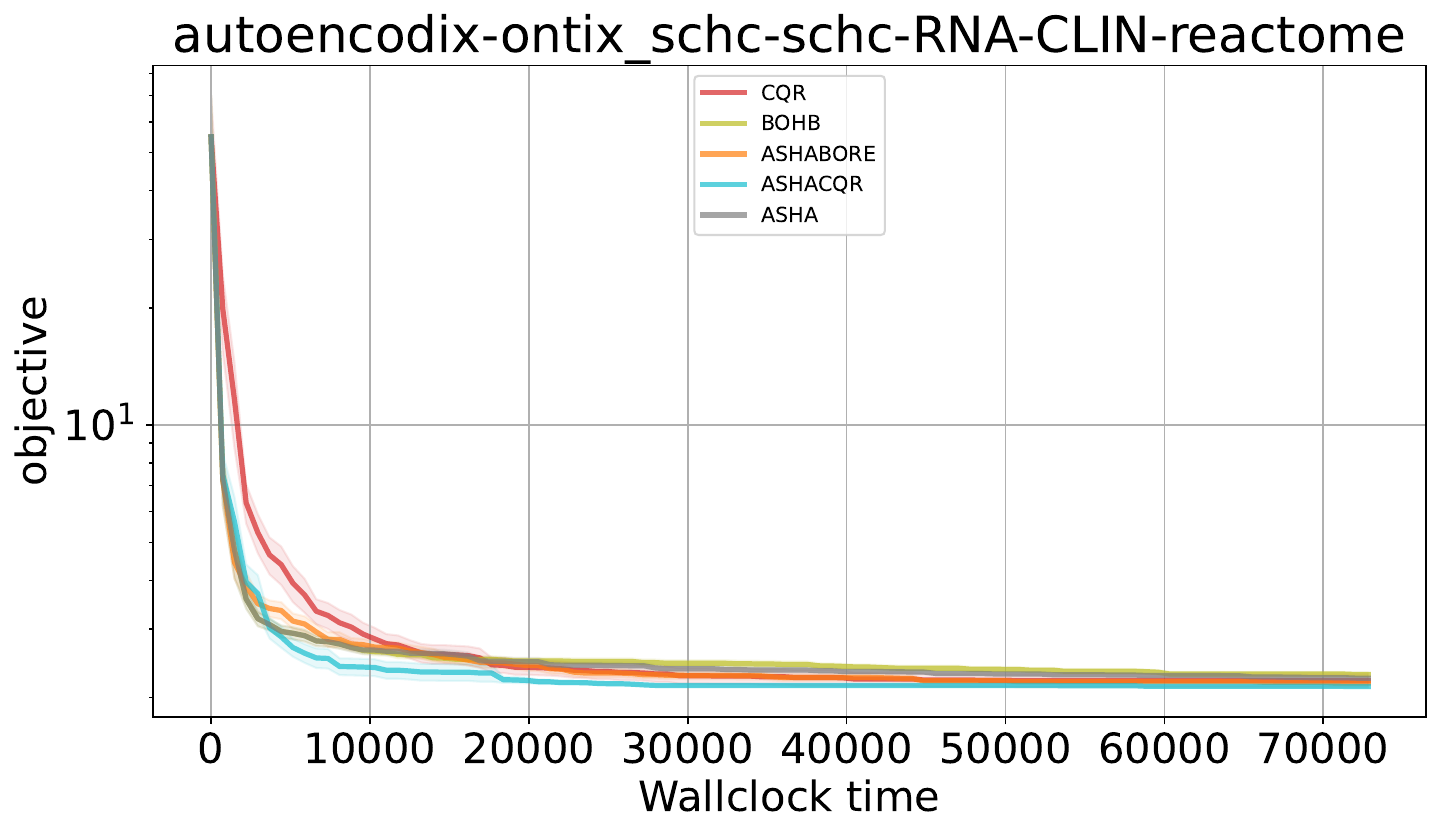} \\
    \includegraphics[width=0.32\textwidth]{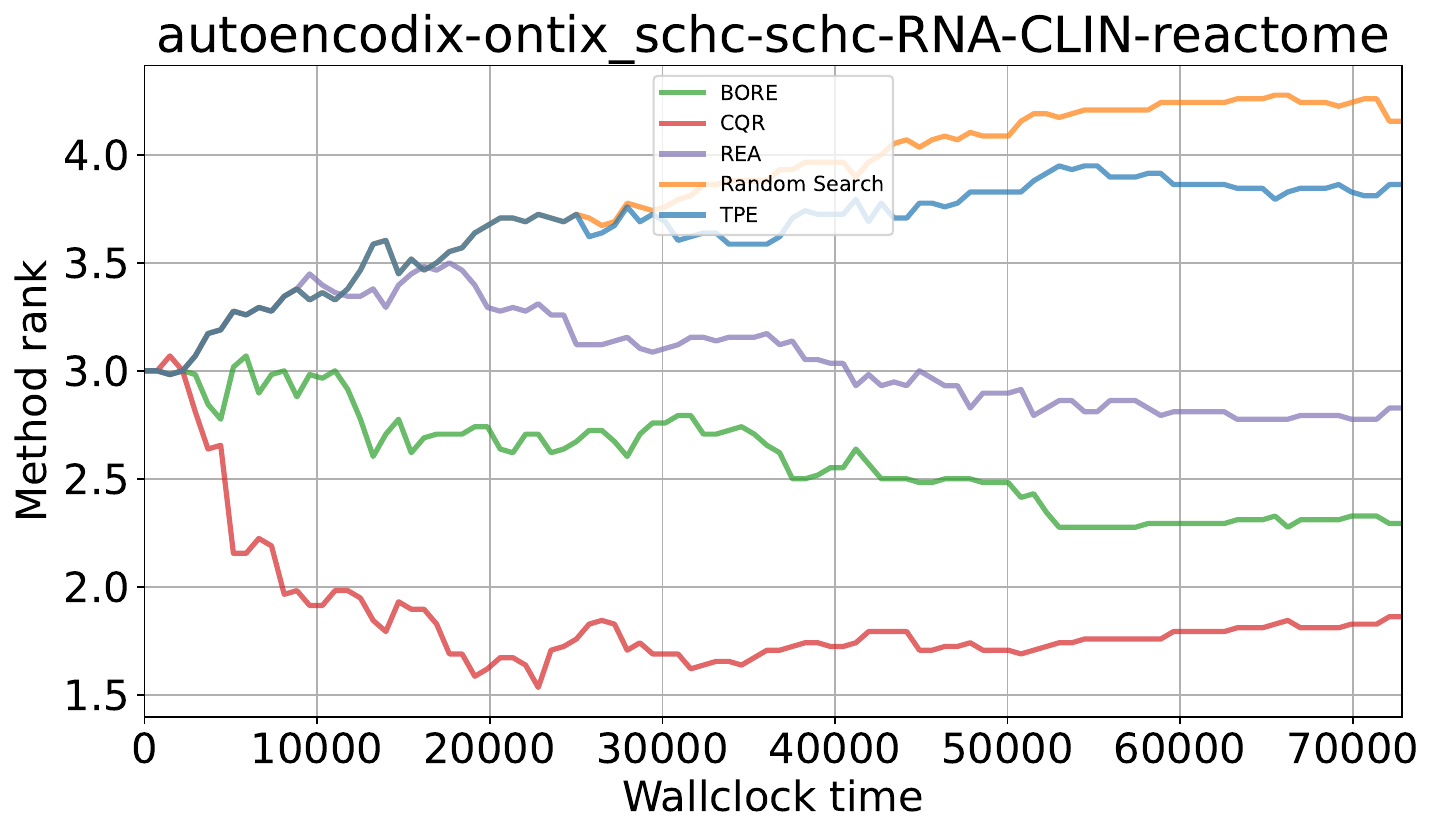} &
    \includegraphics[width=0.32\textwidth]{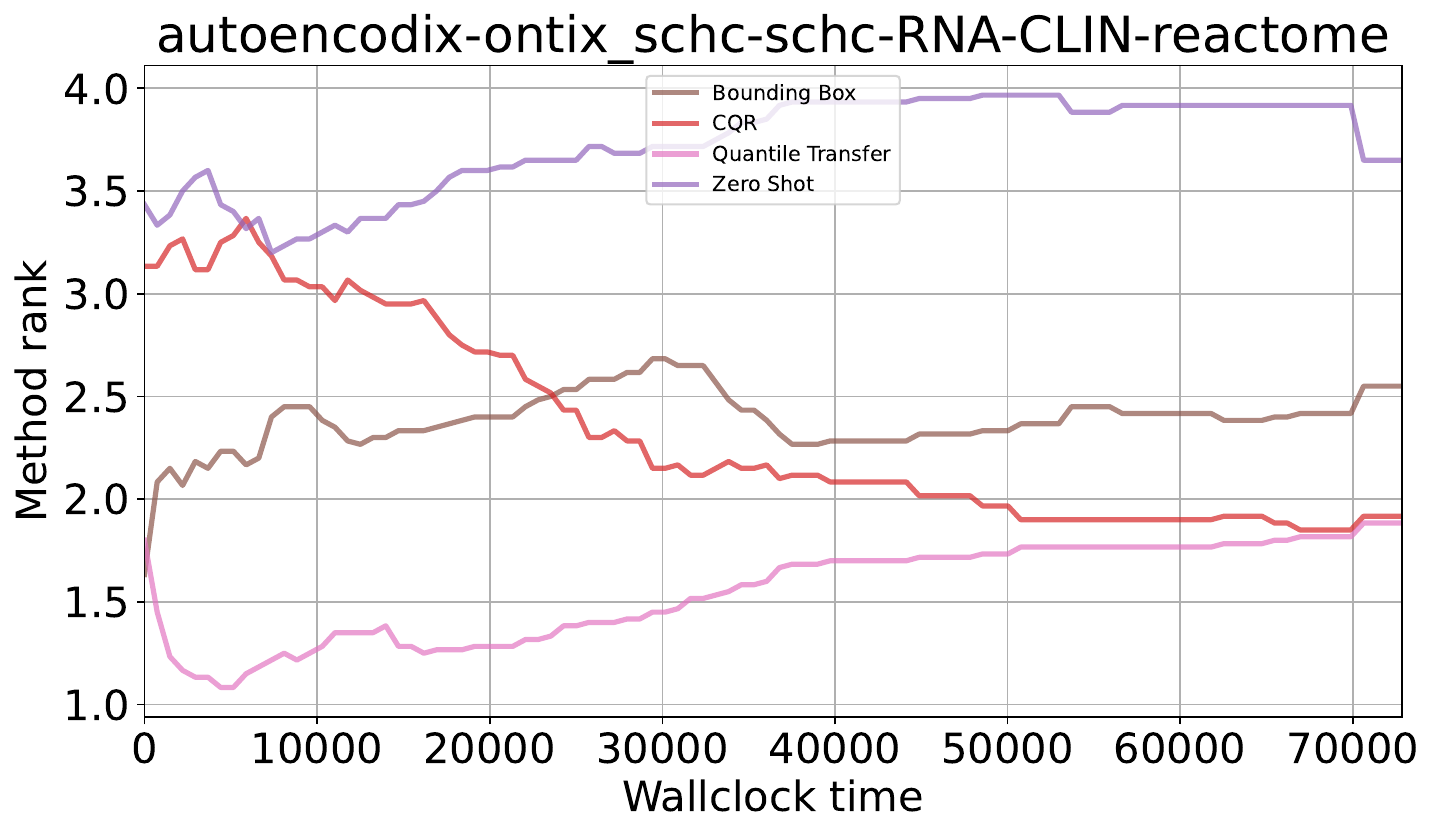} &
    \includegraphics[width=0.32\textwidth]{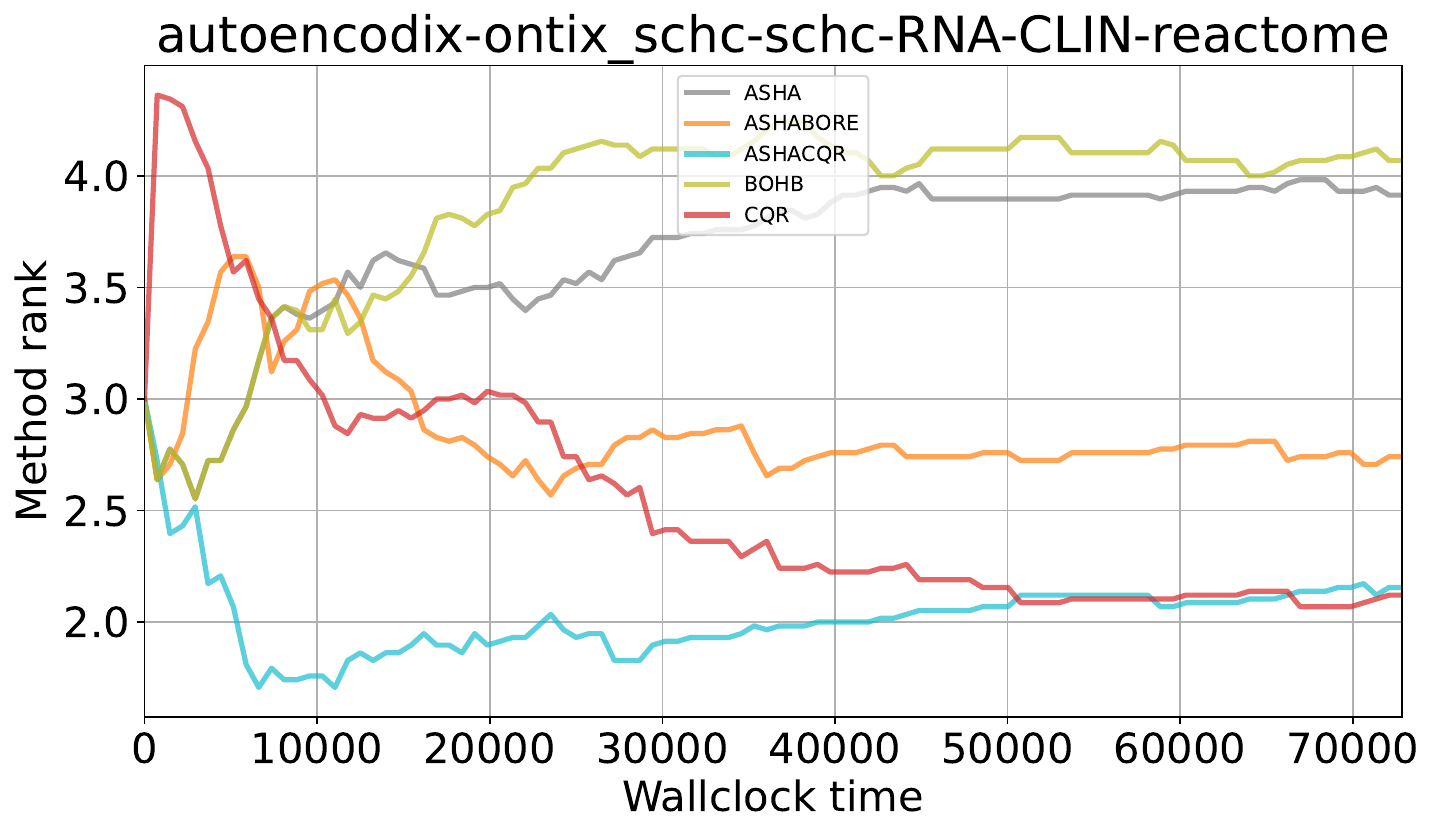} \\
    \midrule
    \multicolumn{3}{c}{\textbf{autoencodix-ontix\_schc-schc-RNA-METH-CLIN-chromosome}} \\
    \includegraphics[width=0.32\textwidth]{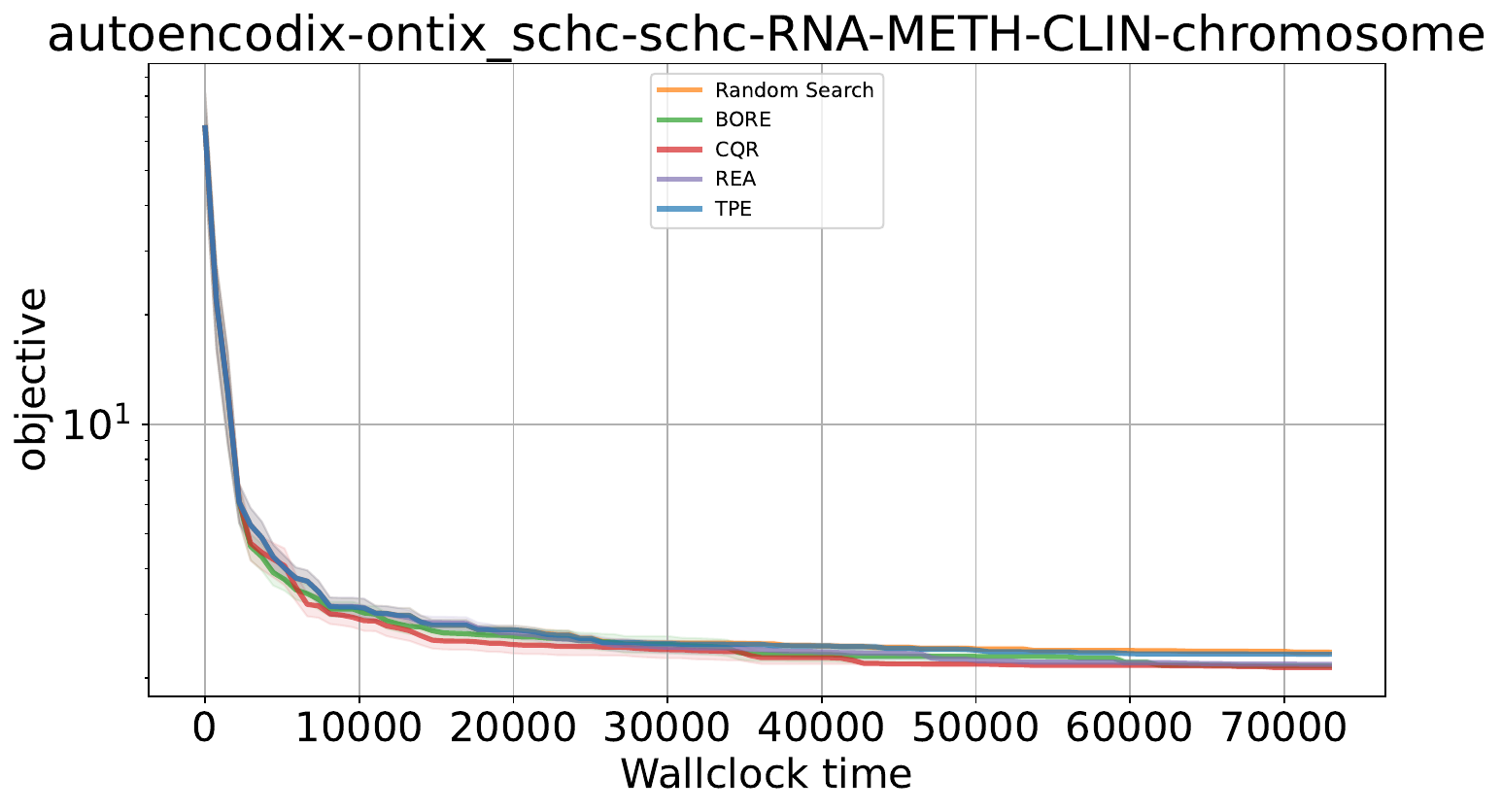} &
    \includegraphics[width=0.32\textwidth]{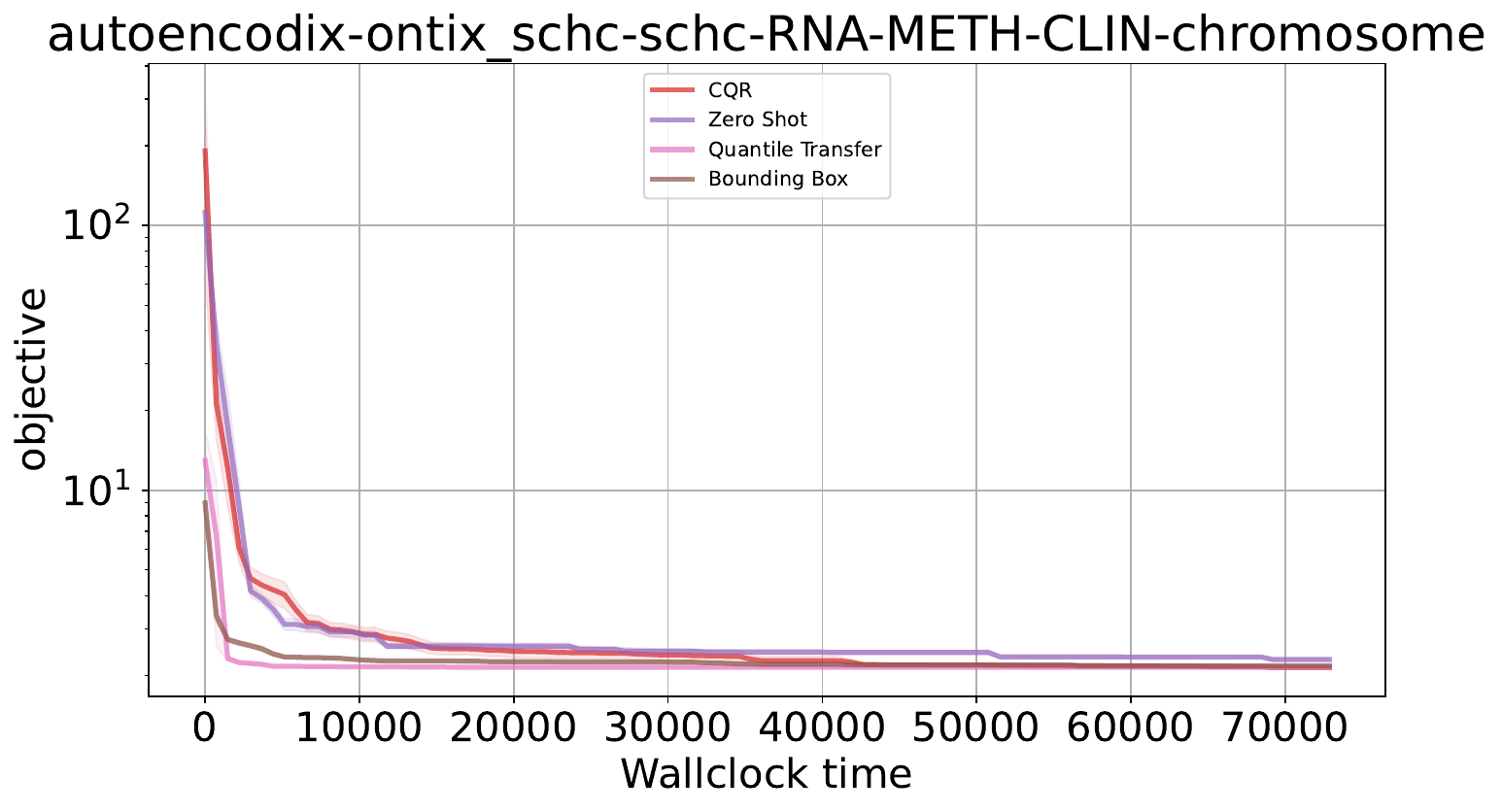} &
    \includegraphics[width=0.32\textwidth]{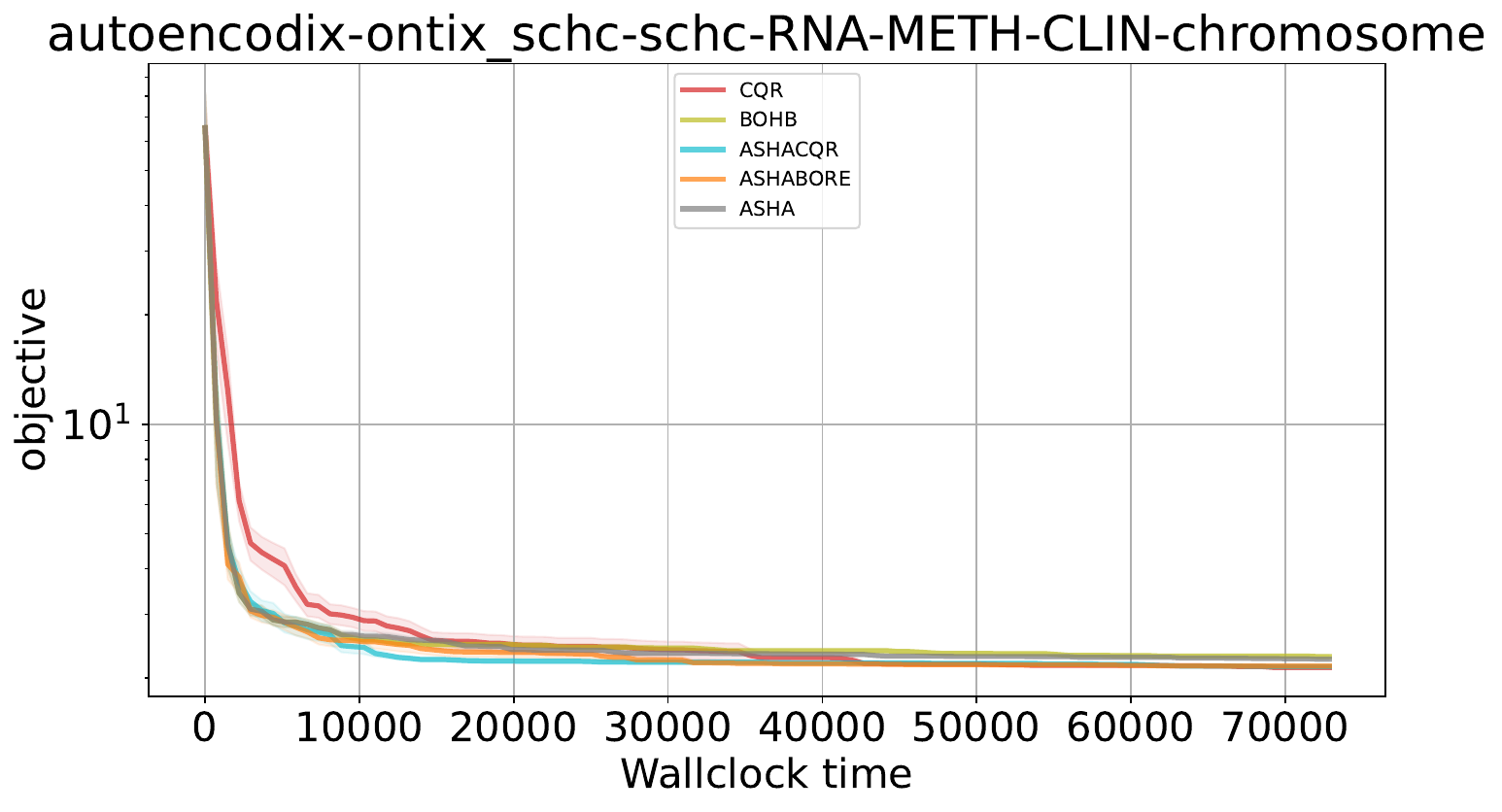} \\
    \includegraphics[width=0.32\textwidth]{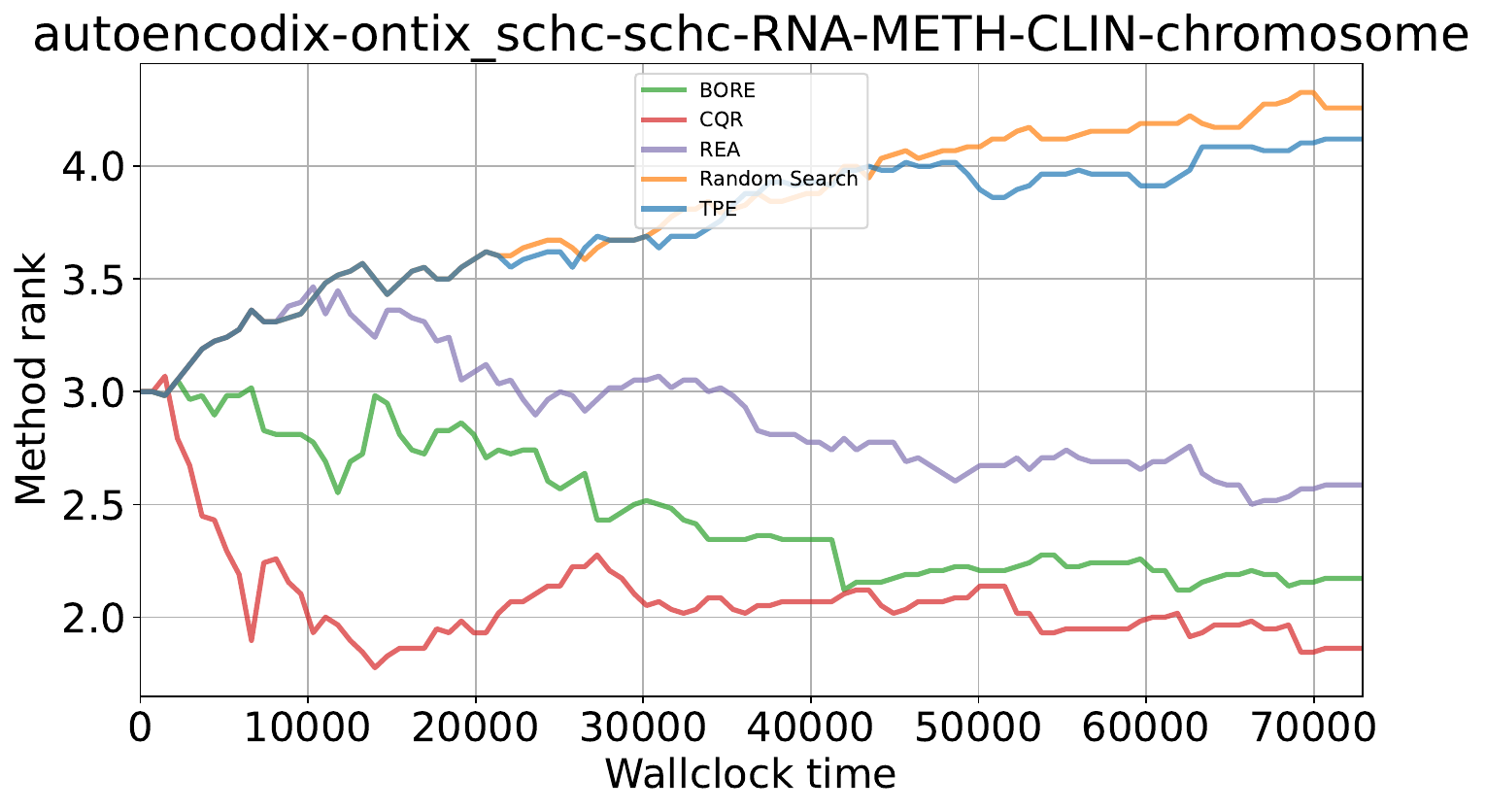} &
    \includegraphics[width=0.32\textwidth]{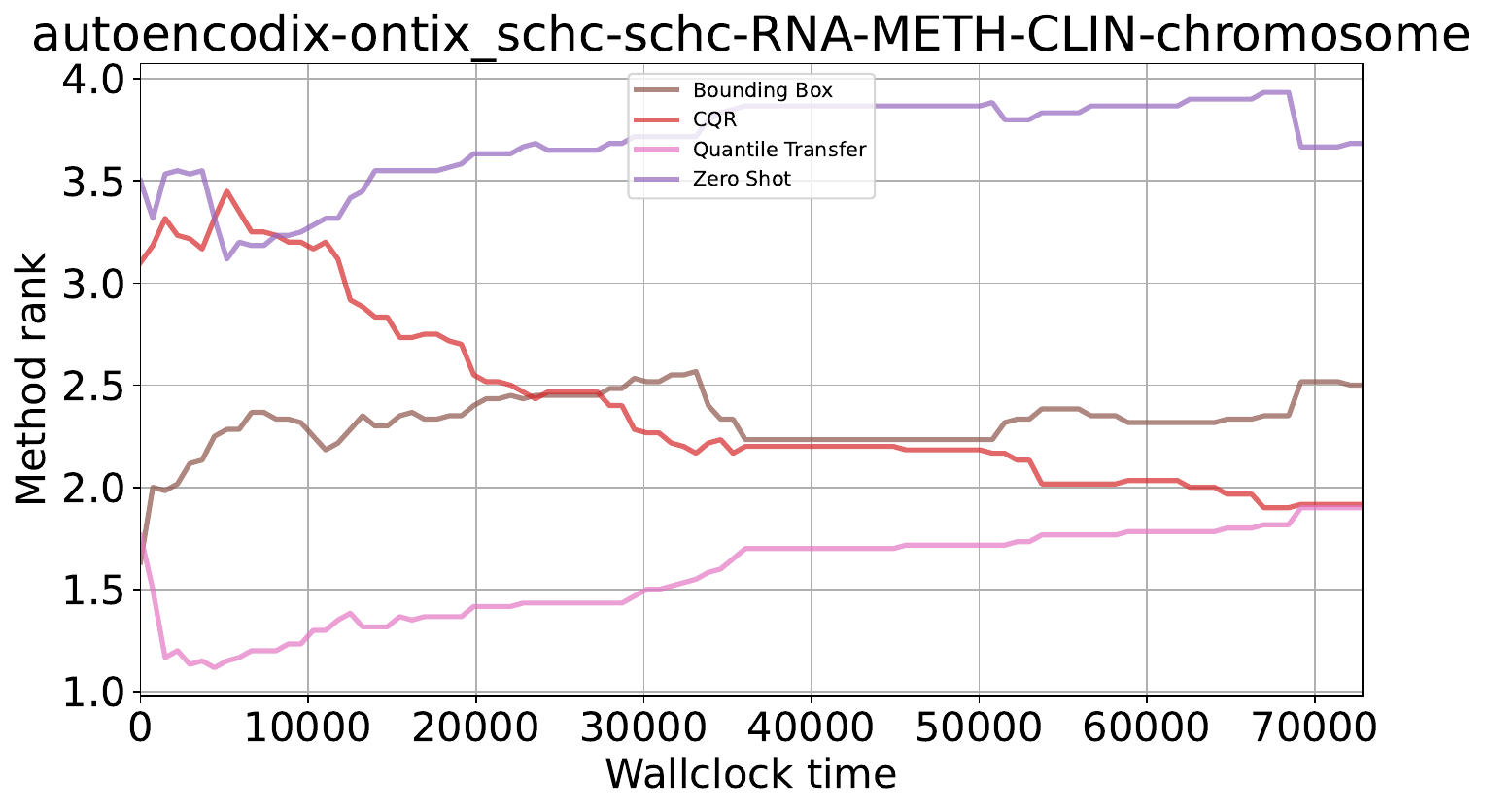} &
    \includegraphics[width=0.32\textwidth]{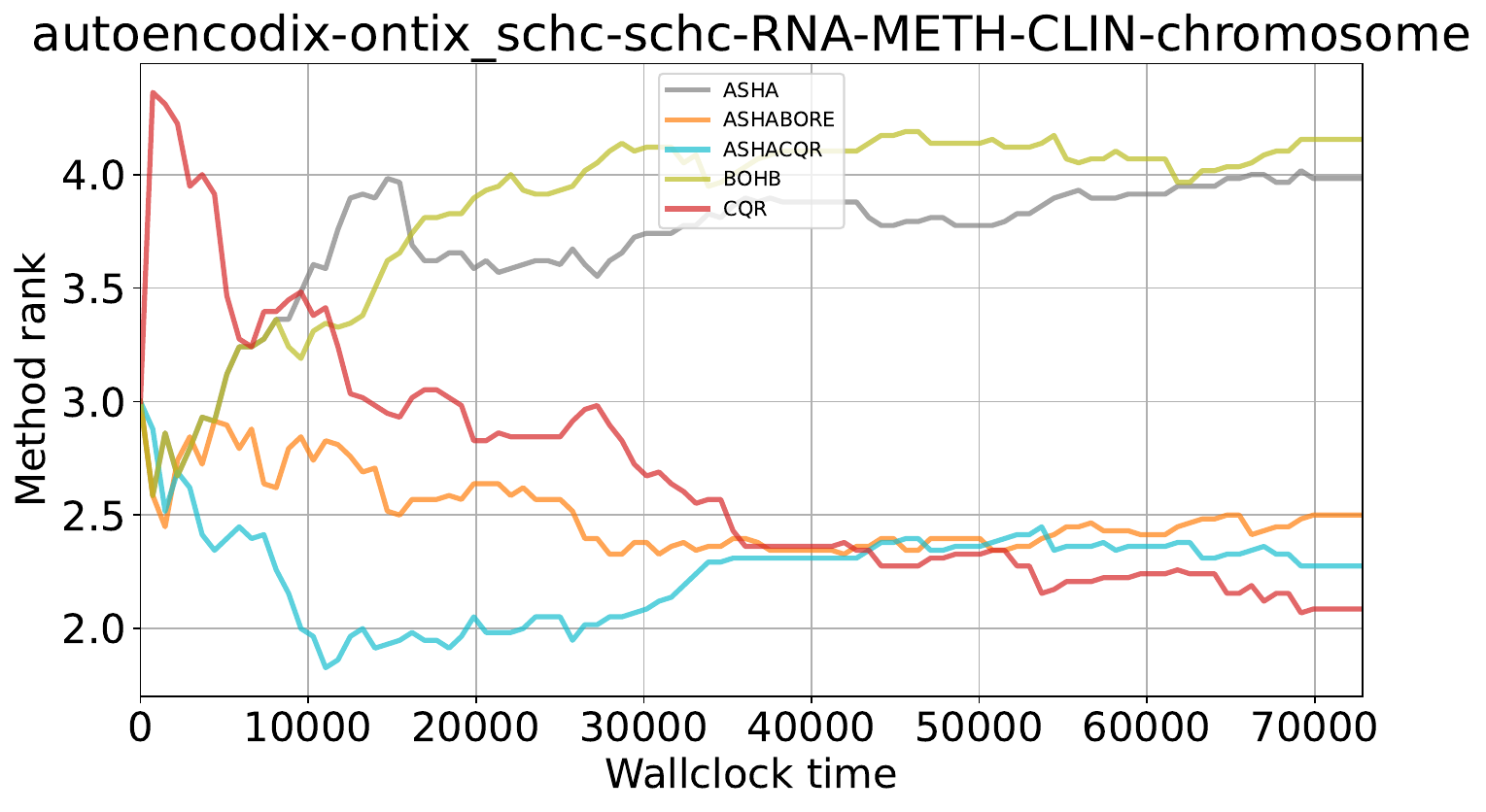} \\
    \midrule
    \multicolumn{3}{c}{\textbf{autoencodix-ontix\_schc-schc-RNA-METH-CLIN-reactome}} \\
    \includegraphics[width=0.32\textwidth]{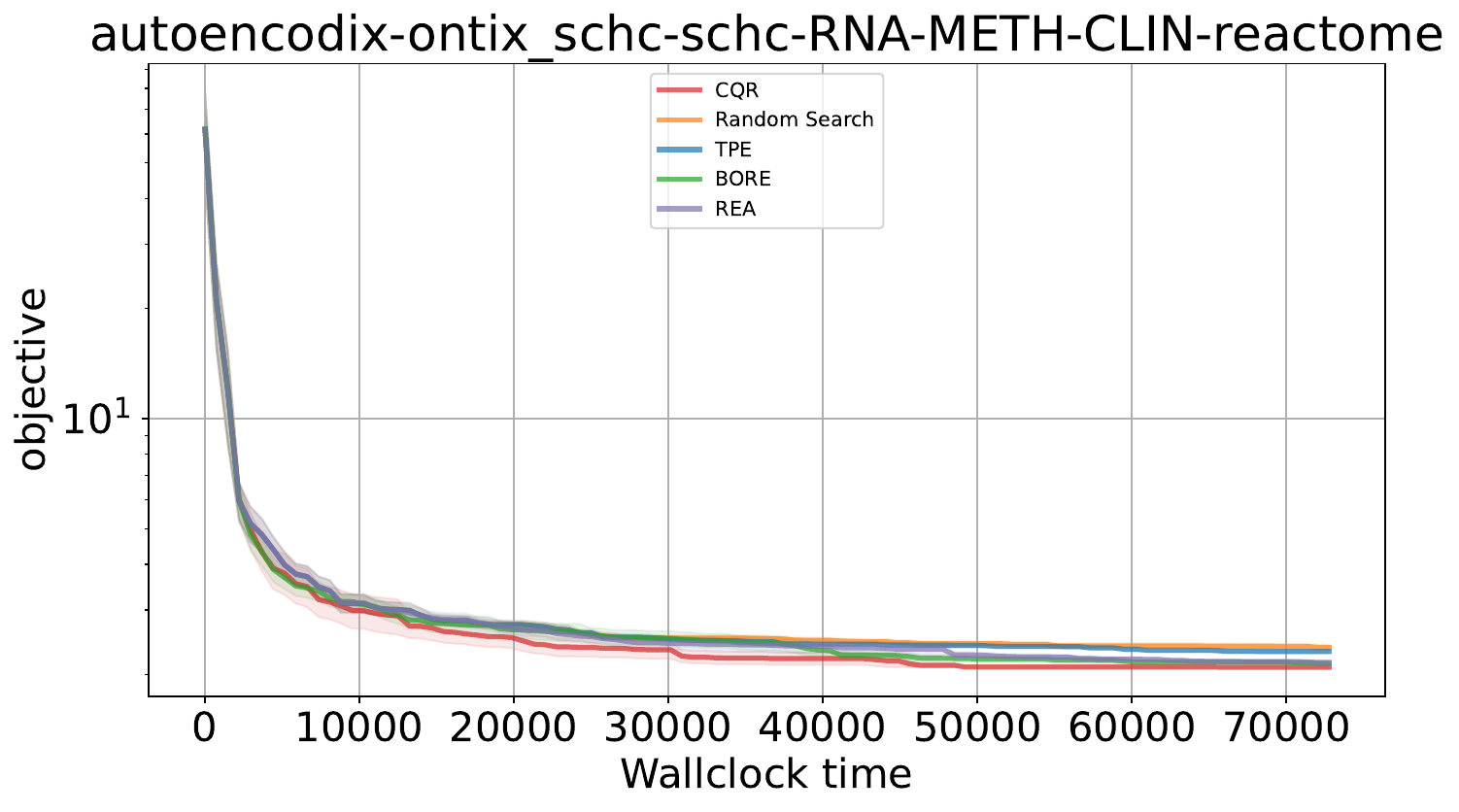} &
    \includegraphics[width=0.32\textwidth]{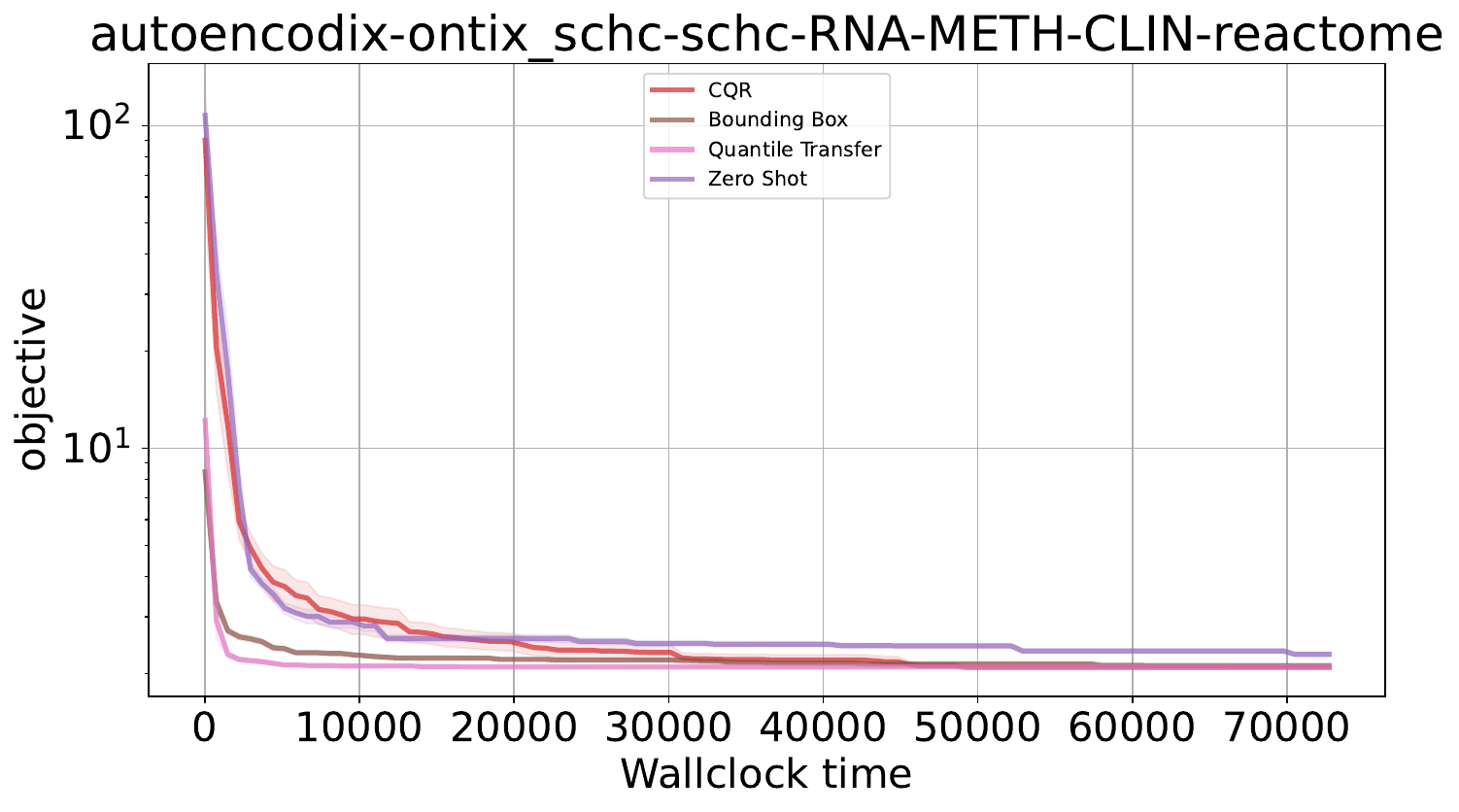} &
    \includegraphics[width=0.32\textwidth]{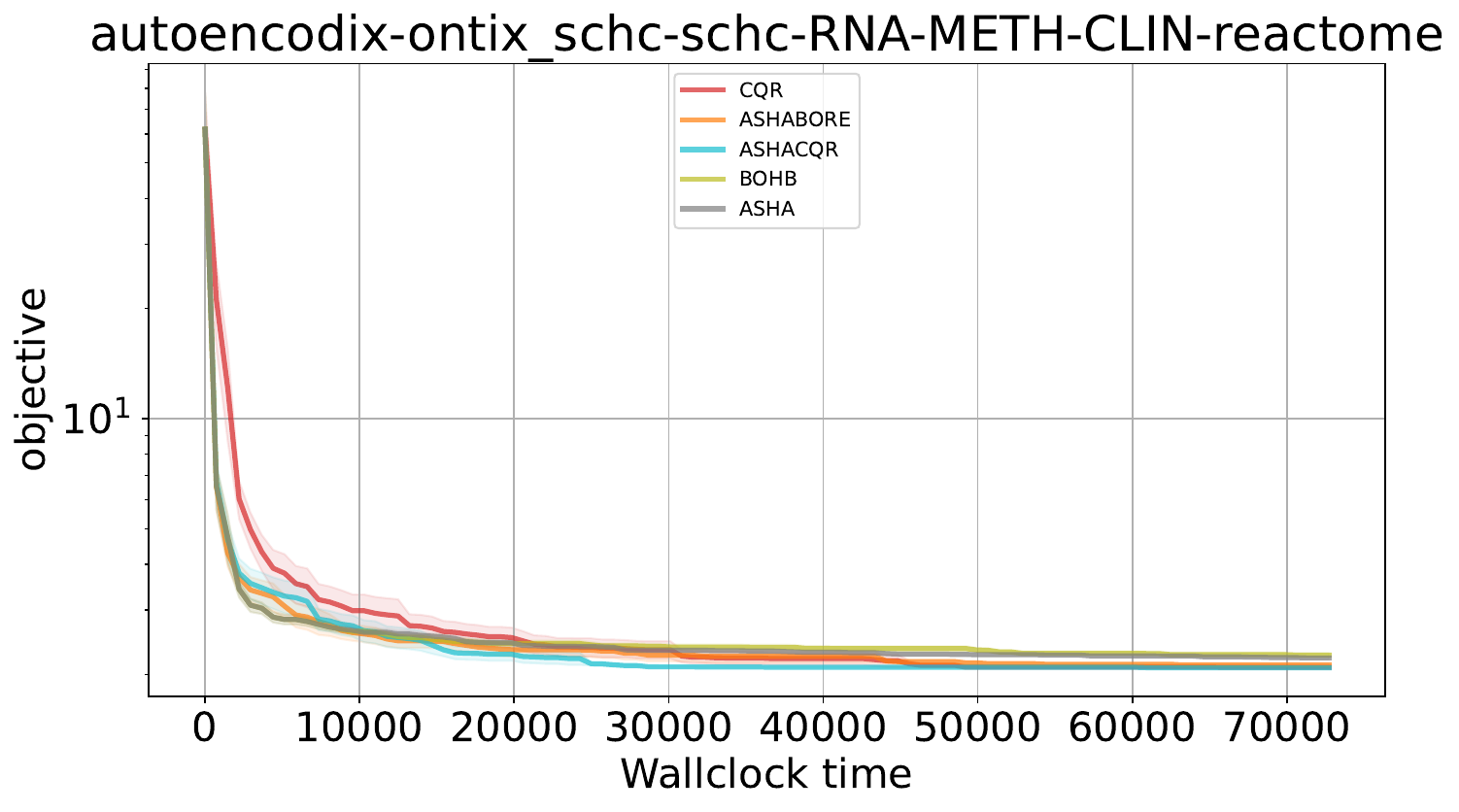} \\
    \includegraphics[width=0.32\textwidth]{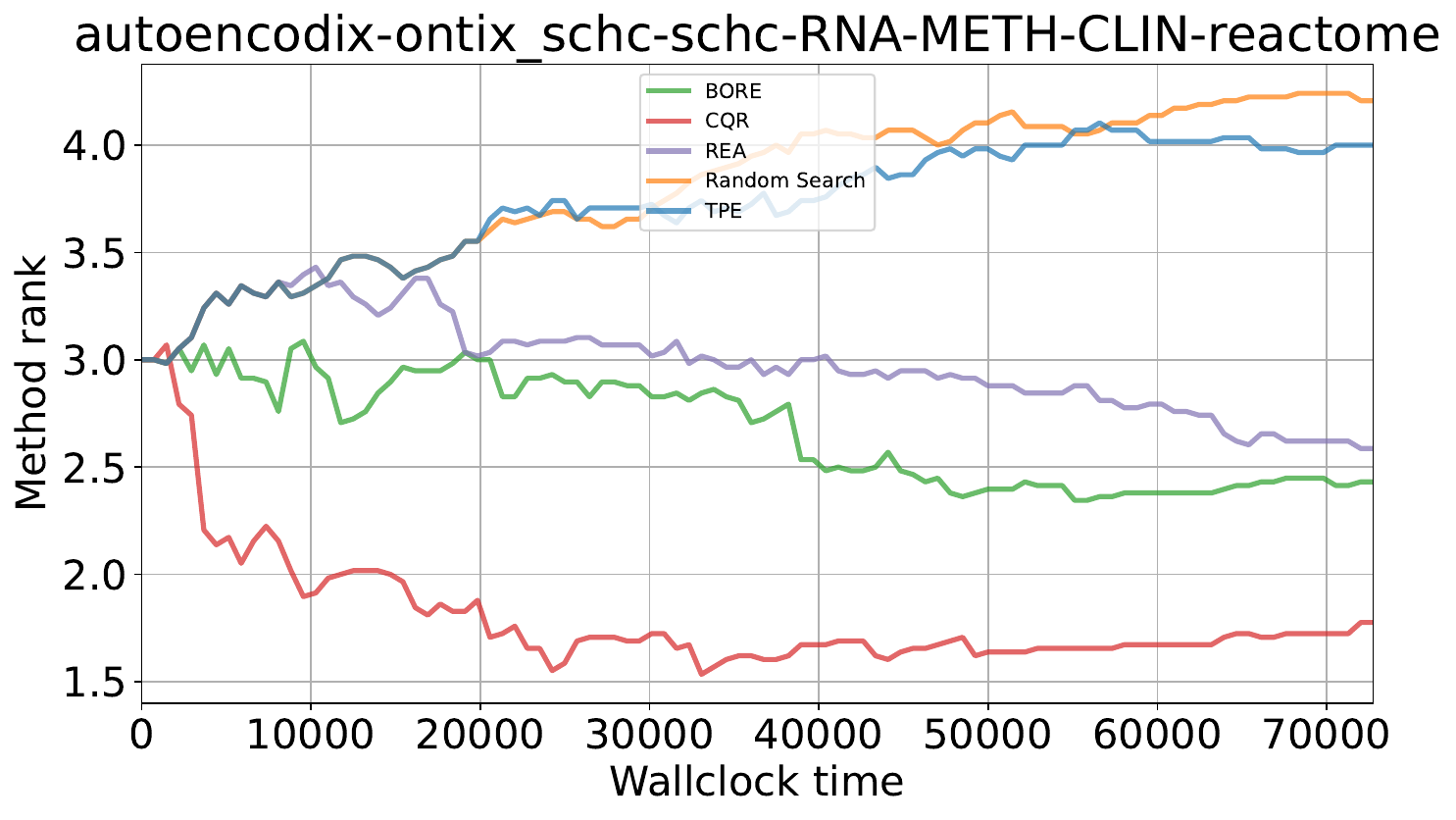} &
    \includegraphics[width=0.32\textwidth]{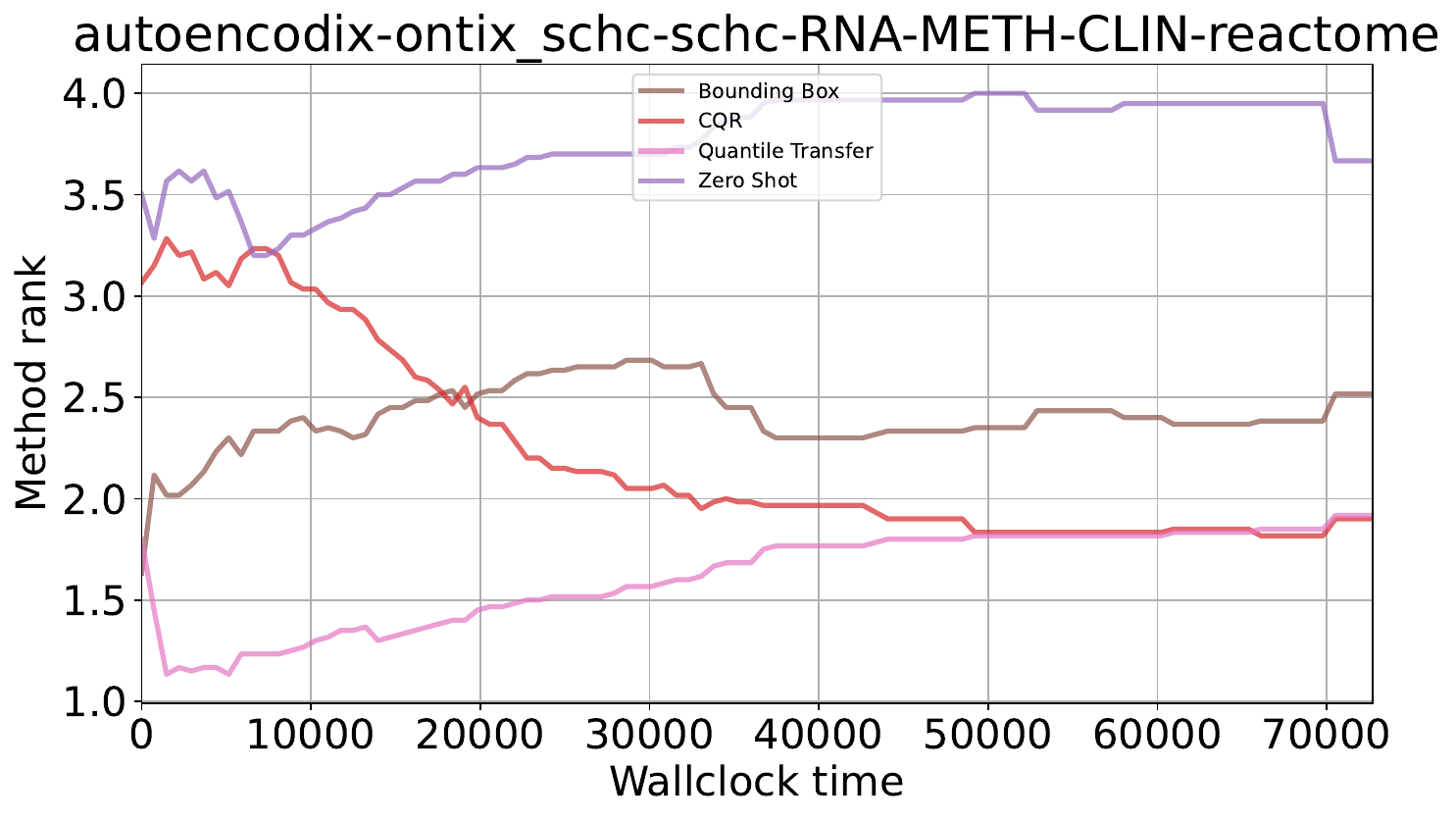} &
    \includegraphics[width=0.32\textwidth]{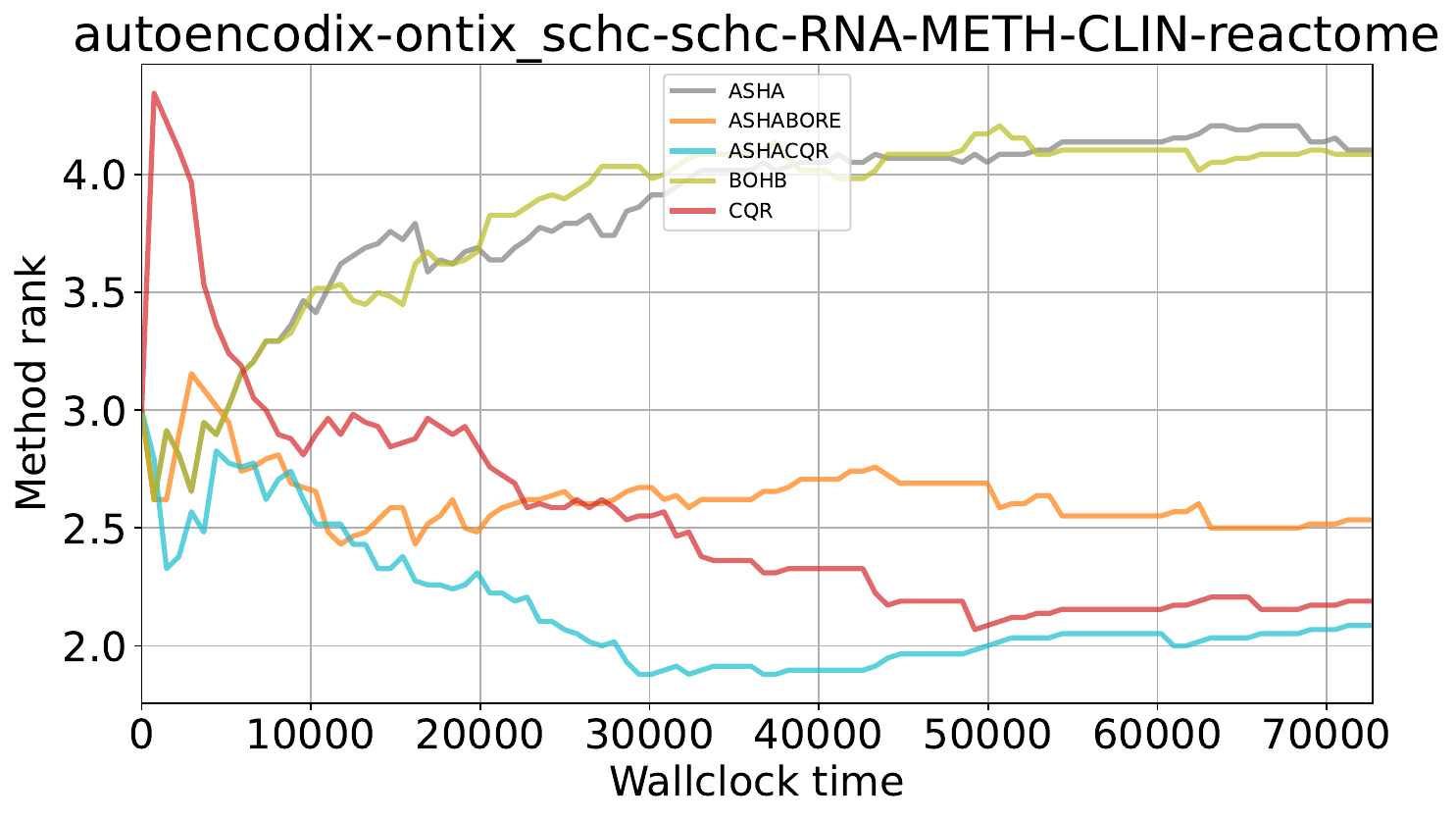} \\
    \end{tabular}
    \caption{Results for Ontix tasks (Part 2).}
    \label{fig:ontix_part2}
\end{figure}

\clearpage

\begin{figure}[htbp]
    \centering
    \setlength{\tabcolsep}{1pt}
    \begin{tabular}{ccc}
    \multicolumn{3}{c}{\textbf{autoencodix-ontix\_tcga-tcga-DNA-CLIN-chromosome}} \\
    \textbf{Single-Fidelity} & \textbf{Transfer Learning} & \textbf{Multi-Fidelity} \\
    \includegraphics[width=0.32\textwidth]{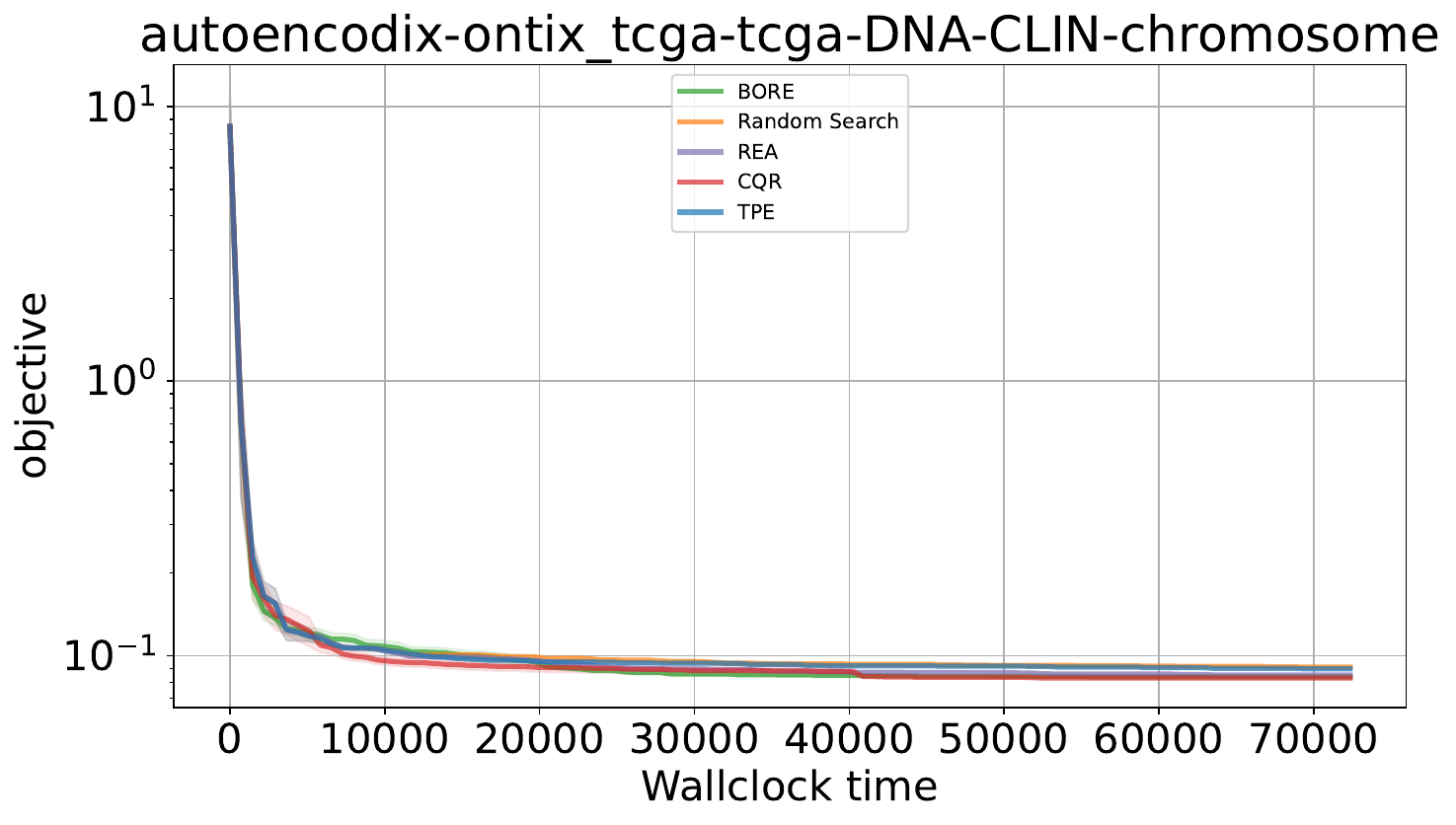} &
    \includegraphics[width=0.32\textwidth]{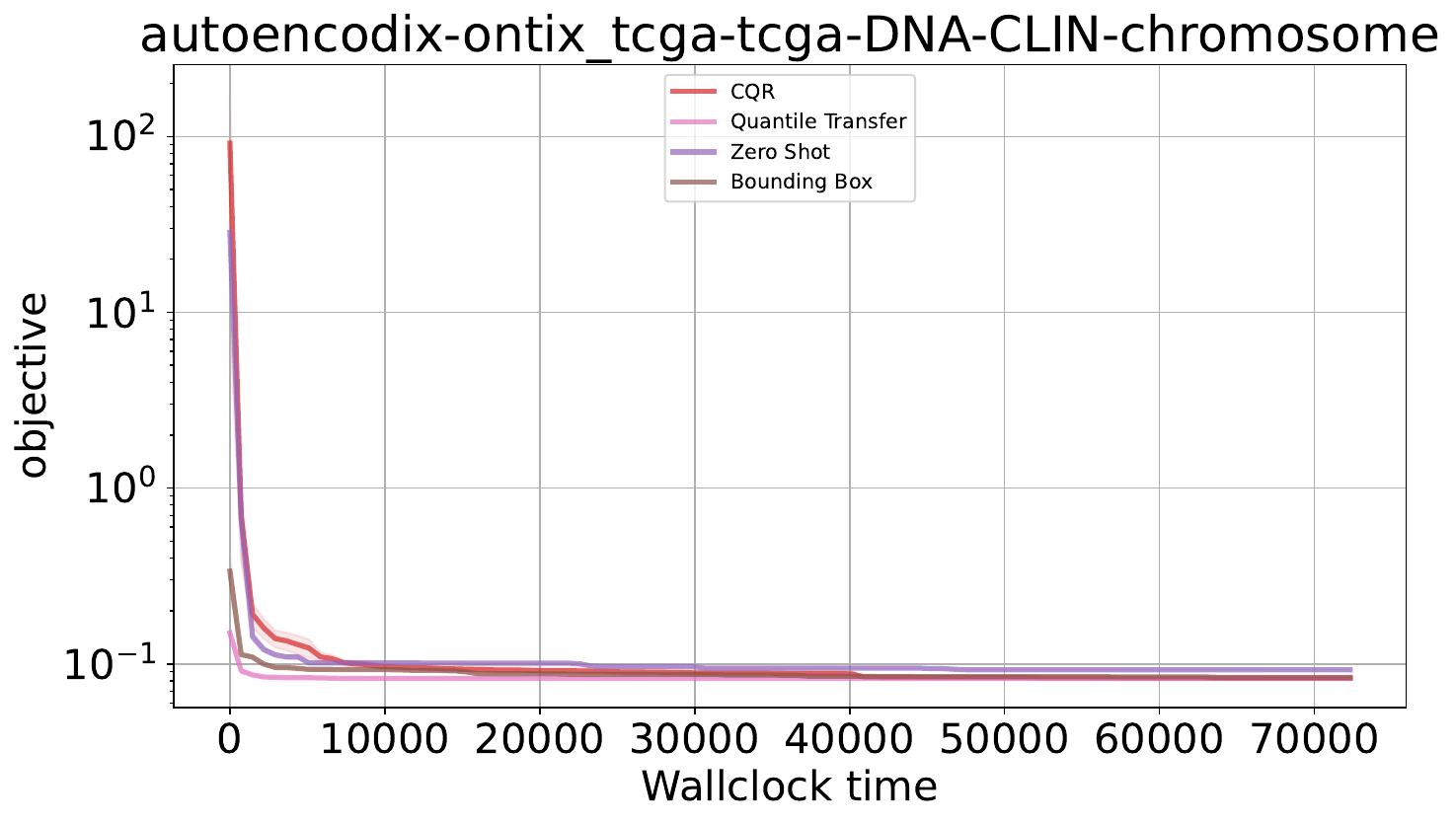} &
    \includegraphics[width=0.32\textwidth]{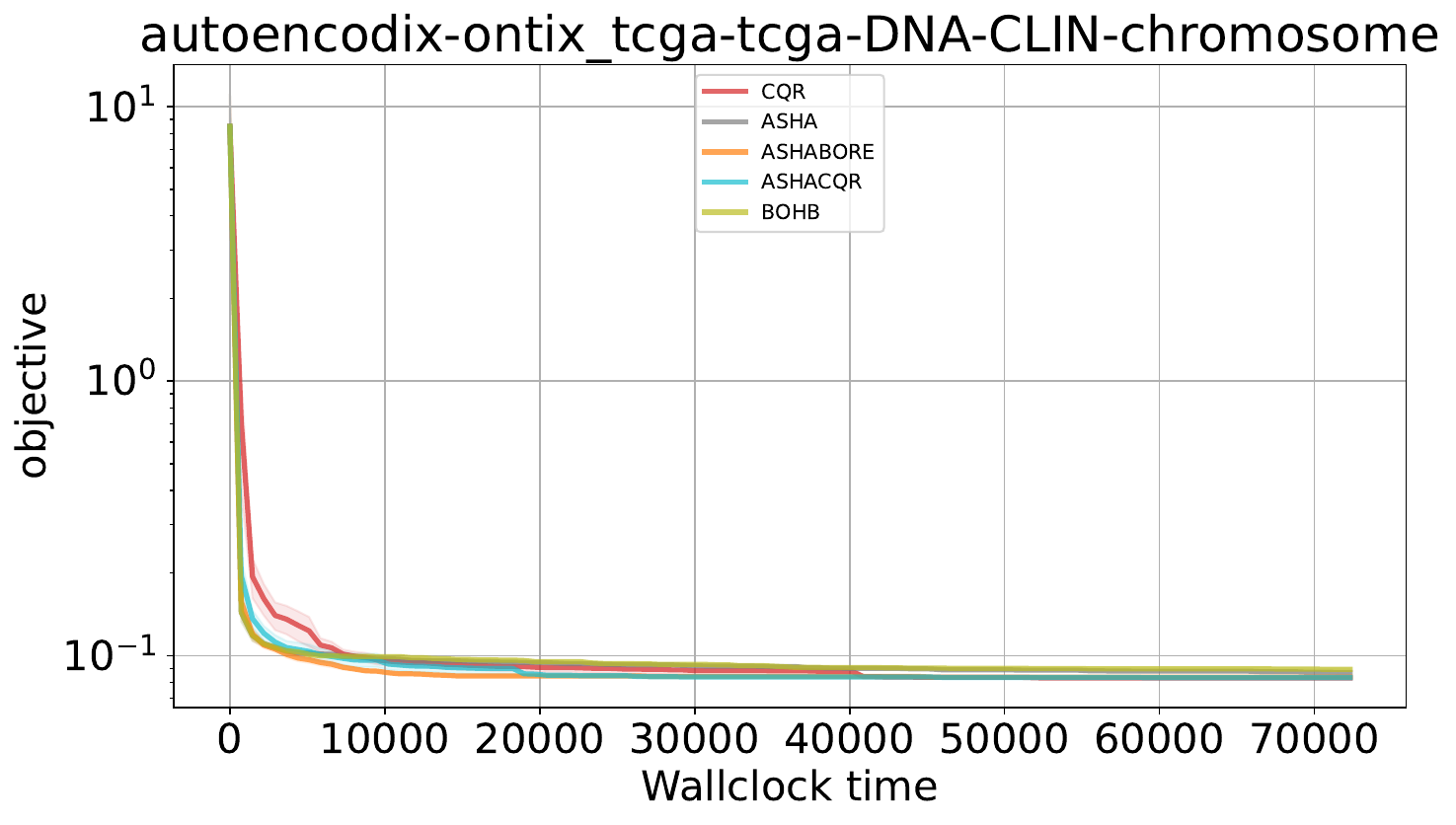} \\
    \includegraphics[width=0.32\textwidth]{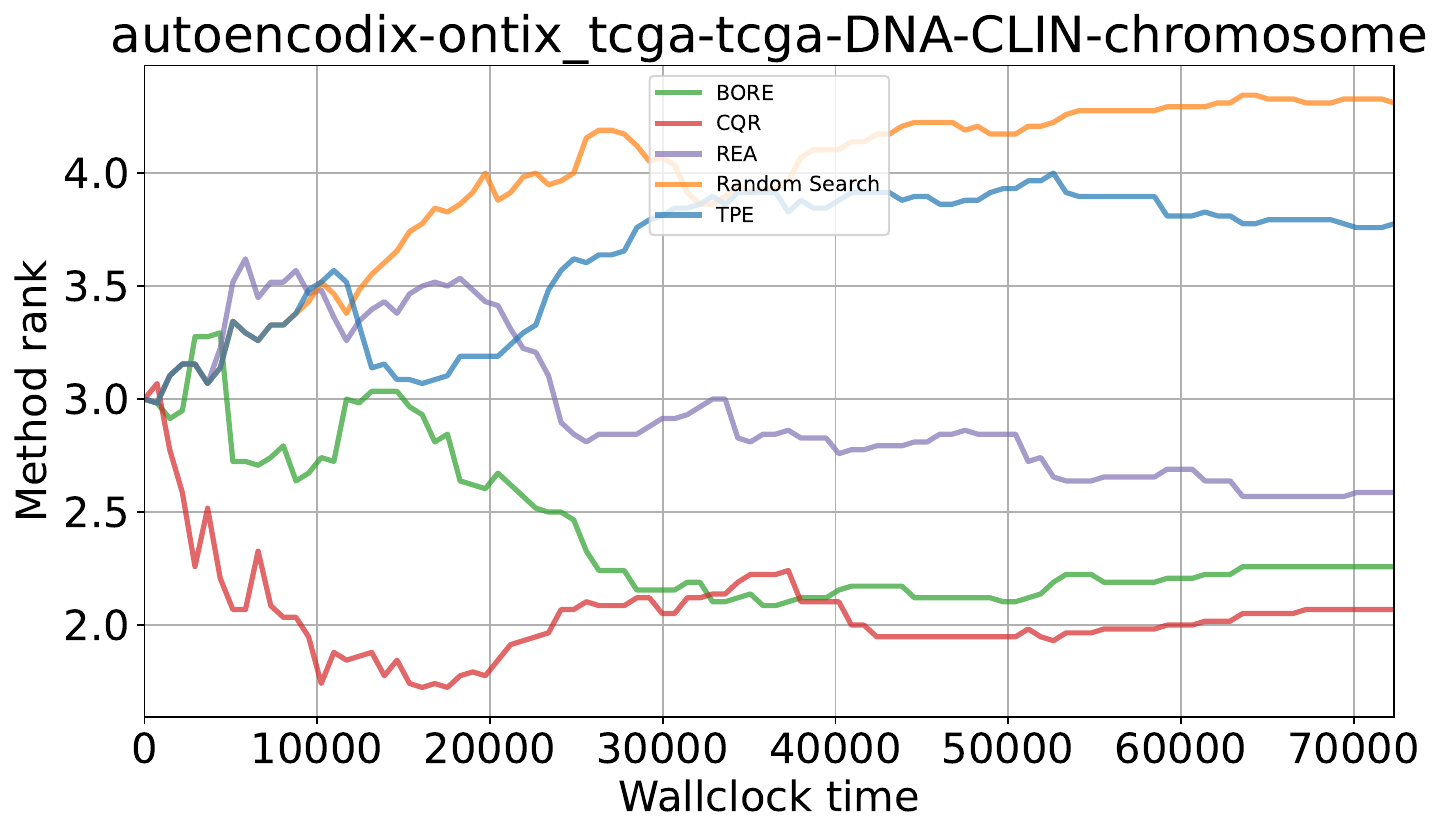} &
    \includegraphics[width=0.32\textwidth]{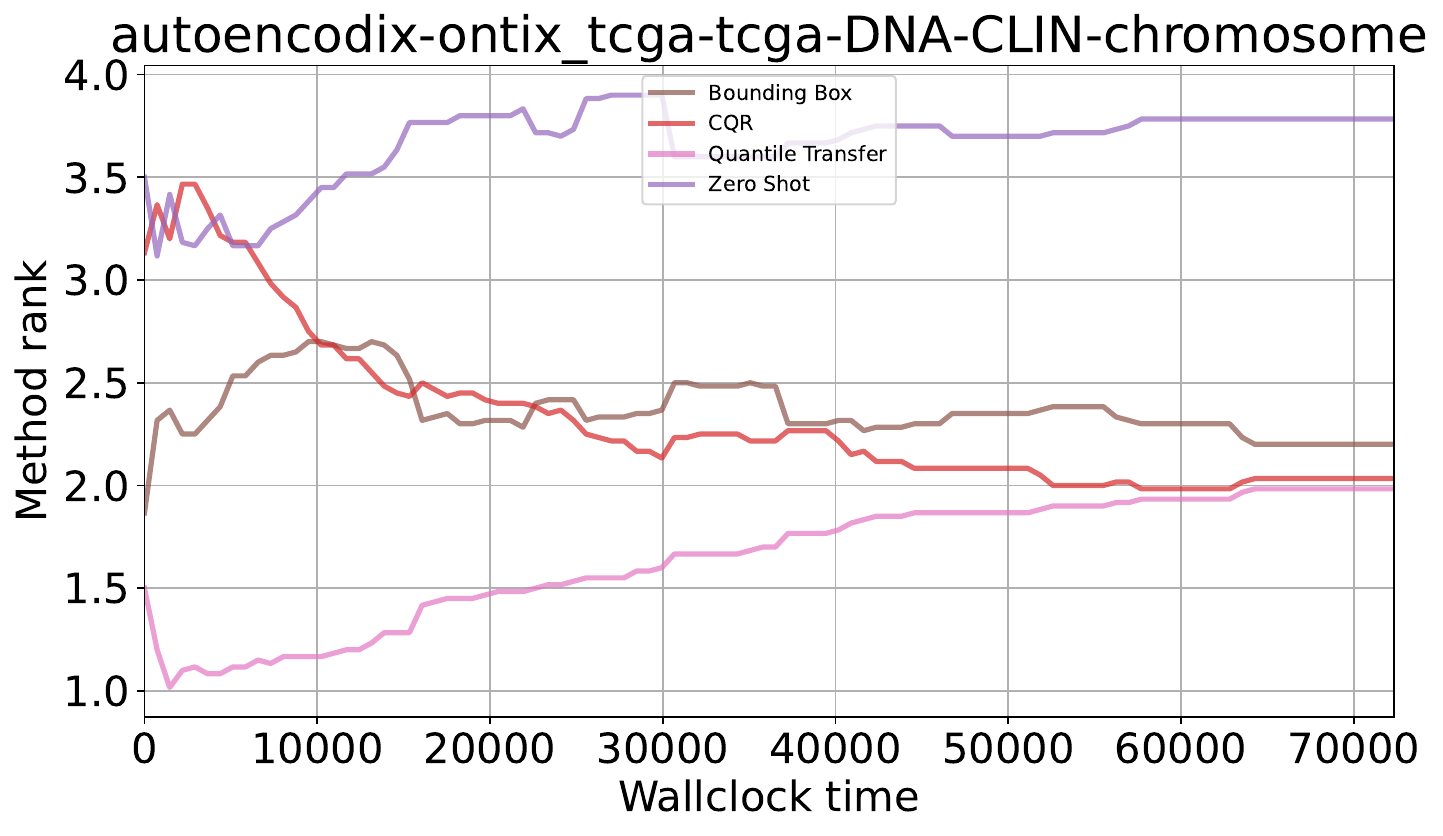} &
    \includegraphics[width=0.32\textwidth]{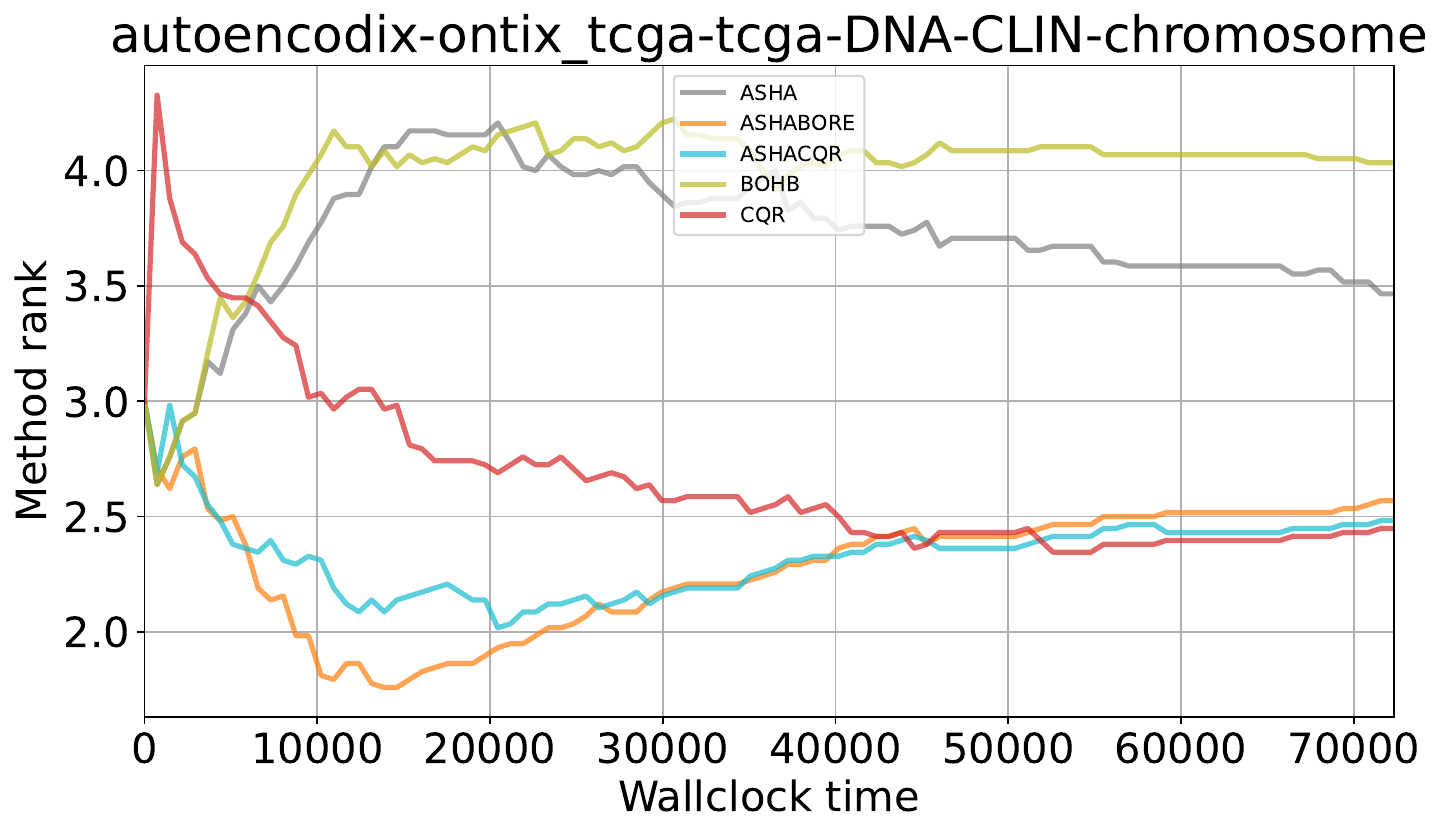} \\
    \midrule
    \multicolumn{3}{c}{\textbf{autoencodix-ontix\_tcga-tcga-DNA-CLIN-reactome}} \\
    \includegraphics[width=0.32\textwidth]{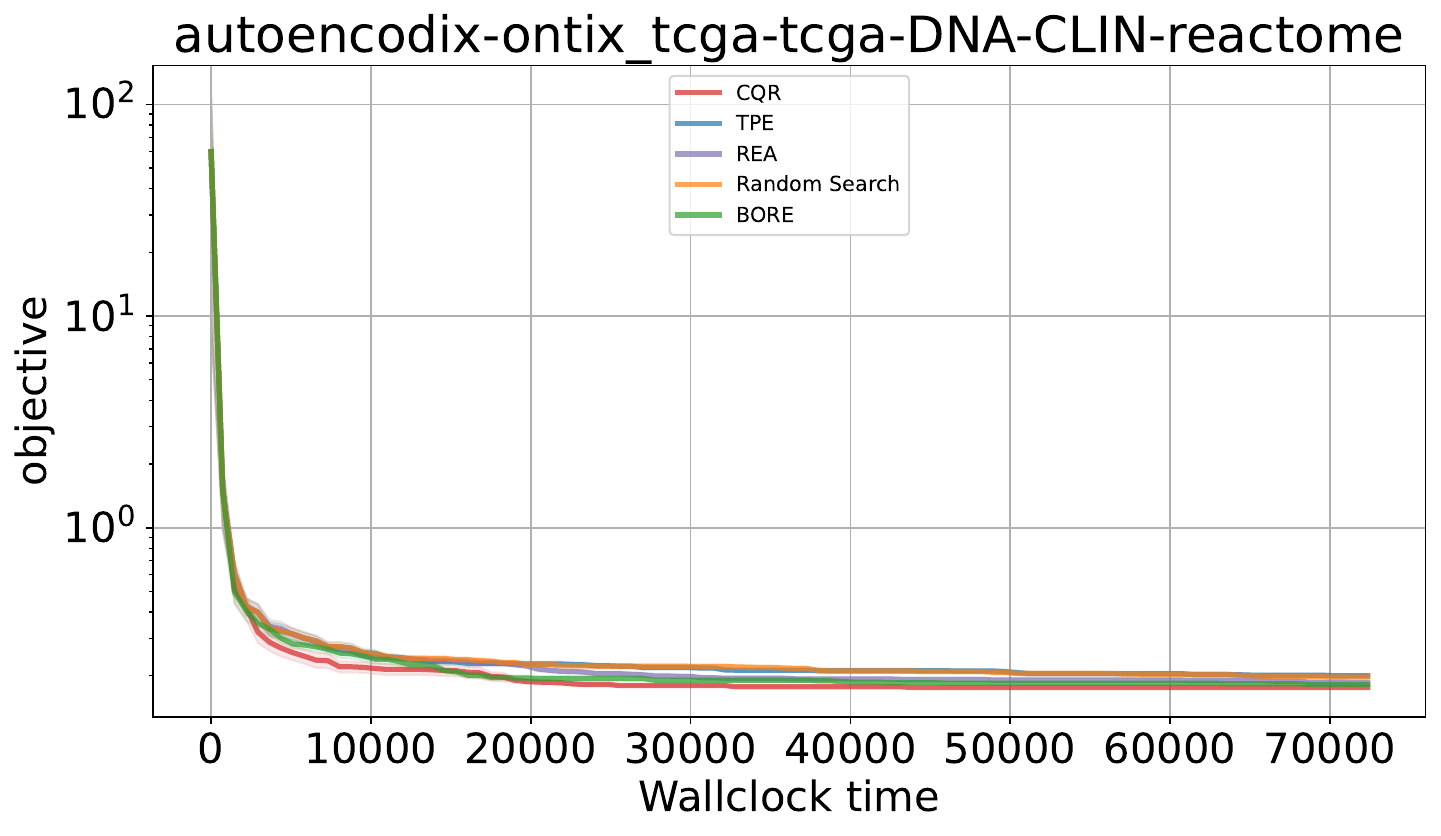} &
    \includegraphics[width=0.32\textwidth]{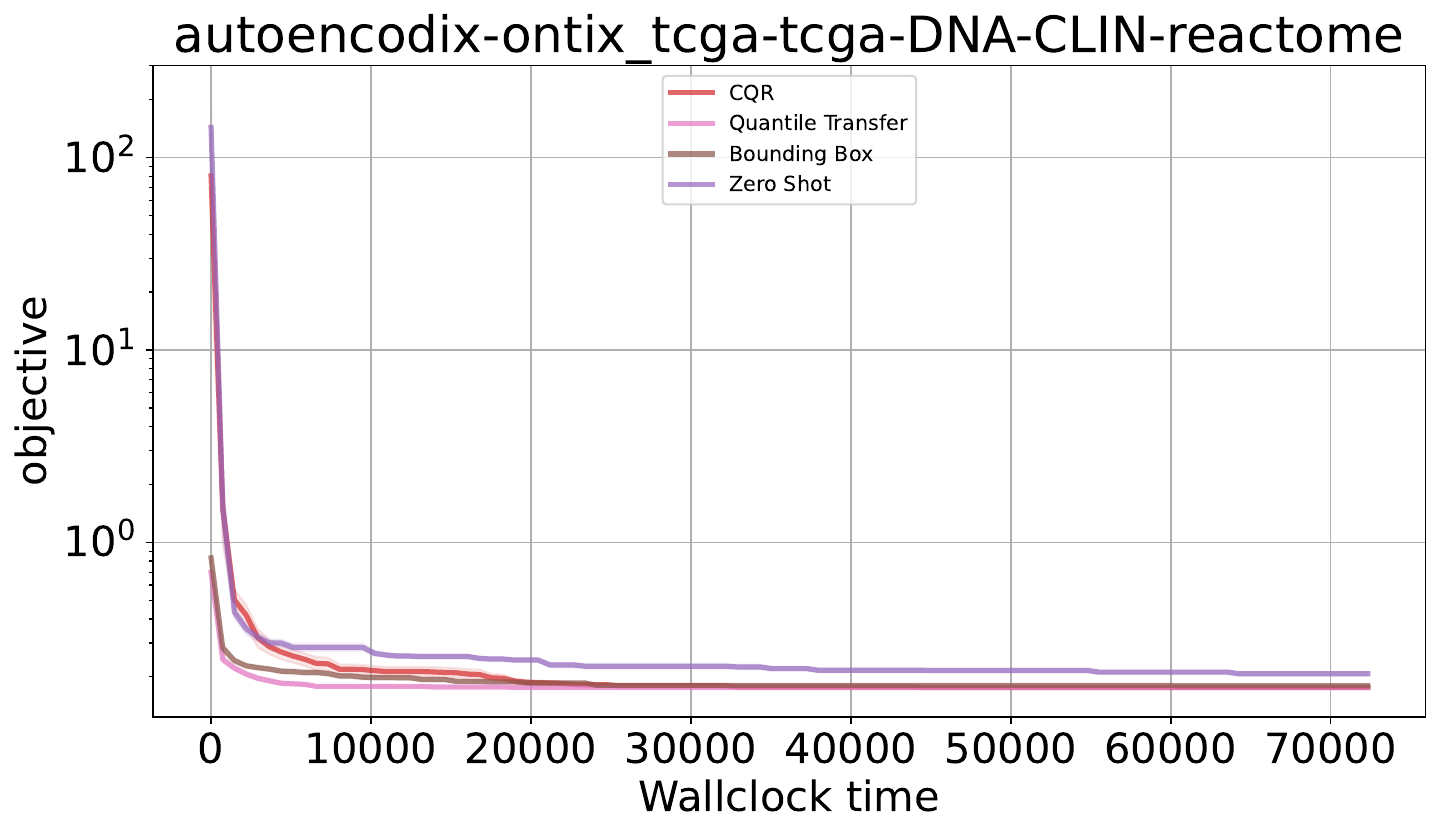} &
    \includegraphics[width=0.32\textwidth]{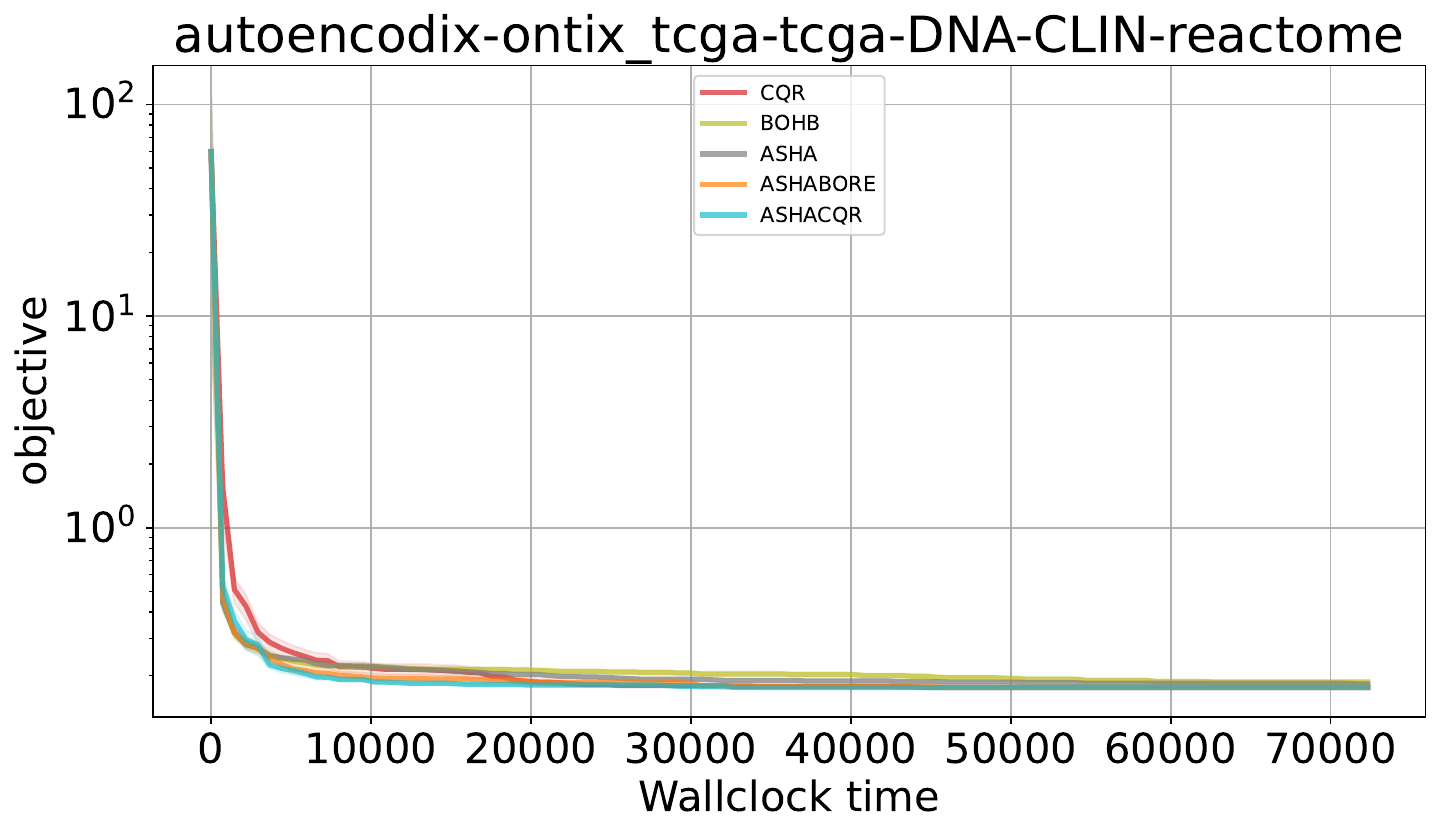} \\
    \includegraphics[width=0.32\textwidth]{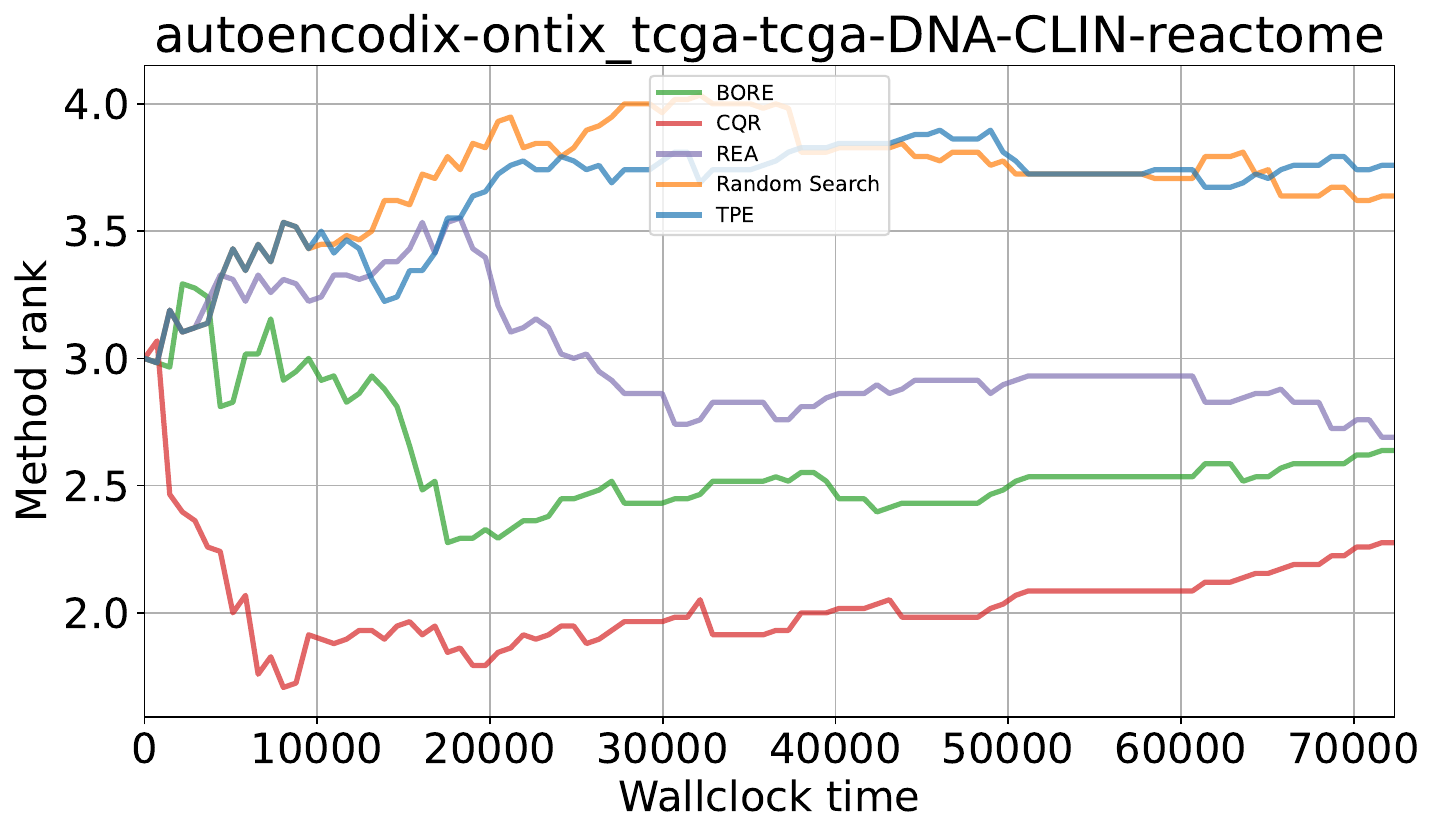} &
    \includegraphics[width=0.32\textwidth]{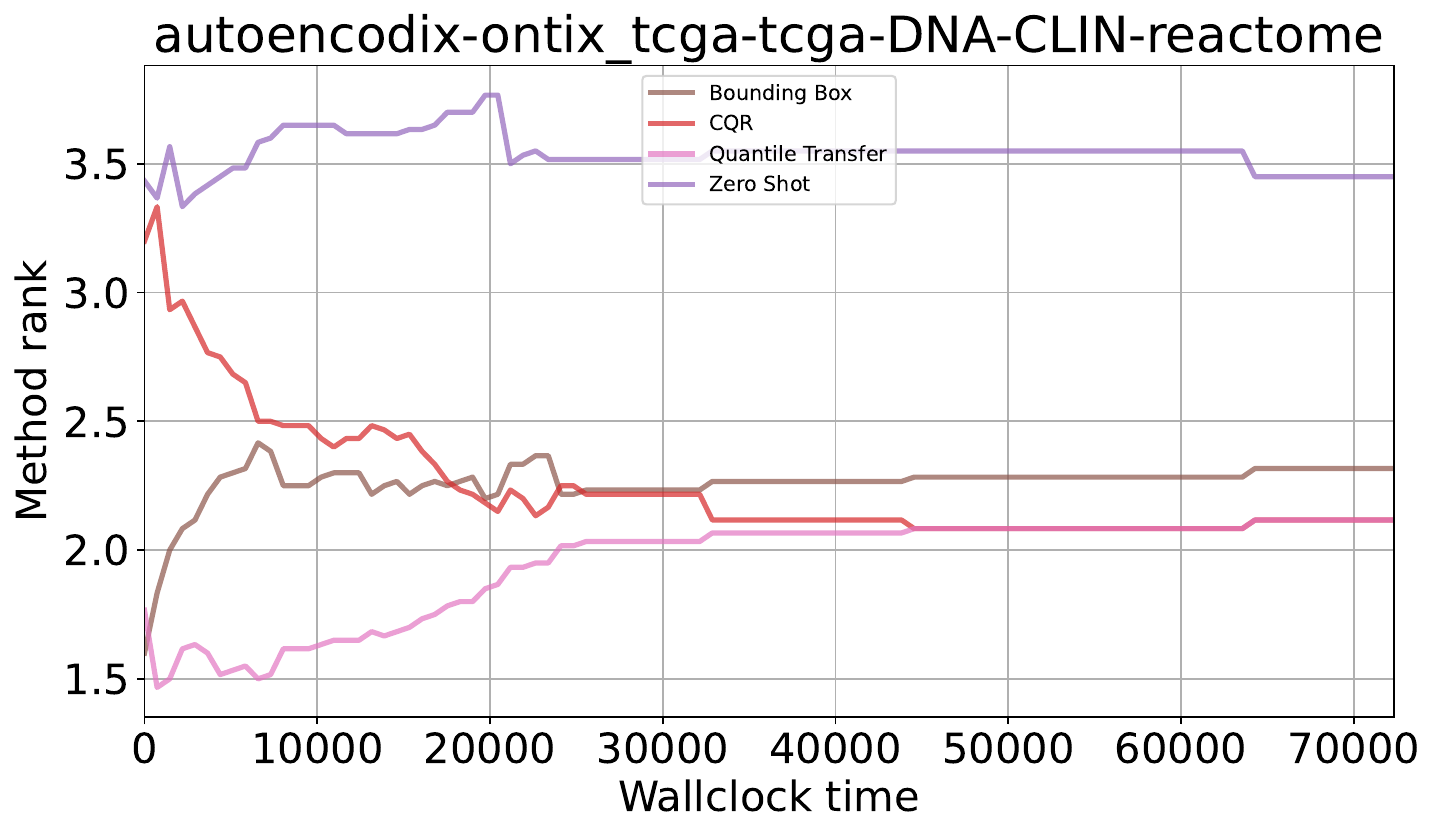} &
    \includegraphics[width=0.32\textwidth]{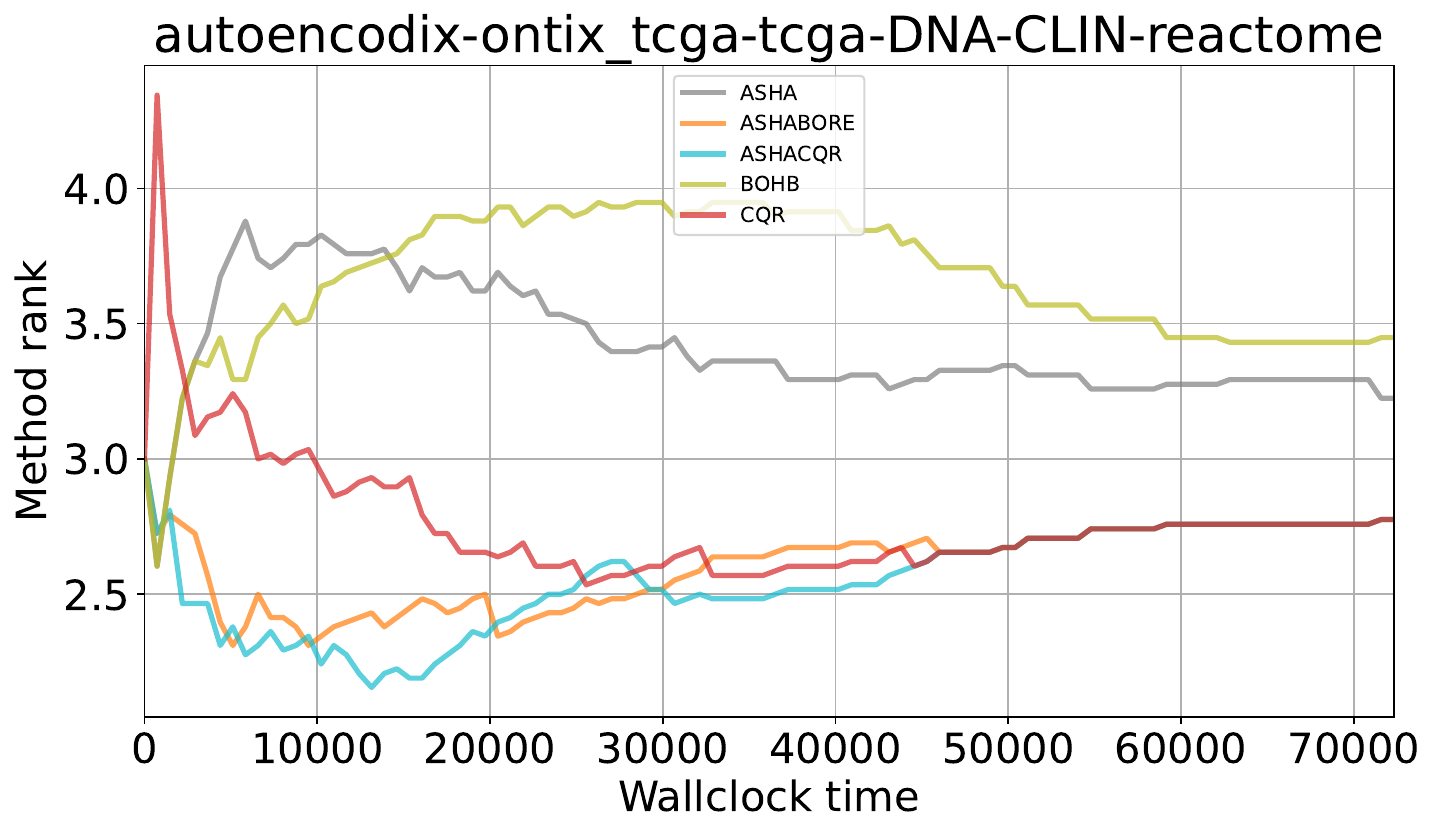} \\
    \midrule
    \multicolumn{3}{c}{\textbf{autoencodix-ontix\_tcga-tcga-METH-CLIN-chromosome}} \\
    \includegraphics[width=0.32\textwidth]{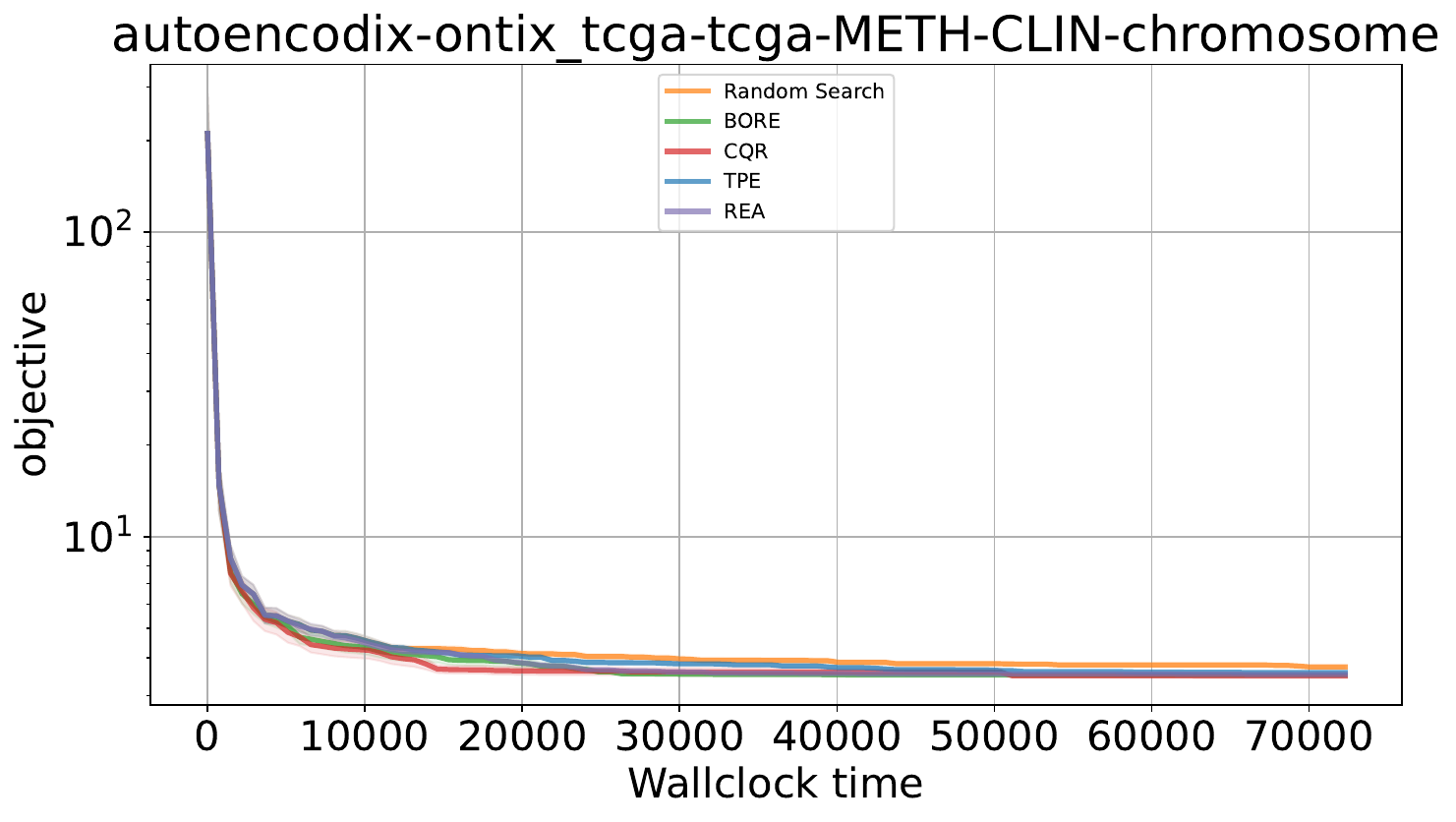} &
    \includegraphics[width=0.32\textwidth]{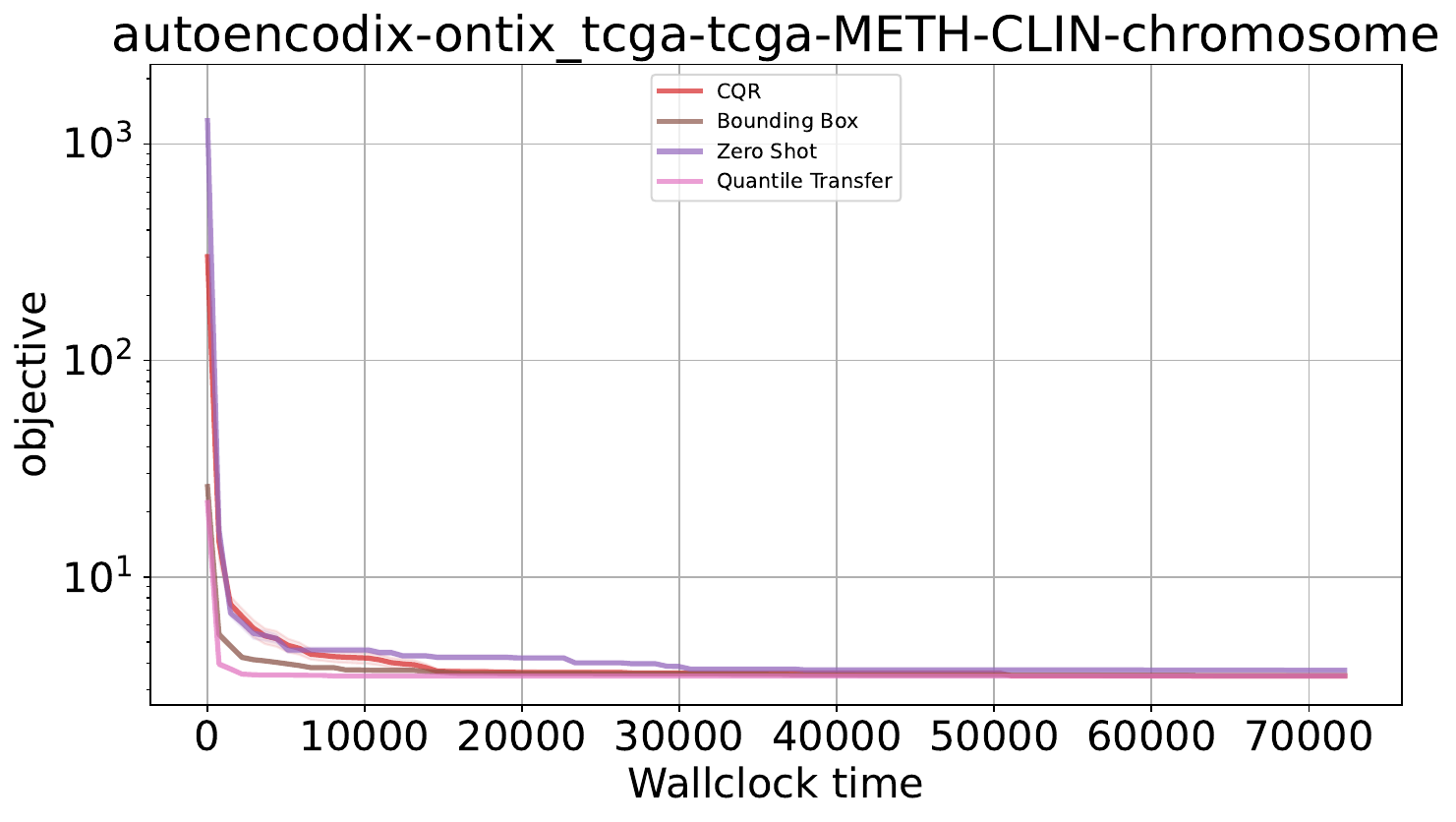} &
    \includegraphics[width=0.32\textwidth]{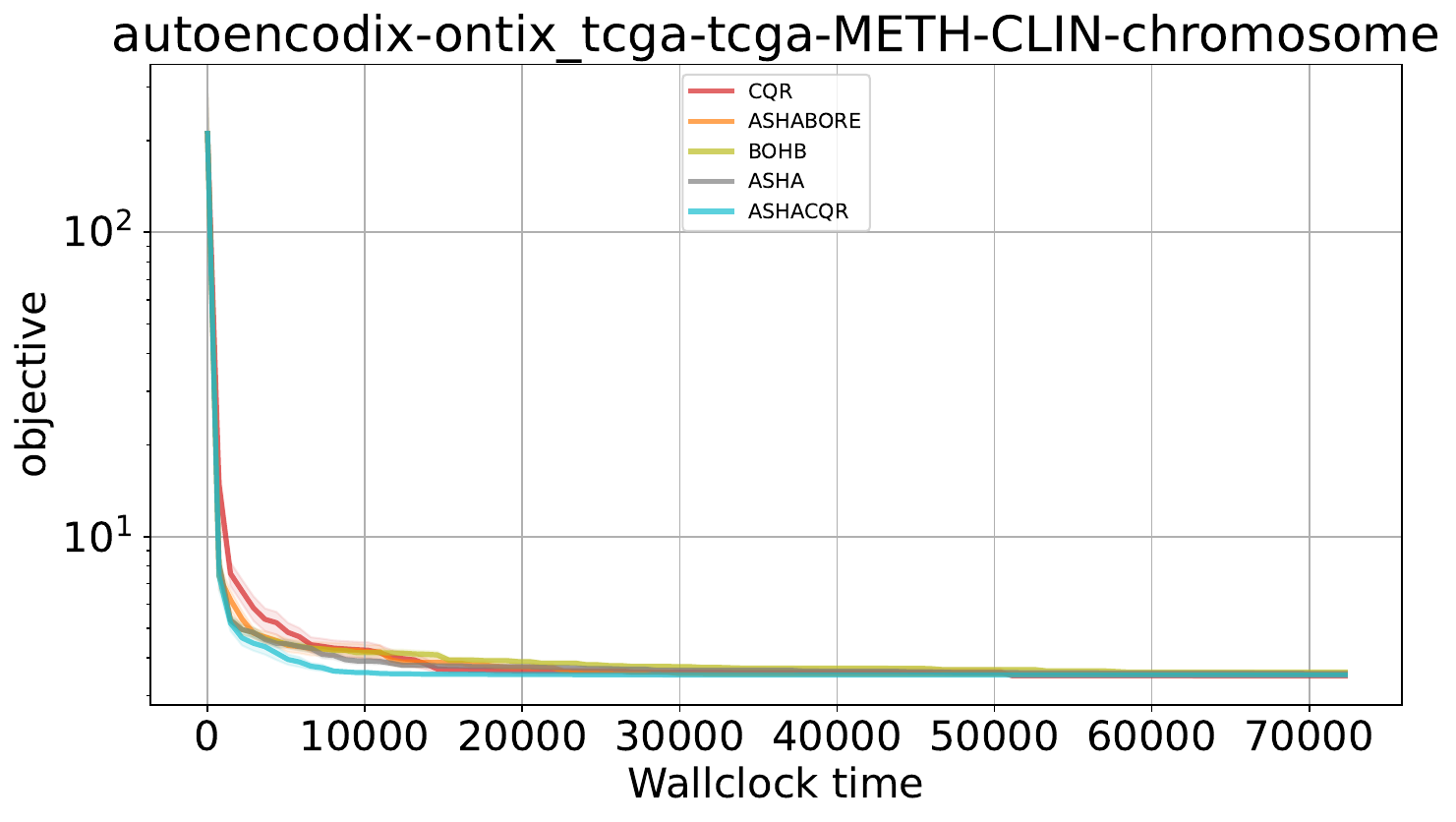} \\
    \includegraphics[width=0.32\textwidth]{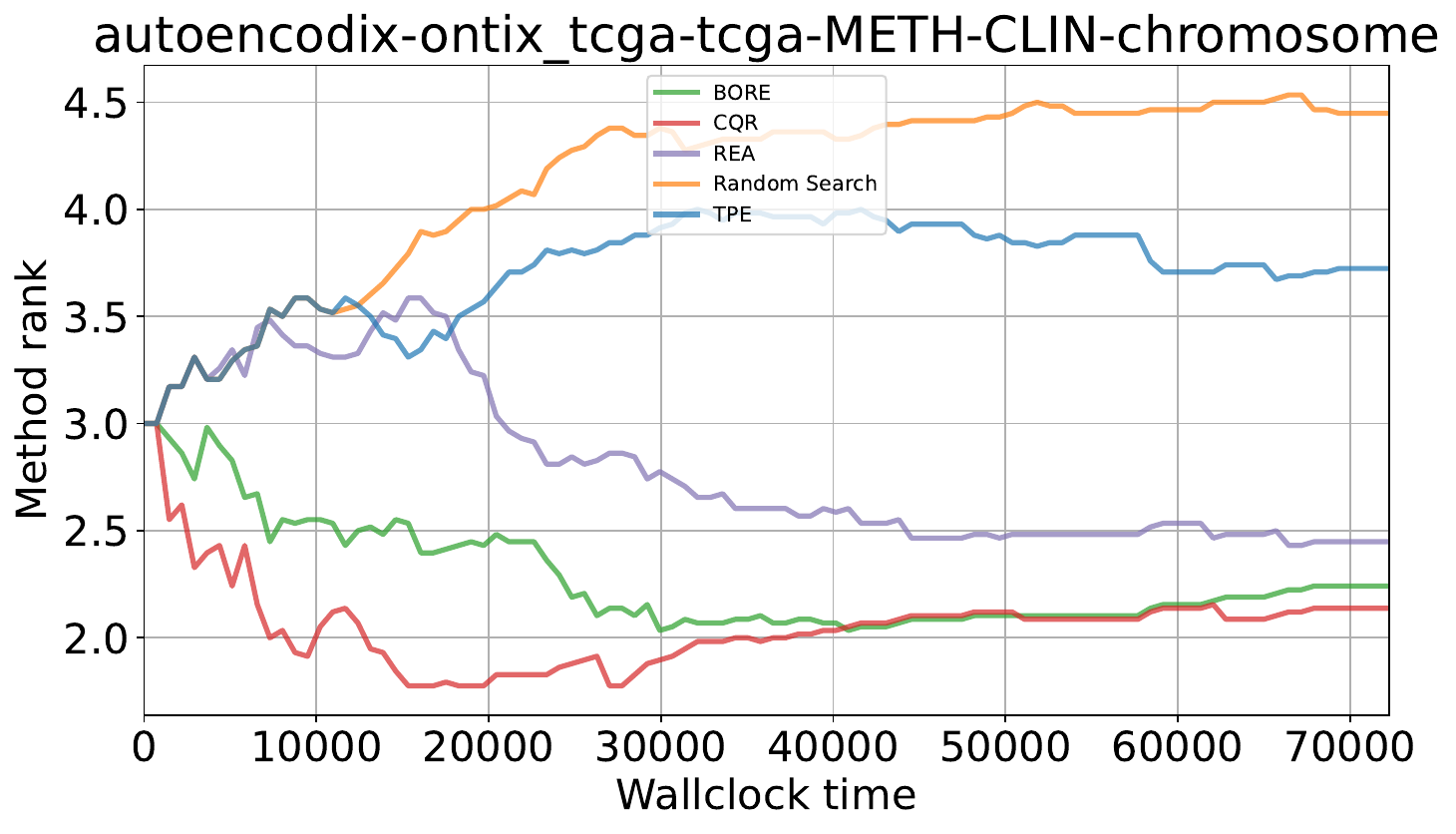} &
    \includegraphics[width=0.32\textwidth]{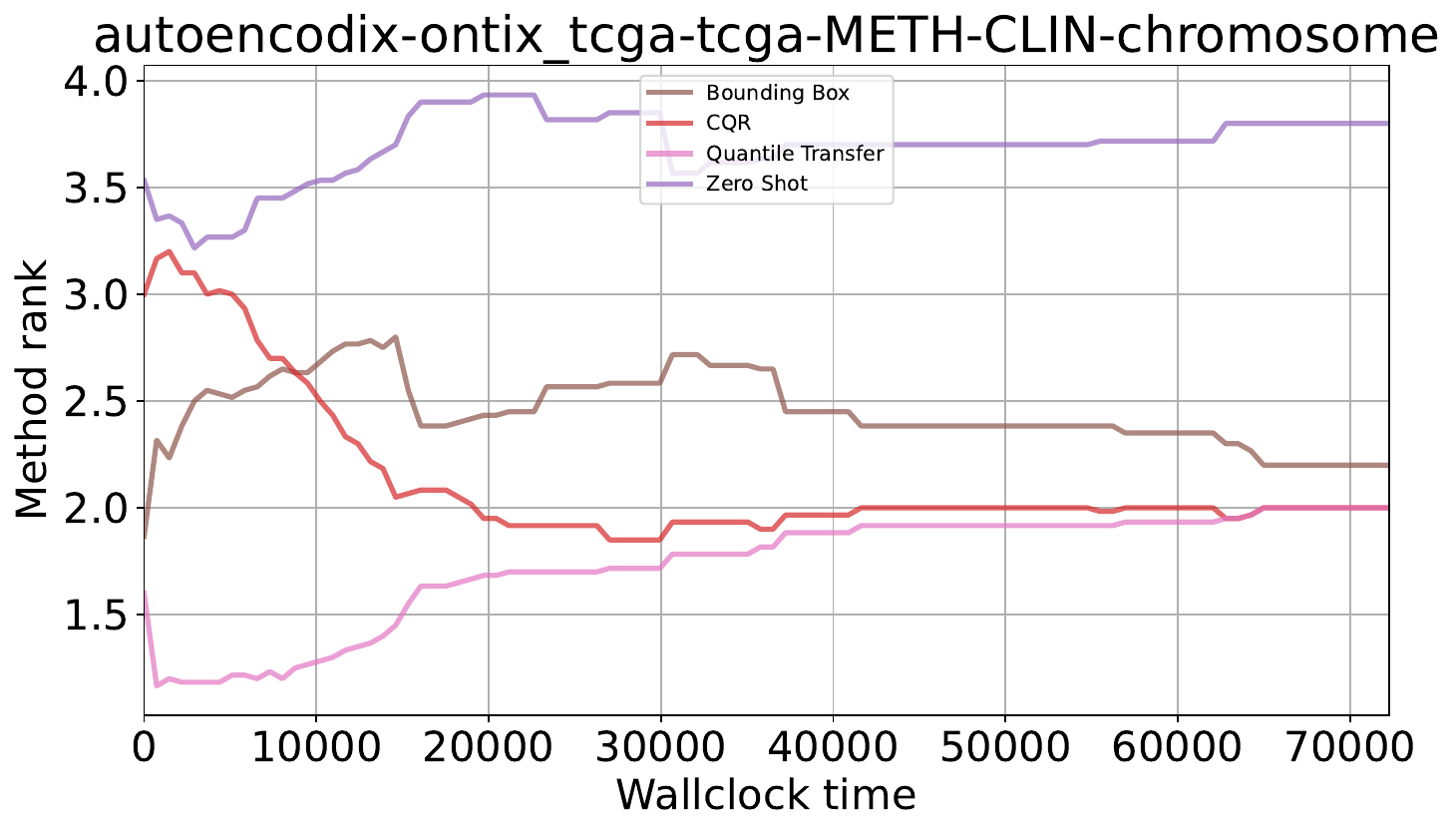} &
    \includegraphics[width=0.32\textwidth]{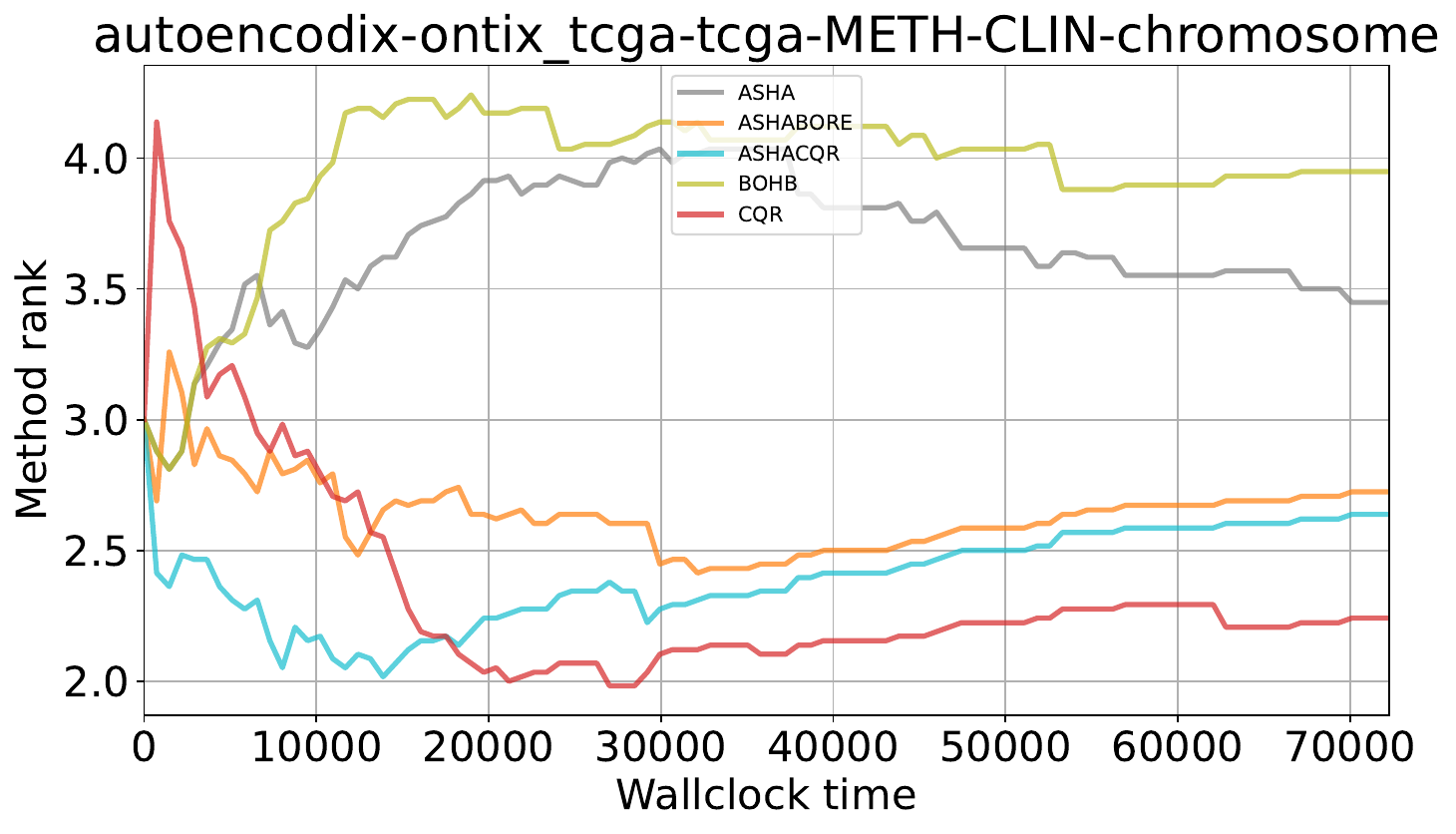} \\
    \end{tabular}
    \caption{Results for Ontix tasks (Part 3).}
    \label{fig:ontix_part3}
\end{figure}

\clearpage

\begin{figure}[htbp]
    \centering
    \setlength{\tabcolsep}{1pt}
    \begin{tabular}{ccc}
    \multicolumn{3}{c}{\textbf{autoencodix-ontix\_tcga-tcga-METH-CLIN-reactome}} \\
    \textbf{Single-Fidelity} & \textbf{Transfer Learning} & \textbf{Multi-Fidelity} \\
    \includegraphics[width=0.32\textwidth]{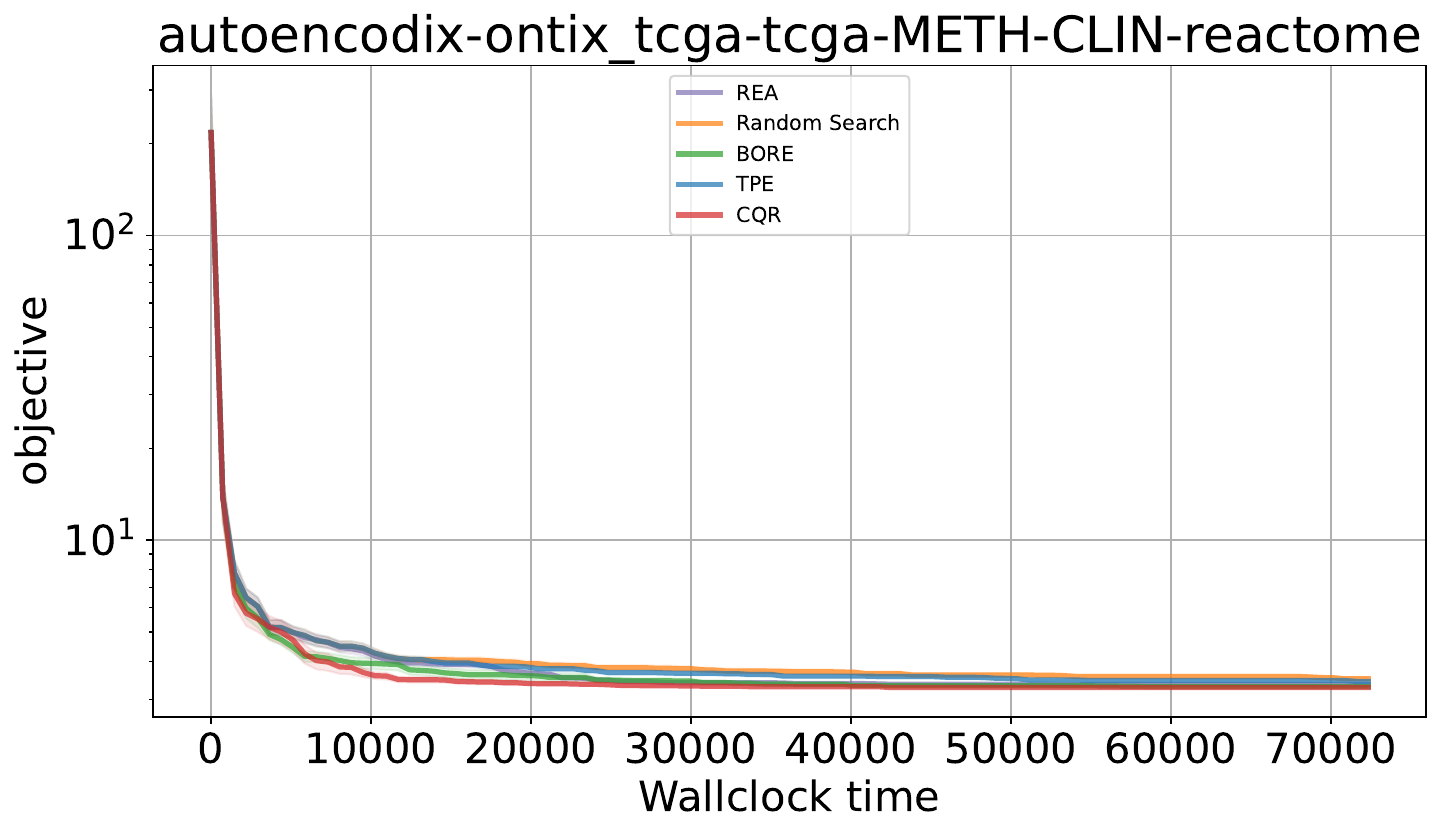} &
    \includegraphics[width=0.32\textwidth]{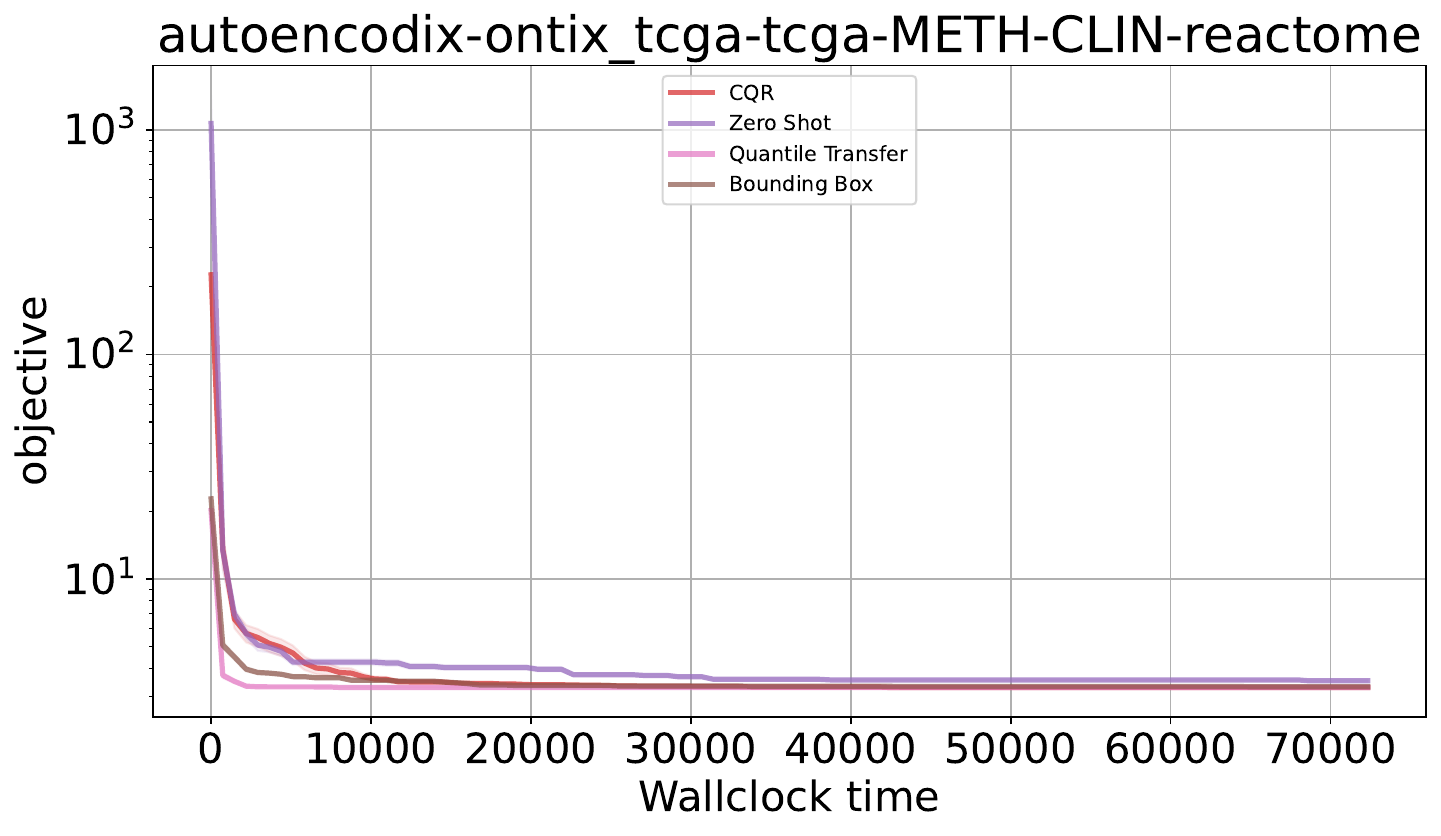} &
    \includegraphics[width=0.32\textwidth]{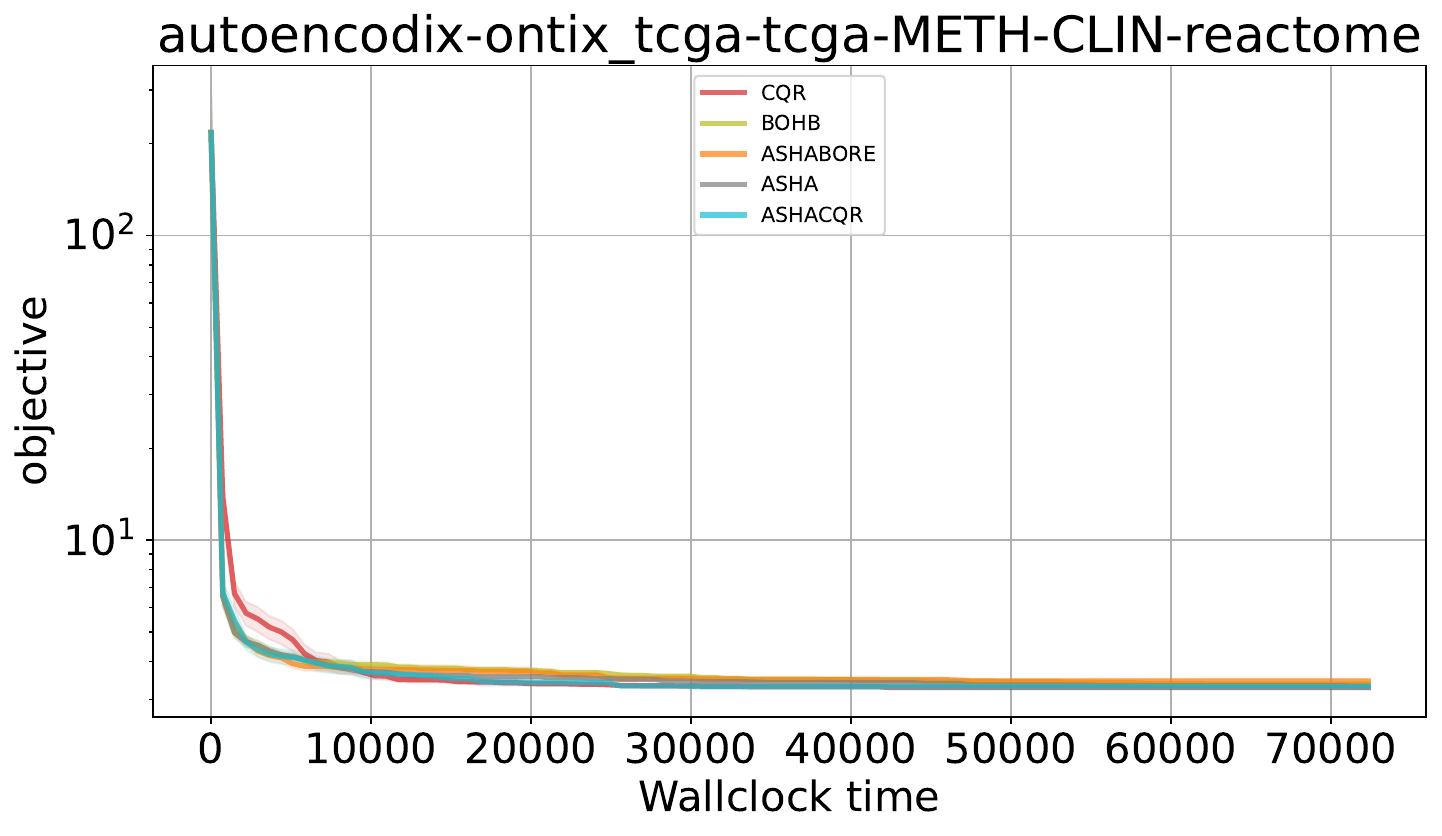} \\
    \includegraphics[width=0.32\textwidth]{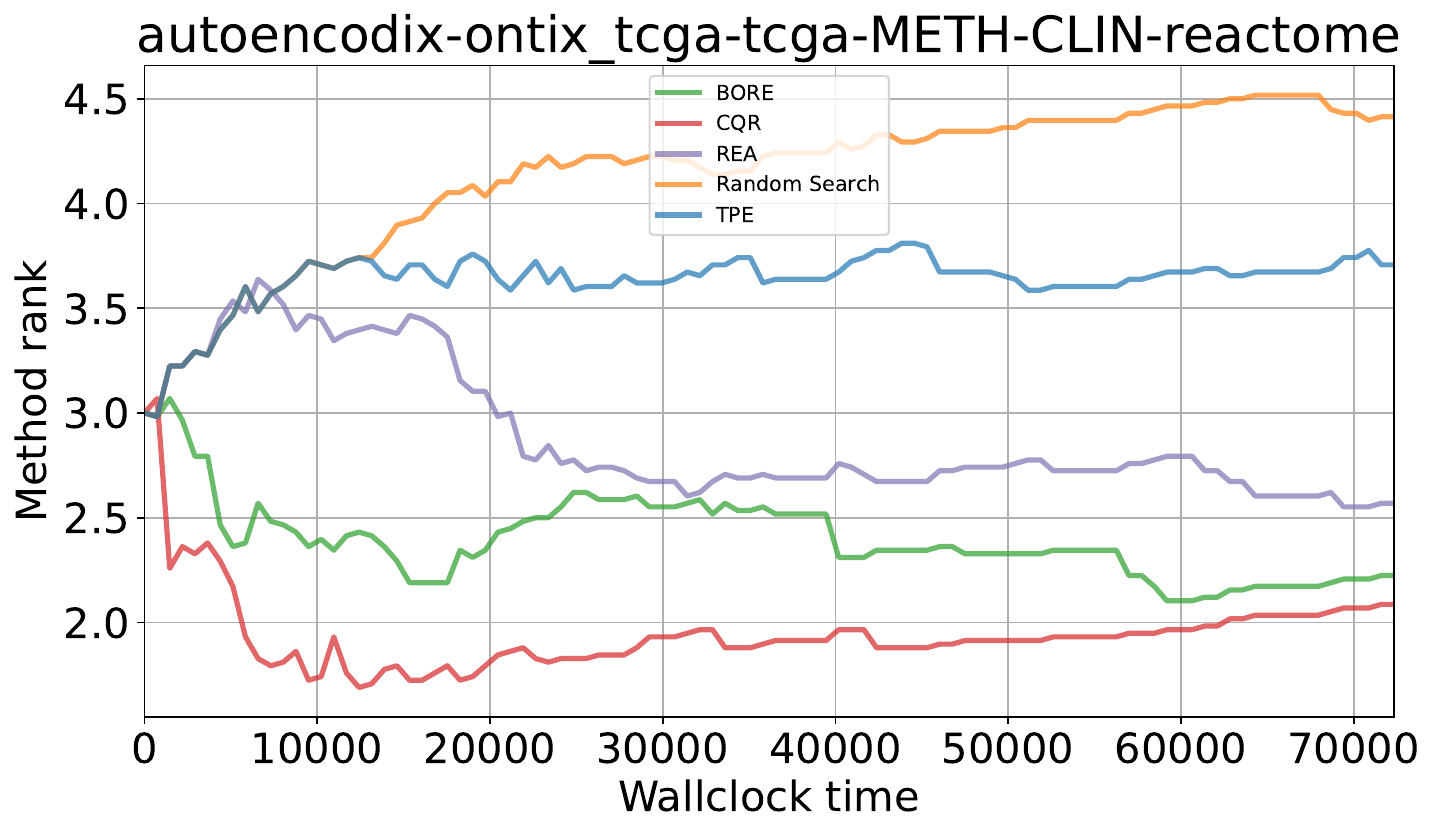} &
    \includegraphics[width=0.32\textwidth]{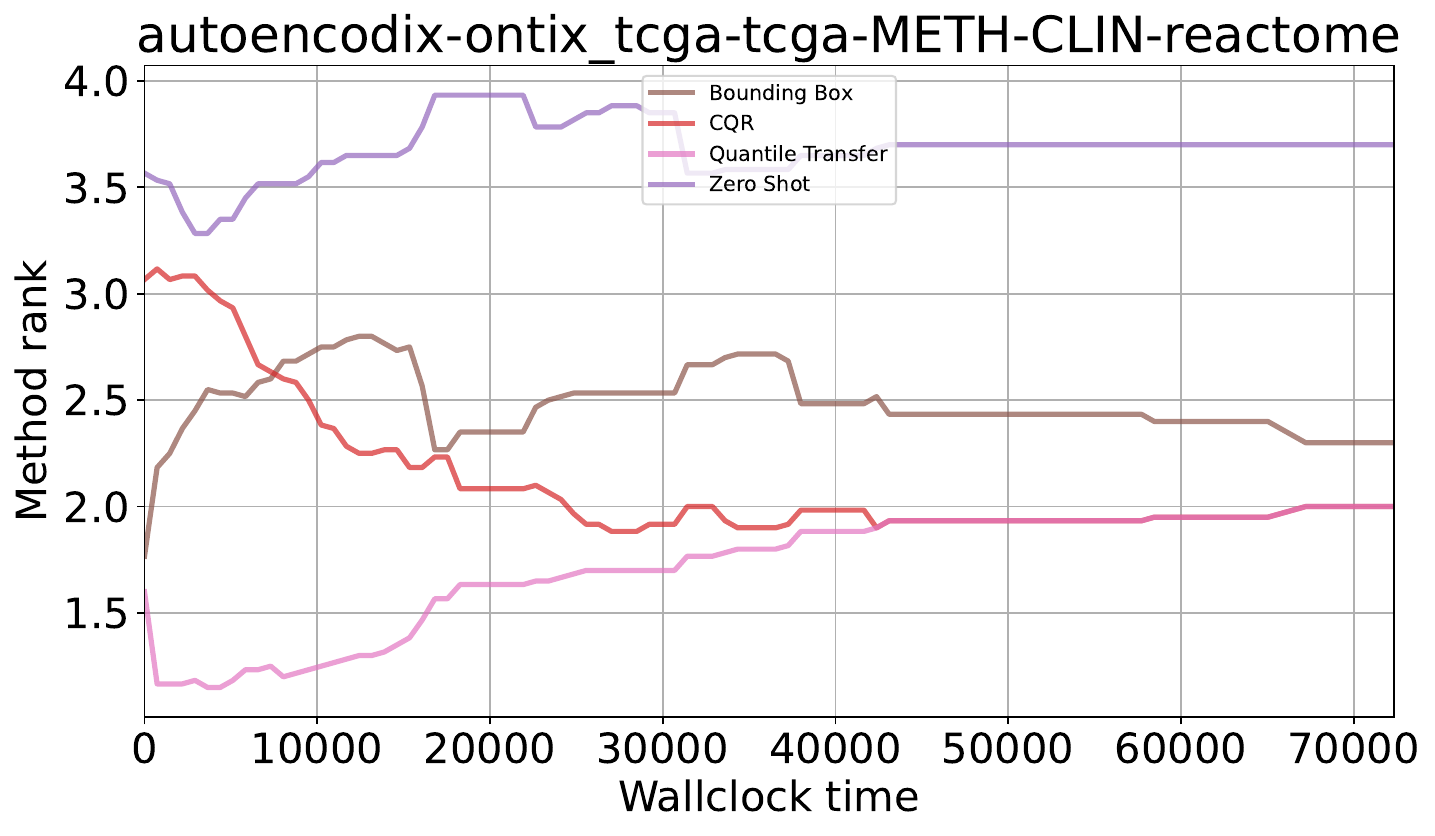} &
    \includegraphics[width=0.32\textwidth]{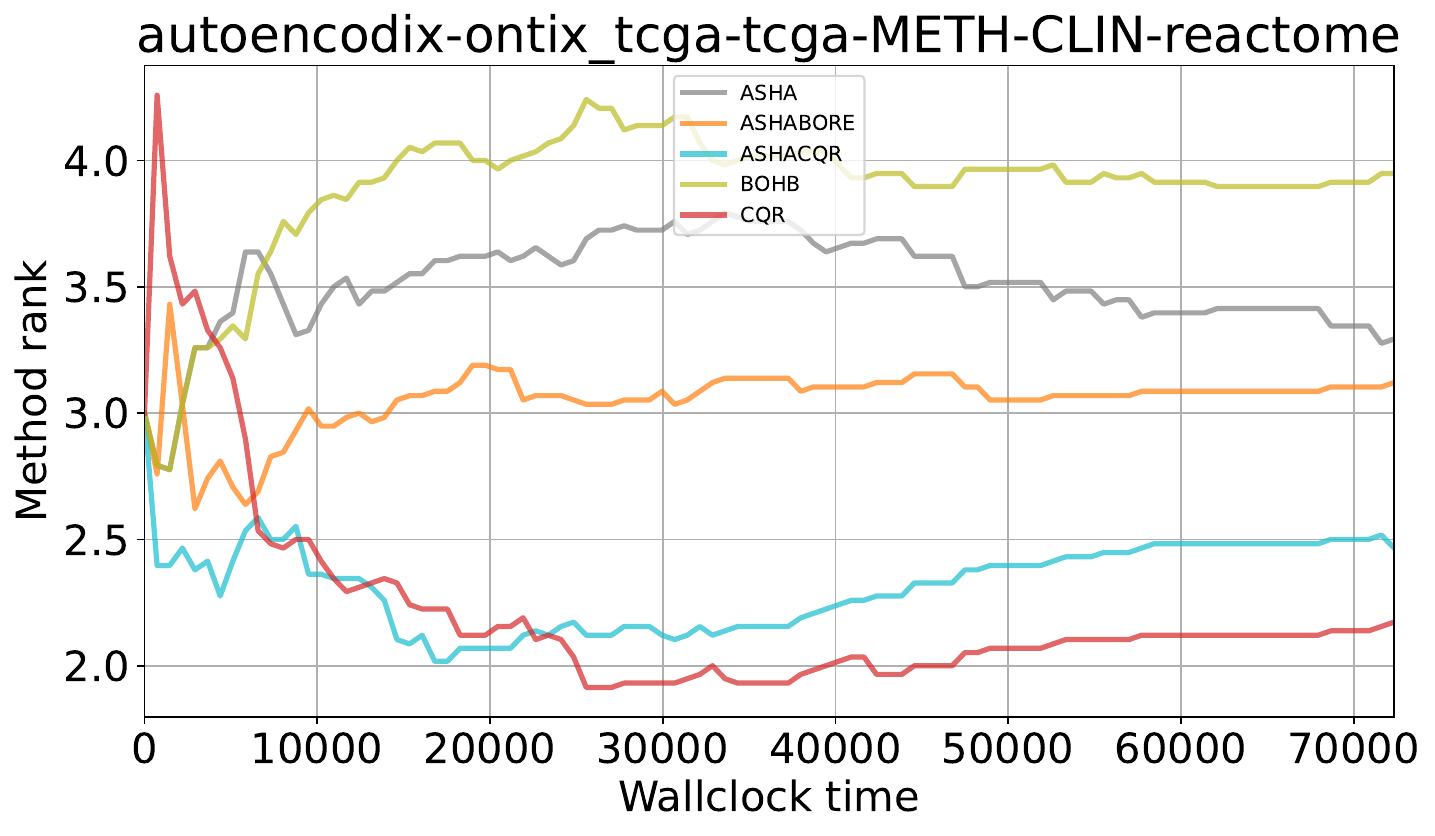} \\
    \midrule
    \multicolumn{3}{c}{\textbf{autoencodix-ontix\_tcga-tcga-RNA-CLIN-chromosome}} \\
    \includegraphics[width=0.32\textwidth]{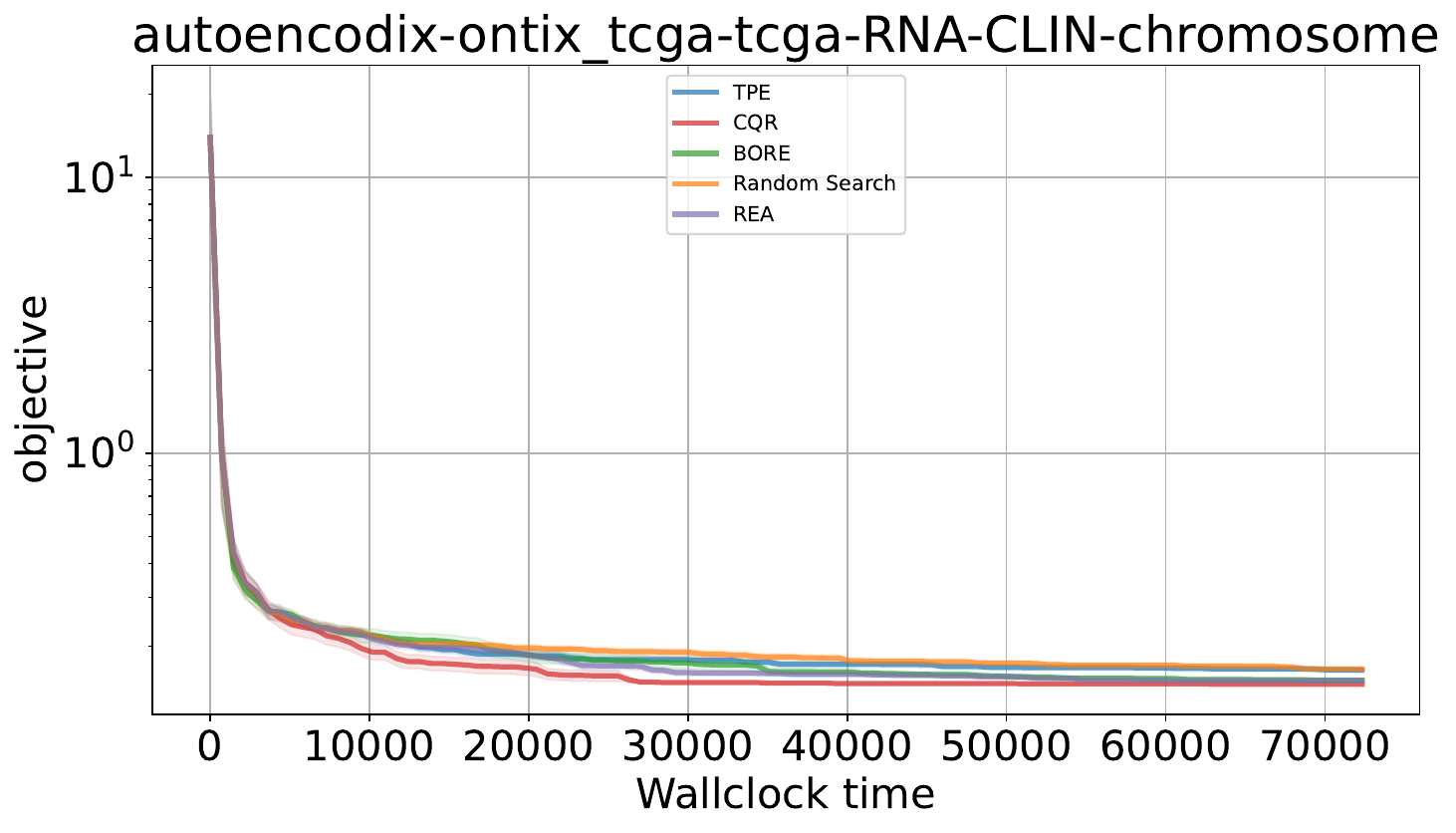} &
    \includegraphics[width=0.32\textwidth]{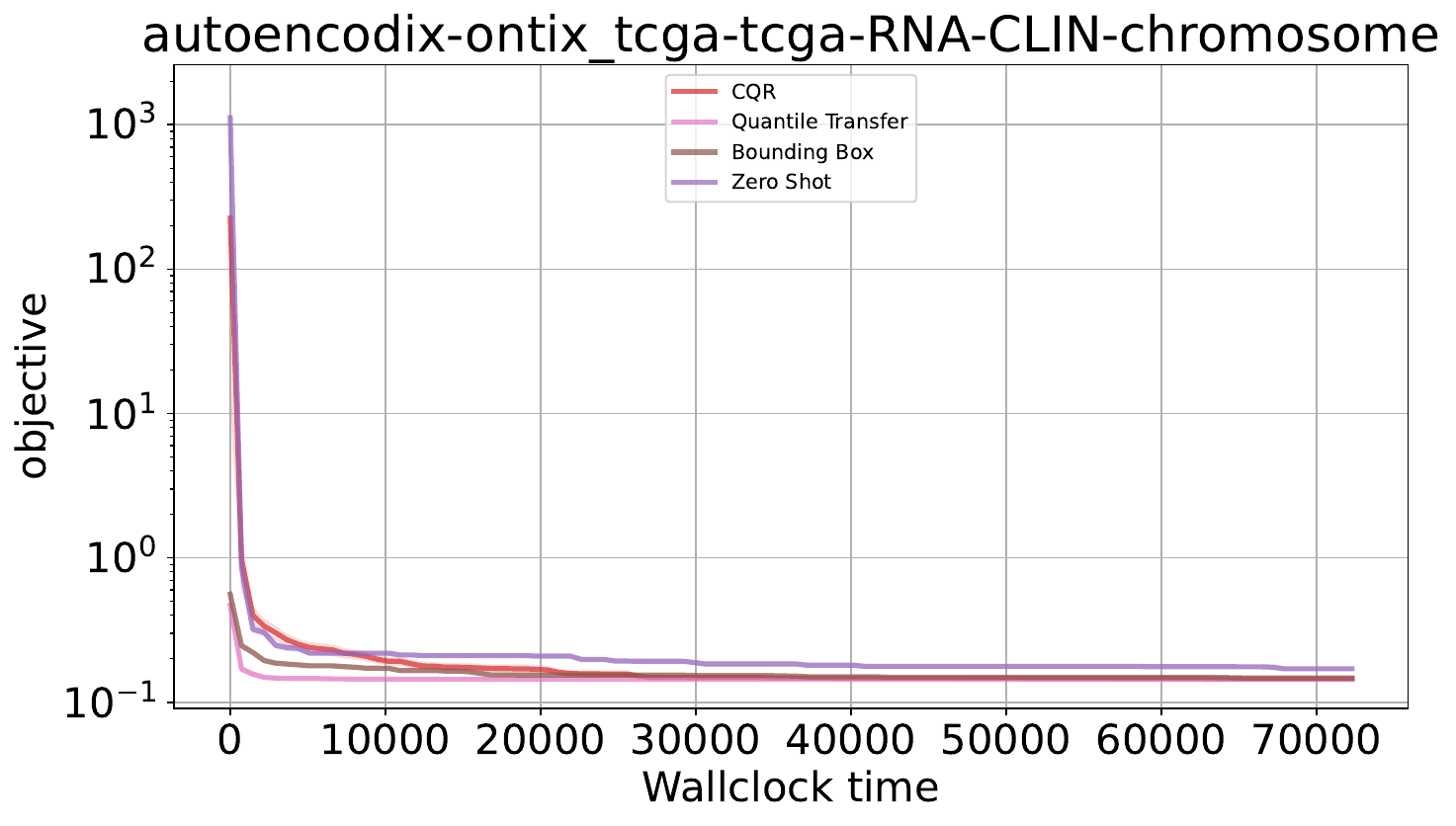} &
    \includegraphics[width=0.32\textwidth]{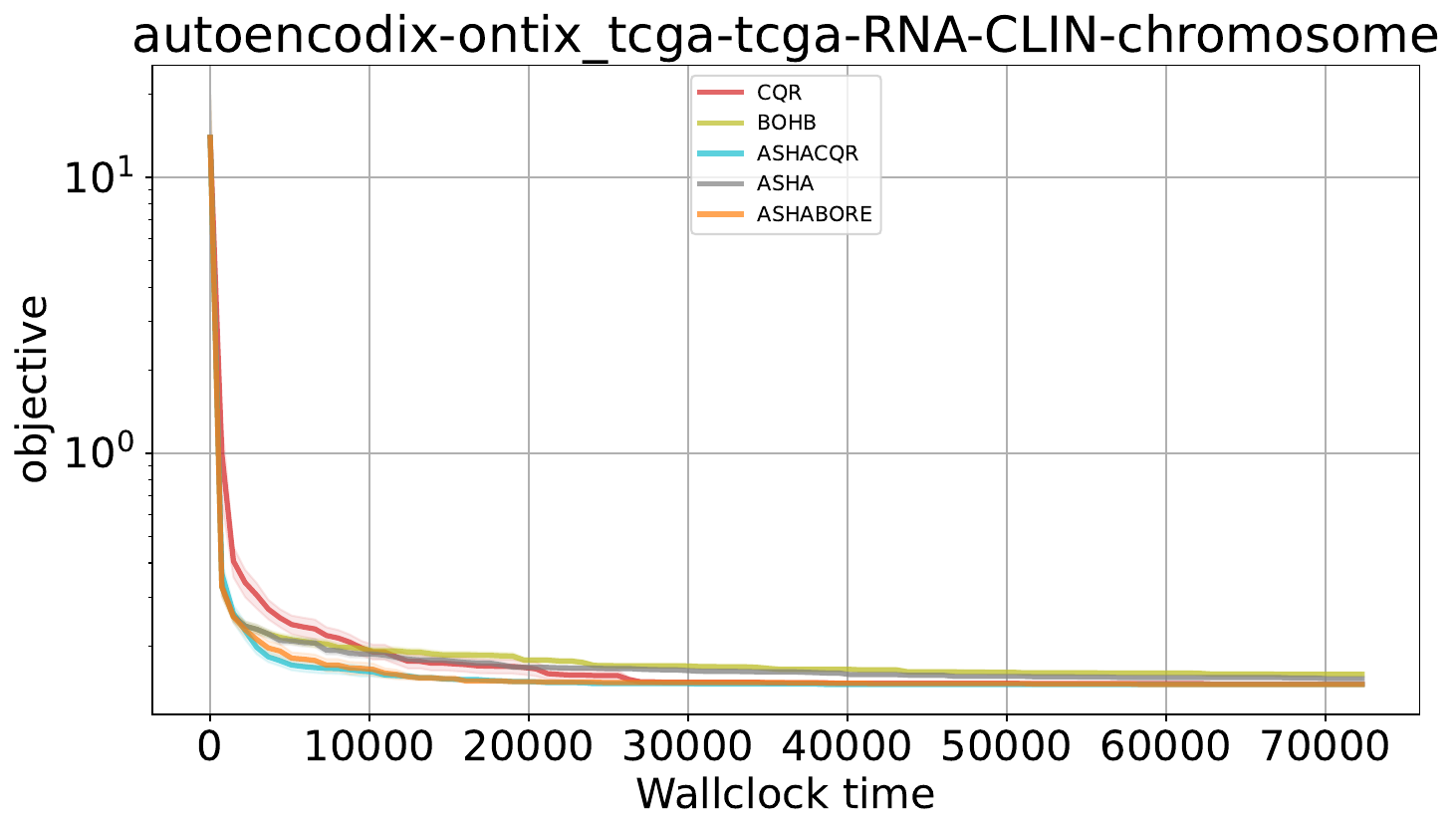} \\
    \includegraphics[width=0.32\textwidth]{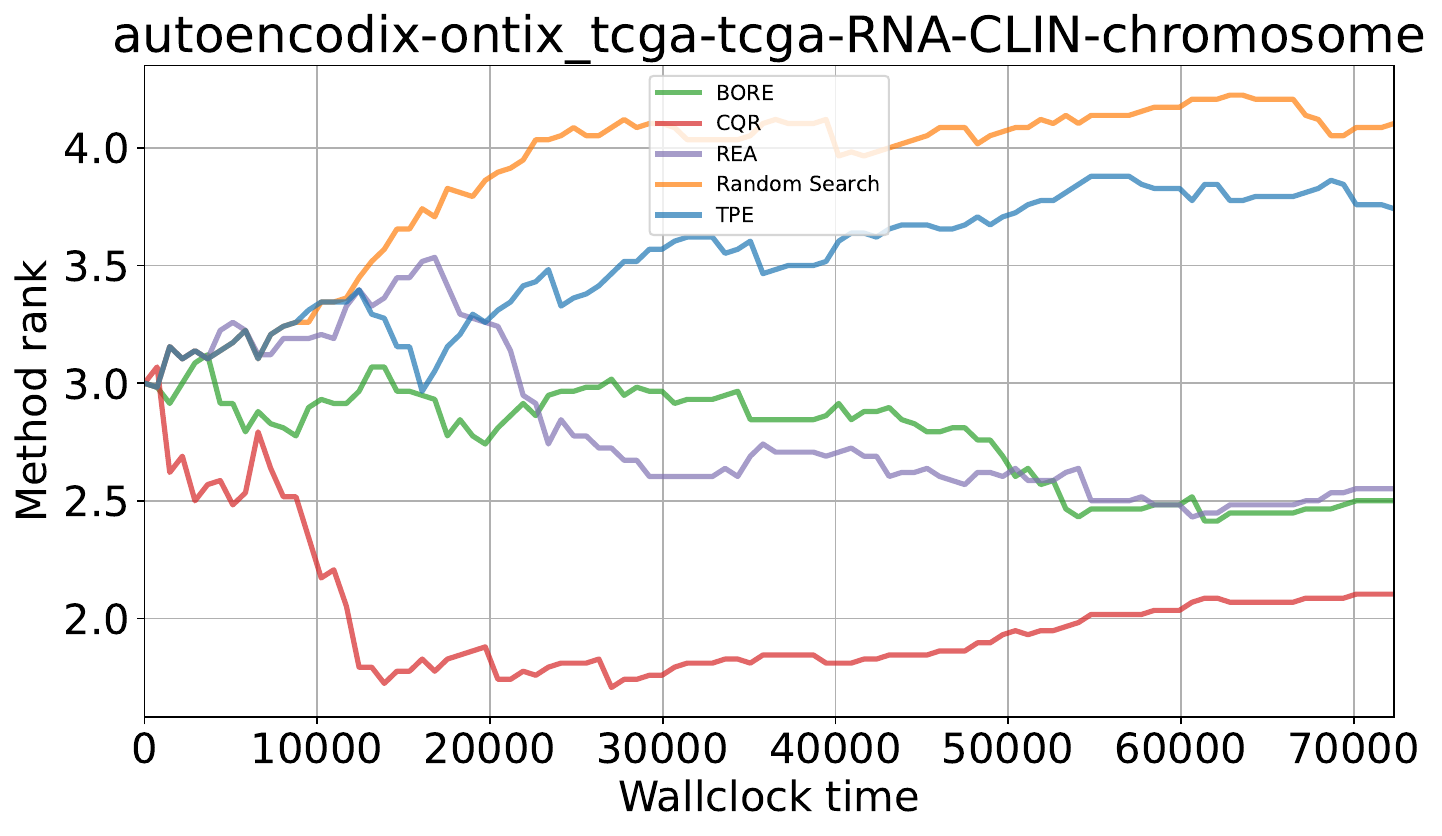} &
    \includegraphics[width=0.32\textwidth]{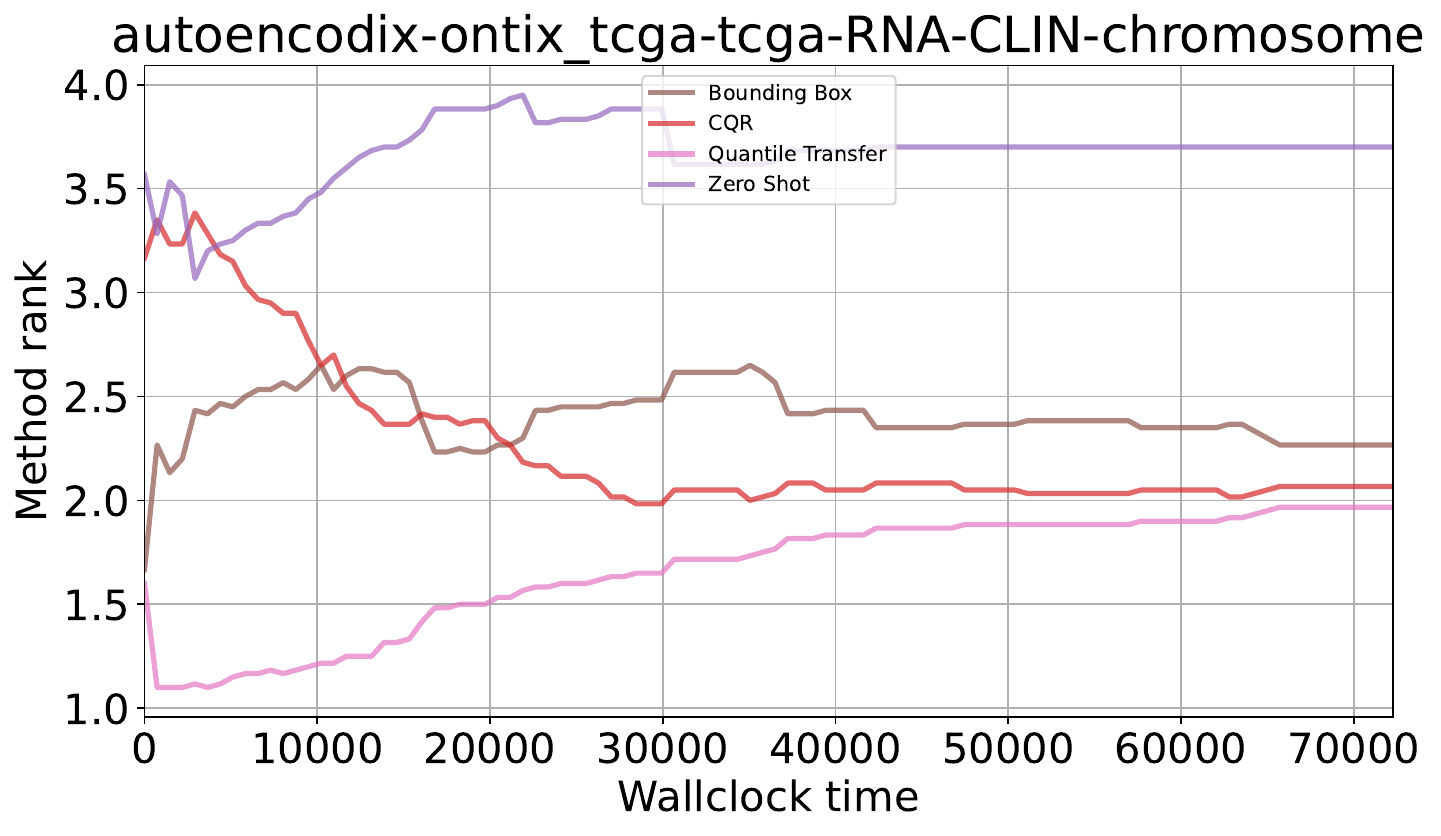} &
    \includegraphics[width=0.32\textwidth]{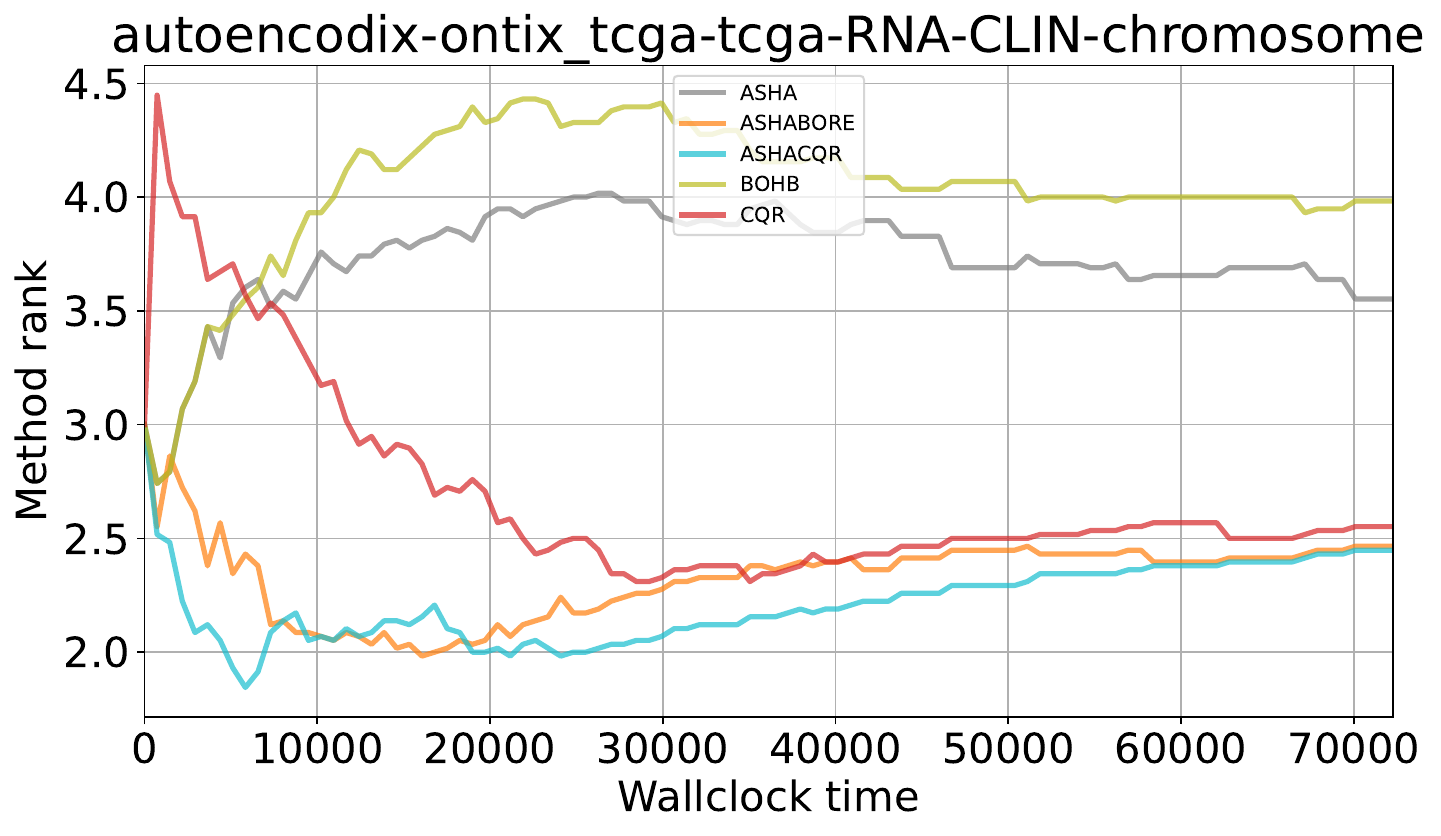} \\
    \midrule
    \multicolumn{3}{c}{\textbf{autoencodix-ontix\_tcga-tcga-RNA-CLIN-reactome}} \\
    \includegraphics[width=0.32\textwidth]{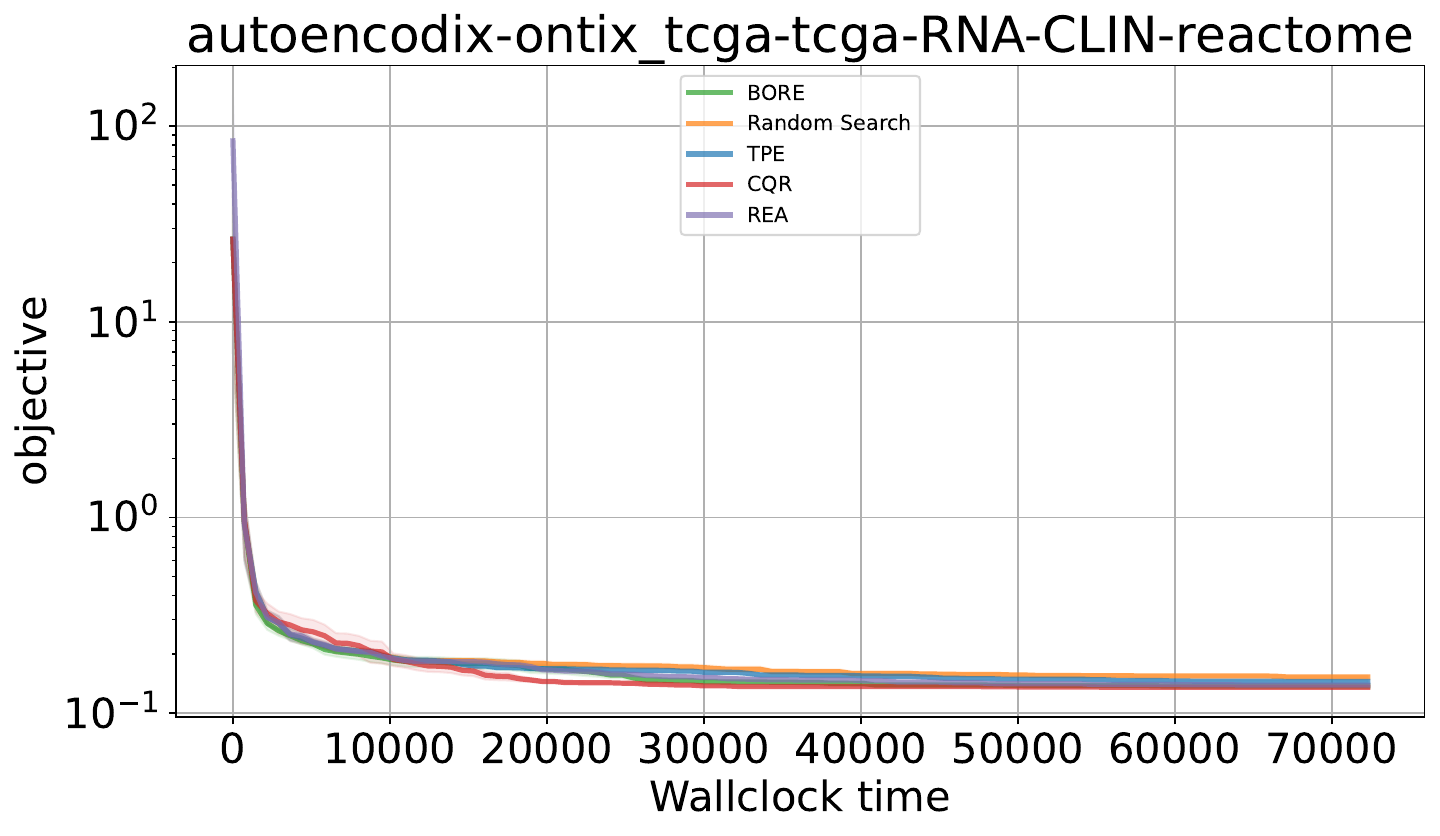} &
    \includegraphics[width=0.32\textwidth]{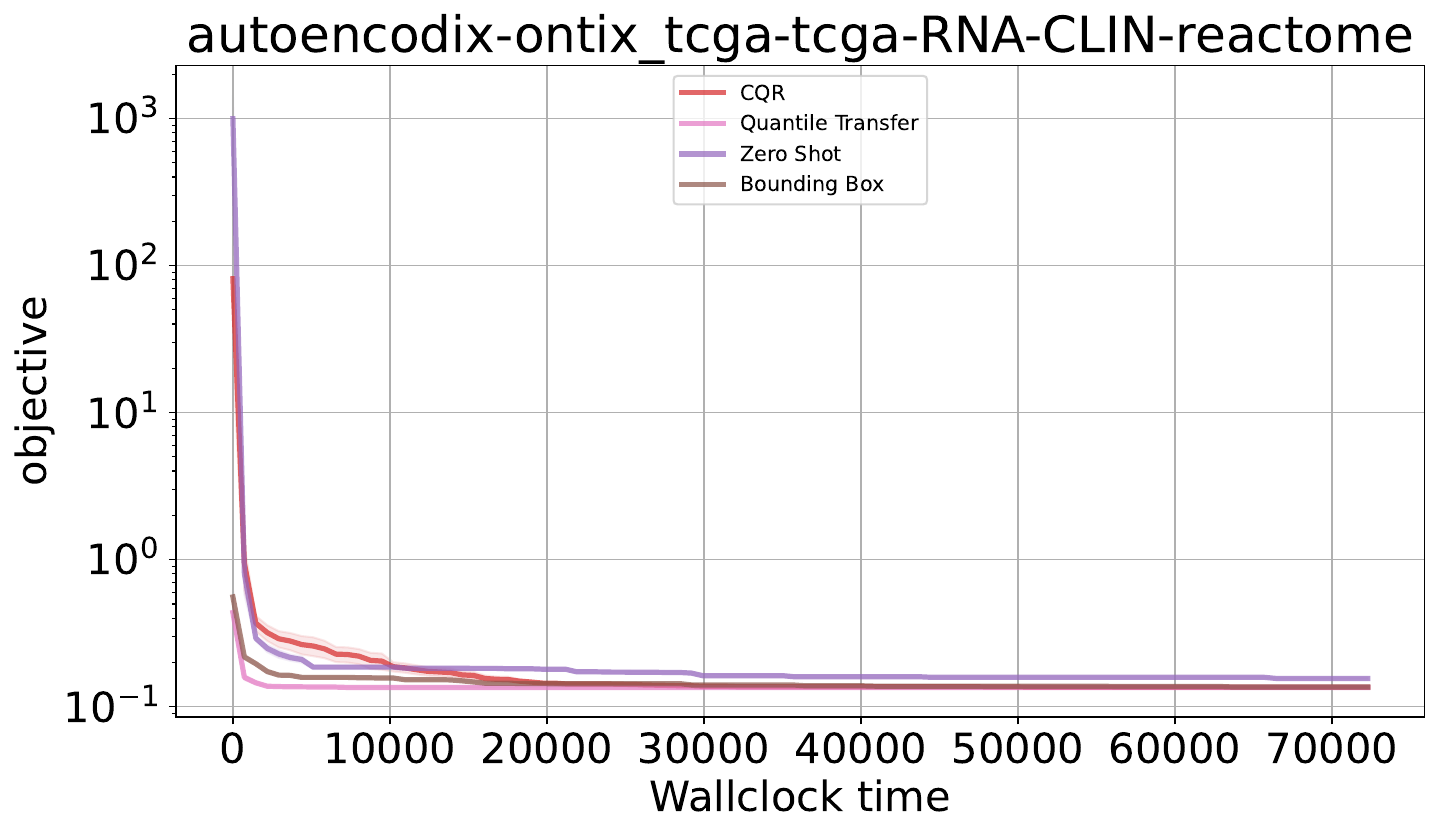} &
    \includegraphics[width=0.32\textwidth]{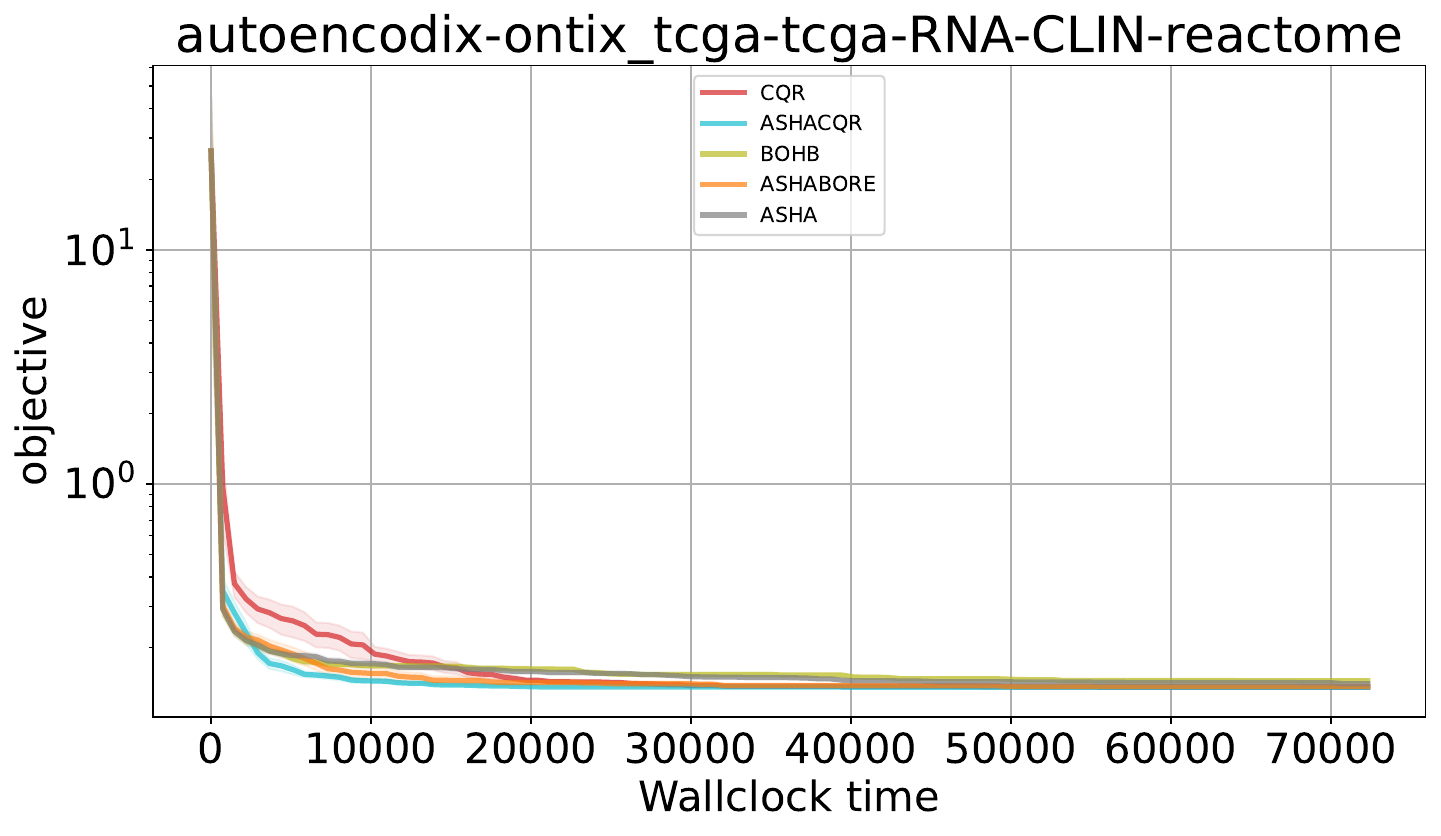} \\
    \includegraphics[width=0.32\textwidth]{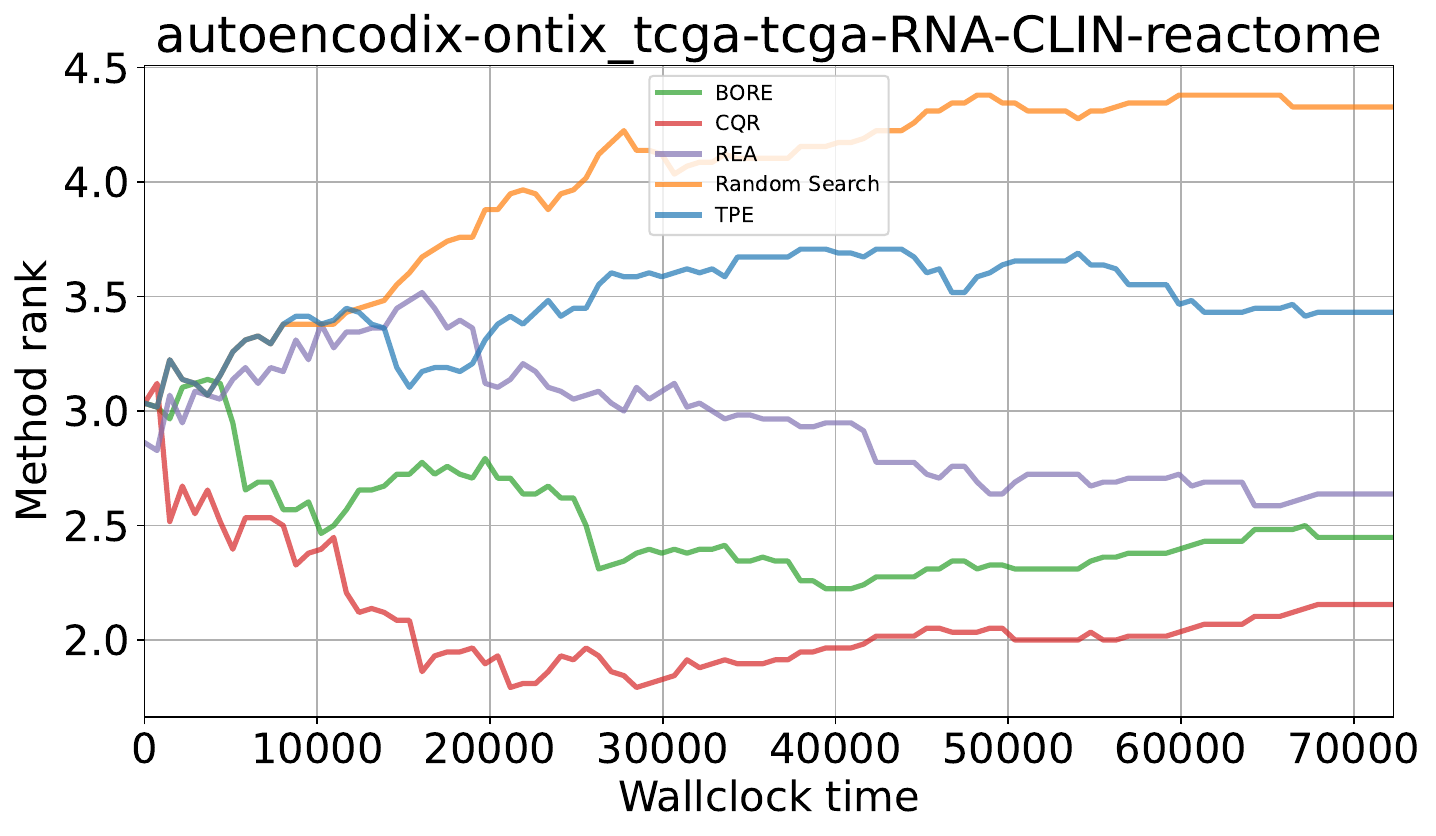} &
    \includegraphics[width=0.32\textwidth]{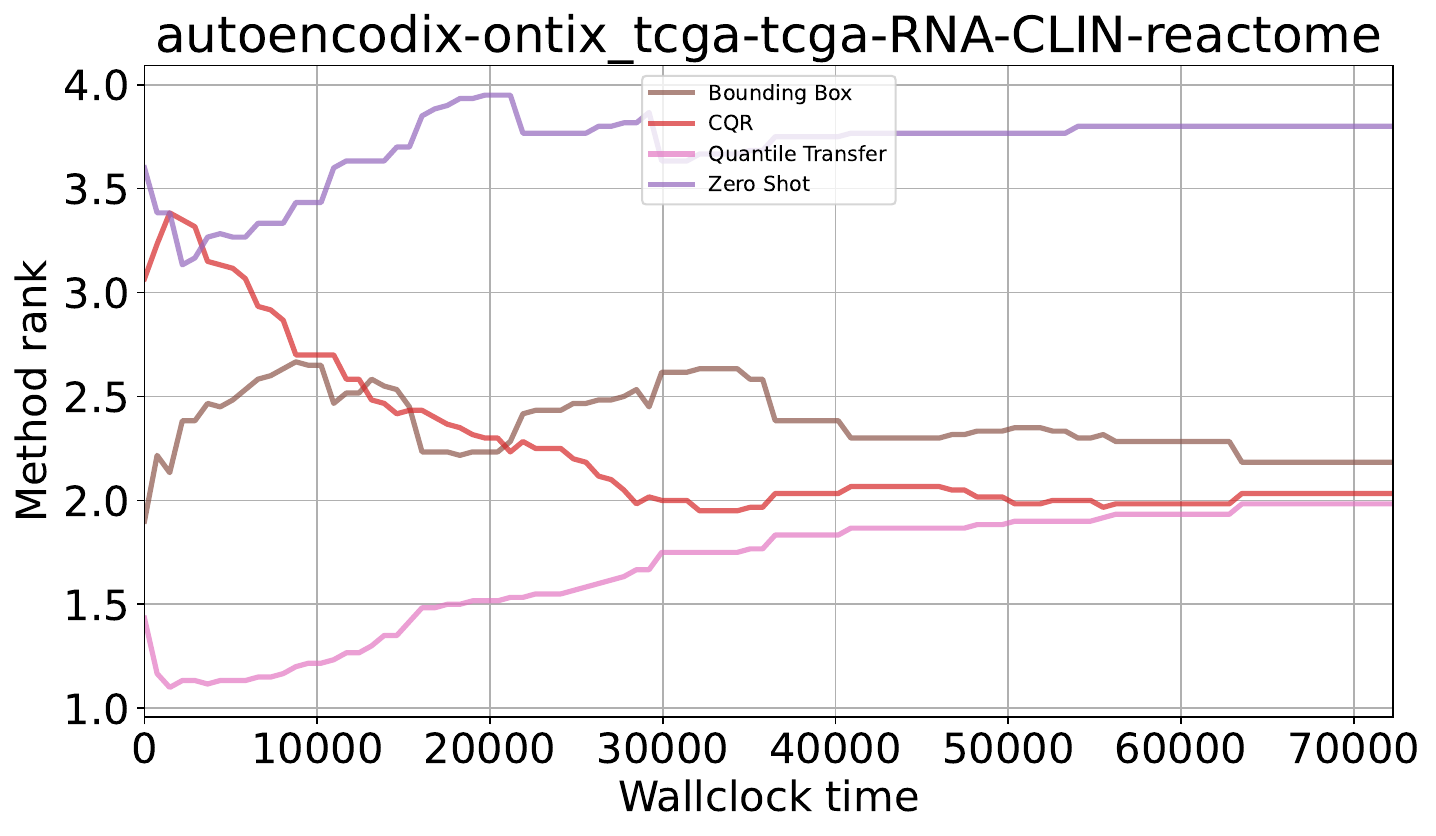} &
    \includegraphics[width=0.32\textwidth]{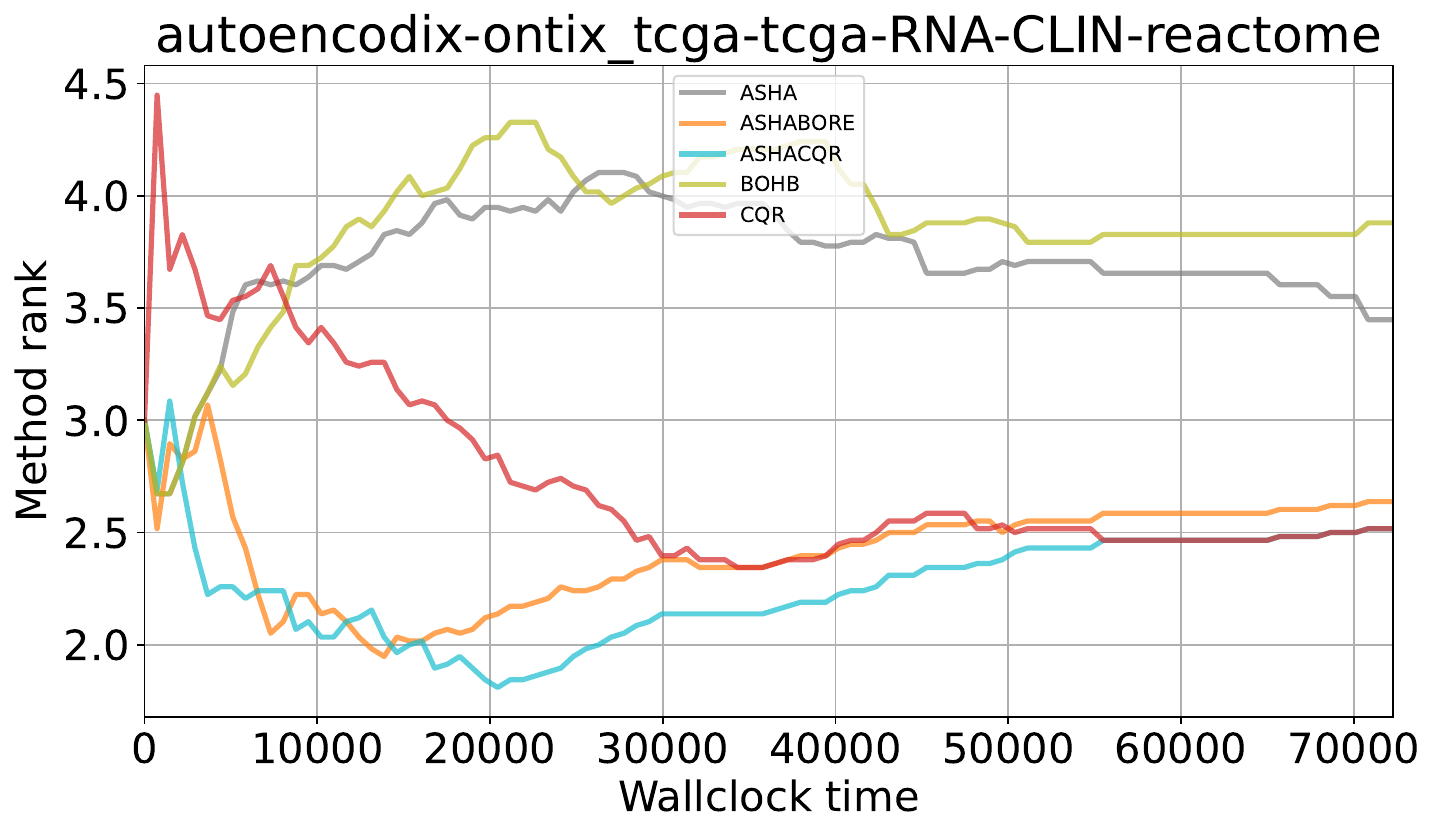} \\
    \end{tabular}
    \caption{Results for Ontix tasks (Part 4).}
    \label{fig:ontix_part4}
\end{figure}

\clearpage

\begin{figure}[htbp]
    \centering
    \setlength{\tabcolsep}{1pt}
    \begin{tabular}{ccc}
    \multicolumn{3}{c}{\textbf{autoencodix-ontix\_tcga-tcga-RNA-DNA-METH-CLIN-chromosome}} \\
    \textbf{Single-Fidelity} & \textbf{Transfer Learning} & \textbf{Multi-Fidelity} \\
    \includegraphics[width=0.32\textwidth]{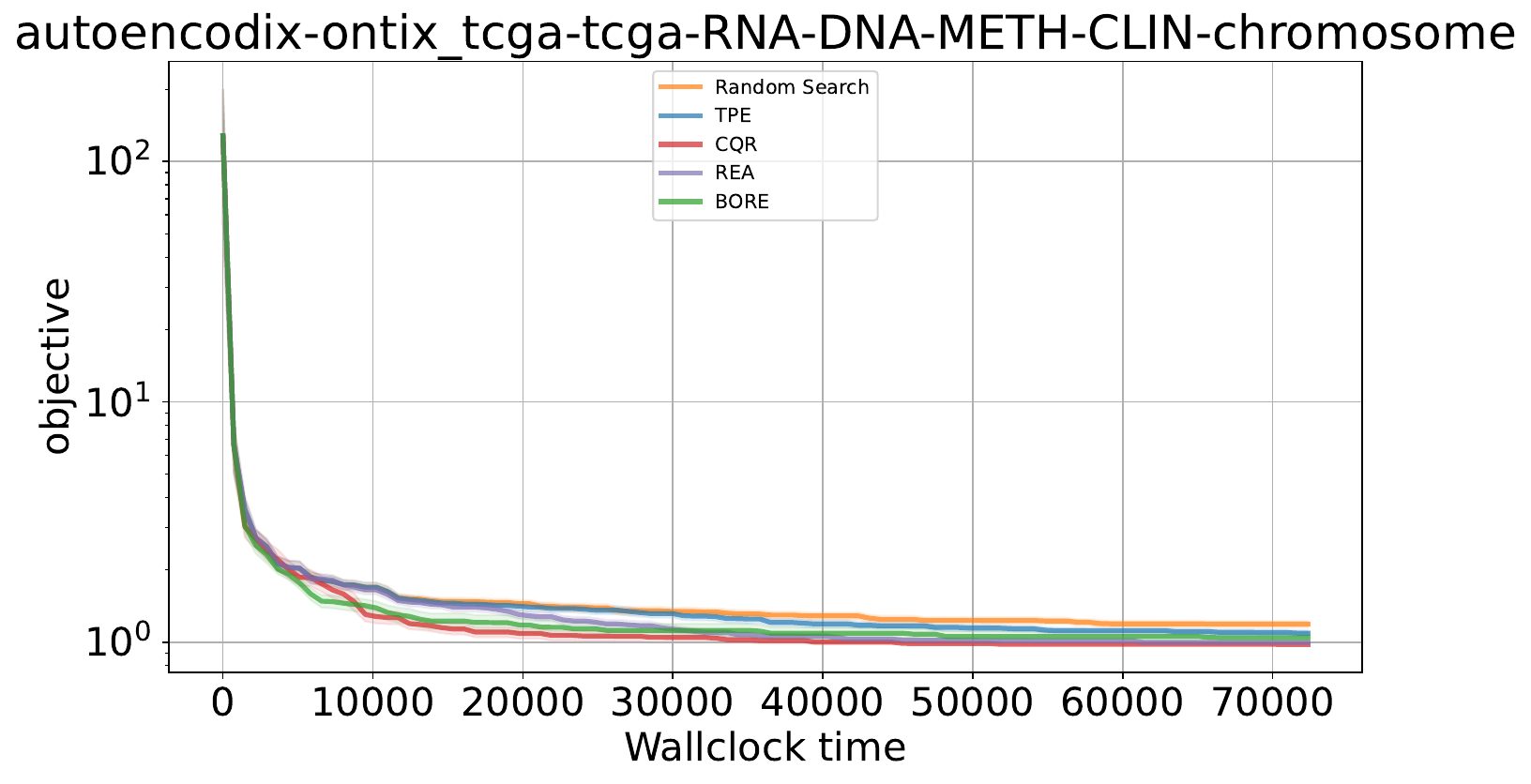} &
    \includegraphics[width=0.32\textwidth]{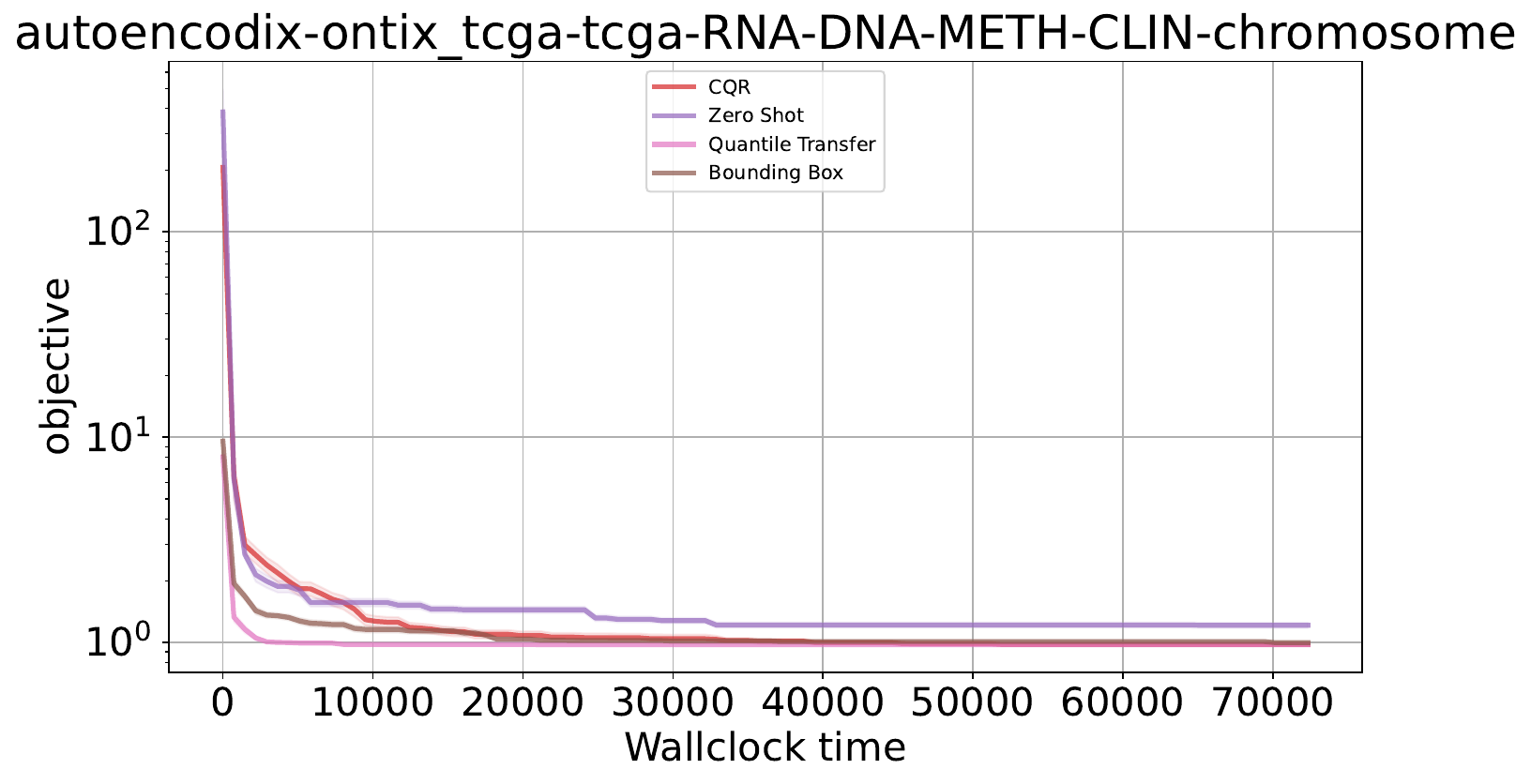} &
    \includegraphics[width=0.32\textwidth]{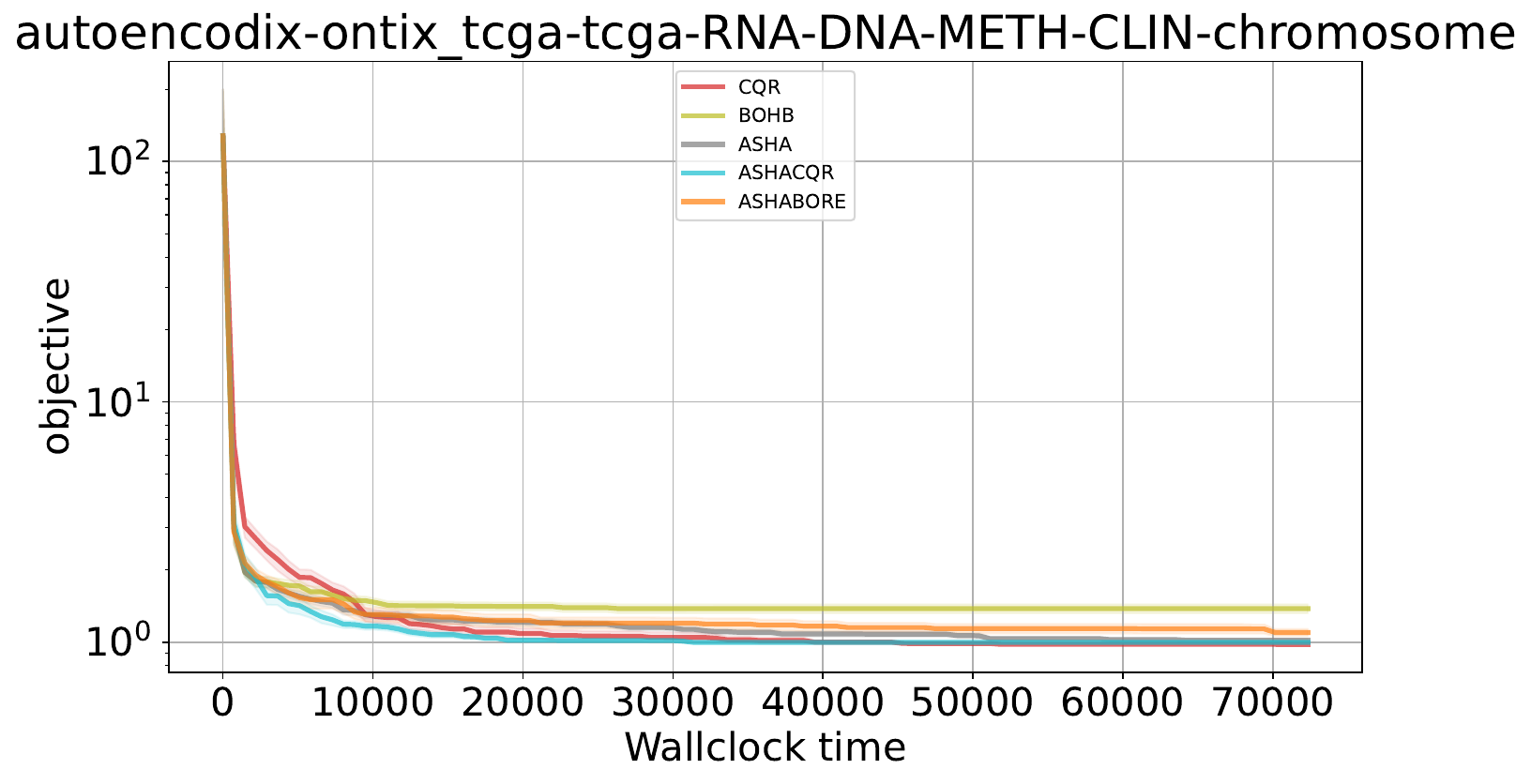} \\
    \includegraphics[width=0.32\textwidth]{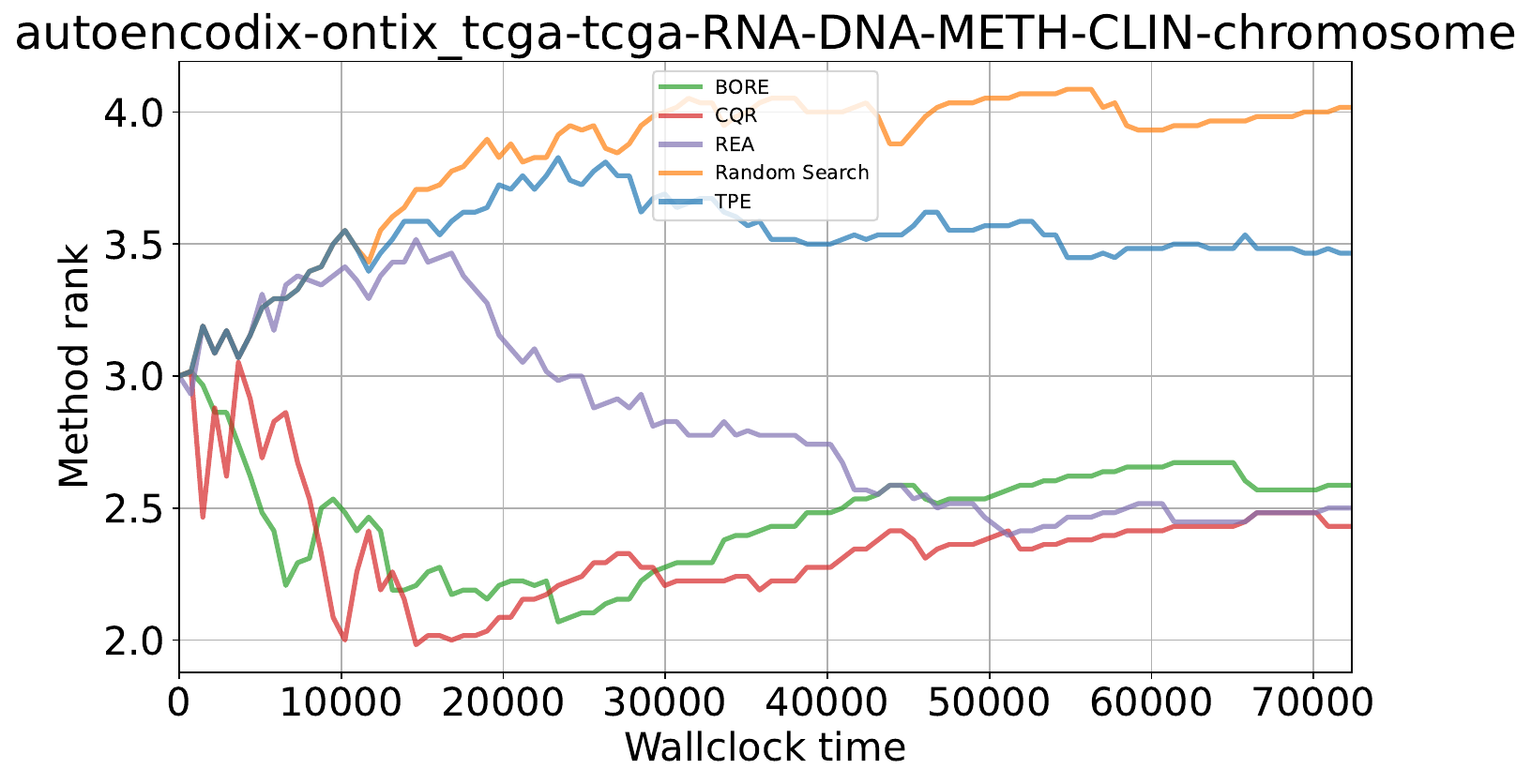} &
    \includegraphics[width=0.32\textwidth]{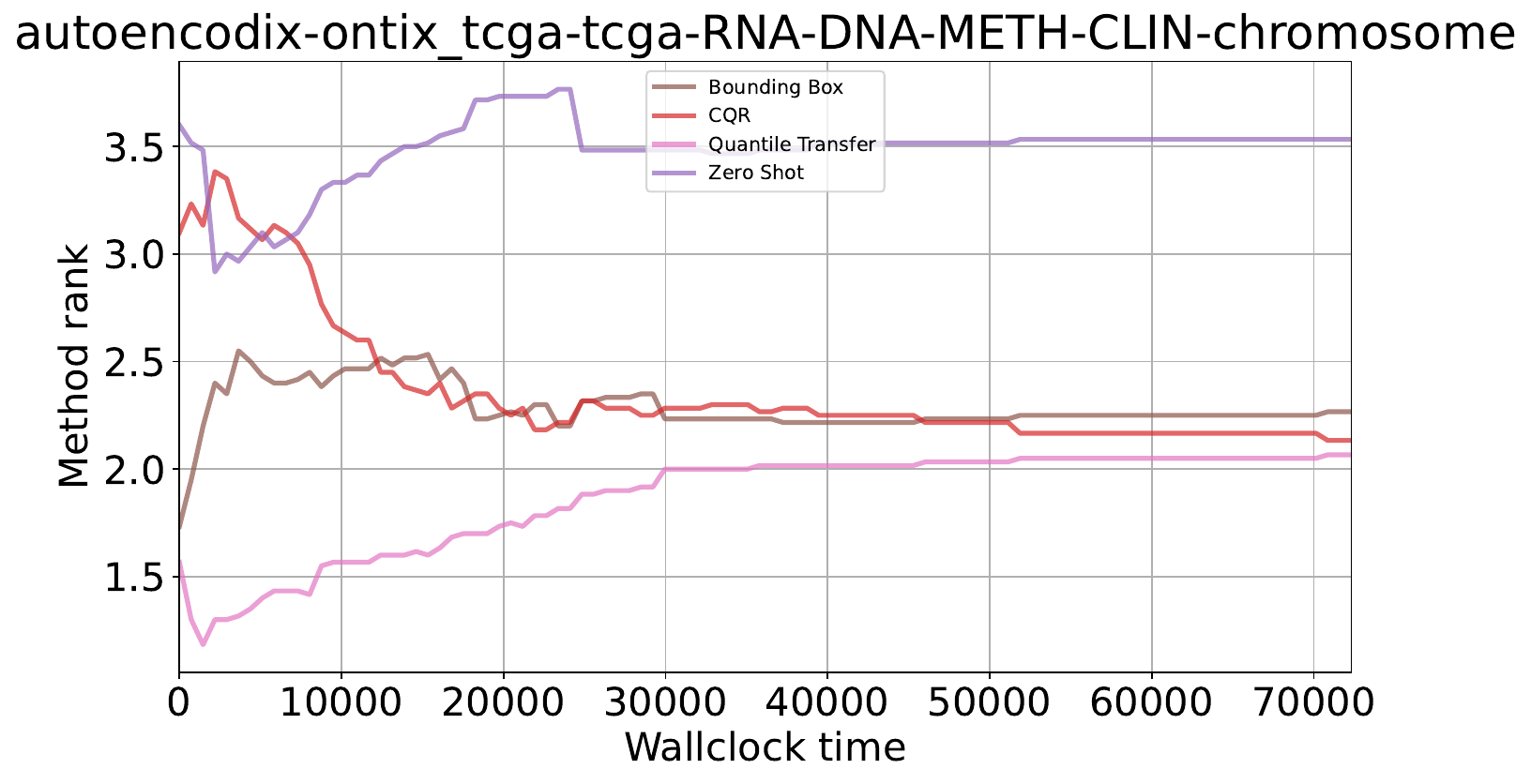} &
    \includegraphics[width=0.32\textwidth]{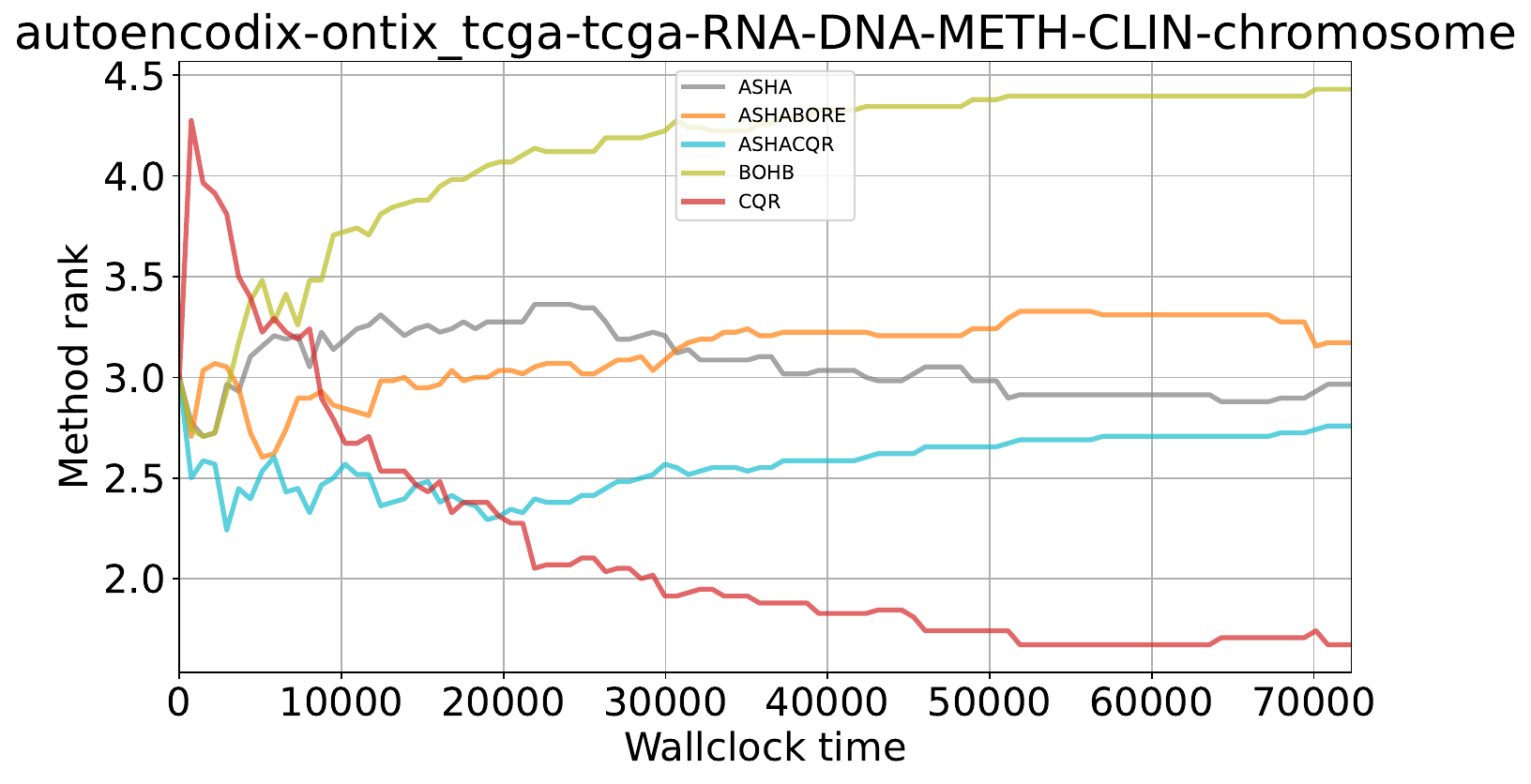} \\
    \midrule
    \multicolumn{3}{c}{\textbf{autoencodix-ontix\_tcga-tcga-RNA-DNA-METH-CLIN-reactome}} \\
    \includegraphics[width=0.32\textwidth]{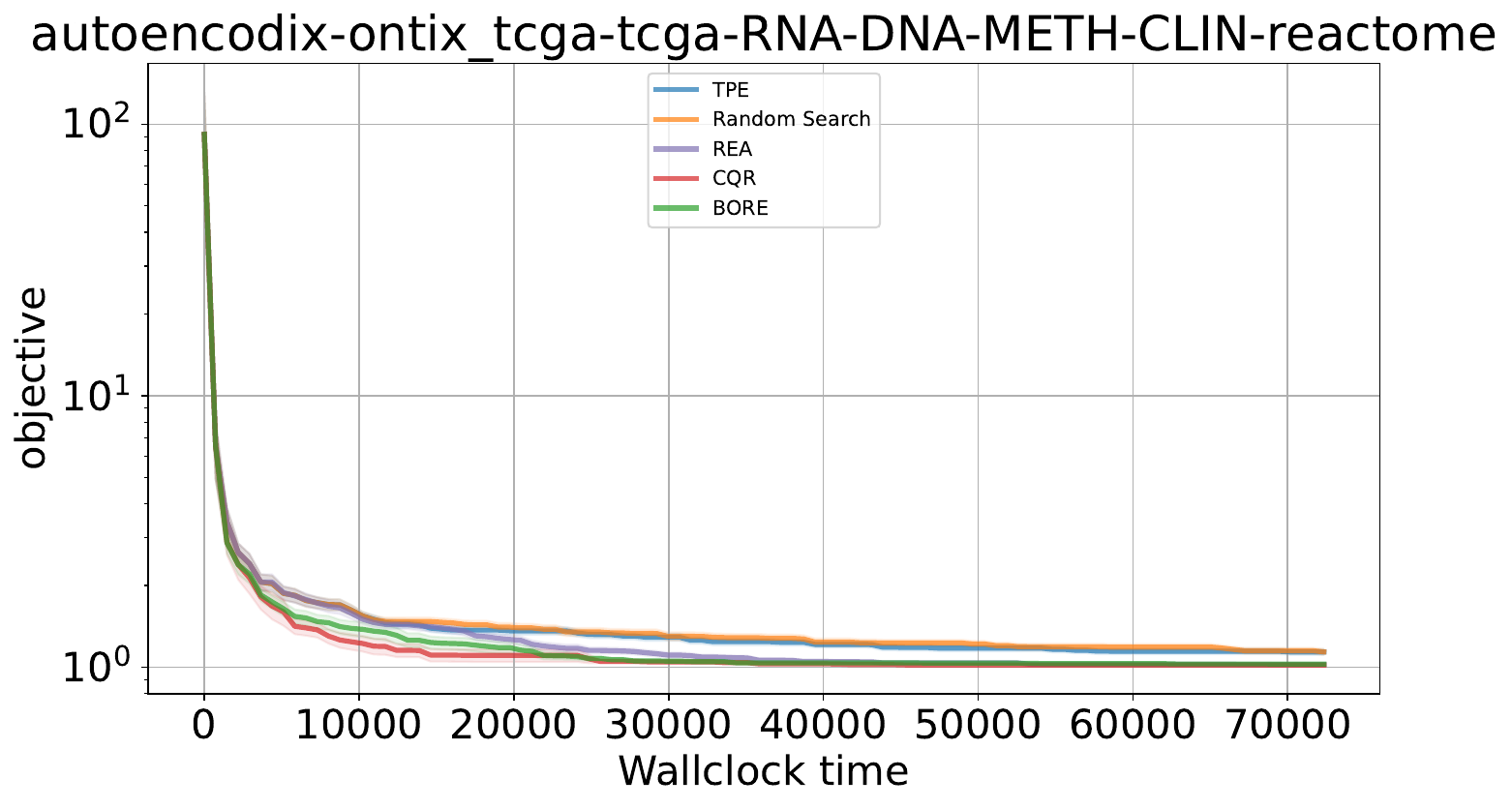} &
    \includegraphics[width=0.32\textwidth]{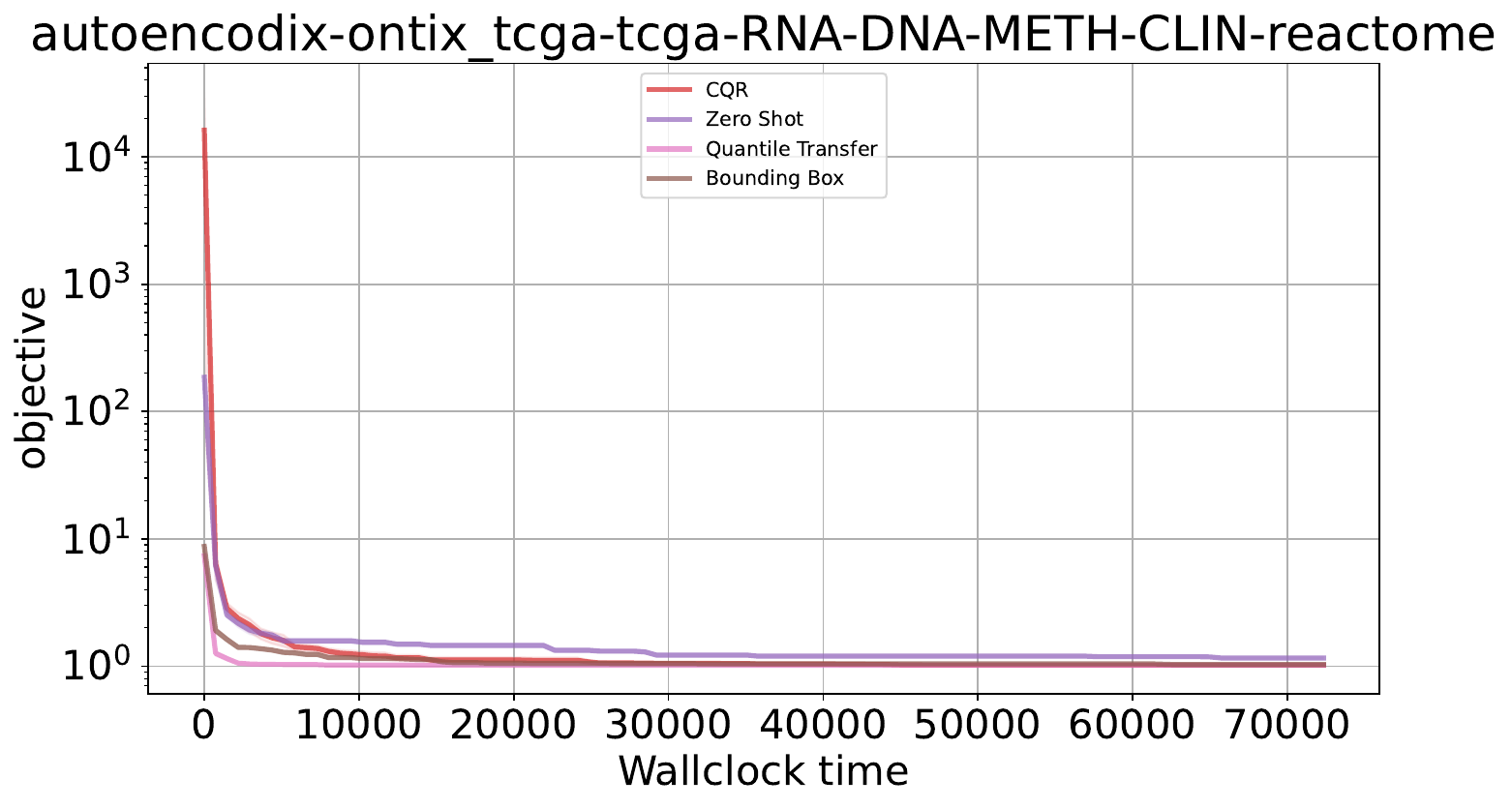} &
    \includegraphics[width=0.32\textwidth]{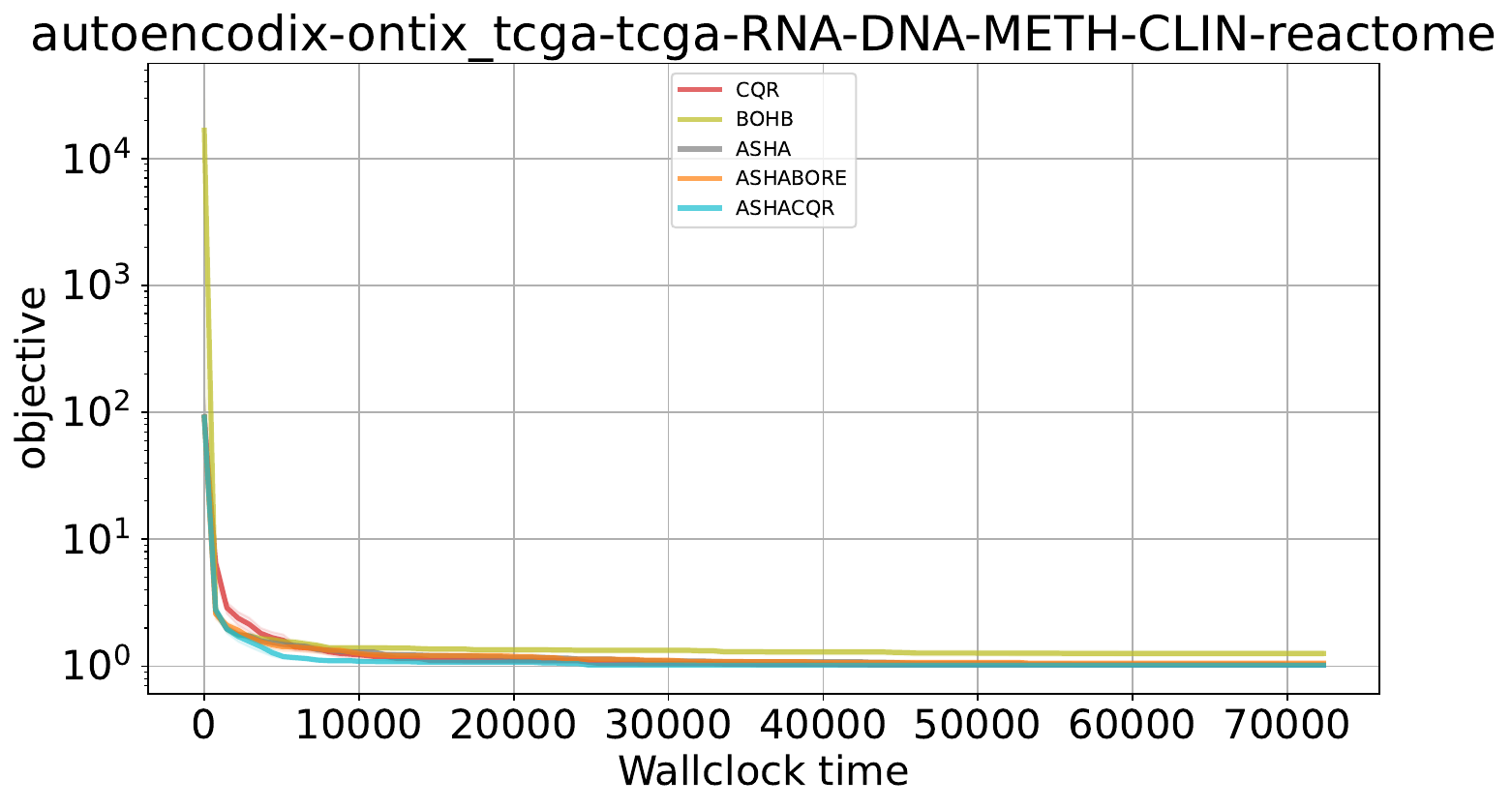} \\
    \includegraphics[width=0.32\textwidth]{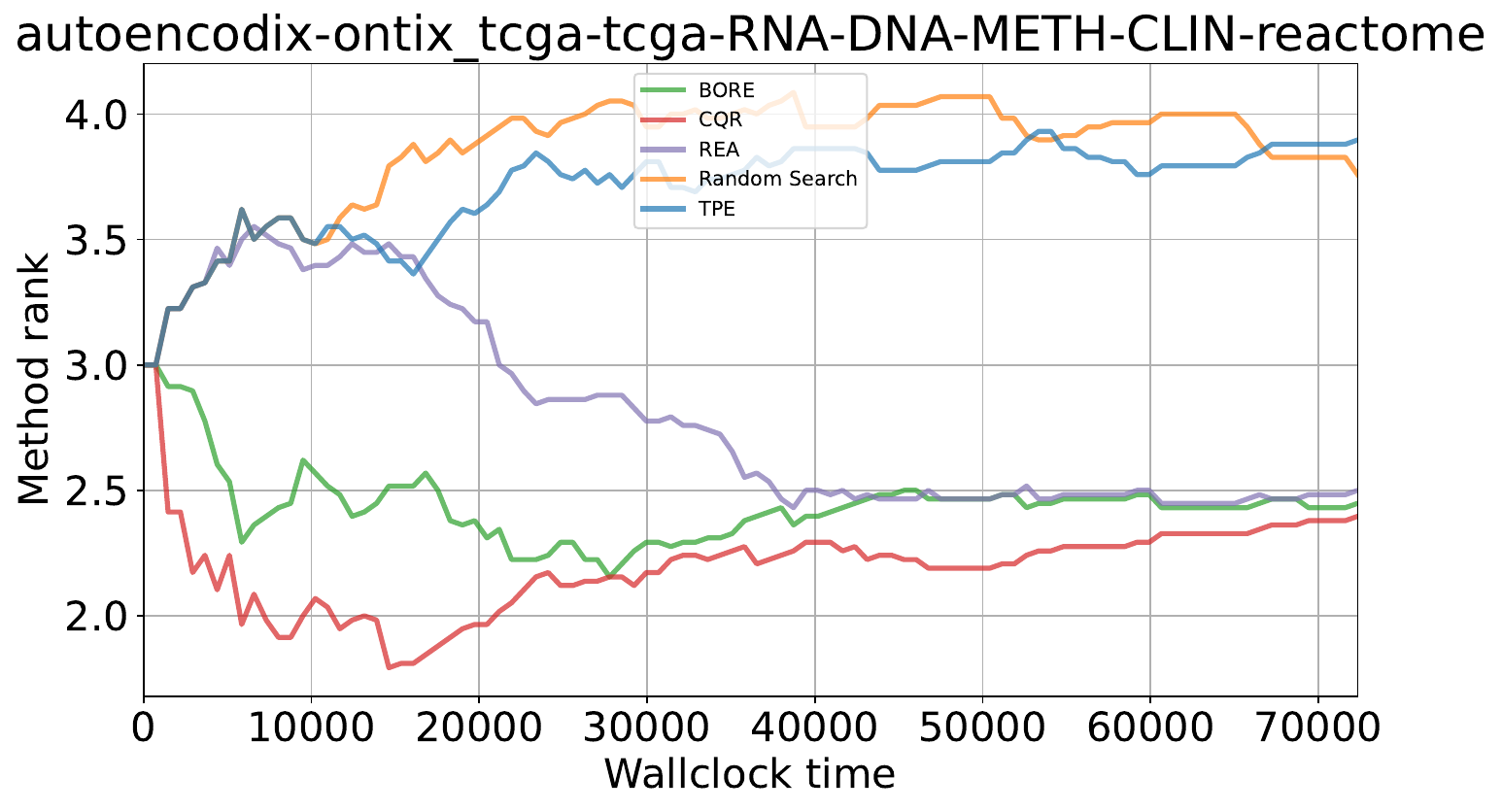} &
    \includegraphics[width=0.32\textwidth]{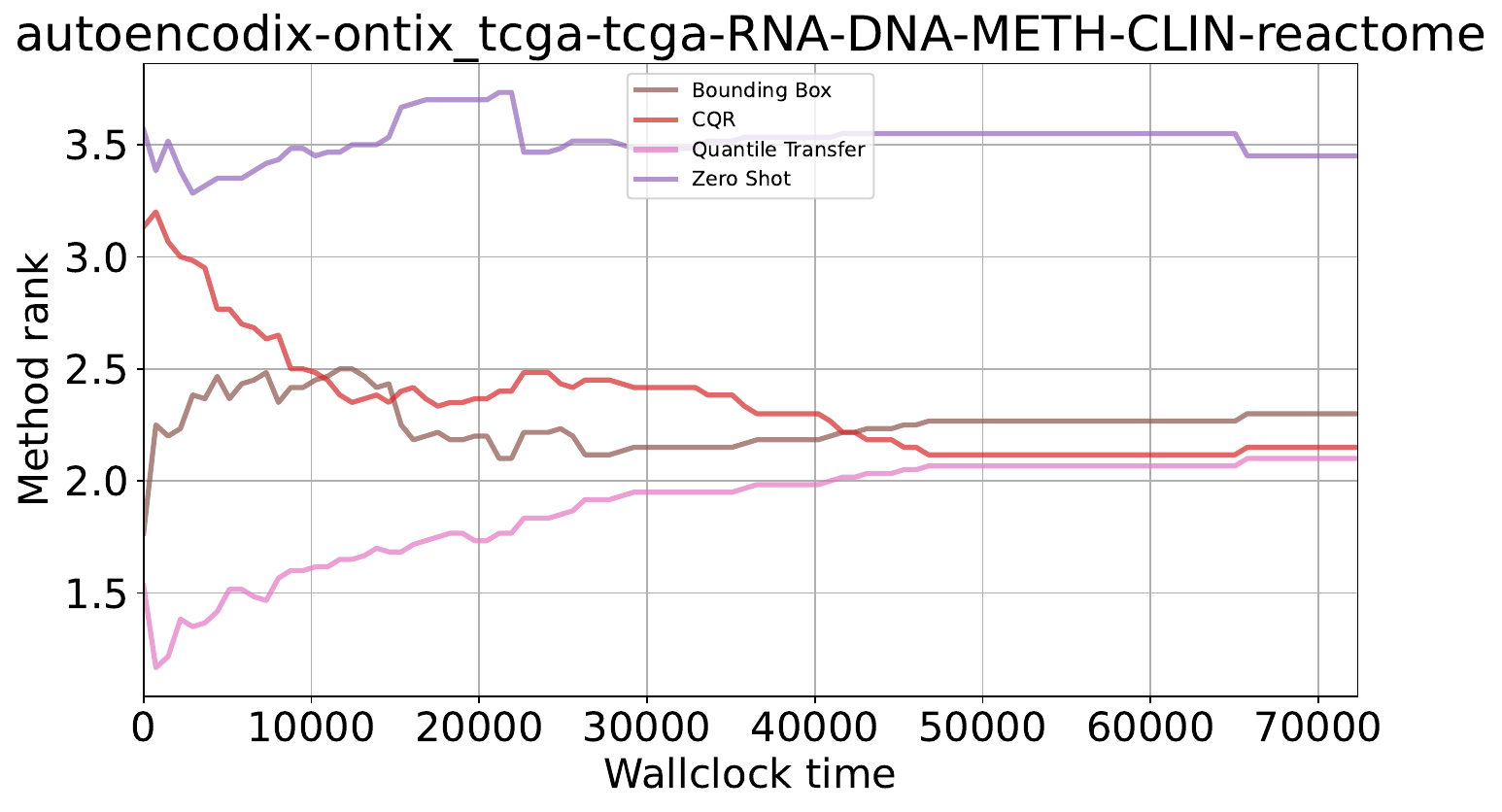} &
    \includegraphics[width=0.32\textwidth]{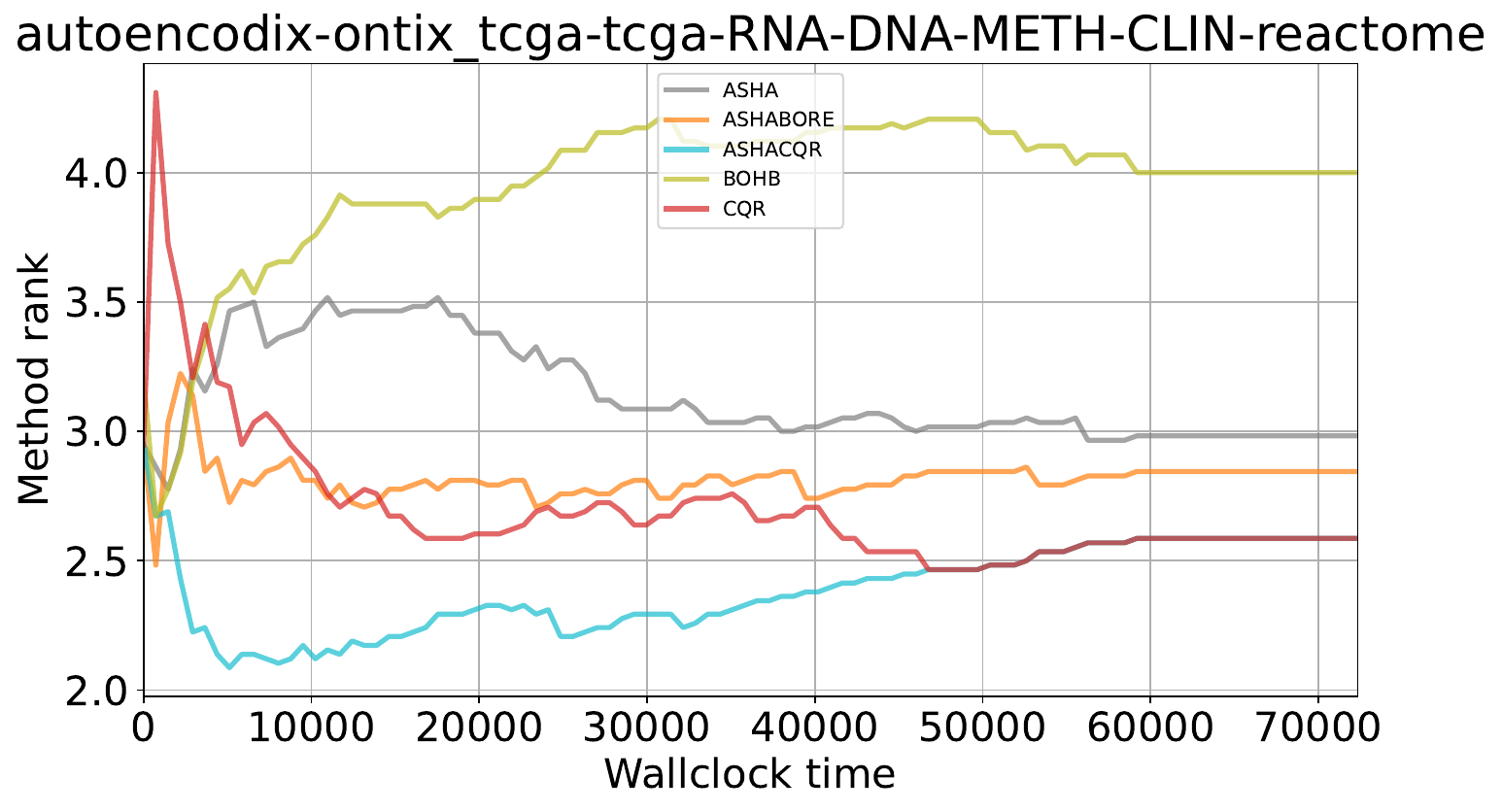} \\
    \end{tabular}
    \caption{Results for Ontix tasks (Part 5).}
    \label{fig:ontix_part5}
\end{figure}

\clearpage

% Model: Vanillix
\begin{figure}[htbp]
    \centering
    \setlength{\tabcolsep}{1pt}
    \begin{tabular}{ccc}
    \multicolumn{3}{c}{\textbf{autoencodix-vanillix\_schc-schc-METH-CLIN}} \\
    \textbf{Single-Fidelity} & \textbf{Transfer Learning} & \textbf{Multi-Fidelity} \\
    \includegraphics[width=0.32\textwidth]{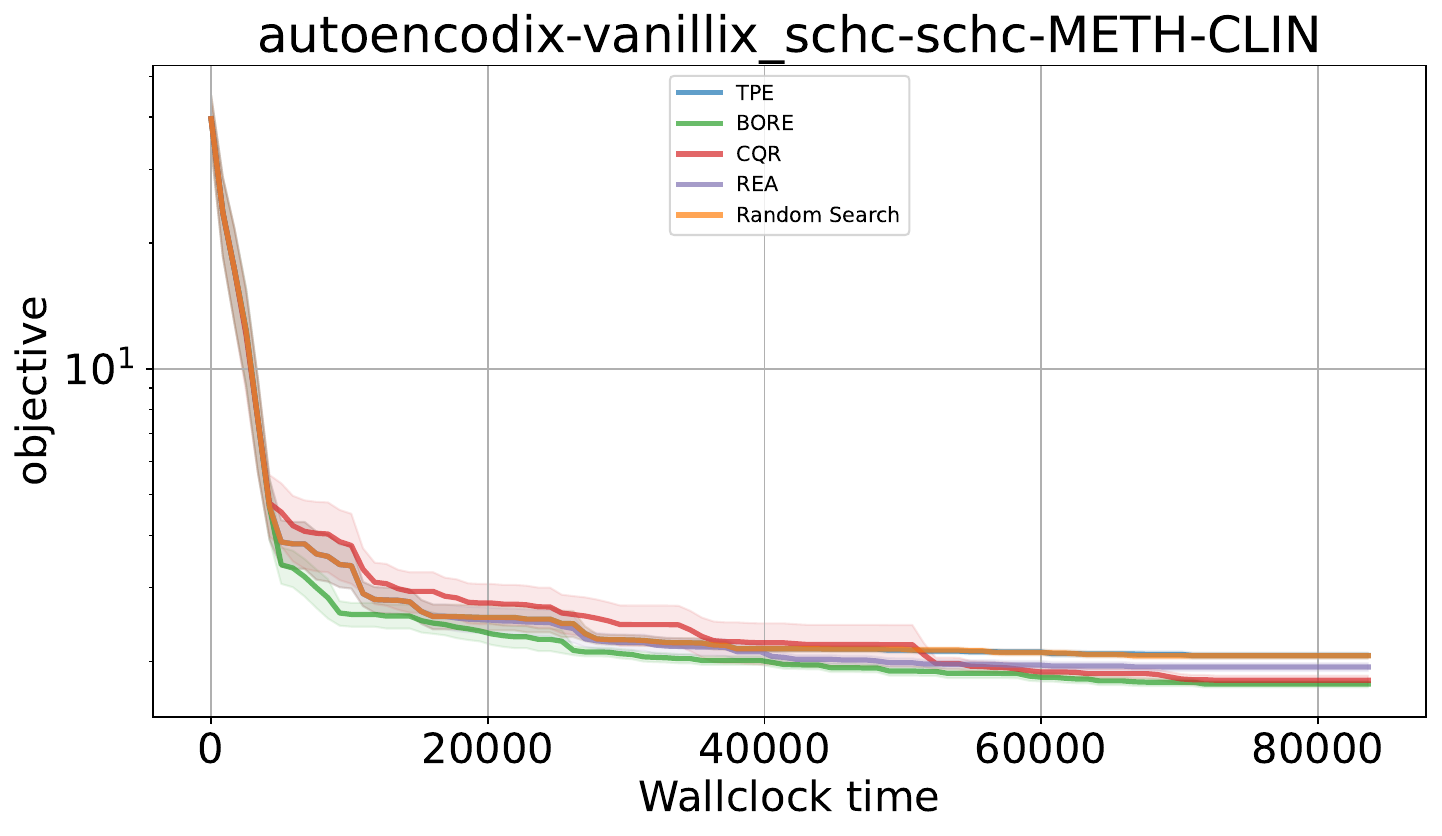} &
    \includegraphics[width=0.32\textwidth]{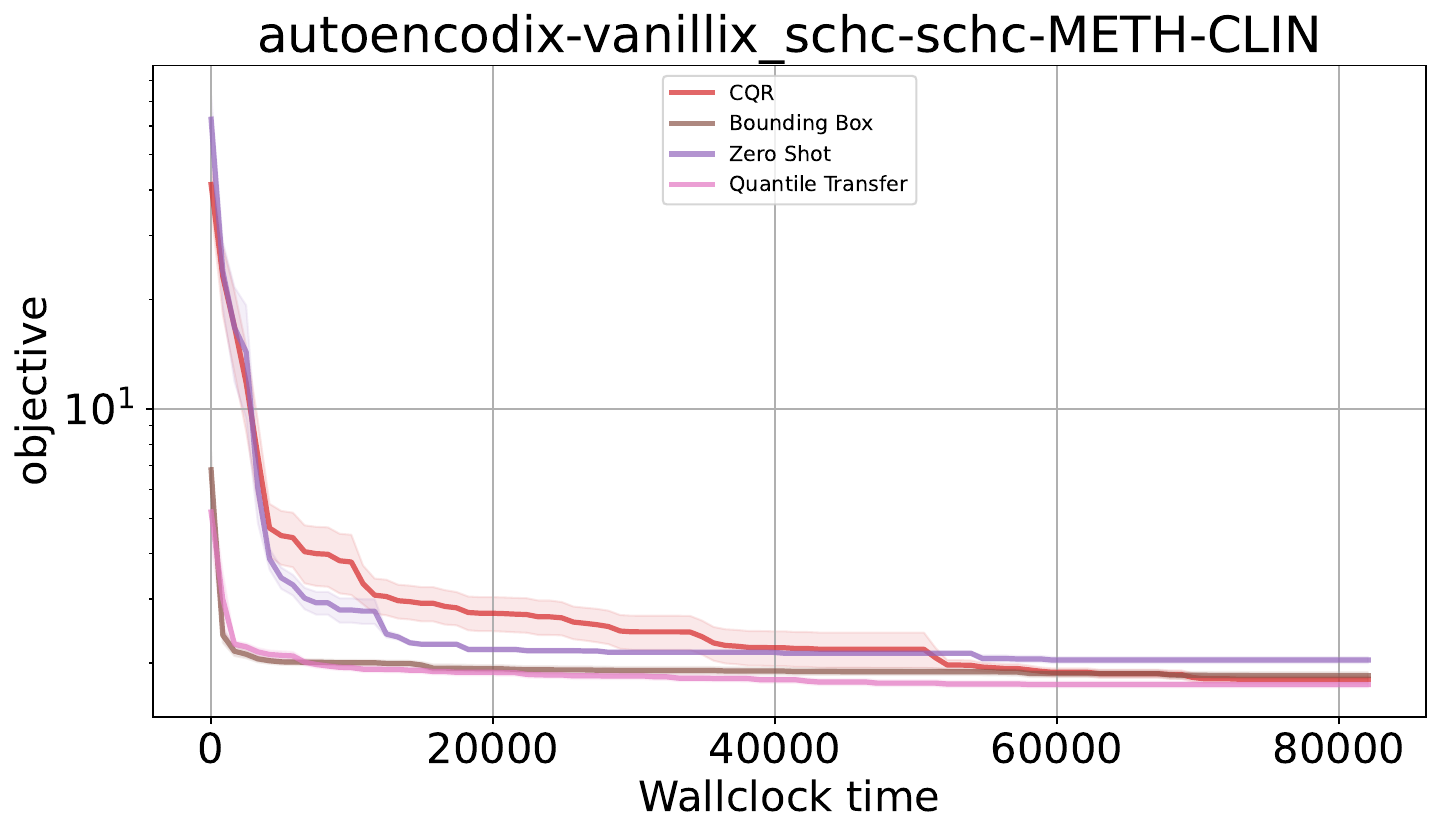} &
    \includegraphics[width=0.32\textwidth]{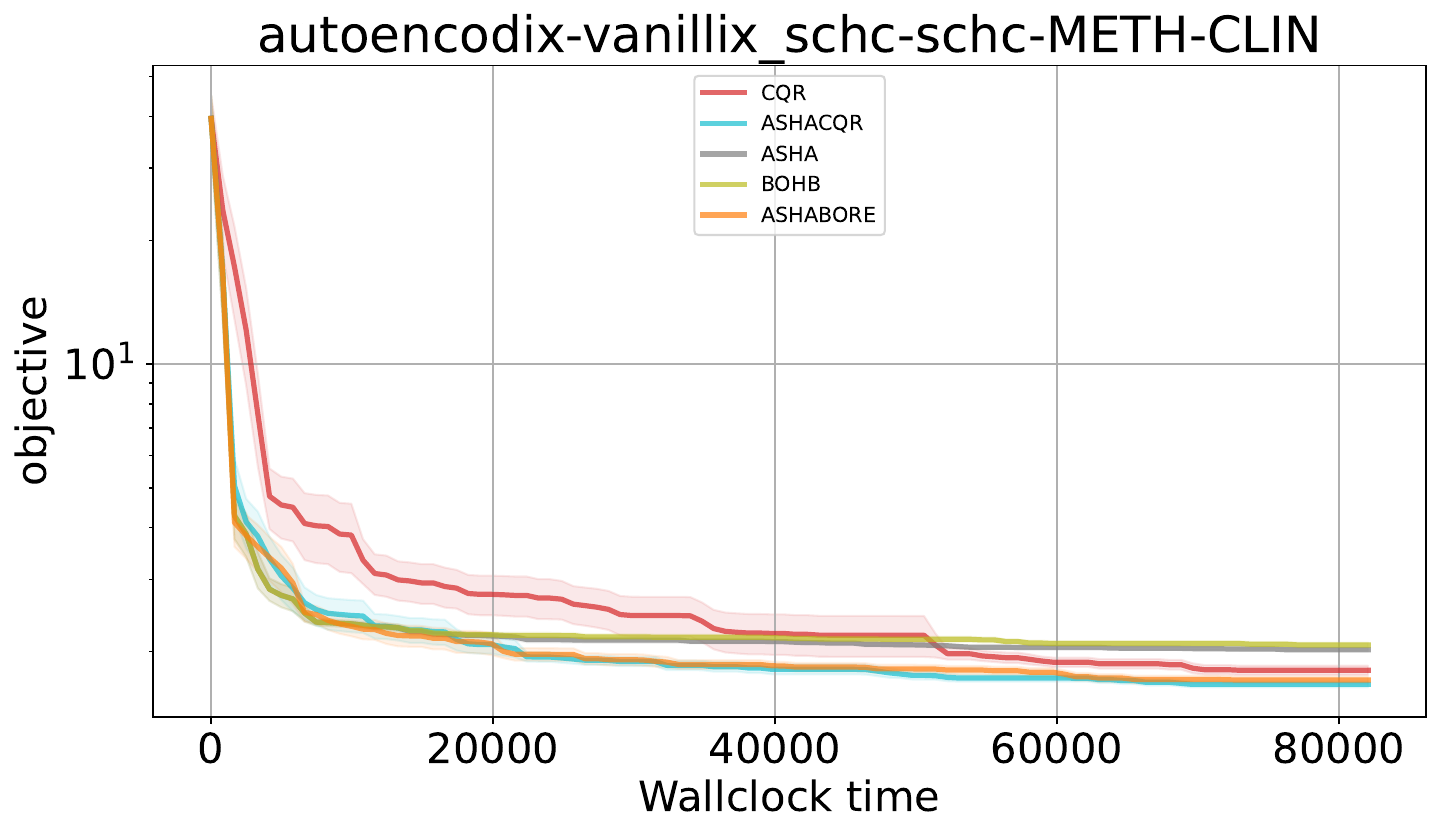} \\
    \includegraphics[width=0.32\textwidth]{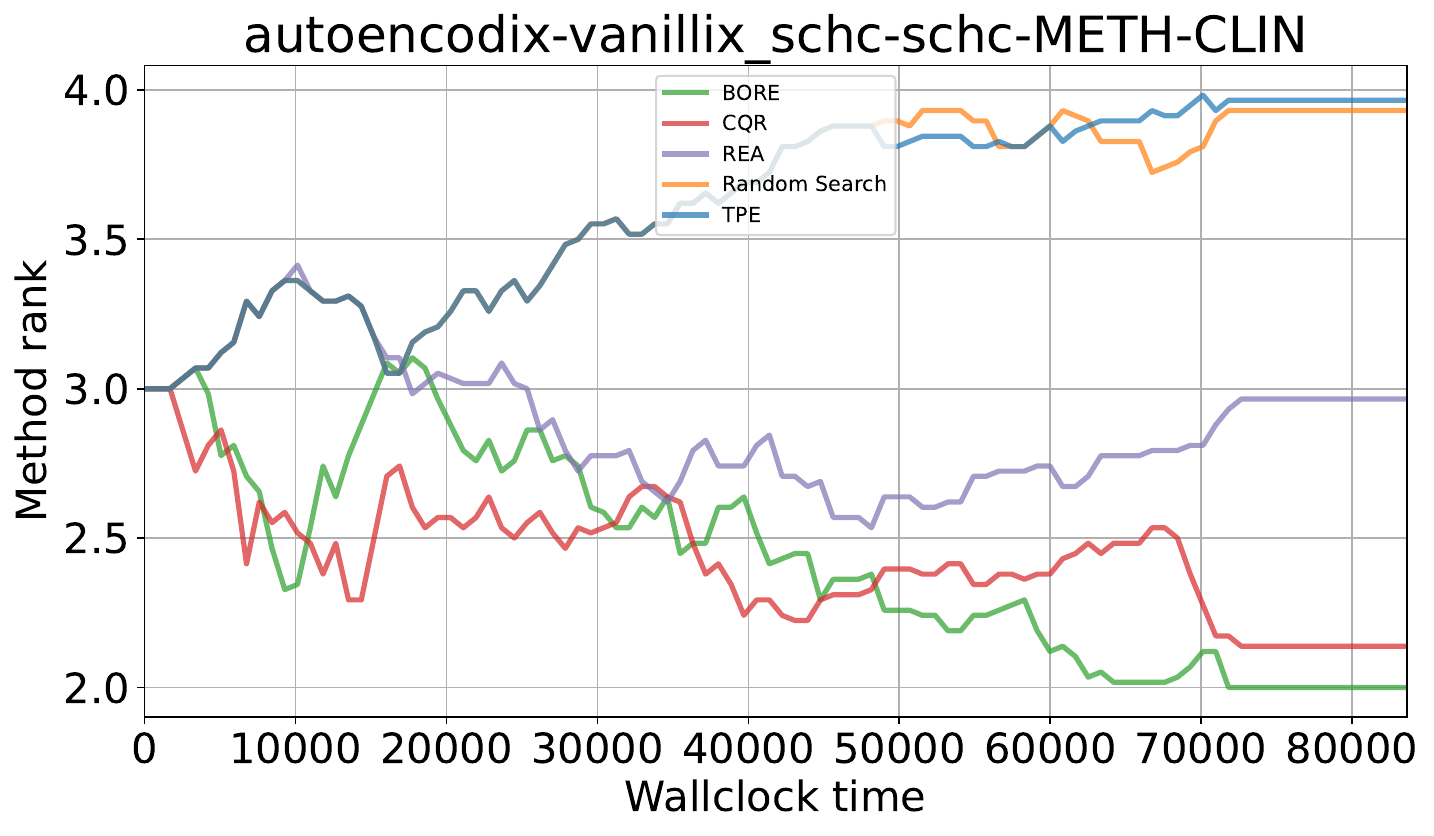} &
    \includegraphics[width=0.32\textwidth]{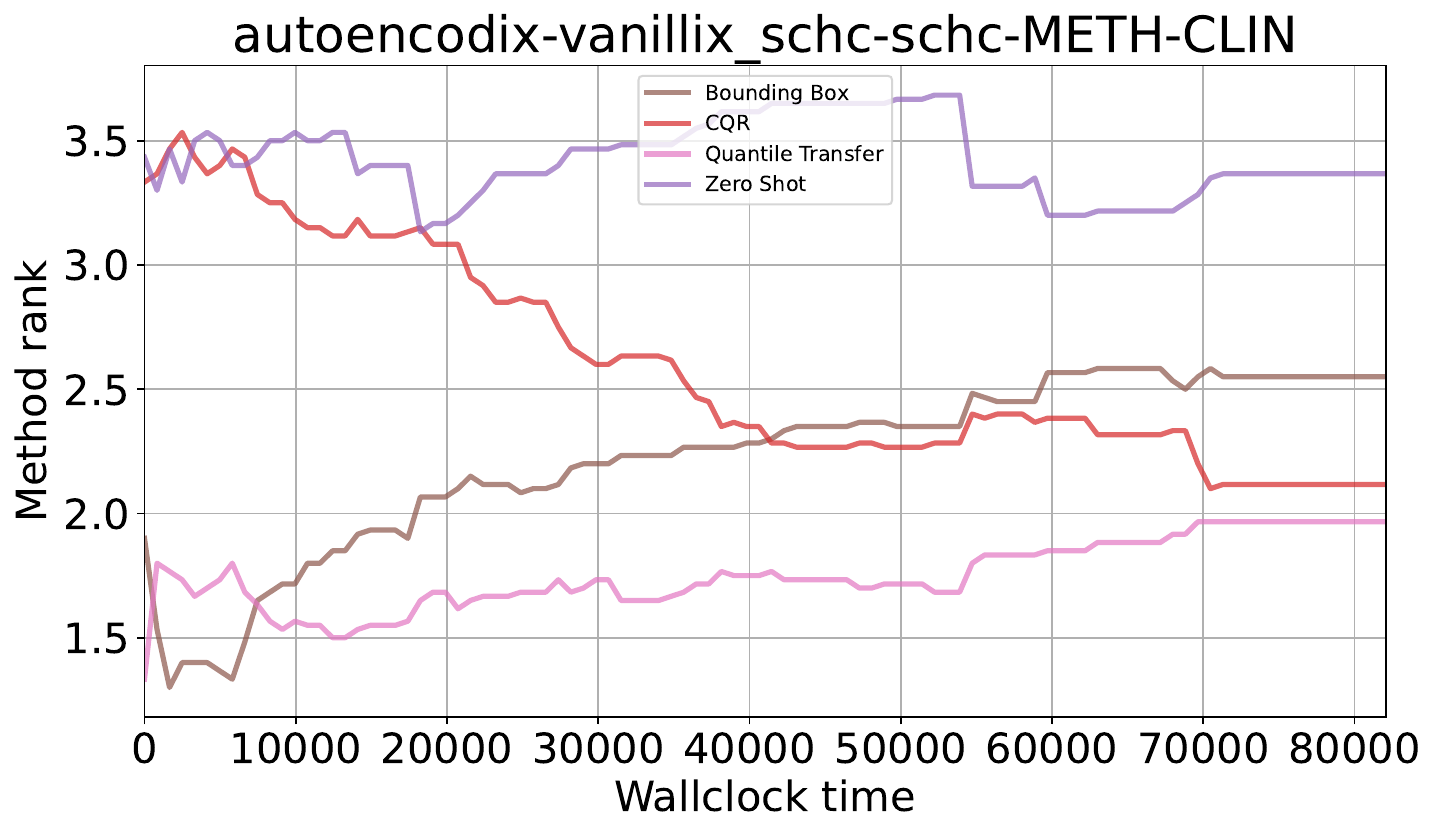} &
    \includegraphics[width=0.32\textwidth]{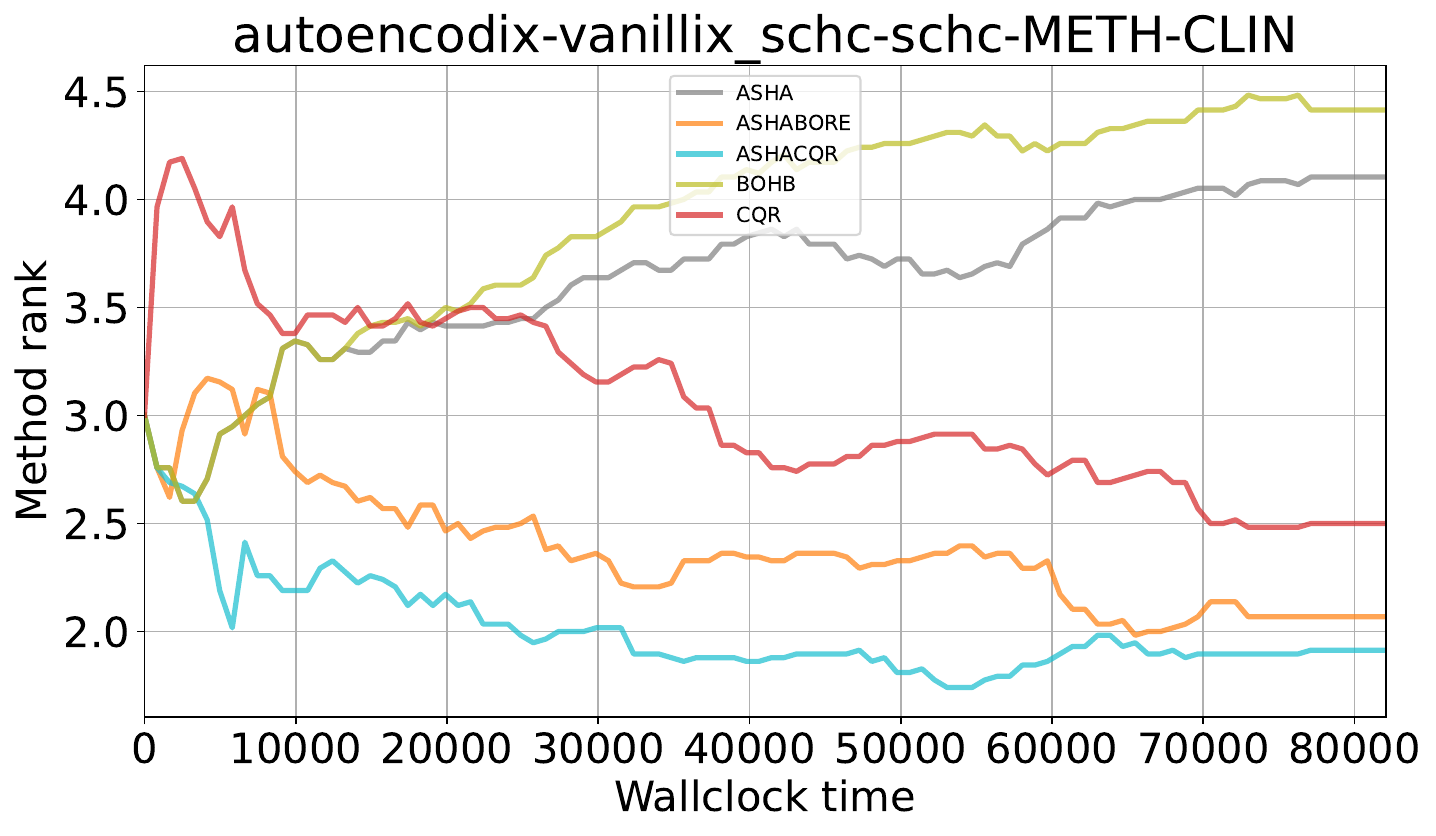} \\
    \midrule
    \multicolumn{3}{c}{\textbf{autoencodix-vanillix\_schc-schc-RNA-CLIN}} \\
    \includegraphics[width=0.32\textwidth]{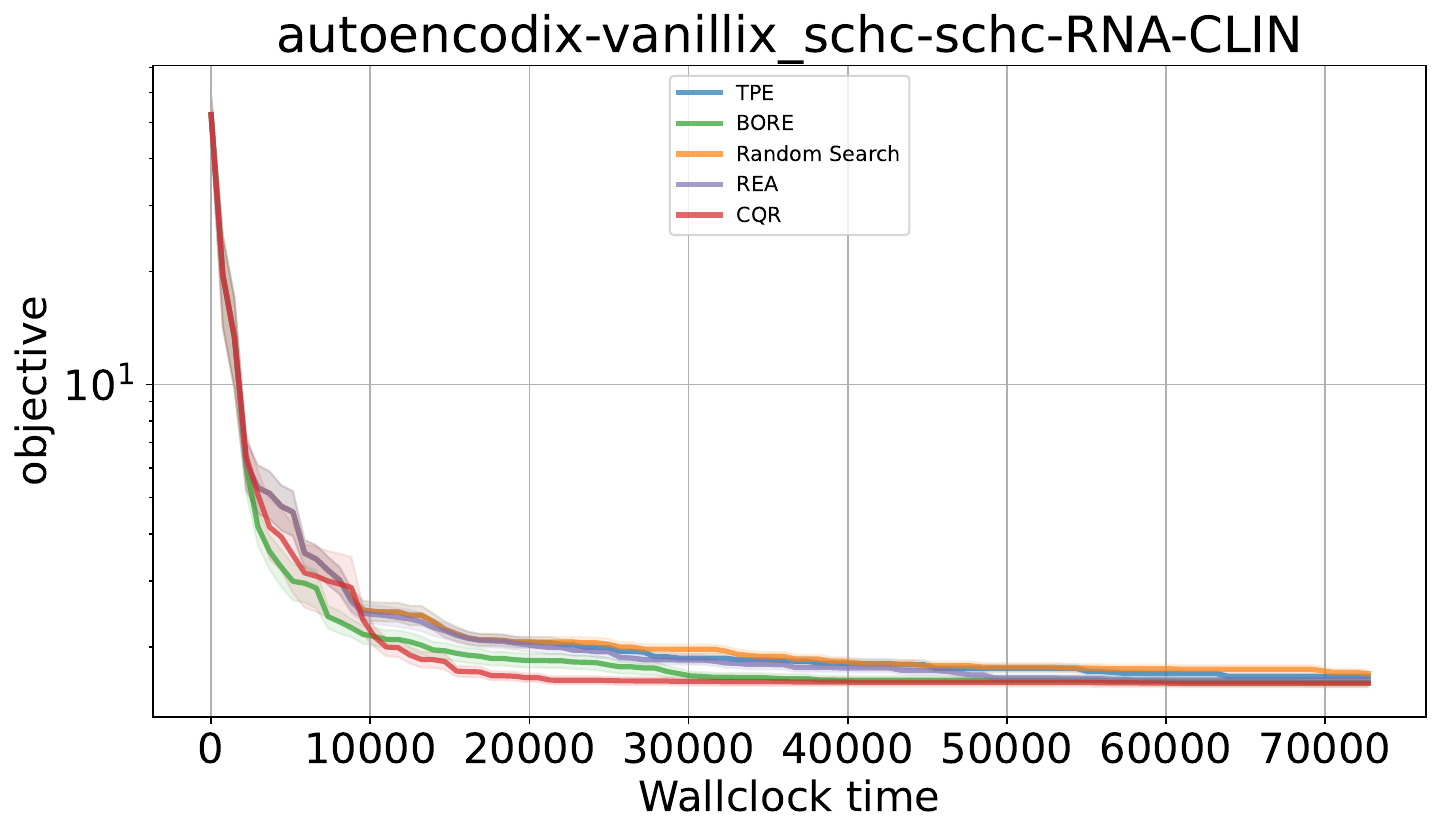} &
    \includegraphics[width=0.32\textwidth]{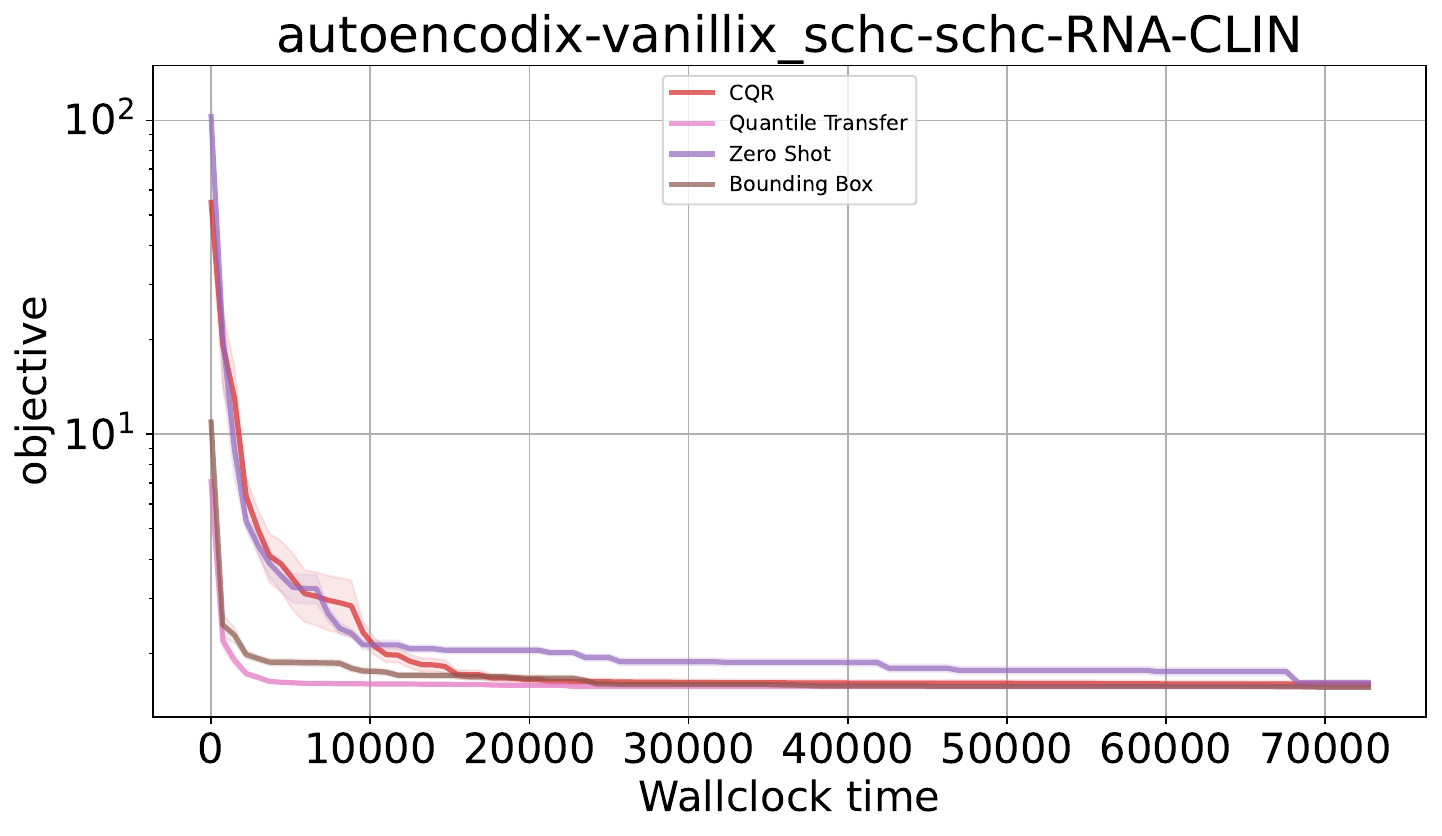} &
    \includegraphics[width=0.32\textwidth]{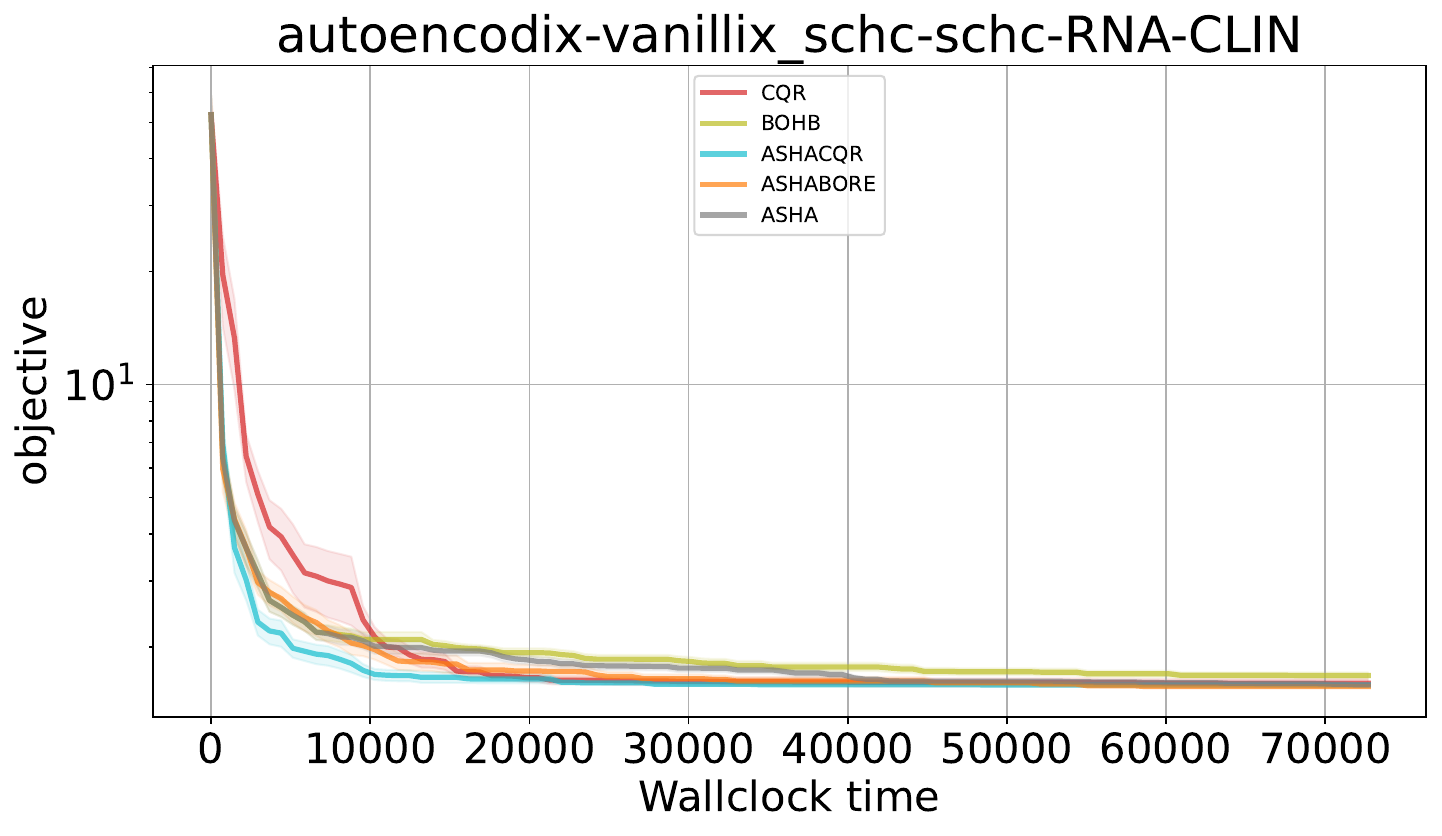} \\
    \includegraphics[width=0.32\textwidth]{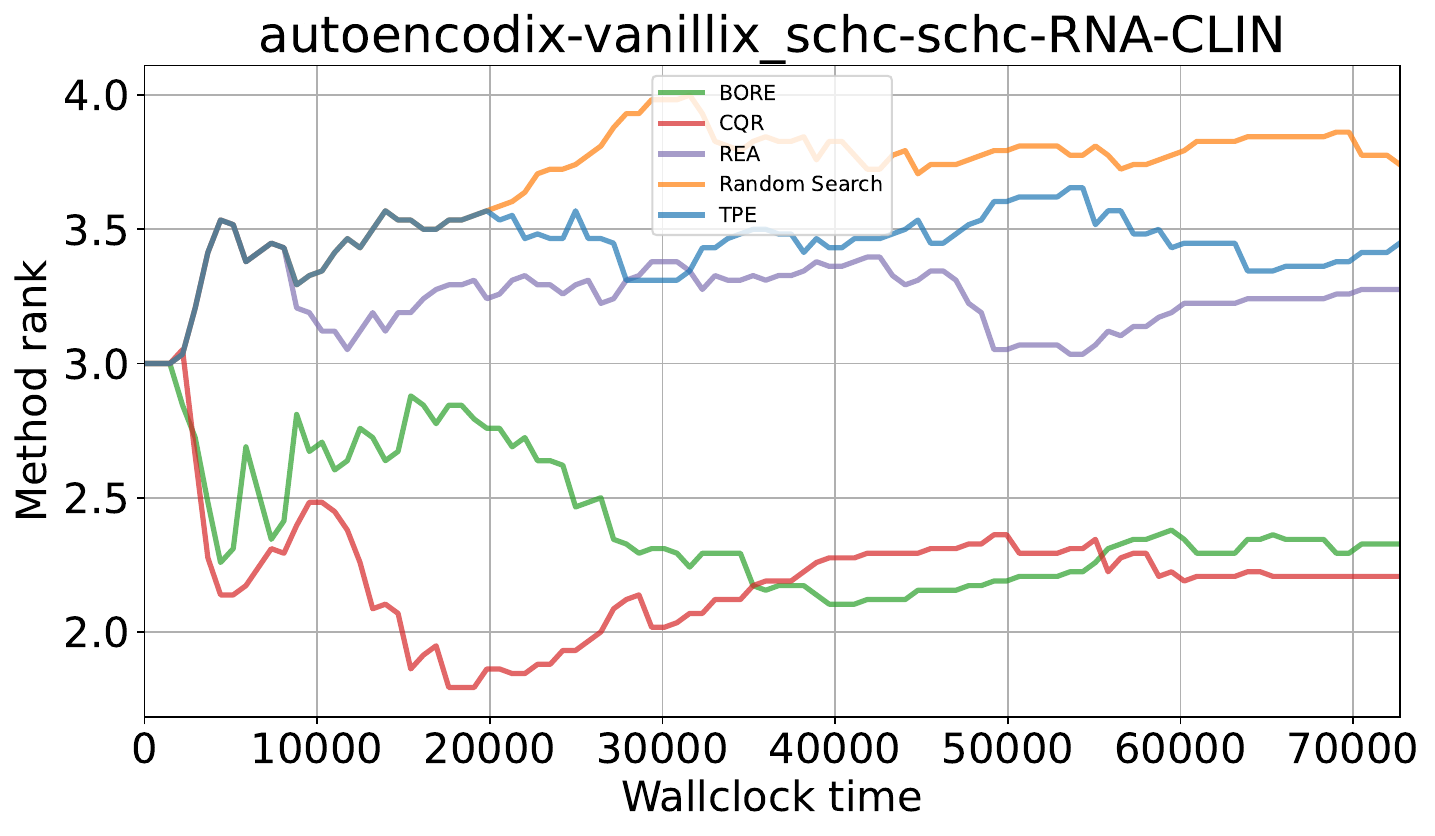} &
    \includegraphics[width=0.32\textwidth]{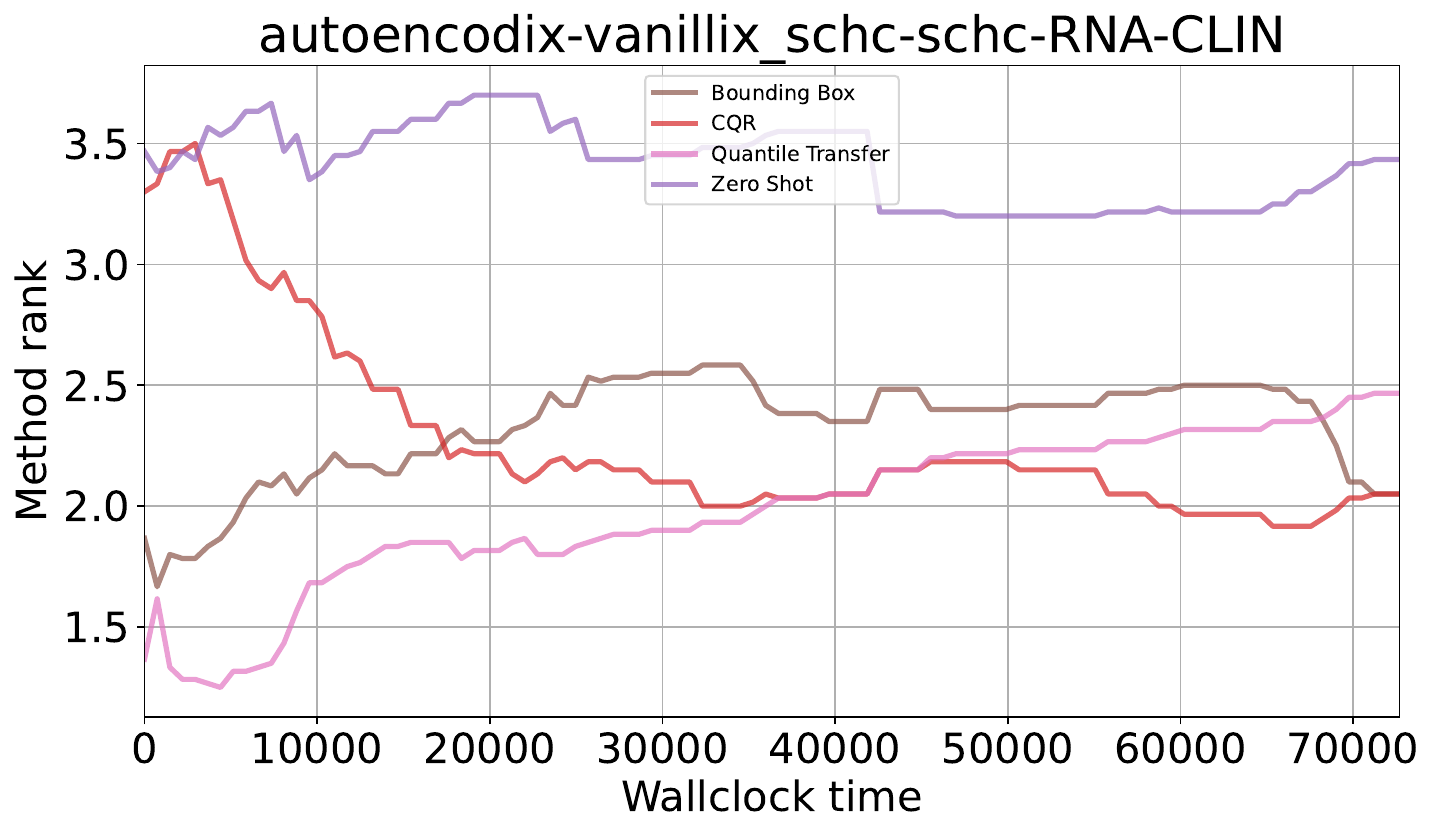} &
    \includegraphics[width=0.32\textwidth]{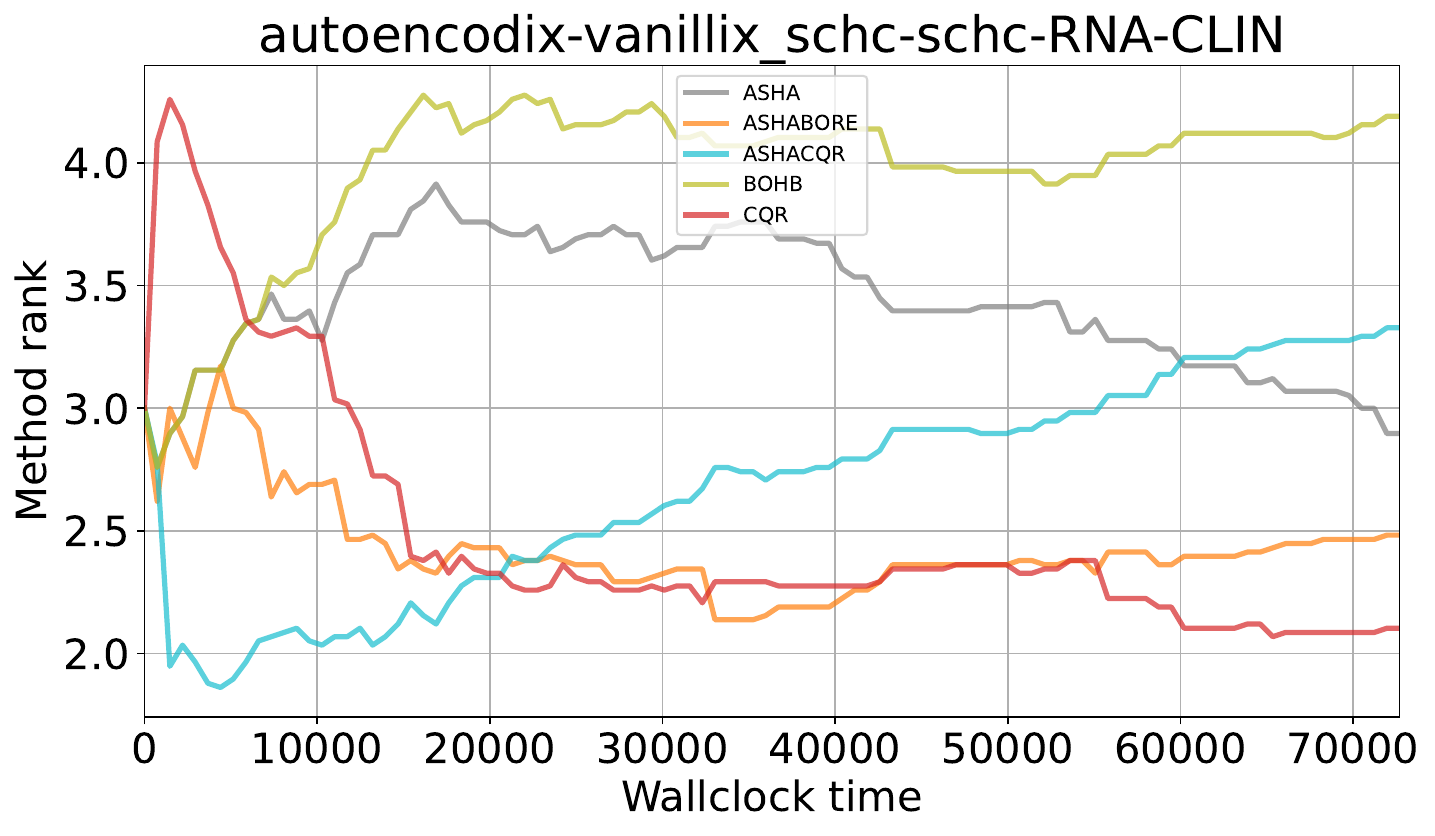} \\
    \midrule
    \multicolumn{3}{c}{\textbf{autoencodix-vanillix\_schc-schc-RNA-METH-CLIN}} \\
    \includegraphics[width=0.32\textwidth]{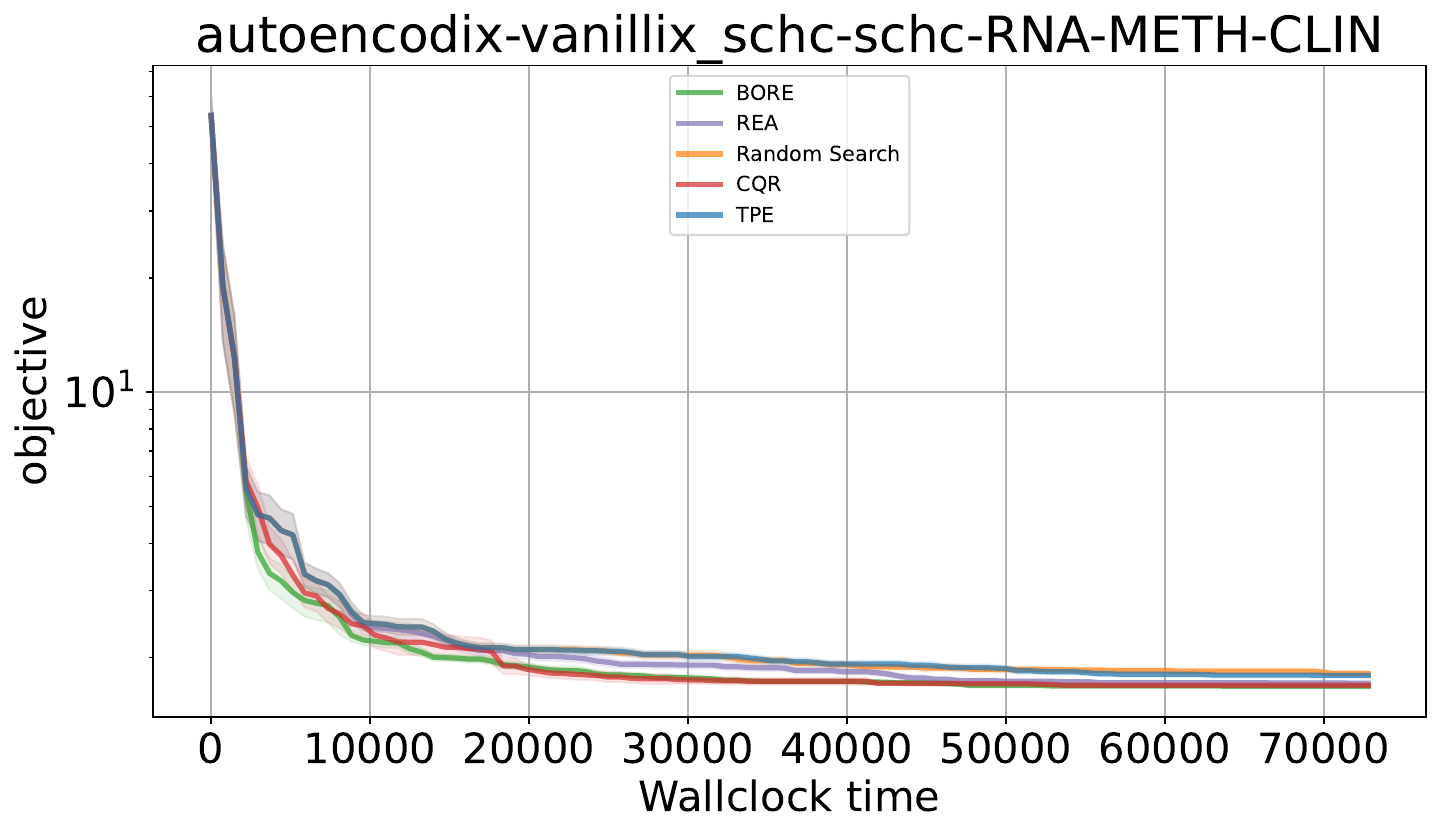} &
    \includegraphics[width=0.32\textwidth]{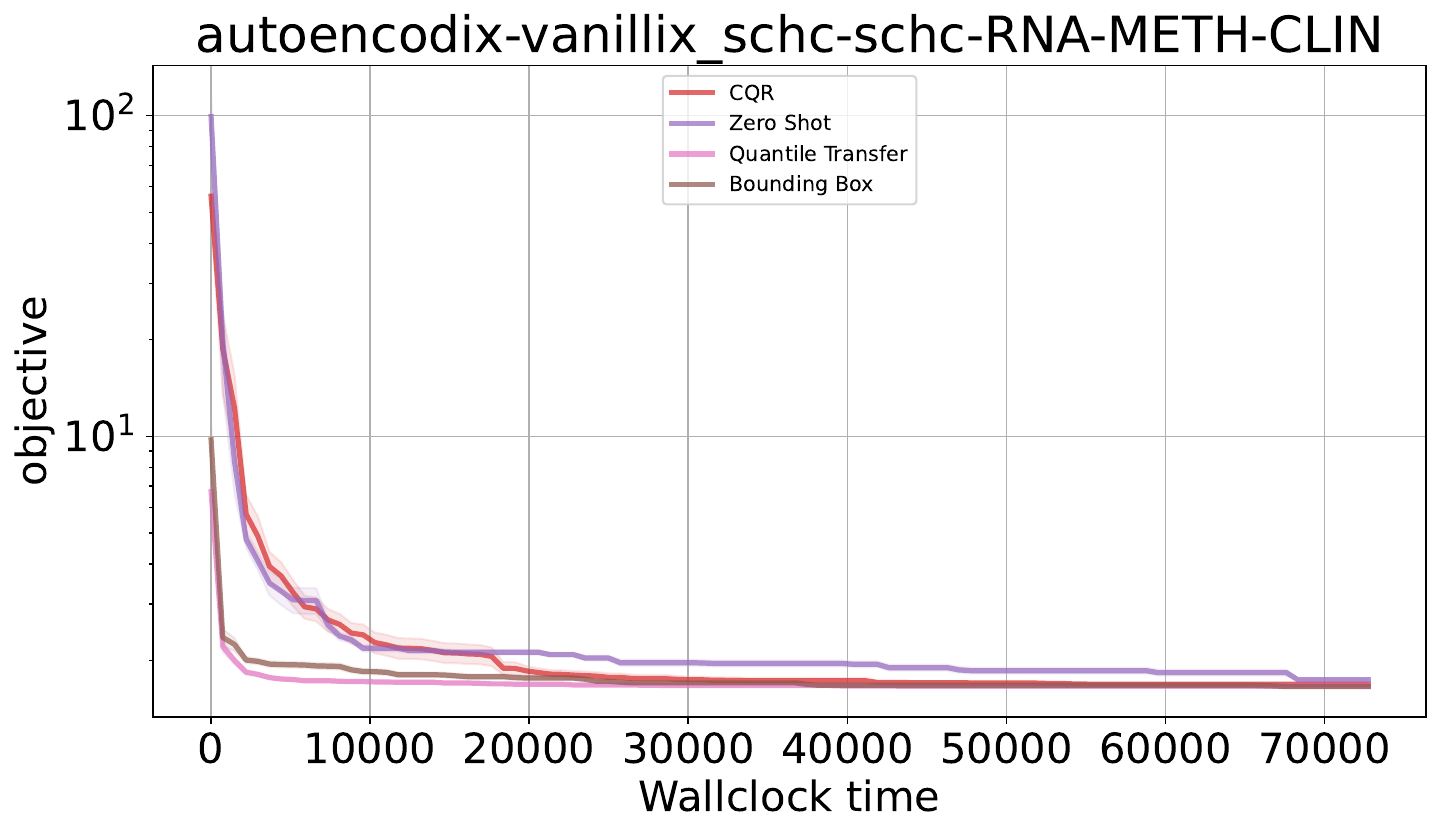} &
    \includegraphics[width=0.32\textwidth]{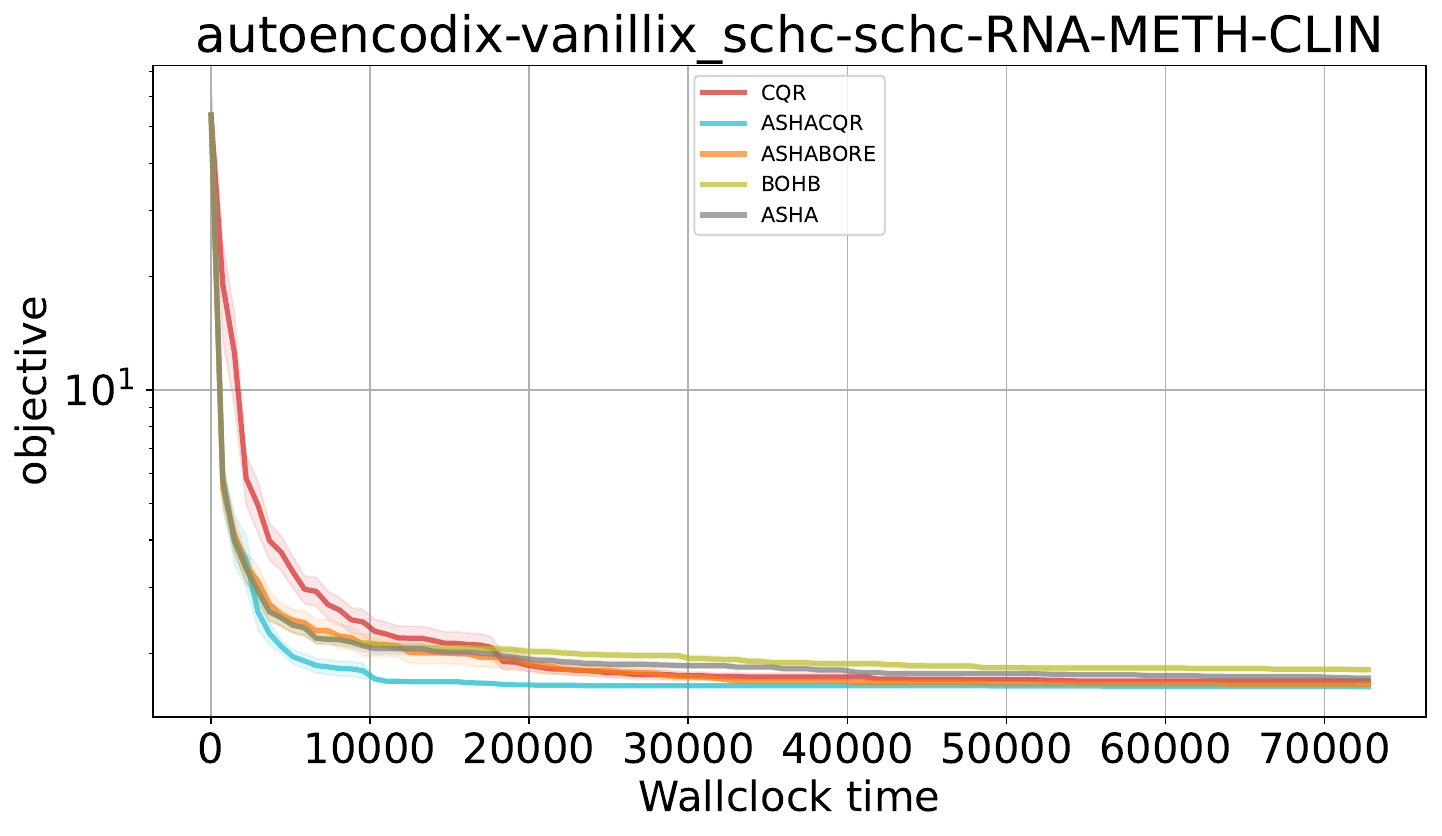} \\
    \includegraphics[width=0.32\textwidth]{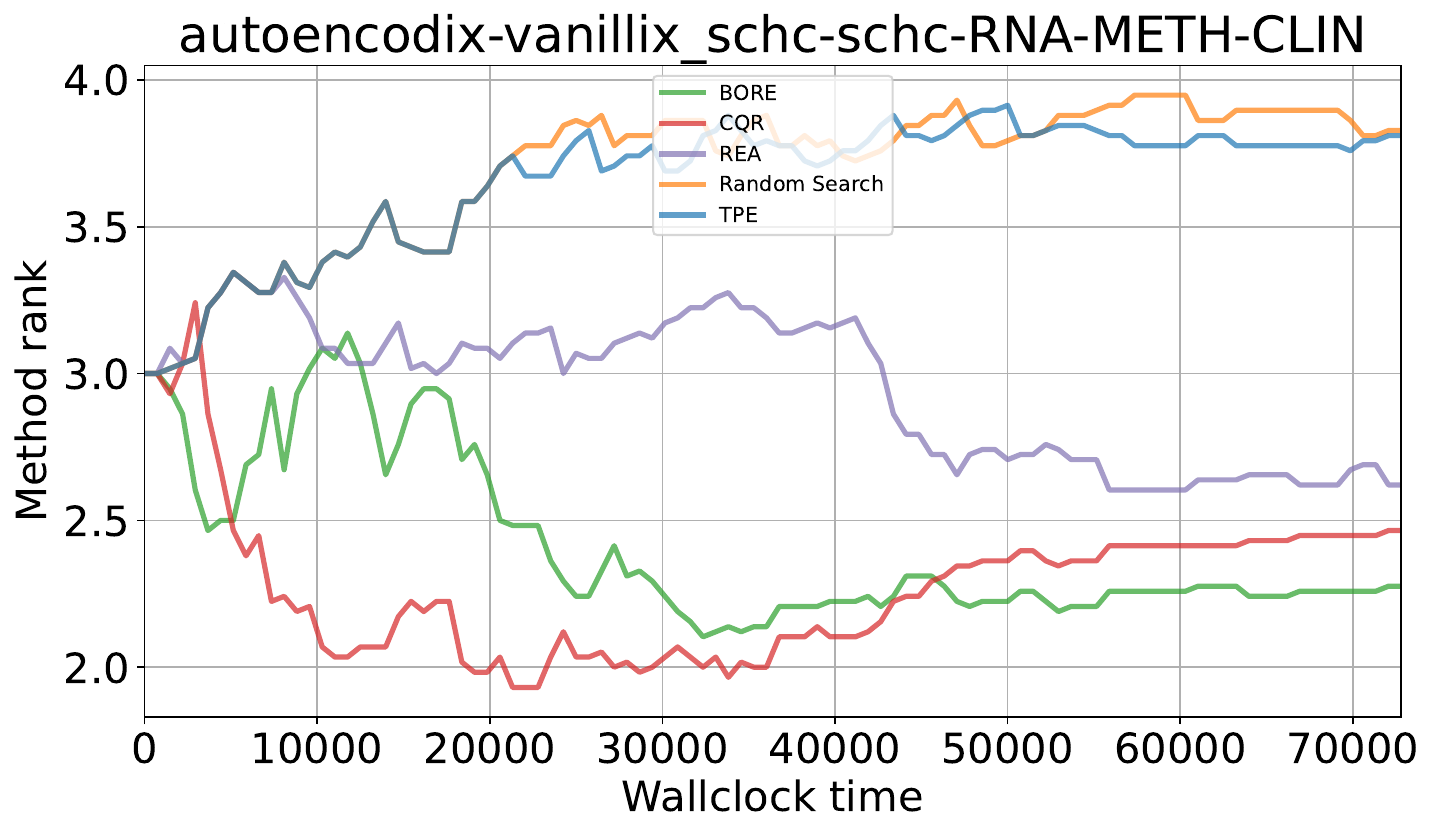} &
    \includegraphics[width=0.32\textwidth]{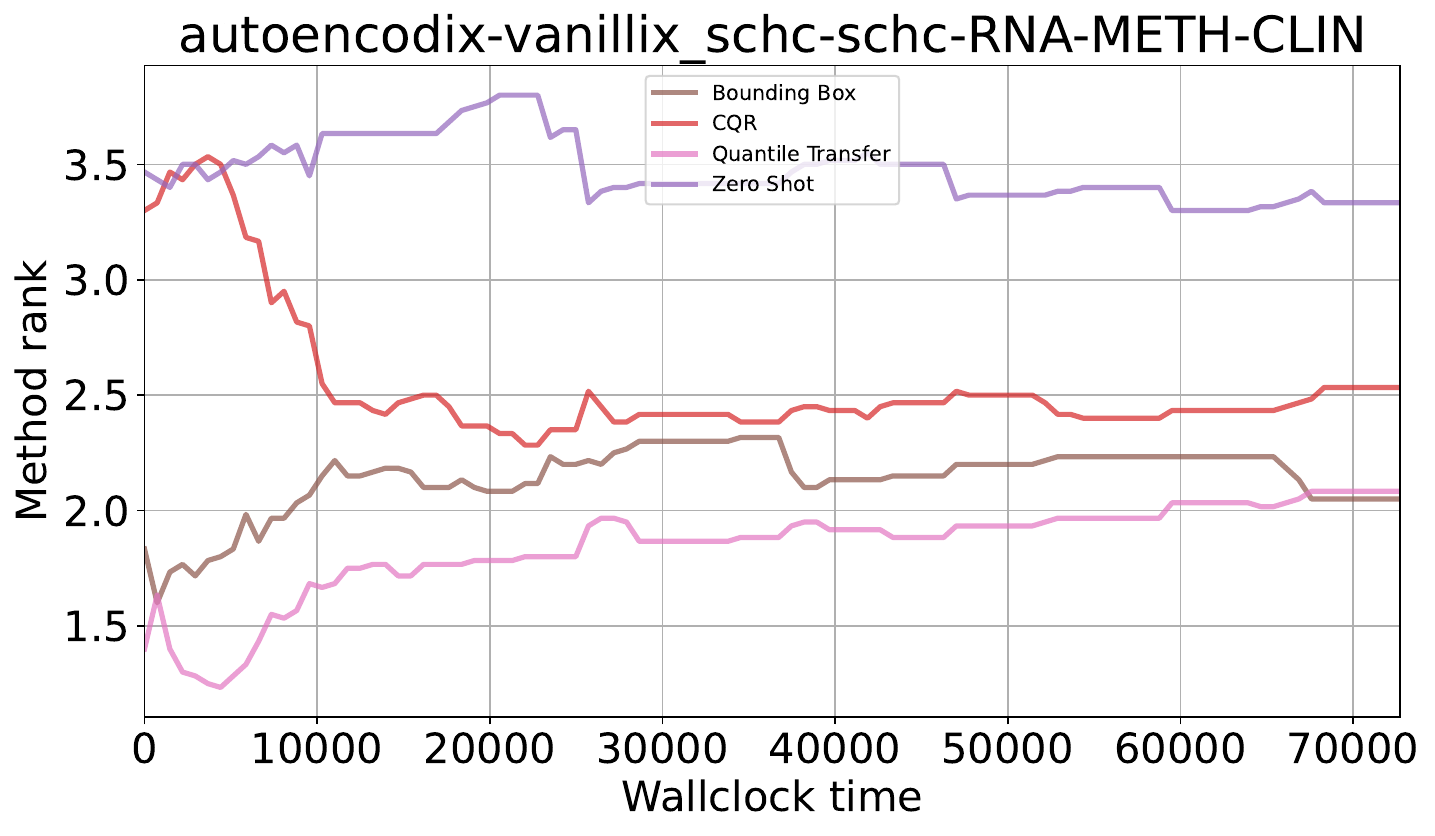} &
    \includegraphics[width=0.32\textwidth]{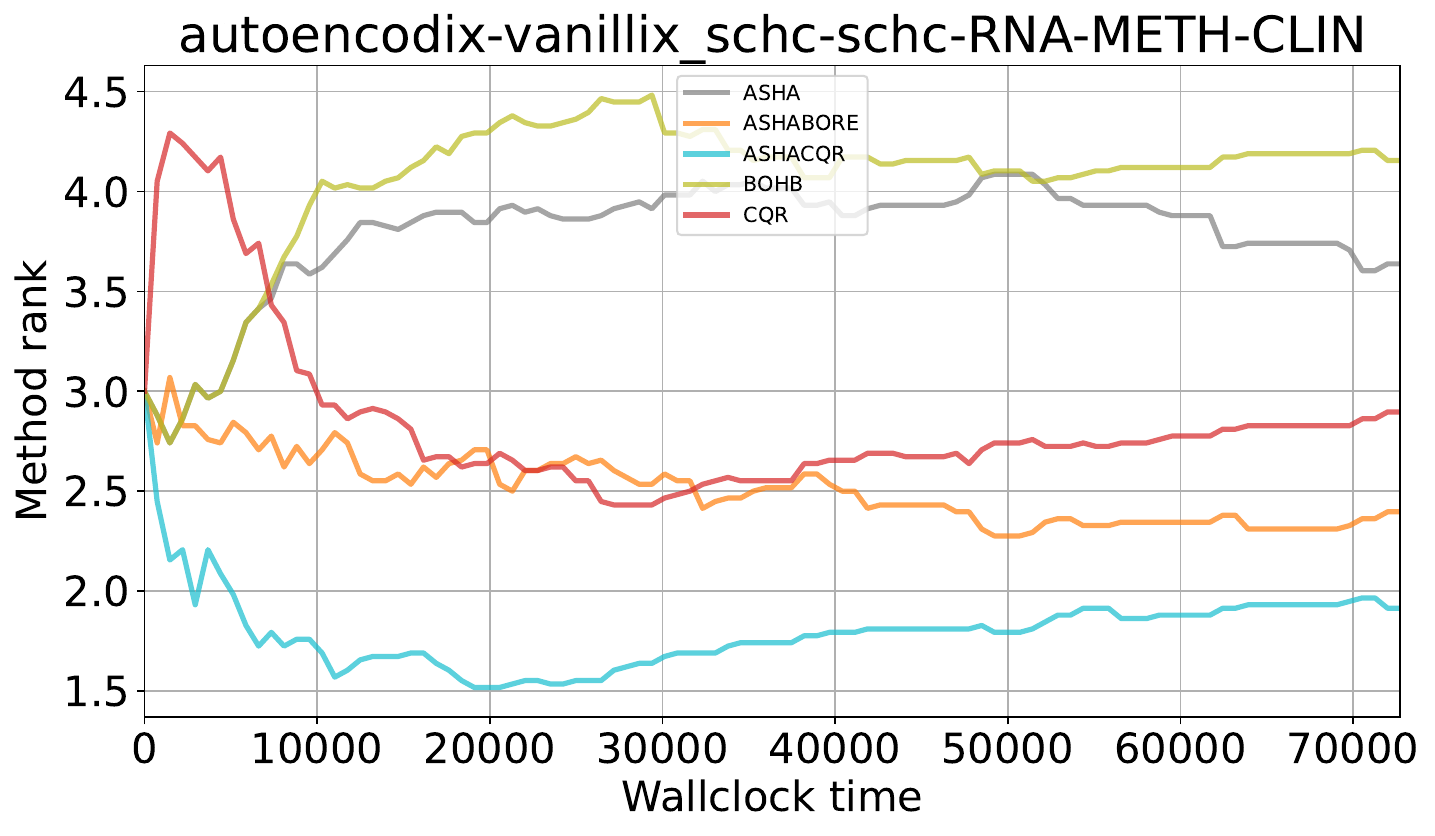} \\
    \end{tabular}
    \caption{Results for Vanillix tasks (Part 1).}
    \label{fig:vanillix_part1}
\end{figure}

\clearpage

\begin{figure}[htbp]
    \centering
    \setlength{\tabcolsep}{1pt}
    \begin{tabular}{ccc}
    \multicolumn{3}{c}{\textbf{autoencodix-vanillix\_tcga-tcga-DNA-CLIN}} \\
    \textbf{Single-Fidelity} & \textbf{Transfer Learning} & \textbf{Multi-Fidelity} \\
    \includegraphics[width=0.32\textwidth]{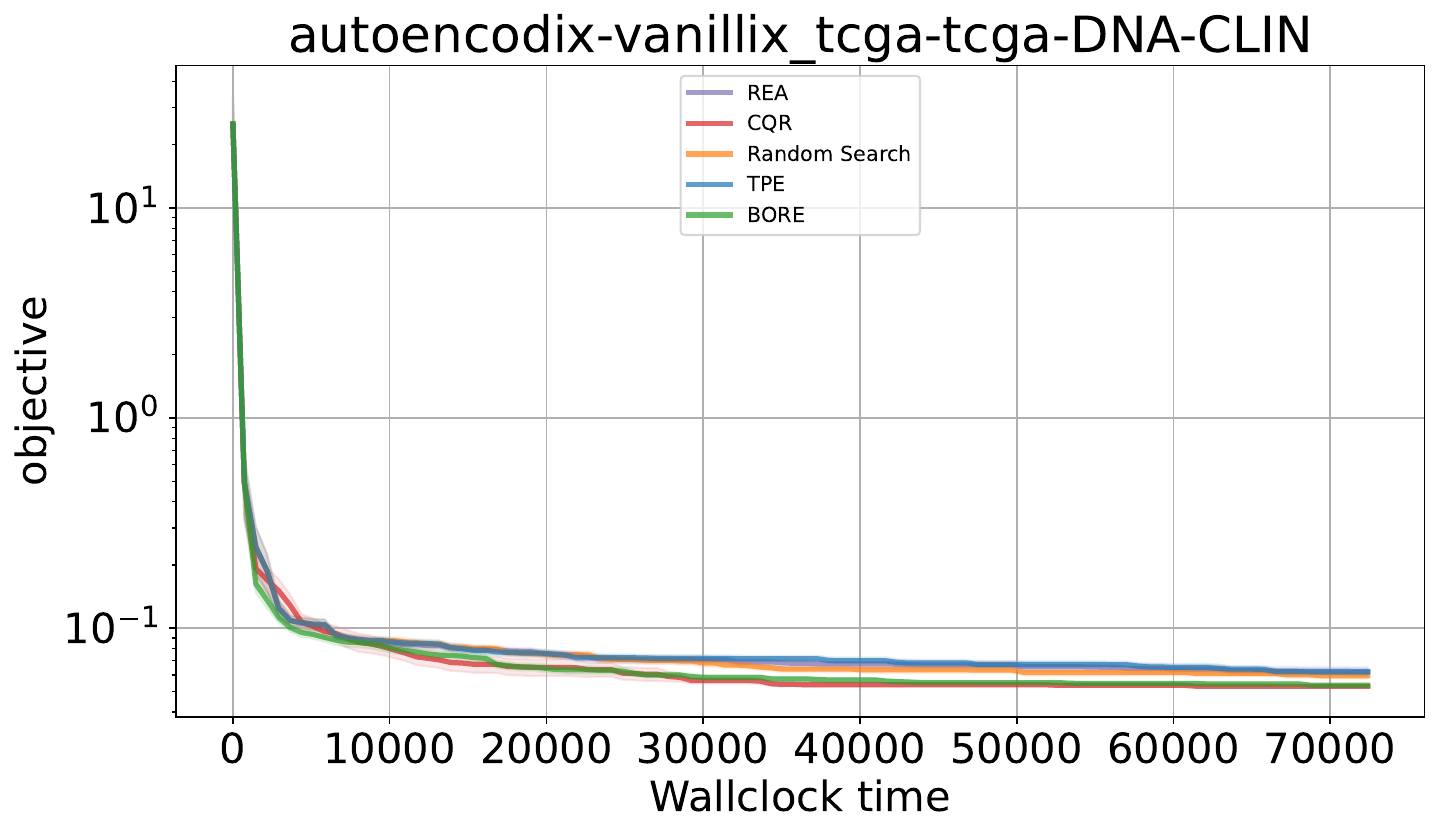} &
    \includegraphics[width=0.32\textwidth]{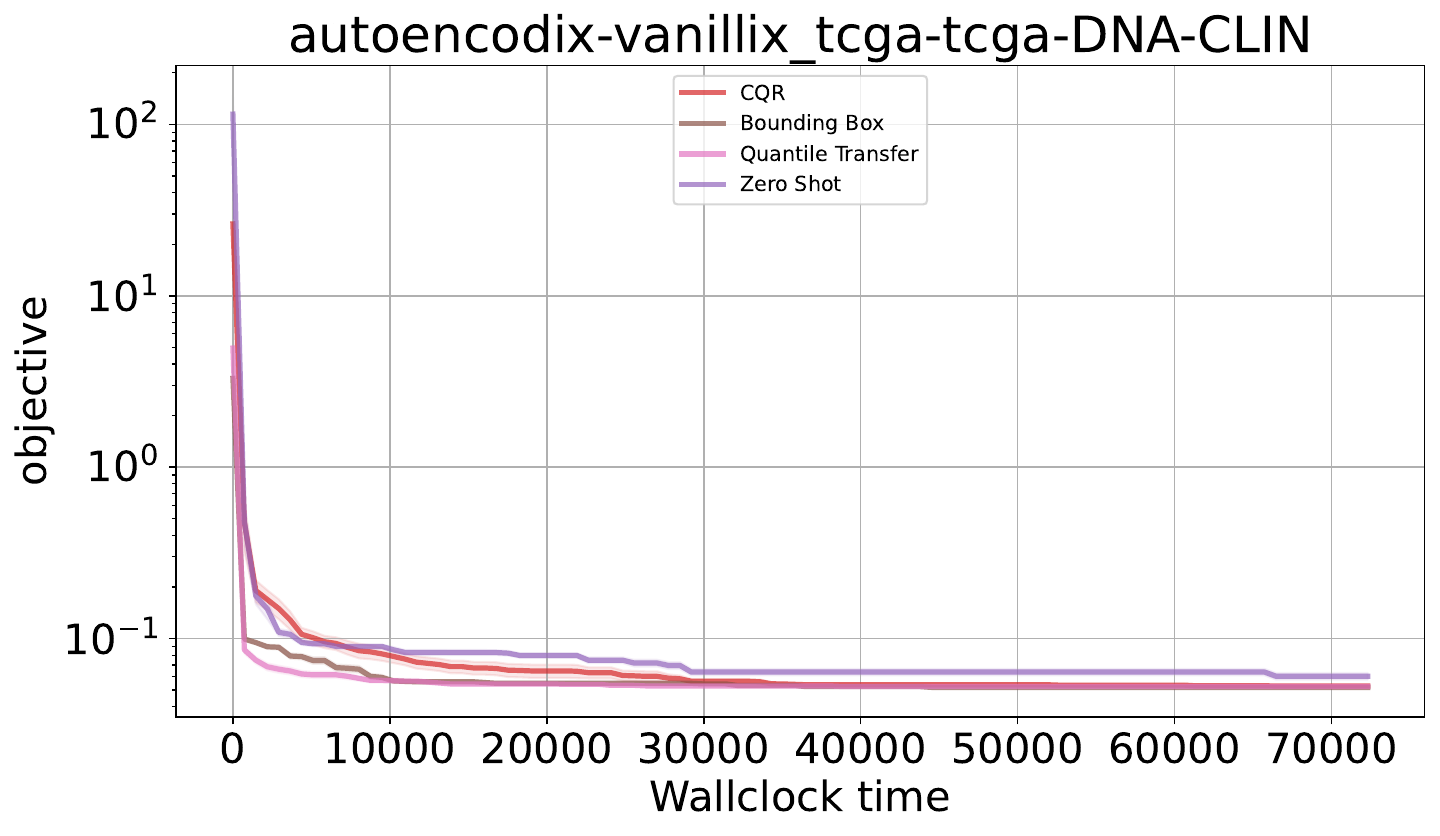} &
    \includegraphics[width=0.32\textwidth]{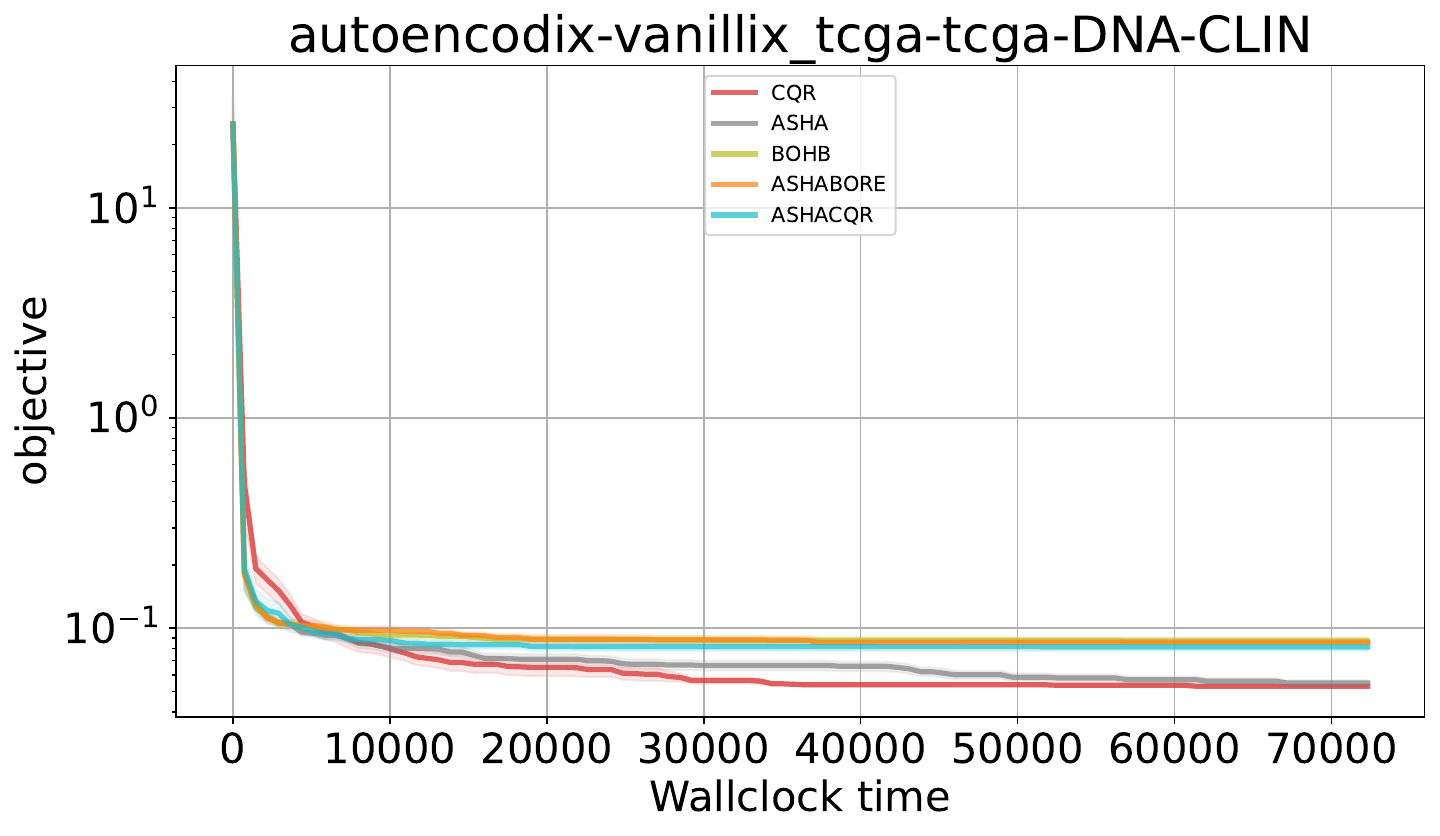} \\
    \includegraphics[width=0.32\textwidth]{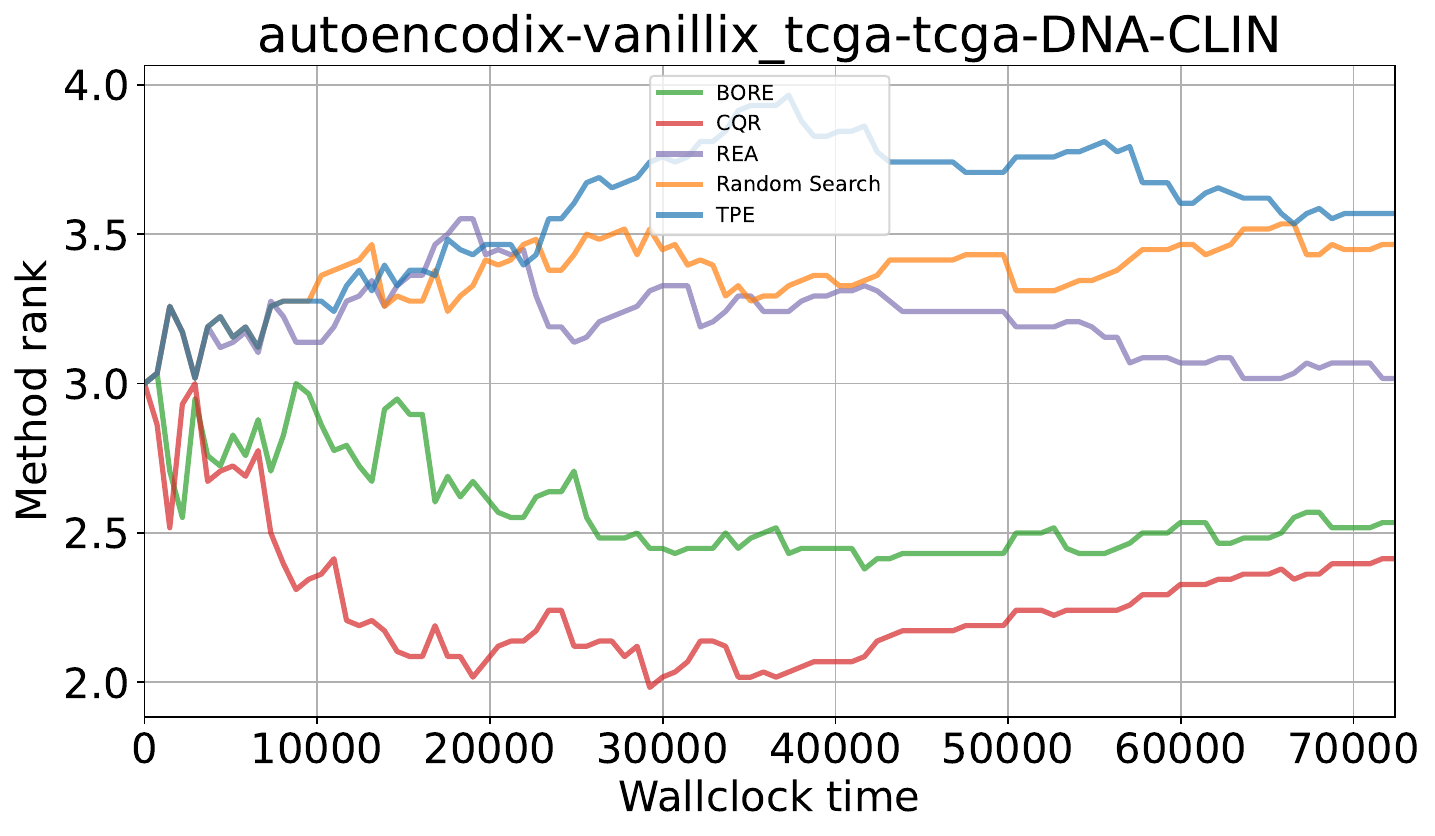} &
    \includegraphics[width=0.32\textwidth]{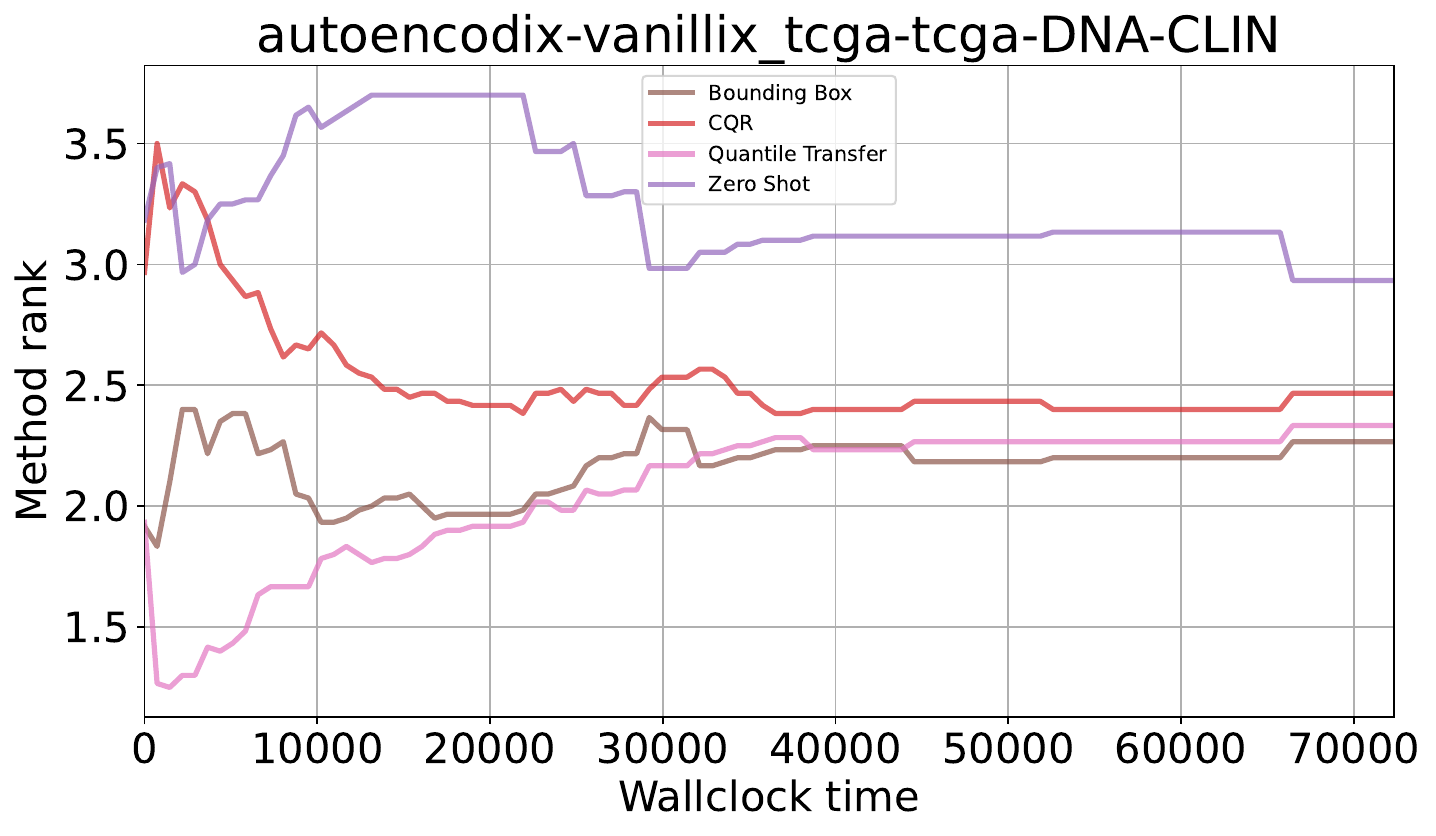} &
    \includegraphics[width=0.32\textwidth]{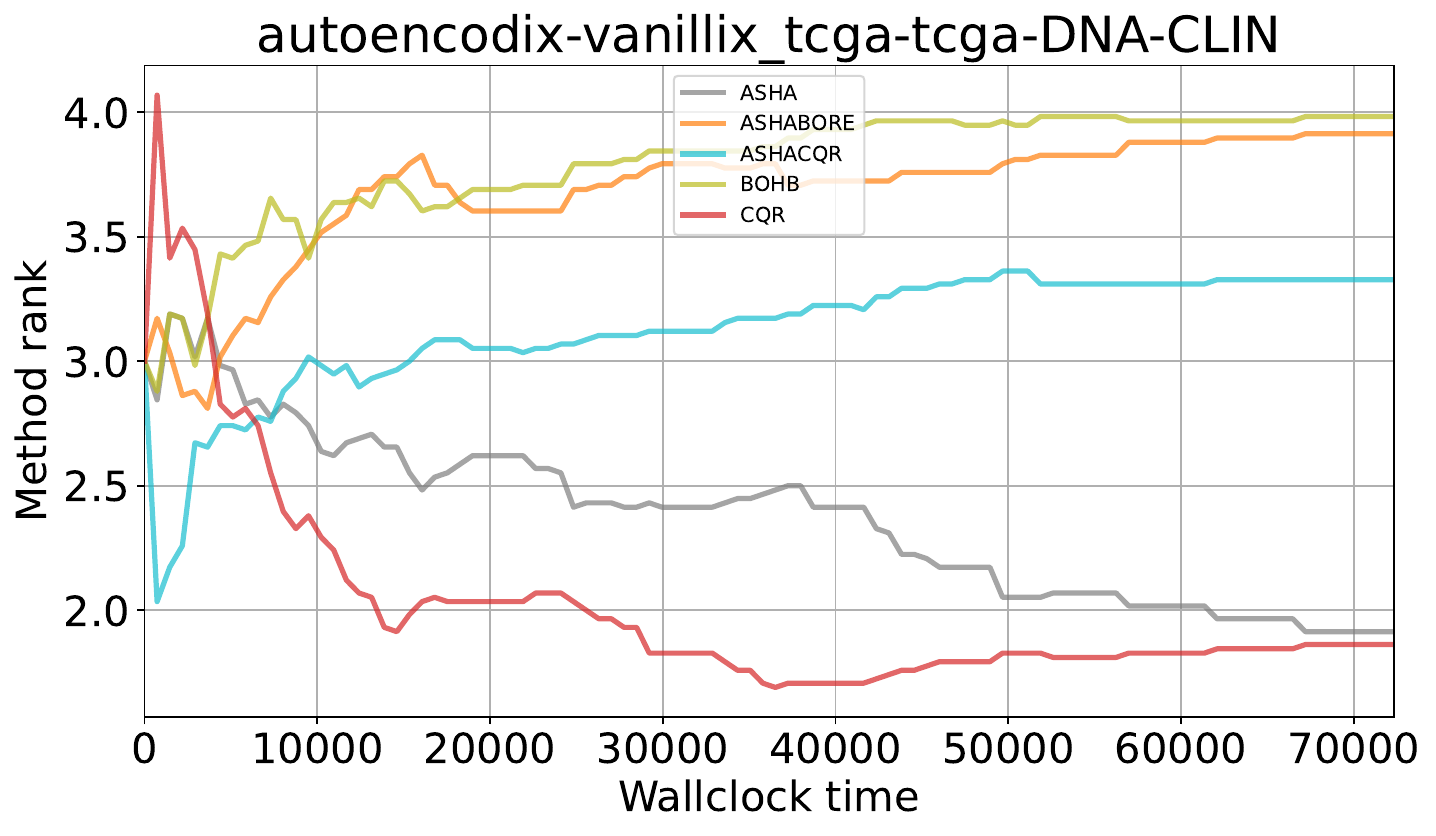} \\
    \midrule
    \multicolumn{3}{c}{\textbf{autoencodix-vanillix\_tcga-tcga-METH-CLIN}} \\
    \includegraphics[width=0.32\textwidth]{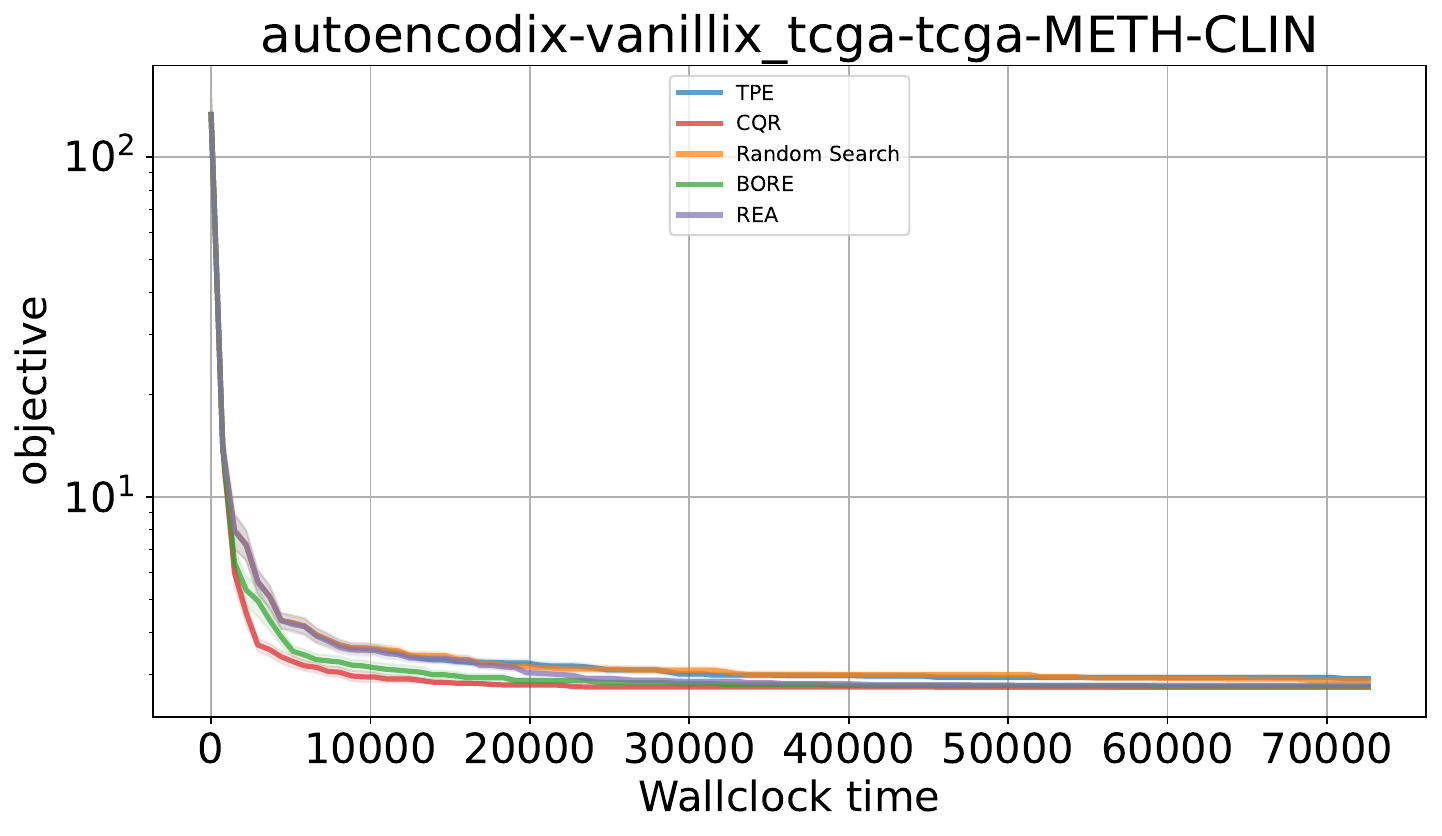} &
    \includegraphics[width=0.32\textwidth]{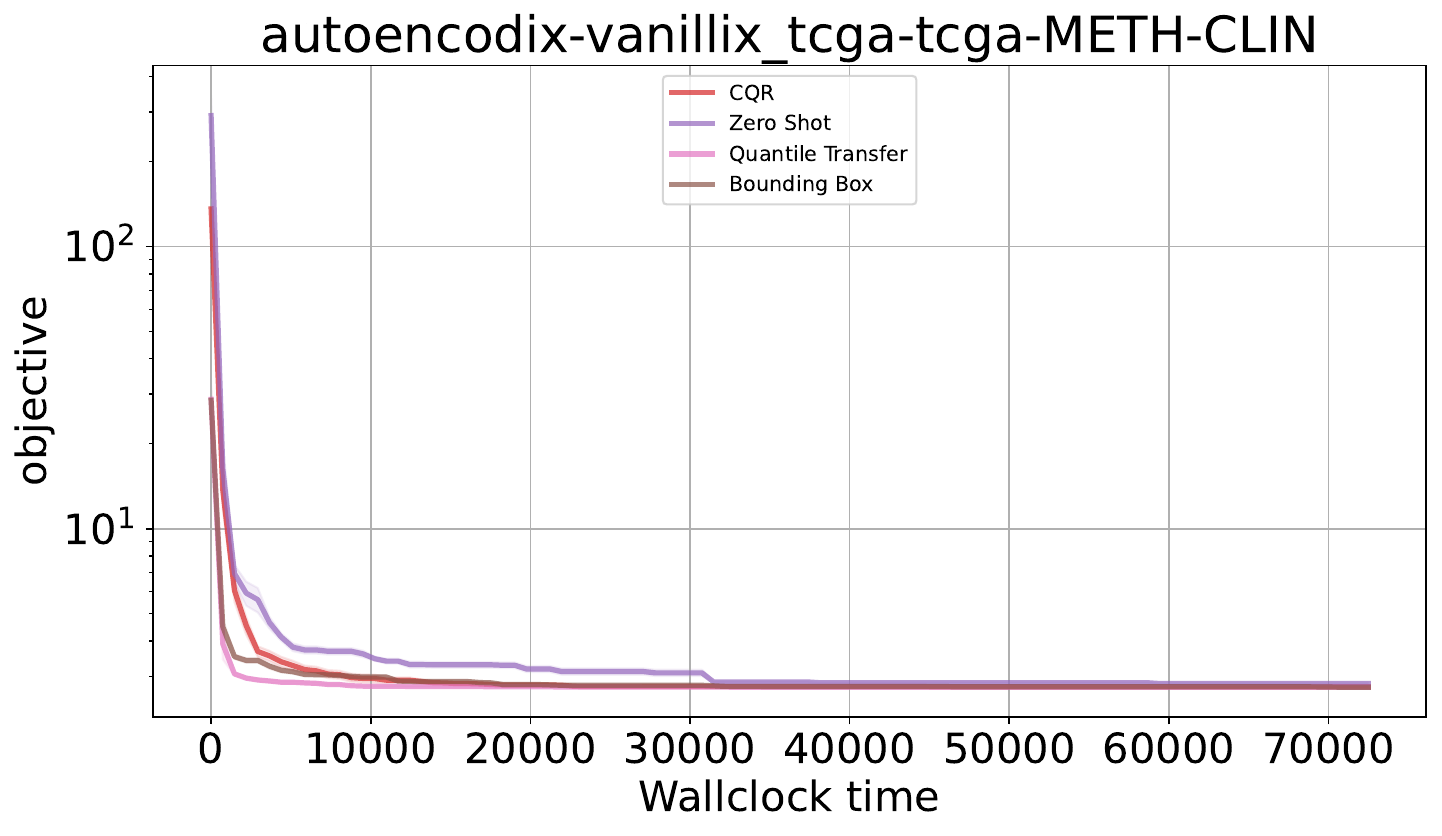} &
    \includegraphics[width=0.32\textwidth]{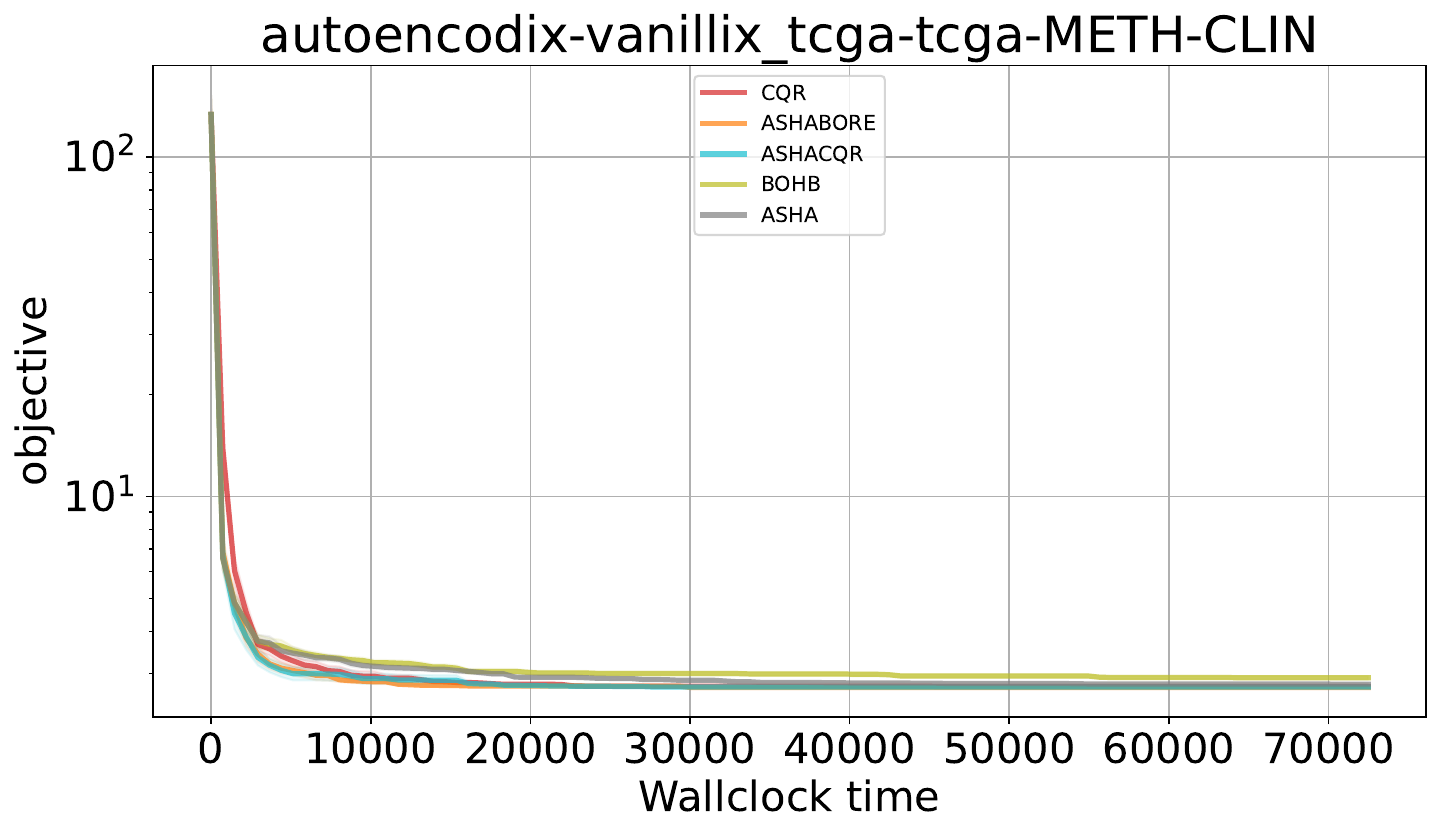} \\
    \includegraphics[width=0.32\textwidth]{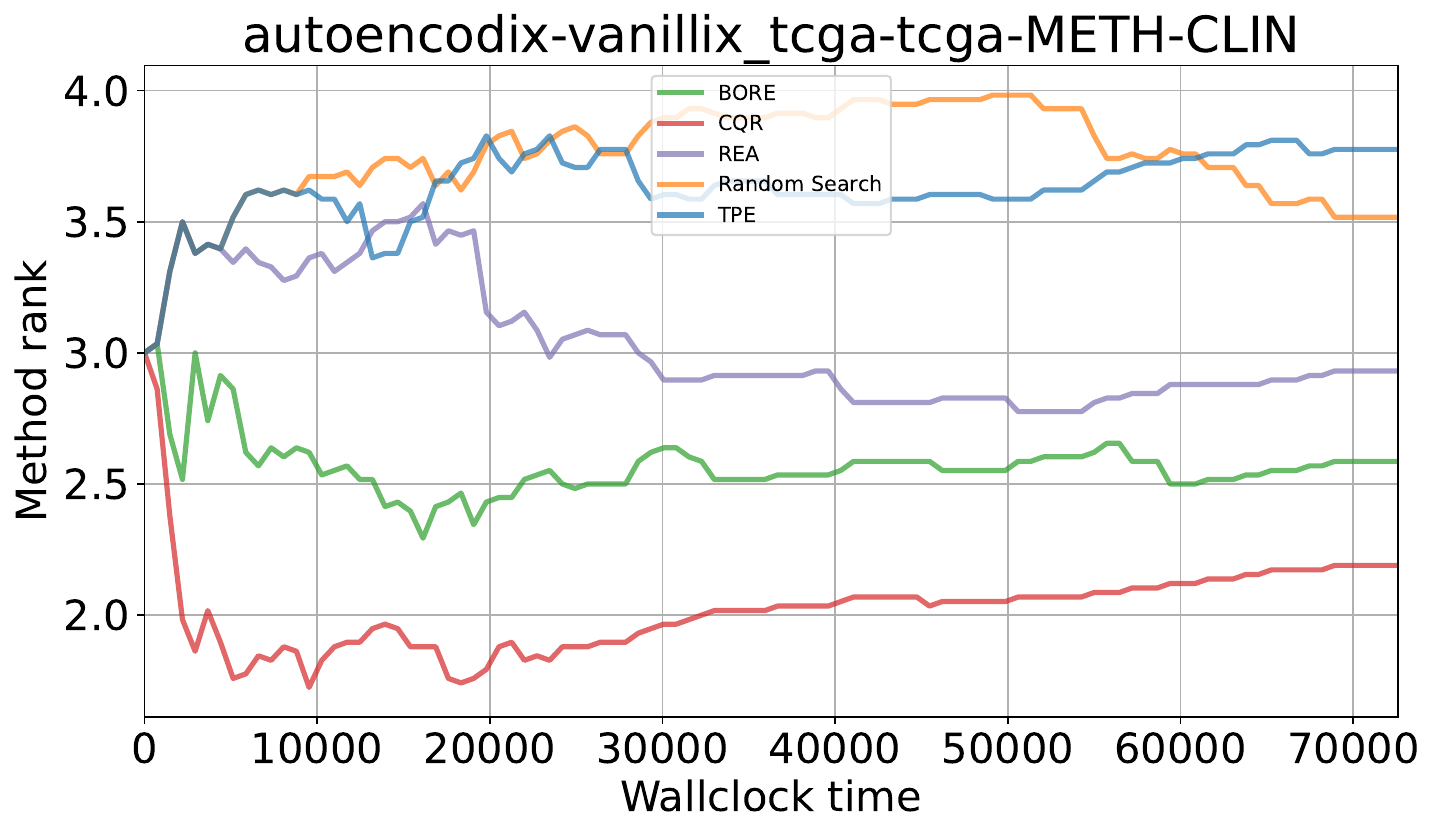} &
    \includegraphics[width=0.32\textwidth]{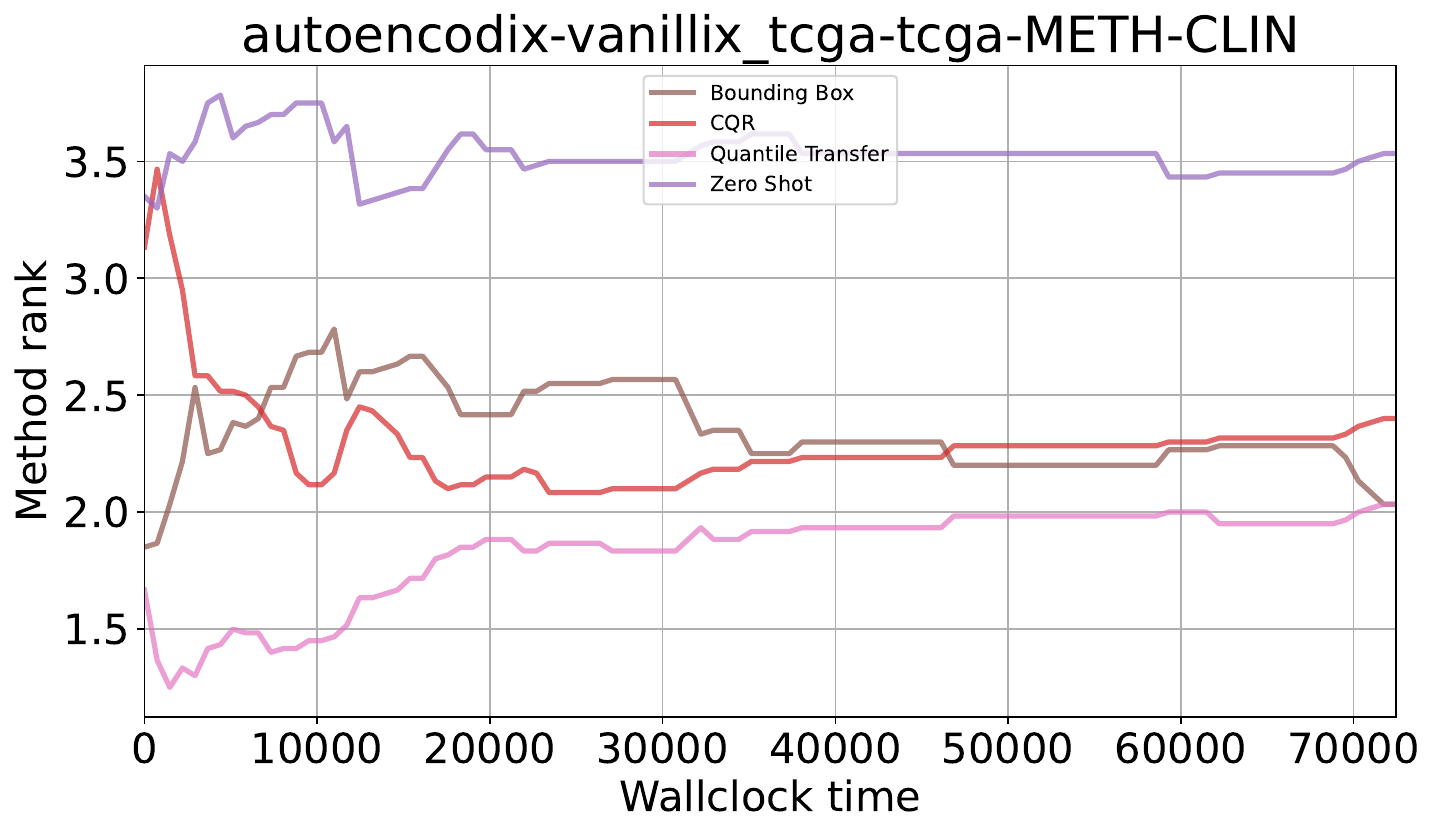} &
    \includegraphics[width=0.32\textwidth]{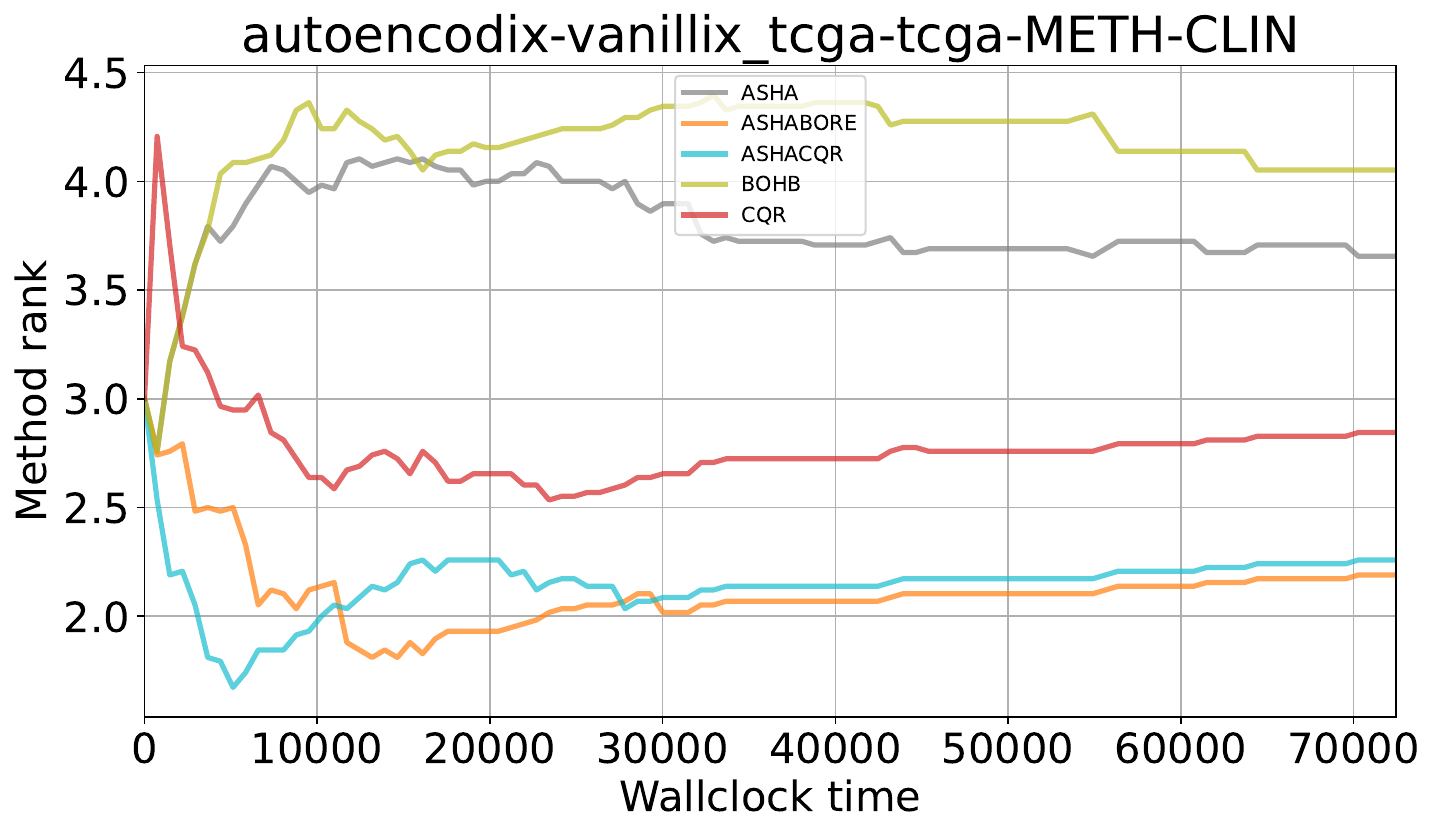} \\
    \midrule
    \multicolumn{3}{c}{\textbf{autoencodix-vanillix\_tcga-tcga-RNA-CLIN}} \\
    \includegraphics[width=0.32\textwidth]{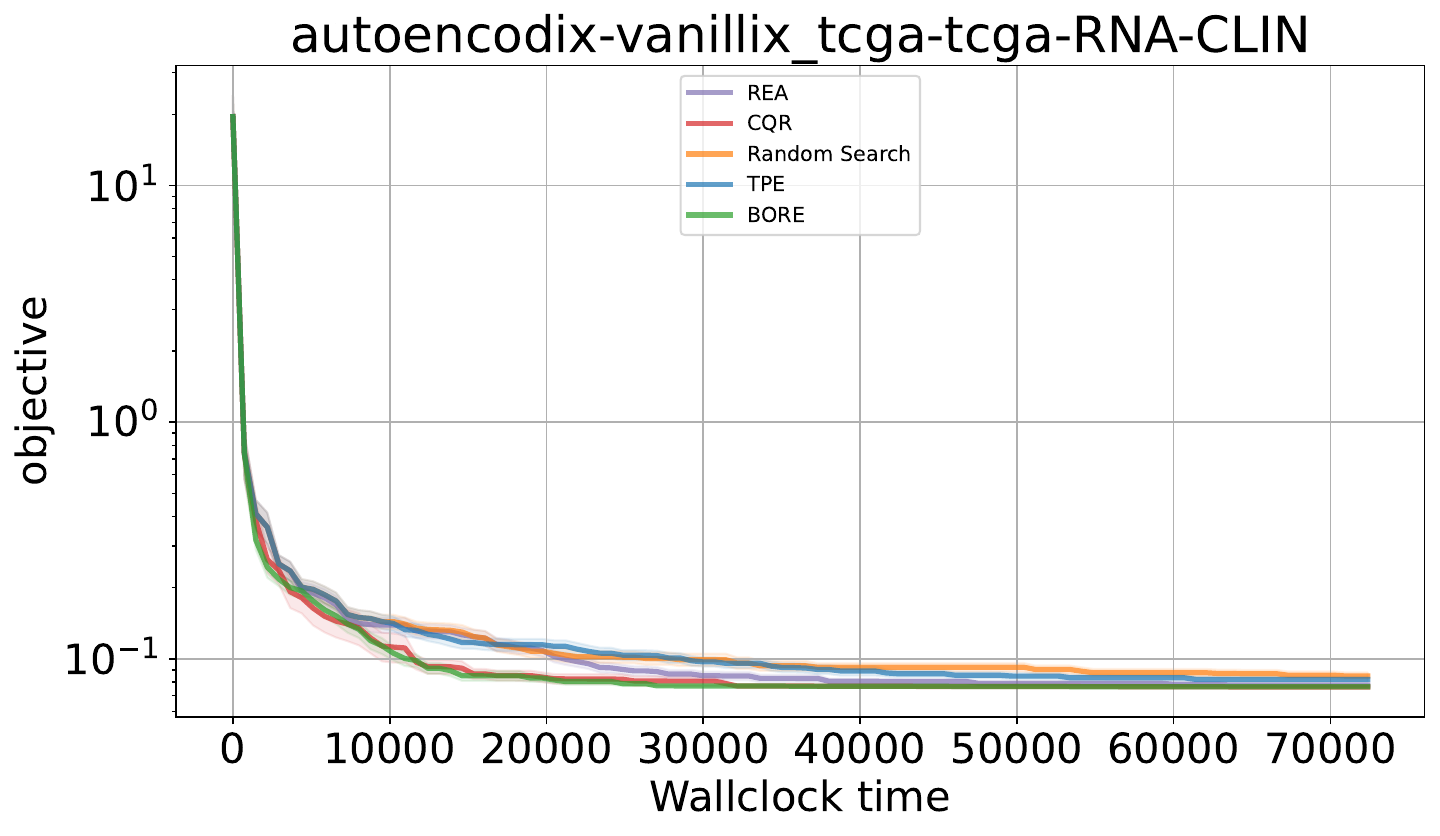} &
    \includegraphics[width=0.32\textwidth]{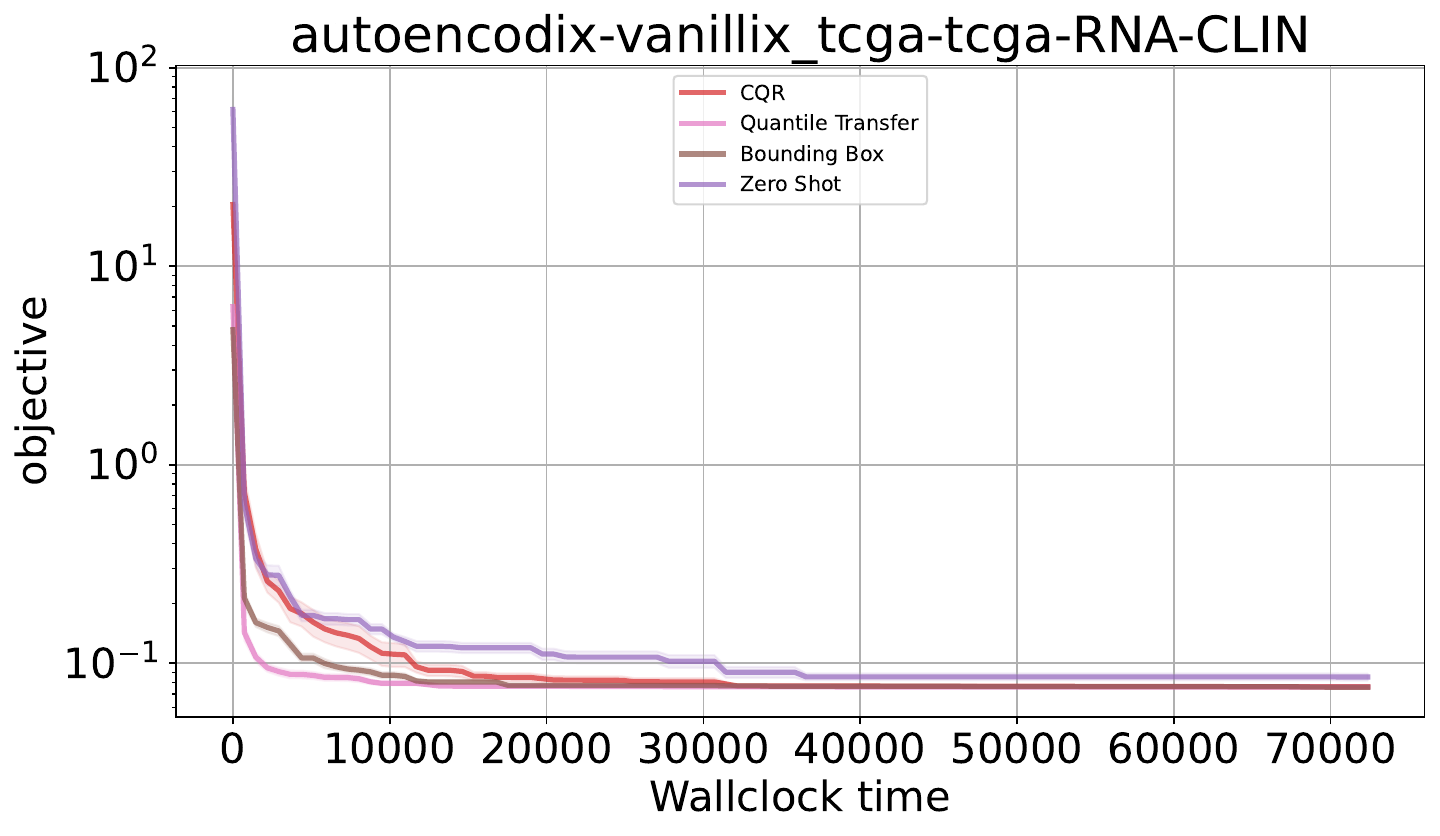} &
    \includegraphics[width=0.32\textwidth]{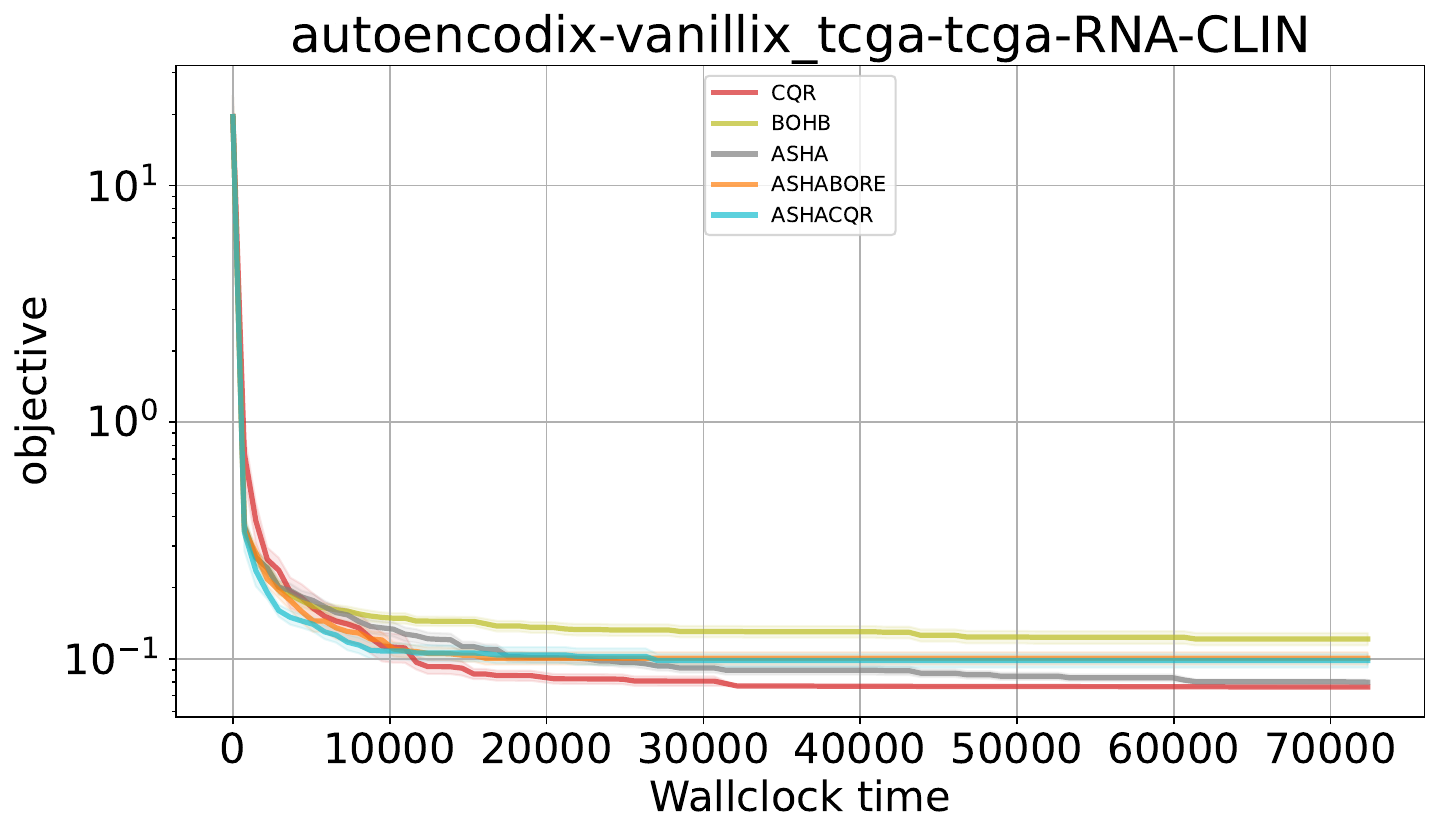} \\
    \includegraphics[width=0.32\textwidth]{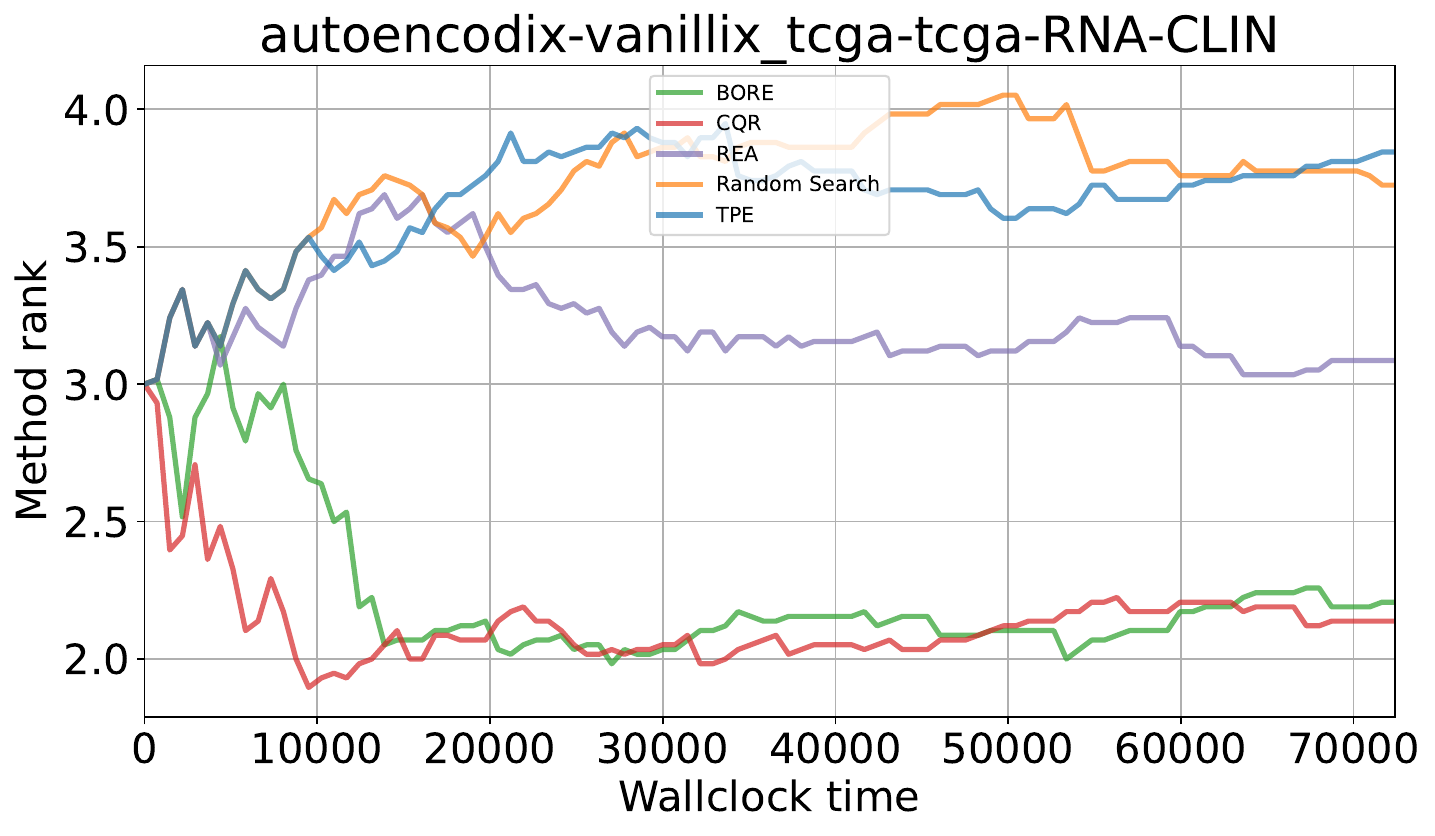} &
    \includegraphics[width=0.32\textwidth]{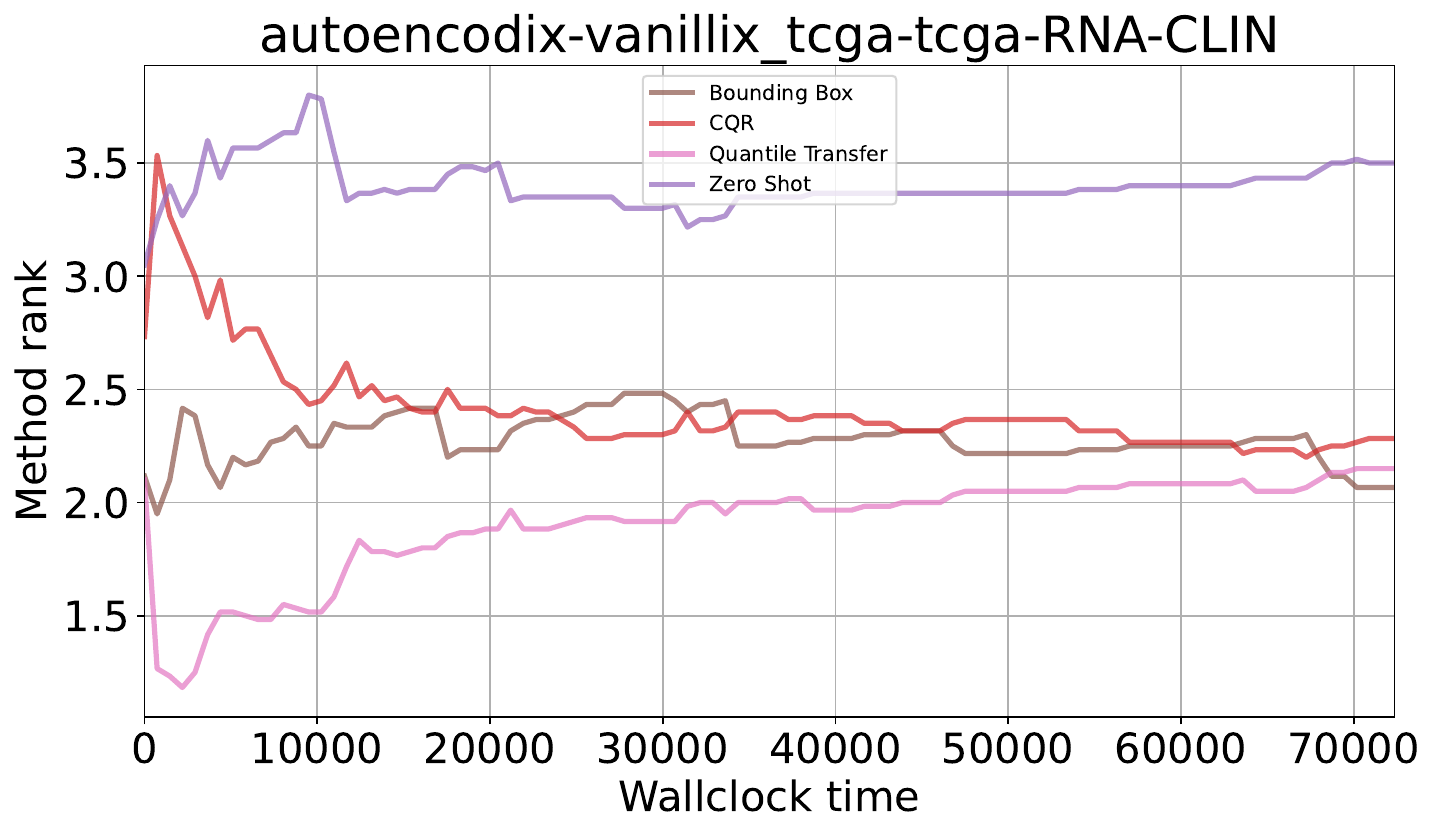} &
    \includegraphics[width=0.32\textwidth]{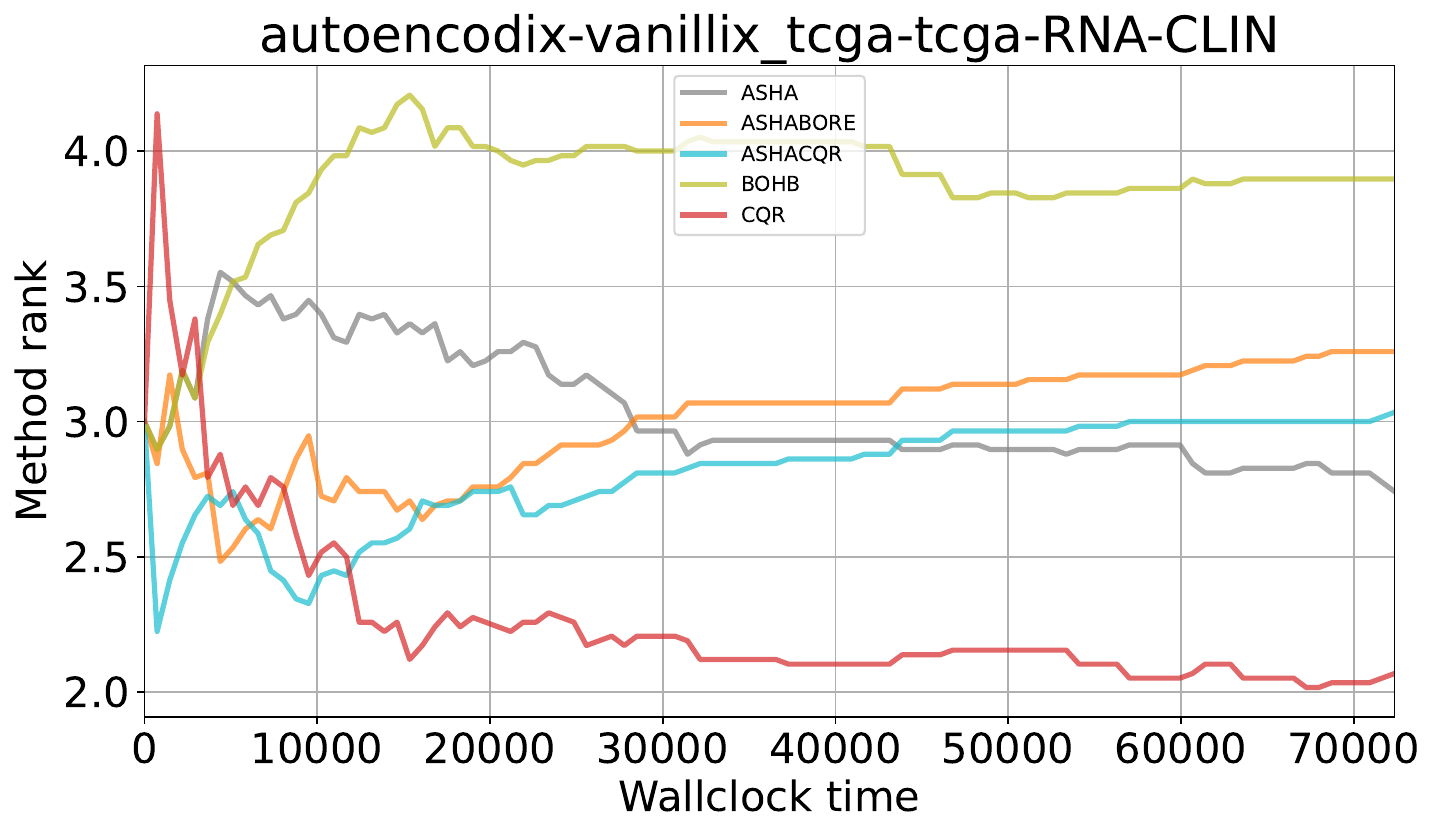} \\
    \end{tabular}
    \caption{Results for Vanillix tasks (Part 2).}
    \label{fig:vanillix_part2}
\end{figure}

\clearpage

\begin{figure}[htbp]
    \centering
    \setlength{\tabcolsep}{1pt}
    \begin{tabular}{ccc}
    \multicolumn{3}{c}{\textbf{autoencodix-vanillix\_tcga-tcga-RNA-DNA-METH-CLIN}} \\
    \textbf{Single-Fidelity} & \textbf{Transfer Learning} & \textbf{Multi-Fidelity} \\
    \includegraphics[width=0.32\textwidth]{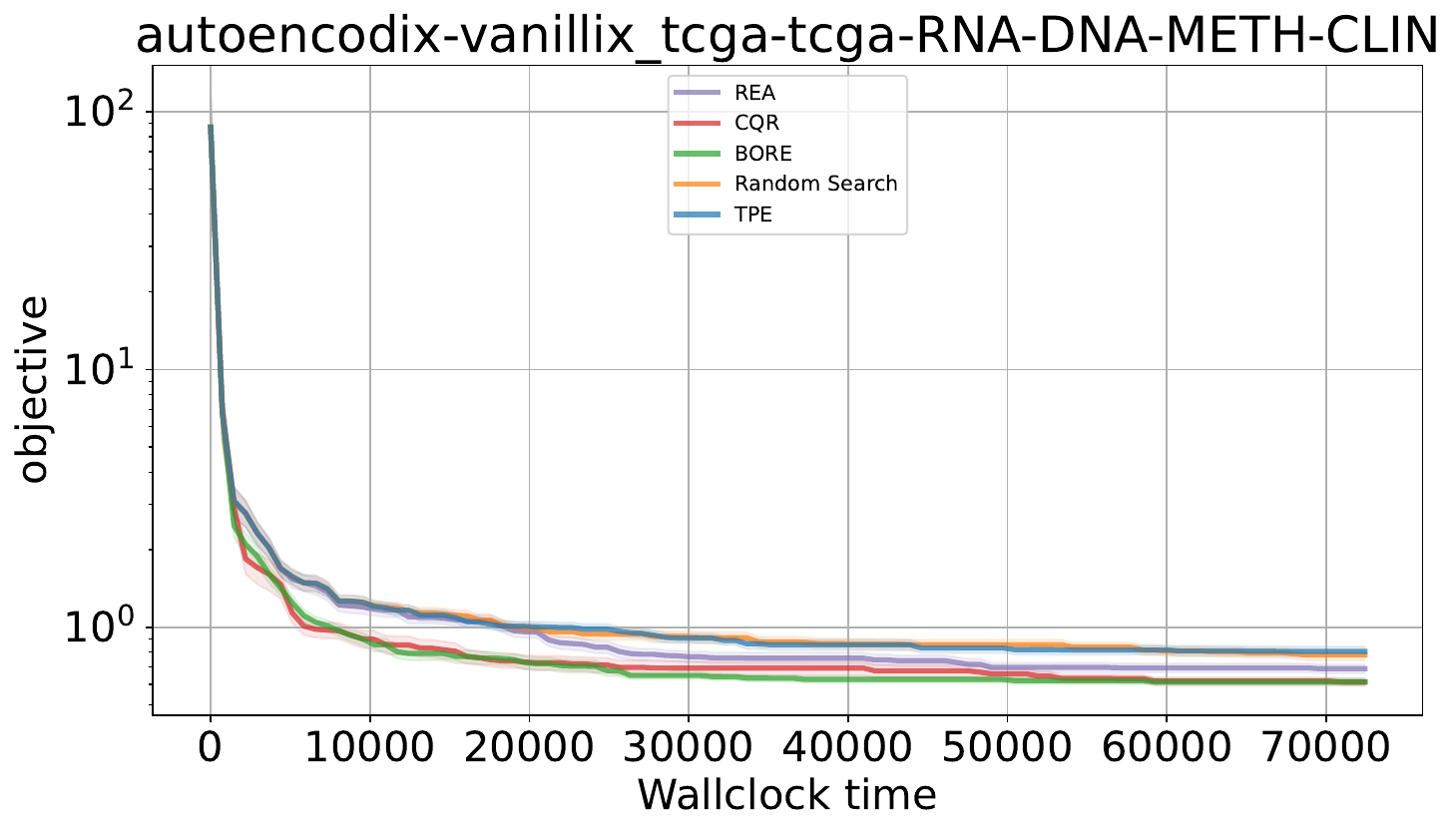} &
    \includegraphics[width=0.32\textwidth]{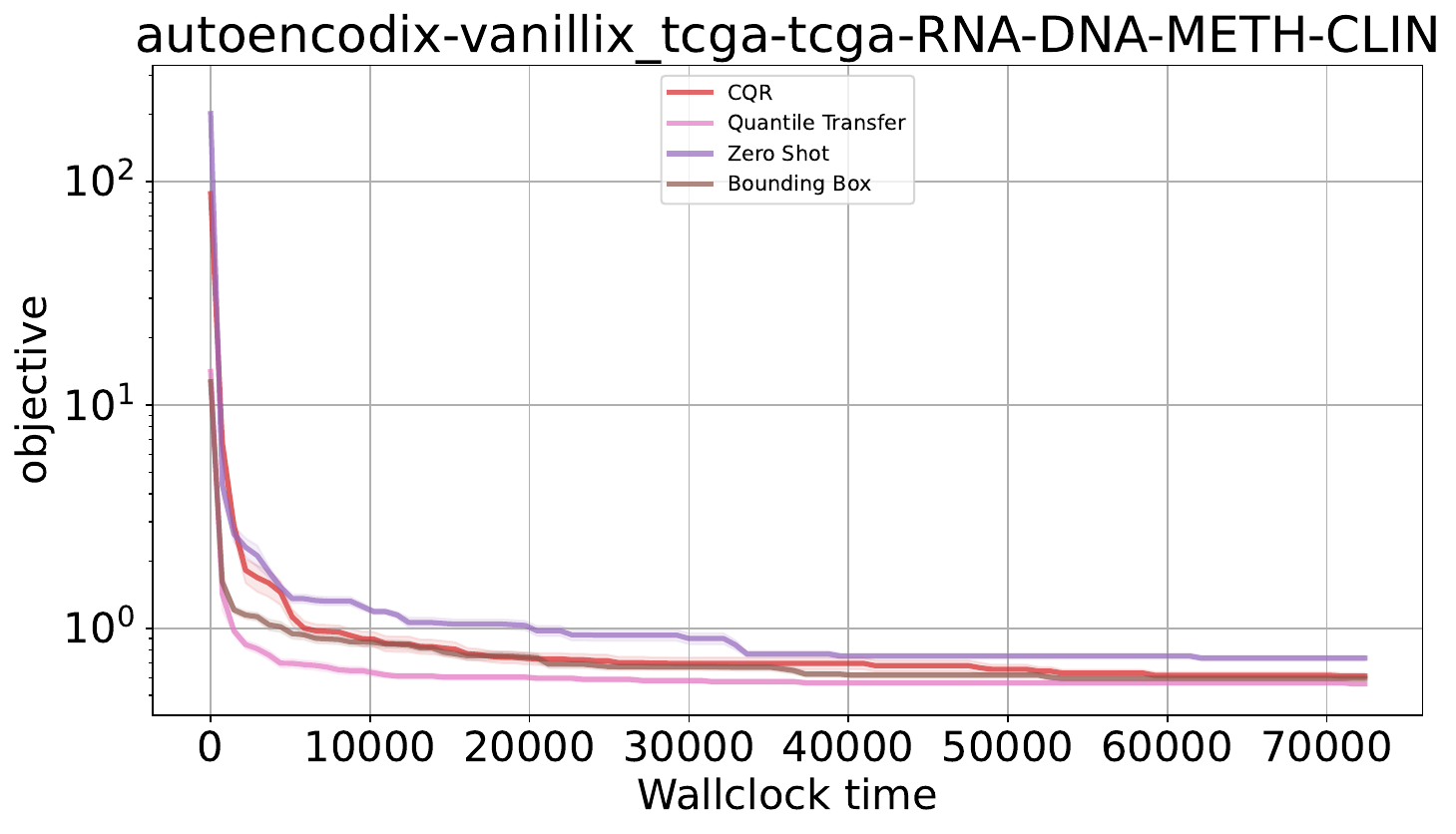} &
    \includegraphics[width=0.32\textwidth]{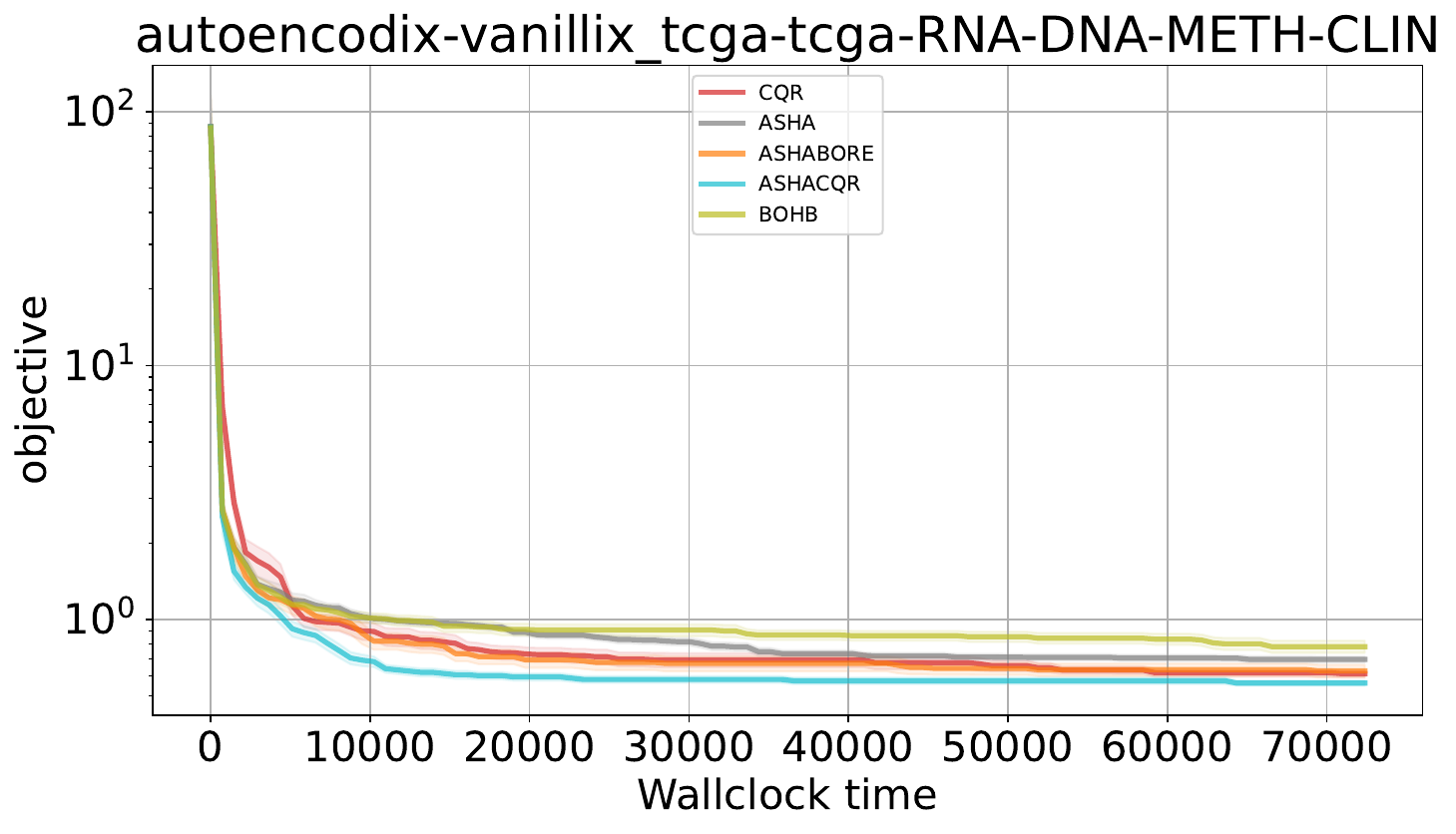} \\
    \includegraphics[width=0.32\textwidth]{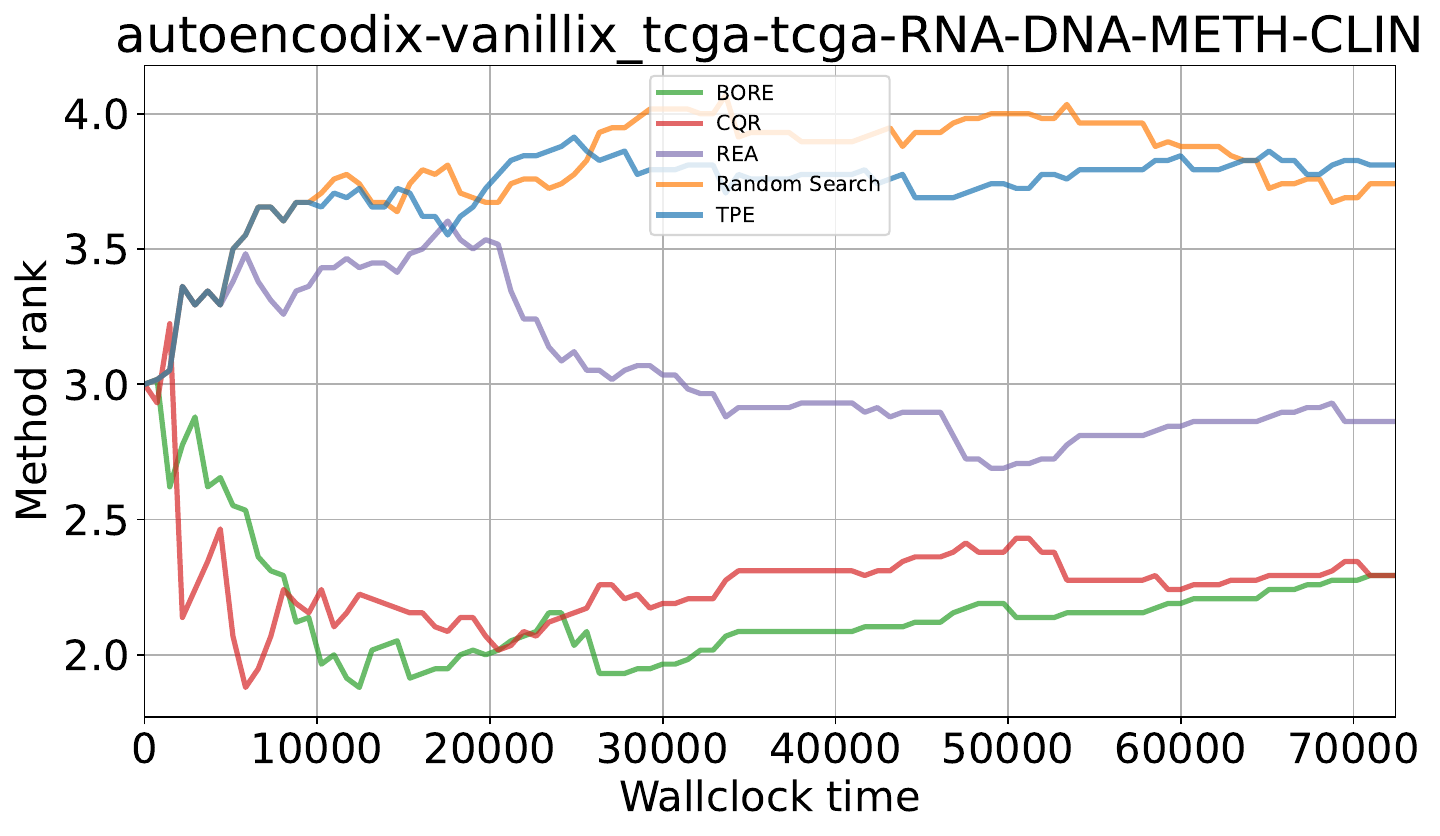} &
    \includegraphics[width=0.32\textwidth]{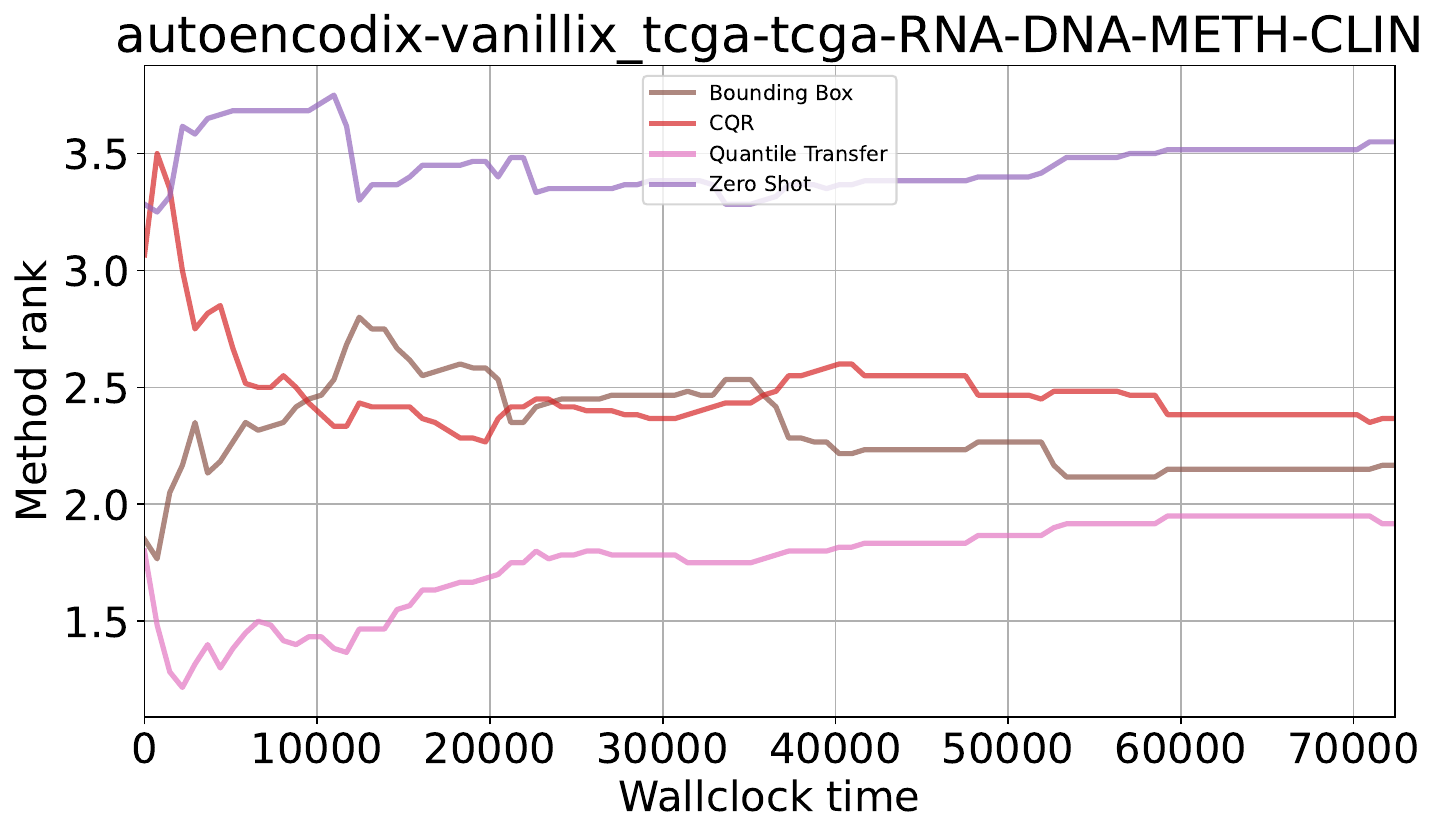} &
    \includegraphics[width=0.32\textwidth]{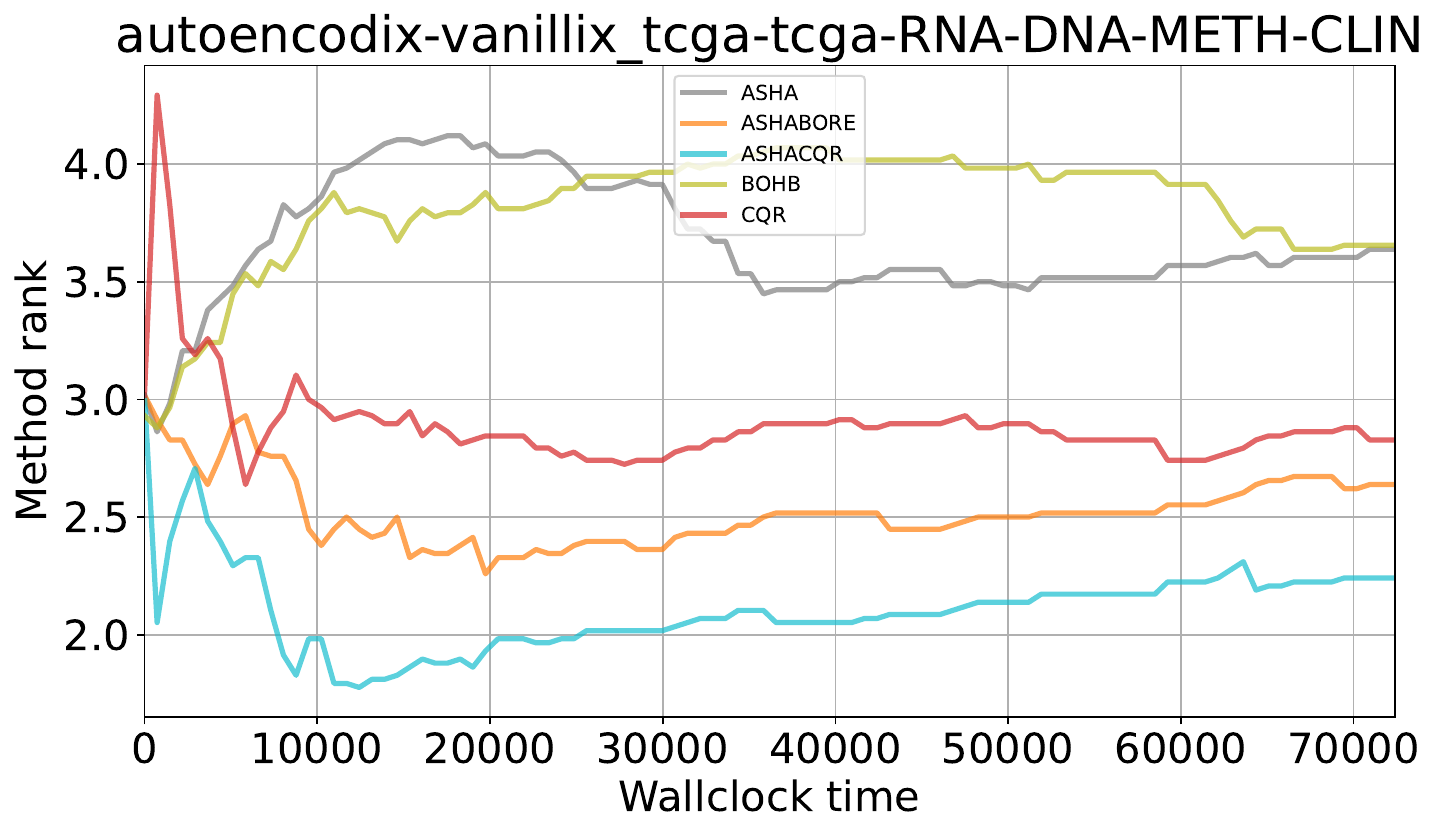} \\
    \end{tabular}
    \caption{Results for Vanillix (autoencodix-vanillix\_tcga-tcga-RNA-DNA-METH-CLIN).}
    \label{fig:vanillix_part3}
\end{figure}

\clearpage

% Model: Varix
\begin{figure}[htbp]
    \centering
    \setlength{\tabcolsep}{1pt}
    \begin{tabular}{ccc}
    \multicolumn{3}{c}{\textbf{autoencodix-varix\_schc-schc-METH-CLIN}} \\
    \textbf{Single-Fidelity} & \textbf{Transfer Learning} & \textbf{Multi-Fidelity} \\
    \includegraphics[width=0.32\textwidth]{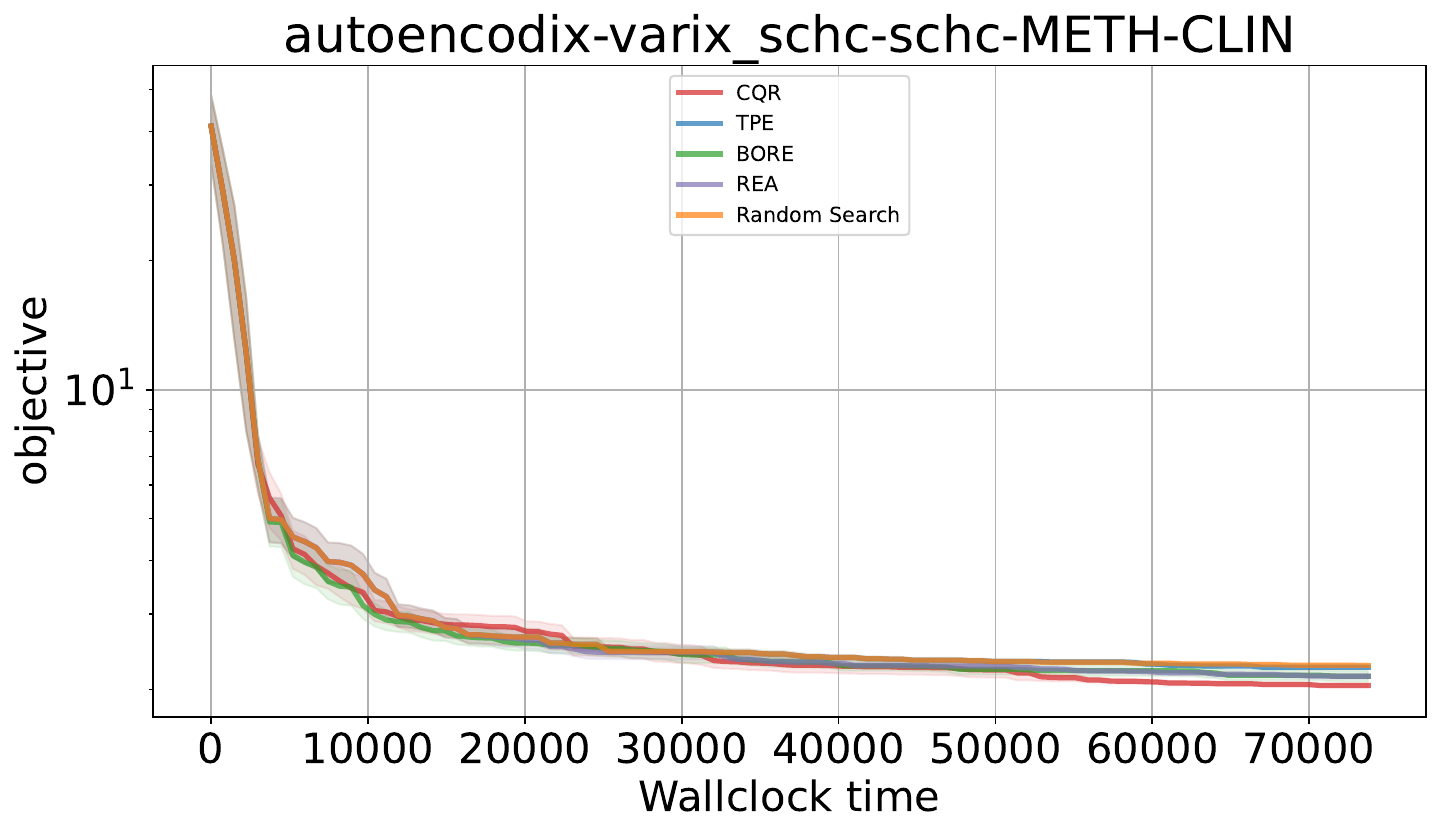} &
    \includegraphics[width=0.32\textwidth]{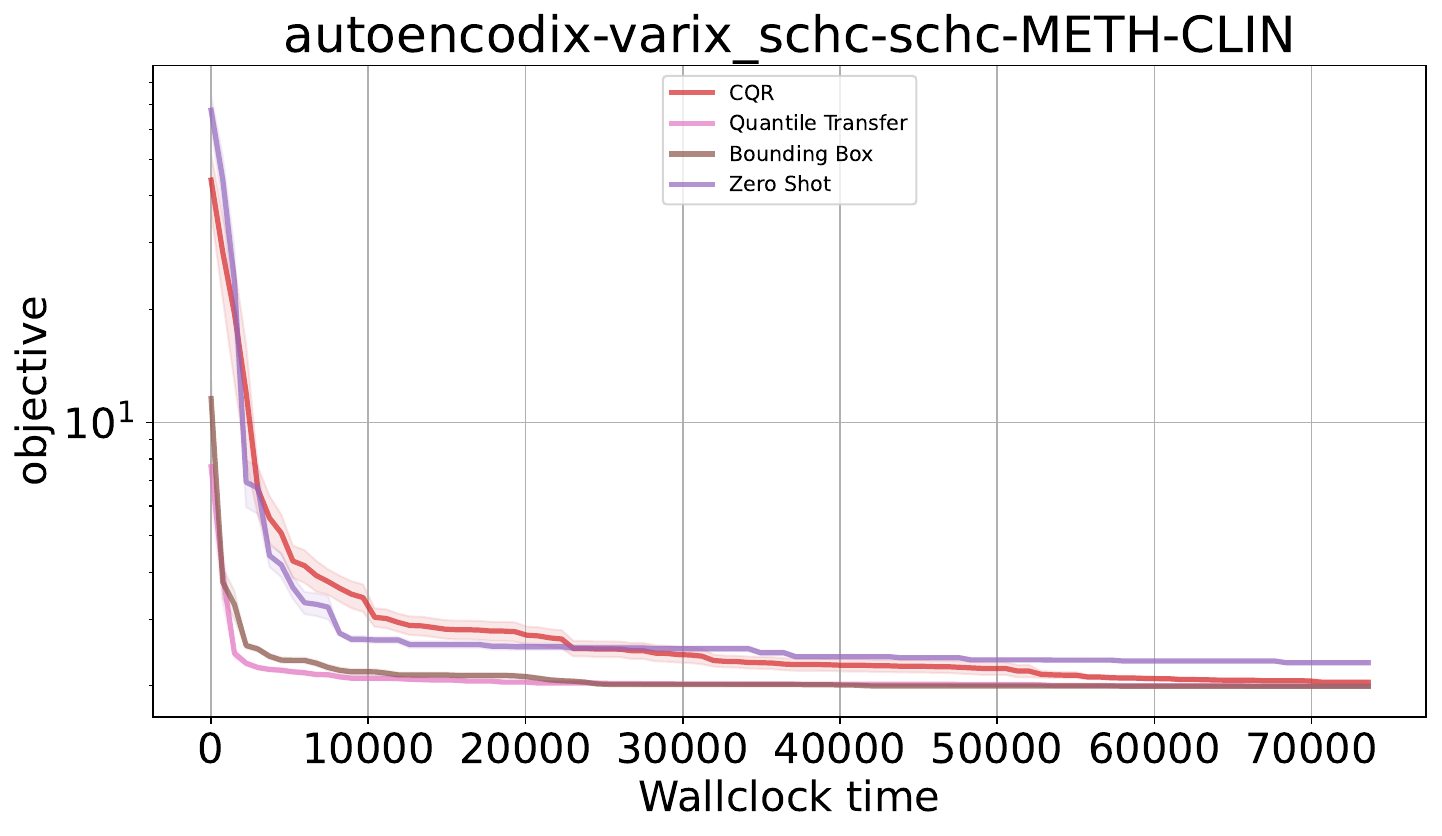} &
    \includegraphics[width=0.32\textwidth]{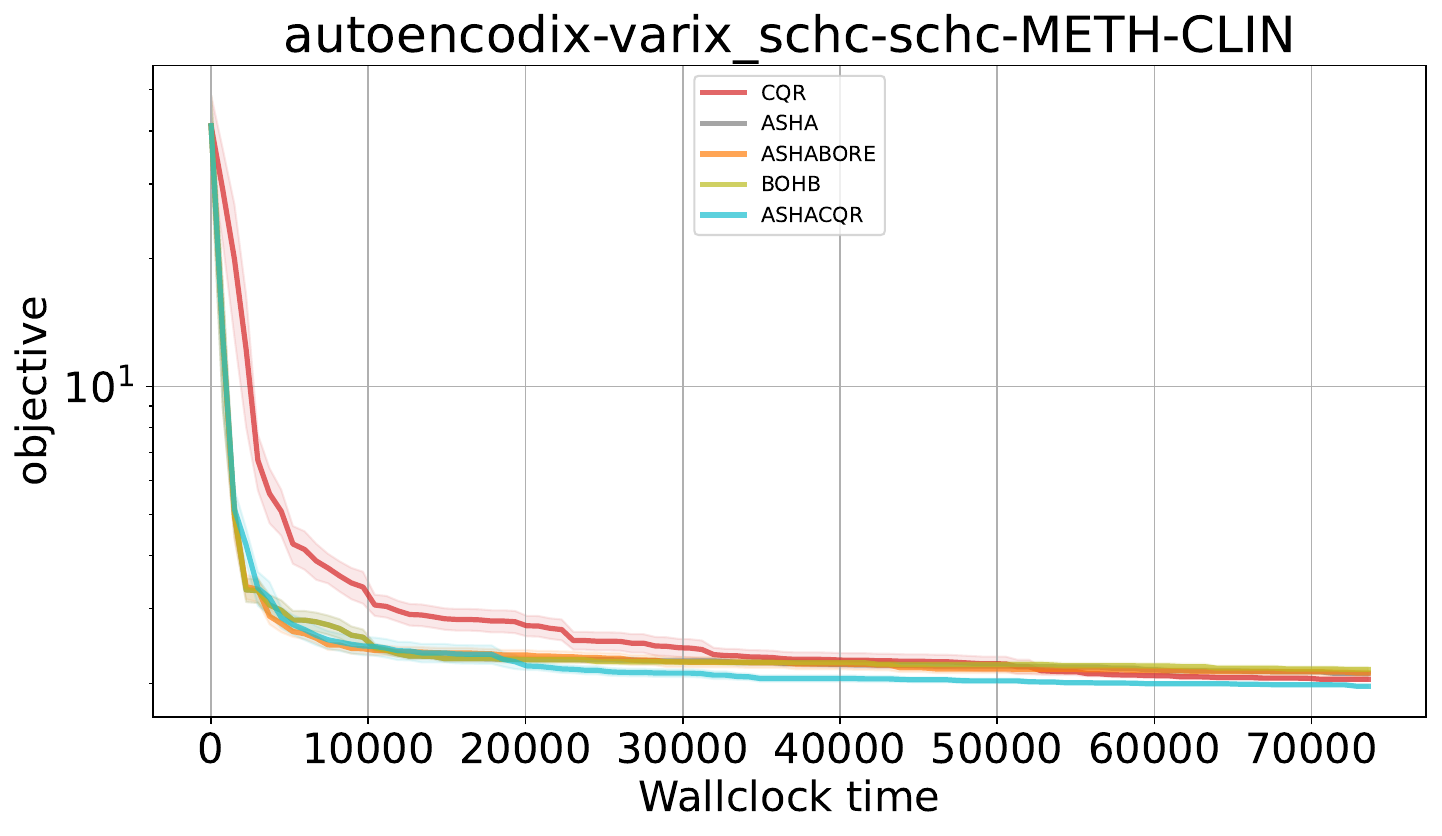} \\
    \includegraphics[width=0.32\textwidth]{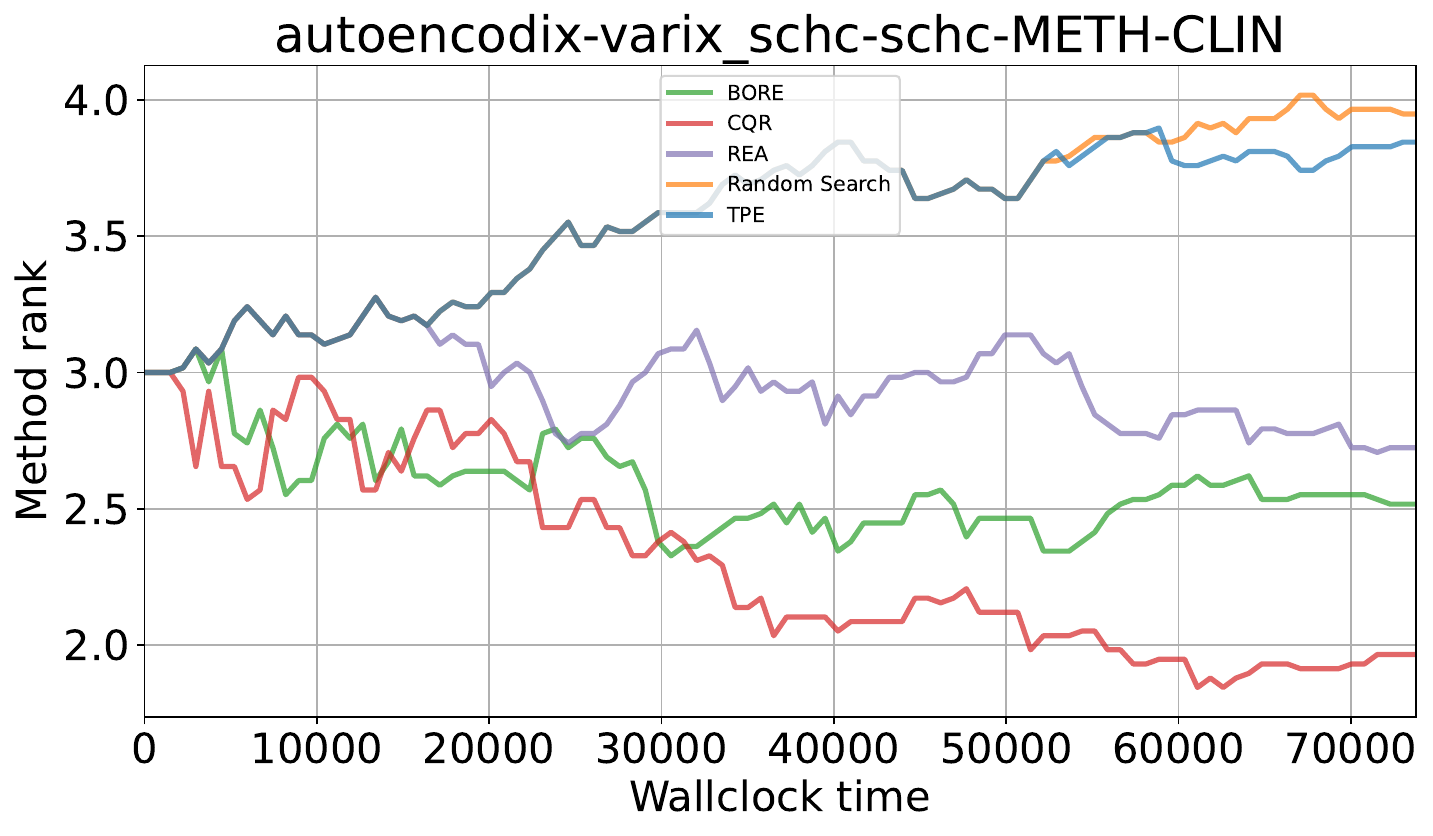} &
    \includegraphics[width=0.32\textwidth]{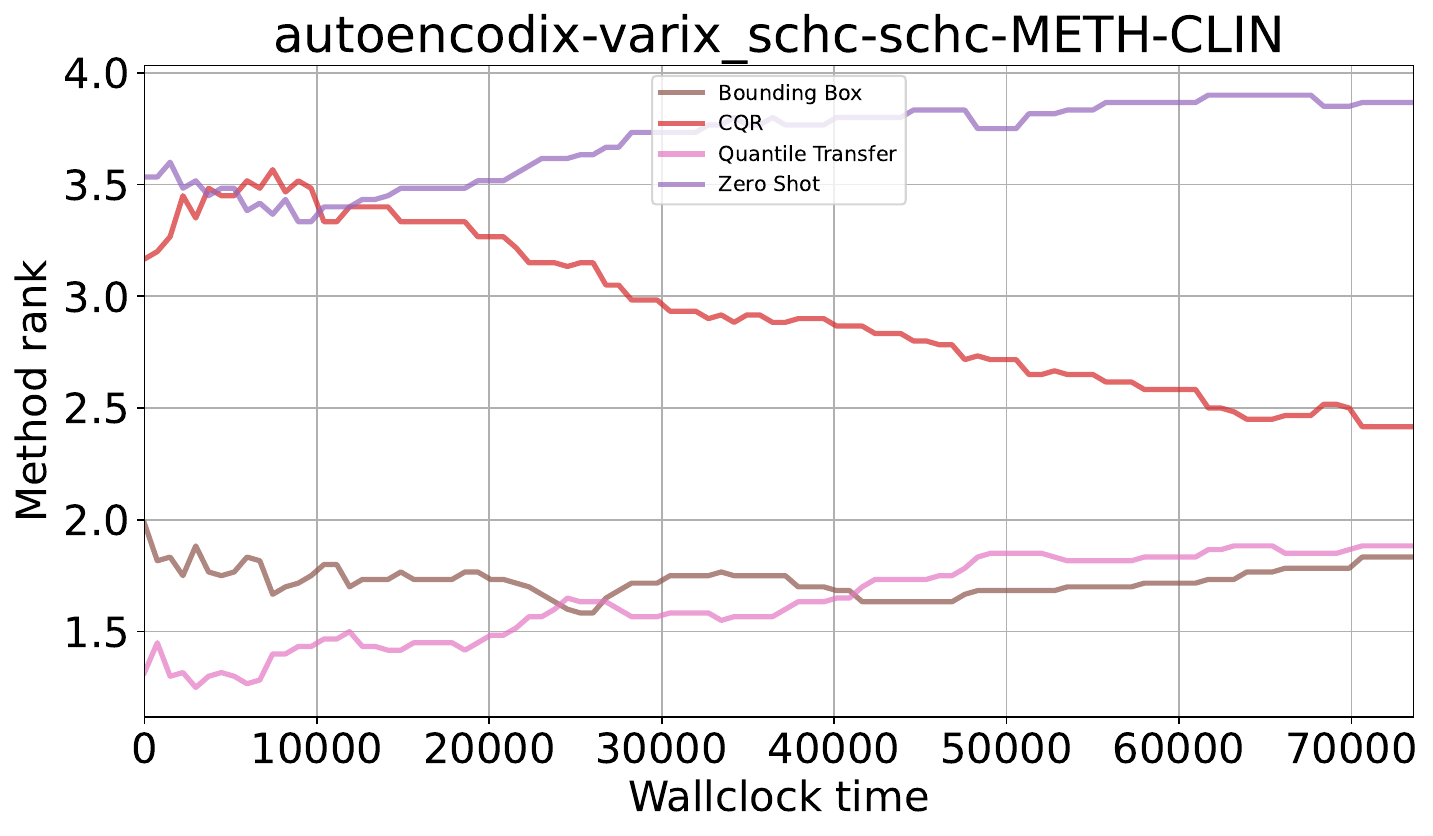} &
    \includegraphics[width=0.32\textwidth]{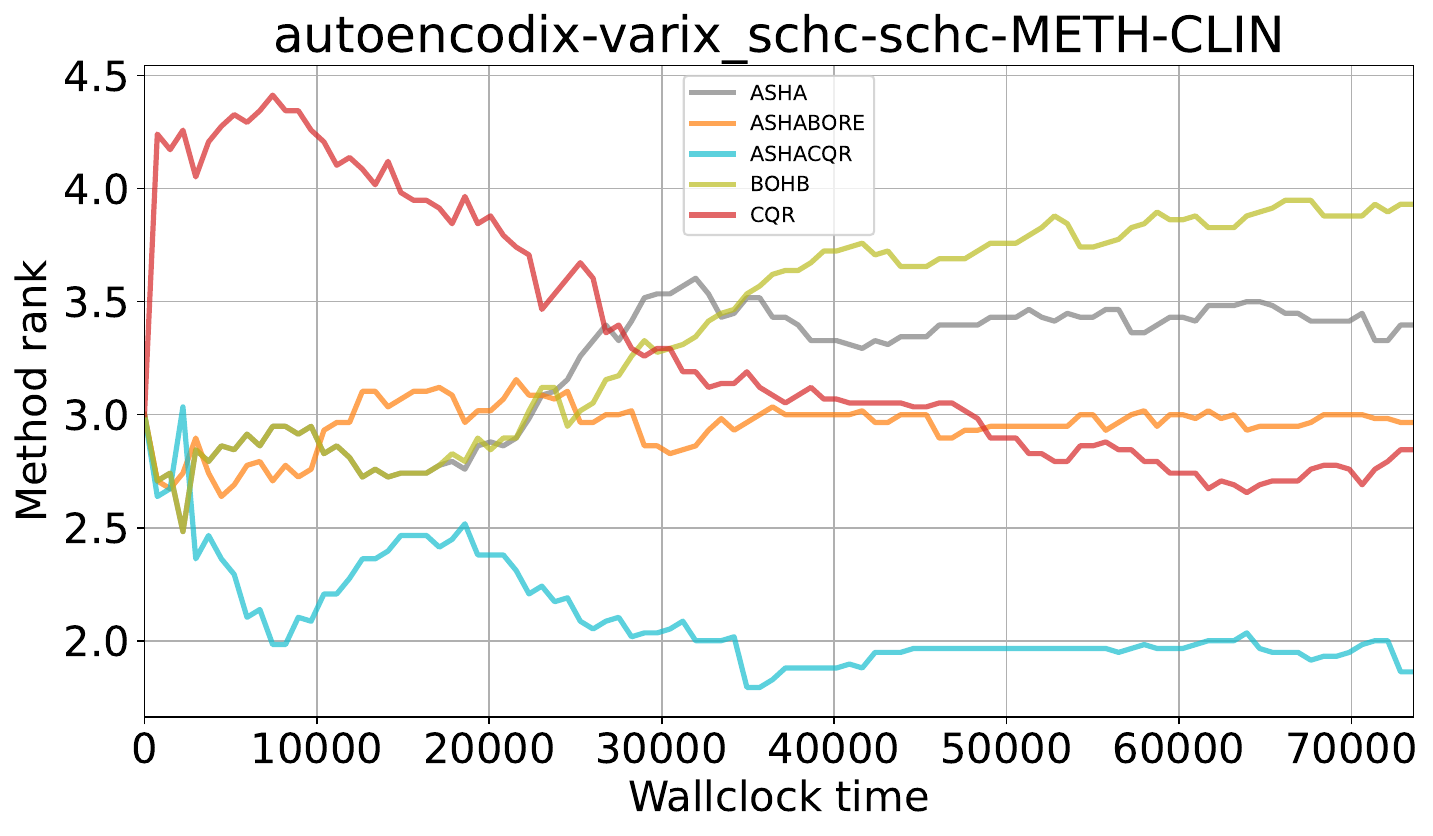} \\
    \midrule
    \multicolumn{3}{c}{\textbf{autoencodix-varix\_schc-schc-RNA-CLIN}} \\
    \includegraphics[width=0.32\textwidth]{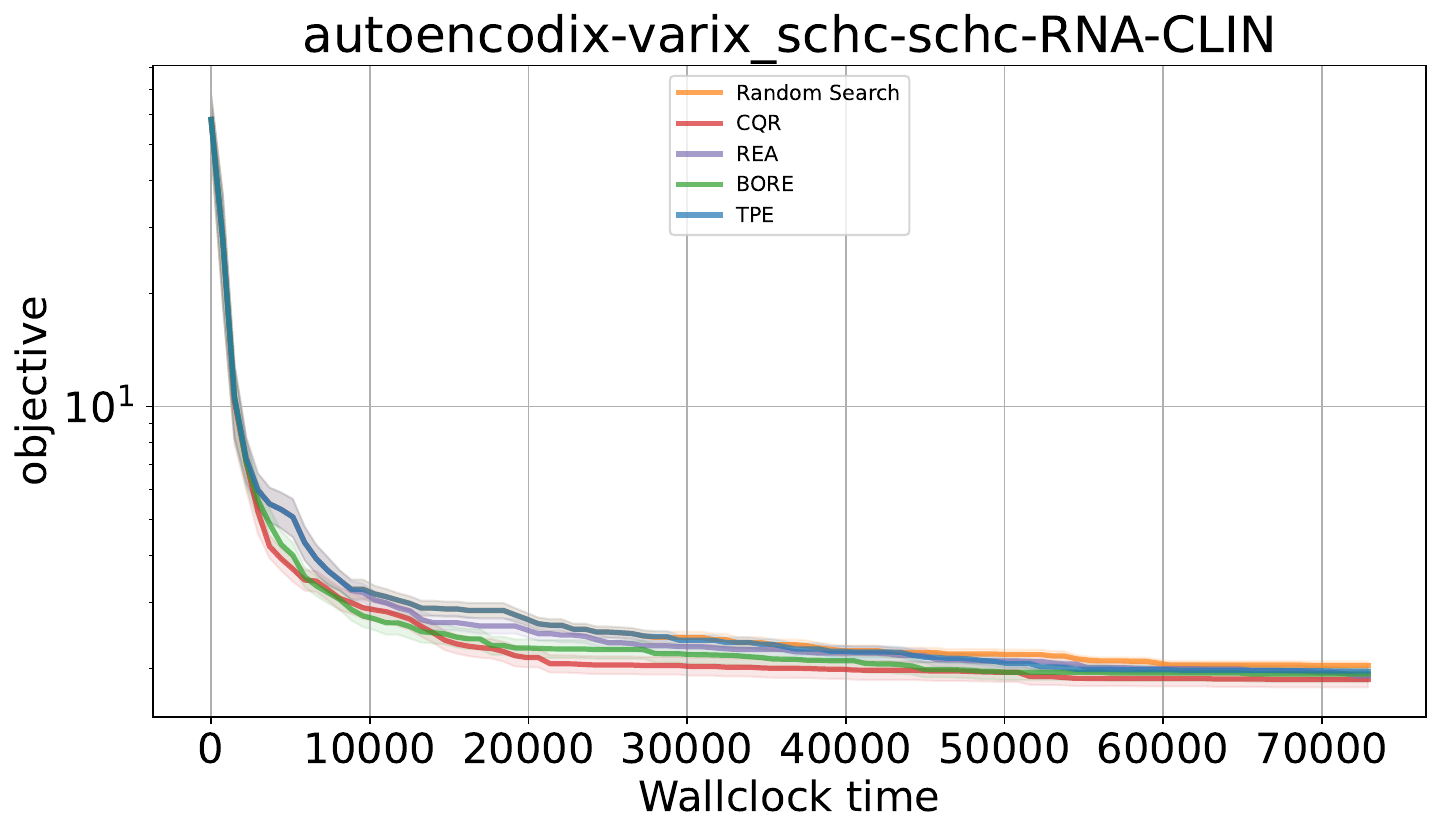} &
    \includegraphics[width=0.32\textwidth]{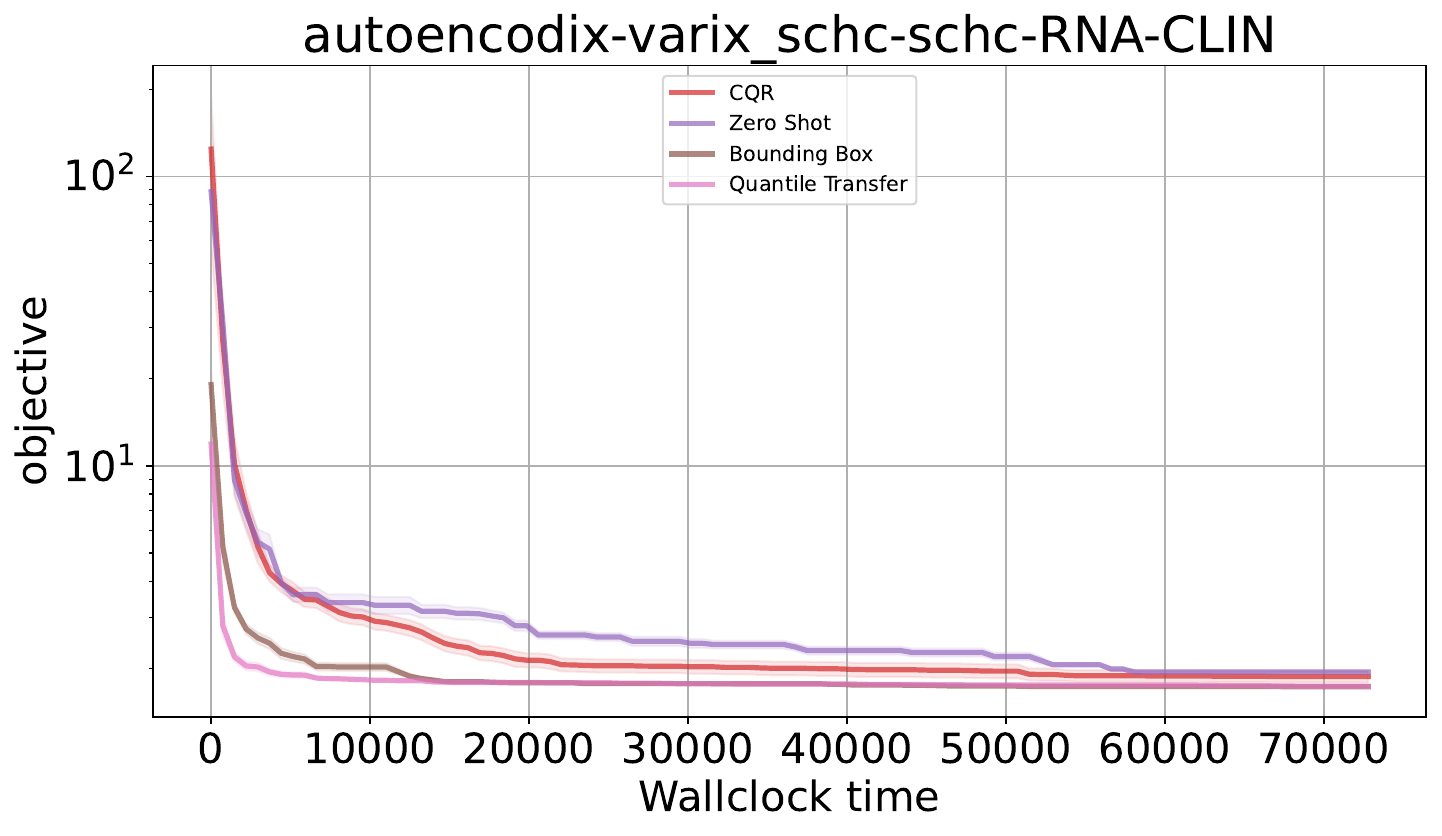} &
    \includegraphics[width=0.32\textwidth]{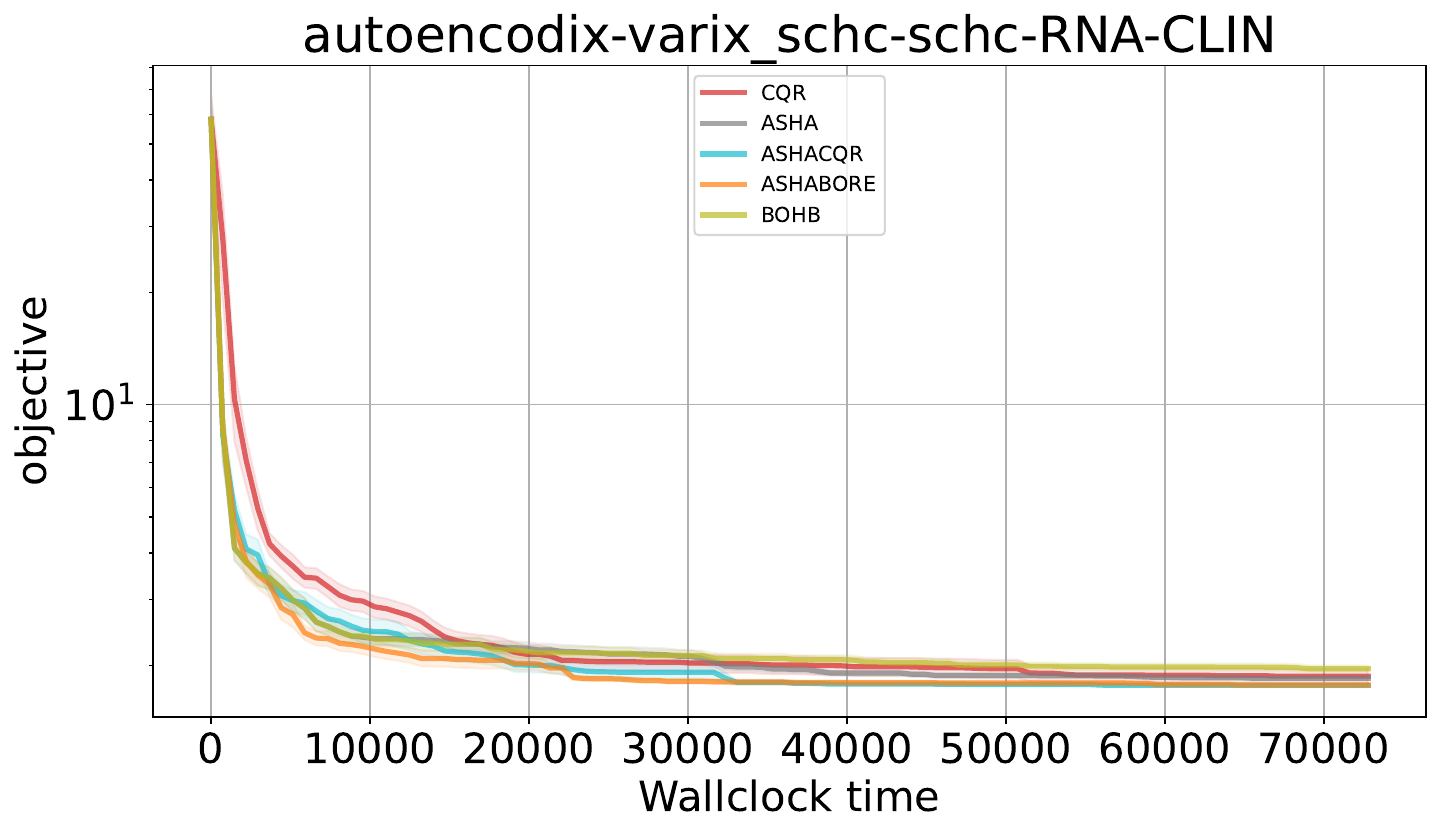} \\
    \includegraphics[width=0.32\textwidth]{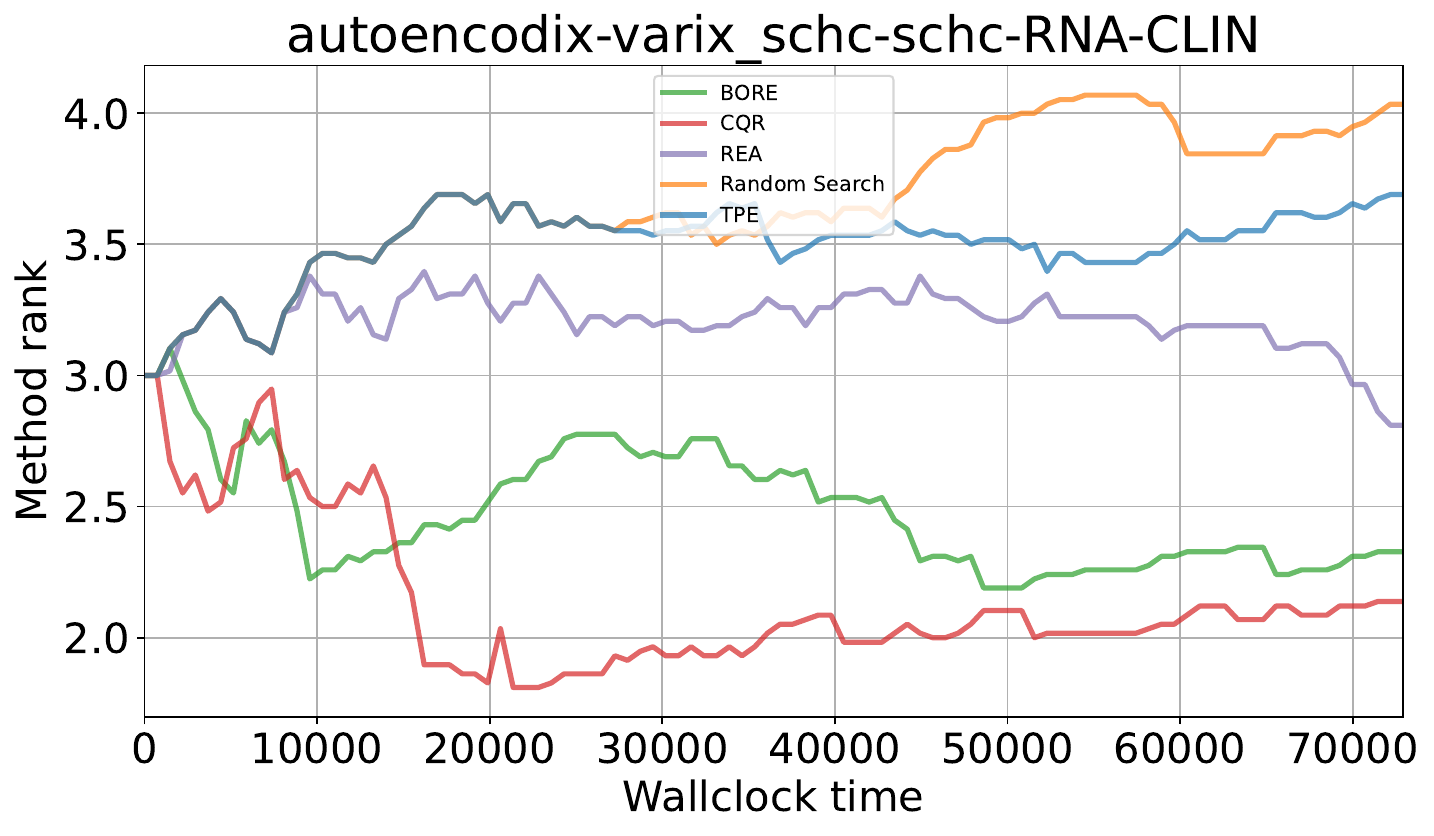} &
    \includegraphics[width=0.32\textwidth]{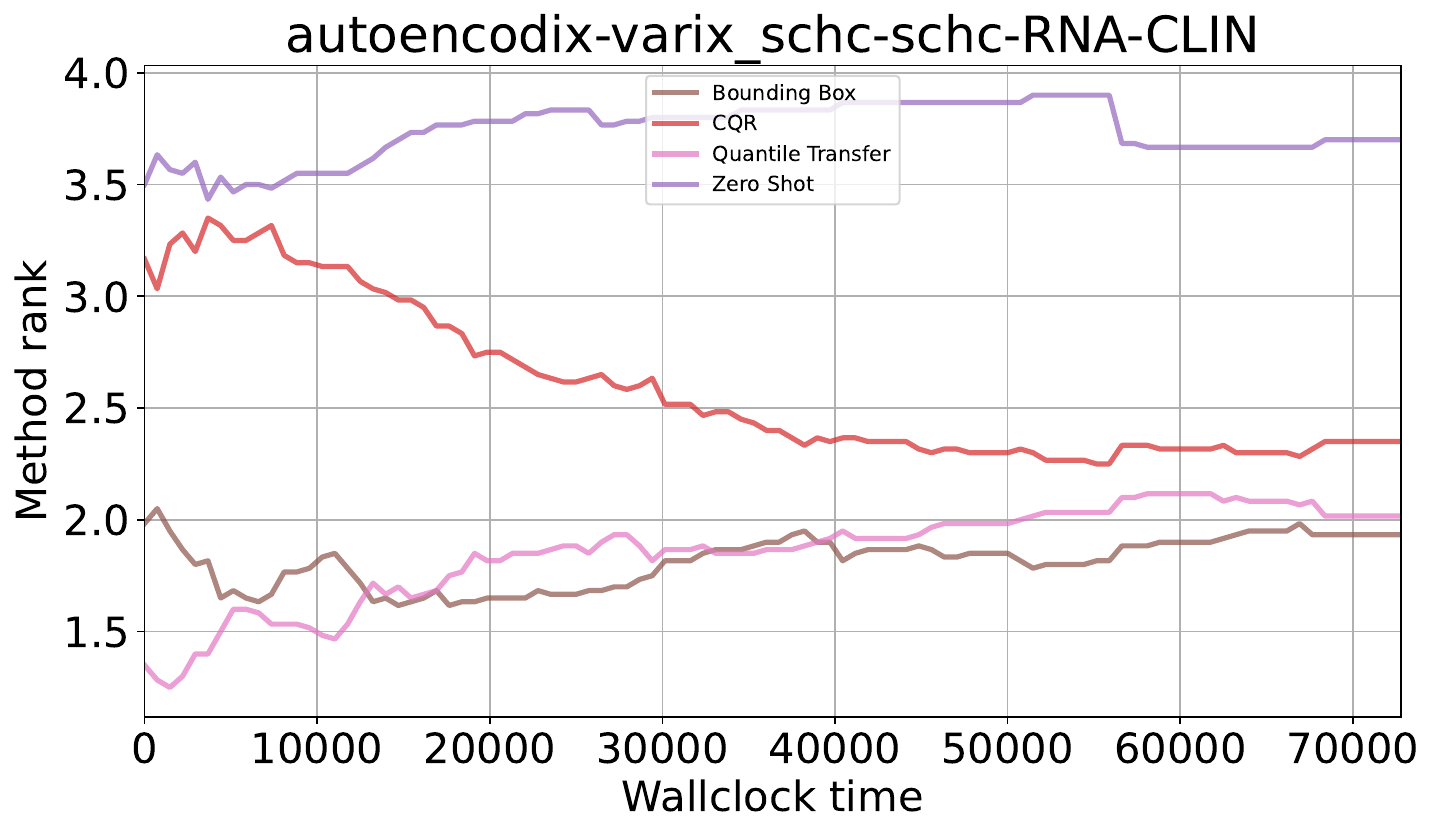} &
    \includegraphics[width=0.32\textwidth]{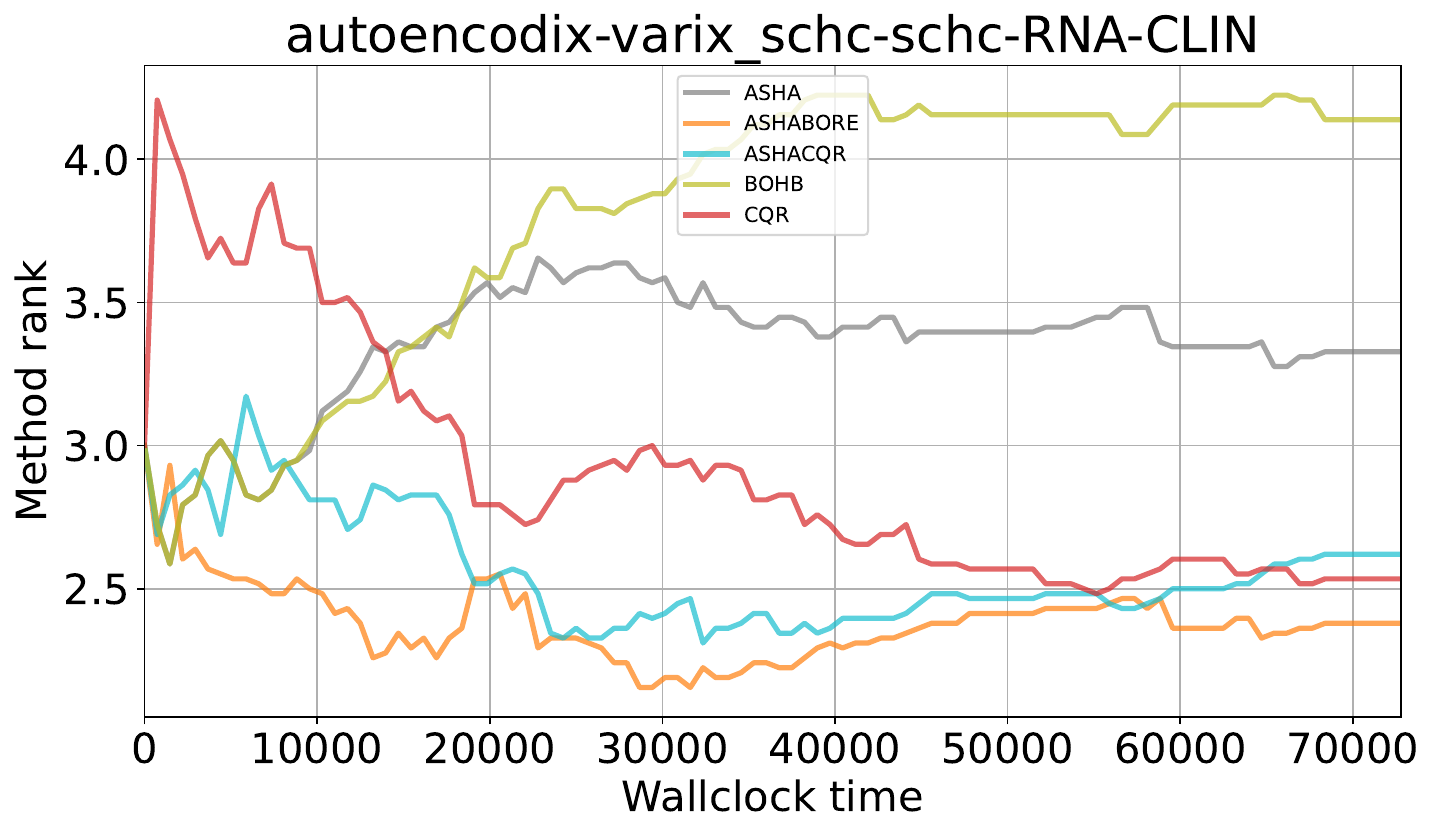} \\
    \midrule
    \multicolumn{3}{c}{\textbf{autoencodix-varix\_schc-schc-RNA-METH-CLIN}} \\
    \includegraphics[width=0.32\textwidth]{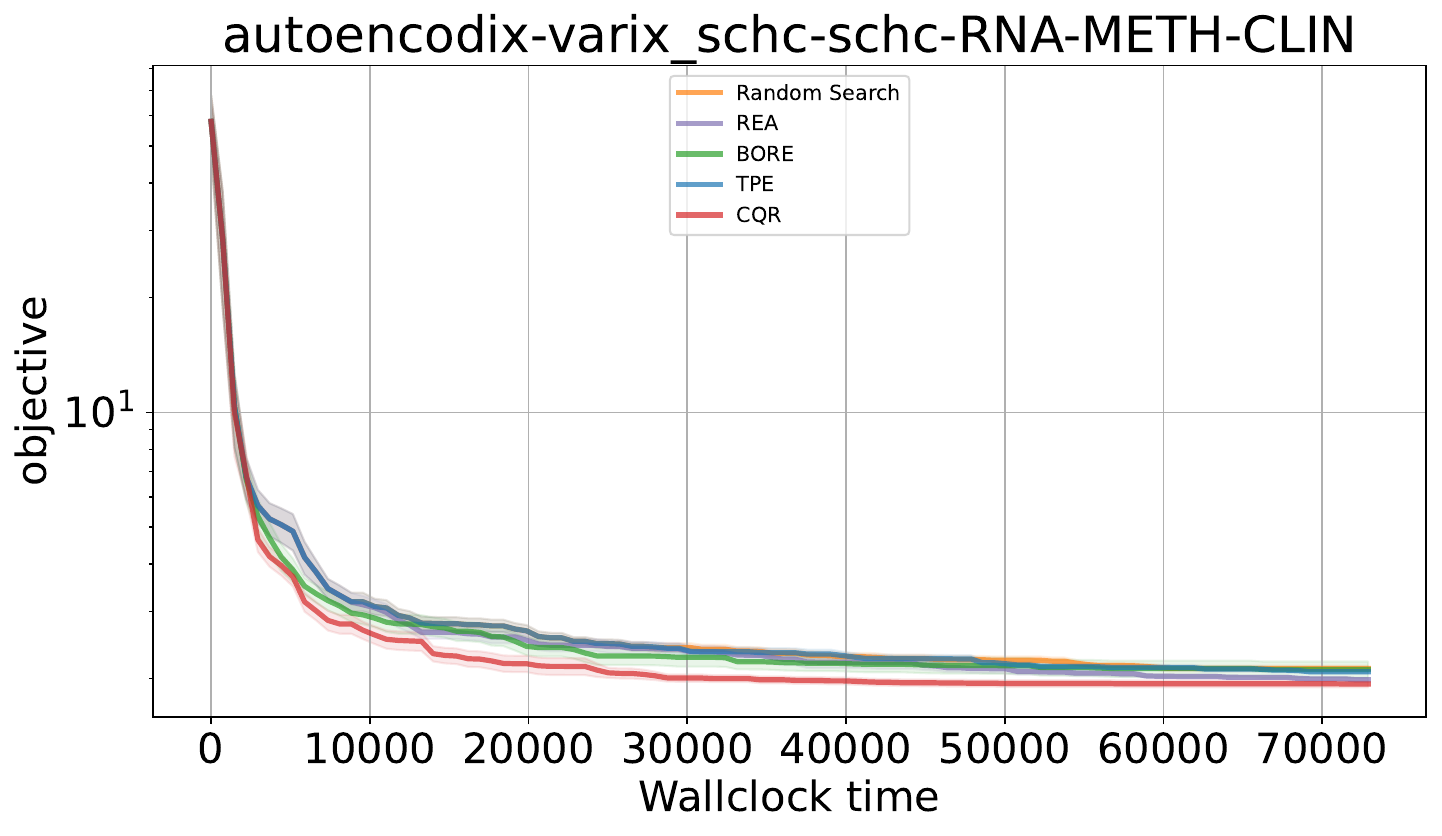} &
    \includegraphics[width=0.32\textwidth]{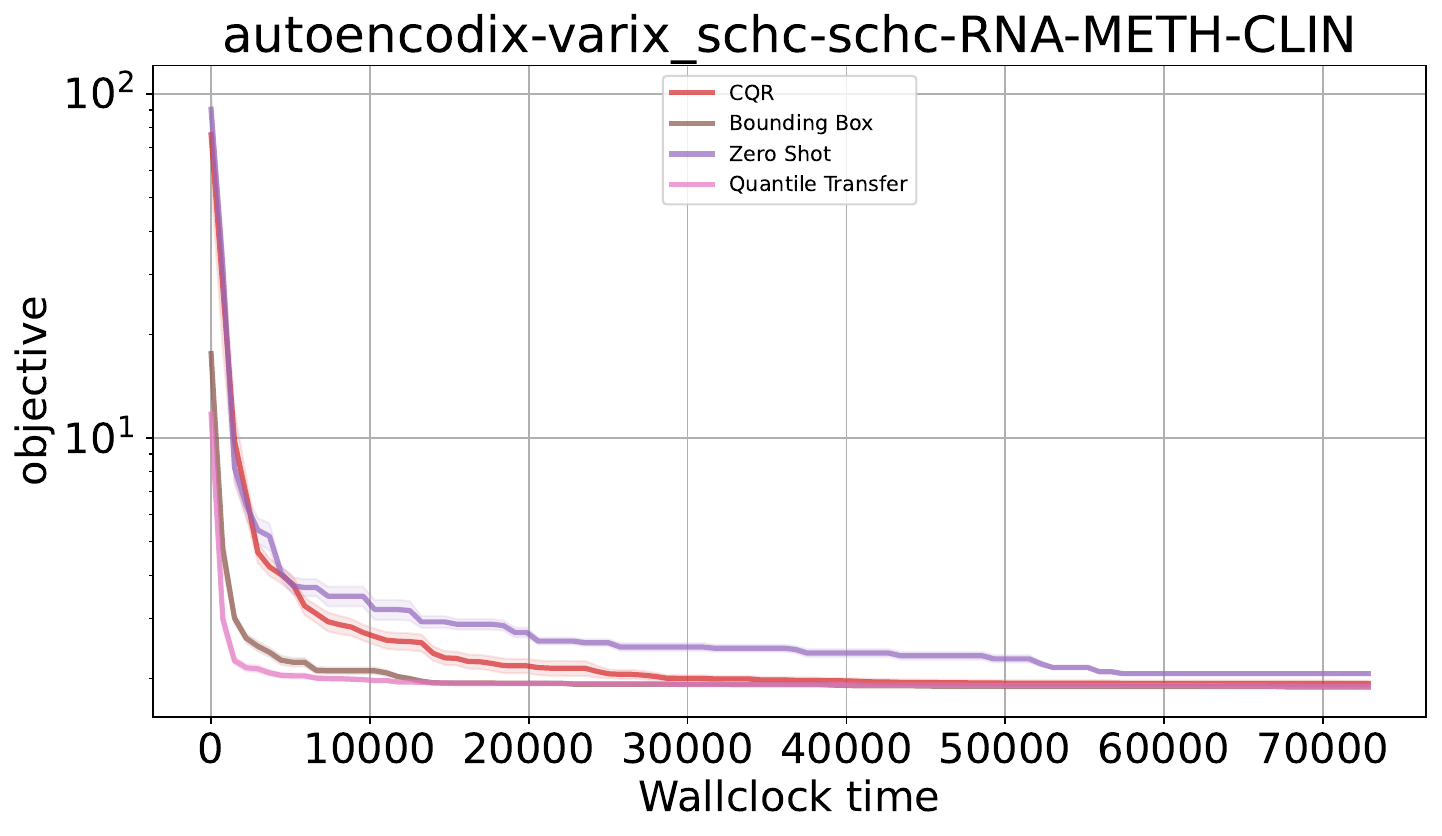} &
    \includegraphics[width=0.32\textwidth]{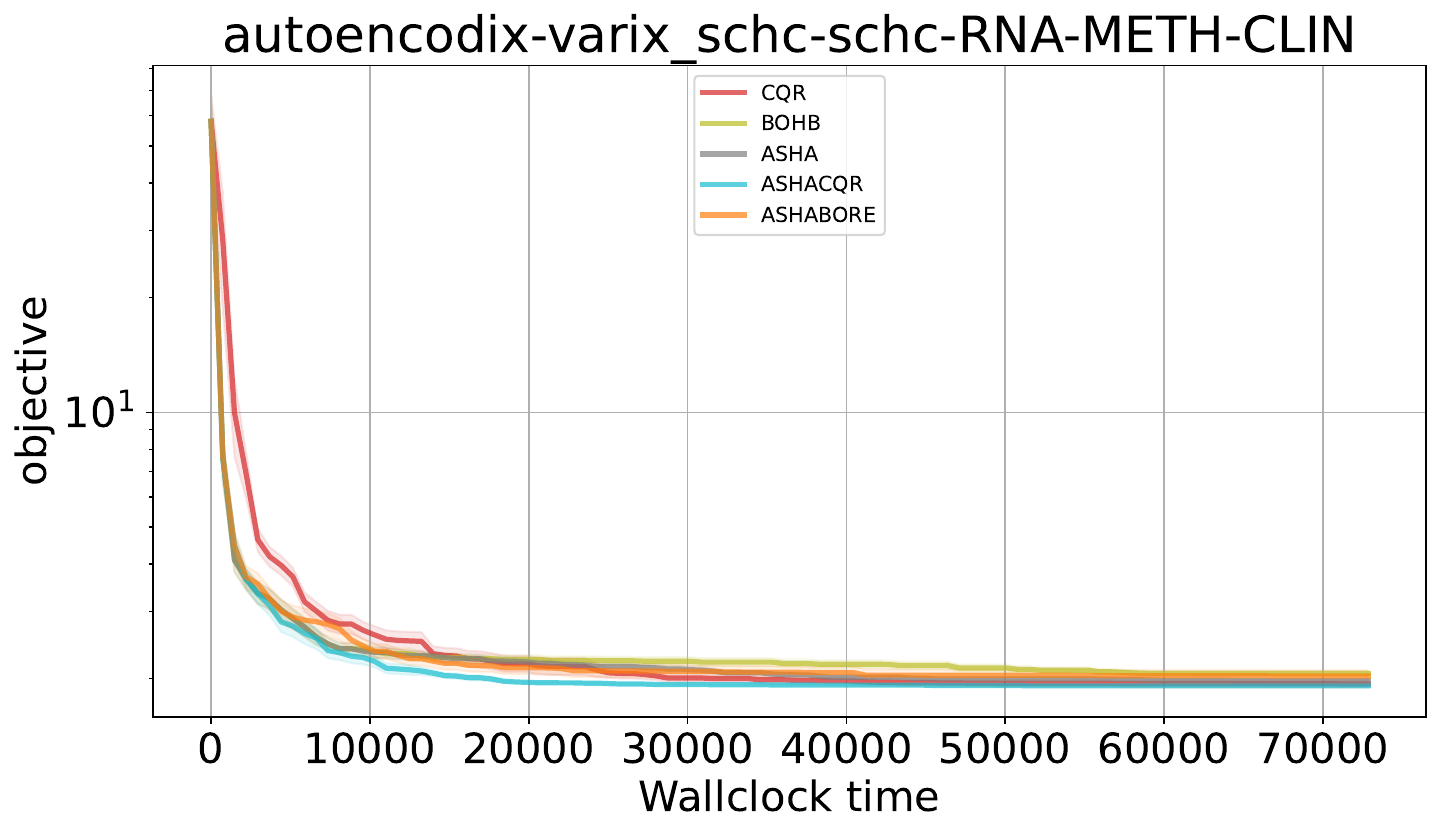} \\
    \includegraphics[width=0.32\textwidth]{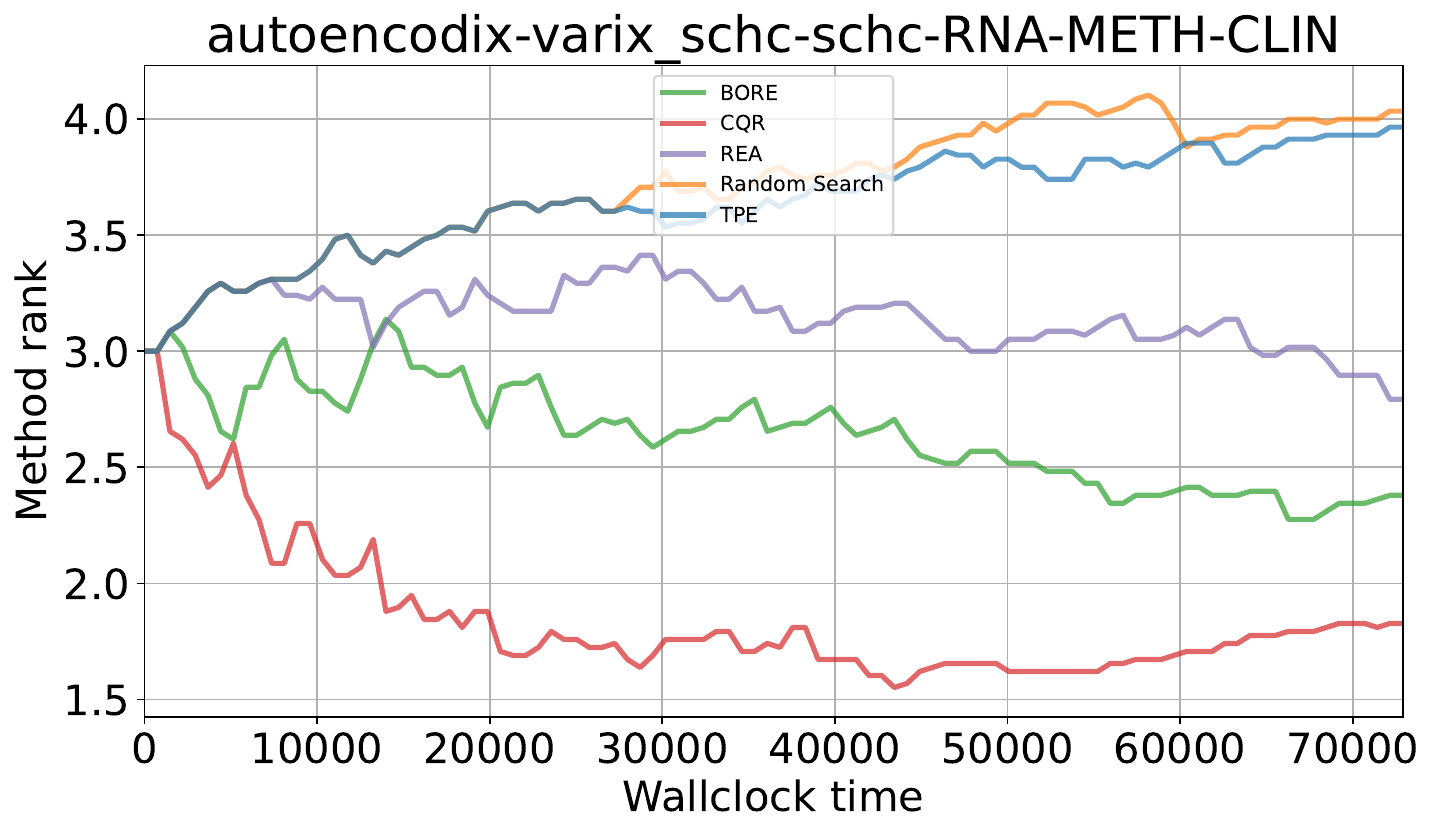} &
    \includegraphics[width=0.32\textwidth]{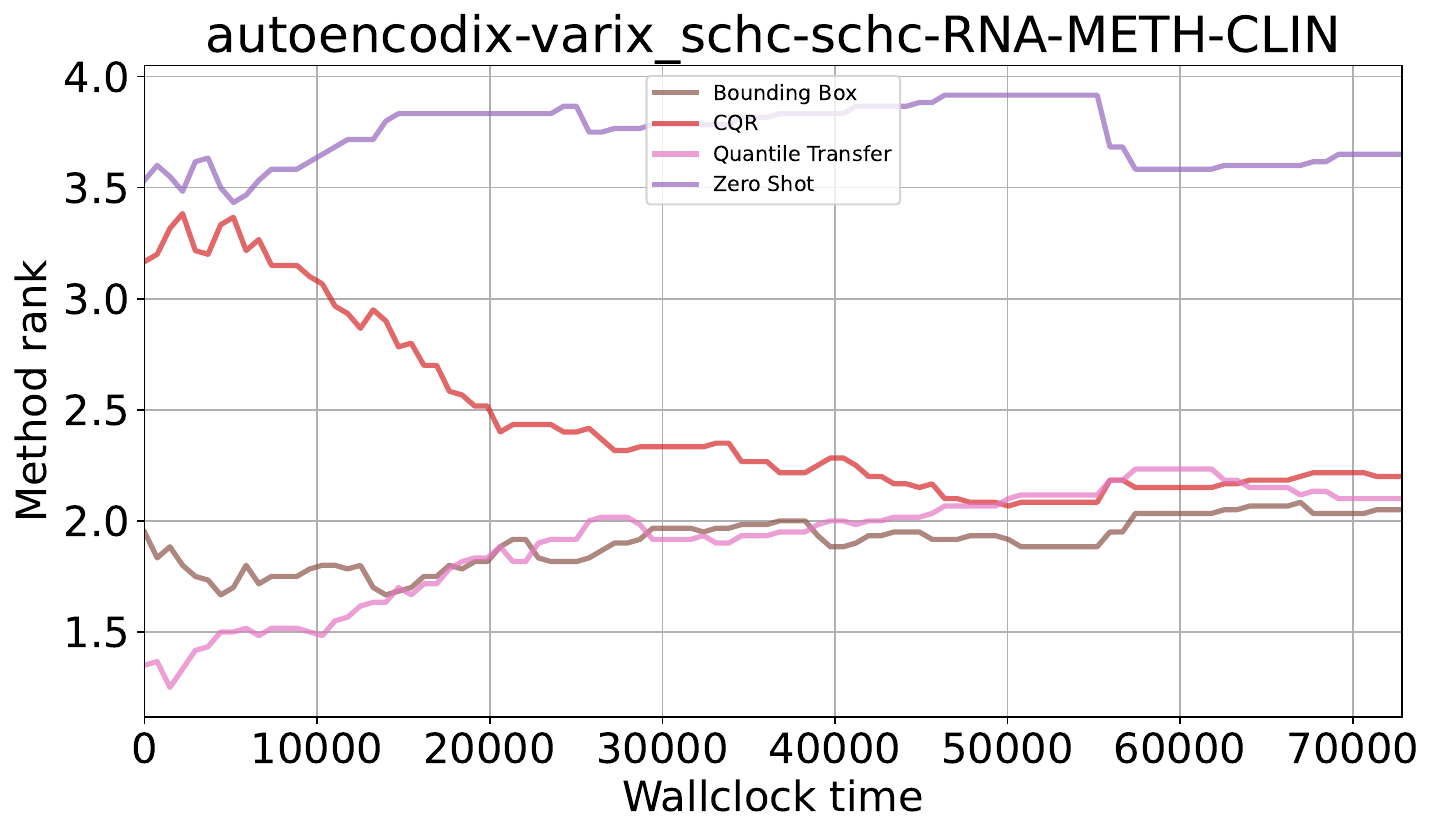} &
    \includegraphics[width=0.32\textwidth]{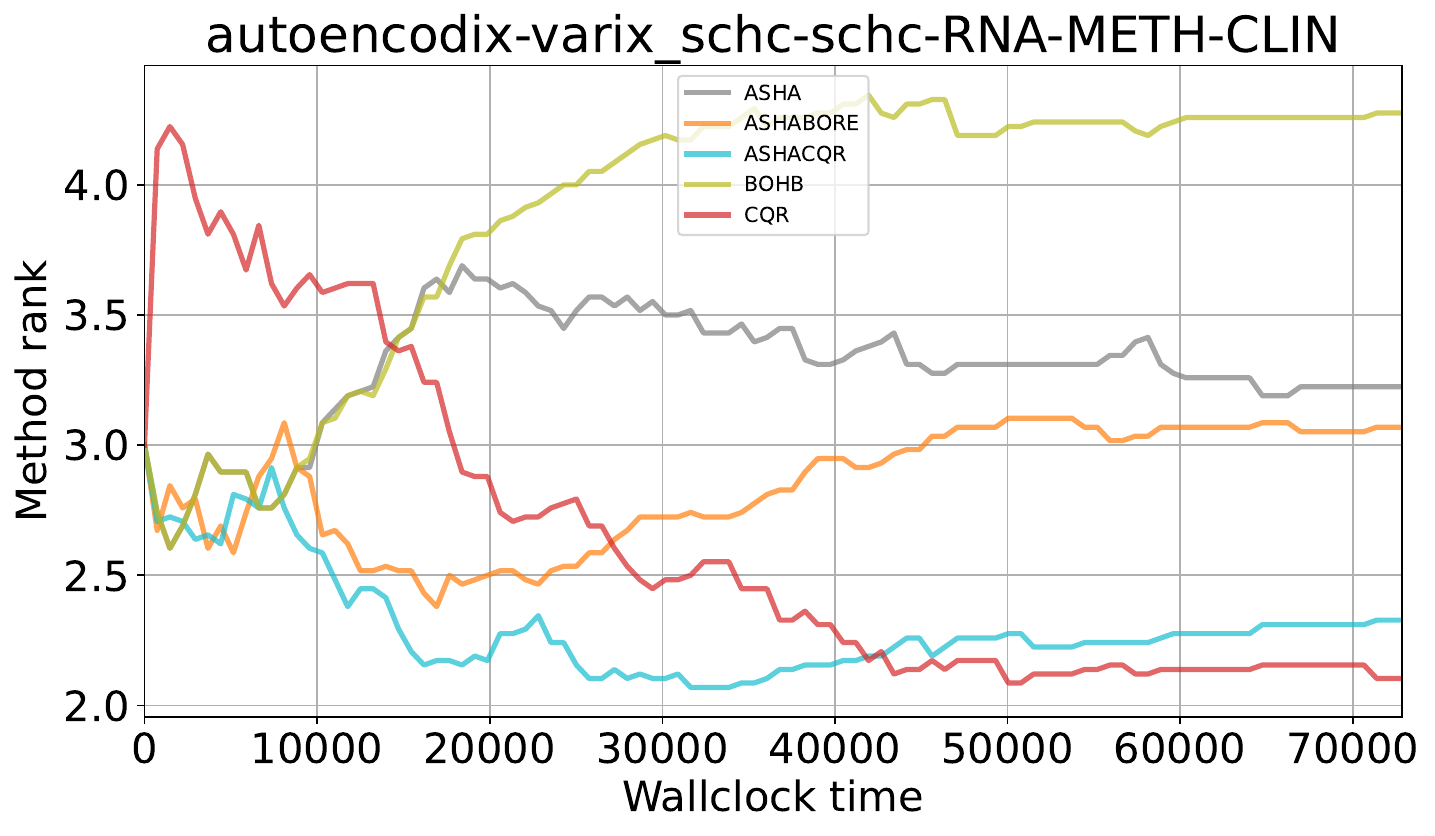} \\
    \end{tabular}
    \caption{Results for Varix tasks (Part 1).}
    \label{fig:varix_part1}
\end{figure}

\clearpage

\begin{figure}[htbp]
    \centering
    \setlength{\tabcolsep}{1pt}
    \begin{tabular}{ccc}
    \multicolumn{3}{c}{\textbf{autoencodix-varix\_tcga-tcga-DNA-CLIN}} \\
    \textbf{Single-Fidelity} & \textbf{Transfer Learning} & \textbf{Multi-Fidelity} \\
    \includegraphics[width=0.32\textwidth]{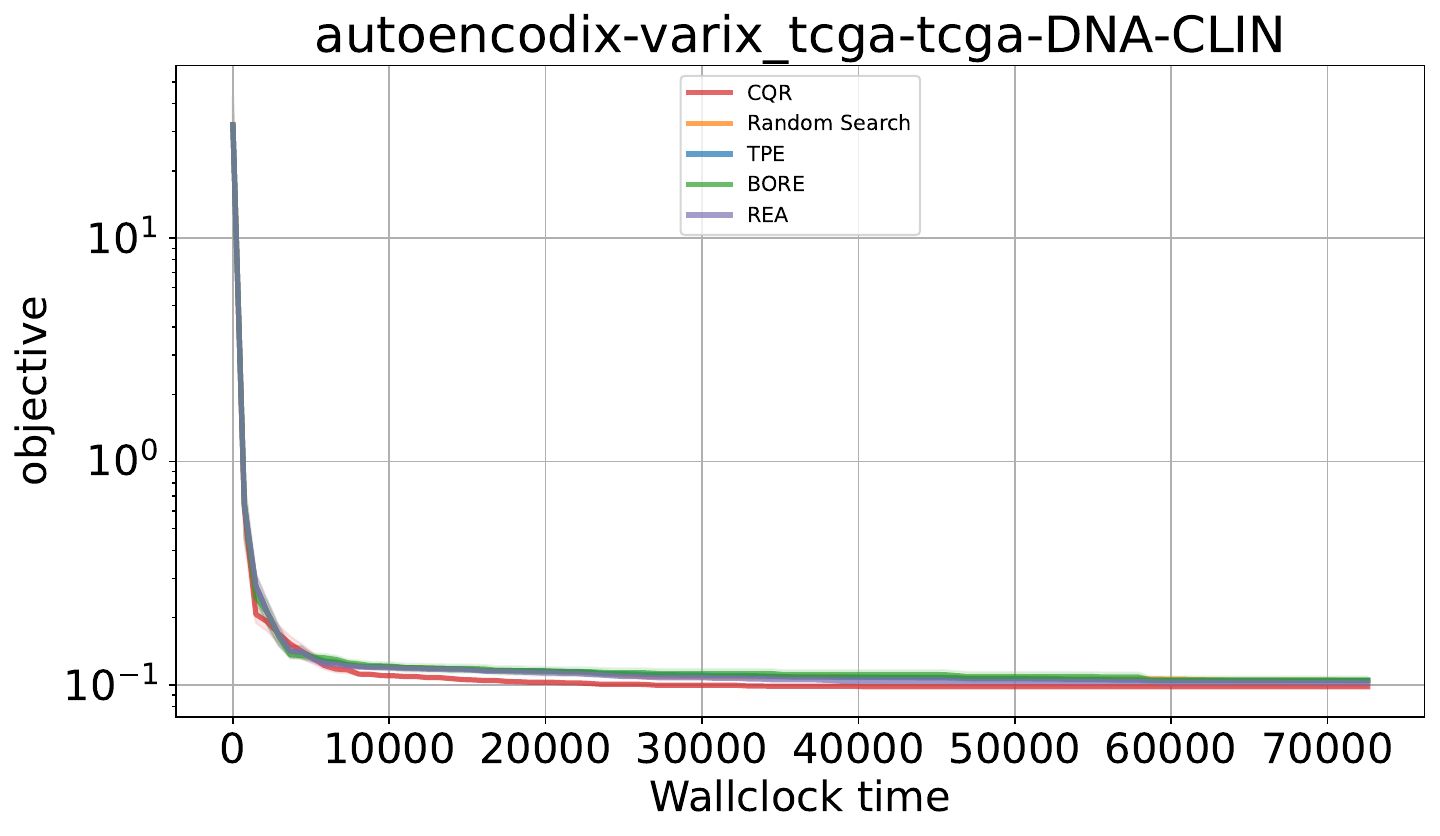} &
    \includegraphics[width=0.32\textwidth]{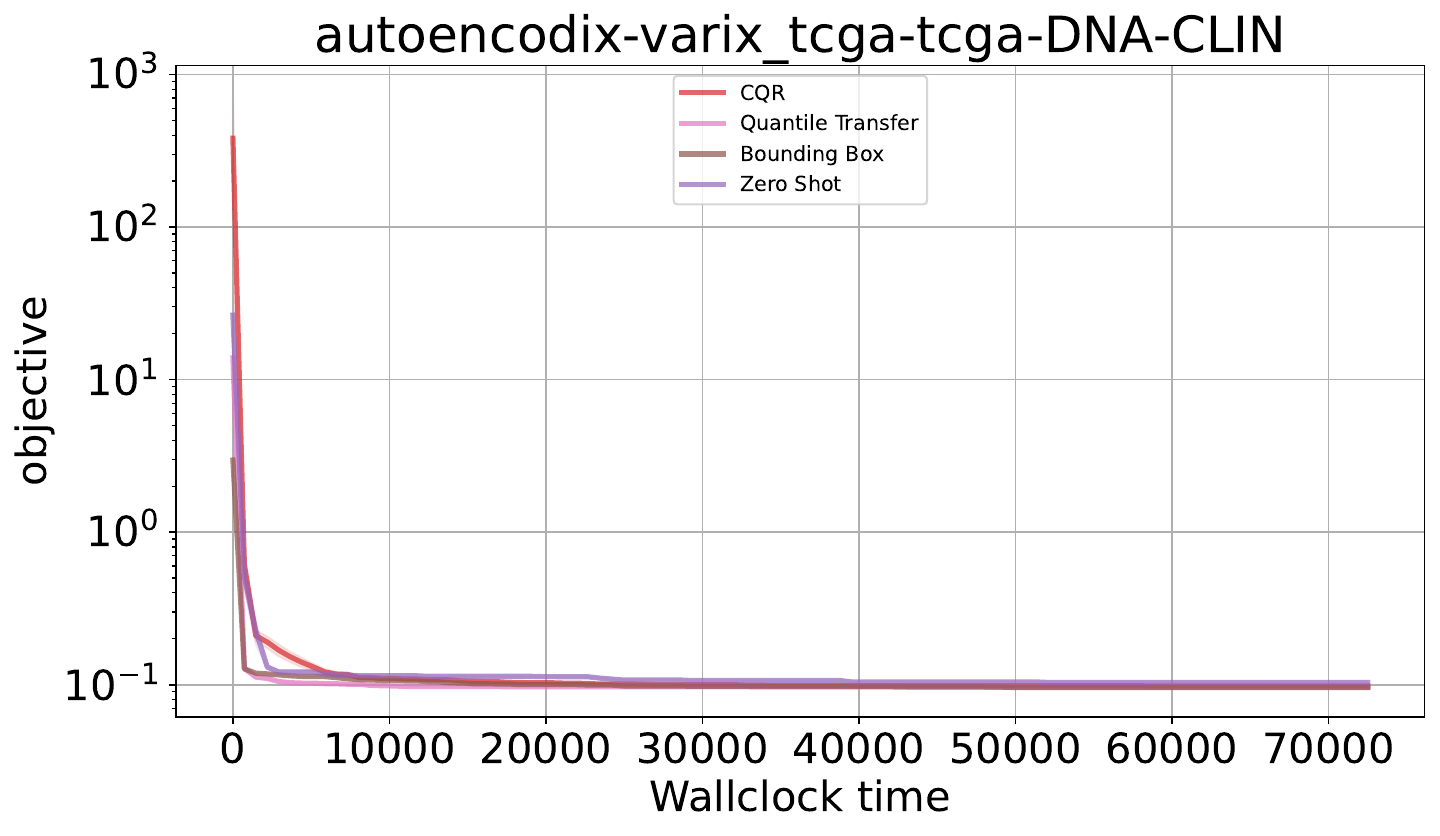} &
    \includegraphics[width=0.32\textwidth]{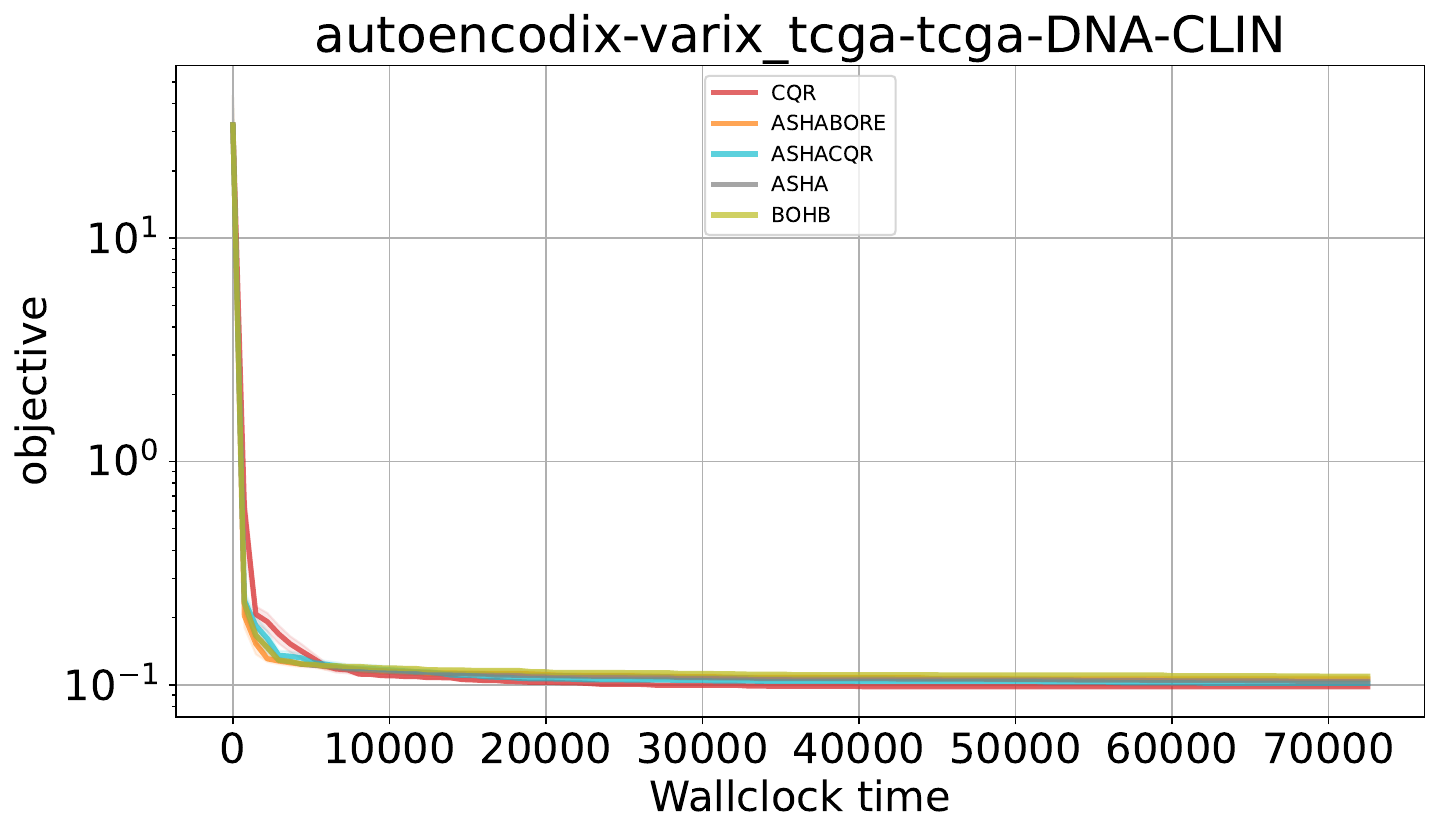} \\
    \includegraphics[width=0.32\textwidth]{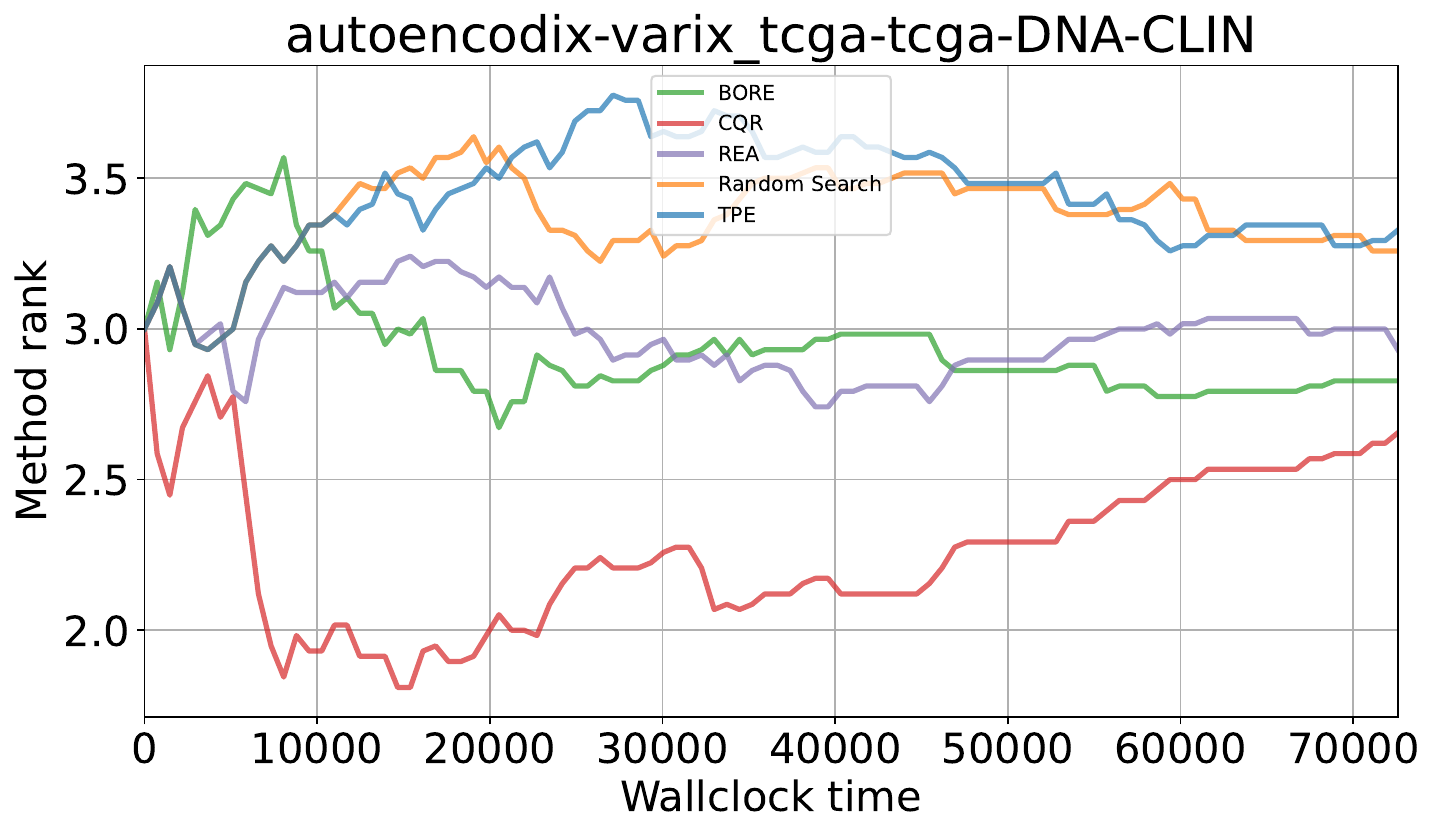} &
    \includegraphics[width=0.32\textwidth]{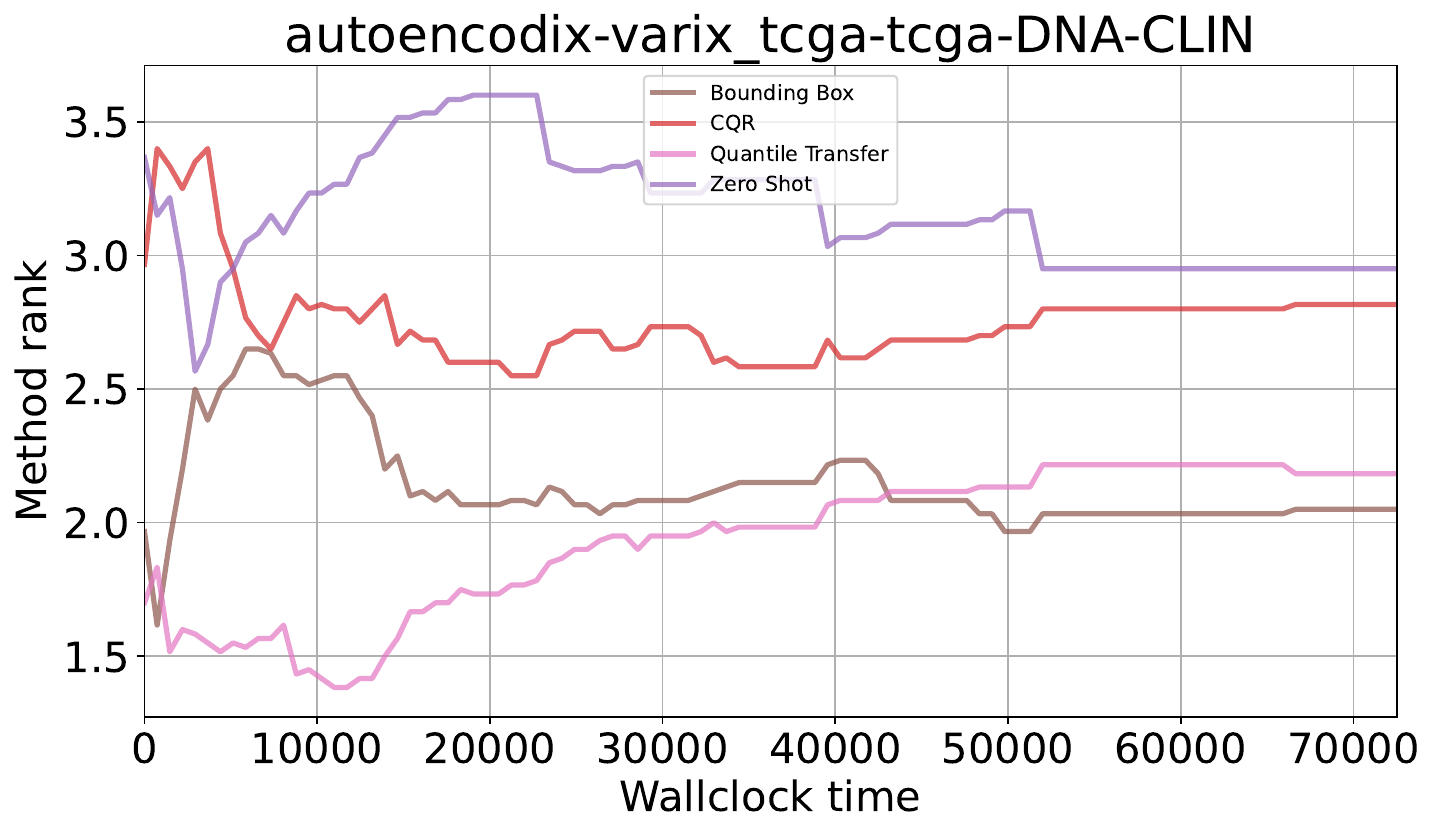} &
    \includegraphics[width=0.32\textwidth]{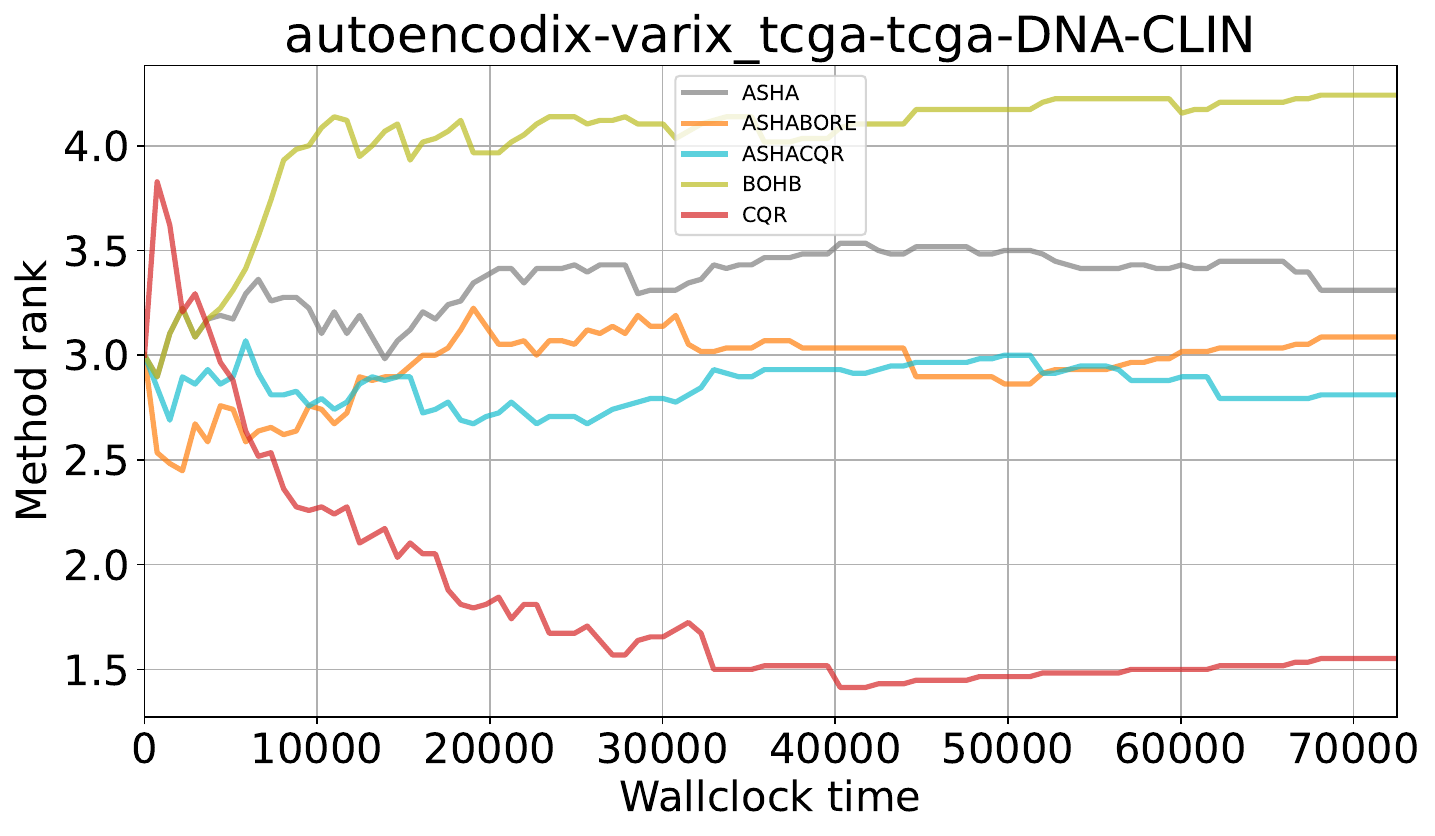} \\
    \midrule
    \multicolumn{3}{c}{\textbf{autoencodix-varix\_tcga-tcga-METH-CLIN}} \\
    \includegraphics[width=0.32\textwidth]{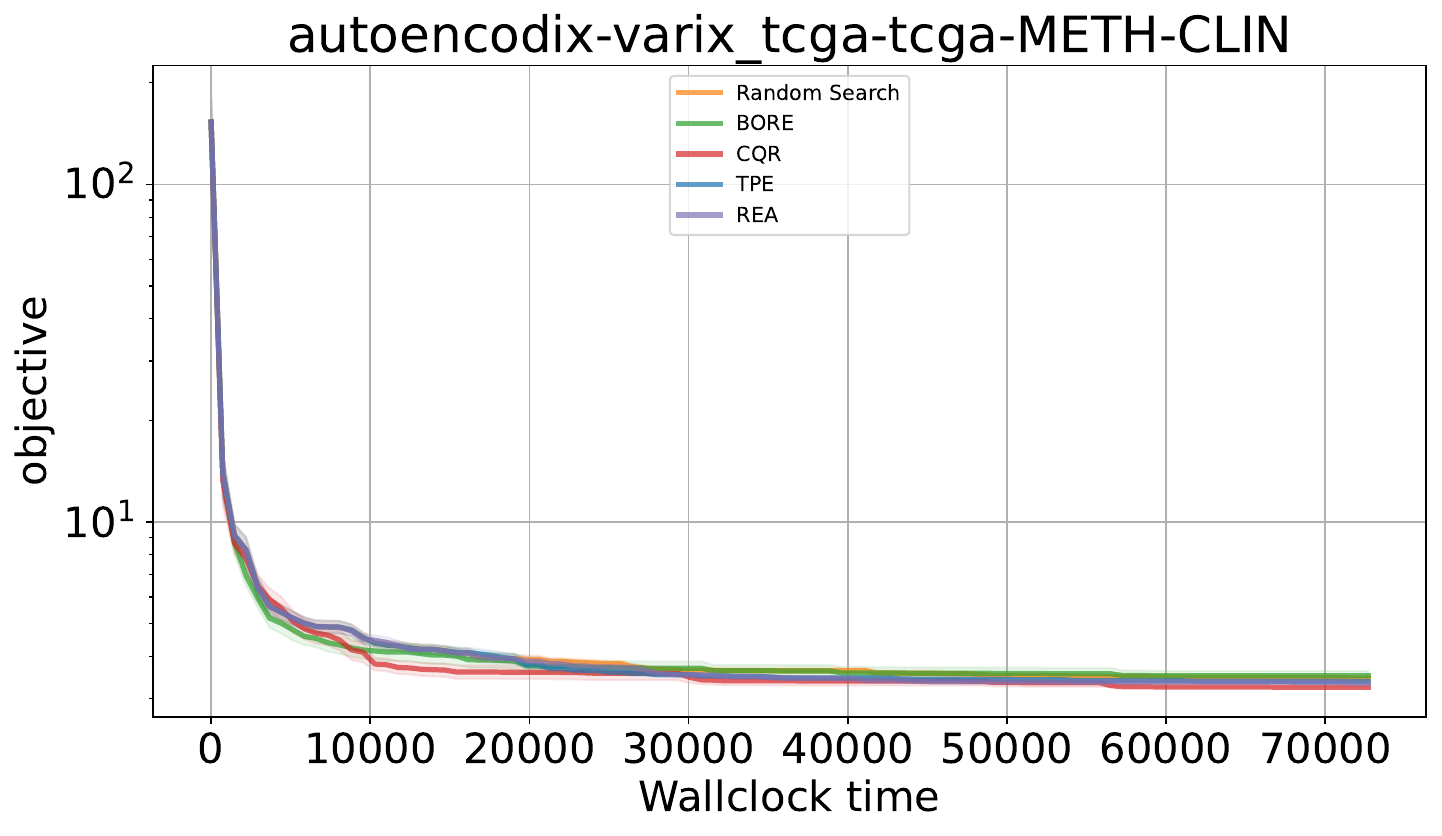} &
    \includegraphics[width=0.32\textwidth]{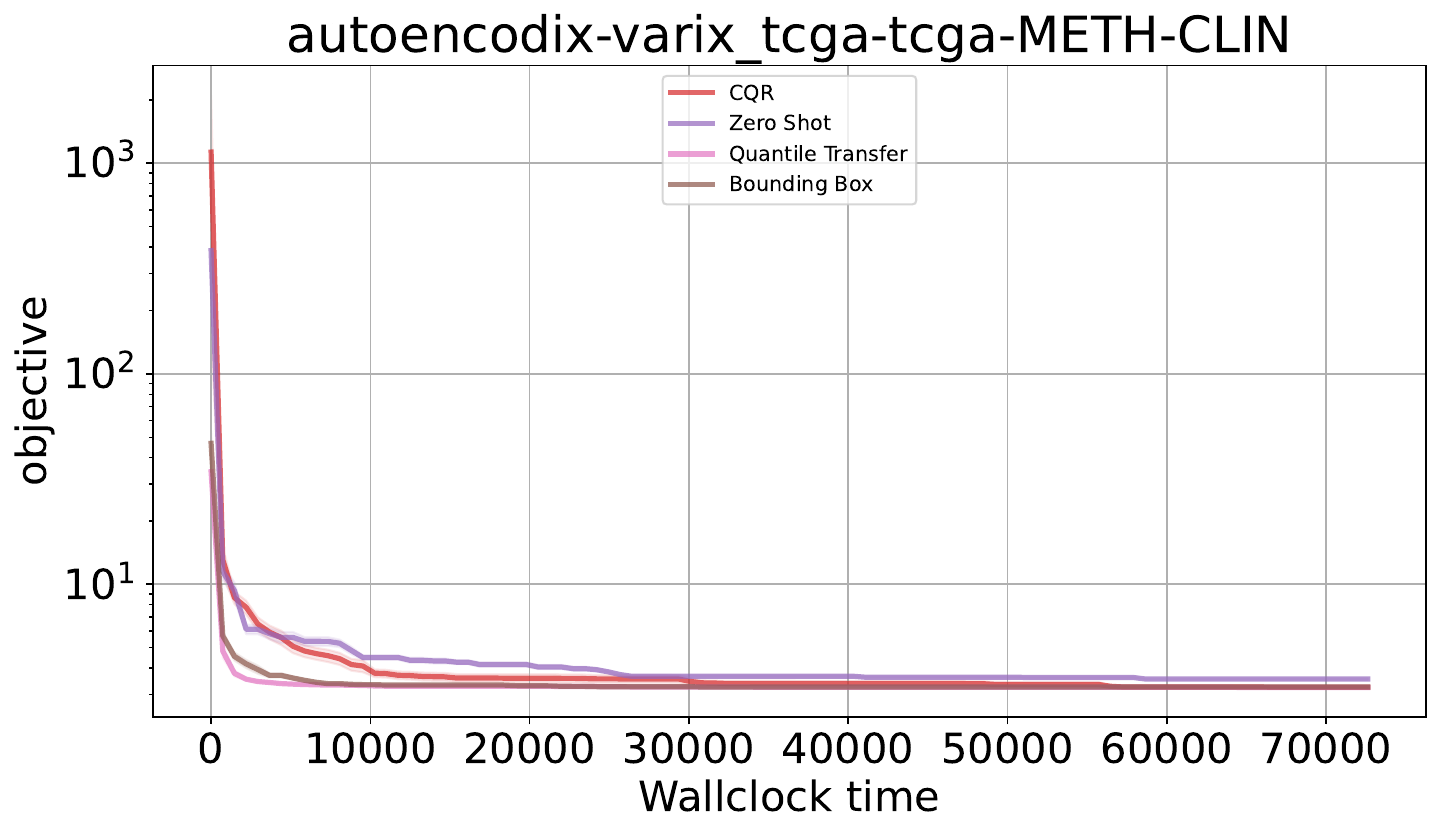} &
    \includegraphics[width=0.32\textwidth]{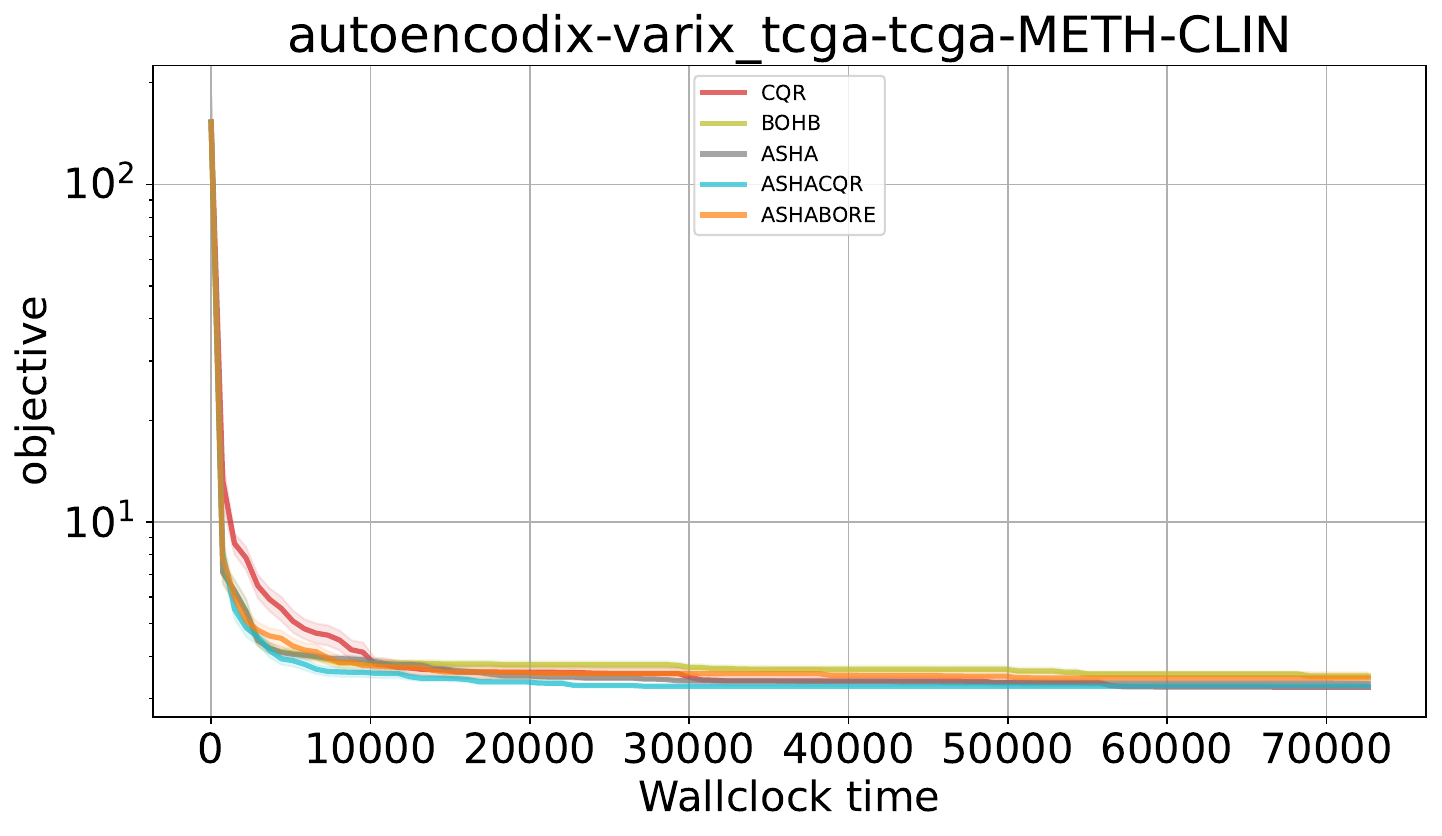} \\
    \includegraphics[width=0.32\textwidth]{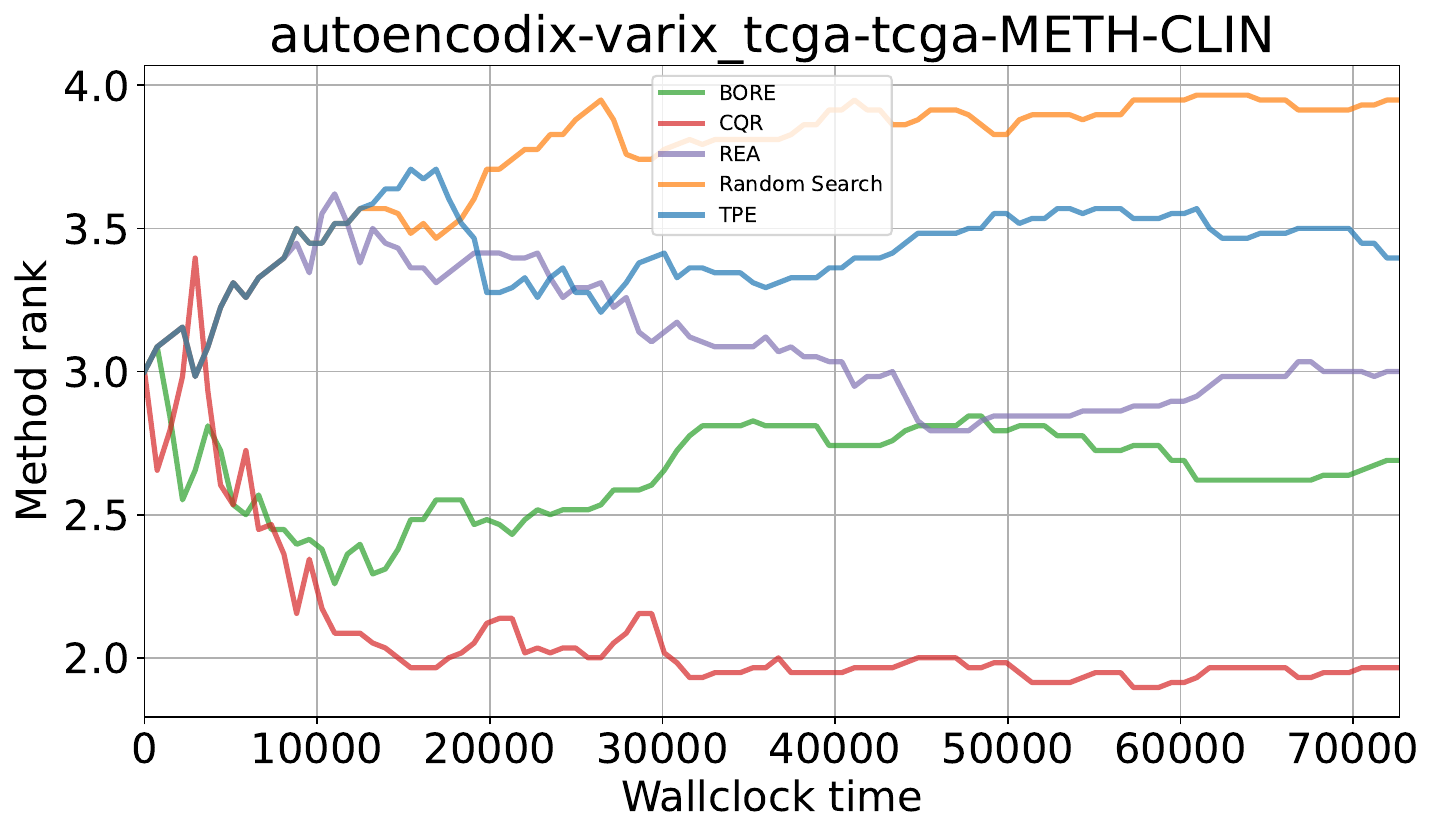} &
    \includegraphics[width=0.32\textwidth]{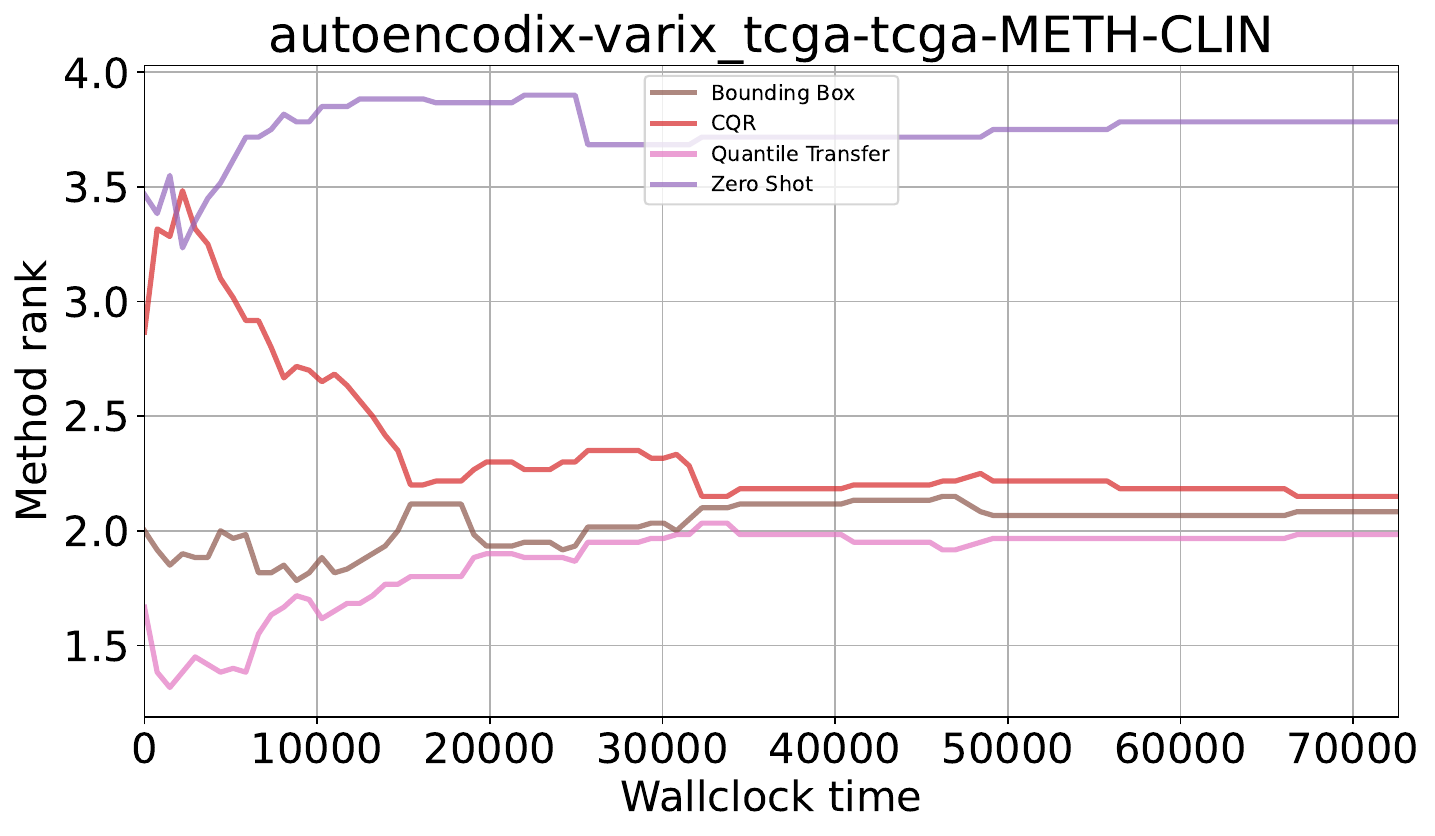} &
    \includegraphics[width=0.32\textwidth]{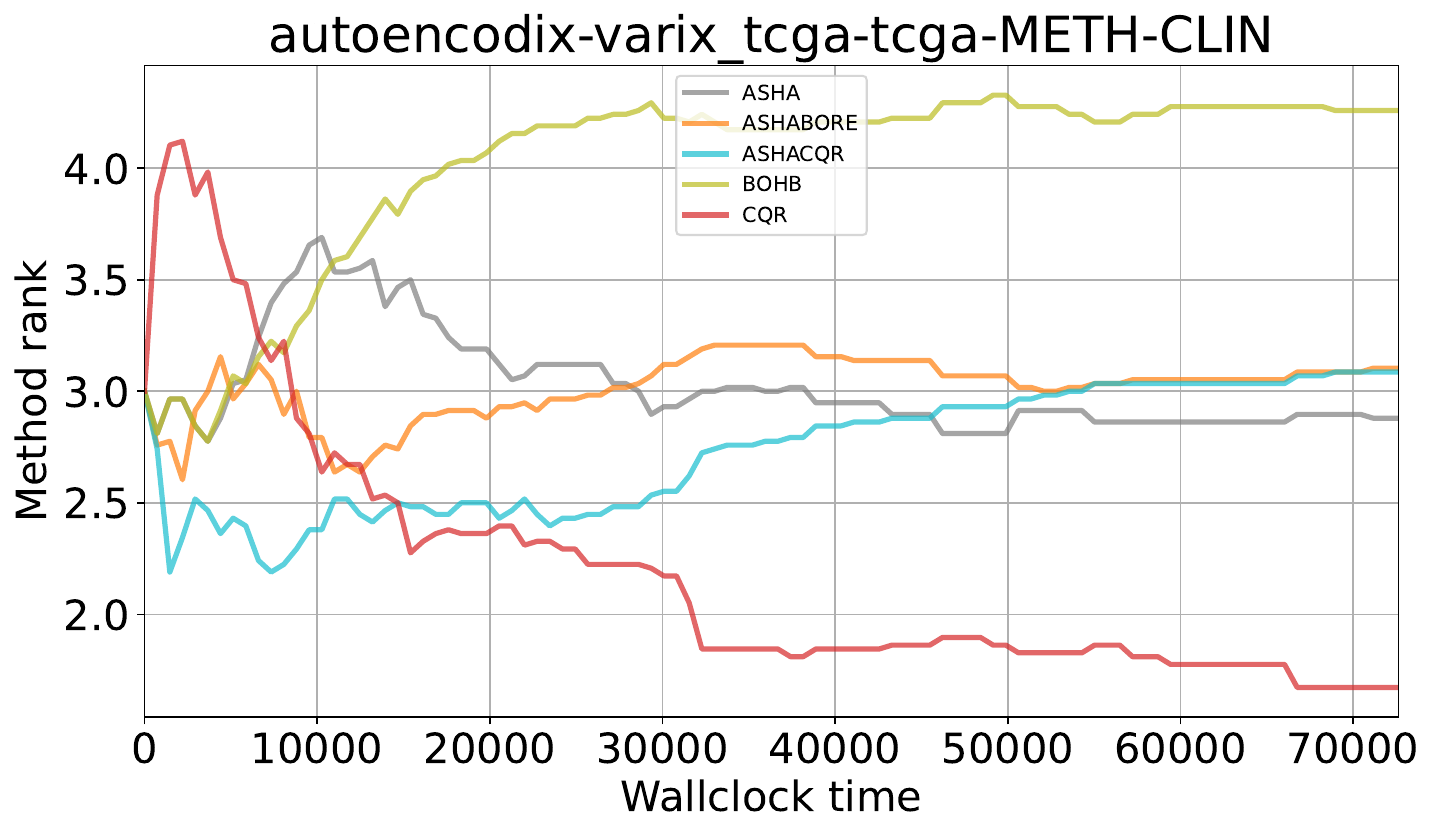} \\
    \midrule
    \multicolumn{3}{c}{\textbf{autoencodix-varix\_tcga-tcga-RNA-CLIN}} \\
    \includegraphics[width=0.32\textwidth]{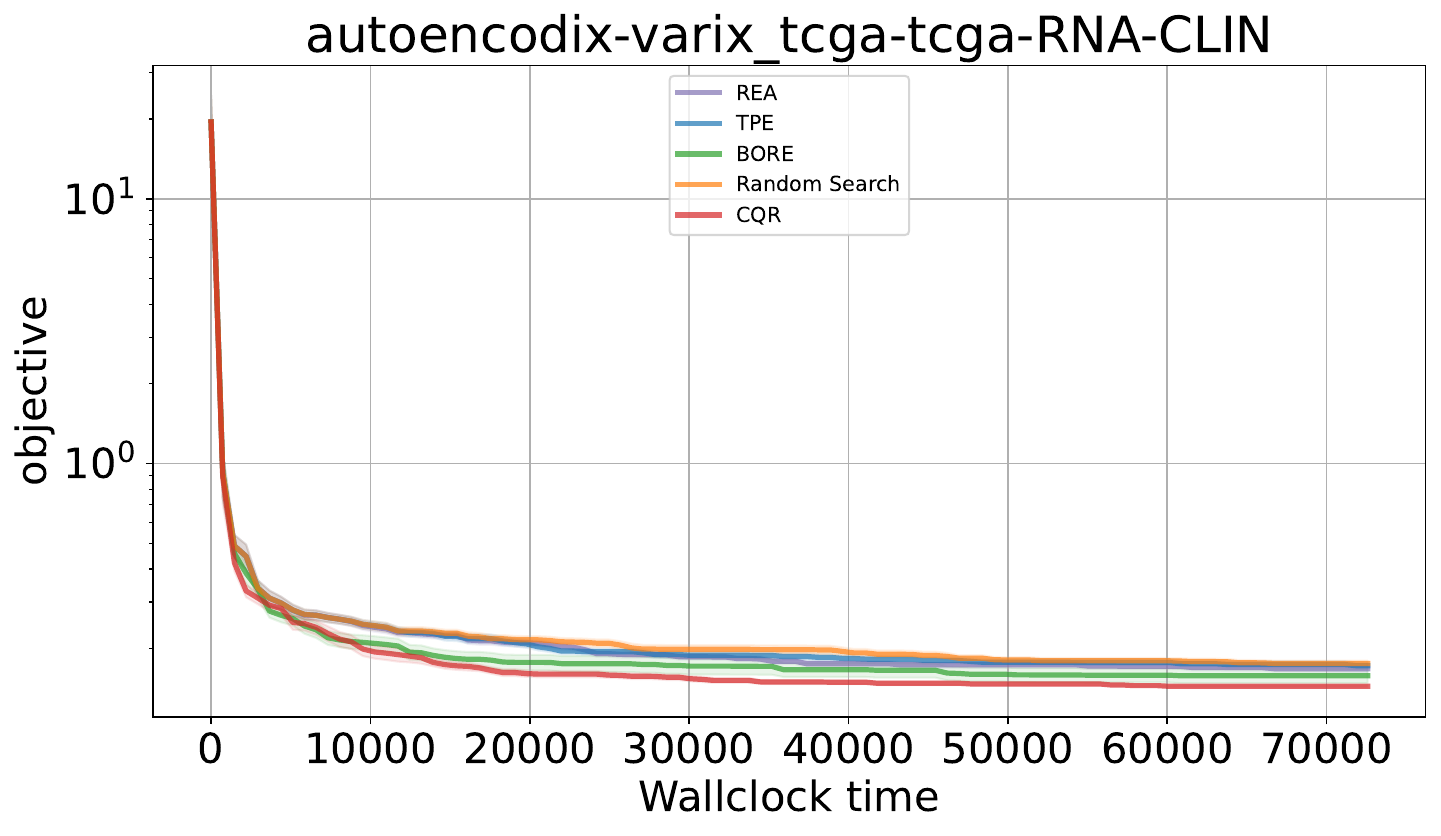} &
    \includegraphics[width=0.32\textwidth]{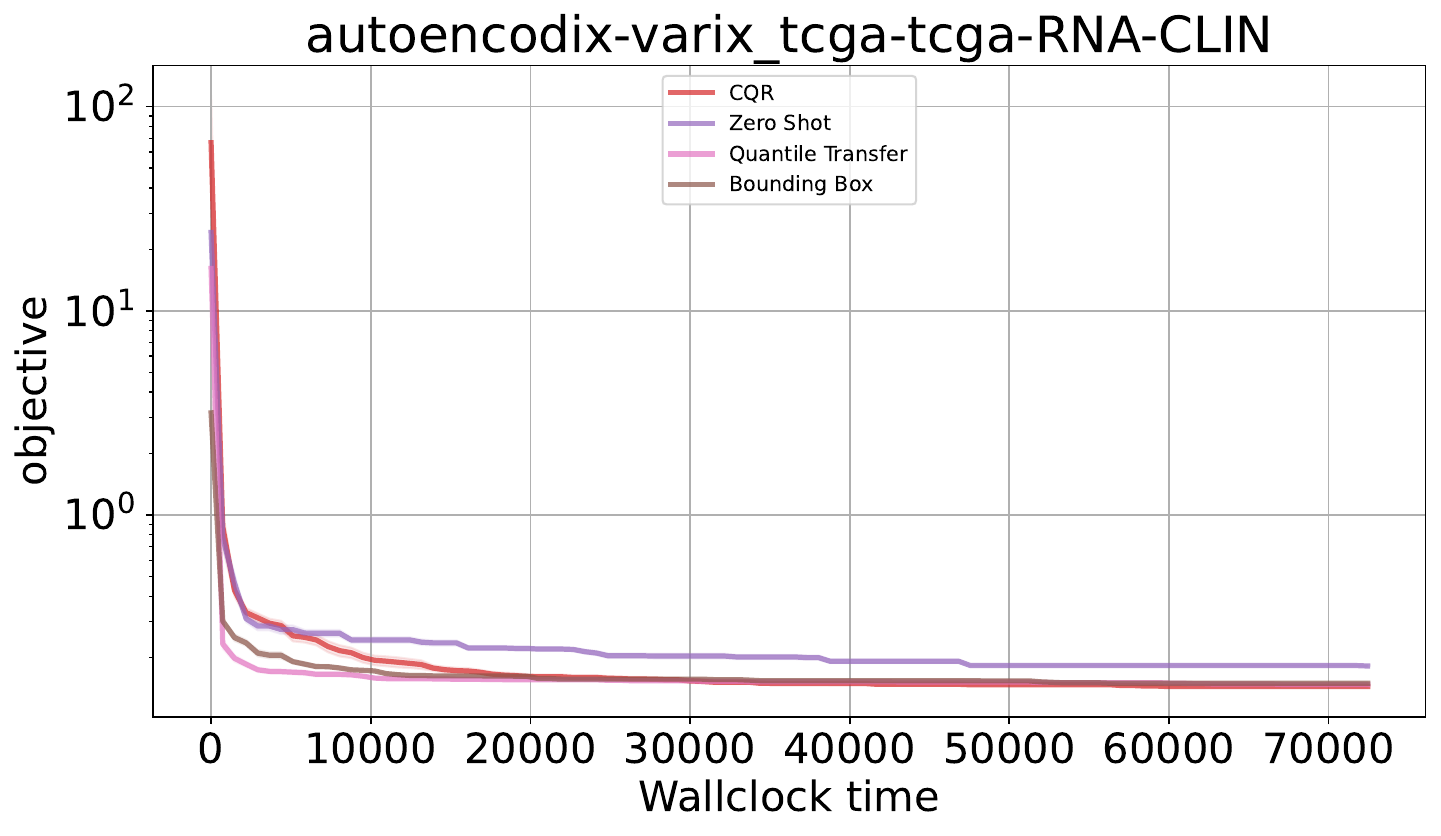} &
    \includegraphics[width=0.32\textwidth]{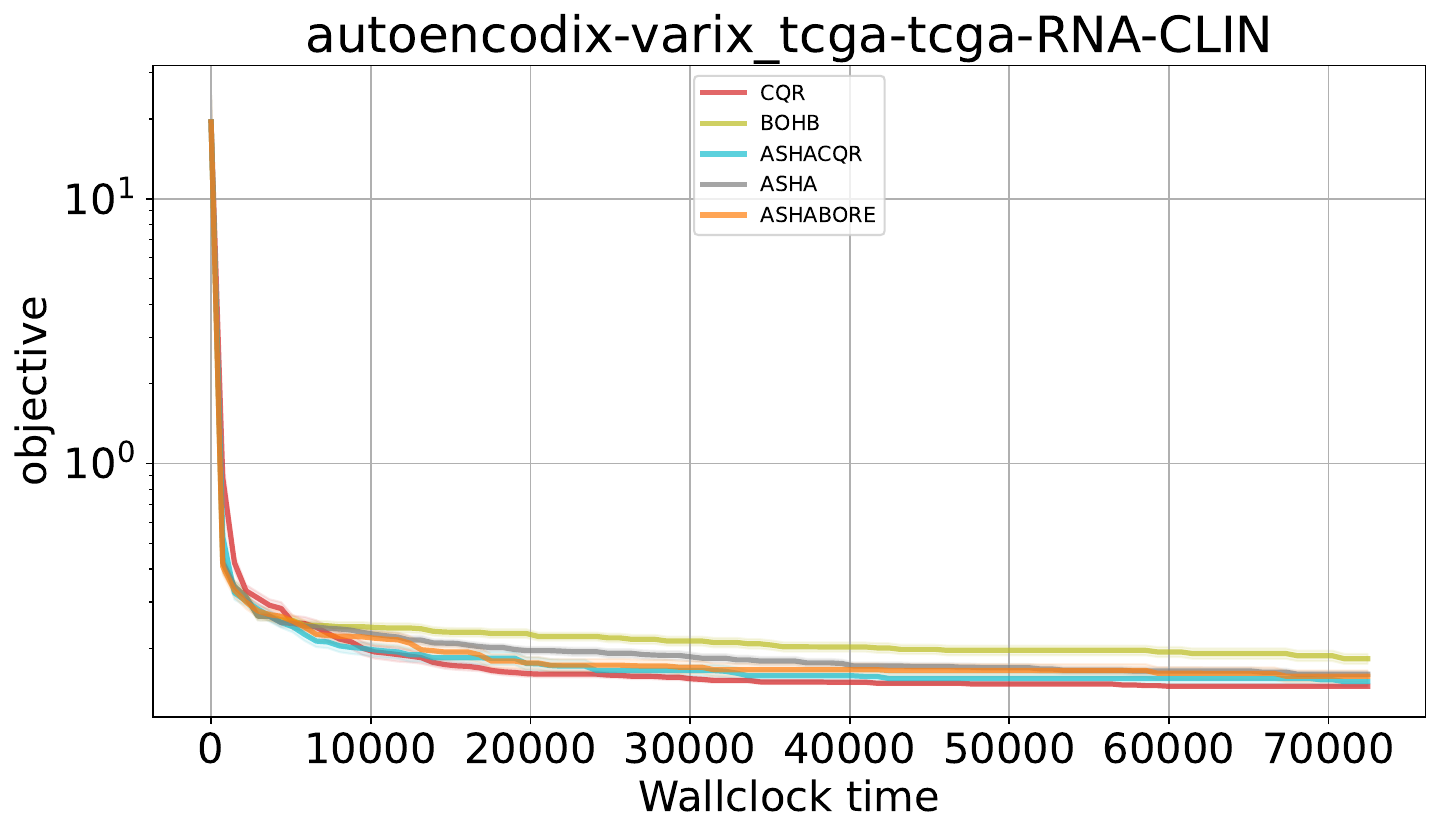} \\
    \includegraphics[width=0.32\textwidth]{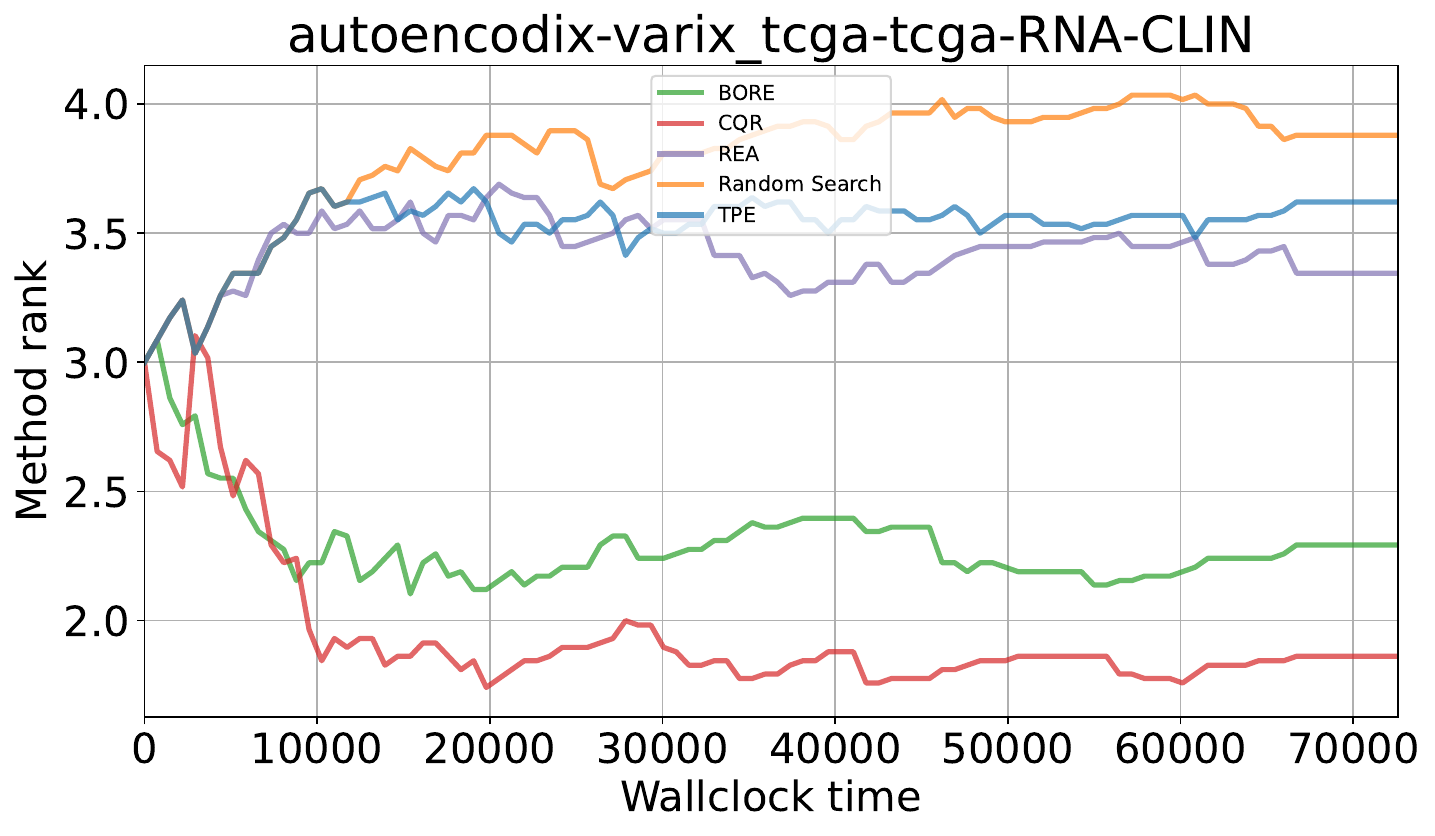} &
    \includegraphics[width=0.32\textwidth]{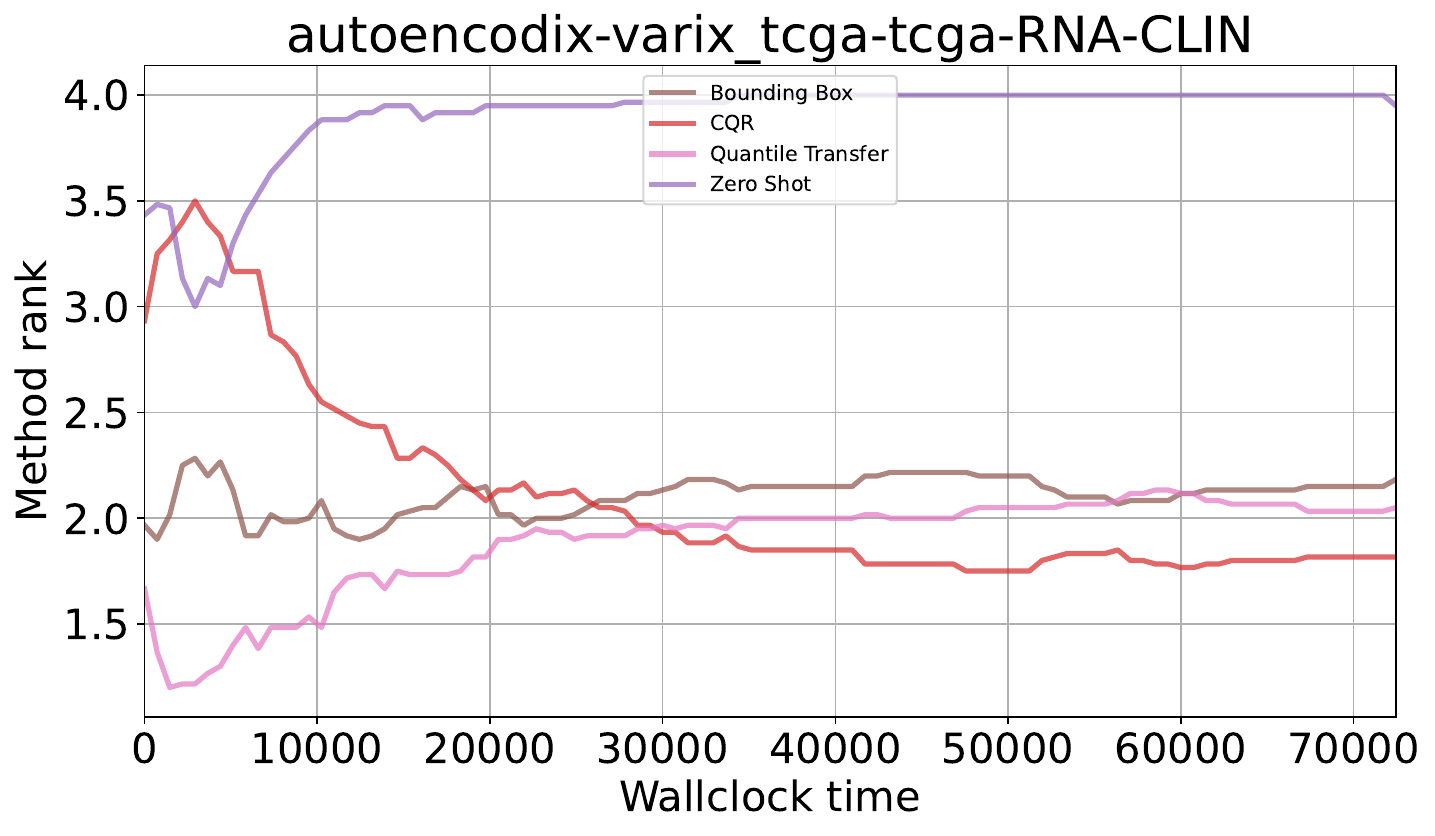} &
    \includegraphics[width=0.32\textwidth]{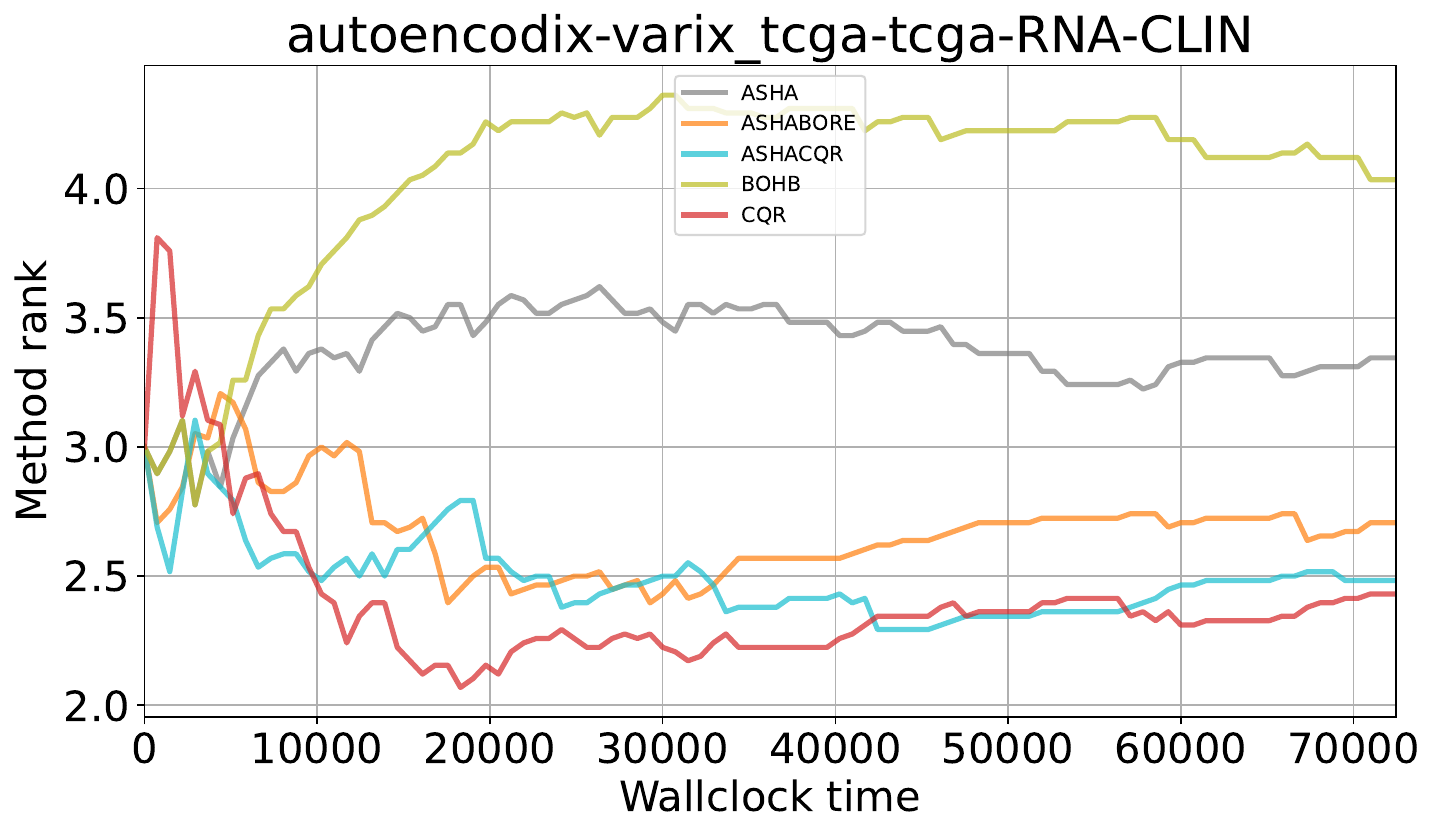} \\
    \end{tabular}
    \caption{Results for Varix tasks (Part 2).}
    \label{fig:varix_part2}
\end{figure}
\clearpage

\begin{figure}[htbp]
    \centering
    \setlength{\tabcolsep}{1pt}
    \begin{tabular}{ccc}
    \multicolumn{3}{c}{\textbf{autoencodix-varix\_tcga-tcga-RNA-DNA-METH-CLIN}} \\
    \textbf{Single-Fidelity} & \textbf{Transfer Learning} & \textbf{Multi-Fidelity} \\
    \includegraphics[width=0.32\textwidth]{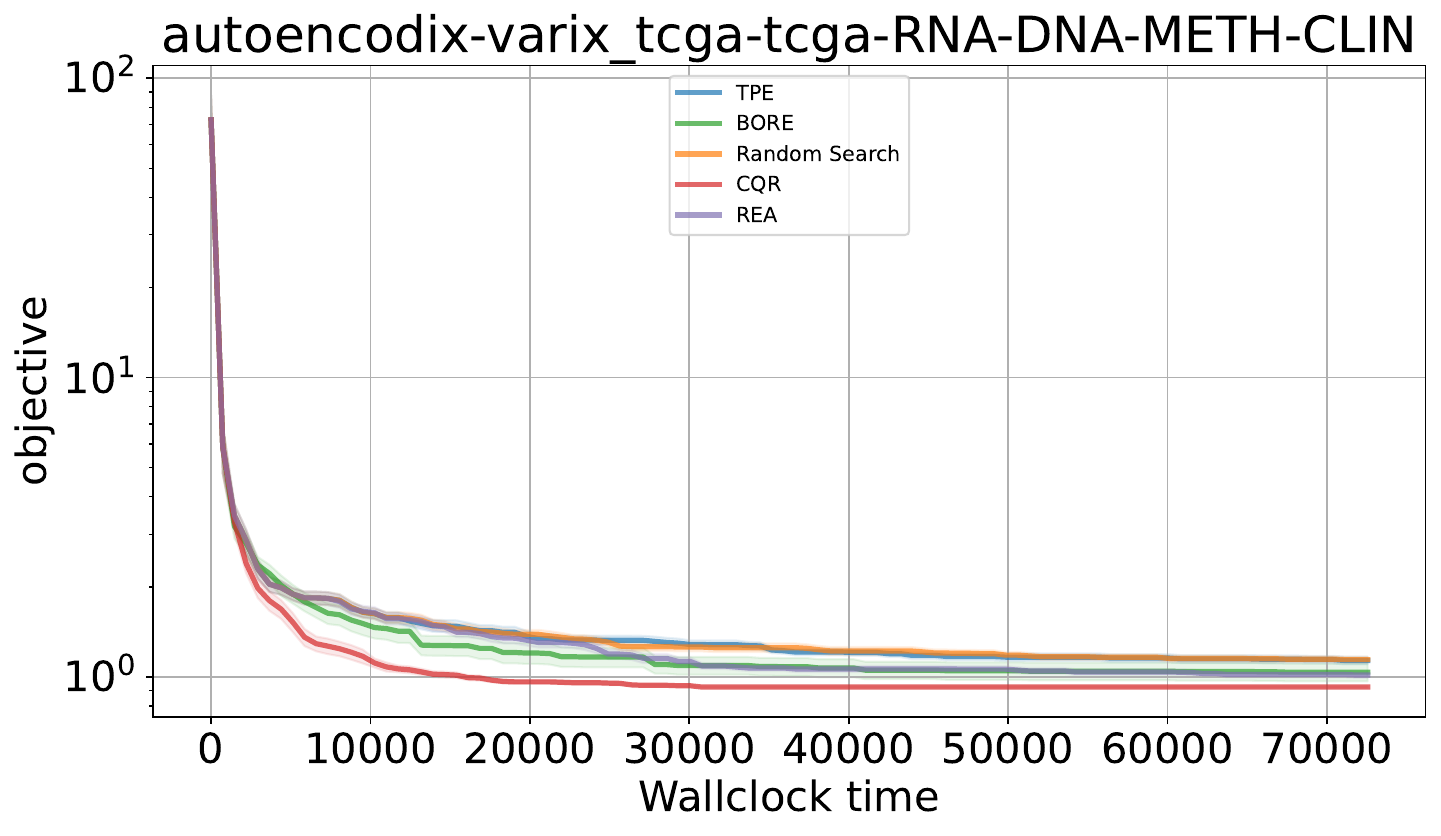} &
    \includegraphics[width=0.32\textwidth]{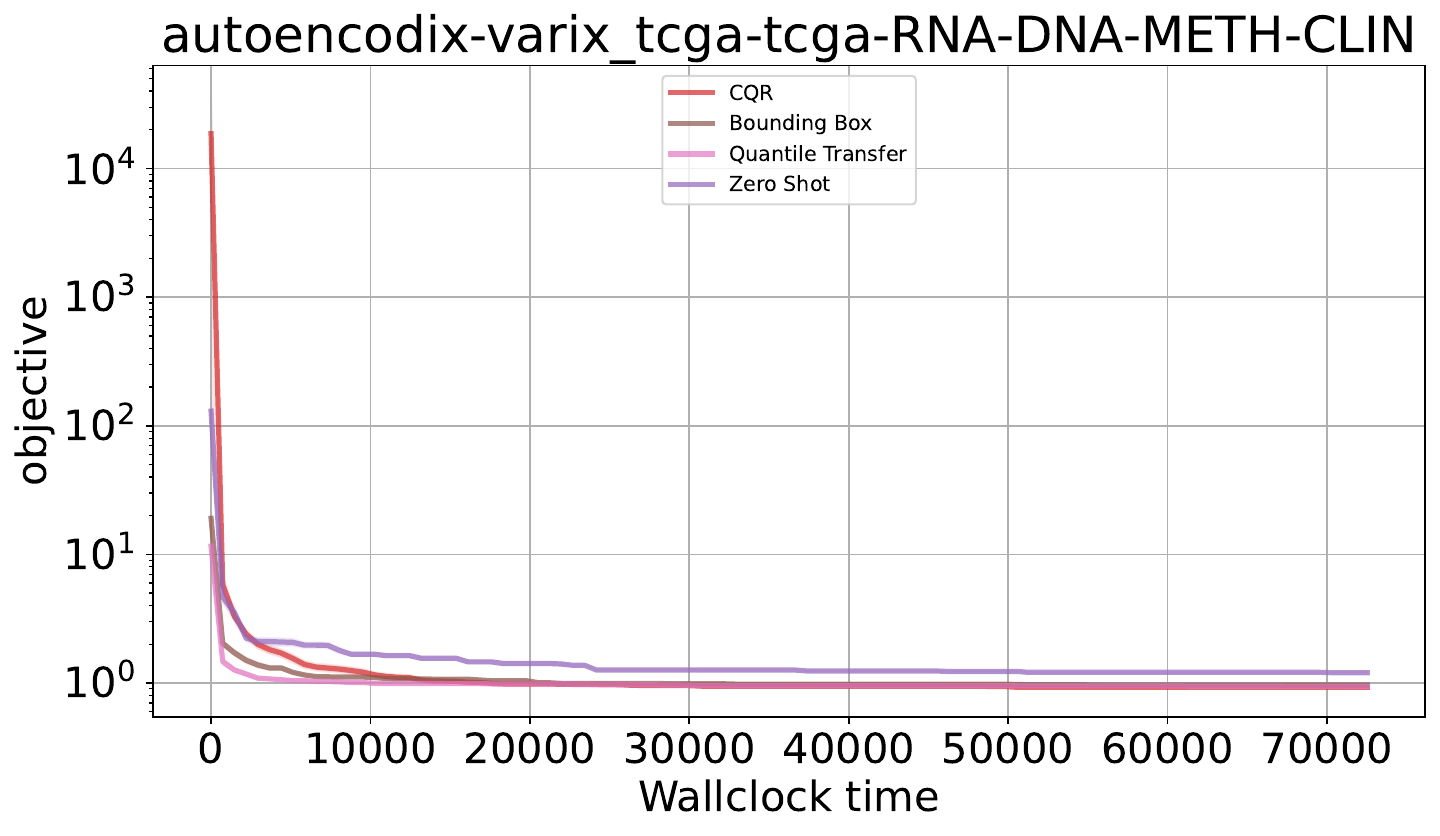} &
    \includegraphics[width=0.32\textwidth]{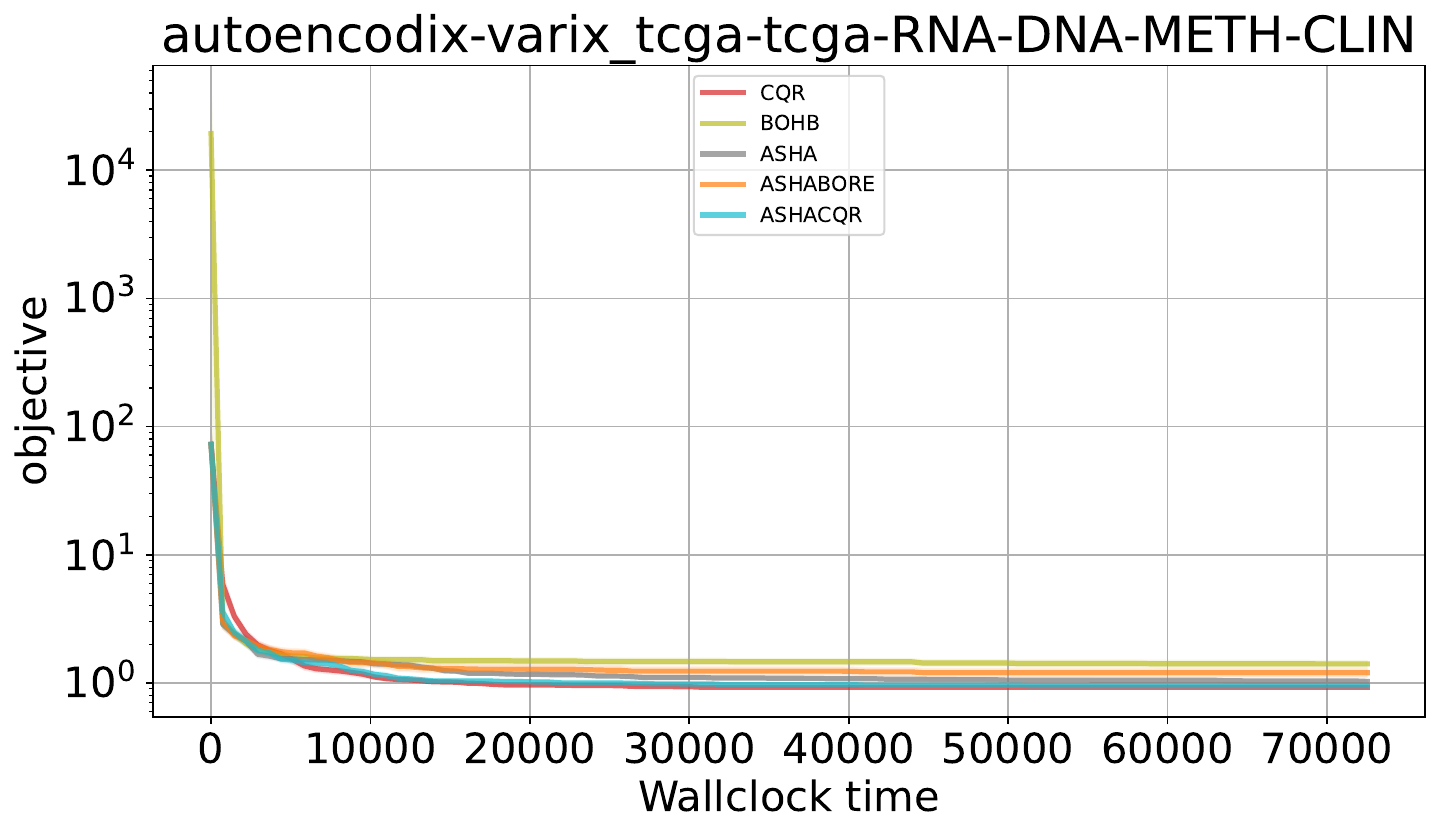} \\
    \includegraphics[width=0.32\textwidth]{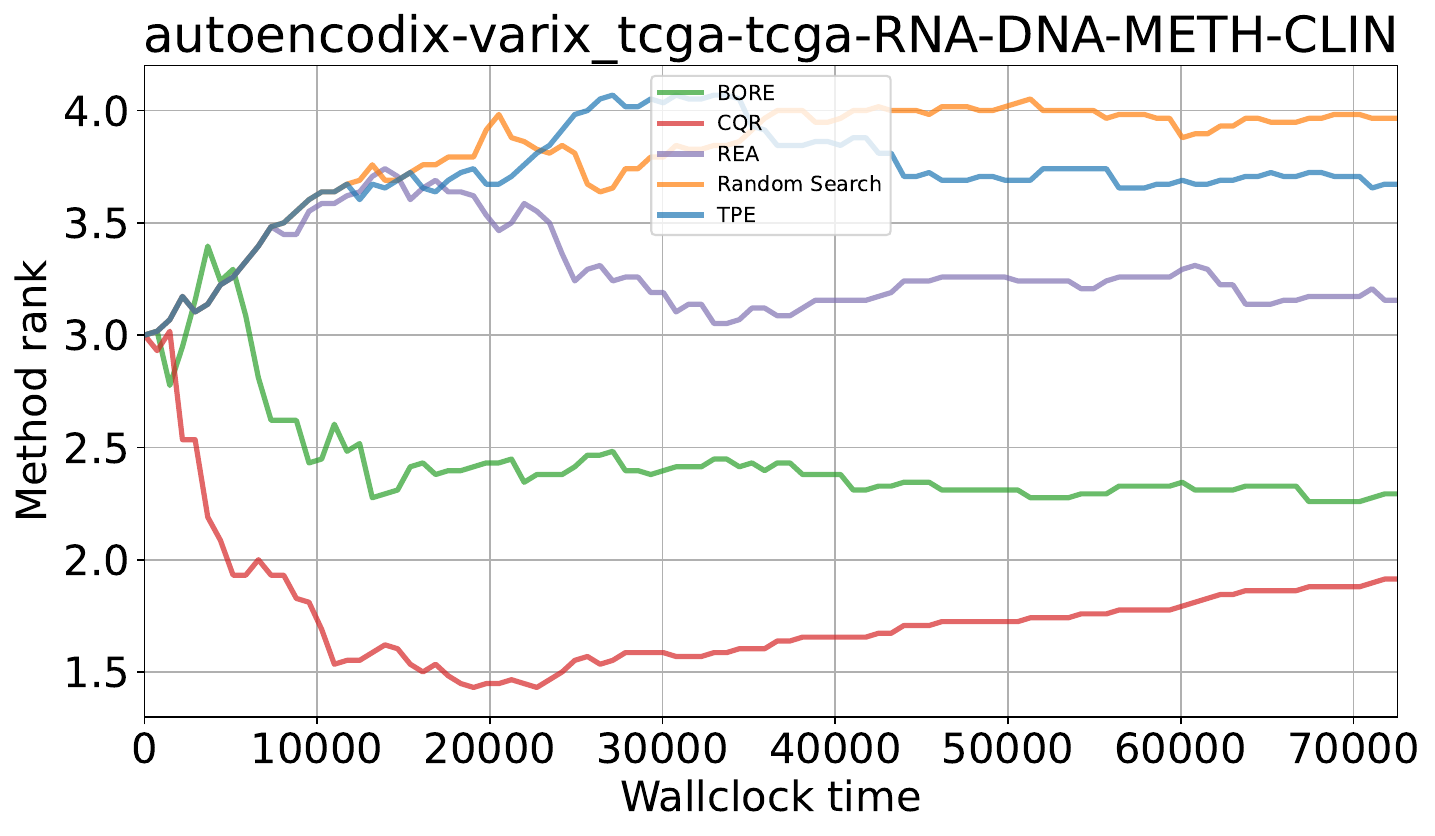} &
    \includegraphics[width=0.32\textwidth]{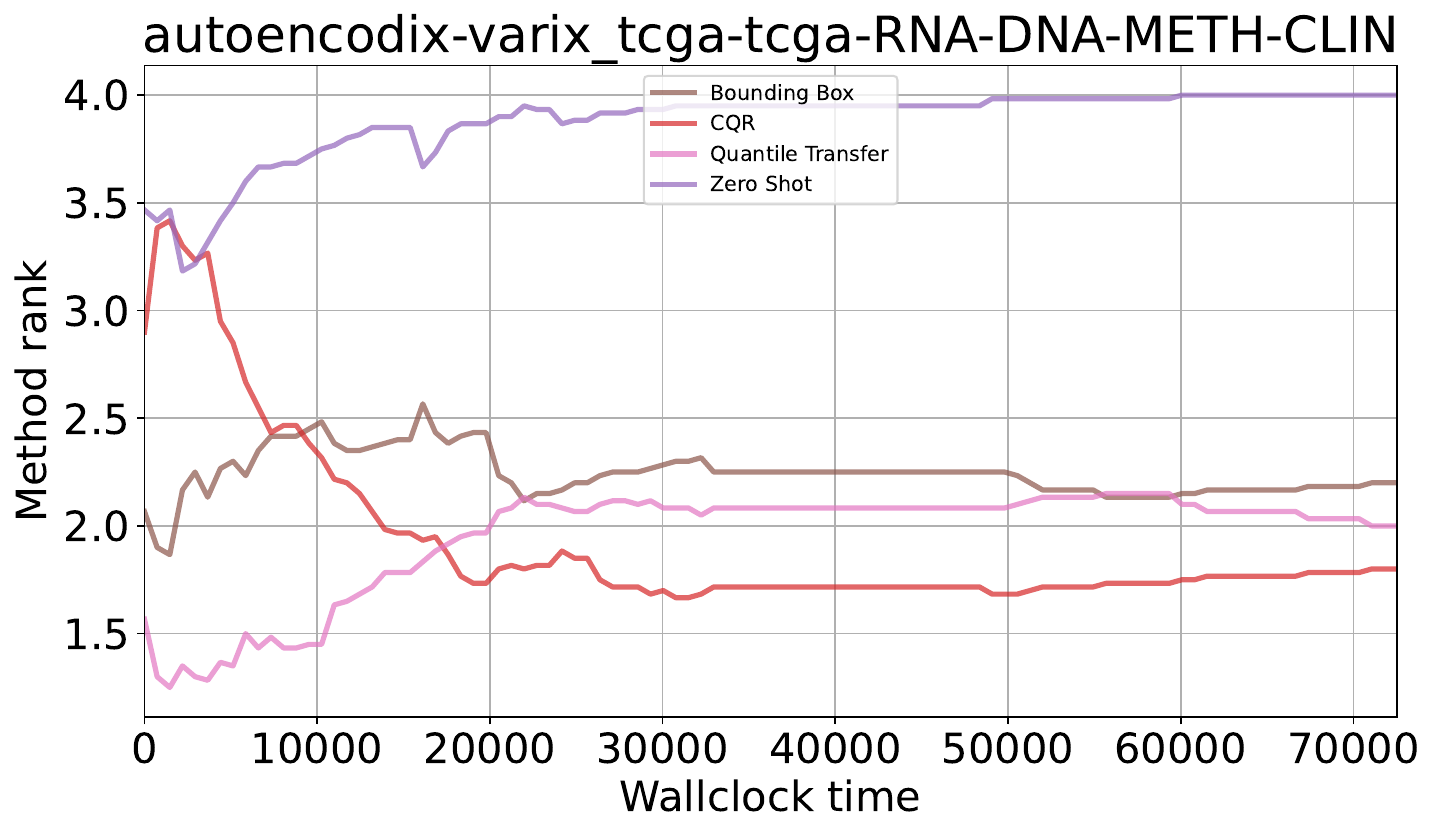} &
    \includegraphics[width=0.32\textwidth]{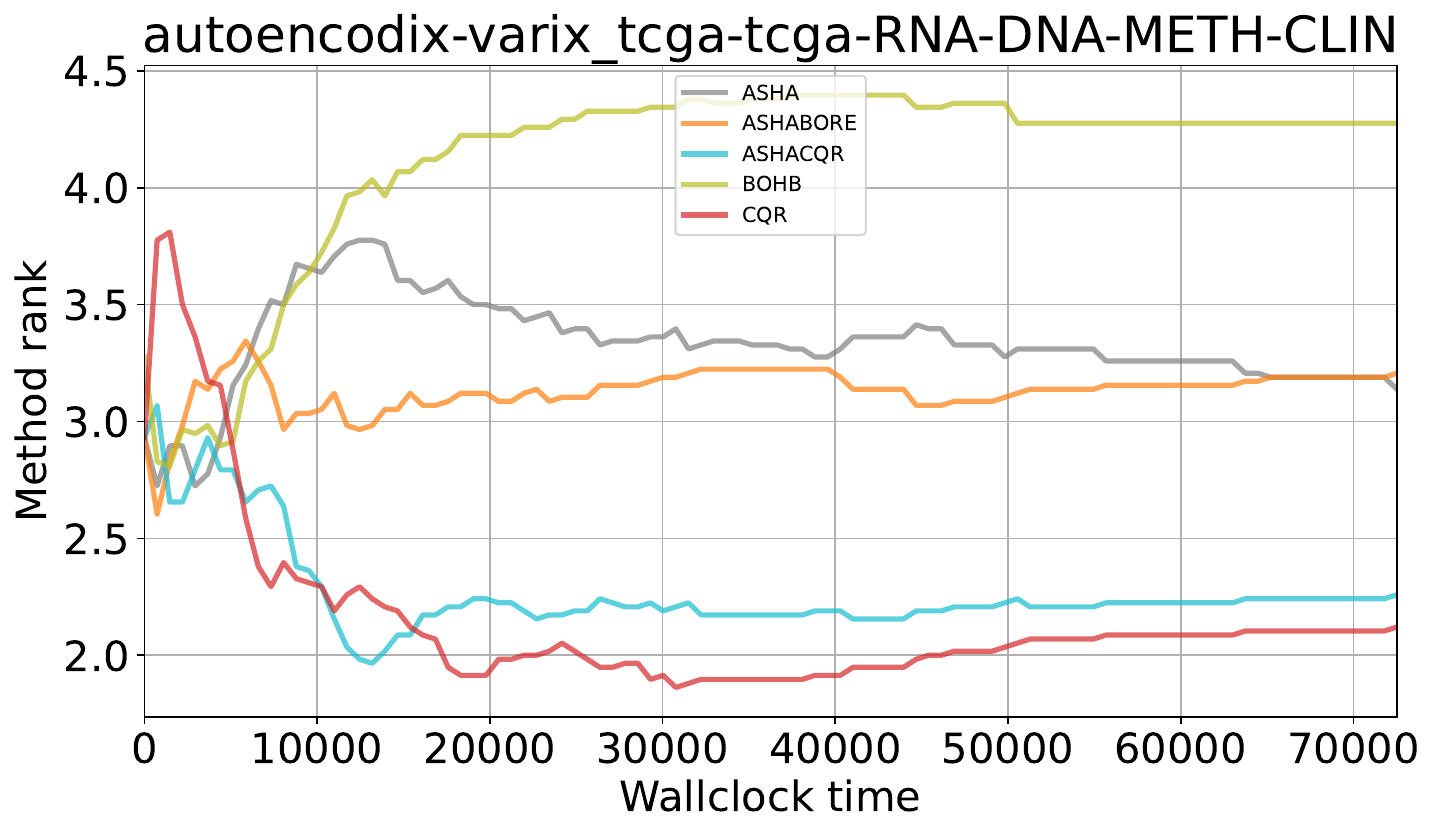} \\
    \end{tabular}
    \caption{Results for Varix (autoencodix-varix\_tcga-tcga-RNA-DNA-METH-CLIN).}
    \label{fig:varix_part3}
\end{figure}
\clearpage

\subsection{Optimization Trajectories of Downstream Performance}
\label{app:maximization-trajectories}
\clearpage
% ── FIGURE 1: DISENTANGLIX ────────────────────────────────────────────────────
\begin{figure*}[t]
    \centering
    \setlength{\tabcolsep}{1pt}
    \begin{tabular}{ccc}
    \textbf{SCHC-METH} & \textbf{SCHC-RNA} & \textbf{SCHC-RNA-METH} \\
    \includegraphics[width=0.32\linewidth]{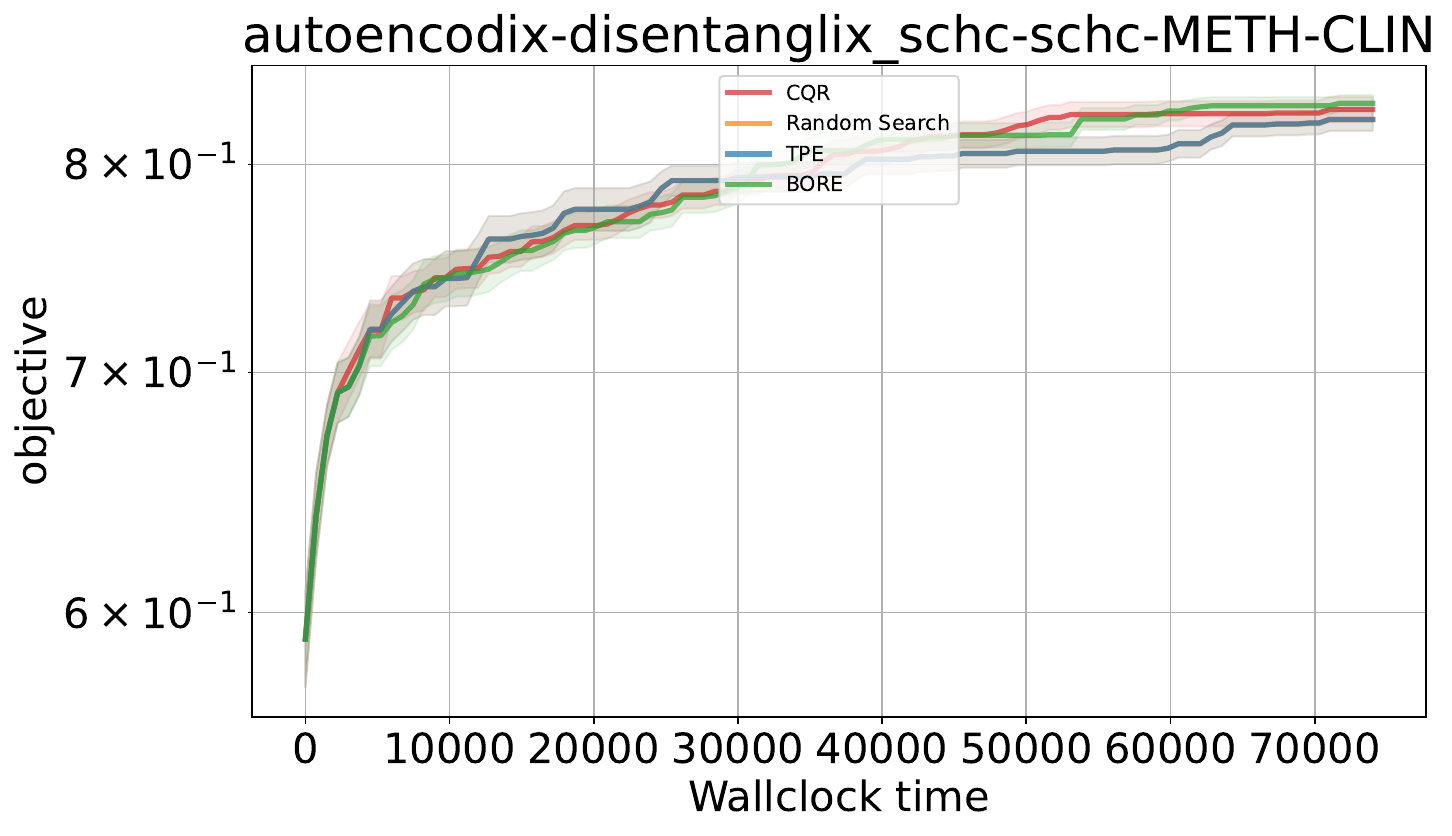} & \includegraphics[width=0.32\linewidth]{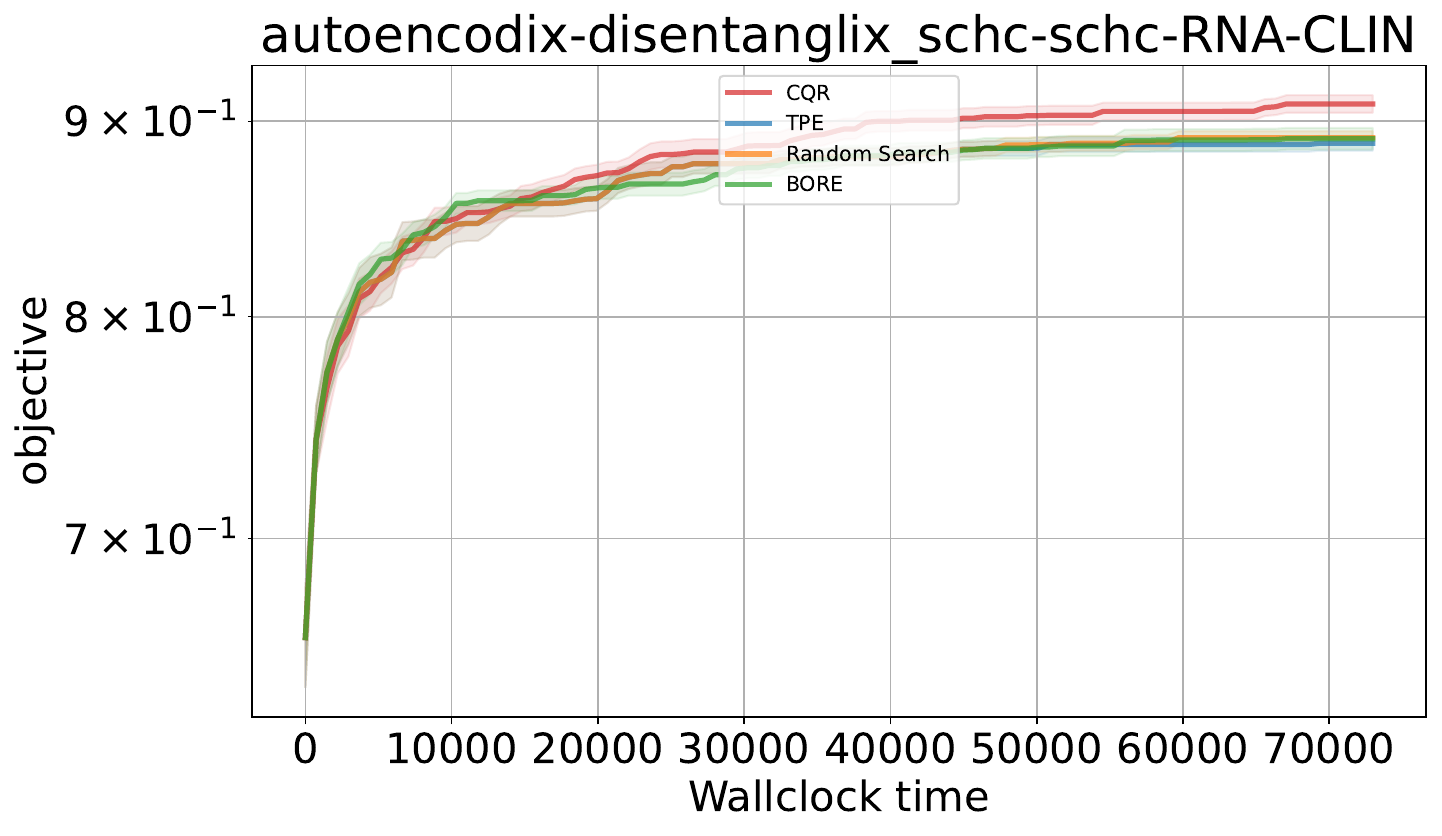} & \includegraphics[width=0.32\linewidth]{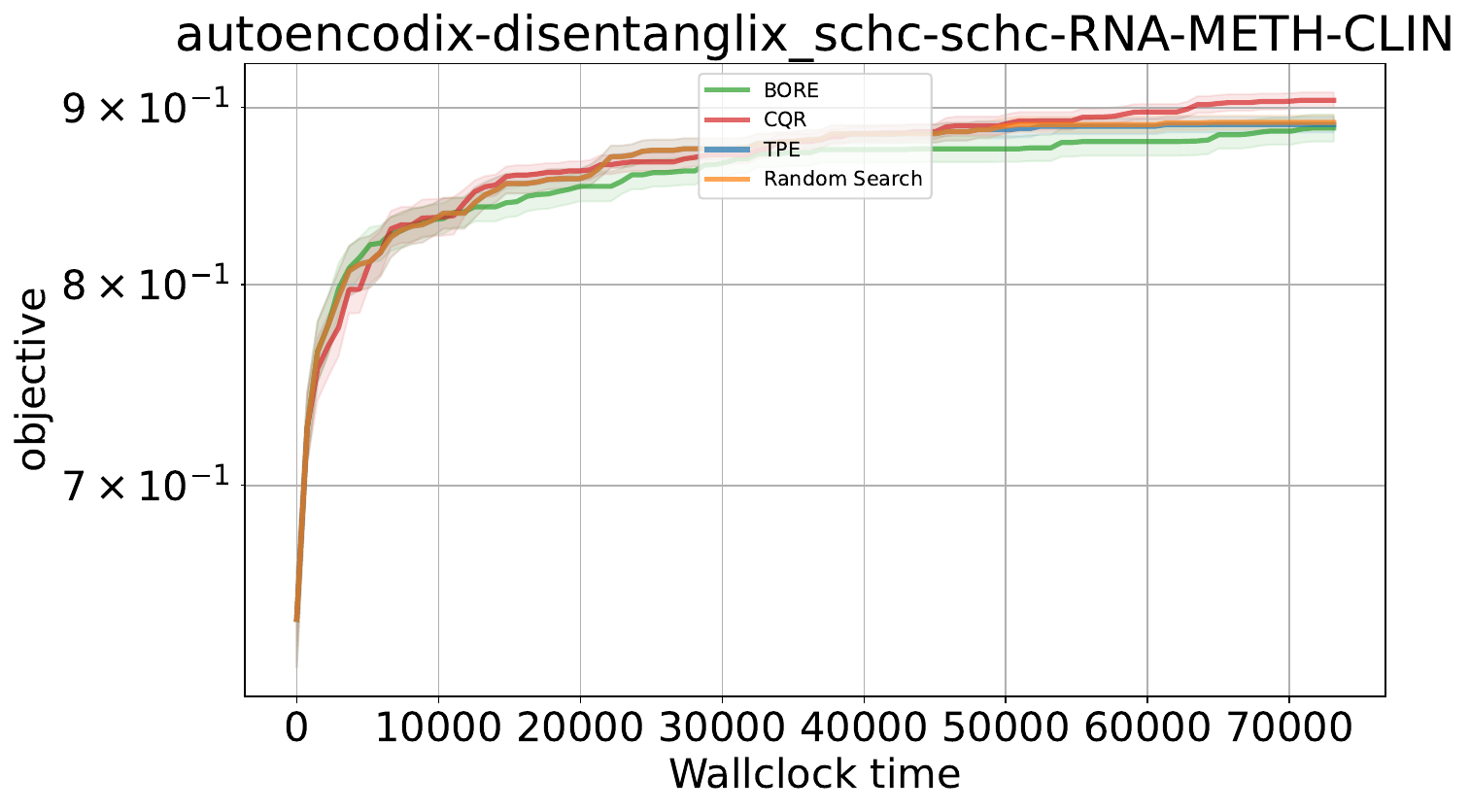} \\
    \midrule
    \textbf{TCGA-DNA} & \textbf{TCGA-METH} & \textbf{TCGA-RNA} \\
    \includegraphics[width=0.32\linewidth]{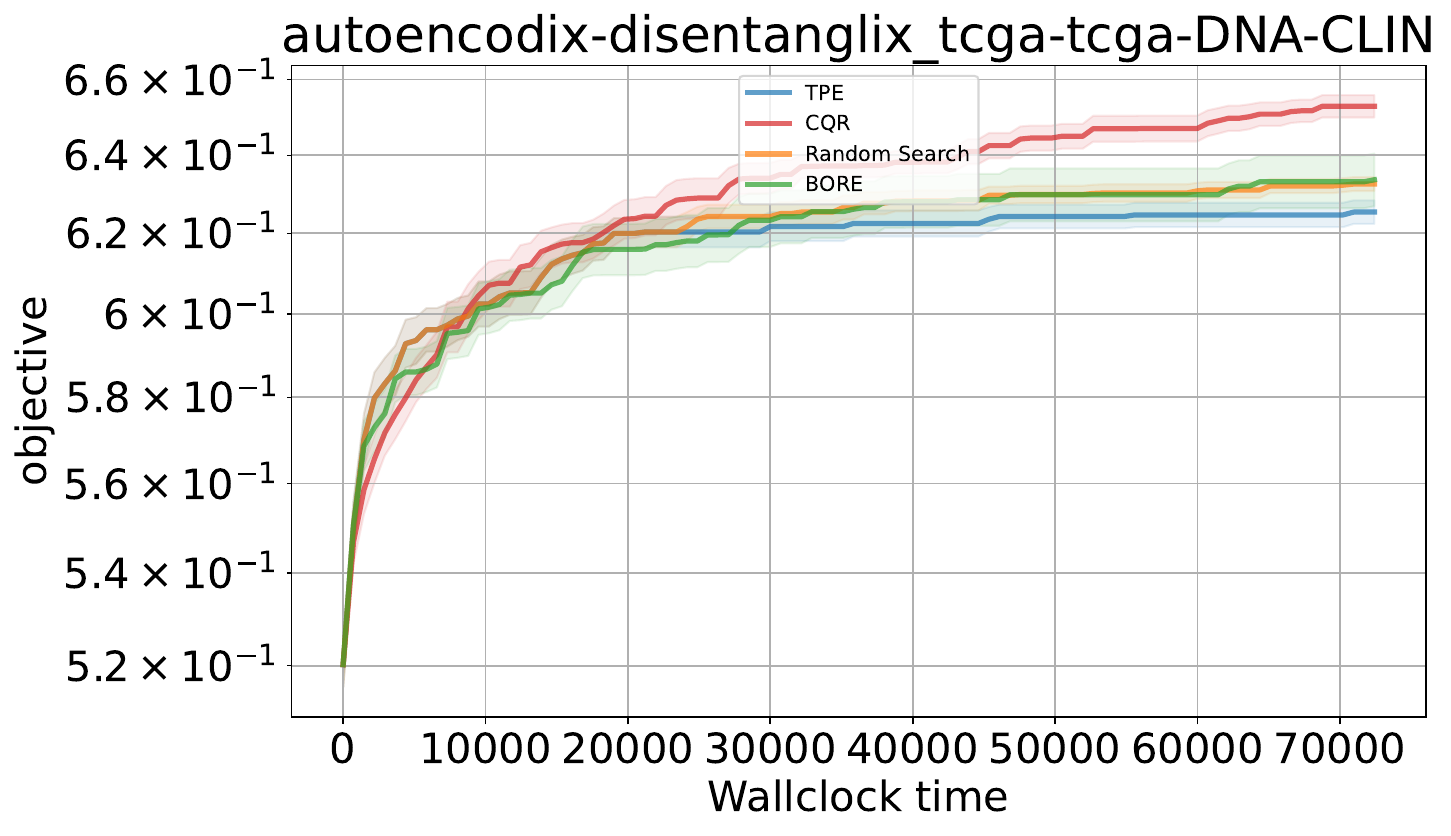} & \includegraphics[width=0.32\linewidth]{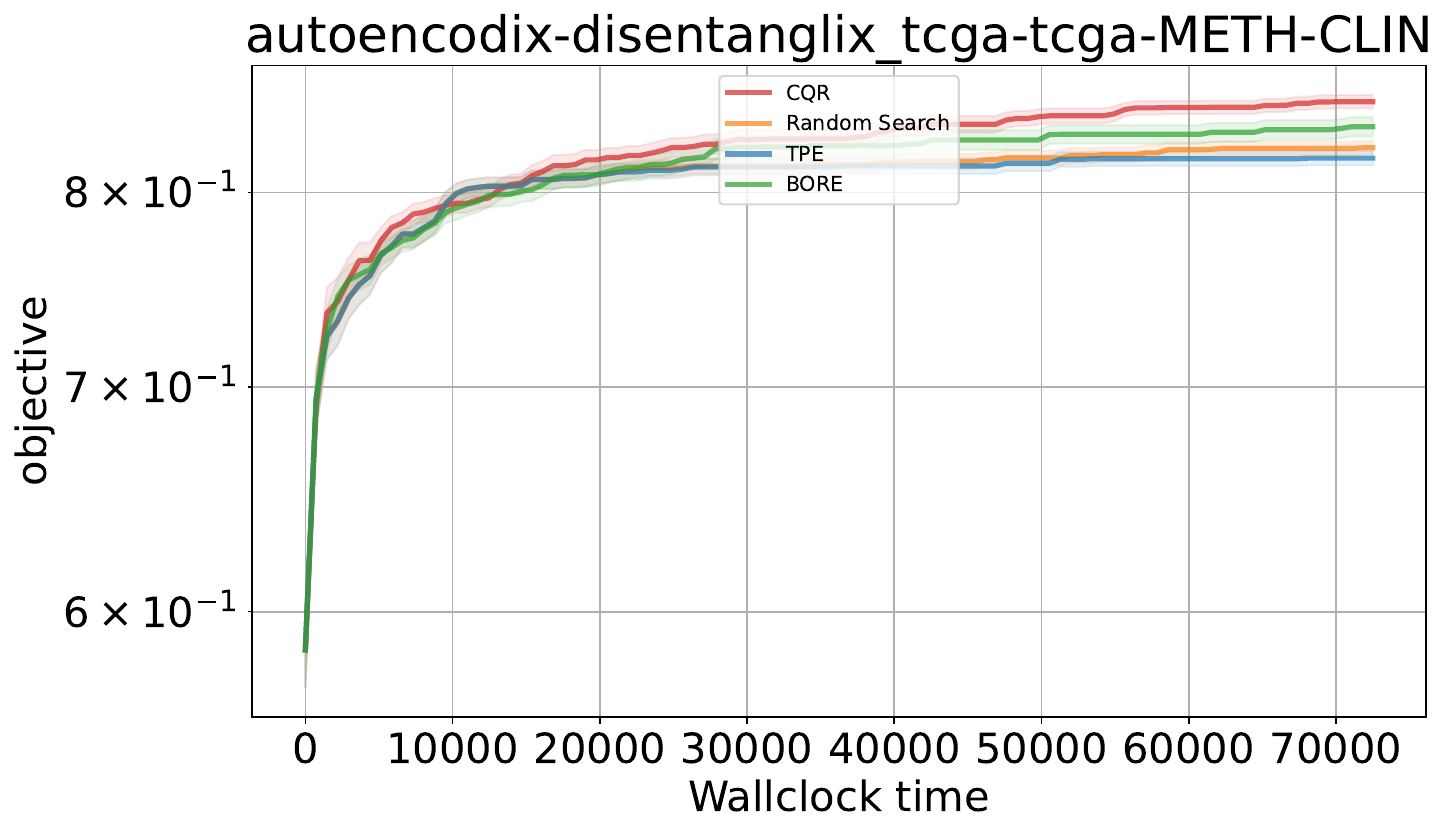} & \includegraphics[width=0.32\linewidth]{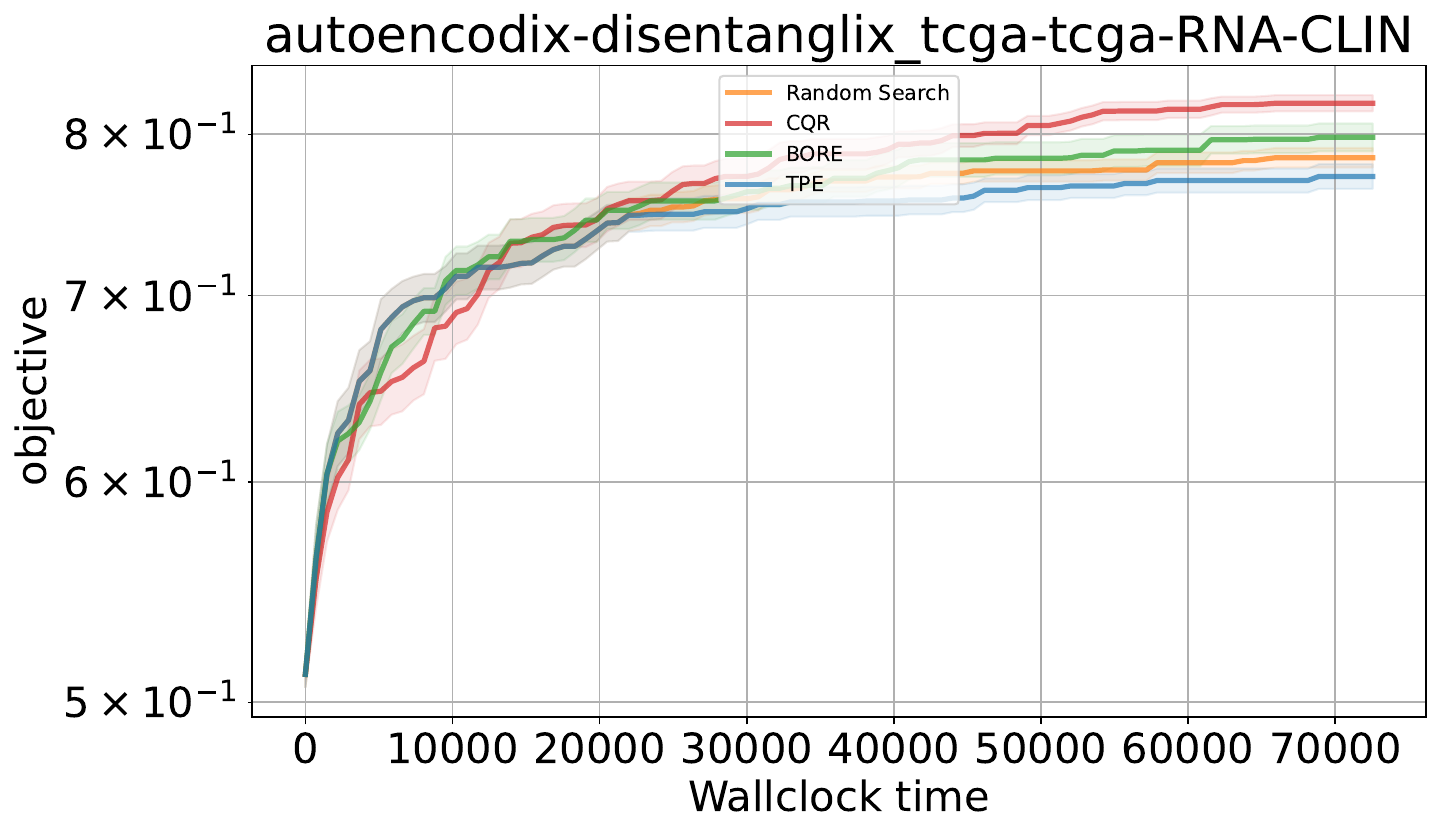} \\
    \midrule
    \textbf{TCGA-RNA-DNA-METH} &  &  \\
    \includegraphics[width=0.32\linewidth]{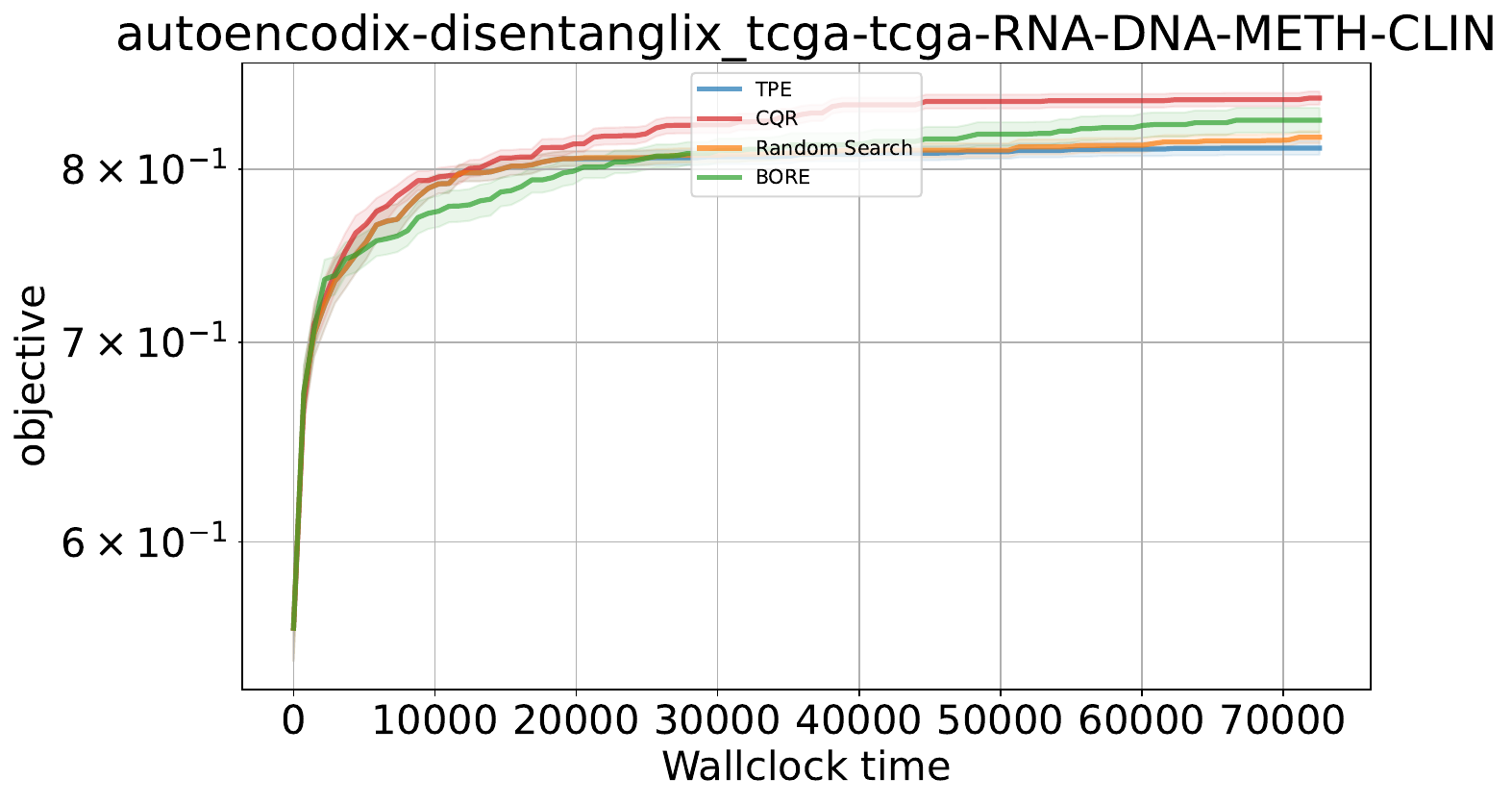} &  &  \\
    \end{tabular}
    \caption{Disentanglix: optimization trajectories maximizing downstream performance}
    \label{fig:optimizers-disentanglix}
\end{figure*}

% ── FIGURE 2: VARIX ───────────────────────────────────────────────────────────
\begin{figure*}[t]
    \centering
    \setlength{\tabcolsep}{1pt}
    \begin{tabular}{ccc}
    \textbf{SCHC-METH} & \textbf{SCHC-RNA} & \textbf{SCHC-RNA-METH} \\
    \includegraphics[width=0.32\linewidth]{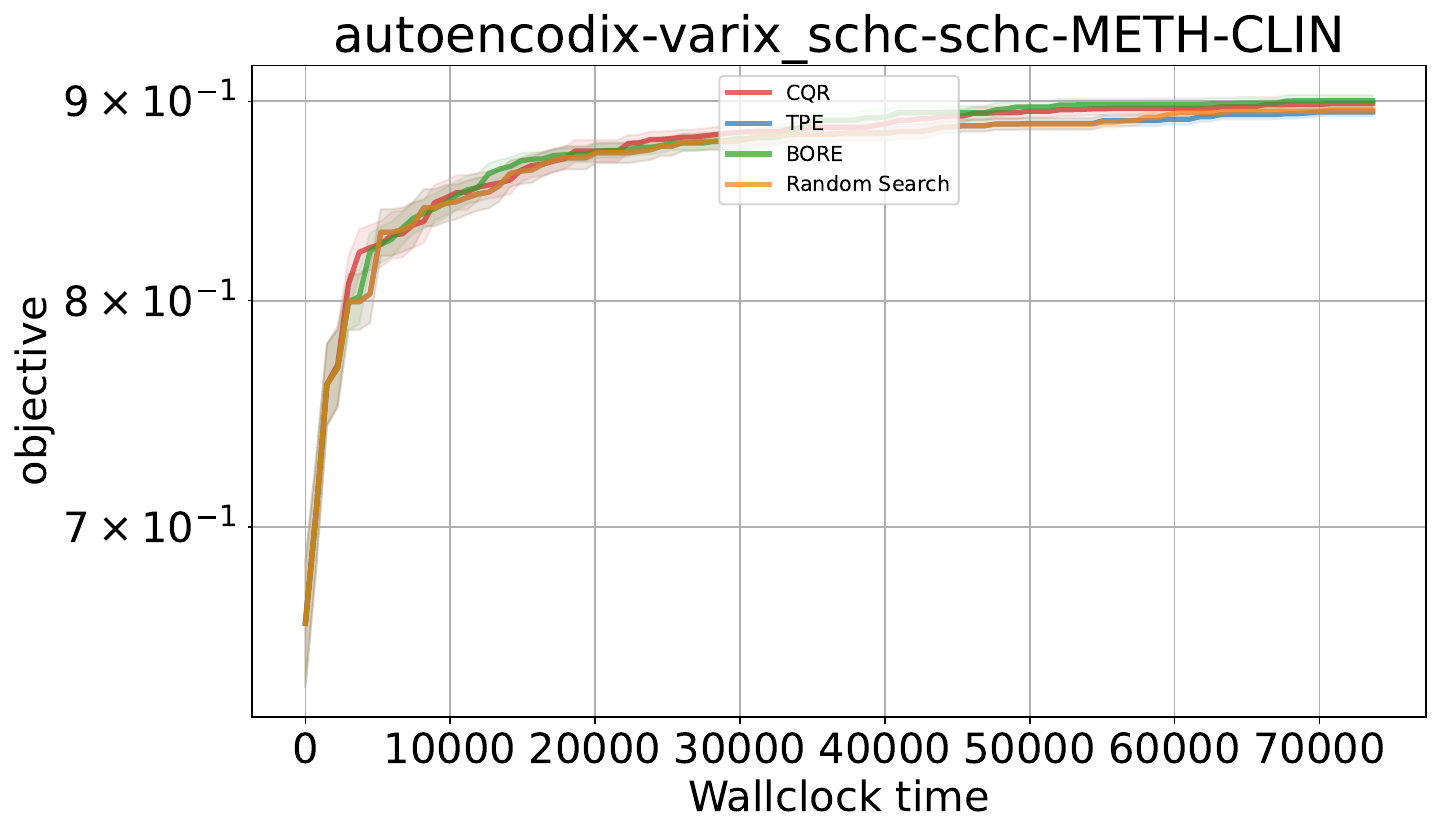} & \includegraphics[width=0.32\linewidth]{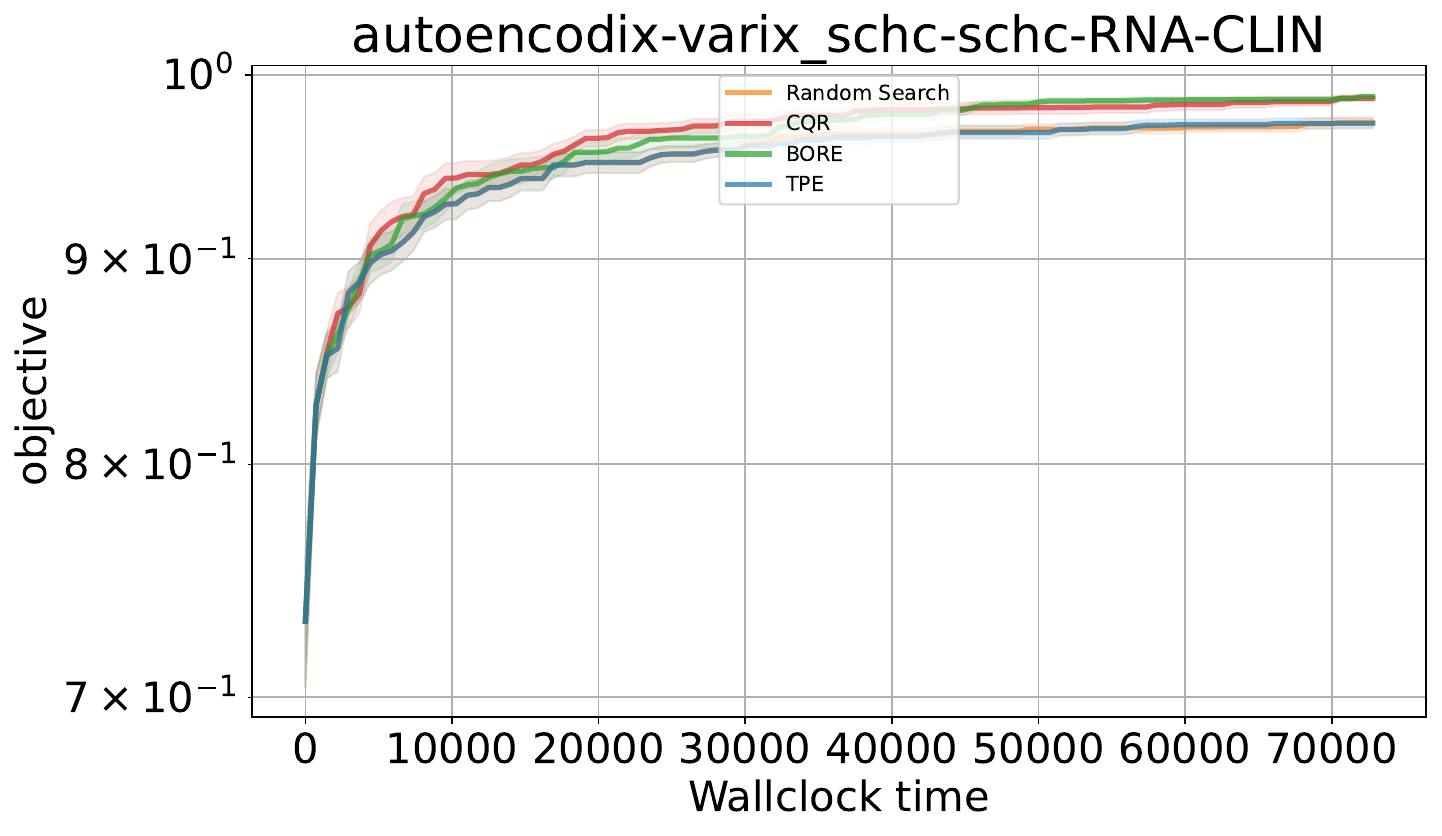} & \includegraphics[width=0.32\linewidth]{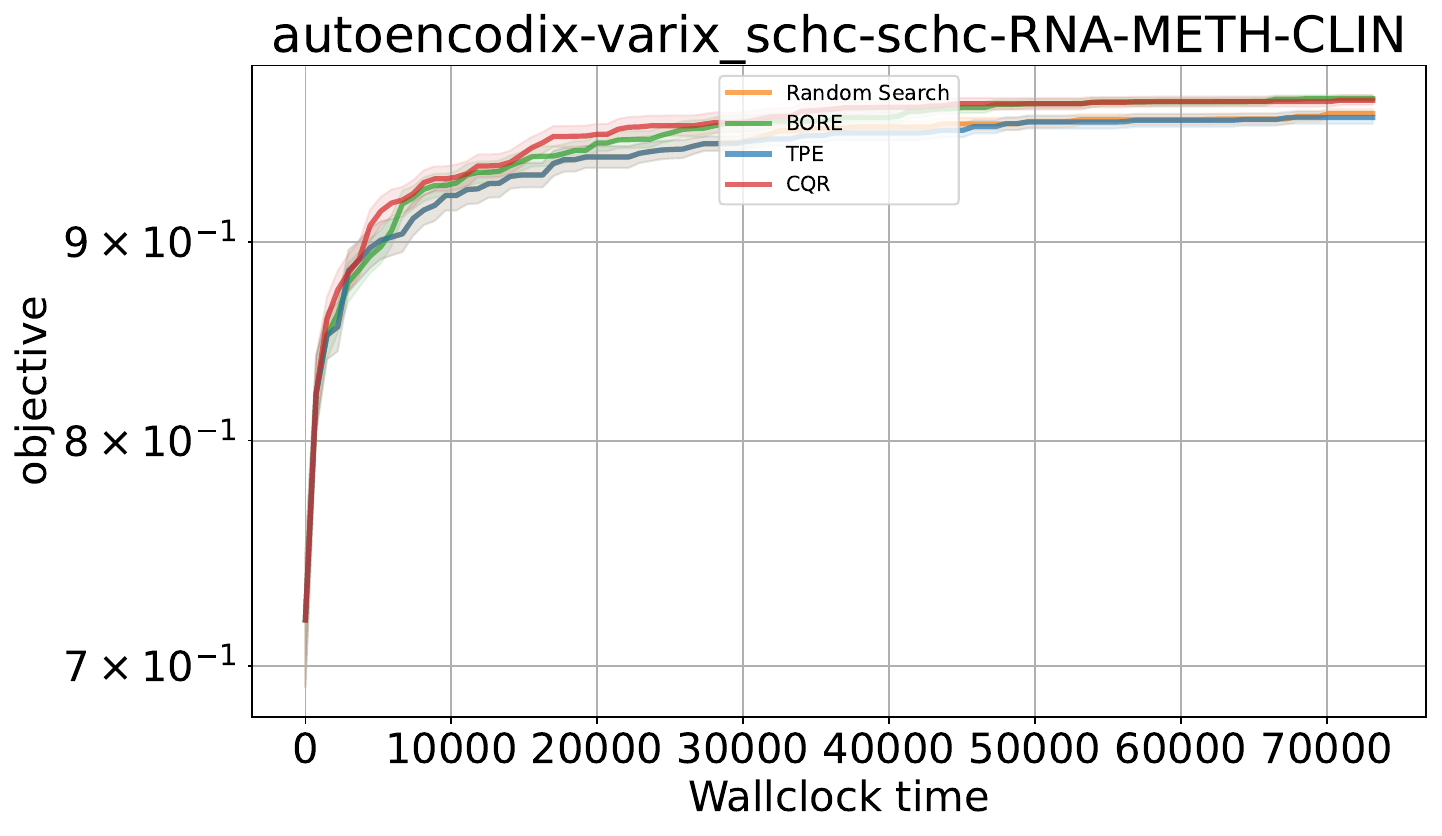} \\
    \midrule
    \textbf{TCGA-DNA} & \textbf{TCGA-METH} & \textbf{TCGA-RNA} \\
    \includegraphics[width=0.32\linewidth]{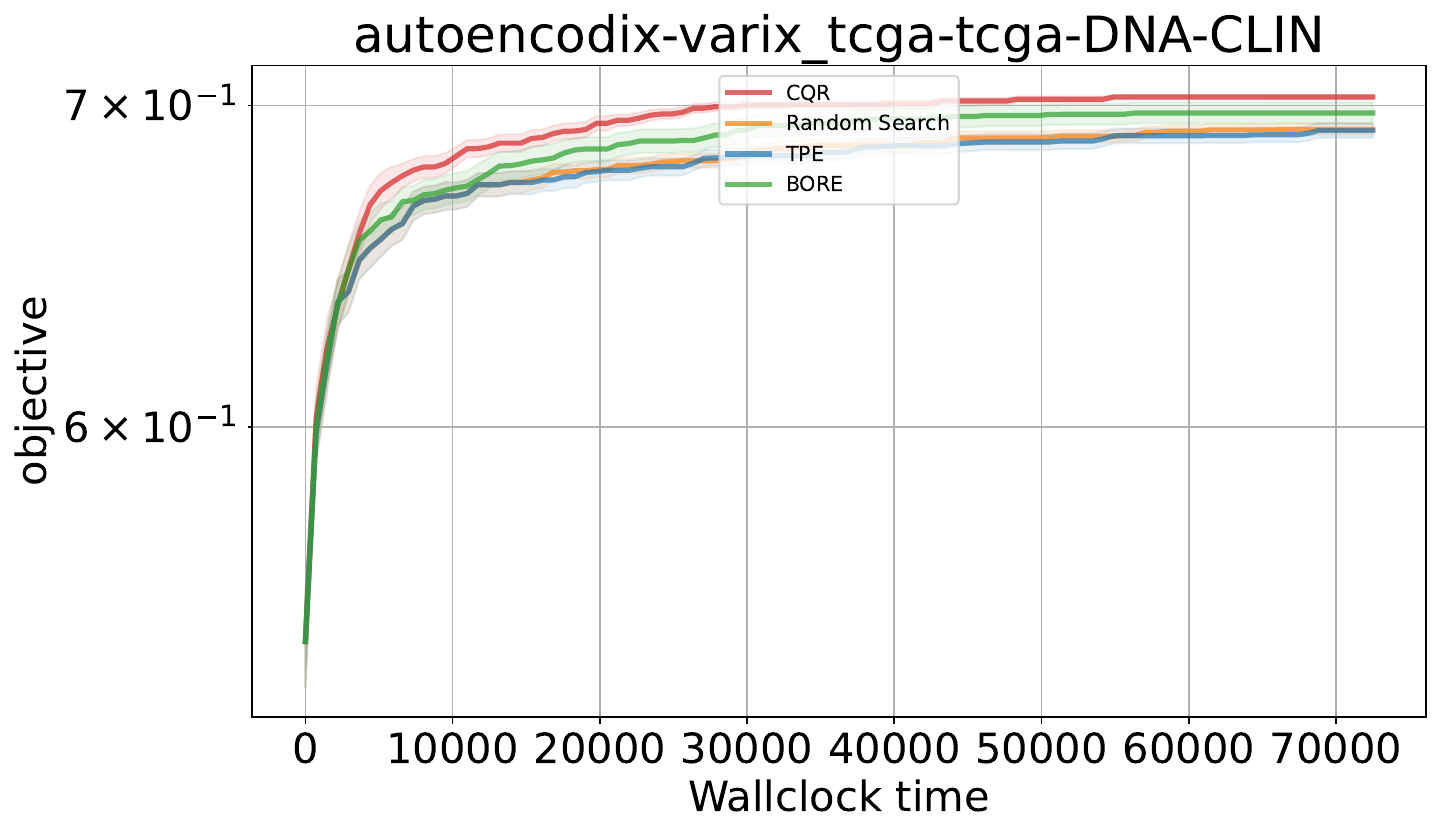} & \includegraphics[width=0.32\linewidth]{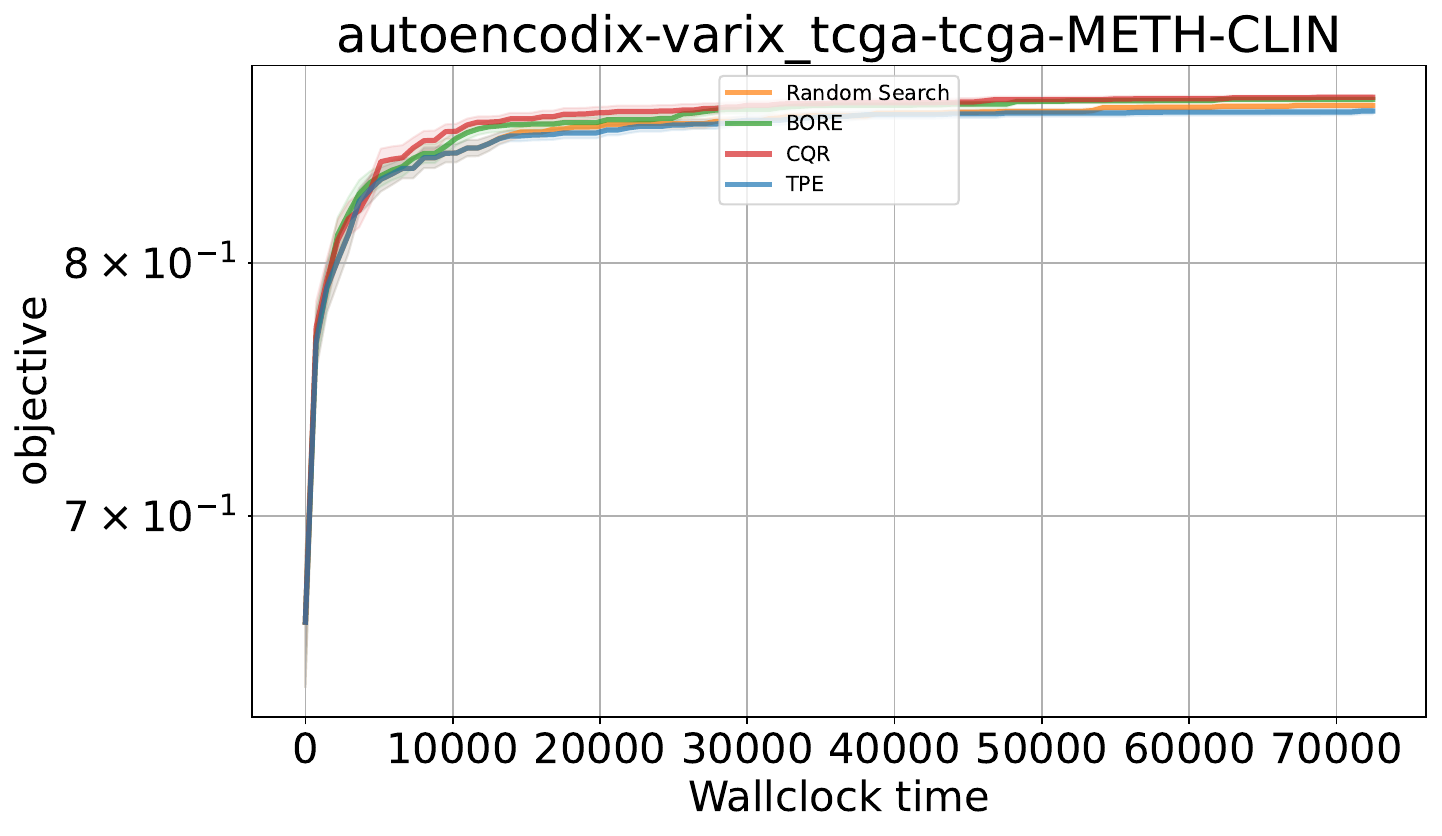} & \includegraphics[width=0.32\linewidth]{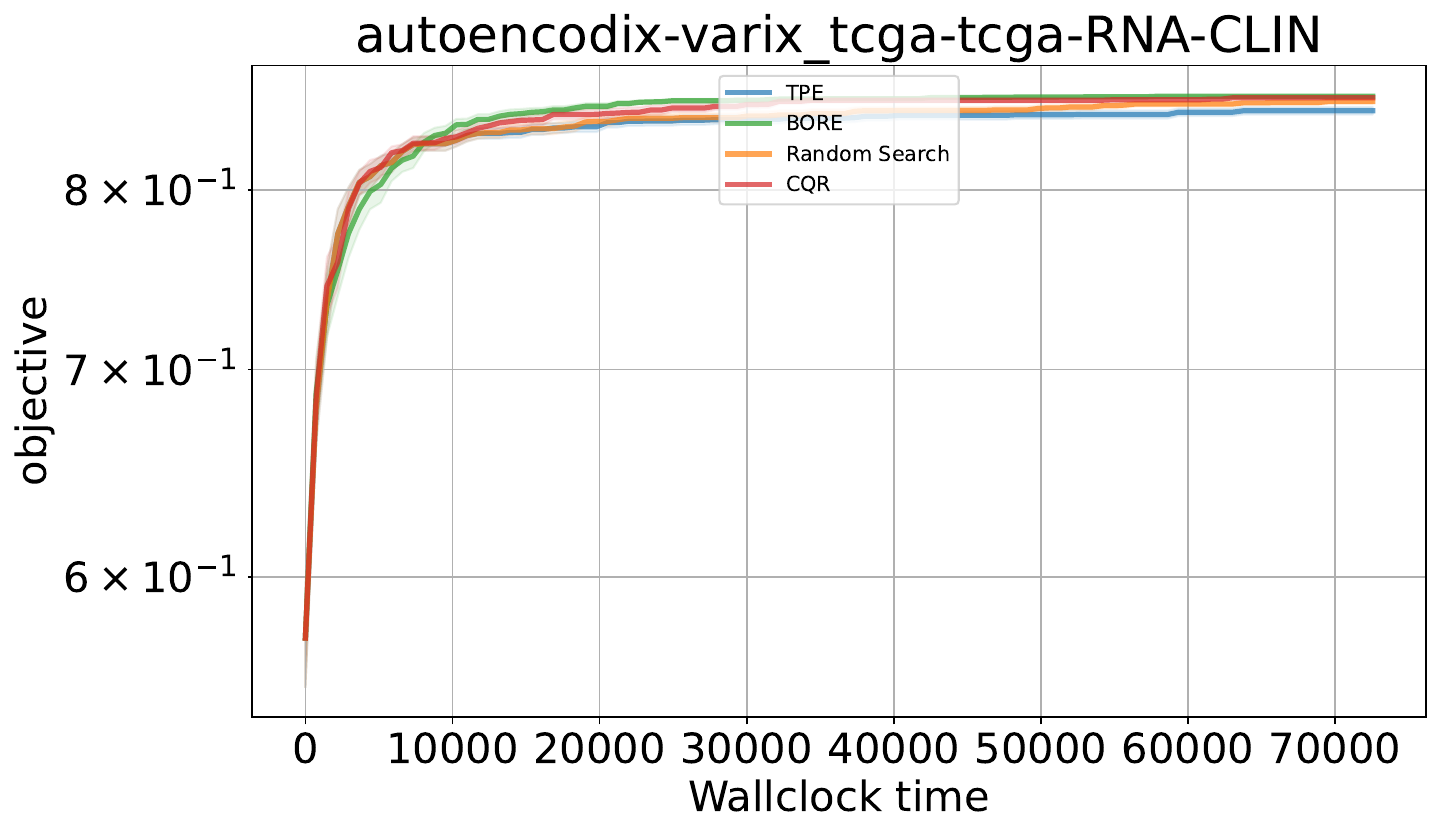} \\
    \midrule
    \textbf{TCGA-RNA-DNA-METH} &  &  \\
    \includegraphics[width=0.32\linewidth]{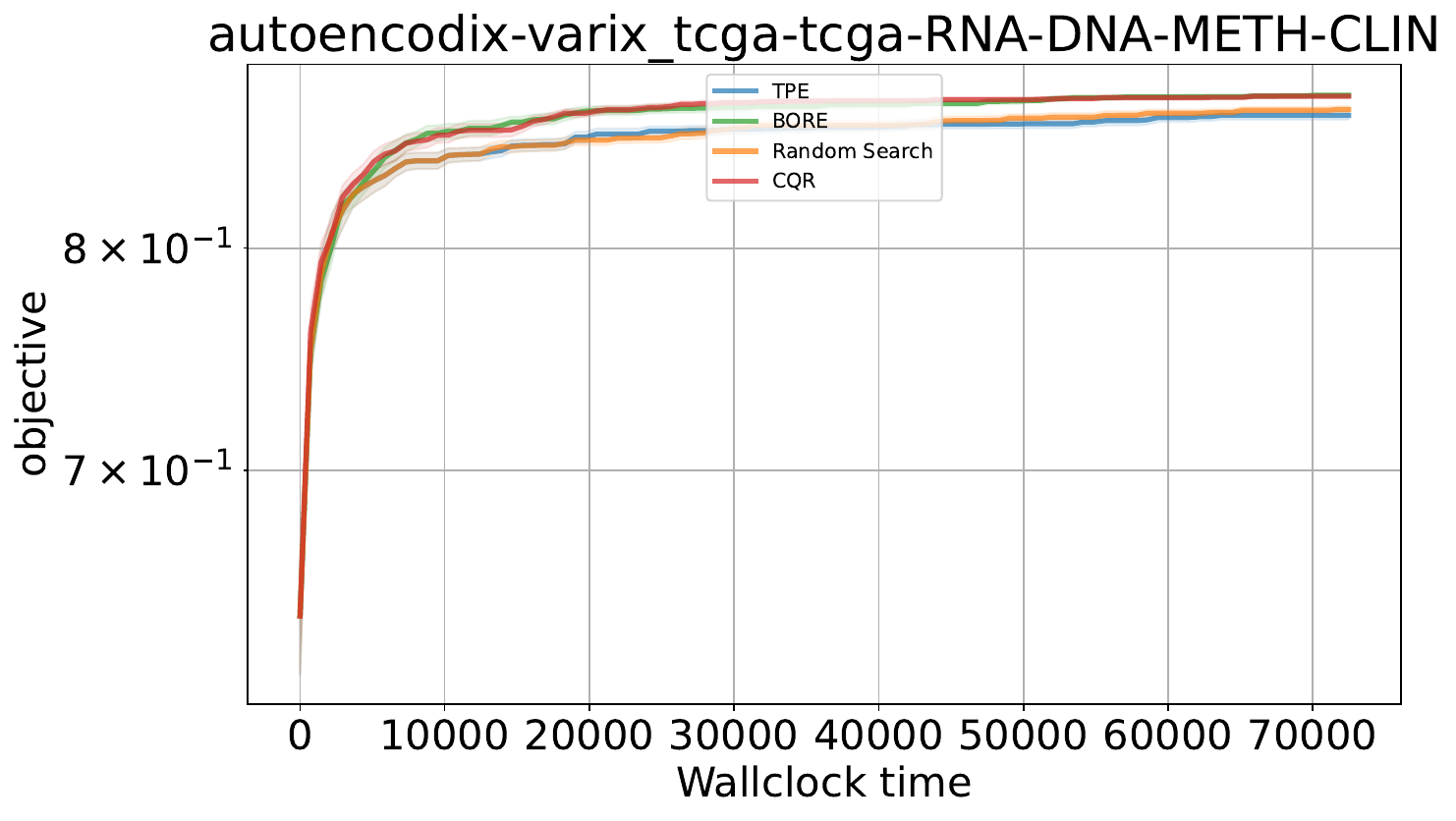} &  &  \\
    \end{tabular}
    \caption{Varix: optimization trajectories maximizing downstream performance}
    \label{fig:optimizers-varix}
\end{figure*}

% ── FIGURE 3: VANILLIX ────────────────────────────────────────────────────────
\begin{figure*}[t]
    \centering
    \setlength{\tabcolsep}{1pt}
    \begin{tabular}{ccc}
    \textbf{SCHC-METH} & \textbf{SCHC-RNA} & \textbf{SCHC-RNA-METH} \\
    \includegraphics[width=0.32\linewidth]{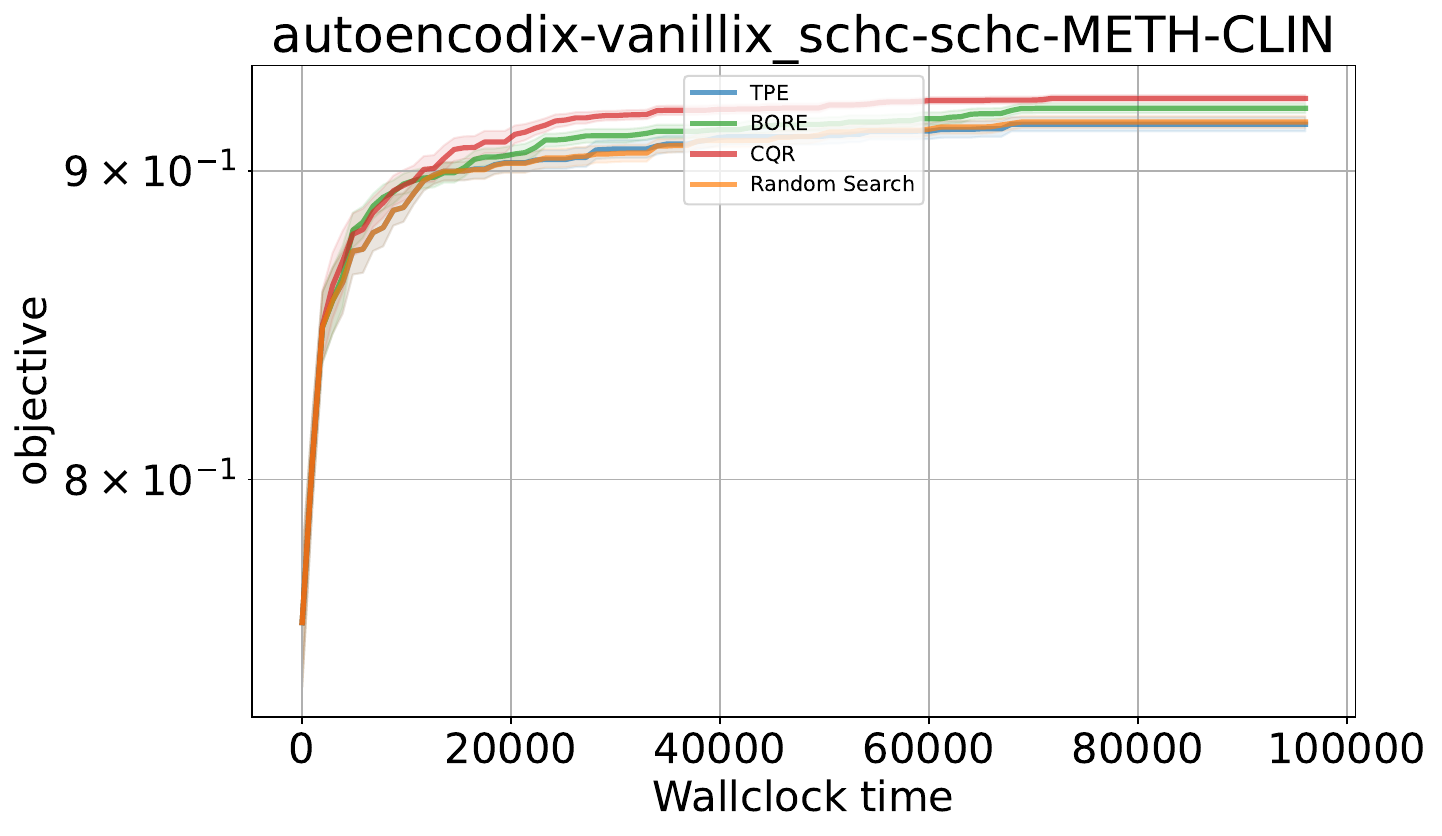} & \includegraphics[width=0.32\linewidth]{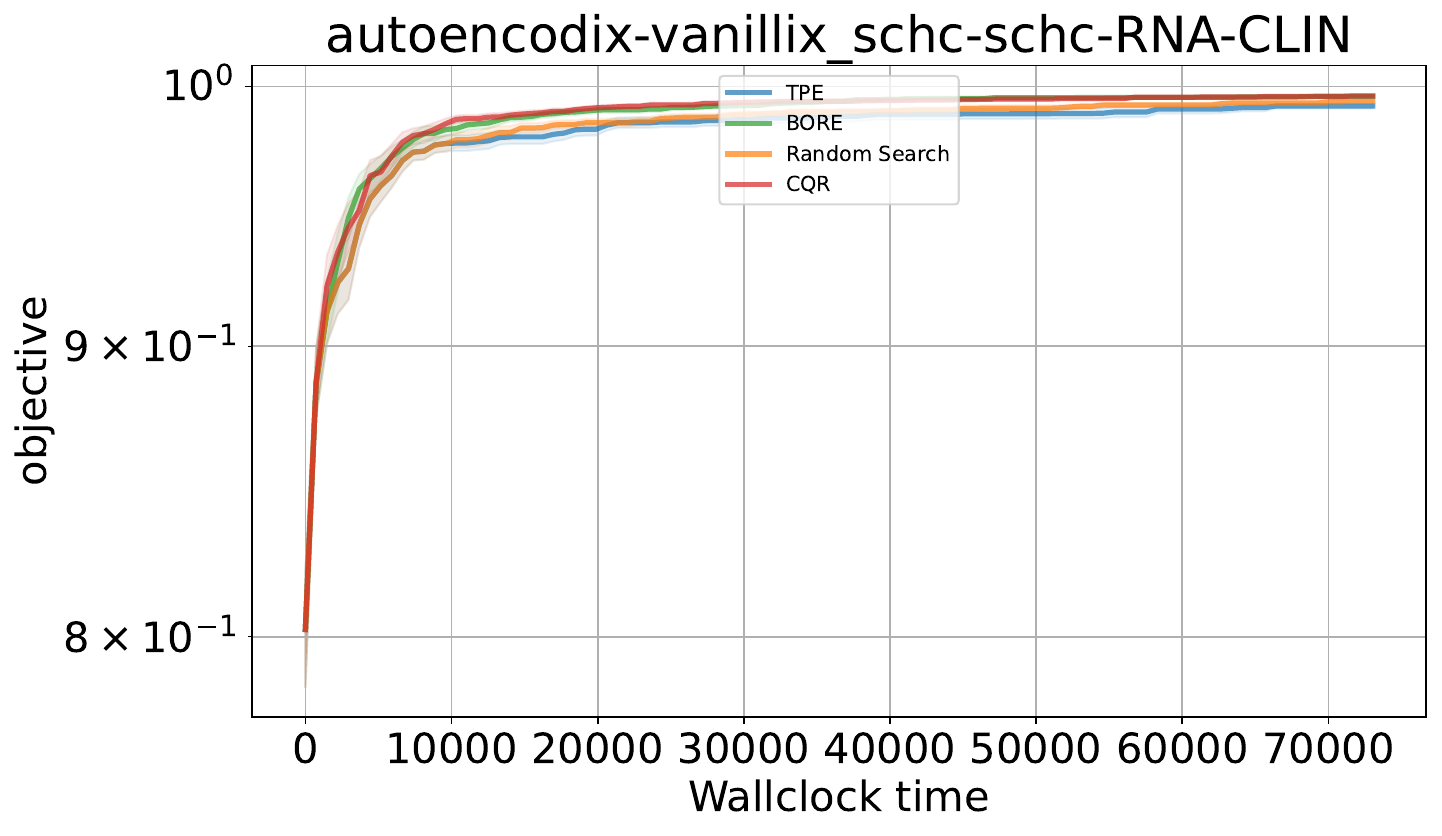} & \includegraphics[width=0.32\linewidth]{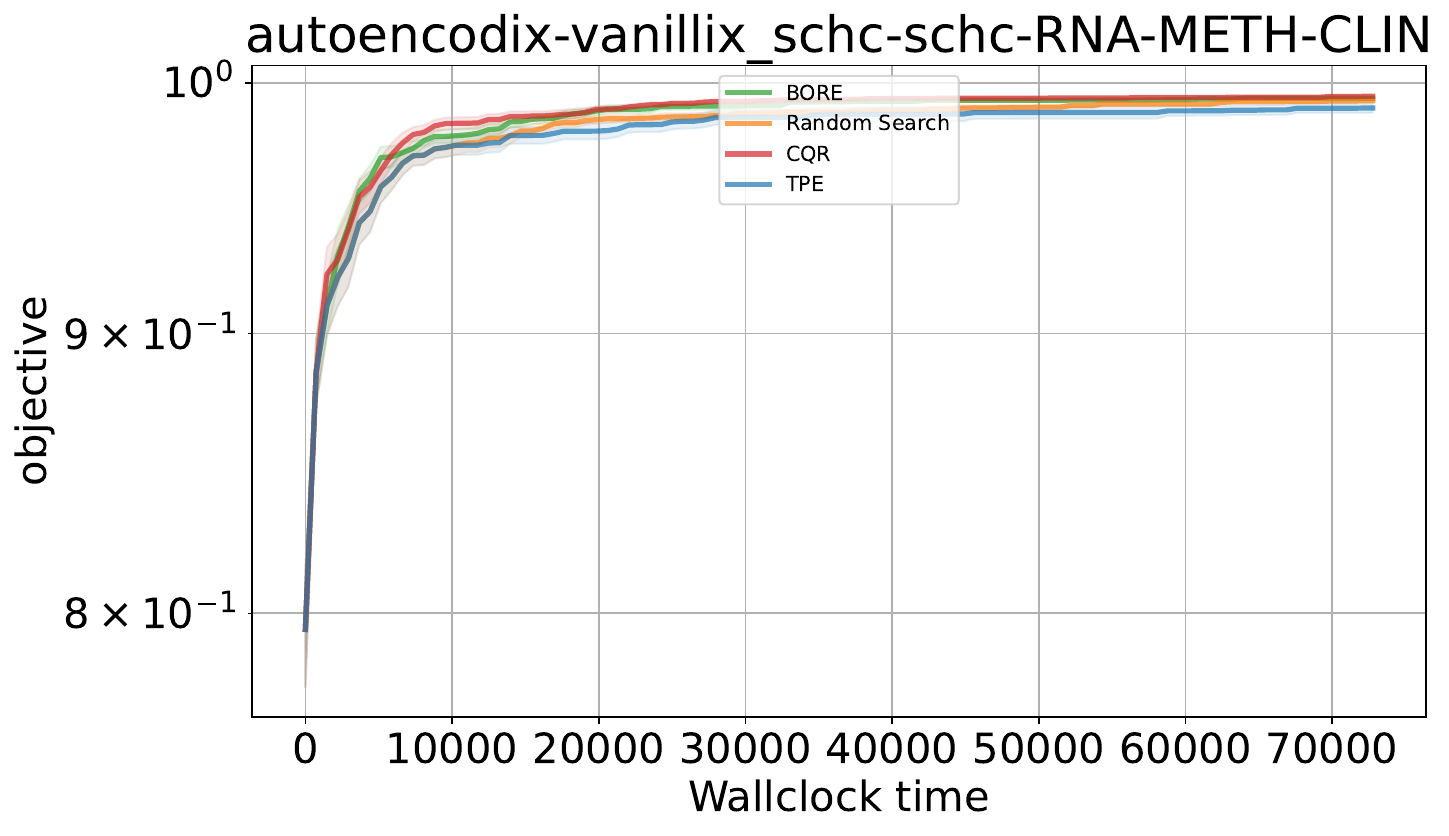} \\
    \midrule
    \textbf{TCGA-DNA} & \textbf{TCGA-METH} & \textbf{TCGA-RNA} \\
    \includegraphics[width=0.32\linewidth]{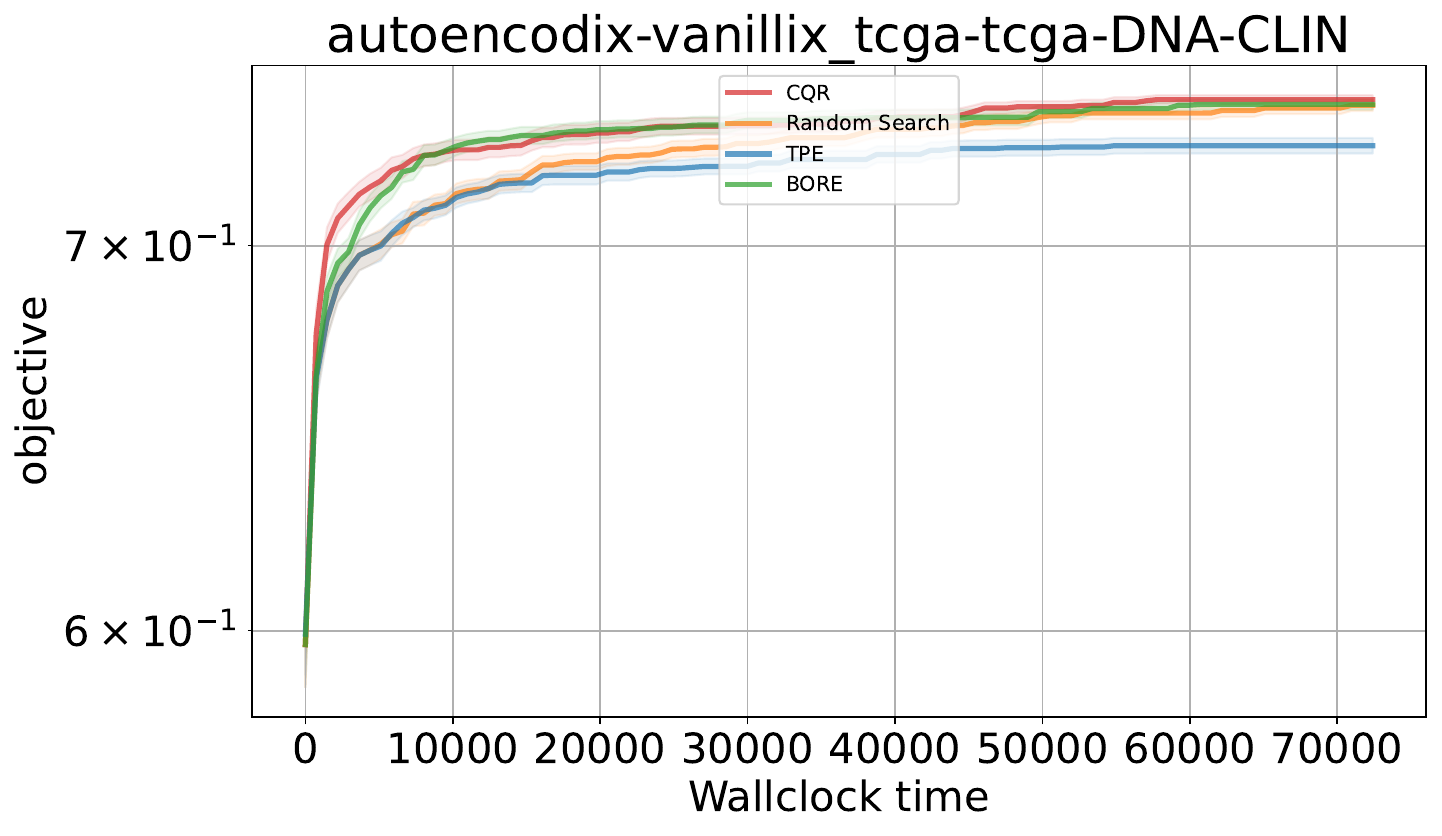} & \includegraphics[width=0.32\linewidth]{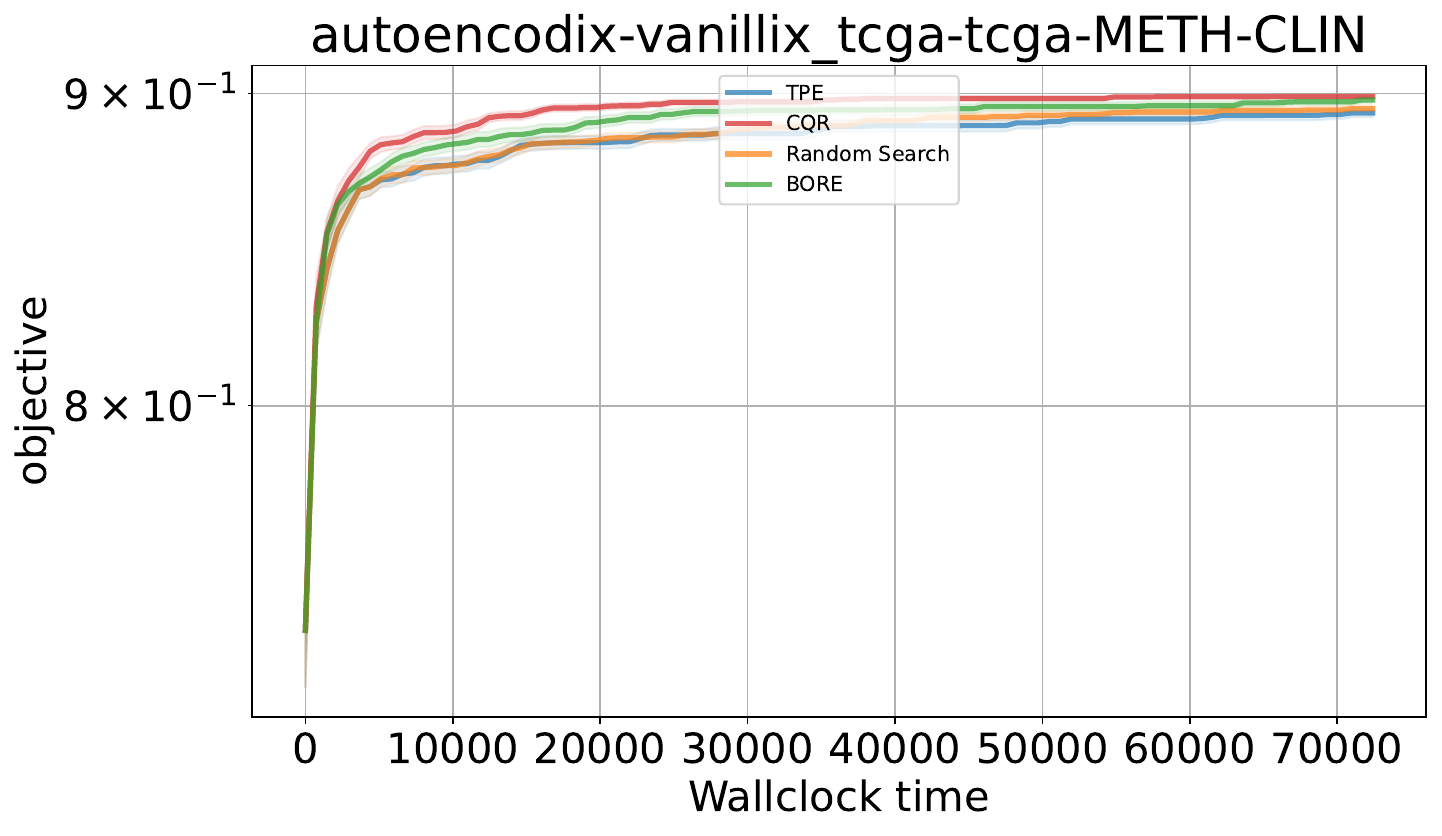} & \includegraphics[width=0.32\linewidth]{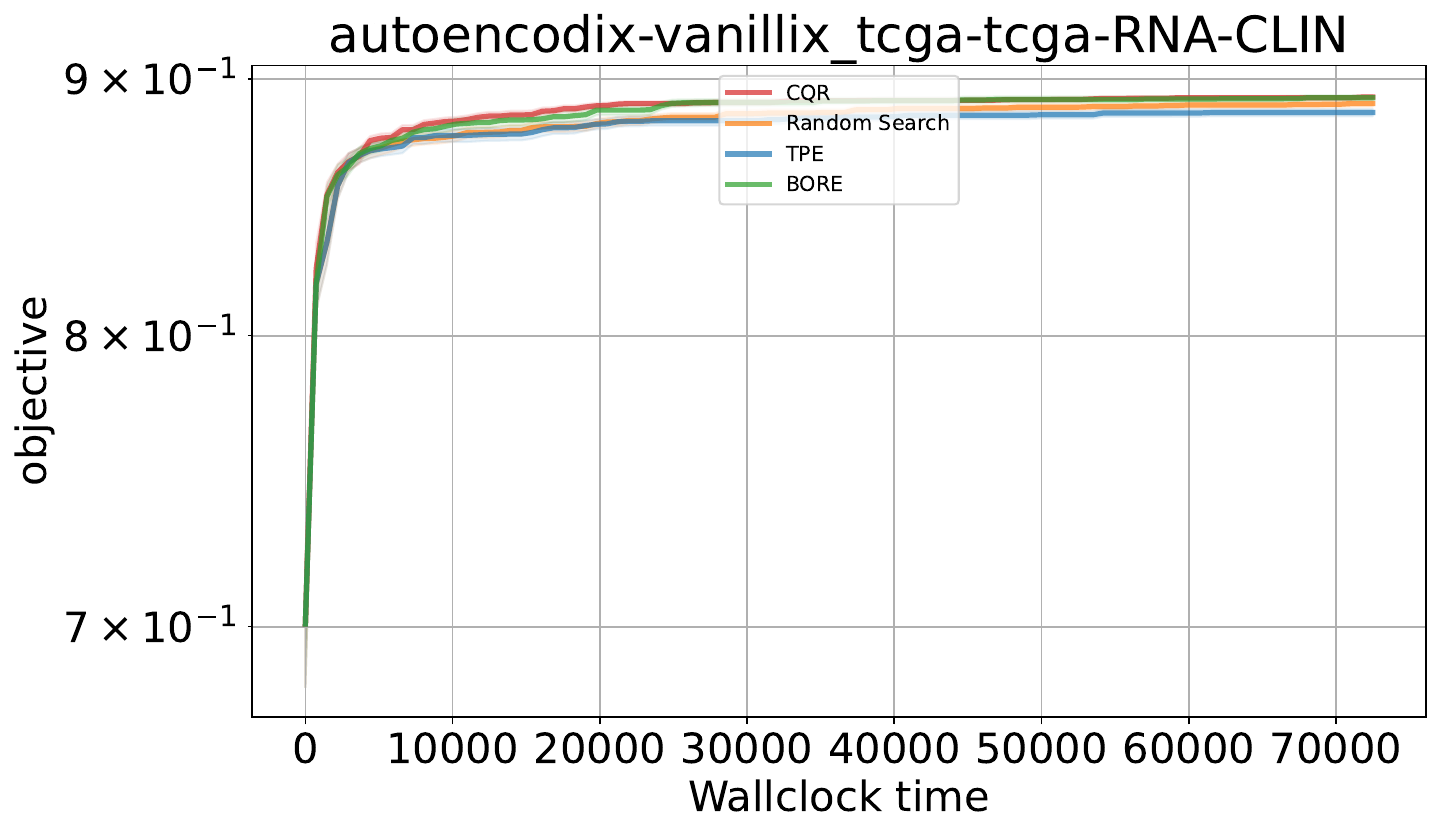} \\
    \midrule
    \textbf{TCGA-RNA-DNA-METH} &  &  \\
    \includegraphics[width=0.32\linewidth]{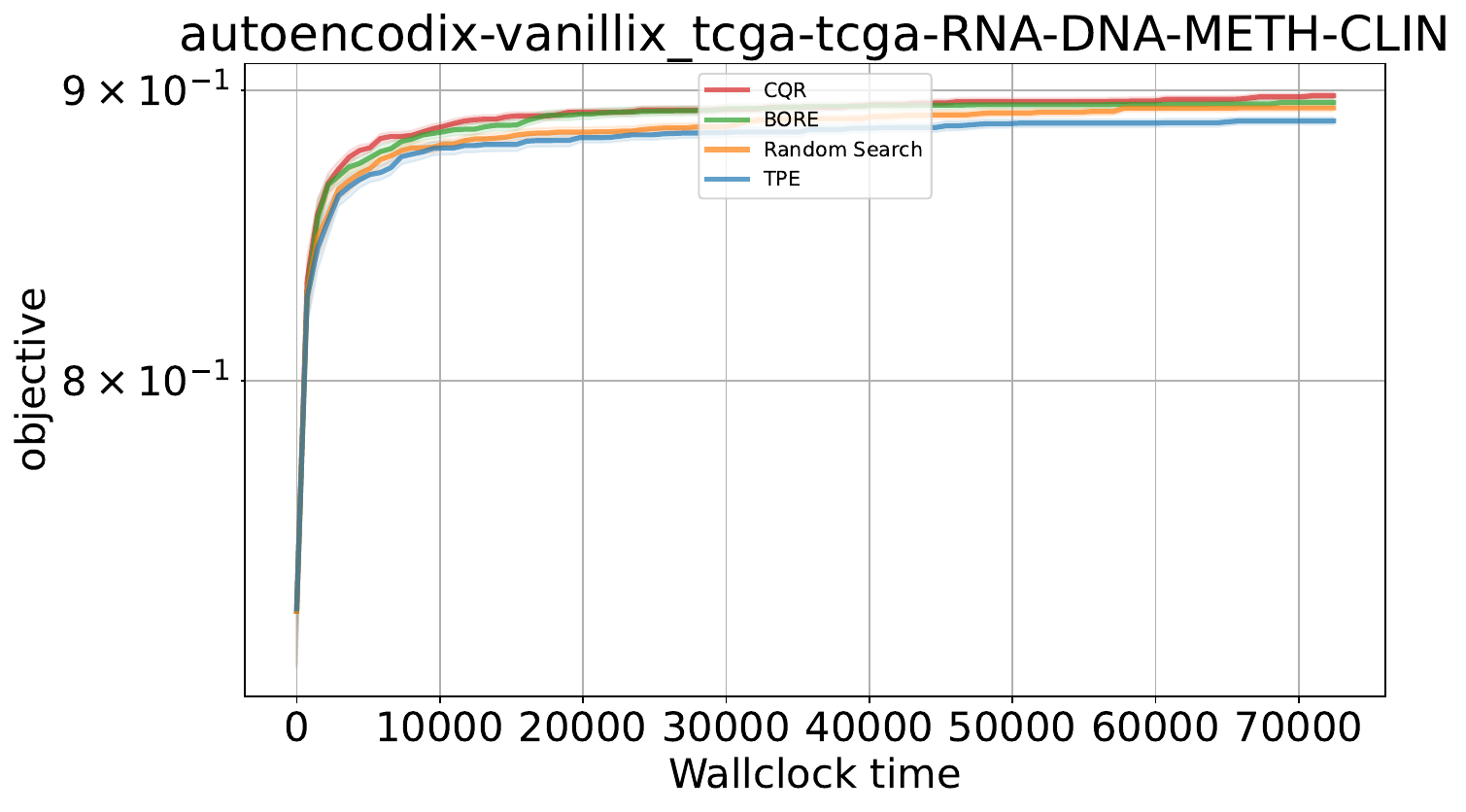} &  &  \\
    \end{tabular}
    \caption{Vanillix: optimization trajectories maximizing downstream performance}
    \label{fig:optimizers-vanillix}
\end{figure*}

% ── FIGURE 4: ONTIX ───────────────────────────────────────────────────────────
\begin{figure*}[t]
    \centering
    \setlength{\tabcolsep}{1pt}
    \begin{tabular}{ccc}
    \textbf{SCHC-METH (chromosome)} & \textbf{SCHC-METH (reactome)} & \textbf{SCHC-RNA (chromosome)} \\
    \includegraphics[width=0.32\linewidth]{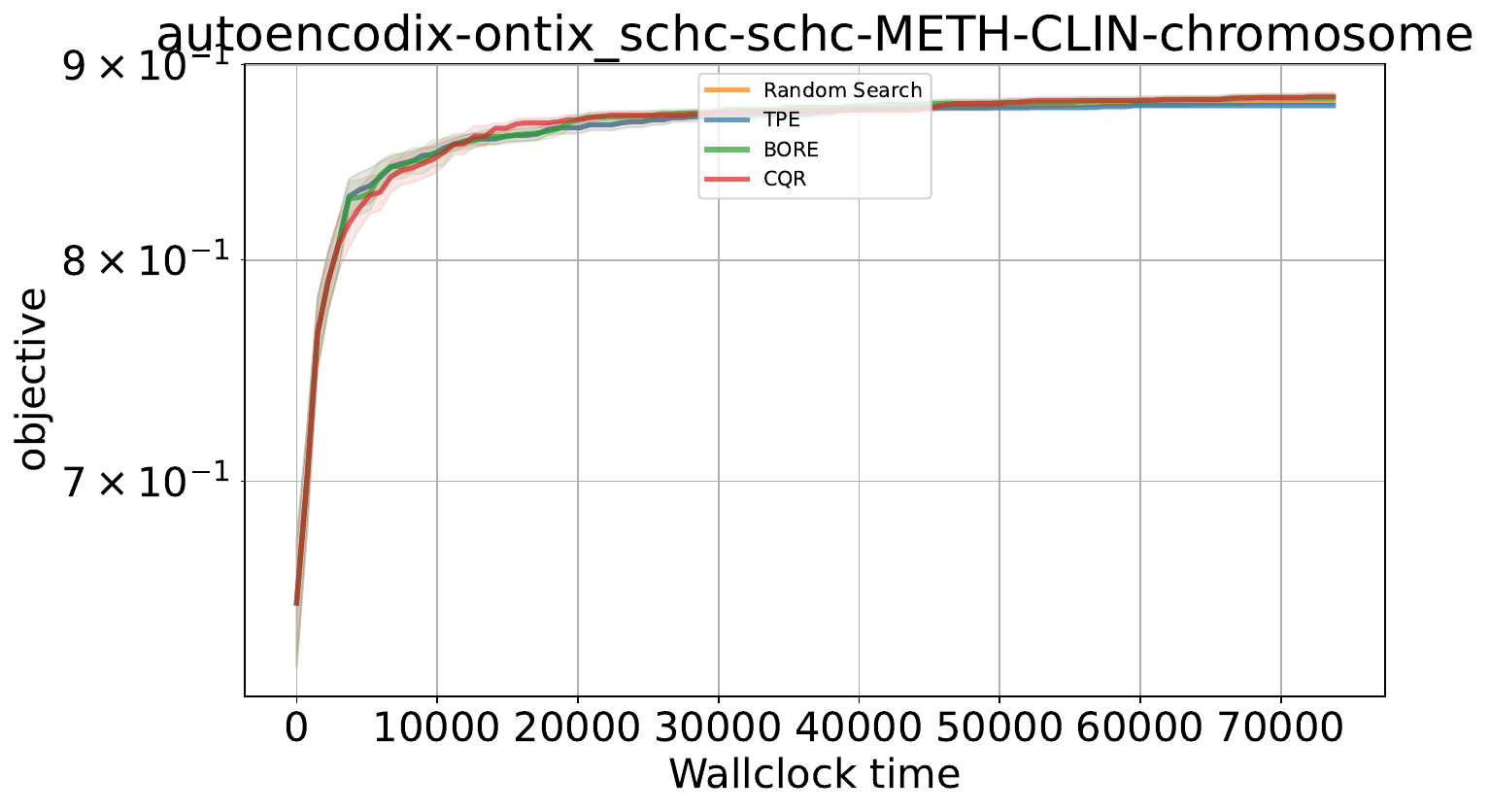} & \includegraphics[width=0.32\linewidth]{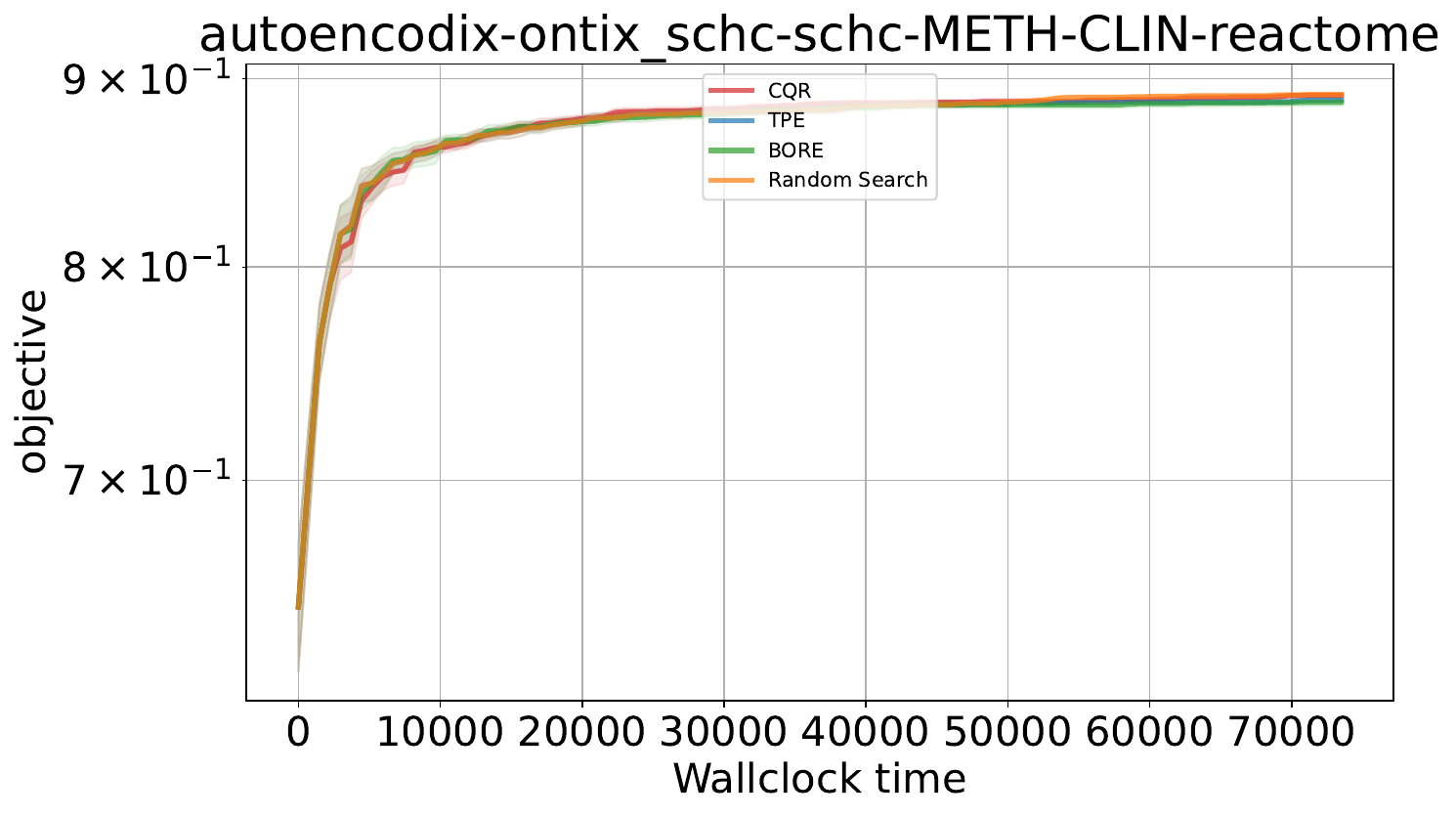} & \includegraphics[width=0.32\linewidth]{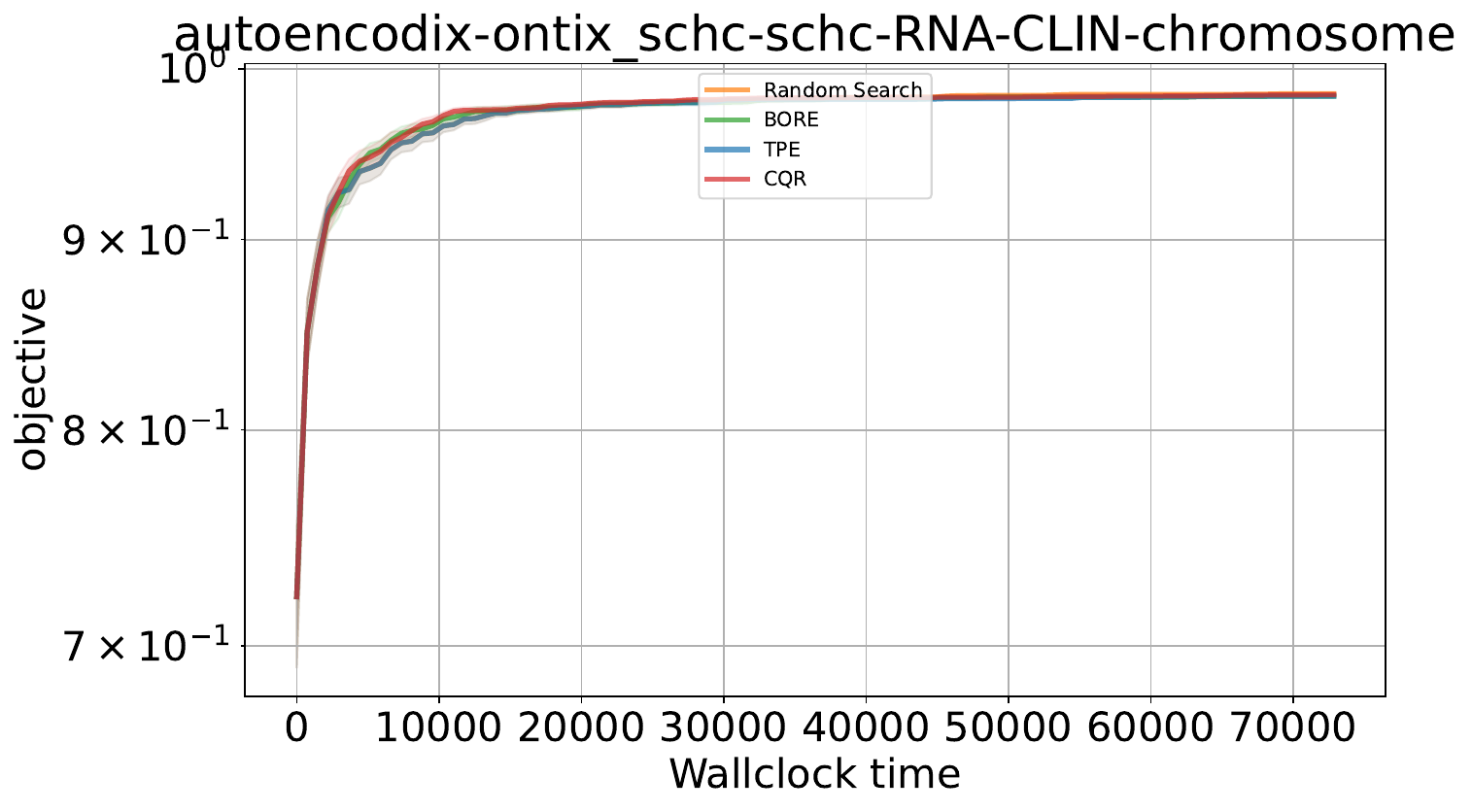} \\
    \midrule
    \textbf{SCHC-RNA (reactome)} & \textbf{SCHC-RNA-METH (chromosome)} & \textbf{SCHC-RNA-METH (reactome)} \\
    \includegraphics[width=0.32\linewidth]{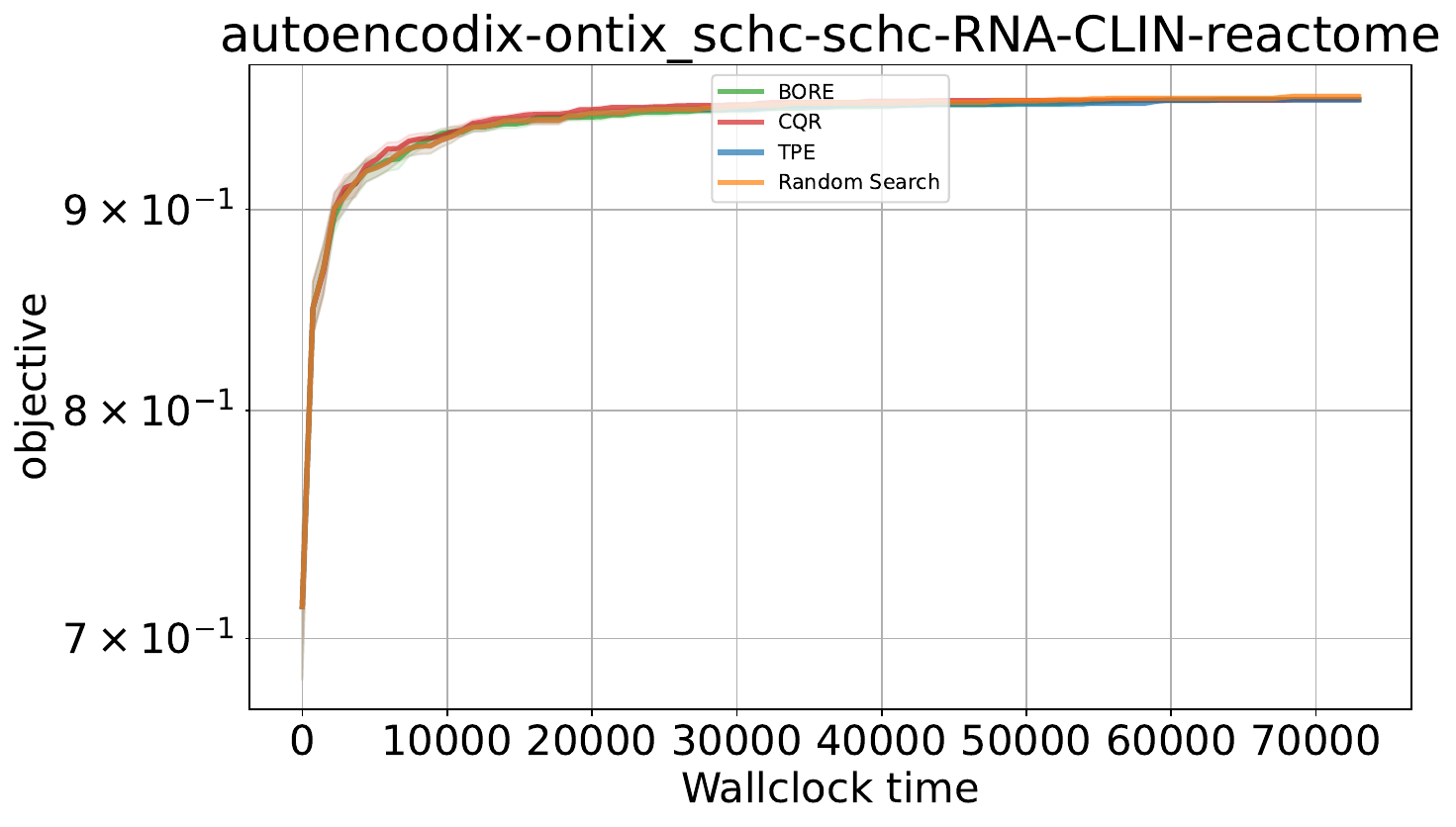} & \includegraphics[width=0.32\linewidth]{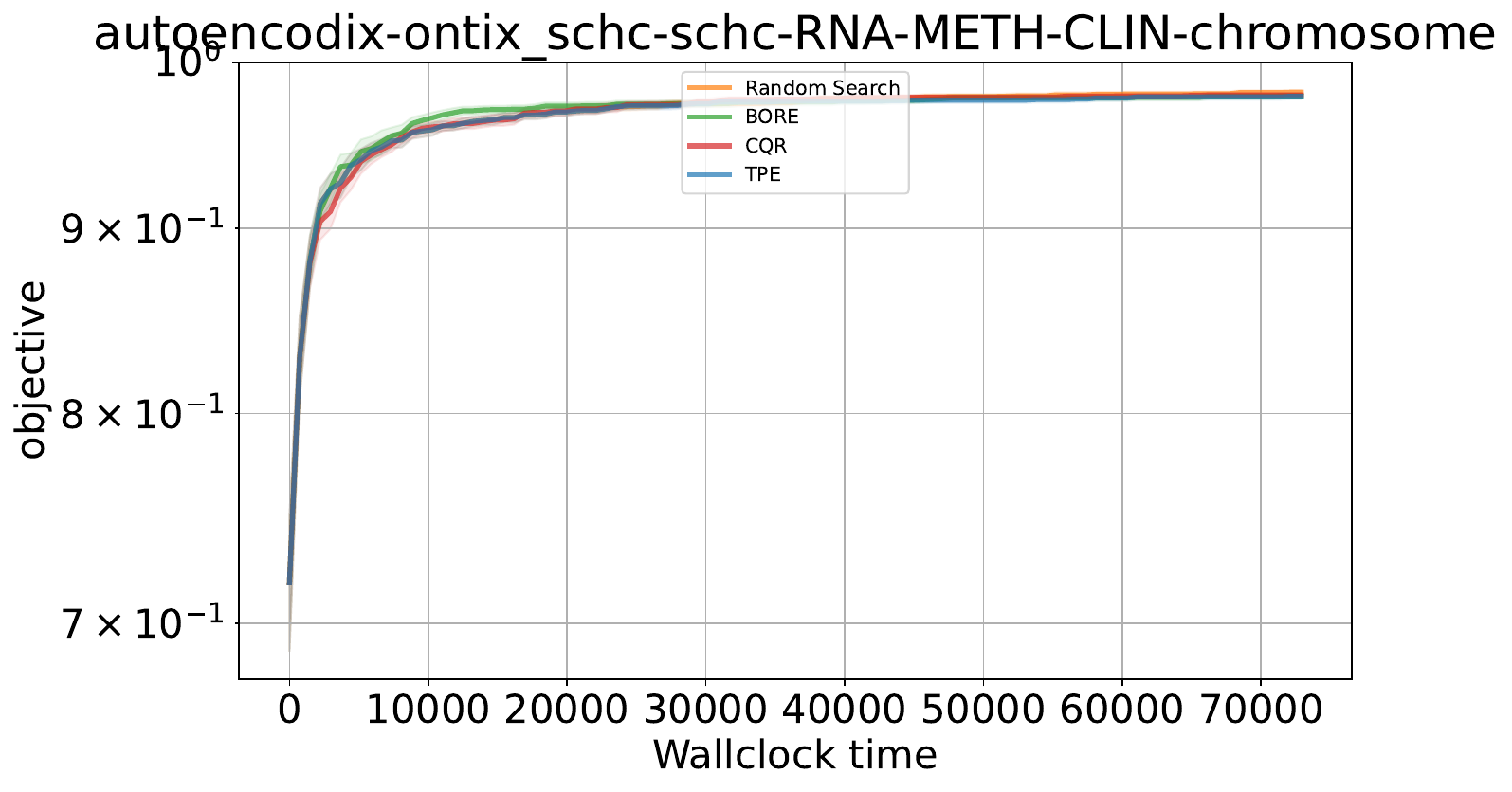} & \includegraphics[width=0.32\linewidth]{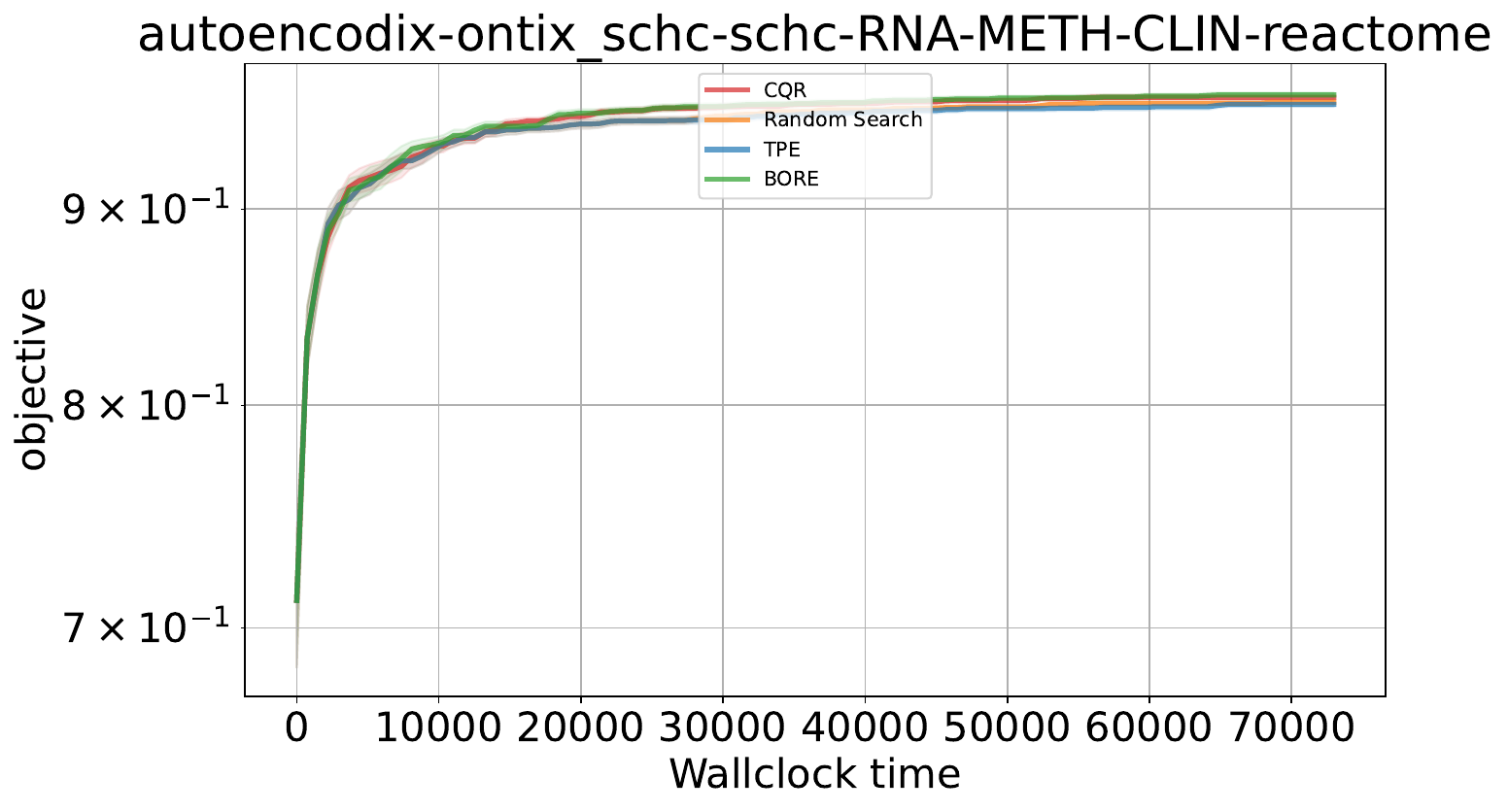} \\
    \midrule
    \textbf{TCGA-DNA (chromosome)} & \textbf{TCGA-DNA (reactome)} & \textbf{TCGA-METH (chromosome)} \\
    \includegraphics[width=0.32\linewidth]{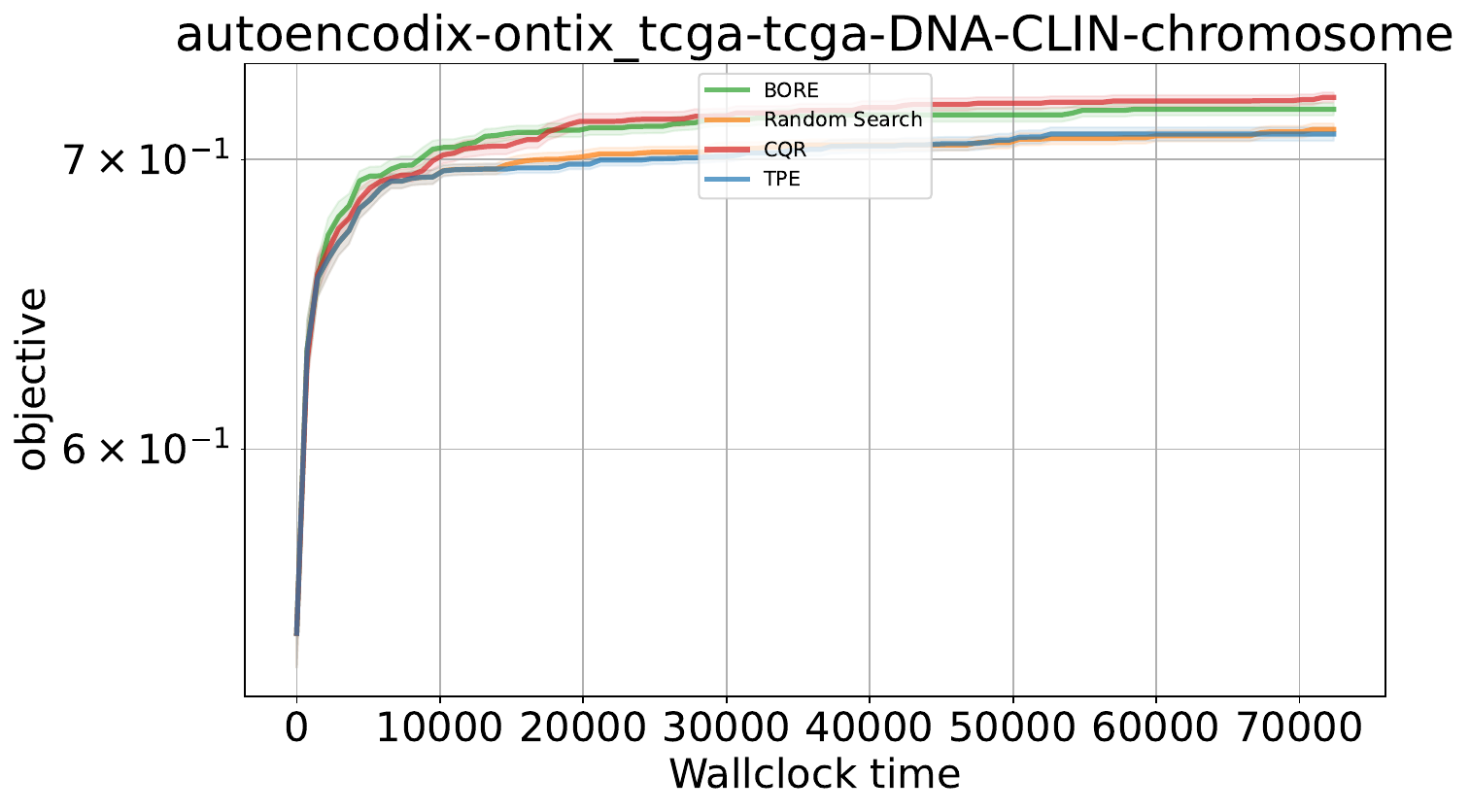} & \includegraphics[width=0.32\linewidth]{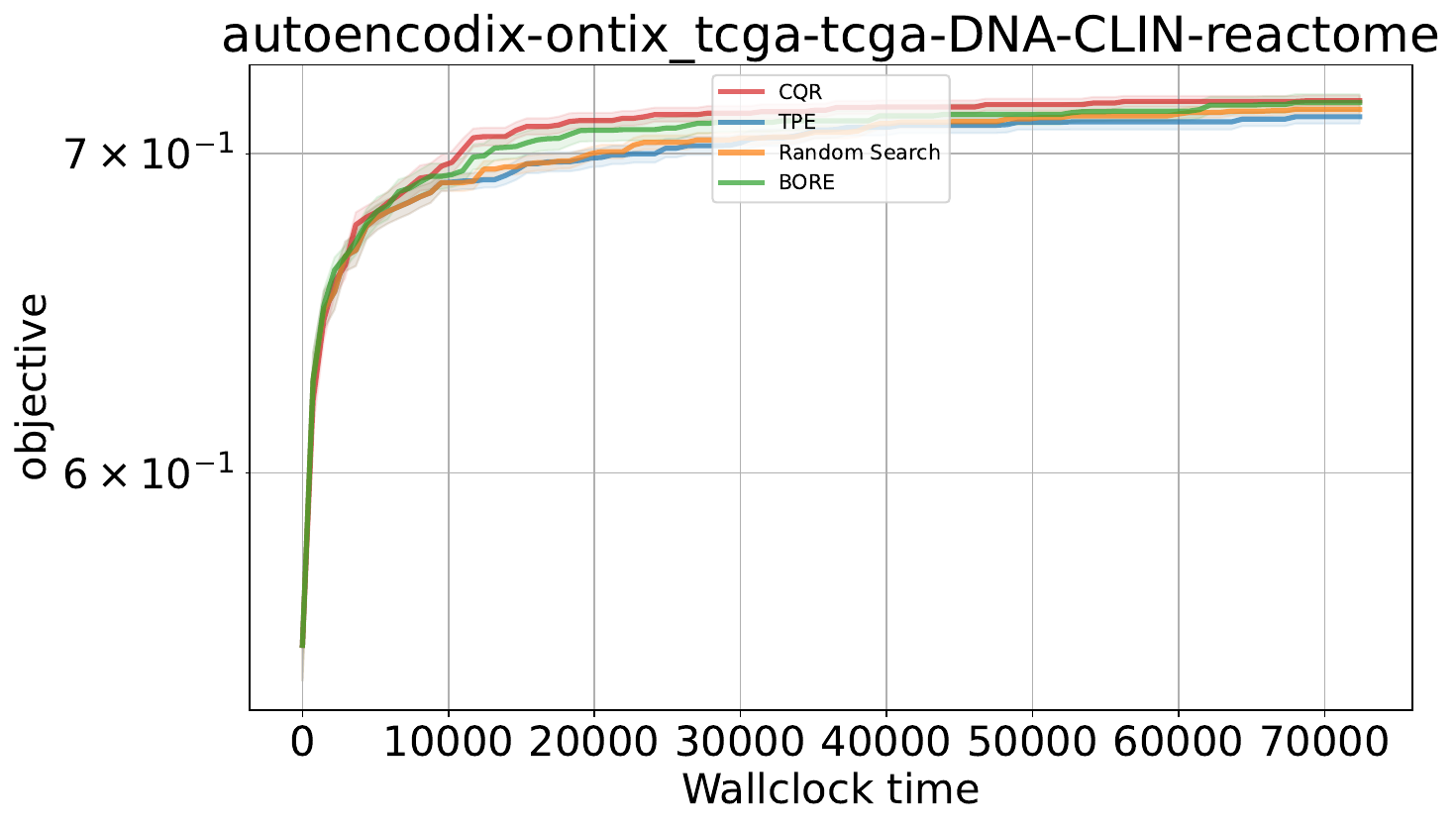} & \includegraphics[width=0.32\linewidth]{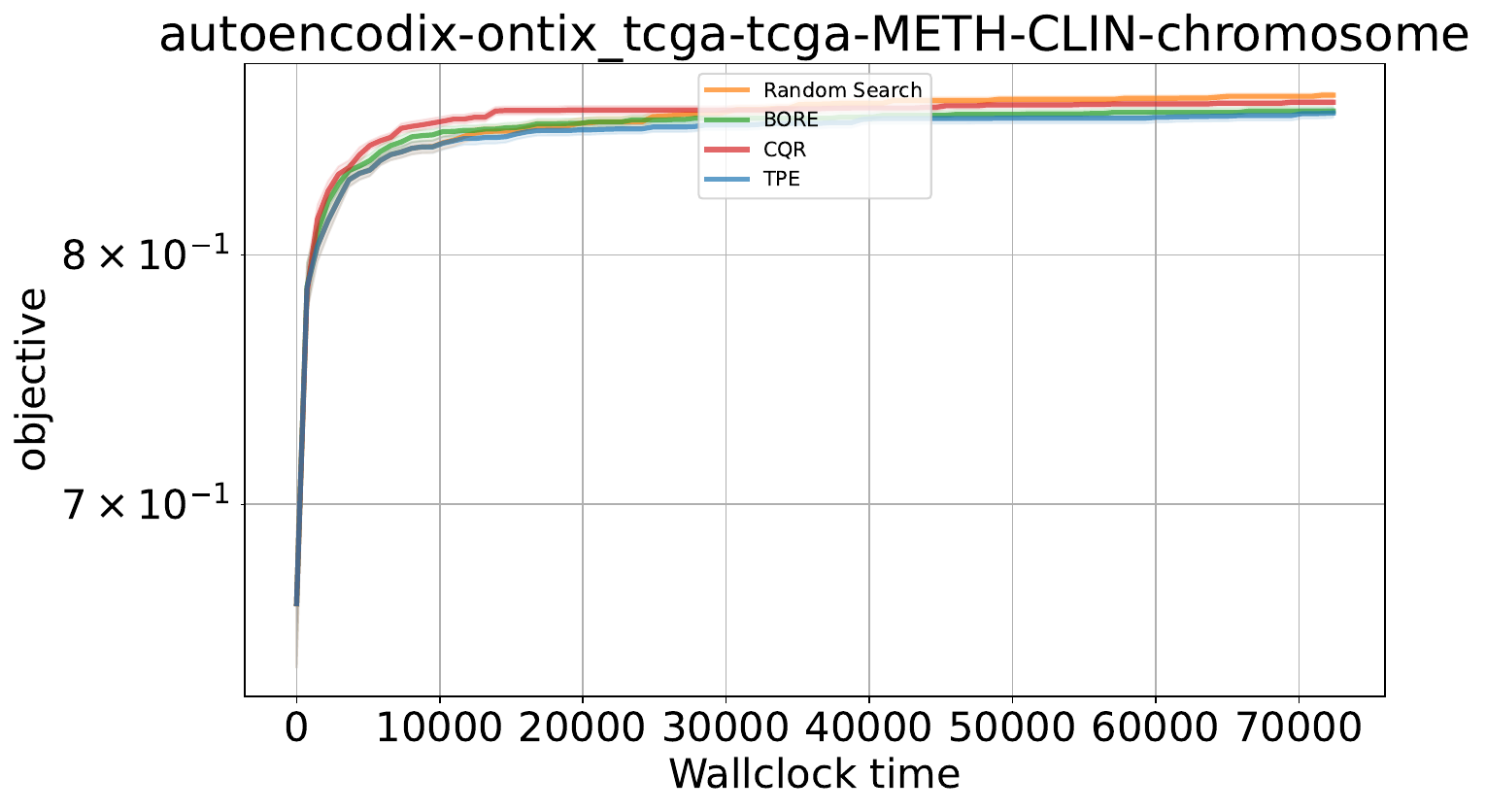} \\
    \midrule
    \textbf{TCGA-METH (reactome)} & \textbf{TCGA-RNA (chromosome)} & \textbf{TCGA-RNA (reactome)} \\
    \includegraphics[width=0.32\linewidth]{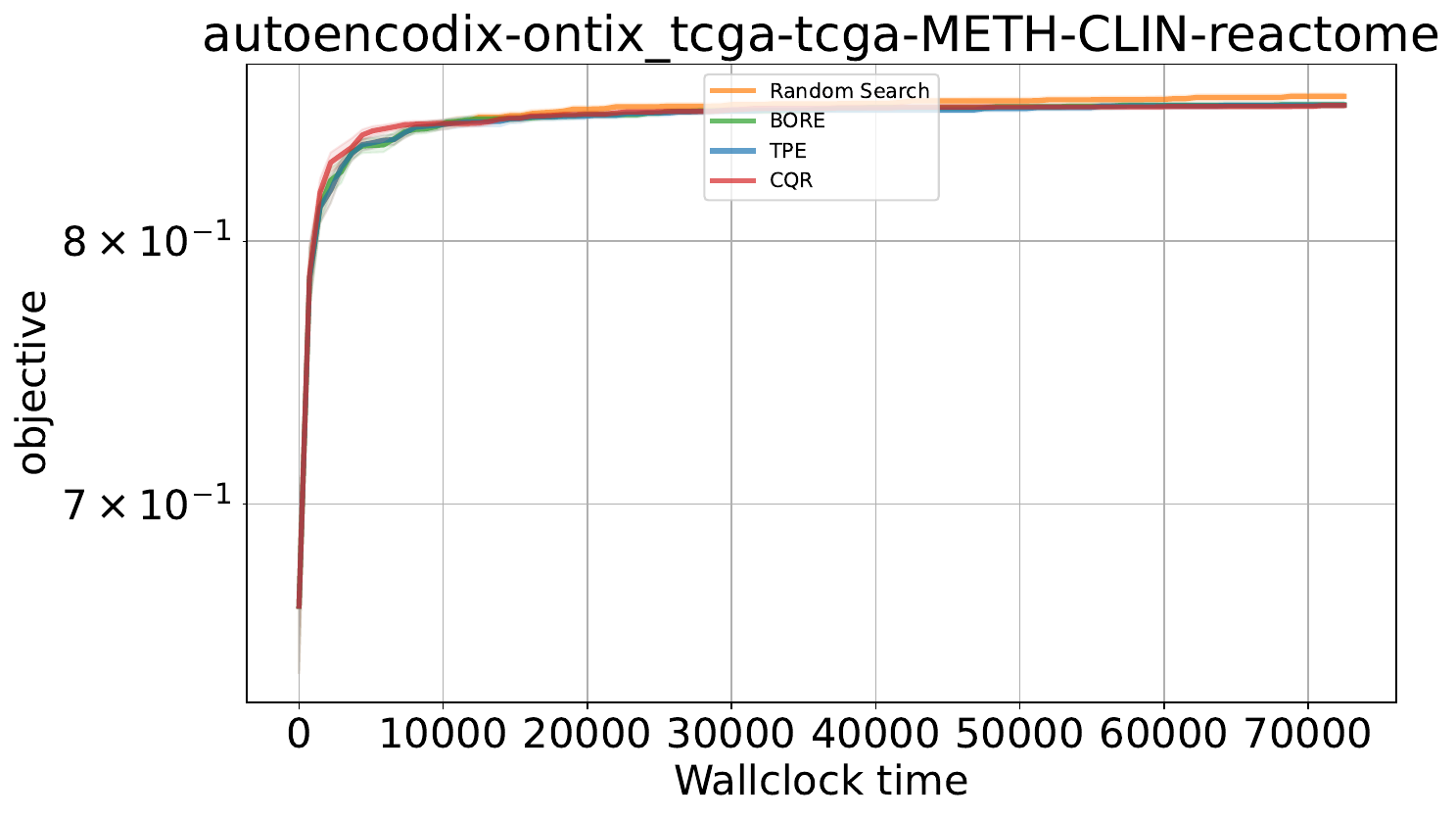} & \includegraphics[width=0.32\linewidth]{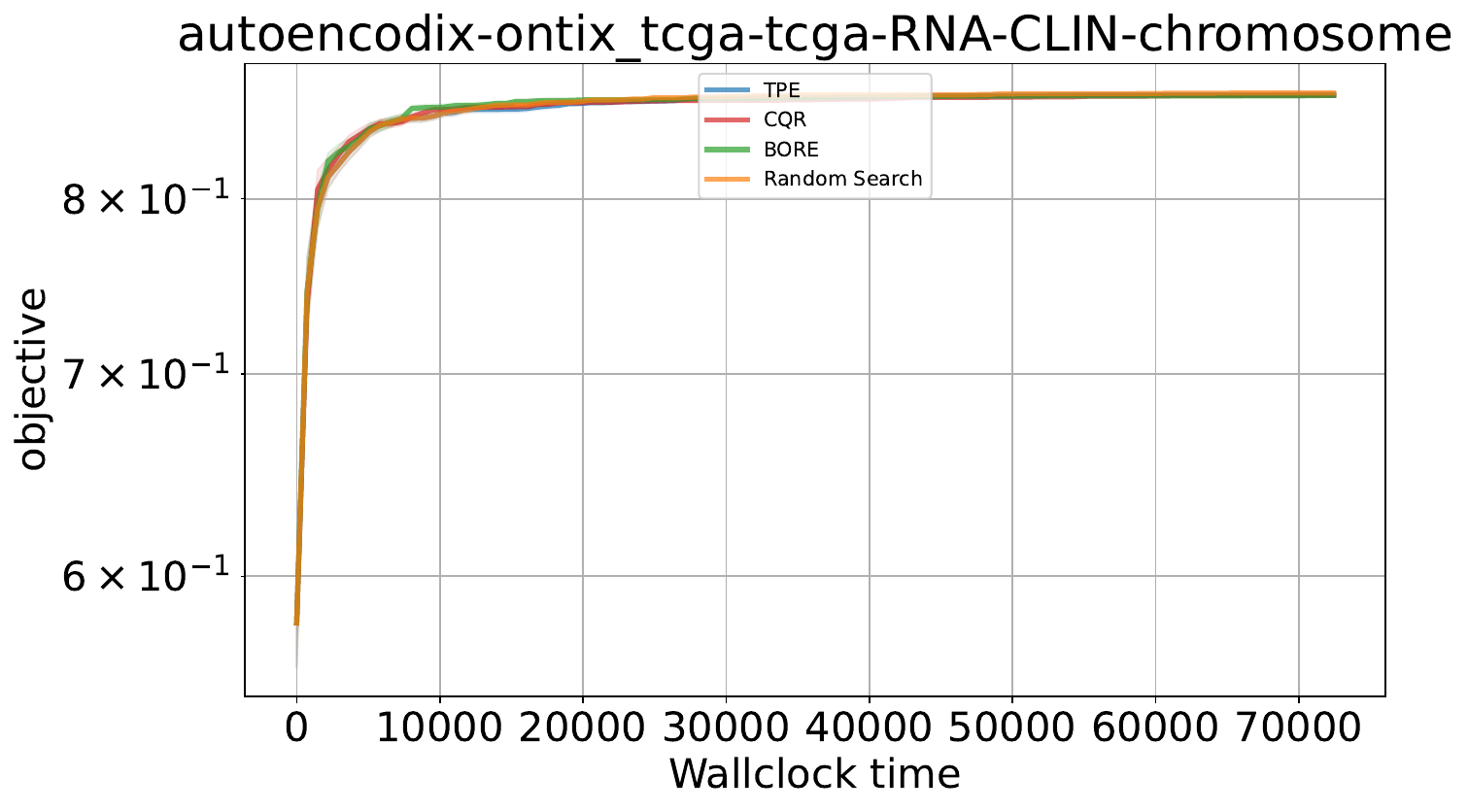} & \includegraphics[width=0.32\linewidth]{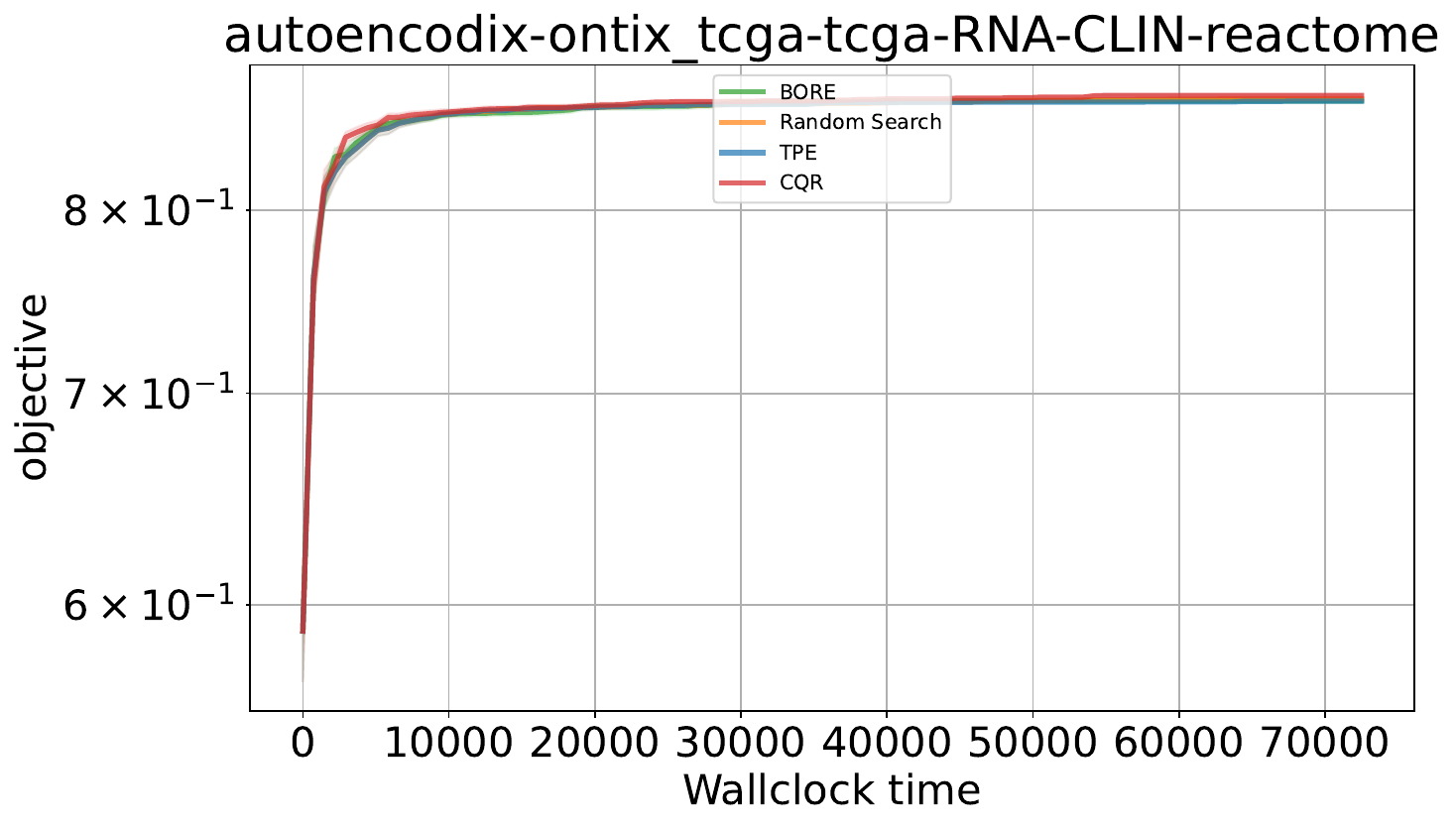} \\
    \midrule
    \textbf{TCGA-RNA-DNA-METH (chromosome)} & \textbf{TCGA-RNA-DNA-METH (reactome)} &  \\
    \includegraphics[width=0.32\linewidth]{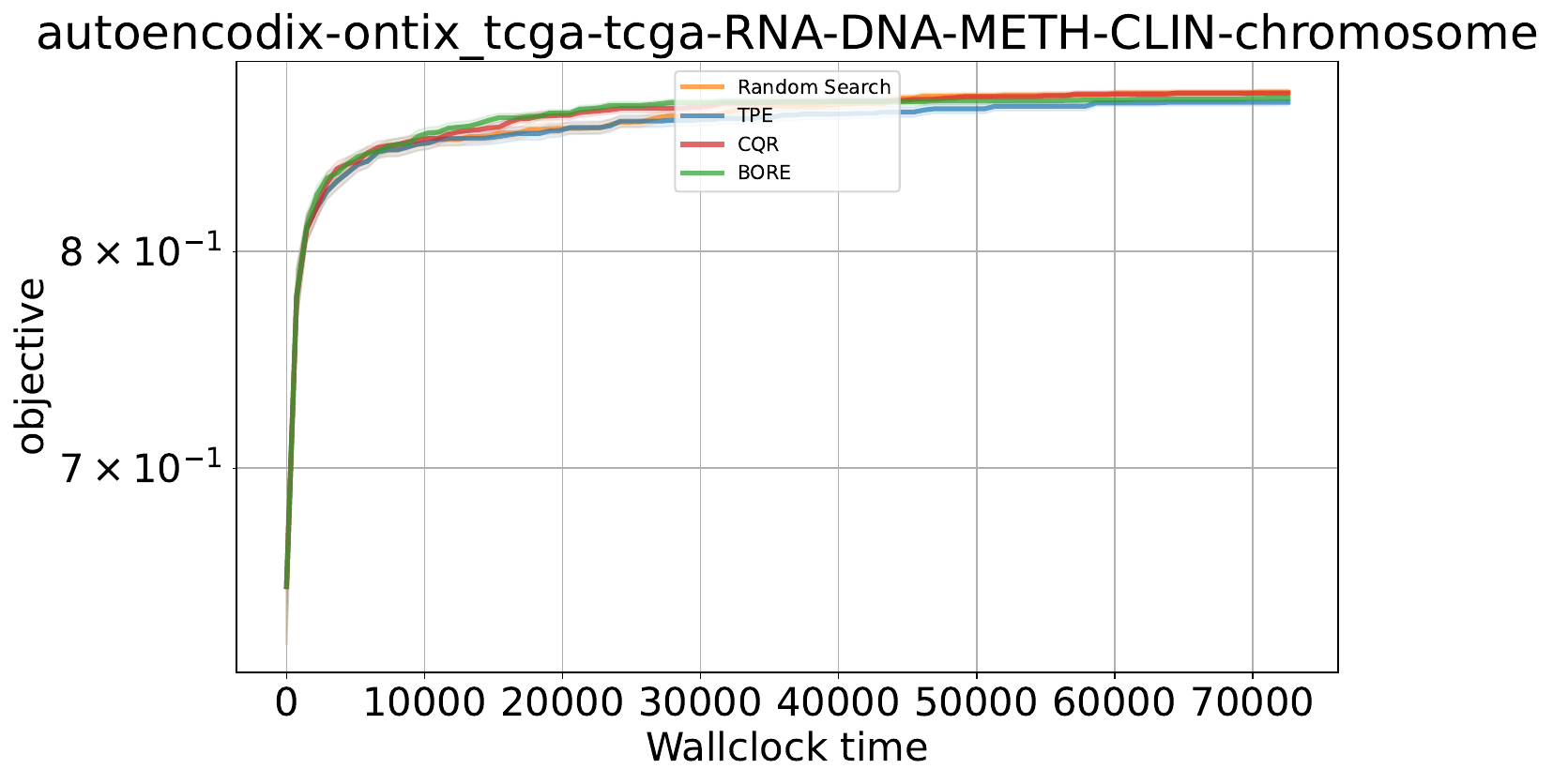} & \includegraphics[width=0.32\linewidth]{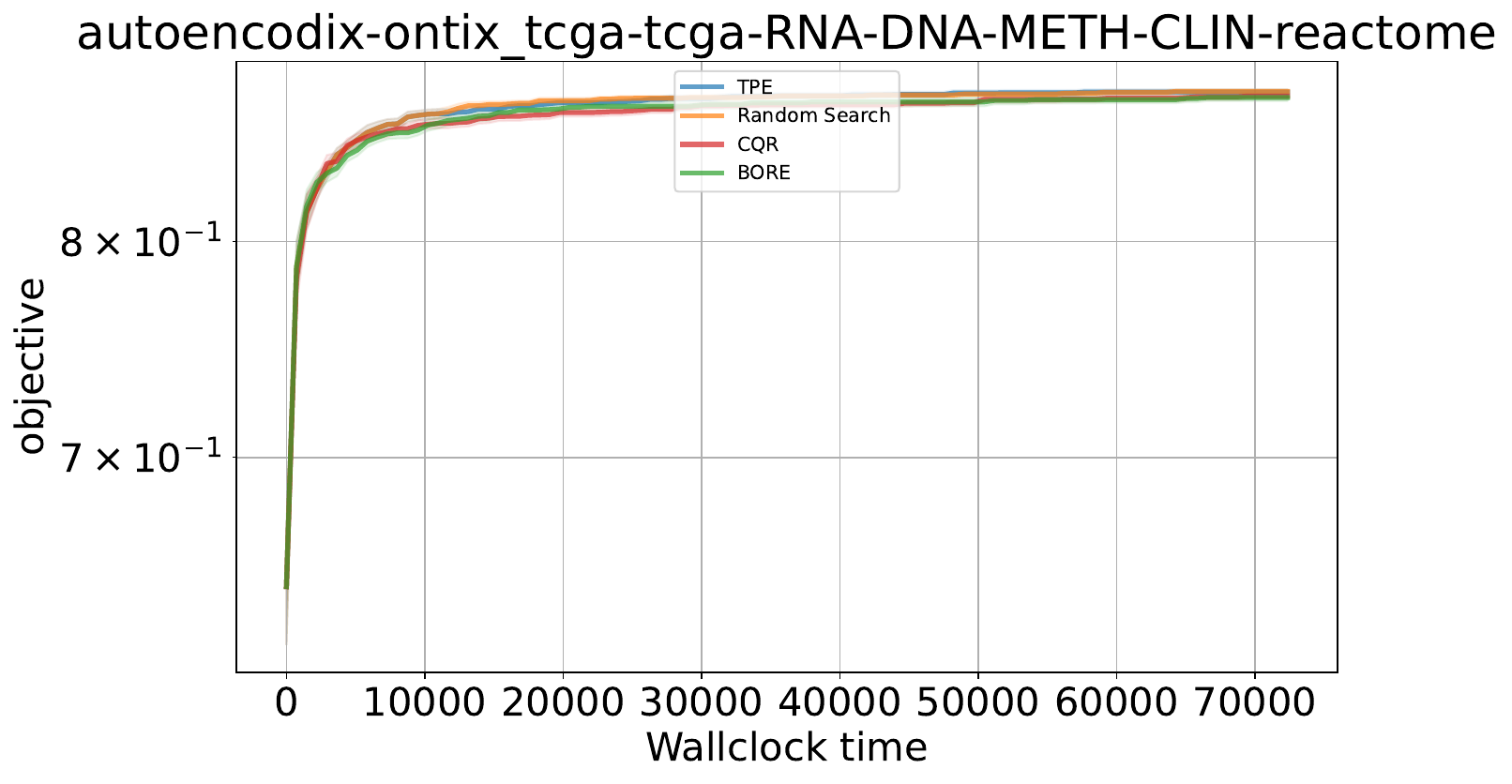} &  \\
    \end{tabular}
    \caption{Ontix: optimization trajectories maximizing downstream performance}
    \label{fig:optimizers-ontix}
\end{figure*}


\begin{thebibliography}{}

\bibitem[Bergstra et~al., 2011]{bergstra2011algorithms}
Bergstra, J., Bardenet, R., Bengio, Y., and K{\'e}gl, B. (2011).
\newblock Algorithms for hyper-parameter optimization.
\newblock {\em Advances in neural information processing systems}, 24.

\bibitem[Bergstra and Bengio, 2012]{bergstra2012random}
Bergstra, J. and Bengio, Y. (2012).
\newblock Random search for hyper-parameter optimization.
\newblock {\em Journal of machine learning research}, 13(2).

\bibitem[Bischl et~al., 2023]{bischl2023hyperparameter}
Bischl, B., Binder, M., Lang, M., Pielok, T., Richter, J., Coors, S., Thomas,
  J., Ullmann, T., Becker, M., Boulesteix, A.-L., et~al. (2023).
\newblock Hyperparameter optimization: Foundations, algorithms, best practices,
  and open challenges.
\newblock {\em Wiley Interdisciplinary Reviews: Data Mining and Knowledge
  Discovery}, 13(2):e1484.

\bibitem[Chen et~al., 2018]{chen2018isolating}
Chen, R.~T., Li, X., Grosse, R.~B., and Duvenaud, D.~K. (2018).
\newblock Isolating sources of disentanglement in variational autoencoders.
\newblock {\em Advances in neural information processing systems}, 31.

\bibitem[Doncevic and Herrmann, 2023]{doncevic2023biologically}
Doncevic, D. and Herrmann, C. (2023).
\newblock Biologically informed variational autoencoders allow predictive
  modeling of genetic and drug-induced perturbations.
\newblock {\em Bioinformatics}, 39(6):btad387.

\bibitem[Dong and Yang, 2020]{dong2020bench}
Dong, X. and Yang, Y. (2020).
\newblock Nas-bench-201: Extending the scope of reproducible neural
  architecture search.
\newblock {\em arXiv preprint arXiv:2001.00326}.

\bibitem[Eggensperger et~al., 2015]{eggensperger2015efficient}
Eggensperger, K., Hutter, F., Hoos, H., and Leyton-Brown, K. (2015).
\newblock Efficient benchmarking of hyperparameter optimizers via surrogates.
\newblock In {\em Proceedings of the AAAI conference on artificial
  intelligence}, volume~29.

\bibitem[Eraslan et~al., 2019]{eraslan2019single}
Eraslan, G., Simon, L.~M., Mircea, M., Mueller, N.~S., and Theis, F.~J. (2019).
\newblock Single-cell rna-seq denoising using a deep count autoencoder.
\newblock {\em Nature communications}, 10(1):390.

\bibitem[Ewald, 2025]{acx2025rawdata}
Ewald, J. (2025).
\newblock Autoencodix raw data for reproducibility.
\newblock \url{https://doi.org/10.5281/zenodo.15518831}.

\bibitem[Falkner et~al., 2018]{falkner2018bohb}
Falkner, S., Klein, A., and Hutter, F. (2018).
\newblock Bohb: Robust and efficient hyperparameter optimization at scale.
\newblock In {\em International conference on machine learning}, pages
  1437--1446. PMLR.

\bibitem[Feurer et~al., 2015]{feurer2015initializing}
Feurer, M., Springenberg, J., and Hutter, F. (2015).
\newblock Initializing bayesian hyperparameter optimization via meta-learning.
\newblock In {\em Proceedings of the AAAI conference on artificial
  intelligence}, volume~29.

\bibitem[Fisher et~al., 2019]{fisher2019all}
Fisher, A., Rudin, C., and Dominici, F. (2019).
\newblock All models are wrong, but many are useful: Learning a variable's
  importance by studying an entire class of prediction models simultaneously.
\newblock {\em Journal of Machine Learning Research}, 20(177):1--81.

\bibitem[Franceschi et~al., 2025]{franceschi-arxiv25}
Franceschi, L., Donini, M., Perrone, V., Klein, A., Archambeau, C., Seeger, M.,
  Pontil, M., and Frasconi, P. (2025).
\newblock Hyperparameter optimization in machine learning.
\newblock {\em arXiv:2410.22854 [stat.ML]}.

\bibitem[Garnett, 2023]{garnett_bayesoptbook_2023}
Garnett, R. (2023).
\newblock {\em {Bayesian Optimization}}.
\newblock Cambridge University Press.

\bibitem[Higgins et~al., 2017]{higgins-iclr17}
Higgins, I., Matthey, L., Pal, A., Burgess, C., Glorot, X., Botvinick, M.,
  Mohamed, S., and Lerchner, A. (2017).
\newblock beta-{VAE}: Learning basic visual concepts with a constrained
  variational framework.
\newblock In {\em International Conference on Learning Representations
  (ICLR'17)}.

\bibitem[Hu and Greene, 2018]{hu2018parameter}
Hu, Q. and Greene, C.~S. (2018).
\newblock Parameter tuning is a key part of dimensionality reduction via deep
  variational autoencoders for single cell rna transcriptomics.
\newblock In {\em BIOCOMPUTING 2019: proceedings of the Pacific symposium},
  pages 362--373. World Scientific.

\bibitem[Jamieson and Talwalkar, 2016]{jamieson-aistats16}
Jamieson, K. and Talwalkar, A. (2016).
\newblock Non-stochastic best arm identification and hyperparameter
  optimization.
\newblock In {\em Proceedings of the 17th International Conference on
  Artificial Intelligence and Statistics (AISTATS'16)}.

\bibitem[Joas et~al., 2025]{joas2025AUTOENCODIX}
Joas, M.~J., Jurenaite, N., Pra{\v{s}}{\v{c}}evi{\'c}, D., Scherf, N., and
  Ewald, J. (2025).
\newblock Autoencodix: a generalized and versatile framework to train and
  evaluate autoencoders for biological representation learning and beyond.
\newblock {\em Nature Computational Science}, pages 1--13.

\bibitem[Klein and Hutter, 2019]{klein2019tabular}
Klein, A. and Hutter, F. (2019).
\newblock Tabular benchmarks for joint architecture and hyperparameter
  optimization.
\newblock {\em arXiv preprint arXiv:1905.04970}.

\bibitem[Kopf and Claassen, 2021]{kopf2021latent}
Kopf, A. and Claassen, M. (2021).
\newblock Latent representation learning in biology and translational medicine.
\newblock {\em Patterns}, 2(3).

\bibitem[Li et~al., 2018]{li2018hyperband}
Li, L., Jamieson, K., DeSalvo, G., Rostamizadeh, A., and Talwalkar, A. (2018).
\newblock Hyperband: A novel bandit-based approach to hyperparameter
  optimization.
\newblock {\em Journal of machine learning research}, 18(185):1--52.

\bibitem[Li et~al., 2020]{li2020system}
Li, L., Jamieson, K., Rostamizadeh, A., Gonina, E., Ben-Tzur, J., Hardt, M.,
  Recht, B., and Talwalkar, A. (2020).
\newblock A system for massively parallel hyperparameter tuning.
\newblock {\em Proceedings of machine learning and systems}, 2:230--246.

\bibitem[Locatello et~al., 2019]{locatello2019challenging}
Locatello, F., Bauer, S., Lucic, M., Raetsch, G., Gelly, S., Sch{\"o}lkopf, B.,
  and Bachem, O. (2019).
\newblock Challenging common assumptions in the unsupervised learning of
  disentangled representations.
\newblock In {\em international conference on machine learning}, pages
  4114--4124. PMLR.

\bibitem[Lotfollahi et~al., 2023a]{lotfollahi2023predicting}
Lotfollahi, M., Klimovskaia~Susmelj, A., De~Donno, C., Hetzel, L., Ji, Y.,
  Ibarra, I.~L., Srivatsan, S.~R., Naghipourfar, M., Daza, R.~M., Martin, B.,
  et~al. (2023a).
\newblock Predicting cellular responses to complex perturbations in
  high-throughput screens.
\newblock {\em Molecular systems biology}, 19(6):MSB202211517.

\bibitem[Lotfollahi et~al., 2023b]{lotfollahi2023biologically}
Lotfollahi, M., Rybakov, S., Hrovatin, K., Hediyeh-Zadeh, S.,
  Talavera-L{\'o}pez, C., Misharin, A.~V., and Theis, F.~J. (2023b).
\newblock Biologically informed deep learning to query gene programs in
  single-cell atlases.
\newblock {\em Nature Cell Biology}, 25(2):337--350.

\bibitem[Mamoshina et~al., 2016]{mamoshina2016applications}
Mamoshina, P., Vieira, A., Putin, E., and Zhavoronkov, A. (2016).
\newblock Applications of deep learning in biomedicine.
\newblock {\em Molecular pharmaceutics}, 13(5):1445--1454.

\bibitem[Mardis, 2008]{mardis2008impact}
Mardis, E.~R. (2008).
\newblock The impact of next-generation sequencing technology on genetics.
\newblock {\em Trends in genetics}, 24(3):133--141.

\bibitem[Milacic et~al., 2024]{milacic2024reactome}
Milacic, M., Beavers, D., Conley, P., Gong, C., Gillespie, M., Griss, J., Haw,
  R., Jassal, B., Matthews, L., May, B., et~al. (2024).
\newblock The reactome pathway knowledgebase 2024.
\newblock {\em Nucleic acids research}, 52(D1):D672--D678.

\bibitem[Ovcharenko et~al., 2025]{ovcharenko2025scssl}
Ovcharenko, O., Barkmann, F., Toma, P., Daunhawer, I., Vogt, J., Schelter, S.,
  and Boeva, V. (2025).
\newblock Scssl-bench: Benchmarking self-supervised learning for single-cell
  data.
\newblock {\em arXiv preprint arXiv:2506.10031}.

\bibitem[Perrone et~al., 2019]{perrone2019learning}
Perrone, V., Shen, H., Seeger, M.~W., Archambeau, C., and Jenatton, R. (2019).
\newblock Learning search spaces for bayesian optimization: Another view of
  hyperparameter transfer learning.
\newblock {\em Advances in neural information processing systems}, 32.

\bibitem[Real et~al., 2019]{real2019regularized}
Real, E., Aggarwal, A., Huang, Y., and Le, Q.~V. (2019).
\newblock Regularized evolution for image classifier architecture search.
\newblock In {\em Proceedings of the aaai conference on artificial
  intelligence}, volume~33, pages 4780--4789.

\bibitem[Reuter et~al., 2015]{reuter2015high}
Reuter, J.~A., Spacek, D.~V., and Snyder, M.~P. (2015).
\newblock High-throughput sequencing technologies.
\newblock {\em Molecular cell}, 58(4):586--597.

\bibitem[Salinas and Erickson, 2023]{salinas2023tabrepo}
Salinas, D. and Erickson, N. (2023).
\newblock Tabrepo: A large scale repository of tabular model evaluations and
  its automl applications.
\newblock {\em arXiv preprint arXiv:2311.02971}.

\bibitem[Salinas et~al., 2023]{salinas2023optimizing}
Salinas, D., Golebiowski, J., Klein, A., Seeger, M., and Archambeau, C. (2023).
\newblock Optimizing hyperparameters with conformal quantile regression.
\newblock In {\em International Conference on Machine Learning}, pages
  29876--29893. PMLR.

\bibitem[Salinas et~al., 2022]{salinas2022syne}
Salinas, D., Seeger, M., Klein, A., Perrone, V., Wistuba, M., and Archambeau,
  C. (2022).
\newblock Syne tune: A library for large scale hyperparameter tuning and
  reproducible research.
\newblock In {\em International Conference on Automated Machine Learning},
  pages 16--1. PMLR.

\bibitem[Salinas et~al., 2020]{salinas2020quantile}
Salinas, D., Shen, H., and Perrone, V. (2020).
\newblock A quantile-based approach for hyperparameter transfer learning.
\newblock In {\em International conference on machine learning}, pages
  8438--8448. PMLR.

\bibitem[Selby et~al., 2025]{selby2025visible}
Selby, D.~A., Jakhmola, R., Sprang, M., Gro{\ss}mann, G., Raki, H., Maani, N.,
  Pavliuk, D., Ewald, J., and Vollmer, S. (2025).
\newblock Visible neural networks for multi-omics integration: a critical
  review.
\newblock {\em Frontiers in Artificial Intelligence}, 8:1595291.

\bibitem[Seninge et~al., 2021]{seninge2021vega}
Seninge, L., Anastopoulos, I., Ding, H., and Stuart, J. (2021).
\newblock Vega is an interpretable generative model for inferring biological
  network activity in single-cell transcriptomics.
\newblock {\em Nature communications}, 12(1):5684.

\bibitem[Simidjievski et~al., 2019]{simidjievski2019variational}
Simidjievski, N., Bodnar, C., Tariq, I., Scherer, P., Andres~Terre, H., Shams,
  Z., Jamnik, M., and Li{\`o}, P. (2019).
\newblock Variational autoencoders for cancer data integration: design
  principles and computational practice.
\newblock {\em Frontiers in genetics}, 10:1205.

\bibitem[Snoek et~al., 2012]{snoek-nips12a}
Snoek, J., Larochelle, H., and Adams, R.~P. (2012).
\newblock Practical {B}ayesian optimization of machine learning algorithms.
\newblock In {\em Proceedings of the 25th International Conference on Advances
  in Neural Information Processing Systems (NeurIPS'12)}.

\bibitem[Tiao et~al., 2020]{tiao2020bayesian}
Tiao, L., Klein, A., Seeger, M., Archambeau, C., Bonilla, E., and Ramos, F.
  (2020).
\newblock Bayesian optimization by density ratio estimation.

\bibitem[Weinstein et~al., 2013]{weinstein2013cancer}
Weinstein, J.~N., Collisson, E.~A., Mills, G.~B., Shaw, K.~R., Ozenberger,
  B.~A., Ellrott, K., Shmulevich, I., Sander, C., and Stuart, J.~M. (2013).
\newblock The cancer genome atlas pan-cancer analysis project.
\newblock {\em Nature genetics}, 45(10):1113--1120.

\bibitem[Whalen et~al., 2022]{whalen2022navigating}
Whalen, S., Schreiber, J., Noble, W.~S., and Pollard, K.~S. (2022).
\newblock Navigating the pitfalls of applying machine learning in genomics.
\newblock {\em Nature Reviews Genetics}, 23(3):169--181.

\bibitem[Wistuba et~al., 2015a]{wistuba2015learning}
Wistuba, M., Schilling, N., and Schmidt-Thieme, L. (2015a).
\newblock Learning hyperparameter optimization initializations.
\newblock In {\em 2015 IEEE international conference on data science and
  advanced analytics (DSAA)}, pages 1--10. IEEE.

\bibitem[Wistuba et~al., 2015b]{wistuba2015sequential}
Wistuba, M., Schilling, N., and Schmidt-Thieme, L. (2015b).
\newblock Sequential model-free hyperparameter tuning.
\newblock In {\em 2015 IEEE international conference on data mining}, pages
  1033--1038. IEEE.

\bibitem[Yang et~al., 2021]{yang2021multi}
Yang, K.~D., Belyaeva, A., Venkatachalapathy, S., Damodaran, K., Katcoff, A.,
  Radhakrishnan, A., Shivashankar, G., and Uhler, C. (2021).
\newblock Multi-domain translation between single-cell imaging and sequencing
  data using autoencoders.
\newblock {\em Nature communications}, 12(1):31.

\bibitem[Zhu et~al., 2023]{zhu2023multi}
Zhu, K., Bendl, J., Rahman, S., Vicari, J.~M., Coleman, C., Clarence, T.,
  Latouche, O., Tsankova, N.~M., Li, A., Brennand, K.~J., et~al. (2023).
\newblock Multi-omic profiling of the developing human cerebral cortex at the
  single-cell level.
\newblock {\em Science advances}, 9(41):eadg3754.

\bibitem[Zimmer et~al., 2021]{zimmer2021auto}
Zimmer, L., Lindauer, M., and Hutter, F. (2021).
\newblock Auto-pytorch: Multi-fidelity metalearning for efficient and robust
  autodl.
\newblock {\em IEEE transactions on pattern analysis and machine intelligence},
  43(9):3079--3090.

\end{thebibliography}
\end{document}